# Some Network Optimization Models under Diverse Uncertain Environments

*A thesis submitted for partial fulfilment of the requirements for the degree of*

**Doctor of Philosophy**

in

Computer Science and Engineering

*by*

**Saibal Majumder**

**Roll No.: 15/CSE/1101, FT, PhD**

**Registration No.: NITD/PhD/CS/2016/00742**

*Under the joint supervision of*

**Tandra Pal**

Professor, Department of Computer Science and Engineering

National Institute of Technology Durgapur, India

and

**Samarjit Kar**

Professor, Department of Mathematics

National Institute of Technology Durgapur, India

# *Dedicated To*

*My parents, who brought me up and without whom I would not have come this far.*

*My respected professors, whose constant guidance motivated me to remain focused on achieving my goal.*

*Millions of poor and downtrodden people of my country, whose sacrifice and tolerance paved the royal road of my education.*

# National Institute of Technology
# Durgapur

# CERTIFICATE

---

It is certified that the work contained in the thesis entitled "**Some Network Optimization Models under Diverse Uncertain Environments**", has been carried out by **Saibal Majumder (Roll No.: 15/CSE/1101, FT, PhD** and **Registration No.: NITD/PhD/CS/2016/00742)** under the guidance of **Prof. Tandra Pal** and **Prof. Samarjit Kar**, the data reported herein is original and this work has not been submitted elsewhere for any other Degree or Diploma.

-------------------------------------

**Saibal Majumder** (Candidate)

Place: ………………………….

Date: …………………………..

This is to certify that the above declaration is true.

| | |
|---|---|
| ------------------------------------- | ------------------------------------- |
| **Prof. Tandra Pal** | **Prof. Samarjit Kar** |
| Place: …………………………. | Place: …………………………. |
| Date: ………………………….. | Date: ………………………….. |

# Acknowledgement

Firstly, I pay deep reverence to the Almighty to have bestowed on me, courage and zeal, to cope up with the challenges in my research. I take this opportunity to express my deep sense of gratitude and respectful regards to all of those people without whose support; the thesis would never have been possible. Since it is impossible to name all of these wonderful people, I will only attempt to mention here a handful of them.

I feel privileged and honour to express my sincere and deepest gratitude to my supervisors, Prof. Tandra Pal and Prof. Samarjit Kar for their invaluable guidance and incessant support during my doctoral research endeavour for the past five years. They have gone beyond their duties to elucidate my doubts, concerns and anxieties, and have helped me to foster the knowledge required for me. I feel fortunate having them as my supervisors.

My heartfelt thank goes to the members of my Doctoral Scrutiny Committee including Prof. G. Sanyal, Prof. S. Roy, Dr. S. Mukhopadhyay and Prof. S. Nandi for providing me insightful comments and timely directions. My earnest thank also goes to all other faculty members of the Department of Computer Science and Engineering, National Institute of Technology Durgapur, for their kind supports throughout my PhD tenure.

The support and cooperation which I have received from my friends are invaluable. I greatly appreciate and acknowledge the necessary support provided by the departments of Computer Science and Engineering and Mathematics, National Institute of Technology Durgapur, to carry out my research work. I extend my thanks to the academic staffs of the institution for their kind cooperation. A special mention of thanks is due to Dr. Pradip Kundu, Dr. Sujit Das, Dr. Prasenjit Dey, Dr. Mohuya B. Kar and Mr. Haresh Sharma for their limitless mental support and cooperation throughout the course. I am extremely thankful to Dr. Kaustuv Nag for his important suggestion which is instrumental in my study.

Being an INSPIRE fellow (DST/INSPIRE Fellowship/2015/IF150410), I am indebted to the Department of Science & Technology (DST), Government of India, for providing me financial support for my PhD work.

A special thanks to my younger brother Bidyut for his love, affection and active support.

Finally, I acknowledge my parents and my aunt for giving me the liberty to select what I desired, and inspiring me to work hard for those I aspire to achieve. I salute you all for giving me unconditional love and the sacrifice you did during the journey to fulfil my dream and aspiration.

**Date**:

**Saibal Majumder**
Roll No.: 15/CSE/1101, FT, PhD
Registration No.: NITD/PhD/CS/2016/00742
Department of Computer Science and Engineering
National Institute of Technology Durgapur
West Bengal-713209, India

# Table of Contents

# List of Tables

# List of Figures

# List of Abbreviations

| | |
|---|---|
| bLAS | Bilinear allocation strategy |
| b-RFQMSTP | bi-objective rough fuzzy quadratic minimum spanning tree problem |
| b-SPP | bi-objective shortest path problem |
| CCM | Chance-constrained model |
| CHC | Cross generational, elitist selection, heterogeneous recombination and cataclysmic mutation |
| DCCM | Dependent chance-constrained model |
| DCM | Dependent-chance model |
| DM | Decision maker |
| EA | Evolutionary algorithm |
| EVM | Expected value model |
| FOU | Footprint of uncertainty |
| FTP | Fixed charge transportation problem |
| GA | Genetic algorithm |
| $GD$ | Generational distance |
| GFN | Gaussian fuzzy number |
| GRFV | Gaussian random fuzzy variable |
| GTrFV | Generalized trapezoidal fuzzy variable |
| $HV$ | Hypervolume |
| $IGD$ | Inverted generational distance |
| iLAS | Improved lifetime allocation strategy |
| IQR | Interquartile range |
| IT2FS | Interval type-2 fuzzy set |
| IT2FV | Interval type-2 fuzzy variable |
| $k$-SPP | $k$-shortest path problem |
| LMF | Lower membership function |
| LPP | Linear programming problem |
| MCDM | Multi-criteria decision making |
| MFGA | Maximum flow genetic algorithm |
| MFP | Maximum flow problem |
| MMFSTPwB | Multi-objective multi-item fixed charge solid transportation problem with budget constraint |
| MOCHC | Multi-objective cross generational, elitist selection, heterogeneous recombination and cataclysmic mutation |
| MOGA | Multi-objective genetic algorithm |
| MOP | Multi-objective optimization problem |
| MP | Mathematical programming |
| MRCCSPP | Multi-criteria rough chance-constrained shortest path problem |
| MRD | Modified rough Dijkstra's |
| $ms$ | Milliseconds |
| MSPP | Multi-criteria shortest path problem |
| MST | Minimum spanning tree |
| MSTP | Minimum spanning tree problem |
| NSGA-II | Nondominated sorting genetic algorithm II |
| OSPF | Open shortest path first |
| OWA | Ordered weighted averaging |
| PF | Pareto front |
| PS | Pareto set |
| PSO | Particle swarm optimization |
| QMST | Quadratic minimum spanning tree |
| QMSTP | Quadratic minimum spanning tree problem |
| RCCP | Rough chance-constrained programming |
| RFCCP | Rough fuzzy chance-constrained programming |
| RFMFP | Random fuzzy maximum flow problem |
| RFMOP | Rough fuzzy multi-objective problem |
| $sd$ | Standard deviation |
| SIFN | Symmetrical intuitionistic fuzzy number |
| SOOP | Single objective optimization problem |
| SPP | Shortest path problem |
| STP | Solid transportation problem |
| T1FS | Type-1 fuzzy set |
| T1FV | Type-1 fuzzy variable |
| T2FS | Type-2 fuzzy set |
| T2FV | Type-2 fuzzy variable |
| TFN | Triangular fuzzy number |
| TP | Transportation problem |
| TrFN | Trapezoidal fuzzy number |
| TrIT2FV | Trapezoidal interval type-2 fuzzy variable |
| TrRFV | Trapezoidal random fuzzy variable |
| UMF | Upper membership function |
| UMMFSTPwB | Uncertain multi-objective multi-item fixed charge solid transportation problem with budget constraint |
| VPGAwIC | Varying population genetic algorithm with indeterminate crossover |
| WCDN | Weighted connected directed network |
| WCN | Weighted connected network |
| WCUN | Weighted connected undirected network |

# Abstract

Network models provide an efficient way to represent many real-life problems mathematically. In the last few decades, the field of network optimization has witnessed an upsurge of interest among researchers and practitioners. The network models considered in this thesis are broadly classified into four types: (i) transportation problem, (ii) shortest path problem, (iii) minimum spanning tree problem and (iv) maximum flow problem.

Quite often, we come across situations, when the decision parameters of network optimization problems are not precise and characterized by various forms of uncertainties arising from the factors, like insufficient or incomplete data, lack of evidence, inappropriate judgements and randomness. Considering the crisp/deterministic environment, there exist several studies on network optimization problems. However, in the literature, not many investigations on single and multi-objective network optimization problems are observed under diverse uncertain frameworks. This thesis proposes seven different network models under different uncertain paradigms. Among those, four network models are formulated as single objective optimization problems and the remaining three network models as multi-objective optimization problems. To formulate all these network models, we have considered type-2 fuzzy, random fuzzy, rough, uncertainty theory and rough fuzzy uncertain environments. Mainly, among the four single objective network models, the solid transportation problem, the shortest path problem, and the minimum spanning tree problem are modelled under type-2 fuzzy environment, and the maximum flow problem is presented under random fuzzy uncertain paradigm. Among the three multi-objective problems, the multi-objective shortest path problem, the multi-objective multi-item fixed charge solid transportation problem with budget constraints and the multi-objective quadratic spanning tree problem are respectively, modelled under rough, uncertainty theory and rough fuzzy uncertain paradigms. In this thesis, the uncertain programming techniques used to formulate the uncertain network models are (i) expected value model, (ii) chance-constrained model and (iii) dependent chance-constrained model. Subsequently, the corresponding crisp equivalents of the uncertain network models are solved using different solution methodologies.

The solution methodologies used in this thesis can be broadly categorized as classical methods and evolutionary algorithms. The classical methods, used in this thesis, are Dijkstra’s and Kruskal’s algorithms, modified rough Dijkstra’s algorithm, goal attainment method, linear weighted method, global criterion method, epsilon-constraint method and fuzzy programming method. Whereas, among the evolutionary algorithms, we have proposed the varying population genetic algorithm with indeterminate crossover and considered two multi-objective evolutionary algorithms:

(i) nondominated sorting genetic algorithm II and (ii) multi-objective cross generational elitist selection, heterogeneous recombination, and cataclysmic mutation.

To illustrate the proposed network models, suitable numerical examples are provided in this thesis. Moreover, the corresponding results of the proposed uncertain network models are compared and analyzed.

# Chapter 1
# Introduction

# Chapter 1

# Introduction

Network optimization is one of the most frequently encountered class of optimization techniques which deals with the optimization of several real-life network problems. A rich contribution in network optimization can be observed in the contemporary fields of operations research, engineering and management. Most of the problems in these fields, including communication systems, electrical networks, computer networks, scheduling problems, shortest paths, maximal flow, shortest tour, supply and demand problems, etc., can be modelled and solved using graph theory techniques. The genesis of network optimization has a connection with the famous Königsberg bridge problem (L. Euler 1736) in graph theory. Network optimization has received remarkable attention among researchers and practitioners after the appearance of the first book on graph theory by D. König (1936). In 1941, F.L. Hitchcock presented the first algorithm to solve the transportation problem. Subsequently, Jr. J.B. Kruskal (1956) proposed the minimum spanning tree algorithm. In the same year, the first algorithm to solve the maximum flow problem (Ford and Fulkerson 1956) in a network was proposed. Moving forward, there has been steady progress in the solution methodologies of network problems due to the technological advancement of the computer system and the development of many efficient algorithms. The acceptability of the network optimization as well as the theoretical significance in the context of complexity theory, which deals with the analysis of algorithms (D. Jungnickel 1999) have been increased.

Considering the practical relevance of the network applications, several researchers have contributed to different types of network optimization problems (R.C. Prim 1957; E.W. Dijkstra 1959; S.E. Dreyfus 1969; Bhatia et al. 1976). All these problems are usually investigated with deterministic problem parameters. Nevertheless, in many real-world network problems, the associated parameters are not always exact or precise. Bellman and Zadeh (1970) mentioned that “Much of the decision making in the real-world takes place in an environment in which the goals, the constraints and the consequences of possible actions are not known precisely.” The reasons behind the existence of uncertainty in the problem parameters are the availability of insufficient information, lack of evidence, uncertainty in judgment, etc. Therefore, a proper representation of the uncertain parameters is necessary while modelling real-world problems. In order to process and represent the imprecise data, many researchers have proposed different theories, like probability theory, fuzzy set (L.A. Zadeh 1965), type-2 fuzzy set (L.A. Zadeh 1975a, b), rough set (Z. Pawlak 1982) and uncertainty theory

(B. Liu 2007). In this thesis, while developing different network models, we have used interval type-2 fuzzy variable (T.-Y. Chen 2013, 2014), random fuzzy variable (B. Liu 2002), rough variable (B. Liu 2002), rough fuzzy variable (B. Liu 2002) and uncertain variable (B. Liu 2007) to represent the related parameters of different models.

Mathematical programming (MP) is considered as an optimization technique in operations research, which optimizes a quantity (or quantities) commonly referred as objective function, with respect to a set of equality and/or inequality constraints. An MP model can be defined as

$$\begin{cases} min\, f(x) \\ subject\ to \\ \quad g_j(x)\ (\leq, =, \geq)\ 0, j = 1,2, \dots, n \\ \quad x \geq 0, \end{cases} \tag{1.1}$$

where $x = (x_1, x_2, \dots, x_q)$ is a $q$-dimentional decision vector and $g_j(x)$ are the constraints, $j = 1,2, \dots n$. Here, all the decision variables are assumed as non-negative, i.e., $x_i \geq 0, i = 1,2, \dots, q$.

The model presented in (1.1) is considered as a single objective optimization problem (SOOP). However, most of the real-world decision making problems have multiple conflicting objectives which are to be optimized simultaneously. Accordingly, model (1.1) can be extended to a multi-objective optimization problem (MOP) as presented in (1.2).

$$\begin{cases} min[f_1(x), f_2(x), \dots f_m(x)] \\ subject\ to \\ \quad g_j(x)\ (\leq, =, \geq)\ 0, j = 1,2, \dots, n \\ \quad x \geq 0, \end{cases} \tag{1.2}$$

where $x = (x_1, x_2, \dots, x_q)$ is a $q$-dimentional decision vector, $f_i(x)$ are the objective functions, $i = 1,2, \dots m$, and $g_j(x)$, $j = 1,2, \dots n$ are the constraints.

The optimization models presented in (1.1) and (1.2) are well-defined as long as the associated parameters of the models are considered as deterministic. However, to model network optimization problems under different types of uncertainty as defined above, the classical optimization models presented in (1.1) and (1.2) become no longer valid. To incorporate uncertain parameters in the optimization models, B. Liu (2002) introduced uncertain programming. The uncertain programming is considered as a branch of MP and can be broadly categorized into three models: (i) expected value model (EVM), which optimizes the expected objective function(s) subject to a set of expected constraints, (ii) chance-constrained model (CCM) (Chance and Cooper 1959), which aims to optimize a return at a particular confidence level to which the uncertain constraints should desirably hold and (iii) dependent-chance model (DCM) (B. Liu 1997), which aims to maximize the chance of an uncertain event. All these models are the possible strategies that a decision maker (DM) can adopt while modelling network problems as uncertain single objective or multi-objective optimization problems.

In the field of optimization, the classical search and optimization techniques are based on single point search, where a solution is improved with iteration. Some studies on the classical approaches used to solve single objective network problems (G.B. Dantzig 1951; Ford and Fulkerson 1956; R.C. Prim 1957; E.W. Dijkstra 1959) can be observed in the literature. Several classical techniques for multi-objective problems, such as linear weighted method (R.T. Eckenrode 1965), epsilon-constraint method (Haimes et al. 1971), goal attainment method (F.W.Gembicki 1974), fuzzy programming method (H.-J. Zimmermann 1978) and global criterion method (S.S. Rao 2006) are observed in the literature. Sutcliffe et al. (1984), Gupta and Warburton (1987) and Pulat et al. (1992) have applied some of these techniques on multi-objective network problems. In the field of optimization, over the last few decades, significant progress in optimization technique has been observed which are based on evolutionary techniques. An evolutionary technique is a population based stochastic optimization technique, which imitates the evolutionary phenomena of nature while driving its search towards optimality. Unlike classical methods, it generates a set of solutions at each iteration. If an optimization problem has a single optimum, then all the population members of the evolutionary methods converge at that optimum. For multiple objectives, it provides multiple optima in its final population. The inimitable characteristic of evolutionary techniques to search multiple solutions in a single execution, essentially makes them unique alternatives to solve MOPs.

In this thesis, we have mainly concentrated on modelling different single and multi-objective network optimization problems under various uncertain paradigms using uncertain programming techniques. These problems are then transformed to their crisp equivalents and are eventually solved using different solutions methodologies, using both classical and evolutionary algorithms.

## 1.1 Literature Study

In this section, we present a survey on the studies related to four network optimization problems: (i) transportation problem, (ii) shortest path problem, (iii) minimum spanning tree problem and (iv) maximum flow problem, which broadly categorize the contributions in our thesis. Here, we have also provided brief discussion of the investigations done on different variants of these network problems. It is to be mentioned that, this survey, by no means, encompasses all the related researches in the literature. However, some studies having significant contributions to these network problems are reviewed below.

**Transportation problem (TP)**

With the development of economic globalization, transportation problem has emerged as one of the important combinatorial optimization problems and becomes very relevant to the ever-expanding worldwide market. TP determines an optimal distribution of the products from each of the sources to various destinations to minimize the total transportation cost. The problem is first introduced by F.L. Hitchcock (1941) and subsequently improved by T.C. Koopmans (1949). Following the contributions of their studies, G.M. Appa (1973) discussed different cases of the unboundedness and infeasibility of the models of TP. C.S. Ramakrishnan (1988) revisited the problem and proposed an improved Vogel's approximation method (Reinfeld and Vogel 1958) as a solution technique of the TP. Arsham and Kahn (1989) proposed an algorithm for TP, which is faster than simplex (G.B. Dantzig 1951), more general than stepping-stone method (W. Shih 1987) and simpler than both. S.I. Gass (1990) analyzed different facets of solution methodologies of TP. Vignaux and Michalewicz (1991) proposed a genetic algorithm as a new solution approach to the problem. Considering multi-objective TP, Aneja and Nair (1979) proposed a parametric search method to solve a bi-objective TP. H. Isermann (1979) proposed an algorithm to determine the nondominated solutions to the problem. G.K. Tayi (1986) solved a transportation problem by minimizing transportation cost and deterioration of an item during a transportation activity. The author developed an approach by explicitly considering trade-offs between the objectives as stated by a decision maker.

An important extension of a classical transportation problem is fixed charge transportation problem, first proposed by Hirsch and Dantzig (1968). In practical applications, the fixed charge problem may include highway toll charges, landing fees at airports, cost for construction of roads or setup costs in production system (Palekar et al. 1990). Kennington and Unger (1976) and Barr et al. (1981) solved the fixed charge TP with branch-and-bound algorithm. Subsequently, different branch-and-bound algorithms with conditional penalties are employed by Cabot and Erenguc (1984, 1986) and Lamar and Wallace (1997), as solution methodologies of the problem. Later, Adlakha and Kowalski (2003) proposed a heuristic algorithm for the problem. Of late, Lotfi and Tavakkoli-Moghaddan (2013) proposed a genetic algorithm with priority-based encoding to solve the problem. Further, Kowalski et al. (2014) developed a simple branching algorithm to generate a global solution of a fixed charge TP. Subsequently, a local search algorithm and an artificial immune algorithm were proposed by Buson et al. (2014) and Altassan et al. (2014), respectively, for solving TP.

The solid transportation problem is yet another variant of the traditional TP. Here, apart from availability and demand constraints, there is one more constraint known as conveyance constraint. There may be multiple modes of conveyances, available at the sources for transporting items. Therefore, the conveyances constraint is used to determine an appropriate mode of transportation so that the total transportation cost is minimized. In this context, K.B. Haley (1962) first introduced the solid TP. The problem was again addressed by Hadley and Whitin (1963). Ever since, solid TP has received much attention among researchers. H.L. Bhatia (1981) determined feasible solutions of a solid TP with indefinite quadratic objective function. Pandian and Anuradha (2010) proposed a new method using the principle of zero point method (Pandian and Natarajan 2010) for finding an optimal solution of solid TP.

In classical transportation models, mentioned above, the associated parameters are considered as deterministic. However, usually there exist some indeterminate factors like insufficient information, fluctuation in financial market, unstable political situation and artificial market crisis. In order to deal with such indeterminacy in transportation problem, A.C. Williams (1963) proposed a TP, where the demands are considered as random variable. D. Wilson (1975) revisited the same problem and solved it using a simple approximation technique. Later, Yang and Feng (2007) proposed three uncertain programming models for bi-objective fixed charge solid TP, where the transportation cost, fixed charge cost and transportation time between a source and destination are considered as random variable. Besides, several researchers (Mahapatra et al. 2010; Romeijn and Sargut 2011; Midya and Roy 2014) contributed to further research on different variants of TP, under stochastic paradigm.

Considering the fuzzy environment, Chanas et al. (1984) proposed a transportation problem with fuzzy availabilities and demands, and solved it using parametric programming technique. Later, Chanas and Kuchta (1996) proposed a transportation problem with fuzzy transportation cost, crisp availabilities and demands, and proposed an algorithm to solve it. Jimenez and Verdegay (1998) proposed two uncertain models of solid TP by considering the parameters, involved in the constraints of the problem, as intervals as well as fuzzy. Yang and Liu (2007) proposed expected value model, chance-constrained model and dependent-chance model based on credibility theory to address a fuzzy fixed charge solid transportation problem. They solved the models with a hybrid intelligent algorithm based on fuzzy simulation and tabu search. Kaur and Kumar (2012) proposed a new algorithm to solve the TP, where the transportation costs are represented by generalized trapezoidal fuzzy numbers. Ojha et al. (2014) proposed a TP by considering the transportation costs and budget at destinations as fuzzy random variable. The authors then solved the crisp equivalent of the proposed model with a genetic algorithm. Besides, Kundu et al. (2014a) addressed a fixed charge transportation problem, where the associated parameters are represented by generalized type-2 fuzzy variable, and solve the corresponding crisp transformations using the optimization software, LINGO. Moreover, Pramanik et al. (2015b) proposed a fixed charge TP for a two-stage supply chain network, where the transportation costs, fixed charges, availabilities and demands are considered as Gaussian type-2 fuzzy variables. In their study, the authors implemented a genetic algorithm (GA) and particle swarm optimization (PSO), as solution methodologies of the proposed problem. Later, Aggarwal and Gupta (2016) proposed the signed distance of symmetrical intuitionistic fuzzy numbers (SIFNs) based on which a new ranking of SIFNs was introduced and employed to solve a solid TP. Under multi-objective domain, Das et al. (1999) solved the transportation problem by fuzzy programming technique, where the transportation costs, supply and demand parameters were represented by interval number. Afterwards, a two-phase fuzzy algorithm was developed by Gao and Liu (2004) to solve a multi-objective fuzzy TP. Ojha et al. (2009) proposed an entropy-based solid TP which minimizes both transportation time and cost. Here, the authors implemented the generalized reduced gradient method to find compromise solution. Kundu et al. (2013c) presented two different models of fuzzy multi-objective multi-item solid TP and determined the compromise solutions of the models using fuzzy programming technique and global criterion method. Pramanik et al. (2015a) proposed a multi-objective fixed charge solid TP, where the associated parameters are considered as random fuzzy variables. The authors then solved the equivalent crisp transformation of the model using the interactive fuzzy satisfaction method (Sakawa et al. 2003). Consequently, Sinha et al. (2016) proposed an expected value model of profit maximization and time minimization solid TP, where the associated input parameters are considered as interval type-2 fuzzy variable, and eventually used the interactive fuzzy satisfaction method to solve the problem.

Under Liu's uncertainty theory framework, Sheng and Yao (2012a) proposed an expected chance model for fixed charge TP, where the direct costs, fixed charges, availabilities and demands are considered as uncertain variable. Subsequently, Sheng and Yao (2012b) and Cui and Sheng (2013) employed the expected chance model to solve the uncertain TP and uncertain solid TP, respectively. Later, Zhang et al. (2016) proposed a hybrid intelligent algorithm based on uncertainty theory and tabu search to solve three uncertain programming models of the proposed uncertain fixed charge solid TP. Considering uncertain multi-objective TP, Mou et al. (2013) proposed a multi-objective uncertain transportation problem to address emergency scheduling in a transportation network. Subsequently, Chen et al. (2017) proposed three uncertain goal programming models for bi-objective solid TP and solved the corresponding deterministic transformations using an optimization software, LINGO. Recently, Majumder et al. (2018) proposed three uncertain programming models of multi-objective multi-item solid fixed charge TP and employed the linear weighted method, global criterion method and fuzzy programming technique to solve each of the corresponding crisp equivalents.

**Shortest path problem (SPP)**

As one of the fundamental and essential problem in network optimization, the shortest path problem aims to determine a path of minimum cost (distance or time) from a specific source vertex to a specific sink vertex of a network. Since the late 1950s, SPP has been studied by many researchers, and in this regard, some efficient algorithms, such as E.W. Dijkstra (1959), R.W. Floyd (1962), S.E. Dreyfus (1969) and Ahuja et al. (1993) have been proposed. These algorithms are very popular in the context of single objective SPP. In the multi-objective domain, the problem is first studied by P. Hansen (1980) for two objectives which is subsequently revisited by M.I. Henig (1986), where the nondominated paths of a bi-objective shortest path problem (b-SPP) are determined using dynamic programming. Mote et al. (1991) proposed an algorithmic approach to solve different non-dominating solutions of the b-SPP. Afterwards, Skriver and Andersen (2000) proposed an algorithm for b-SPP. Iori et al. (2010) proposed an algorithm for multi-objective shortest path problem (MSPP) based on weighted sum aggregated ordering to determine nondominated solutions. Subsequently, Sedeño-Noda and Raith (2015) generalized Dijkstra's algorithm to determine the nondominated paths of b-SPP. Later, Shi et al. (2017b) proposed an exact method for determining the Pareto optimal paths of a multi-objective constrained SPP.

Several contributions on SPP are also observed in different uncertain paradigms. An SPP under uncertain framework, considers the associated parameters of a network as characteristically nondeterministic, which may occur due to different types of uncertainty like lack of evidence and multiple sources of information from different experts. Some researchers considered randomness as nondeterministic phenomena. Therefore, they incorporated probability theory in network optimization problems and used random variable to represent the non-deterministic characteristics of the problem parameters. A random network is first presented by Frank and Hakimi (1965). Since then many researchers (Nie and Wu 2009; Chen et al. 2012; Zockaie et al. 2014) have significantly contributed to the study of random SPP. However, when the observational data are insufficient, then the estimated probability distribution is not appropriate to determine the non-deterministic phenomena (B. Liu 2007; X. Huang 2007a). To circumvent this problem, the most feasible and economic way to estimate data is to consider the opinions of experts. In this regard, the fuzzy set theory is considered as one of the approaches to tackle imprecision. In the literature, there are several studies on fuzzy SPP. Dubois and Prade (1980) first addressed SPP under fuzzy environment. C.M. Klein (1991) proposed an algorithm based on dynamic programming to solve fuzzy SPP models. Okeda and Gen (1994) incorporated Dijkstra's algorithm to solve fuzzy SPP, where the edge weights are represented by interval fuzzy number. Afterwards, S. Okeda (2004) developed a hybridized algorithm to determine the shortest path of a network, based on the degree of possibility of every fuzzy number which is associated with each edge. Ji et al. (2007) proposed three different models: (i) the expected shortest path, (ii) the $\alpha$-shortest path and (iii) the most shortest path model. Moreover, the authors have also proposed a hybrid intelligent algorithm to solve these models using fuzzy simulation and genetic algorithm. Hernandes et al. (2007a) proposed a fuzzy shortest path algorithm based on generic ranking index for comparing the fuzzy numbers associated with a network. Mahdavi et al. (2009) proposed a dynamic programming approach to determine the fuzzy shortest chain problem using a ranking method. Tajdin et al. (2010) proposed a dynamic programming method to determine the shortest path in a network with mixed fuzzy edge weights. Subsequently, T. Hasuike (2010) presented a new risk measure to synthesize the probabilistic conditional Value at Risk and fuzzy credibility measure to model fuzzy random SPP. Dou et al. (2012) investigated a fuzzy shortest path problem in a network having multiple constraints with multi-criteria decision making (MCDM) approach based on the vague similarity measure. Deng et al. (2012) implement fuzzy Dijkstra's algorithm to solve the shortest path of a network, where the edge weights are represented by fuzzy number. Hassanzadeh et al. (2013) presented a genetic algorithm to solve the fuzzy shortest path problem. Ebrahimnejad et al. (2015) used particle swarm optimization to solve the fuzzy shortest path of a network with different types of fuzzy edge weights. Under type-2 fuzzy environment, Anusuya and Sathya (2014) solved the SPP by considering the associated weight of each edge as discrete type-2 fuzzy variable. Later, Kumar et al. (2017) solved the type-2 fuzzy SPP by ranking all the possible paths in a network. Here, the authors considered the associated parameters as generalized type-2 fuzzy variables. Considering multi-objective fuzzy shortest path problem, Mahdavi et al. (2011) proposed two algorithms and a dynamic programming approach to solve the problem. Kumar and Sastry (2013) revisited the problem and proposed an algorithm which can accept both trapezoidal and triangular fuzzy variables as input parameters. Under Liu's uncertainty theory (2007), Y. Gao (2011) proposed two different models of SPP: (i) $\alpha$-shortest path and (ii) most shortest path and solved the crisp equivalents of these two models using Dijkstra's algorithm. Subsequently, Zhou et al. (2014a) addressed an inverse SPP for a network with uncertain edge weights and solved the corresponding crisp equivalent using linear programming. Later, Sheng and Gao (2016) considered a two-fold uncertain hybrid environment while addressing the SPP. Here the authors presented a modified Dijkstra's algorithm to solve the uncertain random SPP.

**Minimum spanning tree problem (MSTP)**

A minimum spanning tree problem is one of the most important problem in combinatorial optimization which has been studied since the beginning of the last century. O. Borüvka (1926) first proposed an algorithm to solve a minimum spanning tree (MST). Since then the problem has been revisited by different researchers including V. Jarník (1930), Jr. J.B. Kruskal (1956) and R.C. Prim (1957). A more detailed development of the problem and its solution methodologies can be found in the studies of Graham and Hell (1985), Karger et al. (1995), Pettie and Ramchandran (2002), and Nešetřil and Nešetřilová (2012). Besides, different variants of MSTP including $k$-smallest spanning trees (H.N. Gabow 1977), Euclidean MST (Agarwal et al. 1991), quadratic MST (Assad and Xu 1992), maximum spanning tree (McDonald et al. 2005), capacitated MST (T. Öncan 2007) and degree constrained MST (J.A. Torkestani 2013), are observed in the literature. All these MSTs mentioned above are developed with the aim to optimize a single objective.

Under the multi-objective domain, the first algorithm to solve a multi-objective MSTP (MMSTP) was introduced by Corley (1985), which is based on Prim's algorithm (R.C. Prim 1957). Thereafter, there have been various developments on the solution methodologies of MSTP including the exact algorithms (M. Ehrgott 2005; Steiner and Radzik 2008; Clímaco and Pascoal 2013; Di Puglia Pugliese et al. 2015), local search algorithms (Andersen et al. 1996; Maia et al. 2013) as well as evolutionary algorithms (Zhou and Gen 1999; Han and Wang 2005; Moradkhan and Browne 2006; Li et al. 2013).

All the above mentioned studies on MSTP and its different variants assume the associated input parameters as deterministic. However, these parameters are not always deterministic. In that case, to tackle nondeterministic or imprecise characteristic of the problem parameters, some improved theories like probability theory, fuzzy set theory and uncertainty theory, are often taken into consideration. In this context, Ishii et al. (1981) first introduced a stochastic spanning tree problem with random edge weights whose probability distributions are not known. A.M. Frieze (1985) determined an MST for a complete graph, where the length of each edge is considered as independent and identically distributed non-negative random variable. Ishii and Matsutomi (1995) extended the work of Ishii et al. (1981), where the parameters of probability distributions are not known in advance, and proposed a polynomial time algorithm to solve the problem. Dhamdhere et al. (2005) discussed two-stage stochastic MSTP and presented an approximation algorithm to solve the problem. Later, Torkestani and Meybodi (2012) proposed a learning automata based heuristic algorithm to determine an MST of a stochastic network, where the probability distribution of an edge weight is unknown.

Under the fuzzy environment, Itoh and Ishii (1996) first proposed a chance-constrained model of an MSTP, where the edge weights of a network are represented by fuzzy variable. Chang and Lee (1999) addressed the fuzzy MSTP by incorporating the concept of ranking index (Chang and Lee 1994) in their proposed algorithm. De Almeida (2005a) proposed an exact algorithm as well as a genetic algorithm to solve the fuzzy MSTP. Gao and Lu (2005) proposed three programming models: (i) expected value model, (ii) chance-constrained model and (iii) dependent-chance model of fuzzy quadratic MSTP and proposed a genetic algorithm to solve the crisp-equivalents of those models. Janiak and Kasperski (2008) implemented the possibility theory to characterize the optimality of edges in a network while selecting a spanning tree with edge costs represented by fuzzy interval. Afterwards, Zhou et al. (2016a) proposed $\alpha$- minimum spanning tree problem based on credibility measure (Liu and Liu 2002) of fuzzy variable and solved the problem with a hybrid intelligent algorithm.

Considering the hybrid environment, where fuzziness and randomness co-exist, Katagiri et al. (2004) first introduced a fuzzy random MSTP and solved it with a polynomial time algorithm. Liu and Yang (2007) proposed the expected value model, chance-constrained model and dependent-chance model for fuzzy random degree constrained MST and presented a hybrid intelligent algorithm to solve the corresponding crisp transformations. Later, Katagiri et al. (2012) proposed a decision making model based on possibility measure and value at risk measure to determine an MST of a network whose edge weights are represented by random fuzzy variable (B. Liu 2002).

Under the environment of Liu's uncertainty, Zhang et al. (2013a) proposed sum-type and minimax-type uncertain programming models for $\alpha$-minimum spanning tree, and solve the crisp equivalents of the models using classical optimization methods. Consequently, Zhang et al. (2013b) proposed a chance-constrained model to solve an inverse spanning tree problem with uncertain edge weights. Again, Zhou et al. (2015) defined the path optimality conditions of uncertain expected MST and uncertain $\alpha$-MST. Moreover, the authors proposed an uncertain most MST for a network with uncertain edge weights and established its relation with uncertain $\alpha$-MST. Zhou et al. (2016c) also proposed an ideal uncertain $\alpha$-MST by extending the definition of uncertain $\alpha$-MST. Further, the authors also proposed the definition of uncertain distribution- minimum spanning tree based on the concept of ideal uncertain $\alpha$-MST. Besides, two different variants of MST with uncertain edge weights, i.e., uncertain quadratic minimum spanning tree problem (Zhou et al. 2014b) and uncertain degree constrained minimum spanning tree problem (Gao and Jia 2017) are also observed in the literature. Considering the uncertain random paradigm, the problem of minimum spanning tree is first proposed by Sheng et al. (2017). Subsequently, Gao et al. (2017) studied the uncertain random degree constrained minimum spanning tree problem.

**Maximum flow problem (MFP)**

The maximum flow problem is one of the fundamental problems in network optimization. MFP aims to determine the maximum permissible quantity to be shipped from source to sink vertex of a directed network subject to non-violation of capacity constraints. Several efficient algorithms of MFP are observed in the literature. Fulkerson and Dantzig (1955) first proposed a computational algorithm based on simplex algorithm to solve the problem. The solution methods for MFP are broadly classified as augmenting path based algorithm and preflow based algorithm. An augmenting path based algorithm pushes a flow along a path from source to sink in a residual network (Ford and Fulkerson 1956, 1962). In this context, Ford and Fulkerson (1956) first proposed augmenting flow based algorithm. Subsequently, E.A. Dinic (1970) introduced the concept of shortest path network called layered network and proposed an algorithm to determine the flow of such network. Edmonds and Karp (1972) improved the algorithm proposed by Ford and Fulkerson (1956) by introducing the concept of the shortest augmenting path method. Conversely, a preflow based algorithm pushes the flow on a single edge instead of an entire augmenting path. A.V. Karzanov (1974) first proposed the concept of preflow in a layered network. Later, Trajan (1984) presented a simplified version of Karzanov's algorithm. Subsequently, Goldberg and Tarjan (1986) incorporated the concept of distance labels in their proposed preflow-push algorithm instead of constructing a layered network. Munakata and Hashier (1993) addressed the MFP using a genetic algorithm, where each solution is represented by a flow matrix. Ericsson et al. (2001) proposed a genetic algorithm to solve the open shortest path first (OSPF) weight setting problem, which is an extension of MFP. Later Gen et al. (2008) proposed a priority based genetic algorithm to solve the MFP. The authors also presented a multi-objective genetic algorithm to solve maximum flow minimum cost model.

The researchers of all the studies of MFP as mentioned above considered networks, where the associated capacities of the edges are represented as deterministic quantity. However, in practice, a flow capacity of an edge may change over time in communication or transportation network (Frank and Hakimi 1965; S. Ding 2015). In this context, Frank and Hakimi (1965) first presented the concept of random network, where the associated capacity of an edge is represented by a random variable and determined the probability that a particular flow between a pair of vertices can be achieved. Further, P. Doulliez (1971) proposed an algorithm to determine the probability distribution function of the capacity for a multi-terminal network, where the capacities of the edges are represented by independent random variable and the demands at different vertices are expressed as increasing function of time. J.R. Evans (1976) investigated some theoretical aspects of MFP for a random network, where an edge capacity is represented by a discrete random variable. Carey and Hendrickson (1984) proposed algorithms to determine the upper and lower bounds of expected maximum flow and benefits in a capacitated network subject to random link failure. Moreover, G.S. Fishman (1987) employed a stochastic optimization technique to estimate the maximum flow distribution of a random network.

Under the fuzzy environment, Kim and Roush (1982) first introduced maximum flow problem and presented some theoretical results related to maximum admissible flow in a directed network with fuzzy capacity. Subsequently, the problem is further developed by the studies of Chanas and Kołodziejczyk (1982, 1984, 1986). Afterwards, Liu and Kao (2004) proposed MFP with fuzzy edge capacity. Here, the authors incorporated Yager ranking index (R.R. Yager 1981) for the crisp transformation of the proposed problem and eventually used linear programming to generate the solution. Later, Ji et al. (2006) proposed a fuzzy chance-constrained model of the problem based on credibility measure, where the edge capacity is characterized by trapezoidal fuzzy number. Additionally, Kumar and Kaur (2010) proposed an algorithm to determine the maximum flow of a fuzzy network, where the edge capacity is represented by generalized trapezoidal fuzzy number. Further, Kumar and Kaur (2011) formulated a fuzzy linear programming model of MFP and used a fuzzy ranking method (Liou and Wang 1992) and fuzzy arithmetic operation (Kaufmann and Gupta 1985) to determine the corresponding crisp equivalent of the problem.

The maximum flow problem is also studied by several researchers under uncertainty theory (B. Liu 2007). Han et al. (2014) first considered the maximum flow problem under the framework of uncertainty theory. Here, the authors employed a 99-method (B. Liu 2010) to determine the expected maximum flow of an uncertain network. Based on the chance-constrained model, S. Ding (2015) proposed an $\alpha$-maximum flow model of an uncertain network and solved the corresponding deterministic transformation of the model with a preflow-push algorithm at different confidence levels of $\alpha$. Considering the two-fold uncertain random network, Sheng and Gao (2014) proposed an expected value model and a chance-constrained model of MFP, and provided some related theorems. Recently, Shi et al. (2017a) also developed an expected value model and a chance-constrained model of uncertain random MFP, and implemented the Ford-Fulkerson algorithm to solve the corresponding deterministic equivalents of the proposed models.

## 1.2 Research Motivation and Objectives

The motivation behind the proposed research works and their objectives are presented below.

### 1.2.1 Motivation

Many network optimization problems, such as transportation problem, shortest path problem and minimum spanning tree problem have captivated researchers due to their real-life applications in communication, manufacturing, traffic, production, logistics and so forth. The parameters of such decision making problems are quite often imprecise or uncertain in nature. Therefore, incorporating the uncertain parameters while modelling such network problems becomes essential. In this context, a linear programming problem (LPP) with fuzzy parameters (H.-J. Zimmermann 1978), can be considered as an effective approach. Application of LPP with type-1 fuzzy set (T1FS) can be observed in Le and Gogne (2010), Cheng et al. (2013) and Chakraborty et al. (2014). However, there is a lack of appropriate methods to deal with LPPs involving interval type-2 fuzzy parameters. J.C. Figueroa-García (2012), and Figueroa-García and Hernández (2014) implemented fuzzy linear programming, where the associated parameters are represented under interval type-2 fuzzy paradigm. However, to the best of our knowledge, there is no investigation on LPP by using the credibility measure of the interval type-2 fuzzy variable (IT2FV). It makes us motivated to formulate an LPP based on chance-constrained programming technique and credibility measure of the IT2FV. The proposed LPP is used to formulate three different uncertain network models with interval type-2 fuzzy parameters: (i) solid transportation problem (STP), (ii) shortest path problem (SPP) and (iii) minimum spanning tree problem (MSTP).

The maximum flow problem (MFP) is an important network optimization problem, which aims to maximize the amount of flow between the two specified nodes of a network. Compared to some popular network problems, like TP, SPP and MSTP, MFP has been studied less under uncertain framework. Kim and Roush (1982) first proposed MFP with fuzzy capacities. Later, Hernandes et al. (2007b) modified the algorithm of Ford and Fulkerson (1962) under fuzzy environment. Considering the two-fold uncertain environment, B. Liu (2002) presented the idea of random fuzzy variable, where the parameters of a random variable are considered as fuzzy. Later, different optimization problems, such as project selection (X. Huang 2007a), facility location-allocation problem (Wen and Iwamura 2008) and MSTP (Katagiri et al. 2012) are developed and solved under random fuzzy environment. However, to the best of our knowledge, a random fuzzy MFP is yet to be studied in the literature. Therefore, we are motivated to conduct a study on random fuzzy MFP.

Most of the real-world problems are naturally recognized with multiple conflicting objectives, which are to be optimized simultaneously. Solving MOPs is a challenge to the researchers. In addition, if the multi-objective problems (MOPs) are to be formulated under uncertain framework, modelling of such problems becomes more complex. In the literature, there exist very few investigations of network models with multiple objectives under uncertain environment.

One of the important network optimization problems, the fuzzy shortest path problem was first studied by Dubois and Prade (1980). Since then, there are very few research works in the field of multi-objective fuzzy SPP. Mahdavi et al. (2011) proposed two algorithms and a dynamic programming model to solve the bi-objective fuzzy SPP. Later, Kumar and Sastry (2013) presented an algorithm to find Pareto optimal solutions of a fuzzy bi-objective shortest path problem. However, to the best of our knowledge, any investigation on SPP with multiple objectives under rough environment has not yet investigated. Therefore, we have been motivated to study multi-objective shortest path problem (MSPP) of a weighted connected directed network (WCDN), whose edge weights are represented by rough variable (B. Liu 2002).

The notion of a classical transportation problem (TP) is to determine an optimal solution (transportation plan) such that the transportation cost is minimized. F.L. Hitchcock (1941) introduced transportation problem by modelling it as a conventional two dimensional optimization problem with respect to supply and demand. Since then, several developments of TP can be observed under different uncertain environments (Kaur and Kumar 2012; Sheng and Yao 2012a,b; Kundu et al. 2014a; Liu et al. 2017). In addition, several studies on uncertain multi-objective TP can also be observed in the literature (Gen et al. 1995; Mou et al. 2013; Kundu et al. 2013a, b; H. Dalman 2016). In spite of all the studies on TP, to the best of our knowledge, none has considered a multi-objective multi-item profit maximization and time minimization fixed charge STP model with budget constraint. It motivates us to study on uncertain multi-objective multi-item fixed charge solid transportation problem with budget constraint (UMMFSTPwB) at destinations under the framework of uncertainty theory (B. Liu 2007).

A quadratic minimum spanning tree problem (QMSTP) is a variant of MSTP, which is NP-hard in nature. The problem is first proposed by Assad and Xu (1992). There are two more significant investigations in the literature on QMSTP under uncertain environment. Gao and Lu (2005) defined a fuzzy QMSTP and solved the problem using genetic algorithm (GA). Later, Zhou et al. (2014b) defined a chance-constrained model for a QMSTP under the framework of uncertainty theory, and solve the problem using an optimization software, LINGO. In decision making, consideration of hybrid uncertain environment is also significant, where fuzziness and roughness co-exist. In this context, the concepts of rough fuzzy sets and fuzzy rough sets, introduced by Dubois and Prade (1990), play an important role. The authors combined the fuzzy set and rough set with an aim to fulfil two different purposes. The first one is to determine the lower and upper approximations of the fuzzy set, and the second one is to replace the equivalence relation of the rough set with the fuzzy similarity relation. Motivated by the study of Dubois and Prade (1990), B. Liu (2002) introduced the concept of rough fuzzy variable, which is defined as a measurable function form the possibility space to a set of rough variables. The application of rough fuzzy variable is not yet observed in any optimization problem. Further, a multi-objective QMSTP under any uncertain framework is not yet considered. It has motivated us to study the bi-objective QMSTP by considering the associated parameters as rough fuzzy variable.

### 1.2.2 Objectives

This thesis provides an empirical study on some network optimization problems for developing and solving different single and multi-objective network models under various uncertain frameworks using different classical and evolutionary algorithms. The main objectives of the thesis are summarized below:

- To develop some single and multi-objective network models which are broadly classified under four different network problem: (i) transportation problem, (ii) shortest path problem, (iii) minimum spanning tree problem and (iv) maximum flow problem.

- To formulate the network models under five different uncertain paradigms: (i) type-2 fuzzy, (ii) random fuzzy, (iii) rough, (iv) rough fuzzy and (v) uncertainty theory.
- To model the network problems using three uncertain programming techniques: (i) expected value model (EVM), (ii) chance-constrained model (CCM) and (iii) dependent chance-constrained model (DCCM).
- To develop the crisp equivalents of the corresponding uncertain programming models of the network problems.
- To implement some classical methods (Dijkstra's and Kruskal's algorithms, modified rough Dijkstra's algorithm, goal attainment method, linear weighted method, global criterion method, epsilon-constraint method and fuzzy programming method) as solution methodologies for some of the proposed uncertain network problems.
- To implement three evolutionary algorithms: (i) varying population GA with improved lifetime allocation strategy (iLAS), (ii) nondominated sorting genetic algorithm II (NSGA-II) (Deb et al. 2002) and (iii) multi-objective cross generational elitist selection, heterogeneous recombination, and cataclysmic mutation (MOCHC) (Nebro et al. 2007) as solution methodologies for some of the proposed uncertain network problems.

## 1.3 Preliminary Concepts

In this section, we discuss some basic concepts, theorems and methodologies which are relevant to the studies presented in chapters 2 through 6.

### 1.3.1 Fuzzy Set

Introduced by L.A. Zadeh (1965), the fuzzy set theory provides an excellent framework to represent the uncertainty and ambiguity prevalent in human interpretation. In contrast to Boolean set or "crisp set", where membership value of an element is restricted to either 1 or 0, the fuzzy set allows the membership value in a range between 0 and 1.

**Definition 1.3.1**: Let $X$ be a collection of elements (universe of discourse) and $x$ an element of $X$, then a fuzzy set $\tilde{A}$ in $X$ is defined by the membership function $\mu_{\tilde{A}}(x)$, $(0 \leq \mu_{\tilde{A}}(x) \leq 1)$, defined on every element $x$ of $\tilde{A}$. The fuzzy set $\tilde{A}$ is denoted by $\tilde{A} = \{(x, \mu_{\tilde{A}}(x)): x \in X\}$ or $\tilde{A} = \sum \frac{\mu_{\tilde{A}}(x)}{x}$ , $x \in X$ or $\tilde{A} = \int_{x \in X} \mu_{\tilde{A}}(x)/x$.

**Definition 1.3.2** (B. Liu 2002): Let $\Theta$ is a non-empty set and $\Gamma(\Theta)$ is a power set of $\Theta$. For each $\mathcal{A} \in \Gamma(\Theta)$, there is a non-negative number $Pos\{\mathcal{A}\} \in \mathfrak{R}$ called the possibility of the fuzzy event $\{\tilde{\zeta} \in \mathcal{A}\}$, such that

(i) $Pos\{\emptyset\} = 0$, $Pos\{\Theta\} = 1$ and

(ii) $Pos\{\cup_k \mathcal{A}_k\} = sup_k Pos\{\mathcal{A}_k\}$ for any arbitrary collection $\{\mathcal{A}_k\}$ in $\Gamma(\Theta)$.

Then, the triplet $(\Theta, \Gamma(\Theta), Pos)$ is known as possibility space and the function $Pos$: $\Theta \mapsto [0,1]$ is known as the possibility measure (P. Wang 1982).

**Definition 1.3.3** (S. Nahmias 1978) A fuzzy variable is defined as a function from a possibility space $(\Theta, \Gamma(\Theta), Pos)$ to a set of real numbers $\mathfrak{R}$ to describe a fuzzy phenomenon.

**Example 1.3.1**: A fuzzy variable $\tilde{\zeta}$ is said to be a trapezoidal fuzzy number (TrFN) if it is represented as a quadruple $\tilde{\zeta} = \mathcal{T}r(\zeta_1, \zeta_2, \zeta_3, \zeta_4)$ of crisp numbers with $\xi_1 < \xi_2 \leq \xi_3 < \zeta_4$, whose membership function $\mu_{\tilde{\zeta}}(x \in \mathfrak{R})$ is given by

$$\mu_{\tilde{\zeta}}(x \in \mathfrak{R}) = \begin{cases} \frac{x-\zeta_1}{\zeta_2-\zeta_1} & ; if\ \zeta_1 \leq x < \zeta_2 \\ 1 & ; if\ \zeta_2 \leq x \leq \zeta_3 \\ \frac{\zeta_4-x}{\zeta_4-\zeta_3} & ; if\ \zeta_3 < x \leq \zeta_4 \\ 0 & ; otherwise. \end{cases}$$

The membership function of a TrFN $\tilde{\zeta} = \mathcal{T}r(4,6,9,11)$ is presented in Fig. 1.1.

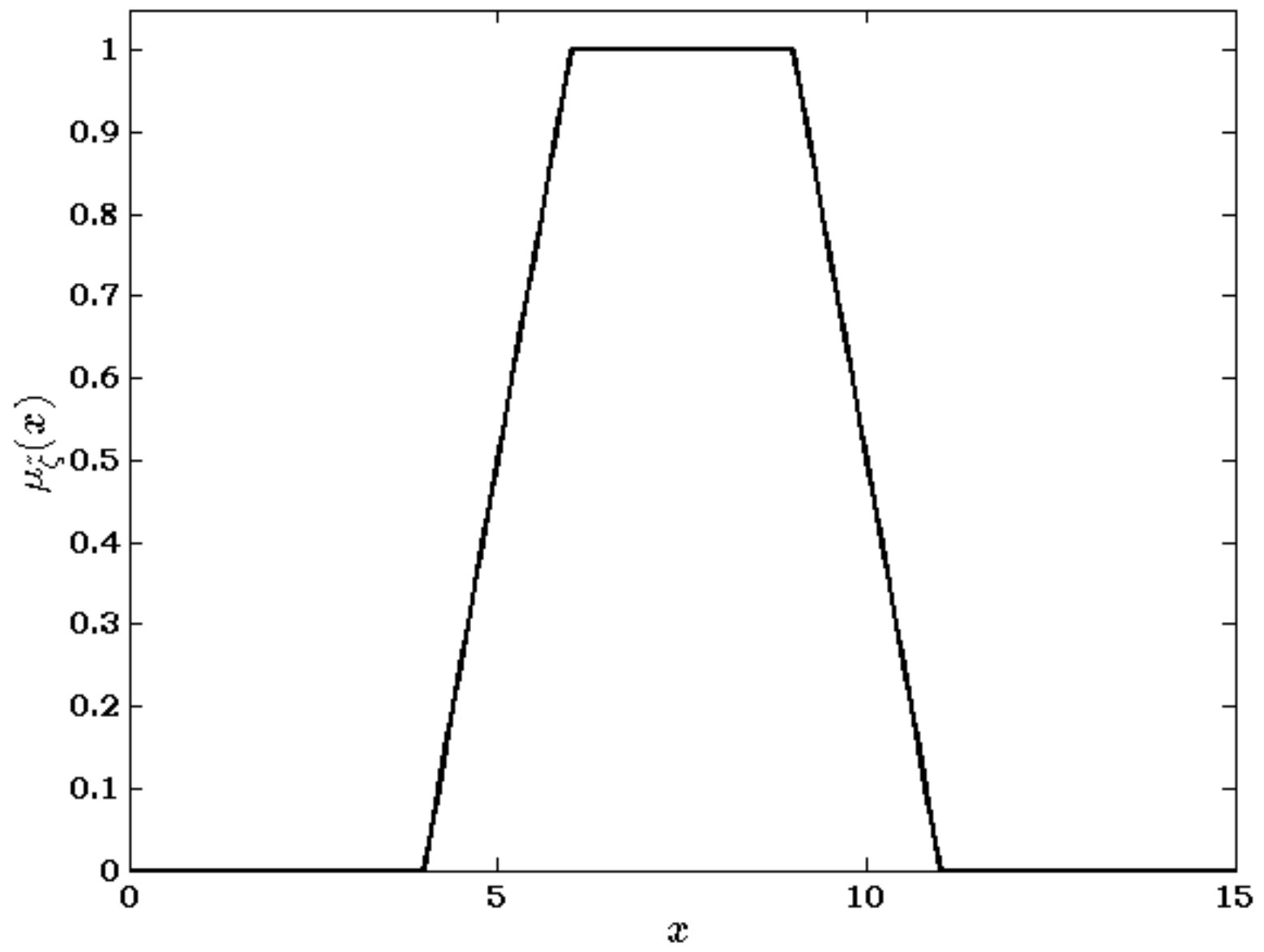

Figure 1.1 Membership function of the TrFN $\tilde{\zeta} = \mathcal{T}r(4,6,9,11)$

**Example 1.3.2**: A triangular fuzzy number (TFN) is a fuzzy variable if it is determined by a triplet $\tilde{\eta} = \mathcal{T}(\eta_1, \eta_2, \eta_3)$ of crisp numbers, such that $\eta_1 < \eta_2 < \eta_3$ and is characterized by a membership function $\mu_{\tilde{\eta}}(x \in \mathfrak{R})$ so that

$$\mu_{\tilde{\eta}}(x \in \mathfrak{R}) = \begin{cases} \frac{x-\eta_1}{\eta_2-\eta_1} & ; if\ \eta_1 \leq x < \eta_2 \\ \frac{\eta_3-x}{\eta_3-\eta_2} & ; if\ \eta_2 \leq x \leq \eta_3 \\ 0 & ; otherwise. \end{cases}$$

For a TFN $\tilde{\eta} = \mathcal{T}(6,8,10)$, the graphical representation of a membership function is presented in Fig. 1.2.

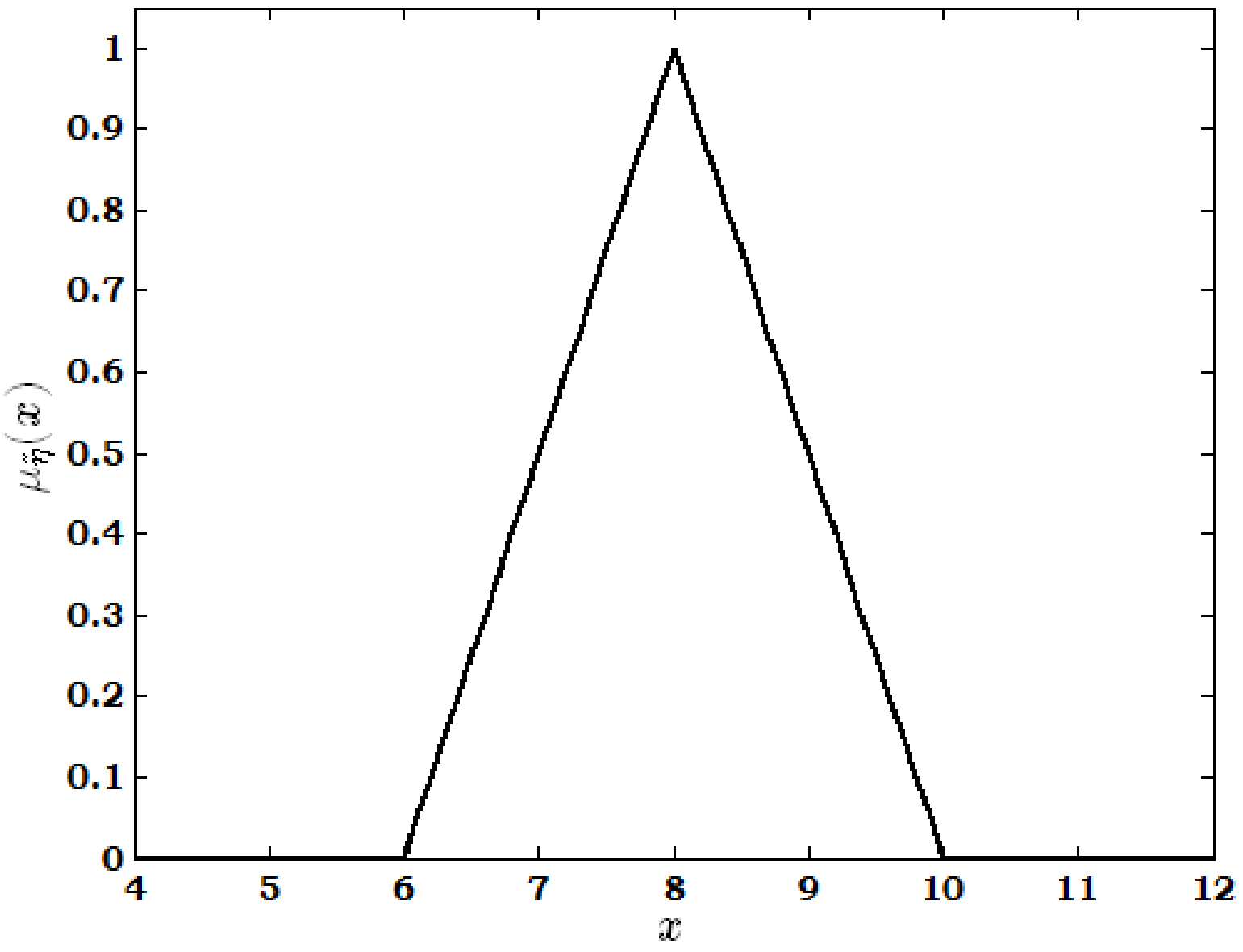

Figure 1.2 Membership function of the TFN $\tilde{\eta} = \mathcal{T}(6,8,10)$

**Example 1.3.3** (Yi et al. 2016; Zhou et al. 2016b): A Gaussian fuzzy number (GFN) is a fuzzy variable if it is determined by $\tilde{\xi} = \mathcal{N}(\rho, \delta)$ of crisp numbers, such that $\rho$, $\delta \in \Re$ and $\delta > 0$ with a membership function $\mu_{\tilde{\xi}}(x \in \Re)$ defined as

$$\mu_{\tilde{\xi}}(x) = exp\left\{-\left(\frac{x-\rho}{\delta}\right)^2\right\}, x \in \Re, \delta > 0.$$

The membership function of $\tilde{\xi} = \mathcal{N}(5, 2)$ is depicted in Fig. 1.3.

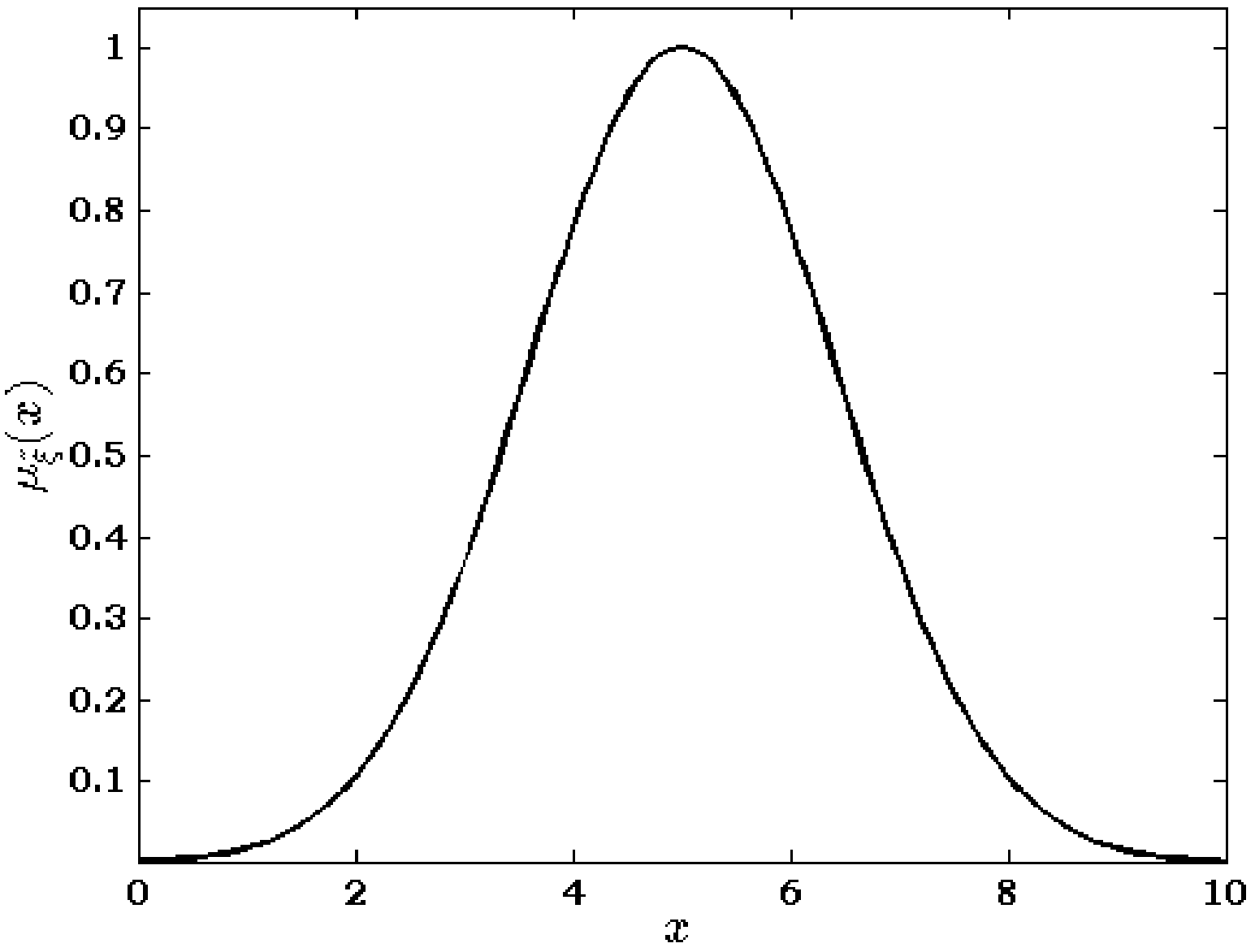

Figure 1.3 Membership function of the GFN $\tilde{\xi} = \mathcal{N}(5, 2)$

**Example 1.3.4**: A generalized trapezoidal fuzzy variable (GTrFV) is a TrFN and represented as $\tilde{\nu} = [\nu_1, \nu_2, \nu_3, \nu_4; w]$, where $\nu_1, \nu_2, \nu_3, \nu_4$ and $w$ are the crisp numbers such that $\nu_1 < \nu_2 \leq \nu_3 < \nu_4$, $0 < w \leq 1$ and the membership function of $\tilde{\nu}$ is given by

$$\mu_{\tilde{v}}(x \in \Re) = \begin{cases} \frac{w(x-v_1)}{v_2-v_1} & ; if\ v_1 \le x < v_2 \\ w & ; if\ v_2 \le x \le v_3,\ 0 < w \le 1 \\ \frac{w(v_4-x)}{v_4-v_3} & ; if\ v_3 < x \le v_4 \\ 0 & ; otherwise. \end{cases}$$

Here, $w$ is called the height of $\tilde{v}$. If $w = 1$ (i.e., normalized), then $\tilde{v}$ becomes a TrFN.

**Definition 1.3.4** (Liu and Liu 2002): Let $\tilde{\zeta}$ be a fuzzy variable on the possibility space $(\Theta,\ \Gamma(\Theta),\ Pos)$, then the expected value of $\tilde{\zeta}$ is defined as below, provided that at least one of the two integrals is finite.

$$E[\tilde{\zeta}] = \int_0^{+\infty} Cr\{\tilde{\zeta} \ge x\}dx - \int_{-\infty}^{0} Cr\{\tilde{\zeta} \le x\}dx \tag{1.3}$$

**Theorem 1.3.1** (Zhou et al. 2016b): Let $\tilde{\zeta}$ be a fuzzy variable with inverse credibility distribution function $\Psi^{-1}$. If the expected value of $\tilde{\zeta}$ exists, then

$$E[\tilde{\zeta}\,] = \int_{\mathbf{0}}^{\mathbf{1}} \Psi^{-1}(\alpha)\, d\alpha \tag{1.4}$$

### 1.3.2 Possibility, Necessity and Credibility Measures

The occurrence of a fuzzy event is determined by two measures, possibility and necessity as defined below.

**Definition 1.3.5** (L.A. Zadeh 1978): The possibility measure ($Pos$) of a fuzzy event $\{\tilde{\xi} \in B\}$, $B \subset \Re$ is defined as $Pos\{\tilde{\xi} \in B\} = sup_{x\in B}\mu_{\tilde{\xi}}(x)$, and necessity measure ($Nec$) which is dual of possibility, is defined as $Nec\{\tilde{\xi} \in B\} = 1 - Pos\{\tilde{\xi} \in B^c\}= 1 - sup_{x\in B^c}\mu_{\tilde{\xi}}(x)$.

According to the above definition, we can obtain the possibility and necessity measures of fuzzy event $\{\tilde{\xi} \ge x\}$ as shown in (1.5) and (1.6), respectively, where $\tilde{\xi} = (\xi_1, \xi_2, \xi_3, \xi_4)$ is a trapezoidal fuzzy number.

$$Pos\{\tilde{\xi} \le x\} = \begin{cases} 1 & ; if\ \xi_2 < x \\ \frac{x-a}{b-a} & ; if\ \xi_1 \le x \le \xi_2 \\ 0 & ; otherwise \end{cases} \tag{1.5}$$

$$Nec\{\tilde{\xi} \le x\} = \begin{cases} 1 & ; if\ \xi_4 < x \\ \frac{x-c}{d-c} & ; if\ \xi_3 \le x \le \xi_4 \\ 0 & ; otherwise. \end{cases} \tag{1.6}$$

**Definition 1.3.6** (Liu and Liu 2002): Let $(\Theta,\ \Gamma(\Theta),\ Pos)$ is a possibility space, and $B \in \Gamma(\Theta)$ be a fuzzy event. Then the credibility measure of $B$ is expressed as

$$Cr\{\tilde{\xi} \in B\} = \frac{1}{2}\left(Pos\{\tilde{\xi} \in B\} + Nec\{\tilde{\xi} \in B\}\right). \tag{1.7}$$

It directly follows from (1.5) that $Cr\{\tilde{\xi} \in B\} = \frac{1}{2}\left(1 + \sup_{x\in B} \mu_{\tilde{\xi}}(x) - \sup_{x\in B^c} \mu_{\tilde{\xi}}(x)\right)$. Here, it is to be noted that this credibility measure is applicable for normalized fuzzy variable $\tilde{\xi}$, i.e., $\sup_{x\in\Re} \mu_{\tilde{\xi}}(x) = 1$.

**Remark 1.3.1**: If the fuzzy variable is not normalized, then the usual credibility measure cannot be used. For this reason, in the case of a generalized fuzzy variable, a generalized credibility measure is employed. A generalized credibility measure $\tilde{C}r$ of a fuzzy event $\{\tilde{\zeta} \in B\},\ B \subset \Re$ is defined as

$$\tilde{C}r\{\tilde{\zeta} \in B\} = \frac{1}{2}\left(\sup_{x\in\Re} \mu_{\tilde{\zeta}}(x) + \sup_{x\in B} \mu_{\tilde{\zeta}}(x) - \sup_{x\in B^c} \mu_{\tilde{\zeta}}(x)\right) \tag{1.8}$$

If $\tilde{\zeta}$ is normalized, then $\tilde{C}r$ coincides with usual credibility measure $Cr$ because then $\sup_{x\in\Re} \mu_{\tilde{\zeta}}(x) = 1$.

**Definition 1.3.7** (B. Liu 2002): The credibility distribution of a TrFN, $\tilde{\zeta} = \mathcal{T}r[\zeta_1, \zeta_2, \zeta_3, \zeta_4]$ is expressed as

$$Cr\{\tilde{\zeta} \le x\} = \begin{cases} 0 & ; if\ x < \zeta_1 \\ \frac{x-\zeta_1}{2(\zeta_2-\zeta_1)} & ; if\ \zeta_1 \le x < \zeta_2 \\ \frac{1}{2} & ; if\ \zeta_2 \le x < \zeta_3 \\ \frac{x+\zeta_4-2\zeta_3}{2(\zeta_4-\zeta_3)} & ; if\ \zeta_3 \le x < \zeta_4 \\ 1 & ; if\ x \ge \zeta_4 \end{cases} \tag{1.9}$$

and

$$Cr\{\tilde{\zeta} \ge x\} = \begin{cases} 1 & ; if\ x < \zeta_1 \\ \frac{2\zeta_2-x-\zeta_1}{2(\zeta_2-\zeta_1)} & ; if\ \zeta_1 \le x < \zeta_2 \\ \frac{1}{2} & ; if\ \zeta_2 \le x < \zeta_3 \\ \frac{\zeta_4-x}{2(\zeta_4-\zeta_3)} & ; if\ \zeta_3 \le x < \zeta_4 \\ 0 & ; if\ x \ge \zeta_4 \end{cases} \tag{1.10}$$

**Remark 1.3.2**: Following Definition 1.3.7, similarly, we can determine the credibility distribution of a TFN $\tilde{\zeta} = \mathcal{T}[\zeta_1, \zeta_2, \zeta_3]$ as well, which are presented in (1.11) and (1.12).

$$Cr\{\tilde{\zeta} \le x\} = \begin{cases} 0 & ; if\ x < \zeta_1 \\ \frac{x-\zeta_1}{2(\zeta_2-\zeta_1)} & ; if\ \zeta_1 \le x < \zeta_2 \\ \frac{x+\zeta_3-2\zeta_2}{2(\zeta_3-\zeta_2)} & ; if\ \zeta_2 \le x < \zeta_3 \\ 1 & ; if\ x \ge \zeta_3 \end{cases} \tag{1.11}$$

and

$$Cr\{\tilde{\zeta} \geq x\} = \begin{cases} 1 & ; if\ x < \zeta_1 \\ \frac{2\zeta_2 - \zeta_1 - x}{2(\zeta_2 - \zeta_1)} & ; if\ \zeta_1 \leq x < \zeta_2 \\ \frac{\zeta_3 - x}{2(\zeta_3 - \zeta_2)} & ; if\ \zeta_2 \leq x < \zeta_3 \\ 0 & ; if\ x \geq \zeta_3 \end{cases} \tag{1.12}$$

**Theorem 1.3.2** (Gao and Lu 2005): Let $\tilde{\zeta} = \mathcal{T}r[\zeta_1, \zeta_2, \zeta_3, \zeta_4]$ is a TrFN and $\alpha$ is the predetermined confidence level, then

(i) while $\alpha \in (0,0.5]$, $Cr\{\xi \leq x\} \geq \alpha$ if and only if $x \geq (1 - 2\alpha)\xi_1 + 2\alpha\xi_2$

(ii) while $\alpha \in (0.5,1]$, $Cr\{\xi \leq x\} \geq \alpha$ if and only if $x \geq (2 - 2\alpha)\xi_3 + (2\alpha - 1)\xi_4$.

**Remark 1.3.3**: If $\tilde{\zeta} = \mathcal{T}[\zeta_1, \zeta_2, \zeta_3]$ is a TFN with a predetermined confidence level $\alpha$, then

(i) while $\alpha \in (0,0.5]$, $Cr\{\xi \leq x\} \geq \alpha$ if and only if $x \geq (1 - 2\alpha)\xi_1 + 2\alpha\xi_2$

(ii) while $\alpha \in (0.5,1]$, $Cr\{\xi \leq x\} \geq \alpha$ if and only if $x \geq (2 - 2\alpha)\xi_2 + (2\alpha - 1)\xi_3$.

**Lemma 1.3.1**: Let $\tilde{\zeta} = \mathcal{T}r[\zeta_1, \zeta_2, \zeta_3, \zeta_4]$ be a TrFN and $\alpha(\in [0,1])$ is the given confidence level, then

(i) if $\alpha \leq \frac{1}{2}$, then $Cr\{\tilde{\zeta} \geq x\} \geq \alpha$ if and only if, $(1 - 2\,\alpha)\zeta_4 + 2\alpha\zeta_3 \geq x$

(ii) if $\alpha > \frac{1}{2}$, then $Cr\{\tilde{\zeta} \geq x\} \geq \alpha$ if and only if, $(2 - 2\,\alpha)\zeta_2 + (2\alpha - 1)\zeta_1 \geq x$.

**Proof**:

(i) When $\alpha \leq \frac{1}{2}$ and $Cr\{\tilde{\zeta} \geq x\} \geq \alpha$, then we have either $\zeta_3 \leq x < \zeta_4$ or $\left\{\frac{\zeta_4 - x}{2(\zeta_4 - \zeta_3)}\right\} \geq \alpha$. Considering, $\left\{\frac{\zeta_4 - x}{2(\zeta_4 - \zeta_3)}\right\} \geq \alpha \Leftrightarrow (1 - 2\,\alpha)\zeta_4 + 2\alpha\zeta_3 \geq x$. Again, $(1 - 2\,\alpha)\zeta_4 + 2\alpha\zeta_3 \geq x$ holds good for $\zeta_2 \leq x < \zeta_3$ since $\alpha \leq \frac{1}{2}$. Conversely, when $\zeta_3 \leq x$, then $Cr\{\tilde{\zeta} \geq x\} \geq \alpha$ if $Cr\{\tilde{\zeta} \geq x\}$ is at least $\frac{1}{2}$. Therefore, $(1 - 2\,\alpha)\zeta_4 + 2\alpha\zeta_3 \geq x \Leftrightarrow \left\{\frac{\zeta_4 - x}{2(\zeta_4 - \zeta_3)}\right\} \geq \alpha \Leftrightarrow Cr\{\tilde{\zeta} \geq x\} \geq \alpha$.

(ii) When $\alpha > \frac{1}{2}$ and $Cr\{\tilde{\zeta} \geq x\} \geq \alpha$, then we observe that either $x \leq \zeta_1$ or $\left\{\frac{2\zeta_2 - x - \zeta_1}{2(\zeta_2 - \zeta_1)}\right\} \geq \alpha$. Considering, each of the observed cases, we have $(2 - 2\,\alpha)\zeta_2 + (2\alpha - 1)\zeta_1 \geq x$. Conversely, when $x < \zeta_1$ then $Cr\{\tilde{\zeta} \geq x\} = 1 \geq \alpha$. Now $(2 - 2\,\alpha)\zeta_2 + (2\alpha - 1)\zeta_1 \geq x \Leftrightarrow \left\{\frac{2\zeta_2 - x - \zeta_1}{2(\zeta_2 - \zeta_1)}\right\} \geq \alpha$, which eventually means $Cr\{\tilde{\zeta} \geq x\} \geq \alpha$.

**Definition 1.3.8** (Zhou et al. 2016b): The credibility distribution of a GFN $\tilde{\zeta} = \mathcal{N}(\rho, \delta)$ is represented as

$$Cr\{\tilde{\zeta} \leq x\} = \begin{cases} \frac{1}{2} exp\left\{-\left(\frac{\rho - x}{\delta}\right)^2\right\} & ; if\ x \leq \rho \\ 1 - \frac{1}{2} exp\left\{-\left(\frac{x - \rho}{\delta}\right)^2\right\} & ; if\ x > \rho \end{cases} \tag{1.13}$$

and

$$Cr\{\tilde{\zeta} \geq x\} = \begin{cases} 1 - \frac{1}{2}exp\left\{-\left(\frac{\rho-x}{\delta}\right)^2\right\} & ; if\ x \leq \rho \\ \frac{1}{2}exp\left\{-\left(\frac{x-\rho}{\delta}\right)^2\right\} & ; if\ x > \rho. \end{cases} \tag{1.14}$$

**Lemma 1.3.2**: Let $\tilde{\zeta} = \mathcal{N}(\rho, \delta)$ be a GFN and $\alpha(\in [0,1])$ is the given confidence level. Then

(i) if $\alpha > \frac{1}{2}$, then $Cr\{\tilde{\zeta} \geq x\} \geq \alpha$ if and only if, $\rho - \delta\sqrt{-ln(2-2\alpha)} \geq x$

(ii) if $\leq \frac{1}{2}$, then $Cr\{\tilde{\zeta} \geq x\} \geq \alpha$ if and only if, $\rho + \delta\sqrt{-ln(2\alpha)} \geq x$.

**Proof**:

(i) When $\alpha > \frac{1}{2}$ and $Cr\{\tilde{\zeta} \geq x\} \geq \alpha$, then we have, $\left[1 - \frac{1}{2}exp\left\{-\left(\frac{\rho-x}{\delta}\right)^2\right\}\right] \geq \alpha$ $\Leftrightarrow \rho - \delta\sqrt{-ln(2-2\alpha)} \geq x$. Conversely, when $x \leq \rho$ then $Cr\{\tilde{\zeta} \geq x\} \geq \alpha$ is satisfied with $\alpha > \frac{1}{2}$. Therefore, $\rho - \delta\sqrt{-ln(2-2\alpha)} \geq x \Leftrightarrow \left[1 - \frac{1}{2}exp\left\{-\left(\frac{\rho-x}{\delta}\right)^2\right\}\right] \geq \alpha \Leftrightarrow$ $Cr\{\tilde{\zeta} \geq x\} \geq \alpha$.

(ii) When $\alpha \leq \frac{1}{2}$ and $Cr\{\tilde{\zeta} \geq x\} \geq \alpha$, then $\left[\frac{1}{2}exp\left\{-\left(\frac{x-\rho}{\delta}\right)^2\right\}\right] \geq \alpha \Leftrightarrow \rho +$ $\delta\sqrt{-ln(2\alpha)} \geq x$. Conversely, when $x > \rho$ then $Cr\{\tilde{\zeta} \geq x\} \geq \alpha$ is satisfied with $\alpha \leq$ $\frac{1}{2}$. Accordingly, $\rho + \delta\sqrt{-ln(2\alpha)} \geq x \Leftrightarrow \left[\frac{1}{2}exp\left\{-\left(\frac{\rho-x}{\delta}\right)^2\right\}\right] \geq \alpha$.

This eventually means, $\rho + \delta\sqrt{-ln(2\alpha)} \geq x \Leftrightarrow Cr\{\tilde{\zeta} \geq x\} \geq \alpha$.

### 1.3.3 Type-2 Fuzzy Set

Some basic concepts related to the type-2 fuzzy set (T2FS) are presented here. A type-2 fuzzy set $\tilde{A}$ in a universe of discourse $X$ is a fuzzy set in which the membership function is also fuzzy, i.e., the membership grade of each element is no longer a crisp value but a fuzzy set. This membership function is known as type-2 membership function.

**Definition 1.3.9** (Mendel and John 2002): A T2FS $\tilde{A}$ is defined as

$$\tilde{A} = \{((x,u), \mu_{\tilde{A}}(x,u)) : \forall x \in X, \forall u \in J_x \subseteq [0,1]\}, \tag{1.15}$$

where $0 \leq \mu_{\tilde{A}}(x,u) \leq 1$ is the type-2 membership function, $J_x$ is the primary membership of $x \in X$ which is the domain of the secondary membership function $\tilde{\mu}_{\tilde{A}}(x)$. $\tilde{A}$ is also expressed as

$$\tilde{A} = \int_{x \in X} \int_{u \in J_x} \mu_{\tilde{A}}(x,u)/(x,u),\ J_x \subseteq [0,1], \tag{1.16}$$

where $\int\int$ denotes union over all admissible $x$ and $u$. For a discrete universe of discourse, $\int$ is replaced by $\sum$.

For each value of $x$, the secondary membership function $\tilde{\mu}_{\tilde{A}}(x)$ is defined as

$$\tilde{\mu}_{\tilde{A}}(x) = \int_{u \in J_x} \mu_{\tilde{A}}(x,u)/u, \tag{1.17}$$

where for a particular value $u = u' \in J_x$, $\mu_{\tilde{A}}(x,u')$ is called secondary membership grade of $(x,u')$.

**Definition 1.3.10** (Mendel et al. 2006): If all the secondary membership grades of a T2FS are 1, i.e., $\mu_{\tilde{A}}(x,u) = 1, \forall\,(x,u)$, then $\tilde{A}$ is called interval type-2 fuzzy set (IT2FS) which is represented in (1.18) and depicted in Fig. 1.4.

$$\tilde{A} = \int_{x \in X} \int_{u \in J_x} 1/(x,u)\,,\; J_x \subseteq [0,1]. \tag{1.18}$$

**Definition 1.3.11** (Mendel and John 2002): An interval type-2 fuzzy set is characterized by the footprint of uncertainty (FOU) which is the union of all primary memberships $J_x$ in a bounded region. If $\tilde{A}$ is an IT2FS, then its corresponding FOU is defined as

$$FOU(\tilde{A}) = \cup_{x \in X} J_x. \tag{1.19}$$

The FOU is bounded by an upper membership function (UMF) $\overline{\mu}_{\tilde{A}}(x)$ and a lower membership function (LMF) $\underline{\mu}_{\tilde{A}}(x)$, both of which are the membership functions of type-1 fuzzy set (T1FS), such that $J_x = \left[\underline{\mu}_{\tilde{A}}(x),\ \overline{\mu}_{\tilde{A}}(x)\right]$, $\forall x \in X$. Therefore, an IT2FS can be represented by $\left(\tilde{A}^U,\ \tilde{A}^L\right)$, where $\tilde{A}^U$ and $\tilde{A}^L$ are the T1FSs (may or may not be normalized). The FOU is a complete description of an IT2FS which is graphically represented in Fig. 1.4.

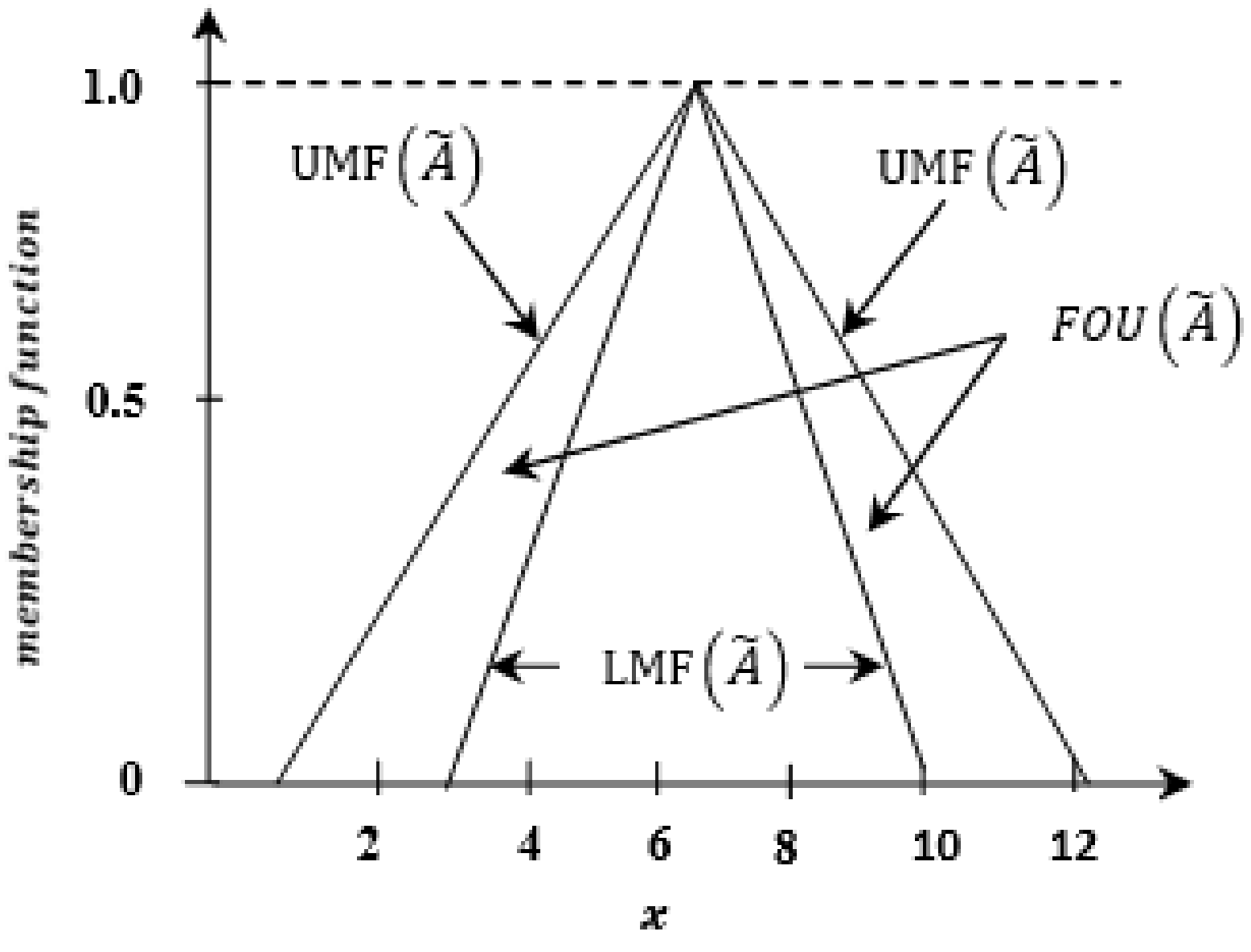

Figure 1.4 Interval type-2 fuzzy set $\tilde{A}$

### 1.3.4 Type-2 Fuzzy Variable

Similar to a type-1 fuzzy variable (T1FV), which is defined as a function from the possibility space to the set of real numbers, a type-2 fuzzy variable (T2FV) (Liu and Liu 2010; Qin et al. 2011) is defined as a function from the fuzzy possibility space to the set of real numbers. If $(\Theta, \Gamma(\Theta), \widetilde{Pos})$ is a fuzzy possibility space (Liu and Liu 2010), then a T2FV $\tilde{\xi}$ is defined as a map from $\Theta$ to $\Re$ such that for any $t \in \Re$, the set $\{\gamma \in \Theta | \tilde{\xi}(\gamma) \leq t\}$ is an element of $p$, i.e., $\{\gamma \in \Theta | \tilde{\xi}(\gamma) \leq t\} \in p$. Then $\tilde{\mu}_{\tilde{\xi}}(x)$ is called secondary possibility distribution function of $\tilde{\xi}$, and is defined as a map $\Re \mapsto \Re[0,1]$ such that $\tilde{\mu}_{\tilde{\xi}}(x) = \widetilde{Pos}\{\gamma \in \Theta | \tilde{\xi}(\gamma) = x\}$, $x \in \Re$. $\mu_{\tilde{\xi}}(x,u)$ is called type-2 possibility distribution function, which is a map $\Re \times J_x \mapsto [0,1]$, and defined as $\mu_{\tilde{\xi}}(x,u) = Pos\,\{\tilde{\mu}_{\tilde{\xi}}(x) = u\}$, $(x,u) \in \Re \times J_x$.

Here, if $\mu_{\tilde{\xi}}(x,u) = 1, \forall (x,u) \in \Re \times J_x$, then $\tilde{\xi}$ is called an interval type-2 fuzzy variable (IT2FV).

**Definition 1.3.12**: A trapezoidal interval type-2 fuzzy variable (TrIT2FV) $\tilde{A}$ in the universe of discourse $X$ is represented as

$$\tilde{A} = (\tilde{A}^U, \tilde{A}^L) = \left((a_1^U, a_2^U, a_3^U, a_4^U; w^U), (a_1^L, a_2^L, a_3^L, a_4^L; w^L)\right),$$

where both $\tilde{A}^U$ and $\tilde{A}^L$ are trapezoidal fuzzy numbers with heights $w^U$ and $w^L$, respectively. As an example, let us consider a TrIT2FV $\tilde{A}$=$\left((2,4,6,8;1),\ (3,4.5,5.5,7;0.8)\right)$ whose membership function is shown in Fig. 1.5.

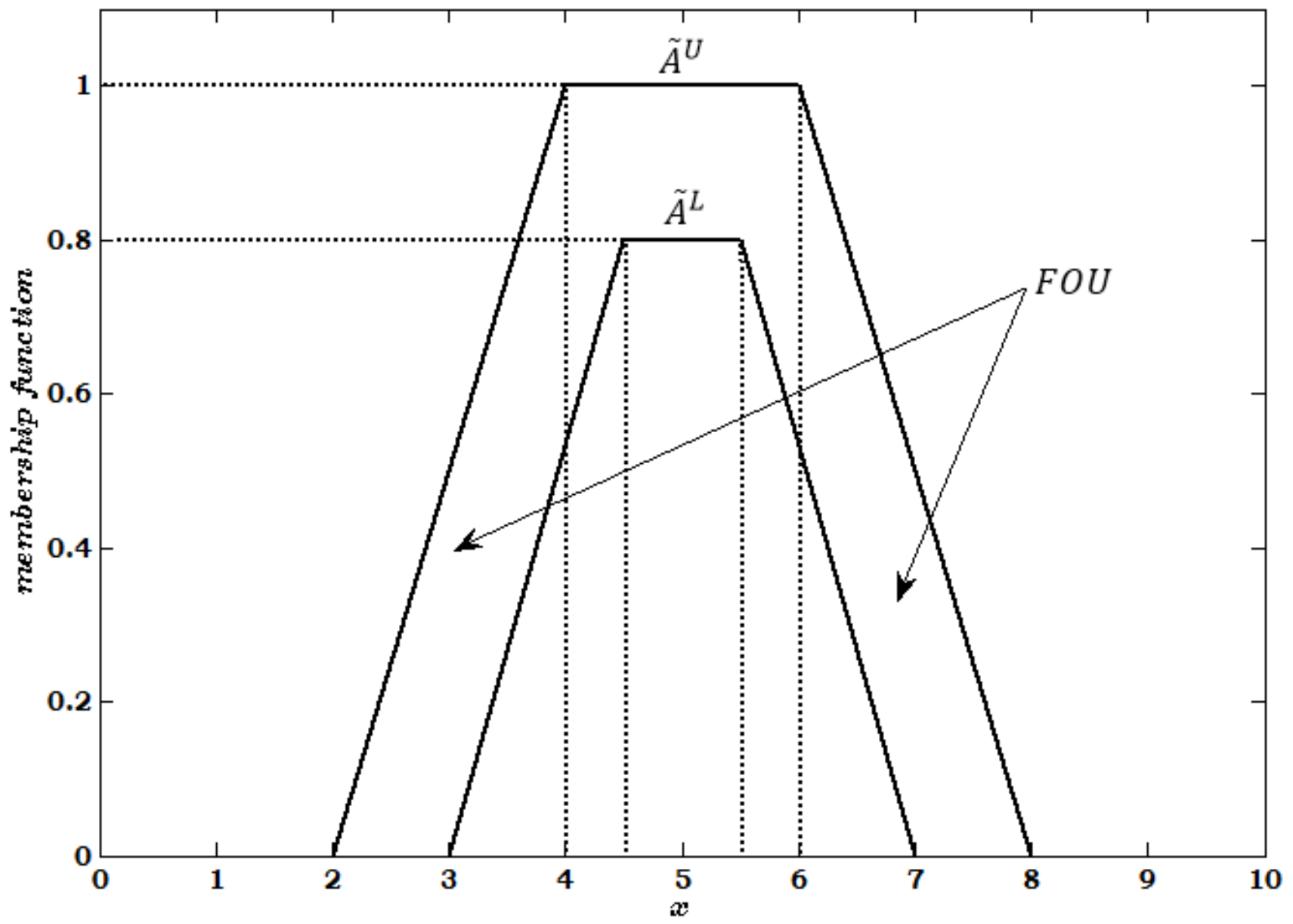

Figure 1.5 Graphical interpretation of the TrIT2FV $\tilde{A}$

### 1.3.5 Random Fuzzy Variable

B. Liu (2001) introduced the concept of a random fuzzy variable, which is a fuzzy variable on the universal set of random variables. In this section, we present some definitions related to the random fuzzy variable.

**Definition 1.3.13** (B. Liu 2001, 2002): A random fuzzy variable is a function from possibility space $(\Theta,\ \Gamma(\Theta),\ Pos)$ to a set of random variables $\mathcal{R}$.

**Definition 1.3.14** (B. Liu 2001, 2002): Let $\tilde{\tilde{\zeta}}_i$ be random fuzzy variables defined on the possibility spaces $(\Theta_i,\ \Gamma(\Theta_i),\ Pos_i)$, $i = 1,2,3 \dots, n$, respectively, and $f: \Re^n \mapsto \Re$ a measurable function. Then, $\tilde{\tilde{\zeta}} = \left(\tilde{\tilde{\zeta}}_1, \tilde{\tilde{\zeta}}_2, \dots, \tilde{\tilde{\zeta}}_n\right)$ is a random fuzzy variable on the product possibility space $(\Theta,\ \Gamma(\Theta),\ Pos)$, defined as

$$\tilde{\tilde{\zeta}}(\theta_1, \theta_2, \dots, \theta_n) = f\left(\tilde{\tilde{\zeta}}_1(\theta_1), \tilde{\tilde{\zeta}}_2(\theta_2), \dots, \tilde{\tilde{\zeta}}_n(\theta_n)\right) \quad (1.20)$$

for all $(\theta_1, \theta_2, \dots, \theta_n) \in \Theta$.

**Definition 1.3.15** (B. Liu 2002; Liu and Liu 2003): Let $\tilde{\tilde{\zeta}}$ be a random fuzzy variable defined on the possibility space $(\Theta,\ \Gamma(\Theta),\ Pos)$. Then the expected value $E\left[\tilde{\tilde{\zeta}}\right]$ of $\tilde{\tilde{\zeta}}$ is defined by

$$E\left[\tilde{\tilde{\zeta}}\right] = \int_0^{+\infty} Cr\left\{\theta \in \Theta | E\left[\tilde{\tilde{\zeta}}(\theta)\right] \geq x\right\} dx - \int_{-\infty}^{0} Cr\left\{\theta \in \Theta | E\left[\tilde{\tilde{\zeta}}(\theta)\right] \leq x\right\} dx \quad (1.21)$$

provided that one of the two integrals is finite.

**Remark 1.3.4** (B. Liu 2002): If the random fuzzy variable $\tilde{\tilde{\zeta}}$ degenerates to a random variable, then the expected value operator becomes

$$E\left[\tilde{\tilde{\zeta}}\right] = \int_0^{+\infty} Pr\left\{\tilde{\tilde{\zeta}} \geq x\right\} dx - \int_{-\infty}^{0} Pr\left\{\tilde{\tilde{\zeta}} \leq x\right\} dx. \quad (1.22)$$

**Remark 1.3.5** (B. Liu 2002): If the random fuzzy variable $\tilde{\tilde{\zeta}}$ degenerates to a fuzzy variable, then the expected value operator becomes

$$E\left[\tilde{\tilde{\zeta}}\right] = \int_0^{+\infty} Cr\left\{\tilde{\tilde{\zeta}} \geq x\right\} dx - \int_{-\infty}^{0} Cr\left\{\tilde{\tilde{\zeta}} \leq x\right\} dx. \quad (1.23)$$

**Definition 1.3.16** (B. Liu 2001, 2002): Let $\tilde{\tilde{\zeta}} = \left(\tilde{\tilde{\zeta}}_1, \tilde{\tilde{\zeta}}_2, \dots, \tilde{\tilde{\zeta}}_n\right)$ be a random fuzzy vector on the possibility space $(\Theta,\ \Gamma(\Theta),\ Pos)$, and $f_p: \Re^n \mapsto \Re$ be continuous functions, $p = 1,2, \dots, k$. Then the primitive chance at predetermined confidence labels $\alpha$ and $\beta$, for a random fuzzy event characterized by $f_p\left(\tilde{\tilde{\zeta}}\right) \leq 0,\ p = 1,2, \dots, k$ is a function from [0,1] to [0,1] and is defined below in (1.24).

$$Ch\left\{f_p\left(\tilde{\bar{\zeta}}\right) \le 0, p = 1,2,\dots,k\right\}(\alpha)=sup\left\{\beta|Pos\left\{\theta \in \Theta|\, Pr\left\{\begin{matrix} f_p\left(\tilde{\bar{\zeta}}(\theta)\right)\le 0 \\ p=1,2,\dots,k \end{matrix}\right\} \ge \beta\right\} \ge \alpha\right\} \quad (1.24)$$

**Remark 1.3.6**: In this thesis, the random fuzzy chance-constrained programming model has been developed using the definition of the primitive chance as defined in (1.25), where a random fuzzy event is measured with a credibility measure and a probability measure, which are represented as $Cr$ and $Pr$, respectively.

$$Ch\left\{f_p\left(\tilde{\bar{\zeta}}\right) \le 0, p = 1,2,\dots,k\right\}(\alpha)=sup\left\{\beta|Cr\left\{\theta \in \Theta|\, Pr\left\{\begin{matrix} f_p\left(\tilde{\bar{\zeta}}(\theta)\right)\le 0 \\ p=1,2,\dots,k \end{matrix}\right\} \ge \beta\right\} \ge \alpha\right\} \quad (1.25)$$

### 1.3.6 Rough Set

Rough set theory, introduced by Z. Pawlak (1982), is a mathematical approach to model imperfect knowledge. The rudimentary conjecture in rough set theory is to recognize every object in the universe on the basis of information (data, knowledge). Rough set models vagueness in a different way altogether. Unlike fuzzy set, rough set expresses the vagueness in terms of boundary region of a set. A rough set is represented by using two approximation sets, the lower approximation set and the upper approximation set.

Let $X$ be the universe of discourse of non-empty finite set of objects and $A$ be the non-empty finite set of attributes, then the pair $S = (X, A)$ is said to be an information system. For any $B \subseteq A$, there exists an indiscernibility (similarity) relation $R(B)$ such that $R(B) = \{(x, y) \in X \times X|\ \forall a \in B, a(x) = a(y)\}$, where $a(x)$ denote the value of an attribute $a$ for the element $x$. $R(B)$ is called the $B$-indiscernibility relation having equivalence classes $[x]_R$. We define the lower approximation and the upper approximation of the set $X$ in order to classify the elements of set $X$.

An information system $S = (X, A)$, $B \subseteq A, T \subseteq X$ can be approximated based on the information of $B$ into $B$-lower approximation and $B$-upper approximation (Z. Pawlak 1982) of $T$, which are defined as

$$\underline{B}(T) = \{x|[x]_B \subseteq T\}$$

and

$$\overline{B}(T) = \{x|[x]_B: [x]_B \cap T \neq \emptyset\}.$$

Note that the objects in $\underline{B}(T)$ become the certain members of $T$, whereas objects in $\overline{B}(T)$ become the possible members of $T$. The $B$ −boundary region of $T$ is defined as

$$BN_B(T) = \overline{B}(T) - \underline{B}(T).$$

Fig. 1.6 depicts a schematic representation of a rough set $S$, which is defined on equivalence classes. The equivalence classes are determined by the small squares in the figure. All the equivalence classes which are fully and partially contained in $S$ represents the upper approximation set $\overline{B}(S)$ of $S$, whereas the equivalence classes entirely belonging to $S$ represent its lower approximation set $\underline{B}(S)$. The boundary region of $S$, $BN_B(S)$ represents the equivalence classes which are in $\overline{B}(S)$ but not in $\underline{B}(S)$.

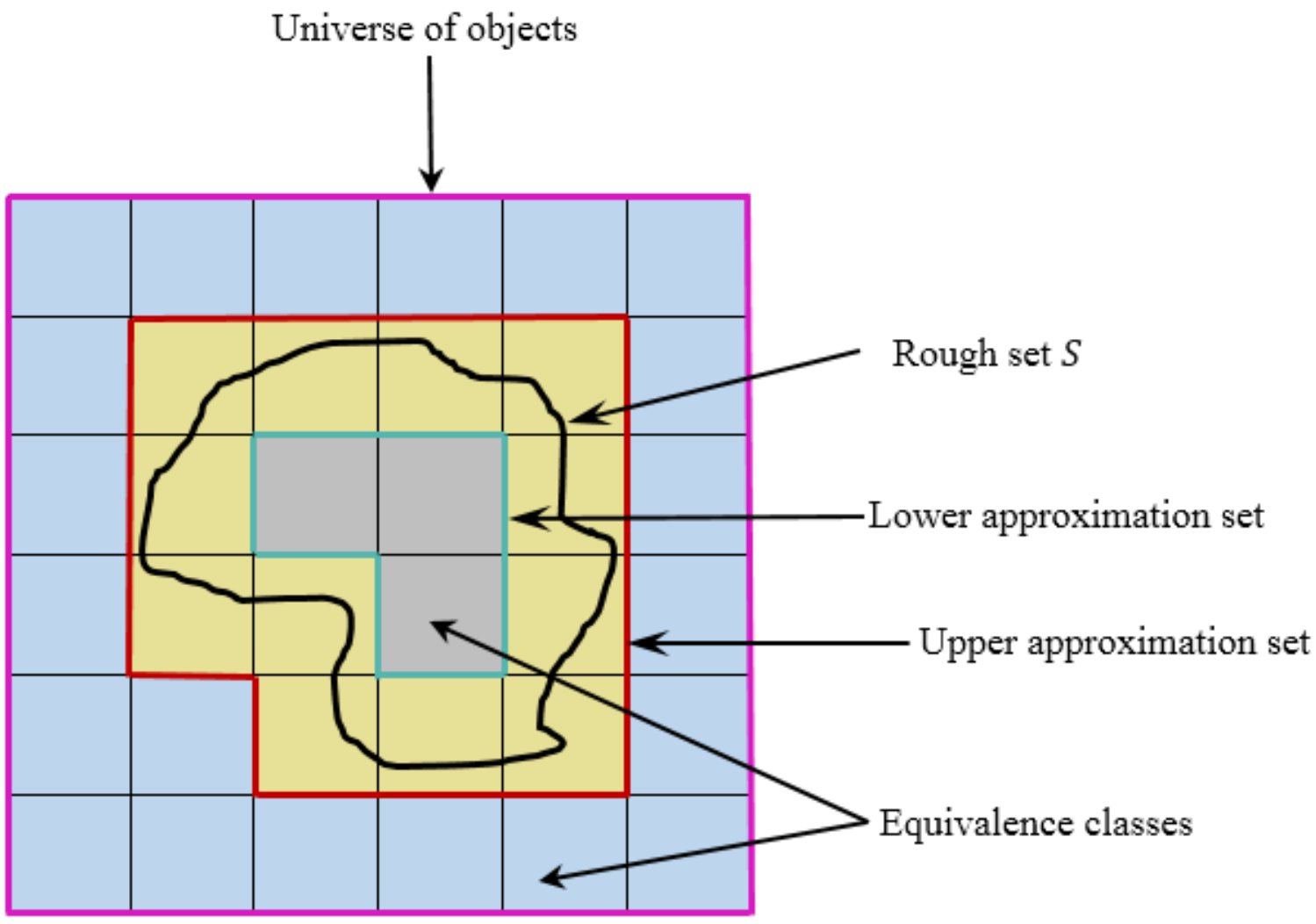

Figure 1.6 A rough set $S$

### 1.3.7 Rough Variable

Słowiński and Vanderpooten (2000) extended the concept of indiscernibility (equivalence) by considering the situation where objects are not significantly distinct. They showed that such a situation could be modelled by defining a binary similarity relation ($\simeq$).

By inheriting the concept of the binary similarity relation as mentioned in the study of Słowiński and Vanderpooten (2000), B. Liu (2002, 2004) put forward the concept of rough space and rough variable and also determine different properties of rough variable.

The binary similarity relation in Słowiński and Vanderpooten (2000) is considered as reflexive, non-symmetric and non-transitive, which unlike equivalence relation does not create partitions on $X$, the universe of discourse. As an example, we consider the similarity relation defined on $\Re$ as "$p$ is similar to $q$ if and only if, $(p - q) \leq 26.1$". Information about similarity can be inferred from similarity classes for each object $p \in X$. The similarity class of $p$, denoted by $R(p)$, determines all the objects similar to $p$ and is expressed as

$$R(p) = \{q \in X | q \simeq p\}.$$

Let $R^{-1}(p)$ be the class of objects to which $p$ is similar. Then $R^{-1}(p)$ can be defined as

$$R^{-1}(p) = \{q \in X|\, p \simeq q\}.$$

The lower and upper approximations of the set based on binary similarity relation $R$, proposed by Słowiński and Vanderpooten (2000), are defined as follows.

**Definition 1.3.17** (Słowiński and Vanderpooten 2000): Considering $U \subseteq X$ and a binary similarity relation $R$ defined on $X$, the lower approximation of $U$, denoted by $\underline{U}$, is defined as

$$\underline{U} = \{x \in X|R^{-1}(x) \subseteq U\},$$

while the upper approximation of $U$, denoted by $\overline{U}$, is defined as

$$\overline{U} = \cup_{x\in U} R(x).$$

Thus, the lower approximation is a subset containing the objects which surely belong to the set $U$, whereas the upper approximation is a superset containing the objects which possibly belong to $U$.

**Theorem 1.3.3** (Słowiński and Vanderpooten 2000): Consider a subset $U \subseteq X$ and a binary similarity relation $R$ defined on $X$, then, $\underline{U} \subseteq U \subseteq \overline{U}$.

**Example 1.3.5**: We consider a set of products which are sequentially arranged in decreasing order of their respective prices. We define the similarity relation $\simeq$ on $\mathfrak{R}$, such that the cost of product $p \simeq$ cost of product $q$ if and only if $c(p) - c(q) \leq \$10$, where $c(\cdot)$ is the manufacturing cost of a product. The cost of a product may vary over a time interval due to the fluctuation in cost of raw materials and transportation costs. The existence of such factors directly influences the fluctuation of the manufacturing cost of the product. In order to capture such fluctuation, $c(\cdot)$ can be represented by intervals. If the cost of the product is around \$80 which is estimated by an expert as the set [60, 90], then the corresponding lower and upper approximations with respect to '$\simeq$' are $\underline{[60,90]} = [70,80]$ and $\overline{[60,90]} = [50,100]$, respectively.

B. Liu (2004) defined some axioms and certain definitions, which are stated below, while introducing the concept of rough variable.

Let $\Lambda$ be a non-empty set, $\mathcal{A}$ be a $\sigma$-algebra of subsets of $\Lambda$, $\Delta$ be an element in $\mathcal{A}$ and $\rho$ be the non-negative real-valued additive set function on $\mathcal{A}$, then the axioms are as follows.

**Axiom 1**: $\rho\{\Lambda\} < +\infty$

**Axiom 2**: $\rho\{\Delta\} > 0$

**Axiom 3**: $\rho\{A\} \geq 0$ for any $A \in \mathcal{A}$

**Axiom 4**: For every countable sequence of mutually disjoint events $\{A_i\}_{i=1}^{\infty}$, we have, $\rho\{\cup_{i=1}^{\infty} A_i\} = \sum_{i=1}^{\infty} \rho\{A_i\}$ .

**Definition 1.3.18** (B. Liu 2002): Let $\Lambda$ be a non-empty set, $\mathcal{A}$ a $\sigma$-algebra of subsets of $\Lambda$, $\Delta$ be an element in $\mathcal{A}$ and $\rho$ be the non-negative real-valued additive set function on $\mathcal{A}$ satisfying Axiom 1 through Axiom 4. Then the quadruple $(\Lambda,\ \Delta, \mathcal{A}, \rho)$ is called a rough space.

**Definition 1.3.19** (B. Liu 2002): A rough variable $\xi$ on a rough space $(\Lambda,\ \Delta, \mathcal{A}, \rho)$ is a measurable function from $\Lambda$ to the set of real numbers $\mathfrak{R}$ such that for every Borel set $\mathcal{B}$ of $\mathfrak{R}$ we have, $\{\lambda \in \Lambda | \zeta(\lambda) \in \mathcal{B}\} \in \mathcal{A}$. Then the lower and upper approximations of the rough variable $\zeta$ are represented as follow.

$$\underline{\zeta} = \{\zeta(\lambda) | \lambda \in \Delta\}$$

and

$$\overline{\zeta} = \{\zeta(\lambda) | \lambda \in \Lambda\}$$

Since $\Delta \subset \Lambda$, it becomes obvious that $\underline{\zeta} \subset \overline{\zeta}$. The extreme cases for $\Delta$ are $\Delta = \emptyset$ and $\Delta = \Lambda$. If $\Delta = \emptyset$, then $\underline{\zeta}$ will be a null set. If $\Delta = \Lambda$, then $\underline{\zeta} = \overline{\zeta}$, and $\zeta$ becomes an ordinary set.

**Example 1.3.6**: Let $\Lambda = \{x | 0 \leq x \leq 80\}$ and $\Delta = \{x | 20 \leq x \leq 50\}$, then the function, $\zeta(x) = x | (x \ mod\ 6) = 0$ is a rough variable defined on $(\Lambda, \Delta, \mathcal{A}, \rho)$.

**Example 1.3.7**: Let us consider a rough variable $\zeta = ([a, b], [c, d])$, $c \leq a \leq b \leq d$, defined on the rough space $(\Lambda, \Delta, \mathcal{A}, \rho)$, where $[a, b]$ and $[c, d]$ are lower and upper approximations of $\zeta$, respectively. This eventually means that the representing elements of $[a, b]$ are certainly the members of $\zeta$ and that belonging to $[c, d]$ are the possible members of $\zeta$. Here $\Lambda = \{x | c \leq x \leq d\}$, $\Delta = \{x | a \leq x \leq b\}$ and $\zeta(x) = x, \forall x \in \Lambda$. $\mathcal{A}$ is the $\sigma$-algebra on $\Lambda$ and $\rho$ is the Lebesgue measure.

### 1.3.8 Trust Measure of Rough Variable

**Definition 1.3.20** (B. Liu 2002): Let $\zeta$ be a rough vector on a rough space $(\Lambda, \Delta, \mathcal{A}, \rho)$ and $\tau_k$: $\mathfrak{R}^n \to \mathfrak{R}$ be the continuous function, $k = 1,2, \ldots, m$. Then the upper trust measure and the lower trust measure of the rough event, represented by $\tau_k(\zeta) \leq 0, k = 1,2, \ldots, m$, are defined as follows.

$$\overline{Tr}\{\tau_k(\zeta) \leq 0, k = 1,2, \ldots, m\} = \frac{\rho\{\lambda \in \Lambda | \tau_k(\zeta(\lambda)) \leq 0,\ \ k = 1,2, \ldots, m\}}{\rho\{\Lambda\}}. \tag{1.26}$$

$$\underline{Tr}\{\tau_k(\zeta) \leq 0, k = 1,2, \ldots, m\} = \frac{\rho\{\lambda \in \Delta | \tau_k(\zeta(\lambda)) \leq 0,\ \ k = 1,2, \ldots, m\}}{\rho\{\Delta\}}. \tag{1.27}$$

If $\rho(\Delta) = 0$, then $\overline{Tr}\{\tau_k(\zeta) \leq 0,\ k = 1,2, \ldots, m\} \equiv \underline{Tr}\{\tau_k(\zeta) \leq 0,\ k = 1,2, \ldots, m\}$.

The trust measure of a rough event is defined as the mean of lower and upper trusts as given below.

$$Tr\{\tau_k(\zeta) \le 0, k = 1,2,\dots,m\} = \frac{\left(\overline{Tr}\{\tau_k(\zeta) \le 0,\ k = 1,2,\dots,m\} + \underline{Tr}\{\tau_k(\zeta) \le 0,\ k = 1,2,\dots,m\}\right)}{2}. \tag{1.28}$$

**Remark 1.3.7** (B. Liu 2002): The value of upper trust, lower trust and trust of a rough event, characterized by $\{\tau_k(\zeta) \le 0, k = 1,2,\dots,m\}$, are within $[0,1]$. A rough event must hold if its trust is 1 and it fails when its trust is 0.

For a given value $r$ and a rough variable $\zeta = ([a,b],[c,d]), c \le a \le b \le d,$ define on a rough space $(\Lambda,\ \Delta, \mathcal{A}, \rho)$, the trust distribution (B. Liu 2002) of $\zeta \le r$ and $\zeta \ge r$ are defined in (1.29) and (1.30), respectively.

$$Tr\{\zeta \le r\} = \begin{cases} 0 & ;if\ r \le c \\ \frac{r-c}{2(d-c)} & ;if\ c \le r \le a \\ \frac{1}{2}\left(\frac{r-a}{b-a} + \frac{r-c}{d-c}\right) & ;if\ a \le r \le b \\ \frac{1}{2}\left(1 + \frac{r-c}{d-c}\right) & ;if\ b \le r \le d \\ 1 & ;if\ r \ge d. \end{cases} \tag{1.29}$$

$$Tr\{\zeta \ge r\} = \begin{cases} 0 & ;if\ r \ge d \\ \frac{d-r}{2(d-c)} & ;if\ b \le r \le d \\ \frac{1}{2}\left(\frac{d-r}{d-c} + \frac{b-r}{b-a}\right) & ;if\ a \le r \le b \\ \frac{1}{2}\left(1 + \frac{d-r}{d-c}\right) & ;if\ c \le r \le a \\ 1 & ;if\ r \le c. \end{cases} \tag{1.30}$$

**Remark 1.3.8** (B. Liu 2002): If the rough variable $\zeta$ degenerates to an interval number $[a,b]$, then the trust distributions of $\zeta \le r$ and $\zeta \ge r$ are shown respectively in (1.31) and (1.32).

$$Tr\{\zeta \le r\} = \begin{cases} 0 & ;if\ r \le a \\ \frac{r-a}{b-a} & ;if\ a \le r \le b \\ 1 & ;if\ b \le r. \end{cases} \tag{1.31}$$

$$Tr\{\zeta \ge r\} = \begin{cases} 0 & ;if\ b \le r \\ \frac{b-r}{b-a} & ;if\ a \le r \le b \\ 1 & ;if\ r \le a. \end{cases} \tag{1.32}$$

**Example 1.3.8**: For a rough variable $\zeta = [6,10],[2,14]$ the distribution of the trust measures $Tr\{\zeta \le r\}$ and $Tr\{\zeta \ge r\}$ are displayed respectively in Fig. 1.7 and Fig. 1.8.

### 1.3.9 Optimistic and Pessimistic Values of Rough Variable

**Definition 1.3.21** (B. Liu 2002): Let $\zeta$ be a rough variable defined on a rough space $(\Lambda, \Delta, \mathcal{A}, \rho)$ and $\alpha \in (0,1]$, then $\alpha$-optimistic and $\alpha$-pessimistic values of $\zeta$, denoted by $\zeta_{sup}(\alpha)$ and $\zeta_{inf}(\alpha)$ are respectively, defined as

$$\zeta_{sup}(\alpha) = sup\{r|Tr\{\zeta \geq r\} \geq \alpha\} \tag{1.33}$$

and

$$\zeta_{inf}(\alpha) = inf\{r|Tr\{\zeta \leq r\} \geq \alpha\}. \tag{1.34}$$

For a rough variable $\zeta = ([a,b],[c,d]), c \leq a \leq b \leq d$ and $\alpha \in [0,1]$, the corresponding $\alpha$-optimistic and $\alpha$-pessimistic values are as follow.

$$\zeta_{sup}(\alpha) = \begin{cases} (1-2\alpha)d + 2\alpha c & ;\ \alpha \leq \frac{d-b}{2(d-c)} \\ 2(1-\alpha)d + (2\alpha-1)c & ;\alpha \geq \frac{2d-a-c}{2(d-c)} \\ \frac{d(b-a)+b(d-c)-2\alpha(b-a)(d-c)}{(b-a)+(d-c)} & ; otherwise. \end{cases} \tag{1.35}$$

$$\zeta_{inf}(\alpha) = \begin{cases} (1-2\alpha)c + 2\alpha d & ;\alpha \leq \frac{a-c}{2(d-c)} \\ 2(1-\alpha)c + (2\alpha-1)d & ;\alpha \geq \frac{b+d-2c}{2(d-c)} \\ \frac{c(b-a)+a(d-c)+2\alpha(b-a)(d-c)}{(b-a)+(d-c)} & ; otherwise. \end{cases} \tag{1.36}$$

**Remark 1.3.9** (B. Liu 2002): If $\zeta$ degenerates to an interval number $[a,b]$, then the corresponding $\alpha$-optimistic and $\alpha$-pessimistic values of $\zeta$ are respectively represented by

$$\zeta_{sup}(\alpha) = \alpha a + (1-\alpha)b$$

and

$$\zeta_{inf}(\alpha) = (1-\alpha)a + \alpha b.$$

**Theorem 1.3.4** (B. Liu 2002): Let $\zeta_{sup}(\alpha)$ and $\zeta_{inf}(\alpha)$ are $\alpha$-optimistic and $\alpha$-pessimistic values of a rough variable $\zeta$, defined on a rough space $(\Lambda, \Delta, \mathcal{A}, \rho)$, and $\alpha \in [0,1]$, then

(i) if $\alpha \leq 0.5$, then $\zeta_{sup}(\alpha) \geq \zeta_{inf}(\alpha)$

(ii) if $\alpha > 0.5$, then $\zeta_{sup}(\alpha) \leq \zeta_{inf}(\alpha)$.

**Example 1.3.9**: We consider the rough variable $\zeta = ([6,10],[2,14])$ as mentioned in Example 1.3.8. Then, for $\alpha = 0.4$ the $\zeta_{sup}(\alpha)$ and $\zeta_{inf}(\alpha)$ values of $\zeta$ are 8.6 and 7.4, respectively. Furthermore, when $\alpha = 0.9$, then $\zeta_{sup}(\alpha) = 4.4$ and $\zeta_{inf}(\alpha) = 11.6$. Here, we observe that $\zeta_{sup}(\alpha) \geq \zeta_{inf}(\alpha)$ when $\alpha$ is less or equal to 0.5, and $\zeta_{sup}(\alpha) \leq \zeta_{inf}(\alpha)$ when $\alpha$ is greater than 0.5.

### 1.3.10 Rough Fuzzy Variable

Initialized by B. Liu (2002), a rough fuzzy variable is a fuzzy variable defined on the universal set of rough variables. In this section, we present some definitions of rough fuzzy variables pertinent to our thesis.

**Definition 1.3.22** (B. Liu 2002, 2004): A rough fuzzy variable $\hat{\xi}$ is a function from a possibility space $(\Theta, \Gamma(\Theta), Pos)$ to a set of rough variables.

**Definition 1.3.23** (B. Liu 2002, 2004): Let $\hat{\xi}$ be a $n$-dimensional rough fuzzy vector. Then $\hat{\xi}$ is a function from possibility space $(\Theta, \Gamma(\Theta), Pos)$ to a set of $n$-dimensional rough vectors.

**Definition 1.3.24** (B. Liu 2002, 2004): Let $u: \Re^n \rightarrow \Re$ be a function and $\hat{\xi}_1, \hat{\xi}_2, \dots, \hat{\xi}_n$ are the rough fuzzy variables on possibility space $(\Theta, \Gamma(\Theta), Pos)$. Then $\hat{\xi} = u(\hat{\xi}_1, \hat{\xi}_2, \dots, \hat{\xi}_n)$ is a rough fuzzy variable and defined as

$$\hat{\xi}(\theta) = u(\hat{\xi}_1(\theta), \hat{\xi}_2(\theta), \dots, \hat{\xi}_n(\theta)), \forall\, \theta \in \Theta. \tag{1.37}$$

**Definition 1.3.25** (B. Liu 2002, 2004): Let $u: \Re^n \rightarrow \Re$ is a continuous function, and $\xi_i$ be a rough fuzzy variable defined on the possibility spaces $(\Theta_i, \Gamma(\Theta_i), Pos_i)$, $i = 1,2,3, \dots, n$, respectively. Then $\hat{\xi} = u(\hat{\xi}_1, \hat{\xi}_2, \dots, \hat{\xi}_n)$ is a rough fuzzy variable defined on the product possibility space $(\Theta, \Gamma(\Theta), Pos)$ such that

$$\hat{\xi}(\theta_1, \theta_2, \dots, \theta_n) = u(\hat{\xi}_1(\theta_1), \hat{\xi}_2(\theta_2), \dots, \hat{\xi}_n(\theta_n)), \forall\, \theta_i \in \Theta. \tag{1.38}$$

**Definition 1.3.26** (B. Liu 2002): Let $u_j: \Re^n \rightarrow \Re, j = 1,2, \dots, m$ are the continuous functions and $\hat{\xi} = (\hat{\xi}_1, \hat{\xi}_2, \dots, \hat{\xi}_n)$ be a rough fuzzy vector defined on possibility space$(\Theta, \Gamma(\Theta), Pos)$, $i = 1,2,3 \dots, n$. Then the primitive chance at predetermined confidence labels $\alpha$ and $\beta$, for a rough fuzzy event characterized by $u_j(\hat{\xi}) \leq 0, j = 1,2, \dots, m$, is a measurable map from [0,1] to [0,1] and is defined below in (1.39).

$$Ch\{u_j(\hat{\xi}) \leq 0, j = 1,2, \dots, m\}(\beta) = \sup\left\{\alpha \middle| Cr\left\{\theta \in \Theta \middle| Tr\left\{\begin{matrix} u_j(\hat{\xi}) \leq 0 \\ j = 1,2, \dots, m \end{matrix}\right\} \geq \alpha\right\} \geq \beta\right\}. \tag{1.39}$$

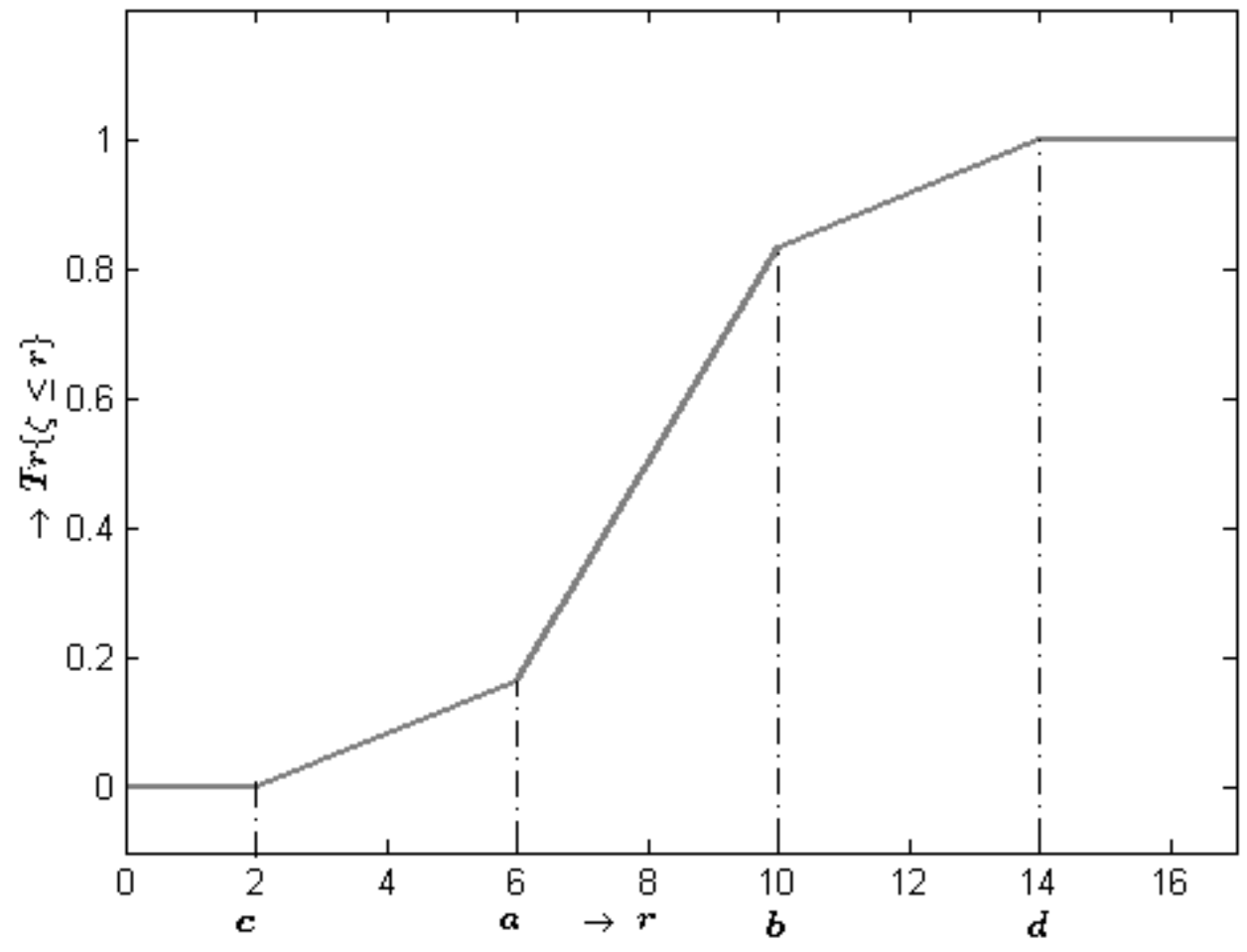

Figure 1.7 Trust distribution of a rough event $\zeta \leq r$

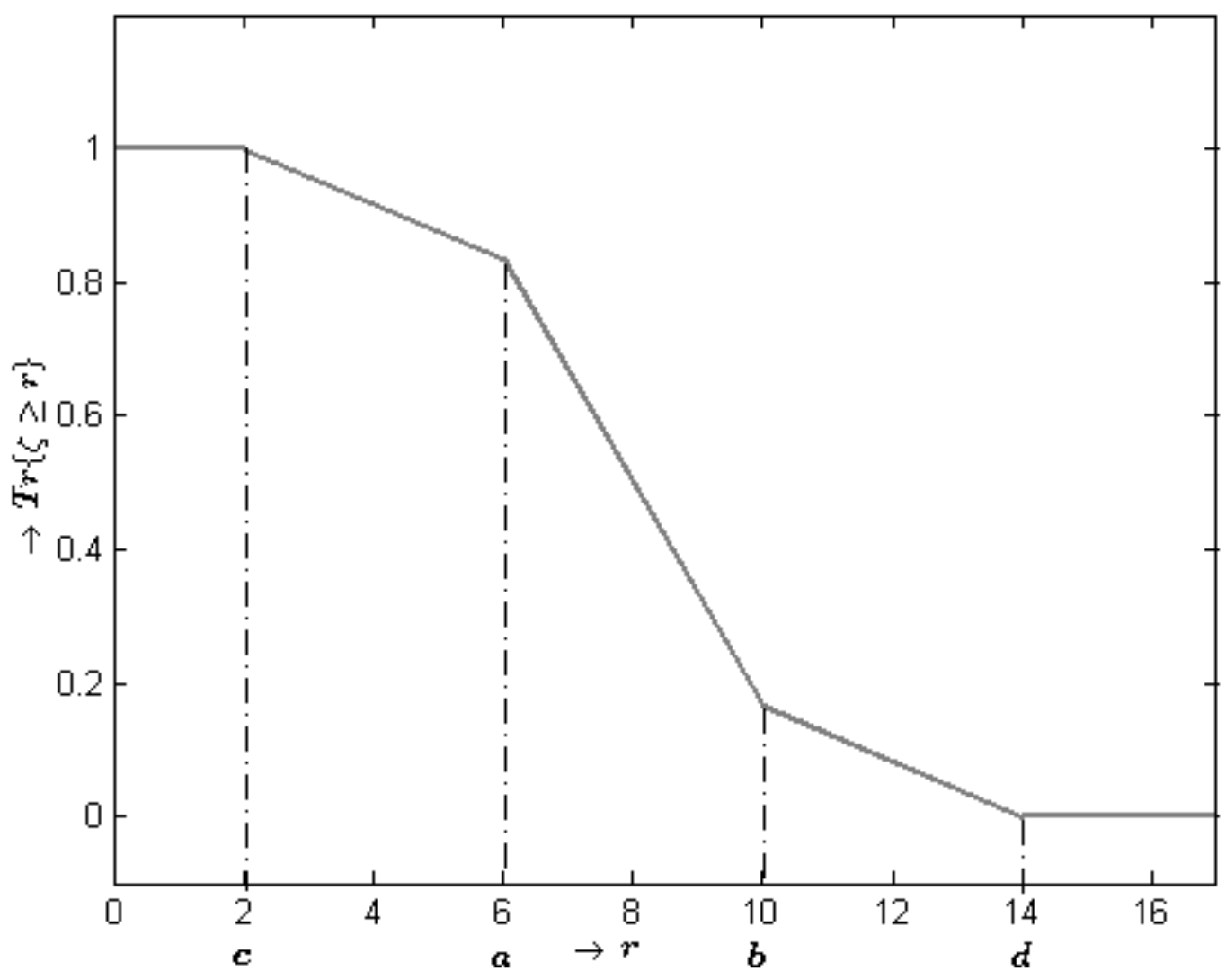

Figure 1.8 Trust distribution of a rough event $\zeta \geq r$

### 1.3.11 Uncertainty Theory

Recently, some recent studies (B. Liu 2012; X. Huang 2012; Chen et al. 2016) reveal that paradoxical results occur if the subjective estimation of an imprecise parameter is represented by a fuzzy variable. To alleviate those paradoxical outcomes, B. Liu (2007) introduced uncertainty theory (B. Liu 2007) and further refined it in B. Liu (2009, 2010). The uncertainty theory is a branch of axiomatic mathematics for modelling human uncertainty. In this section, we recall some basic concepts of uncertainty theory, which are relevant to this thesis.

**Definition 1.3.27** (B. Liu 2007): Let $\Gamma$ be a non-empty set, $\mathcal{L}$ be a $\sigma$-algebra over $\Gamma$ and $\mathcal{M}$ be the uncertain measure of any element and $\Omega$ is contained in $\mathcal{L}$. Then $(\Gamma, \mathcal{L}, \mathcal{M})$ represents an uncertainty space.

In an uncertainty space $(\Gamma, \mathcal{L}, \mathcal{M})$, for each event $\Omega$ in $\mathcal{L}$, $\mathcal{M}$ is the uncertain measure which maps $\mathcal{L}$ to $\Re(0,1)$, i.e., $\mathcal{M}\colon \mathcal{L} \mapsto \Re(0,1)$. In order to define the uncertain measure axiomatically, each event $\Lambda$ of $\mathcal{L}$ is assigned a belief degree $\mathcal{M}\{\Lambda\}$, where $0 \leq \mathcal{M}\{\Lambda\} \leq 1$. Subsequently, to establish certain mathematical properties of $\mathcal{M}\{\Lambda\}$ axiomatically, the following four axioms are proposed by B. Liu (2007).

**Axiom 1.** (*Normality*) $\mathcal{M}\{\Gamma\} = 1$ for the universal set $\Gamma$.

**Axiom 2. (***Self-Duality***)** $\mathcal{M}\{\Omega\} + \mathcal{M}\{\Omega^{c}\} = 1$ for any event $\Omega$.

**Axiom 3.** (*Subadditivity*) For every countable sequence of events $\Omega_1$, $\Omega_2, \ldots$, we have $\mathcal{M}\{\bigcup_{i=1}^{\infty} \Omega_i\} \leq \sum_{i=1}^{\infty} \mathcal{M}\{\Omega_i\}$.

**Axiom 4.** (*Product*) Let $(\Gamma_j, \mathcal{L}_j, \mathcal{M}_j)$ be the uncertainty spaces for $j = 1,2,\ldots$, then the product uncertain measure $\mathcal{M}$ is an uncertain measure satisfying, $\mathcal{M}\{\prod_{j=1}^{\infty} \Omega_j\} = \bigwedge_{j=1}^{\infty} \mathcal{M}_j\{\Omega_j\}$, where $\Omega_j$ is arbitrary chosen event from $\mathcal{L}_j$, for every $j = 1,2,\ldots$, respectively.

**Definition 1.3.28** (B. Liu 2007): An uncertain variable $\zeta$ is a function from an uncertainty space $(\Gamma, \mathcal{L}, \mathcal{M})$ to a set of real numbers $\Re$, i.e., for any Borel set $\mathcal{B}$ of $\Re$ the set $\{\zeta \in \mathcal{B}\}$ in (1.40) is an event.

$$\{\zeta \in \mathcal{B}\} = \{\gamma \in \Gamma | \zeta(\gamma) \in \mathcal{B}\} \tag{1.40}$$

**Definition 1.3.29** (B. Liu 2007): An uncertain variable $\zeta$ is said to be linear uncertain variable if $\zeta$ follows a linear uncertainty distribution as defined below.

$$\Phi(x) = \begin{cases} 0 & ; if\ x < a \\ \frac{x-a}{b-a} & ; if\ a \le x < b \\ 1 & ; if\ x \ge b, \end{cases} \tag{1.41}$$

where $\zeta$ is denoted by $\mathcal{L}(a, b)$ such that $a < b$ and $a, b \in \Re$.

**Example 1.3.10**: The uncertainty distribution of linear uncertainty variable $\mathcal{L}(6,14)$ is depicted in Fig. 1.9.

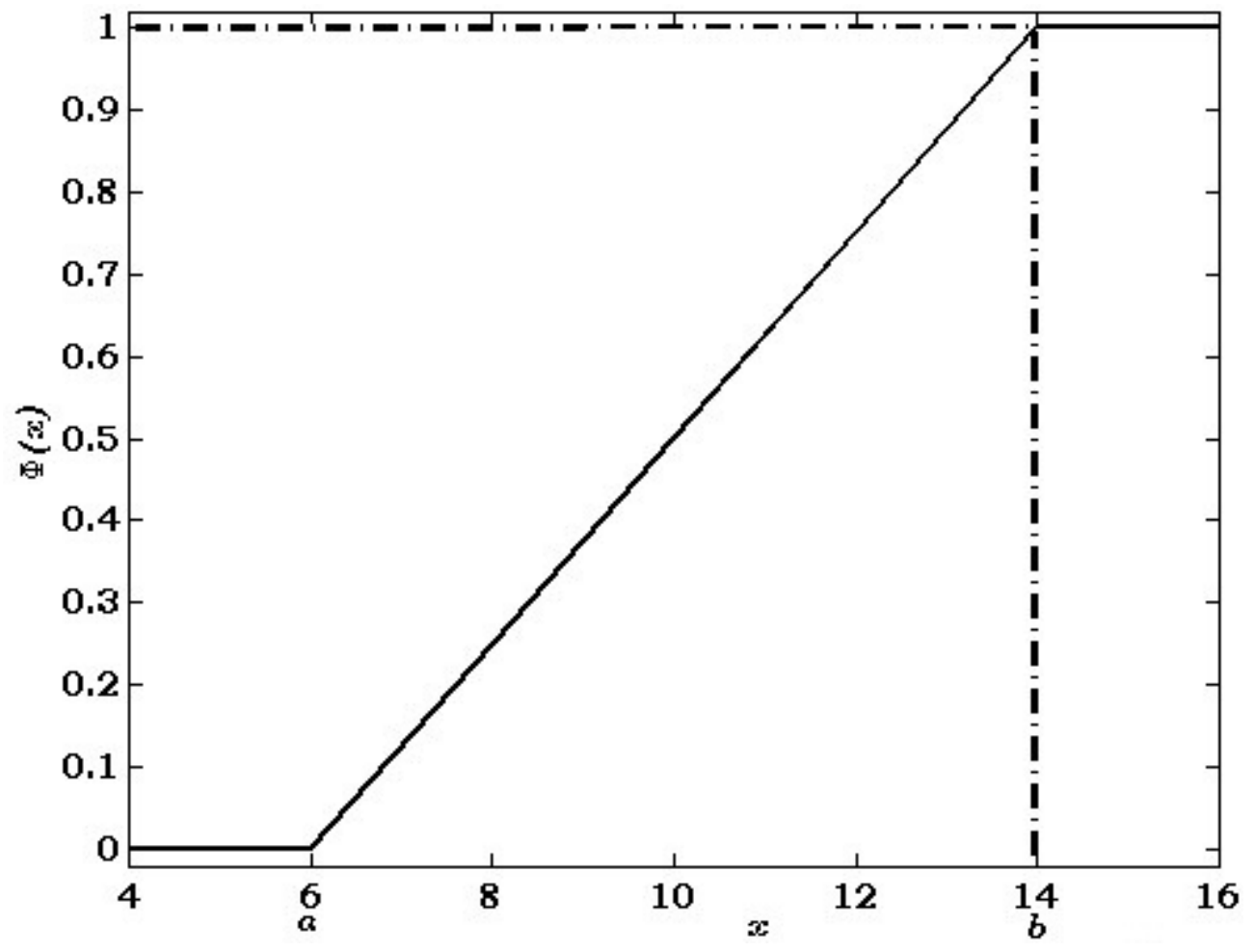

Figure 1.9 Linear uncertain distribution of $\mathcal{L}(6,14)$

**Definition 1.3.30** (B. Liu 2007): An uncertain variable $\zeta$ is said to be zigzag uncertain variable if $\zeta$ follows a zigzag uncertainty distribution as discussed in (1.42).

$$\Phi(x) = \begin{cases} 0 & ; if\ x < a \\ \frac{x-a}{2(b-a)} & ; if\ a \le x < b \\ \frac{(x+c-2b)}{2(c-b)} & ; if\ b \le\ x < c \\ 1 & ; if\ x \ge c, \end{cases} \tag{1.42}$$

where $\zeta$ is denoted by $\mathcal{Z}(a, b, c)$ such that $a < b < c$ and $a, b, c \in \Re$.

**Example 1.3.11**: The uncertainty distribution of zigzag uncertain variable $\mathcal{Z}(10,12,17)$ is shown in Fig. 1.10.

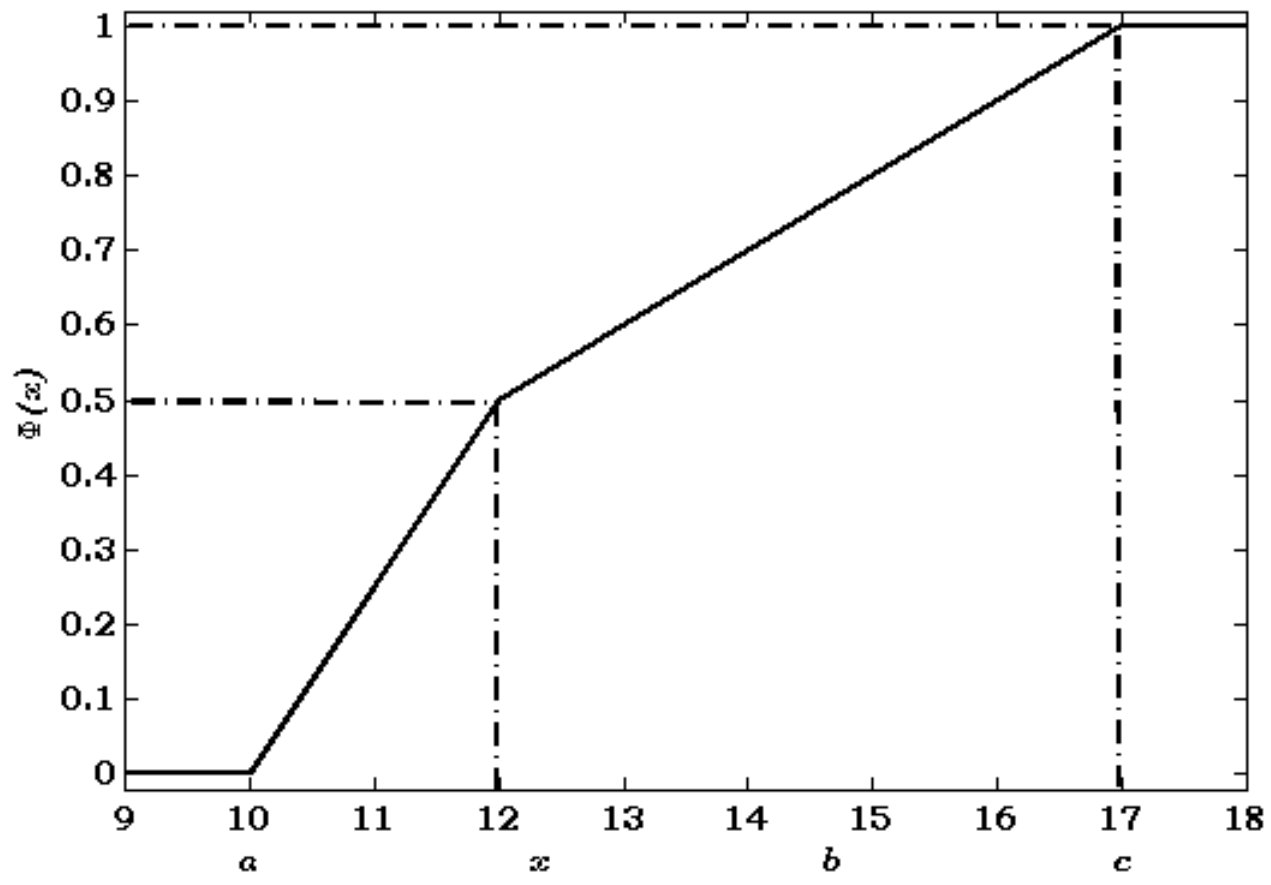

Figure 1.10 Zigzag uncertain distribution of $\mathcal{Z}(10,12,17)$

**Definition 1.3.31** (B. Liu 2007): Let $\zeta$ be an uncertain variable such that $\zeta$ follows a normal uncertainty distribution as defined in (1.43). Then $\zeta$ becomes a normal uncertain variable, such that

$$\Phi(x) = \left(1 + exp\left(\frac{\pi(\rho-x)}{\sigma\sqrt{3}}\right)\right)^{-1}, x \in \Re, \tag{1.43}$$

where $\zeta$ is denoted by $\mathcal{N}(\rho, \sigma)$ such that $\rho, \sigma \in \Re$ and $\sigma > 0$.

**Example 1.3.12**: The uncertainty distribution of a normal uncertain variable $\mathcal{N}(10,2)$ is presented in Fig. 1.11.

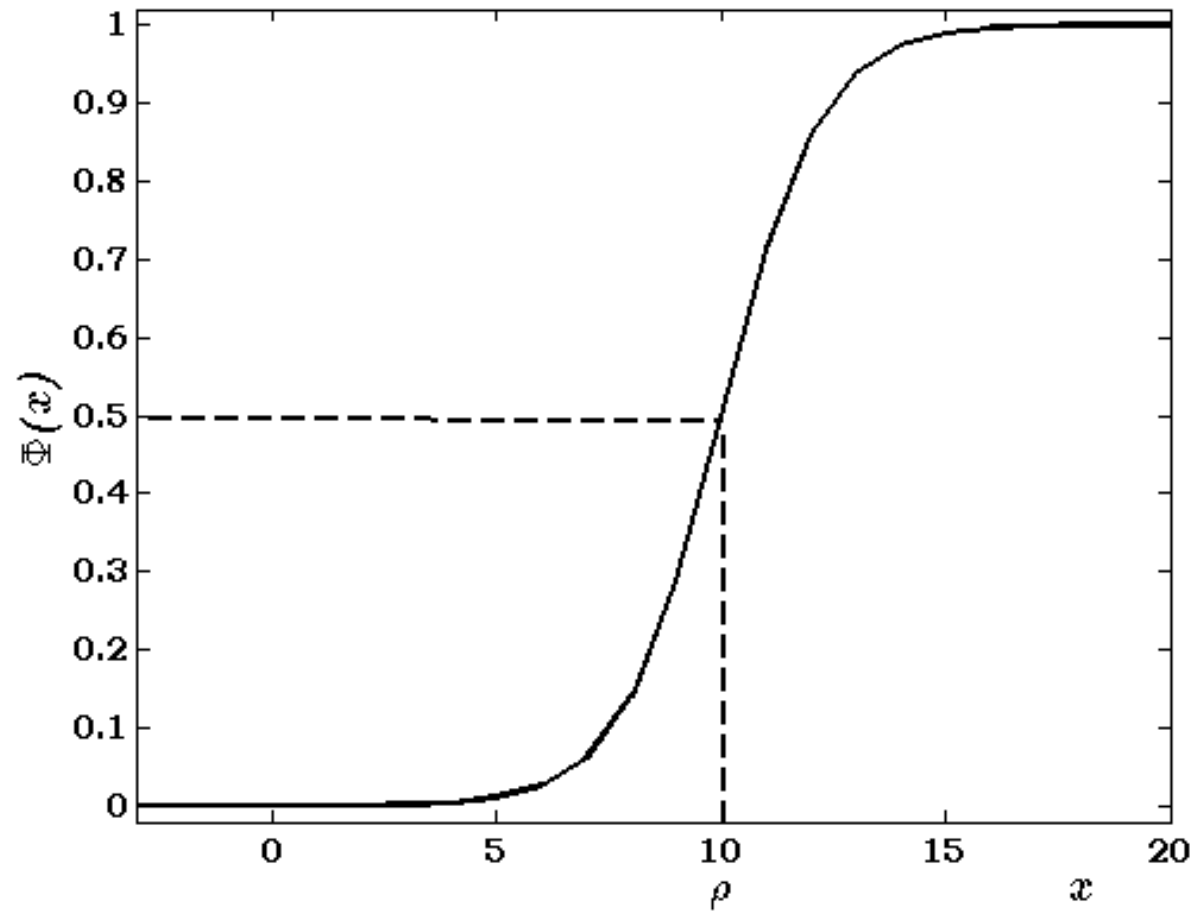

Figure 1.11 Normal uncertain distribution of $\mathcal{N}(10,2)$

**Theorem 1.3.5** (B. Liu 2010): Let $\xi_1$, $\xi_2$, ..., $\xi_n$ be the uncertain variables on the uncertainty space $(\Gamma, \mathcal{L}, \mathcal{M})$ and $f: \Re^n \mapsto \Re$ is a real-valued measurable function. Then $f(\xi_1, \xi_2, \dots, \xi_n)$ is an uncertain variable such that for any Borel set $\mathcal{B}$ of real numbers,

$$\mathcal{M}\{f(\xi) \in \mathcal{B}\} = \mathcal{M}\{\xi \in f^{-1}(\mathcal{B})\} \tag{1.44}$$

**Theorem 1.3.6** (B. Liu 2010): The uncertainty distribution $\Phi: \Re \mapsto [0,1]$ of an uncertain variable $\xi$ is defined as

$$\mathcal{M}\{\xi \le x\} = \Phi(x) \text{ and } \mathcal{M}\{\xi > x\} = 1 - \Phi(x), \forall x \in \Re.$$

**Definition 1.3.32** (B. Liu 2010): An uncertainty distribution $\Phi(x)$ is said to be regular if it is continuous and strictly increasing function with respect to $x$ such that $\Phi(x) \in [0,1]$, and

$$\lim_{x\to -\infty} \Phi(x) = 0 \text{ and } \lim_{x\to +\infty} \Phi(x) = 1.$$

Based on Definition 1.3.32, the inverse uncertainty distribution was proposed by B. Liu (2010) as presented below.

**Definition 1.3.33** (B. Liu 2010): Let $\xi$ be an uncertain variable with regular uncertainty distribution $\Phi(x)$. Then the inverse function $\Phi^{-1}(\alpha)$ is called the inverse uncertainty distribution of $\xi$, where $0 \le \alpha \le 1$.

**Theorem 1.3.7** (B. Liu 2010): A function $\Phi^{-1}$ is an inverse uncertainty distribution of an uncertain variable $\xi$ if and only if

$$\mathcal{M}\{\xi \le \Phi^{-1}(\alpha)\} = \alpha, \ \forall\, \alpha \in [0,1]. \qquad (1.45)$$

**Definition 1.3.34** (B. Liu 2009): The uncertain variables $\xi_1, \xi_2, \dots, \xi_n$ are said to be independent if

$$\mathcal{M}\{\cap_{p=1}^{n}(\xi_p \in \mathcal{B}_p)\} = \wedge_{p=1}^{n} \mathcal{M}\{\xi_p \in \mathcal{B}_p\} \qquad (1.46)$$

for all Borel sets $\mathcal{B}_1, \mathcal{B}_2, \dots, \mathcal{B}_n$ of real numbers.

**Theorem 1.3.8** (B. Liu 2010): Let $\xi_1, \xi_2, \dots, \xi_n$ be the independent uncertain variables with the respective regular uncertainty distributions $\Phi_1, \Phi_2, \dots, \Phi_n$. If $f(\xi_1, \xi_2, \dots, \xi_n)$ is strictly increasing with respect to $\xi_1, \xi_2, \dots, \xi_s$ and strictly decreasing with respect to $\xi_{s+1}, \xi_{s+2}, \dots, \xi_n$, then

$$\Phi^{-1}(\alpha) = f\left(\Phi_1^{-1}(\alpha), \dots, \Phi_s^{-1}(\alpha), \Phi_{s+1}^{-1}(1-\alpha), \dots, \Phi_n^{-1}(1-\alpha)\right) \qquad (1.47)$$

is an inverse uncertainty distribution of the uncertain variable $\xi = f(\xi_1, \xi_2, \dots, \xi_n)$.

**Definition 1.3.35** (B. Liu 2007): Let $\xi$ is an uncertain variable. Then the expected value of $\xi$ is defined as

$$E[\xi] = \int_0^{+\infty} \mathcal{M}\{\xi \ge x\}dx - \int_{-\infty}^{0} \mathcal{M}\{\xi \le x\}dx \qquad (1.48)$$

provided at least one of the two integrals is finite.

**Theorem 1.3.9** (B. Liu 2010): Let $\Phi$ be a regular uncertainty distribution of an uncertain variable $\xi$. Then

$$E[\xi] = \int_0^1 \Phi^{-1}(\alpha)\, d\alpha. \qquad (1.49)$$

**Example 1.3.13**:

(i) Let $\xi = \mathcal{L}(a,b)$ is a linear uncertain variable. Then the expected value of $\xi$ is defined as $\frac{(a+b)}{2}$.

(ii) Let $\xi = \mathcal{Z}(a,b,c)$ is a zigzag uncertain variable. Then the expected value of $\xi$ is defined as $\frac{(a+2b+c)}{4}$.

(iii) Let $\xi = \mathcal{N}(m, \sigma)$ is a normal uncertain variable. Then the expected value of $\xi$ is defined as $m$.

**Theorem 1.3.10** (B. Liu 2010): Let $\xi_1$ and $\xi_2$ are the independent uncertain variables with finite expected values. Then for any real numbers $a$ and $b$, we have

$$E[a\xi_1 + b\xi_2] = aE[\xi_1] + bE[\xi_2]. \tag{1.50}$$

**Theorem 1.3.11** (Liu and Ha 2010): Let $\xi_1$, $\xi_2$,..., $\xi_n$ be the independent uncertain variables with the respective regular uncertainty distributions $\Phi_1$, $\Phi_2$,..., $\Phi_n$. If $\xi = f(\xi_1, \xi_2, \ldots, \xi_n)$ is strictly increasing with respect to $\xi_1$, $\xi_2$,..., $\xi_s$ and strictly decreasing with respect to $\xi_{s+1}$, $\xi_{s+2}$,..., $\xi_n$, then

$$E[\xi] = \int_0^1 f\left(\Phi_1^{-1}(\alpha), \ldots, \Phi_s^{-1}(\alpha), \Phi_{s+1}^{-1}(1-\alpha), \ldots, \Phi_n^{-1}(1-\alpha)\right) d\alpha \tag{1.51}$$

is the expected value of the uncertain variable $\xi = f(\xi_1, \xi_2, \ldots, \xi_n)$.

**Definition 1.3.36** (B. Liu 2007): Let $\zeta$ be an uncertain variable and $\alpha \in (0,1]$ be the confidence level, then $\zeta_{sup}(\alpha)$ and $\zeta_{inf}(\alpha)$ are the respective $\alpha$-optimistic and $\alpha$-pessimistic values of $\zeta$ such that

$$\zeta_{sup}(\alpha) = sup\{t | \mathcal{M}\{\zeta \geq t\} \geq \alpha\} \tag{1.52}$$

and

$$\zeta_{inf}(\alpha) = inf\{t | \mathcal{M}\{\zeta \leq t\} \geq \alpha\} \tag{1.53}$$

**Example 1.3.14**

(i) Let $\zeta$ be a linear uncertain variable $\mathcal{L}(a, b)$, then the corresponding $\alpha$-optimistic and $\alpha$-pessimistic values of $\zeta$ are respectively

$$\zeta_{sup}(\alpha) = \alpha a + (1-\alpha)b$$

and

$$\zeta_{inf}(\alpha) = (1-\alpha)a + \alpha b.$$

(ii) Let $\zeta$ be a zigzag uncertain variable $\mathcal{Z}(a, b, c)$, then $\alpha$-optimistic and $\alpha$-pessimistic values are respectively

$$\zeta_{sup}(\alpha) = \begin{cases} 2\alpha b + (1-2\alpha)c & ; if\ \alpha < 0.5 \\ (2\alpha - 1)a + (2-2\alpha)b & ; if\ \alpha \geq 0.5 \end{cases}$$

and

$$\zeta_{inf}(\alpha) = \begin{cases} (1-2\alpha)a + 2\alpha b & ; if\ \alpha < 0.5 \\ (2-2\alpha)b + (2\alpha - 1)c & ; if\ \alpha \geq 0.5. \end{cases}$$

(iii) Let $\zeta$ be a normal uncertain variable $\mathcal{N}(\rho, \sigma)$ then the respective $\alpha$-optimistic and $\alpha$-pessimistic values of $\zeta$ are respectively

$$\zeta_{sup}(\alpha) = \rho - \frac{\sigma\sqrt{3}}{\pi} ln \frac{\alpha}{1-\alpha}$$

and

$$\zeta_{inf}(\alpha) = \rho + \frac{\sigma\sqrt{3}}{\pi} ln \frac{\alpha}{1-\alpha}.$$

### 1.3.12 Ordered Weighted Averaging (OWA) Operator

In this section, we present the concept of ordered weighted averaging operator and one of its variant, the argument-dependent approach of OWA operator.

**Definition 1.3.37** (R.R. Yager 1988): An OWA operator of dimension $n$ is a mapping, $\text{OWA}: \Re^n \mapsto \Re$, which has an associated weight vector, $\omega\,(= [\omega_1, \omega_2, \dots, \omega_n]^T)$ $\forall \omega_i \in [0,1]$ and $\sum_{i=1}^{n} \omega_i = 1$.

Moreover, $\text{OWA}(b_1, b_2, \dots, b_n) = \sum_{i=1}^{n} \omega_i b_{\pi(i)}$,

where $(b_1, b_2, \dots, b_n)$ is the collection of $n$ arguments and $\{\pi(1), \pi(2), \dots, \pi(n)\}$ is a permutation of $\{1,2, \dots, n\}$ such that $b_{\pi(i-1)} \geq b_{\pi(i)}$ $\forall i \in \{2,3, \dots, n\}$.

**Argument-dependent approach for OWA Weights**

Z. Xu (2006) proposed an argument-dependent approach to determine the OWA weights which are defined as follows.

For an argument list $B = (b_1, b_2, \dots, b_n)$, let $\mu\left(= \frac{\sum_{i=1}^{n} b_i}{n}\right)$ be the mean of $B$ and $\{\pi(1), \pi(2), \dots, \pi(n)\}$ be a permutation of $\{1,2, \dots, n\}$ such that $b_{\pi(i-1)} \geq b_{\pi(i)}$ $\forall i \in \{2,3, \dots, n\}$. Then the similarity degree $s(b_{\pi(i)}, \mu)$ for $i^{th}$ largest argument $b_{\pi(i)}$ is given by

$$s(b_{\pi(i)}, \mu) = 1 - \frac{|b_{\pi(i)} - \mu|}{\sum_{i=1}^{n} |b_i - \mu|} \forall i = 1,2, \dots, n. \tag{1.54}$$

If $\omega\,(= [\omega_1, \omega_2, \dots, \omega_n]^T)$ is the weight vector of OWA operator then we define each $\omega_i$ as

$$\omega_i = \frac{s(b_{\pi(i)}, \mu)}{\sum_{i=1}^{n} s(b_{\pi(i)}, \mu)} \forall i = 1,2, \dots, n. \tag{1.55}$$

Henceforth,

$$\text{OWA}(b_1, b_2, \dots, b_n) = \frac{\sum_{i=1}^{n} (s(b_{\pi(i)}, \mu) \times b_{\pi(i)})}{\sum_{i=1}^{n} s(b_{\pi(i)}, \mu)}. \tag{1.56}$$

**Example 1.3.15**

Consider an argument list $B$ as presented below.

$$B = [37.9647, 23.5676, 47.90, 22.7985, 54.2456]$$

Let $\{\pi(1), \pi(2), \pi(3), \pi(4), \pi(5)\}$ be a permutation of $\{1,2,3,4,5\}$ such that $b_{\pi(i-1)} \geq b_{\pi(i)}$, $i = 2,3, \dots, 5$ and $b_{\pi(i)}$ is an element of $B$. Then, arranging the elements of $B$ as, $b_{\pi(1)} = 54.2456$, $b_{\pi(2)} = 47.90$, $b_{\pi(3)} = 37.9647$, $b_{\pi(4)} = 23.5676$ and $b_{\pi(5)} = 22.7985$. We calculate the mean of the elements of $B$ as

$\mu = \frac{(37.9647+ 23.5676+ 47.90+22.7985+54.2456)}{5} = 37.2953.$

Now we can find all the $\omega_i$ of the corresponding weight vector, $\omega$ for $B$ following (1.50) and (1.51) as follows.

So, $s(b_{\pi(1)}, 37.2953) =$

$$1 - \frac{|54.2456-37.2953|}{(|37.9647-37.2953|+|23.5676-37.2953|+|47.90-37.2953|+|22.7985-37.2953|+|54.2456-37.2953|)}$$

$$= 0.6997$$

Similarly, $s(b_{\pi(2)}, 37.2953) = 0.8121, s(b_{\pi(3)}, 37.2953) = 0.9881,$ $s(b_{\pi(4)}, 37.2953) = 0.7568$ and $s(b_{\pi(5)}, 37.2953) = 0.7432.$

Furthermore, $\omega_1 = \frac{0.6997}{(0.6997+0.8121+0.9881+0.7568+0.7432)} = 0.1749.$

Correspondingly, $\omega_2 = 0.2030, \omega_3 = 0.2470, \omega_4 = 0.1892$ and $\omega_5 = 0.1858.$

Therefore, according to (1.52), $\mathrm{OWA}(37.9647, 23.5676, 47.90, 22.7985, 54.2456)$ $= (0.1749 \times 54.2456 + 0.2030 \times 47.90 + 0.2470 \times 37.9647 + 0.1892 \times 23.5676 + 0.1858 \times 22.7985) = 37.2835.$

### 1.3.13 Single objective Optimization

The problem of optimization concerns with the maximization/minimization of an algebraic or a transcendental function of one or more variables, known as objective function, under some available resources which are represented as constraints. Such a problem is called a single objective optimization problem which is formulated as follows.

$$\begin{cases} Minimize\ z(x) \\ subject\ to \\ g_j(x) \le 0;\ \ j = 1, 2, \dots, m \\ h_k(x) = 0;\ \ k = 1, 2, \dots, p \\ x = (x_1, x_2, \dots, x_n)^T \in \Re^n, \end{cases} \tag{1.57}$$

where $x$ is an $n$-dimensional decision vector, $z(x)$, $g_j(x)$ and $h_k(x)$ are respectively the objective function, set of inequality constraints and equality constraints defined on $x$. A decision vector satisfying all the constraints of the above Single objective optimization problem (SOOP) is called a feasible solution of (1.57). The collection of all such solutions forms the feasible region. The aim of a SOOP problem is to find a feasible solution $x^*$ such that for each feasible solution, $z(x) \le z(x^*)$ for maximization problem and $z(x) \ge z(\ x^*)$ for minimization problem. Here, $x^*$ is called an optimal solution of (1.57).

### 1.3.14 Multi-objective Optimization (MOP)

An MOP optimizes a vector of conflicting objectives, which can be formulated as

$$\begin{cases} Minimize\ \ Z(x) = \big(z_1(x), z_2(x), \dots, z_m(x)\big) \in \Re^m \\ subject\ to \\ \quad g_i(x) \le 0;\ i = 1,2, \dots, p \\ \quad h_j(x) = 0;\ j = p+1, p+2, \dots, q \\ \quad x = (x_1, x_2, \dots, x_n)^T \in \Re^n, \end{cases} \tag{1.58}$$

where $x$ is a $n$-dimensional vector of decision variables, $z_k$ is the $k^{th}$ objective function, $g_i$ is the $i^{th}$ inequality constraint and $h_j$ is the $j^{th}$ equality constraint.

Due to the conflicting nature of the objectives in MOP there exists trade-offs between them, hence improving an objective deteriorates the value of at least one of the remaining objectives. Therefore, instead of a single optimal solution, an MOP always generates a set of solutions. These solutions cannot be compared with each other, i.e., one cannot recommend that a particular solution in the optimal set is better than any of the remaining solutions. The solutions of the optimal set are nondominated solutions in the sense that neither of the solutions can dominate the other. If we consider $r = (r_1, r_2, .., r_n)^T$ and $t = (t_1, t_2, .., t_n)^T \in \Re^n$ as two solutions for the model (1.58), then $r$ dominates $t$ (or $r \prec t$) if and only if

(i) for all objectives $z_k$, $z_k(r) \le z_k(t), \forall\, k \in \{1,2, \dots, m\}$ and

(ii) there exists at least one objective $z_l$ such that $z_l(r) < z_l(t)$, where $l \in \{1,2, \dots, m\}$.

$r$ and $t$ are said to be nondominated to each other if neither $r \prec t$ nor $t \prec r$. A nondominated solution $x^*$ is said to be Pareto optimal solution if there does not exist any other possible solution $x$ in the decision space, such that $Z(x) \prec Z(x^*)$. The set of all Pareto optimal solutions generate the Pareto set (PS), and the set of all objective vectors in the objective space corresponding to PS is known as the Pareto front (PF).

### 1.3.15 Classical Multi-objective Solution Techniques

The classical multi-objective solution techniques are based on single point search, where a single solution at every iteration is considered. The classical optimization methods can at best generate a single compromise solution for a multi-objective optimization problem. In the following subsections, we present five classical methods: (i) linear weighted method, (iii) epsilon-constraint method, (iii) goal attainment method, (iv) fuzzy programming method and (v) global criterion method to solve a multi-objective optimization problem.

#### 1.3.15.1 Linear Weighted Method

The linear weighted method (R.T. Eckenrode 1965) scalarizes a multi-objective problem with an objective vector $Z(x)$ into a single objective problem, where the objective function is a weighted sum of each objective of $Z(x)$. The compromise single objective model of (1.58) using linear weighted method is shown below.

$$\begin{cases} \underset{x \in \Re^n}{Minimize}\ Z(x) = \sum_{i=1}^{n} \omega_i \times z_i(x), i = 1,2,\dots,m \\ subject\ to \\ \quad constraints\ of\ (1.58) \\ \quad \sum_{i=1}^{n} \omega_i = 1 : \omega_i \geq 0 \\ \quad x = (x_1, x_2, \dots, x_n)^T, \end{cases} \tag{1.59}$$

where $\omega_i$ is the weight associated to the $i^{th}$objective function $z_i(x)$.

### 1.3.15.2 Epsilon-constraint Method

Haimes et al. (1971) proposed the epsilon-constraint method to solve a multi-objective optimization problem. In this approach, an MOP is transformed to its equivalent compromise single objective optimization problem, where a single objective is required to be optimized, and the remaining objectives are transformed to the equivalent constraints which are restricted to user-defined values. This method is also proved to be efficient enough to find solutions to the problems in the non-convex region of the objective space. Using epsilon-constraint method, model (1.58) can be transformed into a single objective optimization problem as presented below.

$$\begin{cases} Minimize\ Z(x) = z_k(x) \\ subject\ to \\ z_r(x) \leq \varepsilon_r;\ s = 1,2,\dots,m \text{ and } r \neq k \\ constraints\ of\ (1.58) \\ \quad x = (x_1, x_2, \dots, x_n)^T, \end{cases} \tag{1.60}$$

where $\varepsilon_r$ implies an element of the epsilon vector $\varepsilon$, which represents a value that lies in between the lower and the upper bounds of $z_r(x)$, and is not necessarily a value close to zero.

### 1.3.15.3 Goal Attainment Method

The goal attainment method was first proposed by F.W. Gembicki (1974), which involves determining a set of goals, $Z^* = \{z_1^*, z_2^*, \dots, z_n^*\}$. Here, each $z_i^*$ is considered as the goal of an objective function $z_i(x)$ which constitute the vector of objectives $Z(x) = \{z_1(x), z_2(x), \dots, z_n(x)\}$. The problem formulation allows the objective to be underachieved or overachieved. The relative degree of underachievement or overachievement of these goals are controlled by a weight vector $\omega = \{\omega_1, \omega_2, \dots, \omega_n\}$. The compromise single objective optimization model of (1.58), using goal attainment method is formulated below.

$$\begin{cases} \underset{\lambda \in \Re}{Minimize}\ \lambda \\ subject\ to \\ \quad z_i(x) - \omega_i \lambda \leq z_i^*, \quad i = 1,2,\dots,m \\ \quad constraints\ of\ (1.58) \\ \quad x = (x_1, x_2, \dots, x_n)^T. \end{cases} \tag{1.61}$$

### 1.3.15.4 Fuzzy Programming Method

The fuzzy programming method (H.-J. Zimmermann 1978) is another approach to formulate a compromise single objective optimization problem (SOOP) of a multi-objective problem. To formulate the compromise single objective model of (1.58) using fuzzy programming method, we consider the following steps.

**Step 1**: Solve the model (1.58) as a single objective optimization problem by considering each time only one objective $z_k(x), k = 1,2, \dots, m$ and ignoring the remaining objective functions to obtain the corresponding optimal solution $x^{k*}$.

**Step 2**: Compute the values of all the $m$ objective functions at all of $m$ optimal solutions $x^{k*}(k = 1,2, \dots, m)$, and determine the upper and lower bounds of each objective given by $U^k = Max\{z_k(x^{1*}), z_k(x^{2*}), \dots, z_k(x^{m*})\}$ and $L^k = z_k(x^{k*}),\ k = 1,2, \dots, m.$

**Step 3**: Construct a function (characteristically linear), for each objective function by considering their corresponding upper and lower bounds so that

$$v_k(z_k(x)) = \begin{cases} 1 & ; if\ z_k(x) \leq L^k \\ \left(\frac{U^k - z_k(x)}{U^k - L^k}\right) & ; if\ L^k < Z_2(x) \leq U^k \\ 0 & ; if\ U^k < z_k(x). \end{cases} \tag{1.62}$$

**Step 4**: Formulate the fuzzy programming model of (1.58) as shown below

$$\begin{cases} Maximize\ \lambda \\ subject\ to \\ \left(\frac{U^k - z_k(x)}{U^k - L^k}\right) \geq \lambda \\ constraints\ of\ (1.58) \\ x = (x_1, x_2, \dots, x_n)^T, \end{cases} \tag{1.63}$$

where $\lambda \geq 0$ and $Max\ \lambda$ is the maximum overall satisfactory level of compromise.

### 1.3.15.5 Global Criterion Method

The global criterion method, proposed by S.S. Rao (2006), transforms a multi-objective optimization problem to its equivalent compromise single objective optimization problem (SOOP). In this method, a preselected global criterion, such as the sum of the relative deviations of the individual objective functions from the corresponding feasible ideal solutions, is minimized. The solution of the SOOP is a compromise solution of the corresponding MOOP. The steps for formulating the compromise SOOP for the model (1.58), using global criterion method are presented below.

**Step 1**: Individually solve each objective function of the multi-objective problem, described in (1.58).

**Step 2**: From the results of Step 1, determine the ideal values, $Z_k = z_k(x^{k*})$ of all the objective functions of the model (1.58) corresponding to the optimal solutions $x^{k*}, k = 1,2, \dots, m.$

**Step 3**: Formulate the equivalent compromise model as

$$\begin{cases} Minimize \ \left(\sum_{k=1}^{m}\left(\frac{z_k(x)-Z_k}{Z_k}\right)^{\omega}\right)^{\frac{1}{\omega}} \\ subject\ to \\ \quad constraints\ of\ (1.58) \\ \quad x=(x_1,x_2,\dots,x_n)^T. \end{cases} \tag{1.64}$$

Here, if the value of $\omega( 1 \leq \omega < \infty)$ is considered as 2, then the method is known as the global criterion method in $L_2$ norm*.

### 1.3.16 Genetic Algorithms

Evolutionary algorithm (EA) is a search and optimization algorithm which imitate the natural evolutionary principles to drive its search towards an optimal solution. Unlike classical search and optimization techniques, multiple solutions can be determined from a single execution of EAs due to their population based approach. In each iteration of an EA, a population of solutions are processed. As a result, an EA generates a population of solutions as output. In this context, if an optimization problem has a single optimum solution, then all the candidates of a population are expected to converge to the optimum one. However, for an optimization problem with multiple optimum solutions, EAs are expected to find multiple optimum solutions in its final population.

The concept of genetic algorithm (GA) was first introduced by H.J. Holland (1975). Thereafter, GA has been successfully applied to solve different optimization problems from various domains including science, engineering and management. GA is a population based stochastic search optimization algorithm which is inspired by natural genetics and Darwin's principle of survival of the fittest. A genetic algorithm starts with an initial population consisting of a set of randomly generated solutions which satisfy the constraints of the problem. Each member of a population is known as a chromosome which represents a possible solution of the problem. A chromosome of a GA is represented by a string of symbols. Every chromosome consists of a number of individual structures called genes or alleles, and its respective position is known as locus. The chromosomes are evolved in successive iterations called generations. In each generation, the fitness values of the chromosomes are evaluated using a function called fitness function. In every generation, new chromosomes called offspring/children are generated by the old chromosomes called parents using three genetic operators, selection, crossover and mutation operators. The selection operator chooses chromosomes based on their fitness values. A chromosome with a better fitness value has a higher probability of being selected for the next generation. The crossover operator exchanges information/genes of two chromosomes which imitates biological recombination process between two single chromosome organisms. A mutation operator randomly changes the allele value(s) of chromosome. Finally, after several generations, when a termination criterion (maximum generation, unchanged population for $n$ successive generations, etc.) of a GA is reached, the algorithm is expected to converge to the best chromosome which essentially represents an optimal or sub-optimal solution of the problem.

---

* $L_2$ norm is also known as least square which minimizes the sum of the squares of the differences between the target values and the estimated values.

A genetic algorithm for a particular optimization problem must have the following six components.

(i) Genetic representation of the potential solutions to the problem, called chromosomes.

(ii) A way to generate an initial population of potential solutions by satisfying the problem constraints, may be randomly generated.

(iii) An objective function to evaluate the fitness of each solution/chromosome.

(iv) The genetic operator (selection) which select the chromosomes with better fitness values.

(v) The genetic operators (crossover and mutation) which modify the composition of chromosomes.

(vi) Values of different associated parameters of a genetic algorithm (population size, crossover probability, mutation probability, etc.).

An algorithm for a simple GA is presented below.

**begin**
  $t$ = 0; // first generation
  Generate ($P(0)$) // initial population
  evaluate $\left(P(t)\right)$
  **while** (termination_criterion is not reached) **do**
    $R(t)$ = selection($P(t)$)
    $R(t)$ = crossover($R(t)$)
    $R(t)$ = mutation($R(t)$)
    $P(t) = R(t)$
    evaluate ($P(t)$) // evaluate the fitness of the chromosomes in the
                // population P(t)
    $t = t + 1$
  **end while**
  **return** the best solution
**end**

In this thesis, we have considered both single and multi-objective GAs for some of our studies. Specifically, we have used a modified version of a genetic algorithm (GA) as a solution methodology for a single objective problem (cf. Section 3.4 in Chapter 3). While, for some multi-objective problems, we have employed two multi-objective genetic algorithms (MOGAs): nondominated sorting genetic algorithm II (NSGA-II) (Deb et al. 2002) and multi-objective cross generational elitist selection, heterogeneous recombination and cataclysmic mutation (MOCHC) (Nebro et al. 2007). Accordingly, the working principles of these MOGAs are discused in subsequent subsections.

#### 1.3.16.1 Nondominated Sorting Genetic Algorithm II (NSGA-II)

The nondominated sorting genetic algorithm II (NSGA-II), proposed by Deb et al. (2002), is an elitist multi-objective genetic algorithm, which ensures retaining the fittest candidates in the next population to enhance the convergence. The algorithm starts with initializing a population $P_0$ of $N$ randomly generated solutions. In a particular generation $t$, the genetic operators, i.e., selection, crossover and mutation are applied on the individuals of parent population $P_t$ to generate an offspring population $C_t$ having equal number of candidate solutions as parent population. To ensure elitism, the parent and offspring population are mingled to produce a population $S_t$ of size $2N$. In order to select $N$ best solutions from $S_t$ for next generation, NSGA-II performs following two steps.

**Sorting based on Rank**: It frontifies the individuals of $S_t$, by assigning a rank to each individual solution, i.e., a solution $p$ is assigned a rank $p_{rank}$. Based on their ranks, the solutions in $S_t$ are frontified to different nondominated fronts $N_1, N_2, \dots, N_l$. Each $N_k, k \in \{1,2,\dots,l\}$ represents a set of nondominated solutions of rank $k$. Solutions with the same rank value are placed in a nondominated front. Solutions with lower nondomination ranks is always preferred. In other words, if $p$ and $q$ are the two solutions in $S_t$, and if $p_{rank} < q_{rank}$, then $p$ is superior to $q$. In order to build the next population $P_{t+1}$, the solutions belonging to $N_1$ are considered first. Here, if the size of $N_1$ is less than $N$, all the solutions of $N_1$ are inserted in $P_{t+1}$. The remaining solutions of $P_{t+1}$ are considered from subsequent nondominated fronts in order of their ranking. In this way, the solutions of 2nd front, i.e., $N_2$ are moved next to $P_{t+1}$ followed by solutions of $N_3$ and so on until all the solutions of a nondominated front $N_k$ cannot be fully inserted to $P_{t+1}$.

**Sorting based on Crowding Distance**: If all the solutions of $N_k$ cannot be accommodated in $P_{t+1}$, the solutions are then sorted in descending order based on their corresponding values of crowding distance (a solution $i$ in $N_k$ is assigned $i_{distance}$) as proposed by Deb et al. (2002). Particularly, if $p$ and $q$ are the two nondominated solutions belong to $N_k$ and $p$ has better crowding distance than $q$, i.e., if $p_{rank} = q_{rank}$ and $p_{distance} > q_{distance}$, then $p$ is preferred over $q$ in $N_k$. Accordingly, the solutions with higher crowding distance are eventually selected from $N_k$ to fill the remaining slots of $P_{t+1}$.

In this way, as the formation of $P_{t+1}$ completes, the individuals of $P_{t+1}$ replaces $P_t$ for the next generation. This process is continued until the termination condition (i.e., maximum function evaluations or generations) is reached.

#### 1.3.16.2 Multi-objective Cross Generational Elitist Selection, Heterogeneous Recombination and Cataclysmic Mutation (MOCHC)

Nebro et al. (2007) presented MOCHC as a multi-objective version of cross generational elitist selection, heterogeneous selection and cataclysmic mutation (CHC) algorithm (L.J. Eshelman 1991). MOCHC begins its execution by randomly generating an initial population $P_0$ of size $N$. At every subsequent generation $t$, a set of solutions in $P_t$ is generated by randomly selecting a pair of parent solutions from $P_{t-1}$ and applying recombination operator on them. The selection of solutions for recombination is done in a way to implement the mechanism of incest prevention, i.e., the mating of similar solutions is prohibited. In other words, a pair of solutions can only participate in the recombination process if the Hamming distance between them is greater than a predetermined threshold value $(convergence_count)$. The genetic operator commonly used for recombination mechanism in MOCHC is half uniform crossover (HUX) (L.J. Eshelman 1991). This crossover operator first replicates the common information shared by the parent solutions to their respective offspring, and then copies half of the divergent information of the parents to their offspring in a way to achieve maximum Hamming distance between the offspring and parents. Since no mutation operator is used to generate offspring population $C_t$, the maximum Hamming distance between parents and corresponding offspring ensures the only way to introduce diversity in $C_t$.

Once the offspring are generated, the elite $N$ nondominated solutions of the new population $P_{t+1}$ are selected from the combination of $P_t$ and $C_t$ by using crowding-comparison operator (Deb et al. 2002). As the execution of the algorithm progresses, the population also approaches towards the state of homogeneity with generations. Hence, at any subsequent generation, if the solutions of $P_t$ and $P_{t+1}$ remain unchanged, the threshold value is then progressively decreased. In this way, when the threshold value reaches 0 it is assumed that no new diversity is introduced in the population and the population becomes completely homogeneous. This results in a premature convergence of the population due to elitism and absence of mutation. At this stage, in order to introduce new diversity in the population, the *restart* phase of MOCHC begins. In this phase, certain percent of best nondominated solutions (usually $5\%$) of the present population $P_t$ are preserved, which are selected by crowded-comparison operator and the rest of the solutions of $P_t$ are cataclysmically mutated using bit-flip mutation with a very high mutation probability $(0.35)$ as suggested by L.J. Eshelman (1991). High mutation probability is used to introduce new diversity in the population. Once the *restart* phase is over, the same recombination and elitism mechanisms are applied to the population with the threshold set to its initial value, and the entire process continues until the termination criterion is satisfied.

### 1.3.17 Performance Metrics

In order to provide a quantitative assessment for the performance of MOGAs, some metrics are taken into consideration. In this thesis, four performance matrices are used which are discussed below.

#### 1.3.17.1 Hypervolume

As presented by Zitzler and Thiele (1999), Hypervolume ($HV$) is the volume covered by the set of all nondominated solutions $N_D$ in the approximate front (generated by a MOGA). The hypervolume is the region enclosed by all the solutions of $N_D$ in the objective space with respect to the reference solution/point R, where R is a worst solution in the entire objective space of MOP. $HV$ is calculated as

$$HV = volume\left(\cup_{e \in N_D} h_e\right), \tag{1.65}$$

where $h_e$ is the hypercube for each $e \in N_D$. Larger values of $HV$ is always desirable. The $HV$ metric is used for both the convergence and diversity. More the value of $HV$, closer is the approximate front to the Pareto front (PF), generated by a MOGA.

As an example, let $\mathrm{S}_1$ and $\mathrm{S}_2$ are the two fronts generated by two different MOGAs for the same problem as depicted in Fig. 1.12 (a)-(b). In the figures, the Pareto front of the MOP is represented by PF (red color) in the corresponding objective space. In Fig. 1.12 (a), $\mathrm{S}_1$ is more close to PF as well as more diverse compared to $\mathrm{S}_2$. As a result, the $HV$ for $\mathrm{S}_1$ is more than $\mathrm{S}_2$ with respect to reference solution R. If solutions of $\mathrm{S}_1$ are not well diverse compare to $\mathrm{S}_2$ as depicted in Fig. 1.12 (b), a better $HV$ for $\mathrm{S}_2$ is obtained with respect to R, since in this case $\mathrm{S}_2$ covers more spectrum of PF compare to $\mathrm{S}_1$.

#### 1.3.17.2 Spread

Spread ($\Delta$) determines the diversity of a given nondominated front, generated from a simulation of the MOGA. This metric was first proposed by Deb et al. (2002) for a multi-objective problem (MOP) with two objectives. Later, Zhou et al. (2006) extended the metric for MOPs with more than two objectives. The spread metric is expressed as

$$\Delta = \frac{\sum_{i=1}^{m} d(e_i, N_D) + \sum_{X \in \mathrm{PF}} \left| d(X, N_D) - \overline{d} \right|}{\sum_{i=1}^{m} d(e_i, N_D) + |\mathrm{PF}| \times \overline{d}} \ \forall\, e_i \in \mathrm{PF} \tag{1.66}$$

such that $d(X, N_D) = \underset{Y \in N_D, Y \neq X}{min} \|F(X) - F(Y)\|^2$ and $\overline{d} = \frac{1}{|\mathrm{PF}|} \sum_{X \in \mathrm{PF}} d(X, N_D)$, where $F$ is the set of $m$ objectives, $N_D$ is the set of nondominated solutions in the approximate front, PF is the set of known Pareto optimal solutions to the problem and $e_i (i = 1,2, \dots, m)$ represents extreme solutions in PF. Less the value of $\Delta$, more diverse will be the solution set. In the ideal situation, the value of spread is zero.

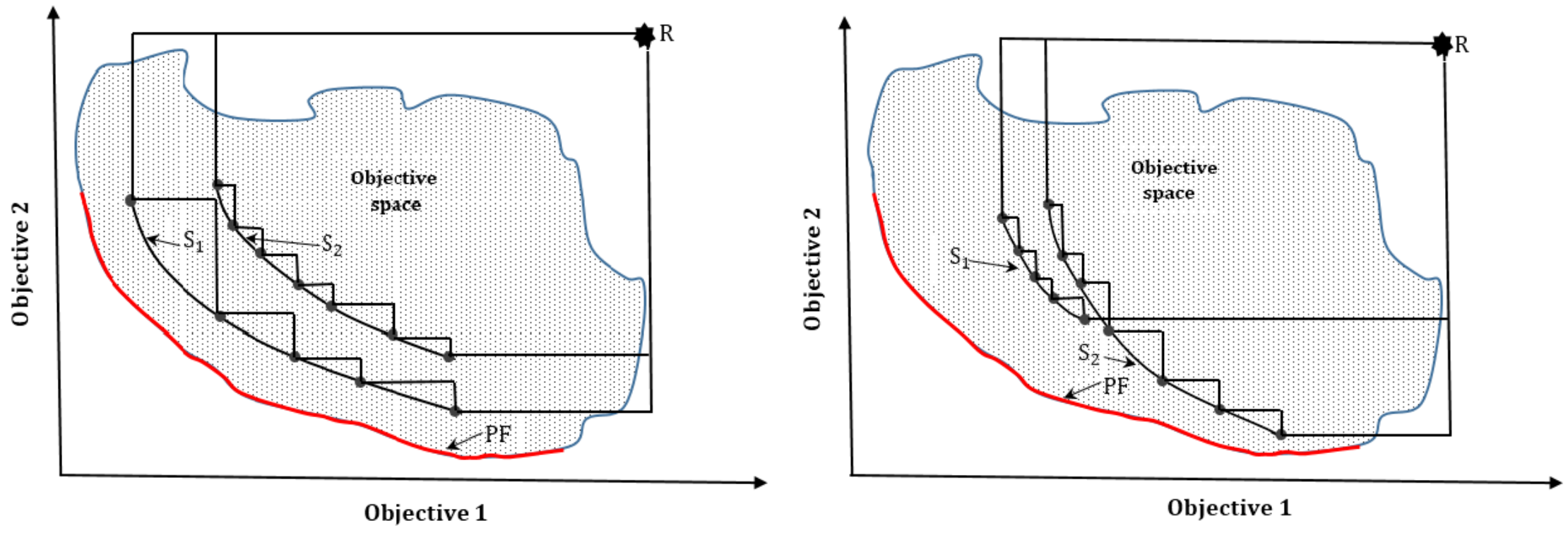

(a) (b)

Figure 1.12 $HV$ for different optimized fronts $S_1$ and $S_2$ with respect to a reference solution $R$. (a) The front $S_1$ is closer to PF than $S_2$, hence shall have higher $HV$ (b) The front $S_2$ is more diversely spread than $S_1$, hence shall have higher $HV$

### 1.3.17.3 Generational Distance

Van Veldhuizen and Lamont (1998) proposed generational distance $(GD)$ which determines the distance in the objective space between the elements of the nondominated solution vectors and their corresponding nearest neighbour in the Pareto front. The formulation of $GD$ metric is presented in (1.67).

$$GD = \frac{\sqrt{\sum_{i=1}^{n} d_i^2}}{n}, \tag{1.67}$$

where total number of solutions in the approximate front is denoted by $n$ and $d_i = \min_j \|F(x_i) - \mathrm{PF}(x_j)\|$ is the distance between the solution $x_i$ and the nearest member $x_j$ in the Pareto front. This metric determines the distance of the approximate front from the Pareto front (PF). In other words, the $GD$ metric ensures the convergence of a MOGA by measuring how far the approximate front of a MOGA is from the Pareto front. At ideal condition $GD$=0.

### 1.3.17.4 Inverted Generational Distance

Inverted generational distance $(IGD)$, presented by Van Veldhuizen and Lamont (1998), measures the distance of the Pareto front from an approximate front, and is represented as follows.

$$IGD = \frac{\sum_{i=1}^{|\mathrm{P}|} d(\rho_i, N_D)}{|\mathrm{P}|} \quad \forall\, \rho_i \in \mathrm{PF}, \tag{1.68}$$

where PF is the Pareto front, $d(\rho_i, N_D)$ is the Euclidian distance between $\rho_i$ and its closest neighbour in $N_D$, and $N_D$ is a set of nondominated solutions in the approximate front. $\rho_i$ represents a solution in PF. $IGD$ assures both convergence and diversity. A lower value of this metric can only be achieved if the solution set $N_D$ is close to PF and nearly covers the spectrum of PF. Therefore, a smaller value of $IGD$ is always preferred.

## 1.4 Scope of the Thesis

In this thesis, we have presented some network optimization problems under various uncertain environments including type-2 fuzzy, rough, rough fuzzy, random fuzzy and uncertainty theory. The thesis is composed of seven chapters, among which the main contribution of the thesis can be observed from Chapter 2 to Chapter 6. These five chapters are further categorized into two parts, Part-I and Part-II. Part-I consists of chapters 2 and 3, which deal with uncertain single objective optimization problems. Whereas, Part-II comprises of chapters 4, 5 and 6, which address uncertain multi-objective optimization problems. The chapters, presented in the thesis, are discussed below in brief.

### 1.4.1 Introduction

In this chapter, we have presented a survey on the studies performed on four different network optimization problems including transportation problem, shortest path problem, minimum spanning tree problem and maximum flow problem, and their variants under different uncertain paradigms. Moreover, some preliminary concepts, definitions, theorems and methodologies pertinent to the thesis are also provided in this chapter.

### 1.4.2 Linear Programming Model with Interval Type-2 Fuzzy Parameters and its Application to some Network Problems

In this chapter, we propose a chance-constrained model to solve constrained linear programming model with interval type-2 fuzzy parameters. It is developed using credibility measure of interval type-2 fuzzy variable. The proposed method is applied to three network optimization problems: (a) solid transportation problem, (b) shortest path problem and (c) minimum spanning tree problem. In the solid transportation problem, the parameters corresponding to the availability at sources, demands at destinations and capacities of the conveyances are considered as trapezoidal interval type-2 fuzzy variables (TrIT2FV). For the shortest path and minimum spanning tree problems, the edge weights of the network are considered as TrIT2FVs. Numerical examples of those problems are provided to illustrate the proposed method.

### 1.4.3 Genetic Algorithm with Varying Population for Random Fuzzy Maximum Flow Problem

In this chapter, we have investigated the uncertain maximum flow problem (MFP) of a network whose capacities are random fuzzy variables. We have developed the EVM and CCM for MFP under random fuzzy environment, and formulate their crisp equivalent models. To solve these models, we have proposed a varying population genetic algorithm with indeterminate crossover (VPGAwIC). In VPGAwIC, selection of a chromosome depends on its lifetime. For this purpose, we have proposed an improved lifetime allocation strategy (iLAS). The ages of the chromosomes are defined linguistically as Young, Middle and Old, which follow the linear and zigzag uncertainty distributions (B. Liu 2007). The crossover probability is considered as indeterminate and it depends on the ages of the parents. The crossover probability is defined by an uncertain rule base. The number of offspring, generated from a population of parents, is determined by reproduction ratio (Z. Michalewicz 1992), which is assumed to be fixed. The population is updated in two ways. Firstly, all the chromosomes with ages greater than their lifetimes are discarded from the population. Secondly, the offspring are combined with their parents for the next generation. The performance of the proposed VPGAwIC on MFP has been compared with that of the genetic algorithm developed by Gen et al. (2008).

### 1.4.4 Multi-criteria Shortest Path Problem on Rough Graph

Shortest path problem (SPP) in real-world applications has to deal with multiple criteria/objectives. This chapter is intended to solve a multi-criteria shortest path problem (MSPP) of a weighted connected directed network (WCDN), where edge weights are represented by rough variable. We have proposed two different approaches to determine the optimum path(s) of MSPP. In the first approach, the rough variable has been incorporated in the Dijkstra's algorithm, which is named as modified rough Dijkstra's (MRD) algorithm. The second approach deals with modelling MSPP using rough chance-constrained programming technique which is eventually solved by two different methods: the goal attainment method and NSGA-II. The proposed methodologies are numerically illustrated for a multi-criteria weighted connected directed network (WCDN) and the simulated results are analyzed using large instances of WCDN to show the efficiency of the proposed algorithms.

### 1.4.5 Uncertain Multi-objective Multi-item Fixed Charge Solid Transportation Problem with Budget Constraint

Modelling real-world applications requires data as input parameter which represent information represented in the state of indeterminacy. In this chapter, we have presented an uncertain multi-objective multi-item fixed charge solid transportation problem with budget constraint (UMMFSTPwB) at each destination. In the proposed model we have optimized two objectives: (i) maximization of profit and (ii) minimization of time. In the proposed UMMFSTPwB, it is considered that items are purchased at different sources with different prices, and accordingly transported to different destinations using different types of vehicles having different capacities. The items are sold to the customers at different selling prices. Here, unit transportation costs, fixed charges, transportation times, supplies at origins, demands at destinations, conveyance capacities and budgets at destinations are assumed to be uncertain variables. For UMMFSTPwB, we develop three uncertain programming models: (i) expected value model (EVM) (ii) chance-constrained model (CCM) and (iii) dependent chance-constrained model (DCCM). These models are formulated under the framework of uncertainty theory. Finally, the equivalent deterministic models are formulated which are solved using three methods: (i) linear weighted method, (ii) global criterion method and (iii) fuzzy programming method. Suitable numerical examples have been used to illustrate each of the models.

### 1.4.6 Multi-objective Rough Fuzzy Quadratic Minimum Spanning Tree Problem

A quadratic minimum spanning tree problem determines a minimum spanning tree of a network whose edges are associated with linear and quadratic weights. Linear weights represent the edge costs, whereas the quadratic weights are the interaction costs between a pair of edges of the graph. In this study, a bi-objective rough fuzzy quadratic minimum spanning tree problem (b-RFQMSTP) has been proposed for a connected graph, where the linear and the quadratic weights are represented by rough fuzzy variable. The proposed model is formulated by using rough fuzzy chance-constrained programming technique. Subsequently, three related theorems are also proposed for the crisp transformation of the proposed model. The crisp equivalent models are solved by a classical multi-objective solution technique, the epsilon-constraint method and two multi-objective evolutionary algorithms: (i) NSGA-II and (ii) MOCHC. A numerical example is provided to illustrate the proposed model when solved by different methodologies. A sensitivity analysis of the example is also performed at different confidence levels. The performance of NSGA-II and MOCHC are analyzed on five randomly generated instances of the proposed model. Finally, a numerical illustration of an application of the proposed model is also presented in this study.

### 1.4.7 Conclusion and Future Scope

In this chapter, we summarize the overall contribution of our studies and their possible future extensions.

# Part-I

## Uncertain Single Objective Optimization Problems

- **Chapter 2**
- **Chapter 3**

# Chapter 2
# Linear Programming Model with Interval Type-2 Fuzzy Parameters and its Application to some Network Problems

# Chapter 2

# Linear Programming Model with Interval Type-2 Fuzzy Parameters and its Application to some Network Problems

## 2.1 Introduction

In practice, decision making is a problem-solving activity, which is used to identify and select alternatives based on the information and preferences of the decision makers (DMs). Making a decision implies that there exist alternative solutions of the given problem and we identify not only as many alternatives as possible but also choose the one that best fits our goals, objectives and preferred values (R. Harris 1998). Uncertainties involved in many decision making problems, in real-life can be broadly classified into two types: (i) probabilistic and (ii) non-probabilistic. If there is enough historical evidence, then probability theory is the only legitimate approach to deal with uncertainty. Conversely, if there is a lack of historical evidence, then we have no option but to invite some domain experts to evaluate the belief degree of the occurrence of an event. Representation of uncertainty by evaluating the belief degrees of domain experts is non-probabilistic. In this regard, type-1 fuzzy set (T1FS) is one of the widely used and acceptable tool to deal with non-probabilistic uncertainty. Non-probabilistic uncertainty usually occur due to the existence of uncertainties like scarcity of information, multiple sources of available data, different perceptions of a concept provided by multiple experts/decision makers and linguistic information. However, it may not always be convenient to represent such imprecise information with the exact membership function of T1FS. In that case, type-2 fuzzy set (T2FS) becomes a useful tool which provides additional degrees of freedom for modelling uncertainties compared to T1FS.

T2FS are the fuzzy sets with fuzzy membership functions proposed by L.A. Zadeh (1975a, b) and developed by Mizumoto and Tanaka (1981), and Mandel and John (2002, 2006). Motivated by the studies of those authors, Liu and Liu (2010) defined type-2 fuzzy variable (T2FV), as a function from the fuzzy possibility space to the set of real numbers. Recently, type-2 fuzzy set has attracted much attention and is applied in various fields, such as data employment analysis (Qin et al. 2011), neural networks (Aliev et al. 2011; Enke and Mehdiyev 2013), portfolio selection problem (Wu and Liu 2012), multi-criteria group decision making (T.-Y. Chen 2013), transportation problem (Liu et al. 2014a; Kundu et al. 2014a; Pramanik et al. 2015b), etc. However, due to the three-dimensional nature of T2FS, the computational overhead involved in the operations of T2FS, is very high. In this case, interval type-2 fuzzy sets (IT2FS) (Mendel et al. 2006; Wu and Mendel 2007) is widely used. IT2FS is a special case of general T2FS and computationally less complex. As an example of IT2FS, let us consider the travelling cost between two cities is “around \$ 80 − \$ 100”. For such an interval, the point values may not be equipossible or are equally important. Therefore, it can be difficult to determine the exact membership value of each point in the range of the travelling cost. To deal with such non-equipossible data, IT2FS can be considered as an alternative.

Many real-life decision making problems in network optimization involve parameters which are characteristically imprecise or uncertain. Under such circumstances, several improved theories like probability theory, fuzzy set theory, rough set theory and uncertainty theory are used to tackle imprecision. However, in this chapter, we have confined our discussion, to the studies of three network optimization problem: (i) STP, (ii) SPP and (iii) MSTP under fuzzy set theory (T1FS or T2FS). Considering fuzzy environment, several studies can be observed in the literature on transportation problem (TP) (Ammar and Youness 2005; Jiménez and Verdegay 1999; Kaur and Kumar 2012; Peidro and Vasant 2011; Ojha et al. 2009), shortest path problem (SPP) (Okada and Gen 1994; Keshavarz and Khorram 2009; Deng et al. 2012) and minimum spanning tree problem (MSTP) (Janiak and Kasperski 2008; De Almeida et al. 2005a, b). In this context, chance-constrained model (CCM) is considered as one of the important approaches while formulating uncertain network problems. Under fuzzy environment, CCM is first proposed by Liu and Iwamura (1998). Thereafter, the application of CCM is observed in fuzzy TP (Yang and Liu 2007; Kundu et al. 2014a; Pramanik et al. 2015a, c), fuzzy SPP (Ji et al. 2007; Dursun and Bozdağ 2014) and fuzzy MSTP (Itoh and Ishii 1996; Gao and Lu 2005; Zhou et al. 2013, 2016a). All these studies, mentioned above are done by considering type-1 fuzzy parameters. However, under type-2 fuzzy environment, relatively few research works have been done for transportation problem, shortest path problem and minimum spanning tree problem. Considering TP, Kundu et al. (2014a) first solved the fixed charge transportation problem (FTP) with generalized triangular and trapezoidal type-2 fuzzy parameters. Consequently, Pramanik et al. (2015b) addressed FTP in a two-stage supply chain network with generalized Gaussian type-2 fuzzy parameters. Under interval type-2 fuzzy environment, Figueroa-García and Hernández (2014) first considered a transportation problem with interval type-2 fuzzy supplies and demands. Subsequently, Sinha et al. (2016) solved an expected value model (EVM) of profit maximization and time minimization solid transportation problem (STP) with interval type-2 fuzzy variables. For SPP, Anusuya and Sathya (2014) first solved the shortest path problem by considering edge weight as discrete

type-2 fuzzy variable. Later, Kumar et al. (2017) addressed the problem using generalized type-2 triangular fuzzy variables. In spite of all the related studies of TP, SPP and MSTP as mentioned above, there are some gaps in the literature as follow.

- A network optimization problem is not yet solved as a chance-constrained model with interval type-2 fuzzy parameters.
- A generalized credibility measure of interval type-2 fuzzy variable is yet to be considered for solving STP, SPP and MSTP under type-2 fuzzy environment.

In order to eliminate the lacunae mentioned above, in this chapter, we have proposed a CCM to solve three different network problems: (i) STP (ii) SPP and (iii) MSTP. For STP, the coefficients of the objective function are considered as crisp, and the constraints of availability, demand and conveyance capacities are represented by interval type-2 fuzzy parameters. Whereas, for SPP and MSTP, the coefficients of the objective function are interval type-2 fuzzy parameters subject to a set of crisp constraints.

The rest of the chapter is organized as follow. In Section 2.2 some results about credibility measure on generalized trapezoidal fuzzy variable are introduced. In Section 2.3, we have developed two chance-constrained models to solve linear programming problems (LPPs) whose associated parameters are considered as IT2FVs. In Section 2.4, the first proposed model is applied to solve an STP, where the availabilities, demands and conveyance capacities are represented by interval type-2 fuzzy parameters. In Section 2.5, the second proposed CCM is used to formulate a fuzzy SPP and a fuzzy MSTP with coefficients of the objective function (edge costs) considered as IT2FVs. Corresponding numerical examples of STP, SPP and MSTP are provided in Section 2.6, and the corresponding results are discussed. Finally, the chapter is concluded in Section 2.7.

## 2.2 Credibility Measure on Generalized Trapezoidal Fuzzy Variable

For a generalized trapezoidal fuzzy variable (GTrFV) (cf. Example 1.3.4) $\tilde{\xi} = (a, b, c, d; w)$, we have

$$\tilde{Cr}\{\tilde{\xi} \le x\} = \frac{1}{2}\left(w + \sup_{r \le x} \mu_{\tilde{\xi}}(r) - \sup_{r > x} \mu_{\tilde{\xi}}(r)\right)$$

$$= \frac{1}{2}(w + 0 - w) = 0, \text{ if } x \le a$$

$$= \frac{1}{2}\left(w + \frac{w(x-a)}{b-a} - w\right) = \frac{w(x-a)}{2(b-a)}, \text{ if } a < x \le b$$

$$= \frac{1}{2}(w + w - w) = \frac{w}{2}, \text{ if } b < x \le c$$

$$= \frac{1}{2}\left(w + w - \frac{w(d-x)}{d-c}\right) = \frac{w(x+d-2c)}{2(d-c)}, \text{ if } c < x \le d$$

$$= \frac{1}{2}(w + w - 0) = w, \text{ if } x > d,$$

$$\text{i.e., } \tilde{C}r\{\tilde{\xi} \leq x\} = \begin{cases} 0 & ; if\ x \leq a \\ \frac{w(x-a)}{2(b-a)} & ; if\ a < x \leq b \\ \frac{w}{2} & ; if\ b < x \leq c \\ \frac{w(x+d-2c)}{2(d-c)} & ; if\ c < x \leq d \\ w & ; if\ x > d. \end{cases} \tag{2.1}$$

**Theorem 2.2.1**: If $\tilde{\xi} = (a, b, c, d; w)$ is a generalized trapezoidal fuzzy variable and $0 < \alpha \leq 1$, then $\tilde{C}r\{\tilde{\xi} \leq x\} \geq \alpha$ is equivalent to

(i) $\frac{1}{w}\big((w - 2\alpha)a + 2\alpha b\big) \leq x$, if $\alpha \leq \frac{w}{2}$

(ii) $\frac{1}{w}(2(w - \alpha)c + (2\alpha - w)d) \leq x$, if $\alpha > \frac{w}{2}$.

**Proof**: For a predetermined value of $\alpha$ and if $a \leq x \leq b$ then it follows from (2.1) that

$$\tilde{C}r\{\tilde{\xi} \leq x\} \geq \alpha \Rightarrow \frac{w(x-a)}{2(b-a)} \geq \alpha$$

$$\Rightarrow \frac{1}{w}\big((w - 2\alpha)a + 2\alpha b\big) \leq x.$$

However, in this case (i.e., $a \leq x \leq b$), the maximum possible value of $\tilde{C}r\{\tilde{\xi} \leq x\}$ can be $\frac{w}{2}$, and the minimum possible value is 0 so that the value of $\alpha$ must be less than or equal to $\frac{w}{2}$, i.e., $\alpha \leq \frac{w}{2}$.

Similarly, for $c \leq x \leq d$ and a predetermined value $\alpha$, it follows from (2.1) that

$$\tilde{C}r\{\tilde{\xi} \leq x\} \geq \alpha \Rightarrow \frac{w(x+d-2c)}{2(d-c)} \geq \alpha$$

$$\Rightarrow \frac{1}{w}(2(w - \alpha)c + (2\alpha - w)d) \leq x.$$

Here, the minimum possible value of $\tilde{C}r\{\tilde{\xi} \leq x\}$ is $\frac{w}{2}$, and the maximum possible value is $w$.

**Corollary 2.2.1**: If $\tilde{\xi} = (a, b, c, d; w)$ is a generalized trapezoidal fuzzy variable and $0 < \alpha \leq 1$, then $\tilde{C}r\{\tilde{\xi} \geq x\} \geq \alpha$ is equivalent to

(i) $\frac{1}{w}\big((w - 2\alpha)d + 2\alpha c\big) \geq x$, if $\alpha \leq \frac{w}{2}$

(ii) $\frac{1}{w}(2(w - \alpha)b + (2\alpha - w)a) \geq x$, if $\alpha > \frac{w}{2}$.

## 2.3 Chance-constrained Model (CCM) to Solve LPP

In this section, we consider two different chance-constrained models: (i) Model-I and (ii) Model-II, of LPP. For Model-I, the parameters involve in the objective functions and the constraints are respectively, crisp and imprecise in nature. Whereas, for Model-II, the corresponding parameters of the objective functions and the constraints are characteristically considered as imprecise and crisp in nature. Here, the imprecise parameters of both the models are considered as IT2FVs. Each of these models is presented in subsequent subsections.

### 2.3.1 Chance-constrained Model with Interval Type-2 Fuzzy Constraints

In this section, we discuss about formulation of CCM of Model–I. Accordingly, we consider an LPP as shown in (2.2) with constraints involving IT2FVs.

$$\begin{cases} Min\ Z = \sum_{j=1}^{n} c_j x_j \\ subject\ to \\ \sum_{j=1}^{n} a_{ij} x_j \leq \tilde{b}_i\ , i = 1,2,\dots,p \\ \sum_{j=1}^{n} d_{kj} x_j \geq \tilde{e}_k\ , k = 1,2,\dots,m \\ x_j \geq 0, \forall j, \end{cases} \tag{2.2}$$

where $c_j$ is the associated crisp value of the objective function, and $\tilde{b}_i$ and $\tilde{e}_k$ are IT2FVs represented by $\tilde{b}_i = \left(\tilde{b}_i^U, \tilde{b}_i^L\right)$ and $\tilde{e}_k = (\tilde{e}_k^U, \tilde{e}_k^L)$. Here, $\tilde{b}_i^U, \tilde{b}_i^L, \tilde{e}_k^U$ and $\tilde{e}_k^L$ are T1FVs. Correspondingly, $\tilde{b}_i^U$ and $\tilde{b}_i^L$ construct the upper and lower membership functions $\bar{\mu}_{\tilde{b}_i^U}$ and $\underline{\mu}_{\tilde{b}_i^L}$ for $\tilde{b}_i$, and for $\tilde{e}_k$, $\tilde{e}_k^U$ and $\tilde{e}_k^L$, respectively construct the upper and lower membership functions $\bar{\mu}_{\tilde{e}_k^U}$ and $\underline{\mu}_{\tilde{e}_k^L}$. In this study, we have considered, the IT2FVs as the generalized trapezoidal IT2FVs, where each T1FV is a GTrFV.

**Steps for formulating CCM for the model (2.2)**

**Step 1**: Construct CCM for the model (2.2), considering the upper membership functions (UMFs) of the IT2FVs using generalized credibility measure as follows.

$$\begin{cases} Min\ Z = \sum_{j=1}^{n} c_j x_j \\ subject\ to \\ \tilde{C}r\left\{\sum_{j=1}^{n} a_{ij} x_j \leq \tilde{b}_i^U\right\} \geq \alpha_i^U, i = 1,2,\dots,p \\ \tilde{C}r\left\{\sum_{j=1}^{n} d_{kj} x_j \geq \tilde{e}_k^U\right\} \geq \beta_k^U, k = 1,2,\dots,m \\ x_j \geq 0, \forall j, \end{cases} \tag{2.3}$$

where $\alpha_i^U$ and $\beta_k^U$ are the predetermined confidence levels of satisfaction of the respective constraints.

**Step 2**: Formulate the deterministic transformation of the model (2.3) as shown in the model (2.4).

$$\begin{cases} Min\ Z = \sum_{j=1}^{n} c_j x_j \\ subject\ to \\ \sum_{j=1}^{n} a_{ij} x_j \leq F_{\tilde{b}_i^U}\ , i = 1,2,\dots,p \\ \sum_{j=1}^{n} d_{kj} x_j \geq F_{\tilde{e}_k^U}\ , k = 1,2,\dots,m \\ x_j \geq 0, \forall j, \end{cases} \tag{2.4}$$

where the deterministic transformations of the two constraints of model (2.4) are obtained by following Corollary 2.2.1 and Theorem 2.2.1, respectively. $F_{\tilde{b}_i^U}$ and $F_{\tilde{e}_k^U}$ are the resulting defuzzified values corresponding to the GTrFVs $\tilde{b}_i^U$ and $\tilde{e}_k^U$.

**Step 3**: Solve model (2.4) and suppose after solving we have $Min\ Z = Z'$.

**Step 4**: Construct CCM for the model (2.2), considering the lower membership functions (LMFs) of the IT2FVs using generalized credibility measure as follows.

$$\begin{cases} Min\ Z = \sum_{j=1}^{n} c_j x_j \\ subject\ to \\ \tilde{C}r\{\sum_{j=1}^{n} a_{ij} x_j \le \tilde{b}_i^L\} \ge \alpha_i^L, i = 1,2,\dots,p \\ \tilde{C}r\{\sum_{j=1}^{n} d_{kj} x_j \ge \tilde{e}_k^L\} \ge \beta_k^L, k = 1,2,\dots,m \\ \quad x_j \ge 0, \forall j, \end{cases} \tag{2.5}$$

where $\alpha_i^L$ and $\beta_k^L$ are the predetermined confidence levels of satisfaction of the respective constraints.

**Step 5**: Formulate the deterministic transformation of the model (2.5) as follows.

$$\begin{cases} Min\ Z = \sum_{j=1}^{n} c_j x_j \\ subject\ to \\ \sum_{j=1}^{n} a_{ij} x_j \le F_{\tilde{b}_i^L}, i = 1,2,\dots,p \\ \sum_{j=1}^{n} d_{kj} x_j \ge F_{\tilde{e}_k^L}, k = 1,2,\dots,m \\ x_j \ge 0, \forall j, \end{cases} \tag{2.6}$$

where the deterministic transformation of the two constraints of the model (2.6) are obtained in the same way as mentioned in Step 2, and $F_{\tilde{b}_i^L}$ and $F_{\tilde{e}_k^L}$ are respectively, the resulting defuzzified values of $\tilde{b}_i^L$ and $\tilde{e}_k^L$.

**Step 6**: Solve model (2.6) and suppose after solving we have $Min\ Z = Z''$.

**Step 7**: Let us denote $Z^{min} = Min(Z', Z'')$, $Z^{max} = Max(Z', Z'')$, $F_{\tilde{b}_i}^{min} = Min\left(F_{\tilde{b}_i^U}, F_{\tilde{b}_i^L}\right)$, $F_{\tilde{b}_i}^{max} = Max\left(F_{\tilde{b}_i^U}, F_{\tilde{b}_i^L}\right)$, $F_{\tilde{e}_k}^{min} = Min\left(F_{\tilde{e}_k^U}, F_{\tilde{e}_k^L}\right)$ and $F_{\tilde{e}_k}^{max} = Max\left(F_{\tilde{e}_k^U}, F_{\tilde{e}_k^L}\right)$. Here, in order to determine a compromise solution within the interval $[Z^{min}, Z^{max}]$, we use fuzzy programming method (H.-J. Zimmermann 1978). Accordingly, we determine the following membership function (for the minimization problem).

$$\mu(Z) = \begin{cases} 1 & ; if\ Z \le Z^{min} \\ \frac{Z^{max} - Z}{Z^{max} - Z^{min}} & ; if\ Z^{min} < Z < Z^{max} \\ 0 & ; if\ Z \ge Z^{max}. \end{cases}$$

Here, maximum $\mu(Z)$ means a better solution for a minimization problem. Now to obtain a compromise solution by ensuring the maximum possible satisfaction of the constraints, we consider an auxiliary variable $\lambda\,(0 \leq \lambda \leq 1)$ and construct the following compromise model.

$$\begin{cases} Max\,\lambda \\ subject\ to \\ \quad \mu(Z) \geq \lambda \Rightarrow Z \leq Z^{max} - \lambda\left(Z^{max} - Z^{min}\right) \\ \quad \sum_{j=1}^{n} a_{ij}x_j \leq F_{\tilde{b}_i}^{max} - \lambda\left(F_{\tilde{b}_i}^{max} - F_{\tilde{b}_i}^{min}\right), i = 1,2,\dots,p \\ \quad \sum_{j=1}^{n} d_{kj}x_j \geq F_{\tilde{e}_k}^{min} + \lambda\left(F_{\tilde{e}_k}^{max} - F_{\tilde{e}_k}^{min}\right), k = 1,2,\dots,m \\ \quad\; x_j \geq 0, \forall j. \end{cases} \tag{2.7}$$

Here, $Max\,\lambda$ acts as a global satisfaction degree of all constraints regarding the best possible objective value. Solving model (2.7), we obtain a compromise solution which lies between $Z^{min}$ and $Z^{max}$.

### 2.3.2 Chance-constrained Model with Interval Type-2 Fuzzy Objective

In this section, we discuss about formulation of CCM of Model–II. In this context, we consider an LPP in (2.8), where the associated parameters of the objective function are regarded as IT2FVs.

$$\begin{cases} Min\,Z = \sum_{j=1}^{n} \tilde{c}_j x_j \\ subject\ to \\ \sum_{j=1}^{n} a_{ij}x_j \leq b_i, i = 1,2,\dots,p \\ \sum_{j=1}^{n} d_{kj}x_j = e_k, k = 1,2,\dots,m \\ \quad x_j \geq 0, \forall j, \end{cases} \tag{2.8}$$

where $\tilde{c}_j$s are the IT2FVs represented by $\tilde{c}_j = \left(\tilde{c}_j^U, \tilde{c}_j^L\right)$ such that $\tilde{c}_j^U$ and $\tilde{c}_j^L$ are GTrFVs.

**Steps for formulating CCM for the model (2.8)**

**Step 1**: Construct CCM for the model (2.8), considering the upper membership functions (UMFs) of the IT2FVs as follows.

$$\begin{cases} Min\,\overline{Z} \\ subject\ to \\ \tilde{C}r\left\{\sum_{j=1}^{n} \tilde{c}_j^U x_j \leq \overline{Z}\right\} \geq \gamma^U \\ \sum_{j=1}^{n} a_{ij}x_j \leq b_i, i = 1,2,\dots,p \\ \sum_{j=1}^{n} d_{kj}x_j = e_k, k = 1,2,\dots,m \\ \quad x_j \geq 0, \forall j, \end{cases} \tag{2.9}$$

where $\gamma^U$ is the predetermined confidence level of satisfaction of the chance-constraint.

**Step 2**: Formulate the deterministic transformation of the model (2.9) as shown in the model (2.10).

$$\begin{cases} Min\ Z_1 = \sum_{j=1}^{n} F_{\tilde{c}_j^U} x_j \\ subject\ to \\ \sum_{j=1}^{n} a_{ij} x_j \le b_i,\ i = 1,2,\dots,p \\ \sum_{j=1}^{n} d_{kj} x_j = e_k,\ k = 1,2,\dots,m \\ x_j \ge 0, \forall j, \end{cases} \tag{2.10}$$

where $F_{\tilde{c}_j^U}$ is the deterministic transformation of the GTrFVs $\tilde{c}_j^U$ of the model (2.9), obtained by following Theorem 2.2.1.

**Step 3**: Solve model (2.10) and suppose after solving we have $Min\ \ Z_1 = Z'$.

**Step 4**: Construct CCM for the model (2.8) by considering the lower membership functions (LMFs) of the IT2FVs as follows.

$$\begin{cases} Min\ \overline{Z} \\ subject\ to \\ \tilde{C}r\{\sum_{j=1}^{n} \tilde{c}_j^L x_j \le \overline{Z}\} \ge \gamma^L \\ \sum_{j=1}^{n} a_{ij} x_j \le b_i,\ i = 1,2,\dots,p \\ \sum_{j=1}^{n} d_{kj} x_j = e_k,\ k = 1,2,\dots,m \\ x_j \ge 0, \forall j, \end{cases} \tag{2.11}$$

where $\gamma^L$ is the predetermined confidence level of satisfaction of the chance-constraint.

**Step 5**: Formulate the deterministic transformation of the model (2.11) as shown in the model (2.12).

$$\begin{cases} Min\ Z_1 = \sum_{j=1}^{n} F_{\tilde{c}_j^L} x_j \\ subject\ to \\ \sum_{j=1}^{n} a_{ij} x_j \le b_i,\ i = 1,2,\dots,p \\ \sum_{j=1}^{n} d_{kj} x_j = e_k,\ k = 1,2,\dots,m \\ x_j \ge 0, \forall j, \end{cases} \tag{2.12}$$

where $F_{\tilde{c}_j^L}$ is the crisp transformation of the GTrFVs $\tilde{c}_j^L$ of the model (2.11), which is obtained by following Theorem 2.2.1.

**Step 6**: Solve model (2.12) and suppose after solving we have $Min\ \ Z_1 = Z''$.

**Step 7**: Let $Z^{min} = Min(Z', Z'')$ and $Z^{max} = Max(Z', Z'')$. Now, to determine a compromise solution within the interval $[Z^{min}, Z^{max}]$ we use linear weighted method (with equal weights) and the corresponding compromise model is shown in (2.13).

$$\begin{cases} Min\ Z^{comp} = \frac{1}{2}(Z^{min} + Z^{max}) \\ subject\ to \\ \sum_{j=1}^{n} a_{ij} x_j \le b_i,\ i = 1,2,\dots,p \\ \sum_{j=1}^{n} d_{kj} x_j = e_k,\ k = 1,2,\dots,m \\ x_j \ge 0, \forall j. \end{cases} \tag{2.13}$$

## 2.4 Formulation of CCM for STP with Availabilities, Demands and Conveyance Capacities as IT2FVs

In this section, we formulate a CCM of solid transportation problem (STP) with availabilities, demands and conveyance capacities as IT2FVs using our proposed Model-I (cf. Section 2.3.1), and present the model in (2.14).

$$\begin{cases} Min\ Z = \sum_{i=1}^{m}\sum_{j=1}^{n}\sum_{k=1}^{K} c_{ijk}x_{ijk} \\ subject\ to \\ \sum_{j=1}^{n}\sum_{k=1}^{K} x_{ijk} \le \tilde{a}_i,\ \ i = 1,2,\dots,m \\ \sum_{i=1}^{m}\sum_{k=1}^{K} x_{ijk} \ge \tilde{b}_j,\ \ j = 1,2,\dots,n \\ \sum_{i=1}^{m}\sum_{j=1}^{n} x_{ijk} \le \tilde{e}_k,\ \ k = 1,2,\dots,K \\ x_{ijk} \ge 0, \forall i,j,k, \end{cases} \tag{2.14}$$

where each $c_{ijk}$ is the unit transportation cost from source $i$ to destination $j$ with conveyance $k$, $\tilde{a}_i$ denotes the amount of available item at the source $i$, $\tilde{b}_j$ is the demand of the item at destination $j$ and $\tilde{e}_k$ is the available transportation capacity of conveyance $k$. $\tilde{a}_i$, $\tilde{b}_j$ and $\tilde{e}_k$ are considered as IT2FVs and represented as follow.

$$\tilde{a}_i=(\tilde{a}_i^U,\tilde{a}_i^L) = \left((a_{i1}^U,a_{i2}^U,a_{i3}^U,a_{i4}^U;w_{a_i}^U),(a_{i1}^L,a_{i2}^L,a_{i3}^L,a_{i4}^L;w_{a_i}^L)\right),$$

$$\tilde{b}_j=(\tilde{b}_j^U,\tilde{b}_j^L)=\left((b_{j1}^U,b_{j2}^U,b_{j3}^U,b_{j4}^U;w_{b_j}^U),(b_{j1}^L,b_{j2}^L,b_{j3}^L,b_{j4}^L;w_{b_j}^L)\right) \text{ and}$$

$$\tilde{e}_k=(\tilde{e}_k^U,\tilde{e}_k^L)=\left((e_{k1}^U,e_{k2}^U,e_{k3}^U,e_{k4}^U;w_{e_k}^U),(e_{k1}^L,e_{k2}^L,e_{k3}^L,e_{k4}^L;w_{e_k}^L)\right).$$

We formulate the chance-constrained model of (2.14) using the steps as mentioned in Section 2.3.1.

**Step 1**: Construct the CCM considering the UMFs of (2.14) as follows.

$$\begin{cases} Min\ Z = \sum_{i=1}^{m}\sum_{j=1}^{n}\sum_{k=1}^{K} c_{ijk}x_{ijk} \\ subject\ to \\ \tilde{C}r\{\sum_{j=1}^{n}\sum_{k=1}^{K} x_{ijk} \le \tilde{a}_i^U\} \ge \alpha_i^U,\ \ i = 1,2,\dots,m \\ \tilde{C}r\{\sum_{i=1}^{m}\sum_{k=1}^{K} x_{ijk} \ge \tilde{b}_j^U\} \ge \beta_j^U,\ \ j = 1,2,\dots,n \\ \tilde{C}r\{\sum_{i=1}^{m}\sum_{j=1}^{n} x_{ijk} \le \tilde{e}_k^U\} \ge \gamma_k^U,\ \ k = 1,2,\dots,K \\ \qquad x_{ijk} \ge 0, \forall i,j,k, \end{cases} \tag{2.15}$$

where $\alpha_i^U$, $\beta_j^U$ and $\gamma_k^U$ are respectively, the predetermined credibility levels of satisfaction of the availability at the source, demand at destination and conveyance capacities.

**Step 2**: The crisp transformation of the model (2.15) is formulated as below.

Following, Corollary 2.2.1, the crisp equivalent of the first constraint of the model (2.15) can be represented as

$$\sum_{j=1}^{n}\sum_{k=1}^{K} x_{ijk} \le F_{\tilde{a}_i}^U, \tag{2.16}$$

where $F_{\tilde{a}_i}^U = \begin{cases} \frac{1}{w_{a_i}^U}\left(\left(w_{a_i}^U - 2\alpha_i^U\right)a_{i4}^U + 2\alpha_i^U a_{i3}^U\right) & ; if\ \alpha_i^U \le \frac{w_{a_i}^U}{2} \\ \frac{1}{w_{a_i}^U}\left(2\left(w_{a_i}^U - \alpha_i^U\right)a_{i2}^U + \left(2\alpha_i^U - w_{a_i}^U\right)a_{i1}^U\right) & ; if\ \alpha_i^U > \frac{w_{a_i}^U}{2}. \end{cases}$

Similarly, from Theorem 2.2.1, the crisp equivalent of the second constraint of the model (2.15) is written as

$$\sum_{i=1}^{m}\sum_{k=1}^{K} x_{ijk} \ge F_{\tilde{b}_j}^U\ , \tag{2.17}$$

where $F_{\tilde{b}_j}^U = \begin{cases} \frac{1}{w_{b_j}^U}\left(\left(w_{b_j}^U - 2\beta_j^U\right)b_{j1}^U + 2\beta_j^U b_{j2}^U\right) & ; if\ \beta_j^U \le \frac{w_{b_j}^U}{2} \\ \frac{1}{w_{a_i}^U}\left(2\left(w_{b_i}^U - \beta_j^U\right)b_{j3}^U + \left(2\beta_j^U - w_{b_j}^U\right)b_{j4}^U\right) & ; if\ \beta_j^U > \frac{w_{b_j}^U}{2}. \end{cases}$

Subsequently, from Corollary 2.2.1, the crisp equivalent of the third constraint of the model (2.15) is determined as

$$\sum_{i=1}^{m}\sum_{j=1}^{n} x_{ijk} \le F_{\tilde{e}_k}^U\ , \tag{2.18}$$

where $F_{\tilde{e}_k}^U = \begin{cases} \frac{1}{w_{e_k}^U}\left(\left(w_{e_k}^U - 2\gamma_k^U\right)e_{k4}^U + 2\gamma_k^U e_{k3}^U\right) & ; if\ \gamma_k^U \le \frac{w_{e_k}^U}{2} \\ \frac{1}{w_{e_k}^U}\left(2\left(w_{e_k}^U - \gamma_k^U\right)e_{k2}^U + \left(2\gamma_k^U - w_{e_k}^U\right)e_{k1}^U\right) & ; if\ \gamma_k^U > \frac{w_{e_k}^U}{2}. \end{cases}$

Therefore, the crisp transformation of the model (2.15) becomes

$$\begin{cases} Min\ Z = \sum_{i=1}^{m}\sum_{j=1}^{n}\sum_{k=1}^{K} c_{ijk}x_{ijk} \\ subject\ to\ the\ contraints\ (2.16), (2.17), (2.18) \\ \quad x_{ijk} \ge 0, \forall i, j, k. \end{cases} \tag{2.19}$$

Here, it is to be noted that model (2.19) generates feasible solution only if $\sum_{i=1}^{m} F_{\tilde{a}_i}^U \ge \sum_{j=1}^{n} F_{\tilde{b}_j}^U$ and $\sum_{k=1}^{K} F_{\tilde{e}_k}^U \ge \sum_{j=1}^{n} F_{\tilde{b}_j}^U$.

**Step 3**: Solve model (2.19) and suppose after solving we have $Min\ \ Z = Z'$.

**Step 4**: Construct the CCM considering the LMFs of (2.14) as follows.

$$\begin{cases} Min\ Z = \sum_{i=1}^{m}\sum_{j=1}^{n}\sum_{k=1}^{K} c_{ijk}x_{ijk} \\ subject\ to \\ \tilde{C}r\left\{\sum_{j=1}^{n}\sum_{k=1}^{K} x_{ijk} \le \tilde{a}_i^L\right\} \ge \alpha_i^L,\ i = 1,2,\dots,m \\ \tilde{C}r\left\{\sum_{i=1}^{m}\sum_{k=1}^{K} x_{ijk} \ge \tilde{b}_j^{\ L}\right\} \ge \beta_j^L,\ \ j = 1,2,\dots,n \\ \tilde{C}r\left\{\sum_{i=1}^{m}\sum_{j=1}^{n} x_{ijk} \le \tilde{e}_k^{\ L}\right\} \ge \gamma_k^L,\ \ k = 1,2,\dots,K \\ \quad x_{ijk} \ge 0, \forall i, j, k, \end{cases} \tag{2.20}$$

where $\alpha_i^L$, $\beta_j^L$ and $\gamma_k^L$ are respectively, the predetermined credibility levels of satisfaction of the availability at the source, demand at destination and conveyance capacities.

**Step 5**: The crisp transformation of model (2.20) is formulated below.

Similar to Step 2, following the Theorem 2.2.1 and Corollary 2.2.1, the crisp equivalent of the first, second and third constraints of the model (2.20) are respectively, shown in (2.21), (2.22) and (2.23).

$$\sum_{j=1}^{n}\sum_{k=1}^{K} x_{ijk} \leq F_{\tilde{a}_i}^{L} \tag{2.21}$$

$$\sum_{i=1}^{m}\sum_{k=1}^{K} x_{ijk} \geq F_{\tilde{b}_j}^{L} \tag{2.22}$$

$$\sum_{i=1}^{m}\sum_{j=1}^{n} x_{ijk} \leq F_{\tilde{e}_k}^{L}, \tag{2.23}$$

where

$$F_{\tilde{a}_i}^{L} = \begin{cases} \frac{1}{w_{a_i}^{L}}\left(\left(w_{a_i}^{L} - 2\alpha_i^{L}\right)a_{i4}^{L} + 2\alpha_i^{L}a_{i3}^{L}\right) & ; if\ \alpha_i^{L} \leq \frac{w_{a_i}^{L}}{2} \\ \frac{1}{w_{a_i}^{L}}\left(2\left(w_{a_i}^{L} - \alpha_i^{L}\right)a_{i2}^{L} + \left(2\alpha_i^{L} - w_{a_i}^{L}\right)a_{i1}^{L}\right) & ; if\ \alpha_i^{L} > \frac{w_{a_i}^{L}}{2}, \end{cases}$$

$$F_{\tilde{b}_j}^{L} = \begin{cases} \frac{1}{w_{b_j}^{L}}\left(\left(w_{b_j}^{L} - 2\beta_j^{L}\right)b_{j1}^{L} + 2\beta_j^{L}b_{j2}^{L}\right) & ; if\ \beta_j^{L} \leq \frac{w_{b_j}^{L}}{2} \\ \frac{1}{w_{a_i}^{L}}\left(2\left(w_{b_i}^{L} - \beta_j^{L}\right)b_{j3}^{L} + \left(2\beta_j^{L} - w_{b_j}^{L}\right)b_{j4}^{L}\right) & ; if\ \beta_j^{L} > \frac{w_{b_j}^{L}}{2} \end{cases}$$

and

$$F_{\tilde{e}_k}^{L} = \begin{cases} \frac{1}{w_{e_k}^{L}}\left(\left(w_{e_k}^{L} - 2\gamma_k^{L}\right)e_{k4}^{L} + 2\gamma_k^{L}e_{k3}^{L}\right) & ; if\ \gamma_k^{L} \leq \frac{w_{e_k}^{L}}{2} \\ \frac{1}{w_{e_k}^{L}}\left(2\left(w_{e_k}^{L} - \gamma_k^{L}\right)e_{k2}^{L} + \left(2\gamma_k^{L} - w_{e_k}^{L}\right)e_{k1}^{L}\right) & ; if\ \gamma_k^{L} > \frac{w_{e_k}^{L}}{2}. \end{cases}$$

Consequently, the crisp transformation of the model (2.20) becomes

$$\begin{cases} Min\ Z = \sum_{i=1}^{m}\sum_{j=1}^{n}\sum_{k=1}^{K} c_{ijk}x_{ijk} \\ subject\ to\ the\ contraints\ (2.21), (2.22), (2.23) \\ \quad x_{ijk} \geq 0, \forall i,j,k. \end{cases} \tag{2.24}$$

Model (2.24) generates feasible solution only if $\sum_{i=1}^{m} F_{\tilde{a}_i}^{L} \geq \sum_{j=1}^{n} F_{\tilde{b}_j}^{L}$ and $\sum_{k=1}^{K} F_{\tilde{e}_k}^{L} \geq \sum_{j=1}^{n} F_{\tilde{b}_j}^{L}$.

**Step 6**: Solve model (2.24) and suppose after solving we have $Min\ \ Z = Z''$.

**Step 7**: Let $Z^{min} = Min(Z', Z'')$, $Z^{max} = Max(Z', Z'')$, $F_{\tilde{a}_i}^{min} = Min\left(F_{\tilde{a}_i}^{U}, F_{\tilde{a}_i}^{L}\right)$, $F_{\tilde{a}_i}^{max} = Min\left(F_{\tilde{a}_i}^{U}, F_{\tilde{a}_i}^{L}\right)$, $F_{\tilde{b}_j}^{min} = Min\left(F_{\tilde{b}_j}^{U}, F_{\tilde{b}_j}^{L}\right)$, $F_{\tilde{b}_j}^{max} = Min\left(F_{\tilde{b}_j}^{U}, F_{\tilde{b}_j}^{L}\right)$, $F_{\tilde{e}_k}^{min} = Min\left(F_{\tilde{e}_k}^{U}, F_{\tilde{e}_k}^{L}\right)$ and $F_{\tilde{e}_k}^{max} = Min\left(F_{\tilde{e}_k}^{U}, F_{\tilde{e}_k}^{L}\right)$.

In order to determine a compromise solution within the interval $\left[Z^{min}, Z^{max}\right]$, we use fuzzy programming method (H.-J. Zimmermann 1978) as mentioned in Step 7 of Section 2.3.1. Accordingly, we construct the following compromise model.

$$\begin{cases} Max\ \lambda \\ subject\ to \\ \quad Z \leq Z^{max} - \lambda(Z^{max} - Z^{min}) \\ \sum_{j=1}^{n}\sum_{k=1}^{K} x_{ijk} \leq F_{\tilde{a}_i}^{max} - \lambda(F_{\tilde{a}_i}^{max} - F_{\tilde{a}_i}^{min}), i = 1,2,\dots,m \\ \sum_{i=1}^{n}\sum_{k=1}^{K} x_{ijk} \leq F_{\tilde{b}_j}^{min} + \lambda\left(F_{\tilde{b}_j}^{max} - F_{\tilde{b}_j}^{min}\right), j = 1,2,\dots,n \\ \sum_{i=1}^{m}\sum_{j=1}^{n} x_{ijk} \leq F_{\tilde{e}_k}^{max} - \lambda(F_{\tilde{e}_k}^{max} - F_{\tilde{e}_k}^{min}), k = 1,2,\dots,K \\ x_{ijk} \geq 0, \forall i,j,k, \end{cases} \tag{2.25}$$

where $\lambda$ is the overall satisfactory level of compromise.

## 2.5 Formulation of CCM for SPP and MSTP with Interval Type-2 Fuzzy Parameters

In this section we address two network problems: (i) SPP and (ii) MSTP under interval type-2 fuzzy paradigm. The corresponding CCM of the problems is formulated in two subsequent subsections.

### 2.5.1 CCM for SPP with Interval Type-2 Fuzzy Parameters

The shortest path problem (SPP) has enticed researchers and engineers due to its wide range of applications. The SPP is encountered quite frequently as sub-problem while solving many combinatorial optimization problems (Ahuja et al. 1993; Gen et al. 2008). In this study, we formulate a CCM of the SPP by considering the associated edge weight as IT2FV.

Let $G = (V_G, E_G)$ be a weighted connected directed network (WCDN), where $V_G = \{v_1, v_2, \dots, v_n\}$ and $E_G$ are the corresponding finite sets of vertices and edges of $G$ of size $n$ and $m$, respectively. Each edge $e_{ij}$ represents a vertex pair $\langle v_i, v_j\rangle$ which connects $v_i$ and $v_j$ and is directed from $v_i$ to $v_j$. The SPP explores the shortest path between two terminal vertices, particularly designated as source $(s)$ and sink $(t)$ vertices. The formulation of SPP under type-2 fuzzy environment is presented below.

$$\begin{cases} Min\ Z = \sum_{i=1}^{n}\sum_{j=1}^{n} \tilde{c}_{ij} x_{ij} \\ subject\ to \\ \quad \sum_{j=1}^{n} x_{ij} - \sum_{k=1}^{n} x_{ki} = \begin{cases} 1 & ; if\ i = 1 \\ 0 & ; if\ i = (2,3,\dots,n-1) \\ -1 & ; if\ i = n \end{cases} \\ \quad x_{ij} \in \{0,1\},\ \forall\ i,j, \end{cases} \tag{2.26}$$

where $n$ is the number of vertices of the network, $\tilde{c}_{ij}$ is the cost (or weight) associated with every $e_{ij}$. The objective function of (2.26) minimizes the total cost incurred while traversing from $v_1$ to $v_n$, which are essentially considered as source $(s)$ and sink $(t)$ vertices of $G$, respectively. The only constraint of (2.26) satisfies the flow conservation law at all the vertices except $s$ and $t$. Every decision variable $x_{ij}$ of (2.26) accepts either 0 or 1. If $x_{ij} = 1$ then the corresponding edge $e_{ij}$ is included in the shortest path from $s$ to $t$, whereas if $x_{ij} = 0$, $e_{ij}$ is excluded from the shortest path. Here, each $\tilde{c}_{ij}$ associated with an edge $e_{ij}$ is considered as an IT2FV and represented as

$$\tilde{c}_{ij} = \left(c_{ij}^U, c_{ij}^L\right) = \left(\left(c_{ij1}^U, c_{ij2}^U, c_{ij3}^U, c_{ij4}^U; w_{c_{ij}}^U\right), \left(c_{ij1}^L, c_{ij2}^L, c_{ij3}^L, c_{ij4}^L; w_{c_{ij}}^L\right)\right).$$

We formulate the chance-constrained model of the (2.26) following the steps as mentioned in Section 2.3.2.

**Step 1**: Construct CCM for the model (2.26), considering the UMFs of the IT2FVs as follows.

$$\begin{cases} Min\ \overline{Z} \\ subject\ to \\ \tilde{C}r\left\{\sum_{i=1}^{n}\sum_{j=1}^{n}\tilde{c}_{ij}^U x_{ij} \le \overline{Z}\right\} \ge \delta^U \\ constraints\ of\ (2.26), \end{cases} \quad (2.27)$$

where $\delta^U$ is the predetermined confidence level of the chance-constraint.

**Step 2**: Since $\tilde{c}_{ij}^U$ is a generalized trapezoidal fuzzy variable and $x_{ij}$ is a binary variable, so according to Theorem 2.2.1, the constraint $\tilde{C}r\left\{\sum_{i=1}^{n}\sum_{j=1}^{n}\tilde{c}_{ij}^U x_j \le \overline{Z}\right\} \ge \delta^U$ equivalently becomes $\sum_{i=1}^{n}\sum_{j=1}^{n} F_{\tilde{c}_{ij}^U} x_{ij} \le \overline{Z}$, where $F_{\tilde{c}_{ij}^U}$ is given by

$$F_{\tilde{c}_{ij}^U} = \begin{cases} \frac{1}{w_{c_{ij}}^U}\left(\left(w_{c_{ij}}^U - 2\delta^U\right)c_{ij1}^U + 2\delta^U c_{ij2}^U\right); & if\ \delta^U \le \frac{w_{c_{ij}}^U}{2} \\ \frac{1}{w_{c_{ij}}^U}\left(2\left(w_{c_{ij}}^U - \delta^U\right)c_{ij3}^U + \left(2\delta^U - w_{c_{ij}}^U\right)c_{ij4}^U\right); & if\ \delta^U > \frac{w_{c_{ij}}^U}{2}\ . \end{cases}$$

Therefore, the crisp transformation of the model (2.27) becomes

$$\begin{cases} Min\ Z_1 = \sum_{i=1}^{n}\sum_{j=1}^{n} F_{\tilde{c}_{ij}^U} x_{ij} \\ subject\ to \\ constraints\ of\ (2.26). \end{cases} \quad (2.28)$$

**Step 3**: Solve model (2.28) and suppose after solving we have $Min\ \ Z_1 = Z'$.

**Step 4**: Formulate CCM for the model (2.26), considering the LMFs of the IT2FVs as follows.

$$\begin{cases} Min\ \overline{Z} \\ subject\ to \\ \tilde{C}r\left\{\sum_{i=1}^{n}\sum_{j=1}^{n}\tilde{c}_{ij}^L x_{ij} \le \overline{Z}\right\} \ge \delta^L \\ constraints\ of\ (2.26), \end{cases} \quad (2.29)$$

where $\delta^L$ is the predetermined confidence level of the chance-constraint.

**Step 5**: Since $\tilde{c}_{ij}^L$ is a generalized trapezoidal fuzzy variable and $x_{ij}$ is a binary variable, then, following Theorem 2.2.1, the constraint $\tilde{C}r\left\{\sum_{i=1}^{n}\sum_{j=1}^{n}\tilde{c}_{ij}^L x_j \leq \overline{Z}\right\} \geq \delta^L$ equivalently becomes $\sum_{i=1}^{n}\sum_{j=1}^{n}F_{\tilde{c}_{ij}^L}x_{ij} \leq \overline{Z}$, where

$$F_{\tilde{c}_{ij}^L} = \begin{cases} \frac{1}{w_{c_{ij}}^L}\left(\left(w_{c_{ij}}^L - 2\delta^L\right)c_{ij1}^L + 2\delta^L c_{ij2}^L\right) & ; if\ \delta^L \leq \frac{w_{c_{ij}}^L}{2} \\ \frac{1}{w_{c_{ij}}^L}\left(2\left(w_{c_{ij}}^L - \delta^L\right)c_{ij3}^L + \left(2\delta^L - w_{c_{ij}}^L\right)c_{ij4}^L\right) & ; if\ \delta^L > \frac{w_{c_{ij}}^L}{2}. \end{cases}$$

Therefore, the crisp transformation of the model (2.29) becomes

$$\begin{cases} Min\ Z_1 = \sum_{i=1}^{n}\sum_{j=1}^{n}F_{\tilde{c}_{ij}^L}x_{ij} \\ subject\ to \\ \quad constraints\ of\ (2.26). \end{cases} \tag{2.30}$$

**Step 6**: Solve model (2.30) and suppose after solving we have $Min\ \ Z_1 = Z''$.

**Step 7**: Let $Z^{min} = Min(Z', Z'')$ and $Z^{max} = Max(Z', Z'')$. Now, following Step 7 of Section 2.3.2, we determine a compromise solution within the interval $\left[Z^{min}, Z^{max}\right]$ using linear weighted method (with equal weights) and the corresponding compromise model is shown in (2.31).

$$\begin{cases} Min\ Z^{comp} = \frac{1}{2}\left(\sum_{i=1}^{n}\sum_{j=1}^{n}F_{\tilde{c}_{ij}^U}x_{ij} + \sum_{i=1}^{n}\sum_{j=1}^{n}F_{\tilde{c}_{ij}^L}x_{ij}\right) \\ subject\ to \\ \quad constraints\ of\ (2.26). \end{cases} \tag{2.31}$$

### 2.5.2 CCM for MSTP with Interval Type-2 Fuzzy Parameters

A minimum spanning tree problem (MSTP) deals with determining a minimally connected acyclic tree which spans all the vertices of a connected network. The MSTP is a popular combinatorial optimization problem which has received colossal attention from researchers and engineers mainly due to its diverse application domains including roadways or railways construction, laying of utility networks for gas, water and electric supplies, designing of local access networks (Ahuja et al. 1993; Gen et al. 2008), etc. In this study, we have addressed the MSTP whose edge weights are represented by IT2FV.

Let $G = (V_G, E_G)$ is a weighted connected undirected network (WCUN) with no parallel edges, where $V_G = \{v_1, v_2, \ldots, v_n\}$ and $E_G$ are the corresponding finite vertex and edge sets of $G$ with size $n$ and $m$, respectively. Each edge $e_{ij}$ is represented by a vertex pair $\langle v_i, v_j\rangle$ which connects $v_i$ and $v_j$. The minimum spanning tree problem under type-2 fuzzy environment is formulated in (2.32).

$$\begin{cases} Min\ Z = \sum_{i=1}^{n}\sum_{j=1}^{n}\tilde{d}_{ij}x_{ij} \\ subject\ to \\ \sum_{i=1}^{|E_G|}\sum_{j=1,i\neq j}^{|E_G|}x_{ij} = |V_G| - 1 \\ \sum_{i,j\in E_{\kappa_G},i\neq j}x_{ik} \leq |\kappa_G| - 1, |\kappa_G| \geq 3 \\ \quad x_{ij} \in \{0,1\},\ \forall\, i,j, \end{cases} \tag{2.32}$$

where $E_{\kappa_G}$ is a set of edges in the subgraph of $G$ induced by the vertex set $\kappa_G$, $n$ is the number of vertices of $G$, $\tilde{d}_{ij}$ is the associated cost of each $e_{ij}$. The cost $\tilde{d}_{ij}$ is represented by an IT2FV and $x_{ij}$ is the binary decision variable. In the model (2.32), the first constraint is the cardinality constraint which determines that exactly $|V_G| - 1$ edges are present in the minimum spanning tree (MST). The second constraint ensures that there exists no cycle in the MST. Here, every $x_{ij}$ either accepts 0 or 1. If $x_{ij} = 1$, then the corresponding edge $e_{ij}$ is contained in the MST. Whereas, if $x_{ij} = 0$, then $e_{ij}$ is not considered in the MST. In this study, each $\tilde{d}_{ij}$ corresponding to an edge $e_{ij}$ is represented as

$$\tilde{d}_{ij}=\left(d_{ij}^U, d_{ij}^L\right)=\left(\left(d_{ij1}^U, d_{ij2}^U, d_{ij3}^U, d_{ij4}^U; w_{d_{ij}}^U\right), \left(d_{ij1}^L, d_{ij2}^L, d_{ij3}^L, d_{ij4}^L; w_{d_{ij}}^L\right)\right).$$

To solve the model (2.32) we follow the same procedure as described in Section 2.5.1. Accordingly, the CCMs can be formulated by considering the UMFs and the LMFs of all $\tilde{d}_{ij}$ for the model (2.32) and eventually their crisp equivalent models can be determined as shown in models (2.33) and (2.34), respectively.

$$\begin{cases} Min\ Z_1 = \sum_{i=1}^{n}\sum_{j=1}^{n}F_{\tilde{d}_{ij}^U}x_{ij} \\ subject\ to \\ \quad constraints\ of\ (2.32) \end{cases} \tag{2.33}$$

and

$$\begin{cases} Min\ Z_1 = \sum_{i=1}^{n}\sum_{j=1}^{n}F_{\tilde{d}_{ij}^L}x_{ij} \\ subject\ to \\ \quad constraints\ of\ (2.32), \end{cases} \tag{2.34}$$

where $F_{\tilde{d}_{ij}^U}=\begin{cases} \frac{1}{w_{d_{ij}}^U}\left(\left(w_{d_{ij}}^U - 2\delta^U\right)d_{ij1}^U + 2\delta^U d_{ij2}^U\right), & if\ \delta^U \leq \frac{w_{d_{ij}}^U}{2} \\ \frac{1}{w_{d_{ij}}^U}\left(2\left(w_{d_{ij}}^U - \delta^U\right)d_{ij3}^U + \left(2\delta^U - w_{d_{ij}}^U\right)d_{ij4}^U\right), & if\ \delta^U > \frac{w_{d_{ij}}^U}{2} \end{cases}$,

and $F_{\tilde{d}_{ij}^L} = \begin{cases} \frac{1}{w_{d_{ij}}^L}\left(\left(w_{d_{ij}}^L - 2\delta^L\right)d_{ij1}^L + 2\delta^L d_{ij2}^L\right), & if\ \delta^L \leq \frac{w_{d_{ij}}^L}{2} \\ \frac{1}{w_{d_{ij}}^L}\left(2\left(w_{d_{ij}}^L - \delta^L\right)d_{ij3}^L + \left(2\delta^L - w_{d_{ij}}^L\right)d_{ij4}^L\right), & if\ \delta^L > \frac{w_{d_{ij}}^L}{2} \end{cases}$.

Finally, a compromise model of (2.33) and (2.34) can be formulated using linear weighted method (with equal weights), as shown in (2.35).

$$\begin{cases} Min\ Z^{comp} = \frac{1}{2}\left(\sum_{i=1}^{n}\sum_{j=1}^{n} F_{\tilde{d}_{ij}^{U}} x_{ij} + \sum_{i=1}^{n}\sum_{j=1}^{n} F_{\tilde{d}_{ij}^{L}} x_{ij}\right) \\ subject\ to \\ \quad constraints\ of\ (2.26). \end{cases} \tag{2.35}$$

Here, $Z^{comp}$ determine a compromise solution within the interval $[Z^{min}, Z^{max}]$, where $Z^{min}$ and $Z^{max}$ for MSTP are calculated in the similar way as mentioned in Step 7 in Section 2.5.1.

## 2.6 Numerical Illustrations

This section consists of three subsections, where we discuss the results of the proposed models of STP, SPP and MSTP. Particularly, in Subsection 2.6.1 we provide the results of the proposed model of STP followed by the Subsection 2.6.2, where the related results of the proposed CCM of SPP are discussed. Finally, the results of the proposed model of MSTP are presented in Subsection 2.6.3. The numerical results of all the models are solved using an optimization software, LINGO 11.0. The basic idea of CCM (Charnes and Cooper 1959; B. Liu 2002; X. Huang 2007a) is that it allows violation of constraints. However, a decision maker needs to ensure that the constraints should hold at some chance level (confidence level). Due to this unique characteristic, the model cannot be compared with any uncertain models of STP, SPP and MSTP with interval type-2 fuzzy parameters.

### 2.6.1 STP with Availabilities, Demands and Conveyance Capacities as IT2FVs

We consider a solid transportation problem with two sources, two destinations and two conveyances, i.e., $i,j,k$=1,2. Here, the availability at sources, demand at destinations and the capacity of conveyances are considered as imprecise and are represented by IT2FV. The unit transportation costs are considered as crisp values. All these associated parameters of STP are reported in Table 2.1-Table 2.4.

Table 2.1 Transportation cost of the product from source $i$ to destination $j$ via conveyance $k$

| $c_{ij1}$ | 1 | 2 | $c_{ij2}$ | 1 | 2 |
|---|---|---|---|---|---|
| 1 | 10 | 8 | 1 | 11 | 14 |
| 2 | 13 | 9 | 2 | 15 | 12 |

Table 2.2 Availability of the product at different sources represented by IT2FV

| $i$ | 1 | 2 |
|---|---|---|
| $\tilde{a}_i$ | $\begin{pmatrix}(38, 40, 43, 45; 1.0),\\(39,41,42,44; 0.8)\end{pmatrix}$ | $\begin{pmatrix}(29, 32, 35,37; 1.0),\\(30,33,34,36; 0.8)\end{pmatrix}$ |

Table 2.3 Demand of the product at different destinations represented by IT2FV

| $j$ | 1 | 2 |
|---|---|---|
| $\tilde{b}_j$ | $\begin{pmatrix}(28, 29, 30, 32; 1.0),\\(29,29.5,30.5,31; 0.8)\end{pmatrix}$ | $\begin{pmatrix}(31, 33, 34,36; 1.0),\\(32,33.5,34,35; 0.8)\end{pmatrix}$ |

Table 2.4 Capacity of different types of conveyances represented by IT2FV

| $k$ | 1 | 2 |
|---|---|---|
| $\tilde{e}_k$ | $\begin{pmatrix}(29,31,34,36;1.0),\\(30,31.5,33.5,35;0.8)\end{pmatrix}$ | $\begin{pmatrix}(39,41,44,47;1.0),\\(40,42,43,46;0.8)\end{pmatrix}$ |

To solve the corresponding CCM of the STP, we consider the predetermined confidence levels as $\alpha_1^U = \alpha_2^U = 0.9$, $\beta_1^U = \beta_2^U = 0.9$ and $\gamma_1^U = \gamma_2^U = 0.9$. Now, considering the UMFs of the associated IT2FVs, and following model (2.19), the crisp equivalent model of the proposed STP is shown in (2.36).

$$\begin{cases} Min\ Z = \sum_{i=1}^{2}\sum_{j=1}^{2}\sum_{k=1}^{2} c_{ijk}x_{ijk} \\ subject\ to \\ \sum_{j=1}^{2}\sum_{k=1}^{2} x_{ijk} \le F_{\tilde{a}_i}^U, i = 1,2 \\ \sum_{i=1}^{2}\sum_{j=1}^{2} x_{ijk} \ge F_{\tilde{b}_j}^U,\ j = 1,2 \\ \sum_{i=1}^{2}\sum_{j=1}^{2} x_{ijk} \le F_{\tilde{e}_k}^U,\ k = 1,2 \\ x_{ijk} \ge 0,\ \forall i,j,k, \end{cases} \tag{2.36}$$

where $F_{\tilde{a}_i}^U$, $F_{\tilde{b}_j}^U$ and $F_{\tilde{e}_k}^U$ are respectively, determined from the eqns. (2.16), (2.17) and (2.18), and are given as $F_{\tilde{a}_1}^U = 38.4000$, $F_{\tilde{a}_2}^U = 29.6000$, $F_{\tilde{b}_1}^U = 31.6000$, $F_{\tilde{b}_2}^U = 35.6000$, $F_{\tilde{e}_1}^U = 29.4000$ and $F_{\tilde{e}_2}^U = 39.4000$. Subsequently, solving (2.36) we get $Min\ Z = Z' = 679.8000$.

$$\begin{cases} Min\ Z = \sum_{i=1}^{2}\sum_{j=1}^{2}\sum_{k=1}^{2} c_{ijk}x_{ijk} \\ subject\ to \\ \sum_{j=1}^{2}\sum_{k=1}^{2} x_{ijk} \le F_{\tilde{a}_i}^L, i = 1,2 \\ \sum_{i=1}^{2}\sum_{j=1}^{2} x_{ijk} \ge F_{\tilde{b}_j}^L,\ j = 1,2 \\ \sum_{i=1}^{2}\sum_{j=1}^{2} x_{ijk} \le F_{\tilde{e}_k}^L,\ k = 1,2 \\ x_{ijk} \ge 0,\ \forall i,j,k, \end{cases} \tag{2.37}$$

where $F_{\tilde{a}_i}^L$, $F_{\tilde{b}_j}^L$ and $F_{\tilde{e}_k}^L$ are respectively, determined following eqns. (2.21), (2.22) and (2.23), and provided as follow. $F_{\tilde{a}_1}^L = 39.5000$, $F_{\tilde{a}_2}^L = 30.7500$, $F_{\tilde{b}_1}^L = 30.8750$, $F_{\tilde{b}_2}^L = 34.7500$, $F_{\tilde{e}_1}^L = 30.3750$ and $F_{\tilde{e}_2}^L = 40.5000$. Consequently, solving model (2.24), we get $Min\ Z = Z'' = 656.8750$. Thereafter, to generate a compromise solution within the interval transportation cost $[656.8750, 679.8000]$, we solve model (2.25) for the problem as given below. The compromise solution along with its compromise transportation plan is reported in Table 2.5.

$$\begin{cases} Max\ \lambda \\ subject\ to \\ \quad Z \le 679.8000 - \lambda(679.8000 - 656.8750) \\ \quad \sum_{j=1}^{2}\sum_{k=1}^{2} x_{ijk} \le 39.5000 - \lambda(39.5000 - 38.4000) \\ \quad \sum_{j=1}^{2}\sum_{k=1}^{2} x_{ijk} \le 30.7500 - \lambda(30.7500 - 29.6000) \\ \quad \sum_{i=1}^{2}\sum_{k=1}^{2} x_{ijk} \le 30.8750 + \lambda(31.6000 - 30.8750) \\ \quad \sum_{i=1}^{2}\sum_{k=1}^{2} x_{ijk} \le 34.7500 + \lambda(35.6000 - 34.7500) \\ \quad \sum_{i=1}^{2}\sum_{j=1}^{2} x_{ijk} \le 30.3750 - \lambda(30.3750 - 29.4000) \\ \quad \sum_{i=1}^{2}\sum_{j=1}^{2} x_{ijk} \le 40.5000 - \lambda(40.5000 - 39.4000) \\ \quad x_{ijk} \ge 0, \forall i,j,k. \end{cases} \tag{2.38}$$

Table 2.5 Compromise solution generated by the model (2.38)

| $\lambda$ | $Z$ | Transportation plan |
|---|---|---|
| 0.5 | 668.3375 | $x_{121} = 7.7125, x_{221} = 22.1750, x_{112} = 31.2375, x_{222} = 5.2875$ |

We have also generated a compromise solution of the model (2.25), at different predetermined confidence levels. The corresponding results are presented in Table 2.6. Here, it is observed that a compromise transportation cost increases progressively as the confidence levels are increased. A graphical interpretation of the generated compromise solution at different confidence levels is shown in Fig. 2.1.

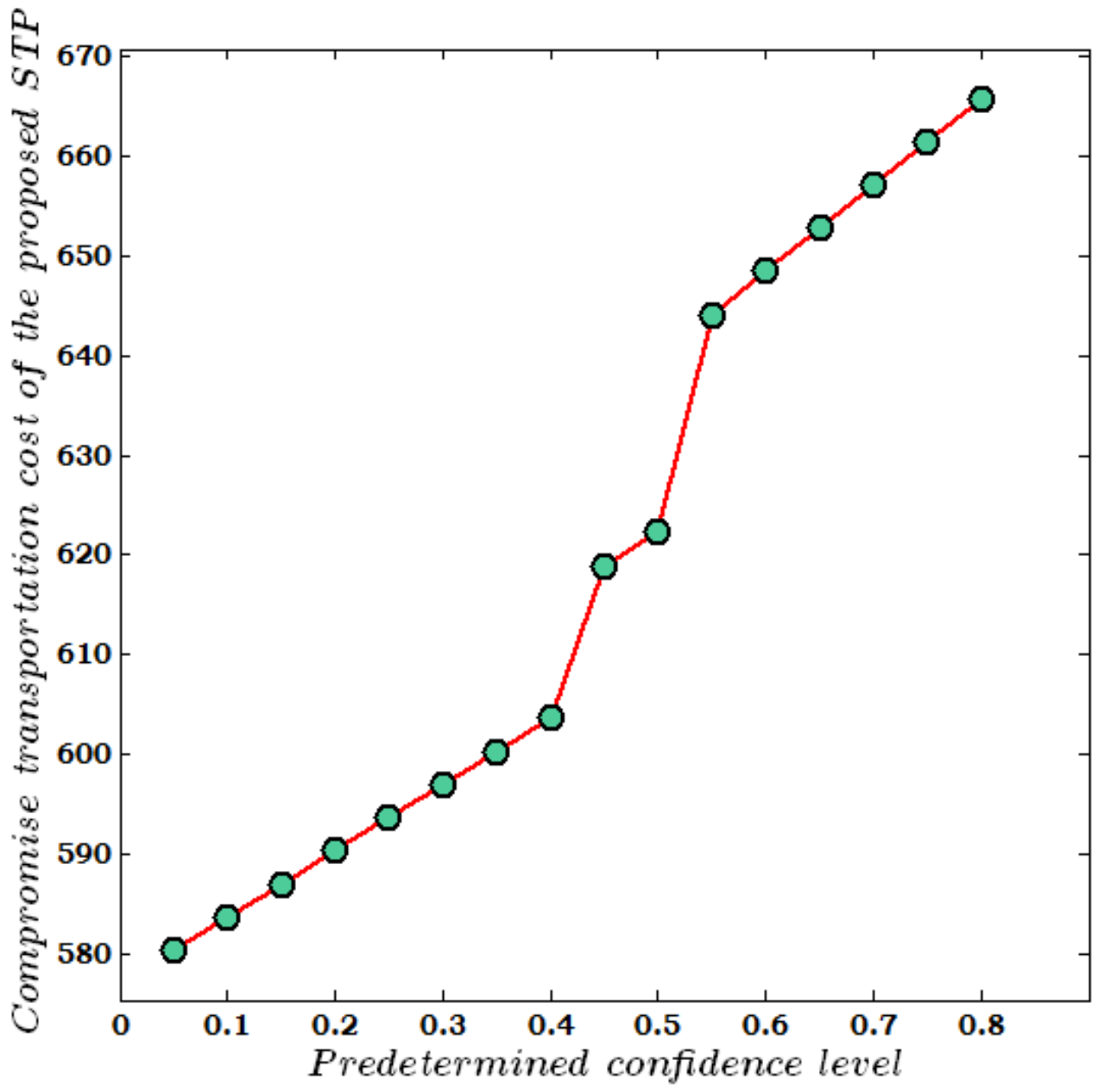

Figure 2.1 Compromise transportation cost for the proposed STP generated by solving model (2.25) at different confidence levels

Table 2.6 Compromise cost of the STP at different predetermined confidence levels

| Predetermined confidence levels | Compromise cost of the proposed STP | Interval cost of the proposed STP | Predetermined confidence levels | Compromise cost of the proposed STP | Interval cost of the proposed STP |
|---|---|---|---|---|---|
| 0.05 | 580.3313 | [568.6000, 592.0625] | 0.45 | 618.8028 | [597.4000, 641.5625] |
| 0.10 | 583.6625 | [572.2000, 595.1250] | 0.50 | 622.3237 | [601.0000, 644.6250] |
| 0.15 | 586.9937 | [575.8000, 598.1875] | 0.55 | 644.1437 | [640.6000, 47.6875] |
| 0.20 | 590.3250 | [579.4000, 601.2500] | 0.60 | 648.4750 | [646.2000, 650.7500] |
| 0.25 | 593.6562 | [583.0000, 604.3125] | 0.65 | 652.8062 | [651.8000, 653.8125] |
| 0.30 | 596.9875 | [586.6000, 607.3750] | 0.70 | 657.1375 | [656.8750, 657.4000] |
| 0.35 | 600.3188 | [590.2000, 610.4375] | 0.75 | 661.4688 | [659.9375, 663.0000] |
| 0.40 | 603.6500 | [593.8000, 613.5000] | 0.80 | 665.8000 | [663.0000, 668.6000] |

*For the model (2.25), the predetermined confidence levels of $\alpha_1^U$, $\alpha_1^L$, $\alpha_2^U$,$\alpha_2^L$, $\beta_1^U$,$\beta_1^L$,$\beta_2^U$ ,$\beta_2^L$ ,$\gamma_1^U$,$\gamma_1^L$,$\gamma_2^U$ and $\gamma_2^L$ are considered the same, and increased by 0.05 for each execution of the model (2.25).

### 2.6.2 SPP with Interval Type-2 Fuzzy Parameters

We consider a WCDN $G = (V_G, E_G)$ with 11 vertices and 21 edges. The cost associated with each edge $e_{ij}$ is considered as imprecise and represented by IT2FV. The source $(s)$ and the sink $(t)$ vertices of $G$ are denoted by $v_1$ and $v_{11}$, respectively. The network, $G$ is depicted in Fig. 2.2 and the associated edge costs of $G$ are shown in Table 2.7.

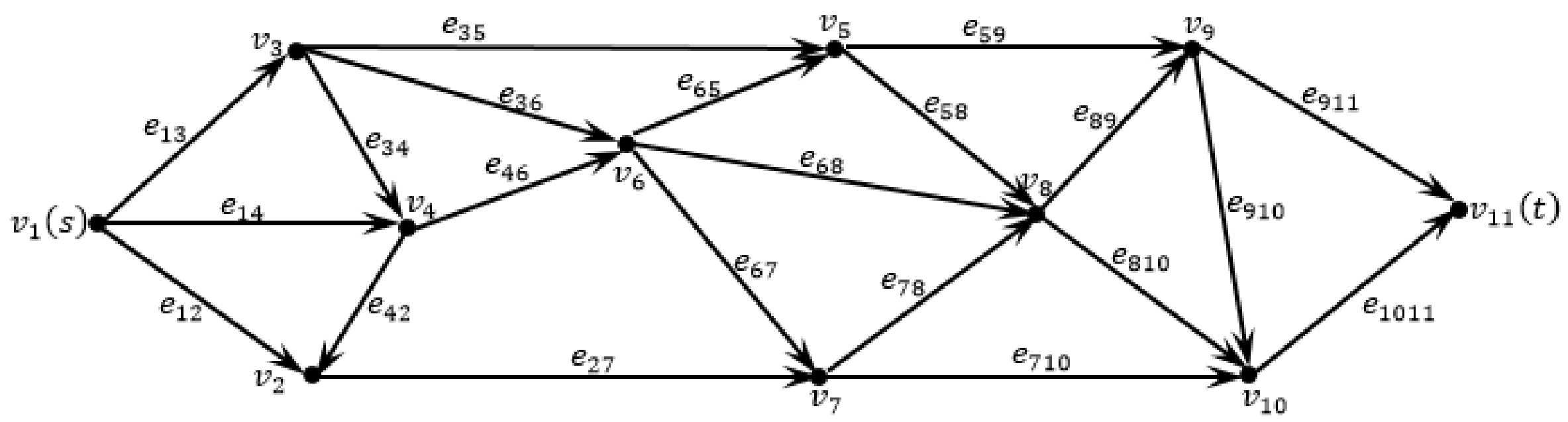

Figure 2.2 A WCDN $G$ with interval type-2 fuzzy edge costs

Table 2.7 Interval type-2 fuzzy edge costs of the WCDN $G$

| Edges | Edge cost $\tilde{c}_{ij} = \left(c_{ij}^U, c_{ij}^L\right)$ | Edge | Edge cost $\tilde{c}_{ij} = \left(c_{ij}^U, c_{ij}^L\right)$ |
|---|---|---|---|
| $e_{12}$ | $((67,69.2,70,70.9; 1.0), (68, 68.9,69.5,70; 0.9))$ | $e_{65}$ | $((71,72,74,76; 1.0), (72,72.9, 73, 75; 0.9))$ |
| $e_{13}$ | $((65,66,67.2,68.5; 1.0), (66,66.9,67,67.5; 0.9))$ | $e_{67}$ | $((70,70.9,74,75; 1.0), (71,72.5,73,74.2; 0.9))$ |
| $e_{14}$ | $((61,63.6,65,66; 0.1), (62, 63.8,63.9,64.2; 0.9))$ | $e_{68}$ | $((60,63,65,66; 1.0), (60.9,62,63,64; 0.9))$ |
| $e_{27}$ | $((59,59.7,61.8,62.6; 1.0), (60,60.9,61.2,61.6; 0.9))$ | $e_{78}$ | $((62,63,64.8,66; 1.0), (63,63.6,64.2,65; 0.9))$ |
| $e_{34}$ | $((63,63.8,64.9,66; 1.0), (63.7,64,64.7,65; 0.9))$ | $e_{710}$ | $((63,64,68,69.6; 1.0), (64, 65, 66, 68; 0.9))$ |
| $e_{35}$ | $((70,71.9,72.3,73.9; 1.0), (71.6,72.3,73.2,73; 0.9))$ | $e_{89}$ | $((60,63,66,68; 1.0), (61,63.6,64,67; 0.9))$ |
| $e_{36}$ | $((65,66.3,67.6,68.7; 1.0), (66,66.9,67,67.9; 0.9))$ | $e_{810}$ | $((56,59,62,65; 1.0), (58,59.7,61.3,64; 0.9))$ |
| $e_{42}$ | $((61,62.8,65,66; 1.0), (62,63.5,64,65.3; 0.9))$ | $e_{910}$ | $((55,57,59,61; 1.0), (56,57.3,58,60; 0.9))$ |
| $e_{46}$ | $((60,62,65,66.3; 1.0), (61,63,64,65.2; 0.9))$ | $e_{911}$ | $((62,63,67,69; 1.0), (63,64,66,67; 0.9))$ |
| $e_{58}$ | $((69,69.2,71,73; 1.0), (70,70.2,70.9,72; 0.9))$ | $e_{1011}$ | $((70,72,73.9,75; 1.0), (71,72.3,73,74; 0.9))$ |
| $e_{59}$ | $((70,70.2,73.2,73.9; 1.0), (71.2,71.9,72.5,73; 0.9))$ | $--$ | $--$ |

In order to solve the corresponding CCM of the STP of $G$, we consider the predetermined confidence levels $\delta^U$ and $\delta^L$ as 0.9, for both the UMFs and the LMFs of the associated IT2FVs. Then, the crisp equivalent models (2.28) and (2.30) are solved, respectively for the CCM with UMFs and LMFs of the proposed SPP. Consequently, the optimized cost between $s$ and $t$ of $G$ are generated as $277.2200$ and $273.6000$ by solving the models (2.28) and (2.30), respectively. Thereafter, to determine the compromise cost of the SPP for $G$, within the interval $[273.6000, 277.2200]$, model (2.31) is solved. Table 2.8 shows the corresponding results generated by the models (2.28), (2.30) and (2.31) for $G$. Here, we observe that the shortest paths generated by the models (2.28), (2.30) and (2.31), are the same when solved for $G$, however the corresponding path costs are different for the corresponding models of $G$.

To verify the results generated by the models shown in Table 2.8, we calculate $F_{\tilde{c}_{ij}^U}$ and $F_{\tilde{c}_{ij}^L}$ for every $e_{ij}$ in $G$, as defined in Step 2 and Step 5, respectively in Section 2.5.1. Here, for both $F_{\tilde{c}_{ij}^U}$ and $F_{\tilde{c}_{ij}^L}$, the predetermined confidence levels ($\delta^U$ and $\delta^L$) are set as 0.9. Moreover, in order to determine a compromise value ($F^{comp}$) between $F_{\tilde{c}_{ij}^U}$ and $F_{\tilde{c}_{ij}^L}$, we calculate $\left(F_{\tilde{c}_{ij}^U} + F_{\tilde{c}_{ij}^L}\right)/2$ for each $e_{ij}$. Subsequently, we consider three independent executions of Dijkstra's algorithm (E.W. Dijkstra 1959) by considering the corresponding values of $F_{\tilde{c}_{ij}^U}$ , $F_{\tilde{c}_{ij}^L}$ and $F^{comp}$ of $G$ in each run. Accordingly, for the network $G$, we observe that the results generated by the Dijkstra's algorithm with respect to $F_{\tilde{c}_{ij}^U}$, $F_{\tilde{c}_{ij}^L}$ and $F^{comp}$ matches exactly with the corresponding results of the models (2.28), (2.30) and (2.31), as reported in Table 2.8.

Table 2.8 Results generated by models (2.28), (2.30) and (2.31) for the WCDN $G$

| Model | Minimum path cost | Shortest Path |
|---|---|---|
| (2.28) | 277.2200 | $e_{12} - e_{27} - e_{710} - e_{1011}$ |
| (2.30) | 273.6000 | $e_{12} - e_{27} - e_{710} - e_{1011}$ |
| (2.31) | 275.4100 | $e_{12} - e_{27} - e_{710} - e_{1011}$ |

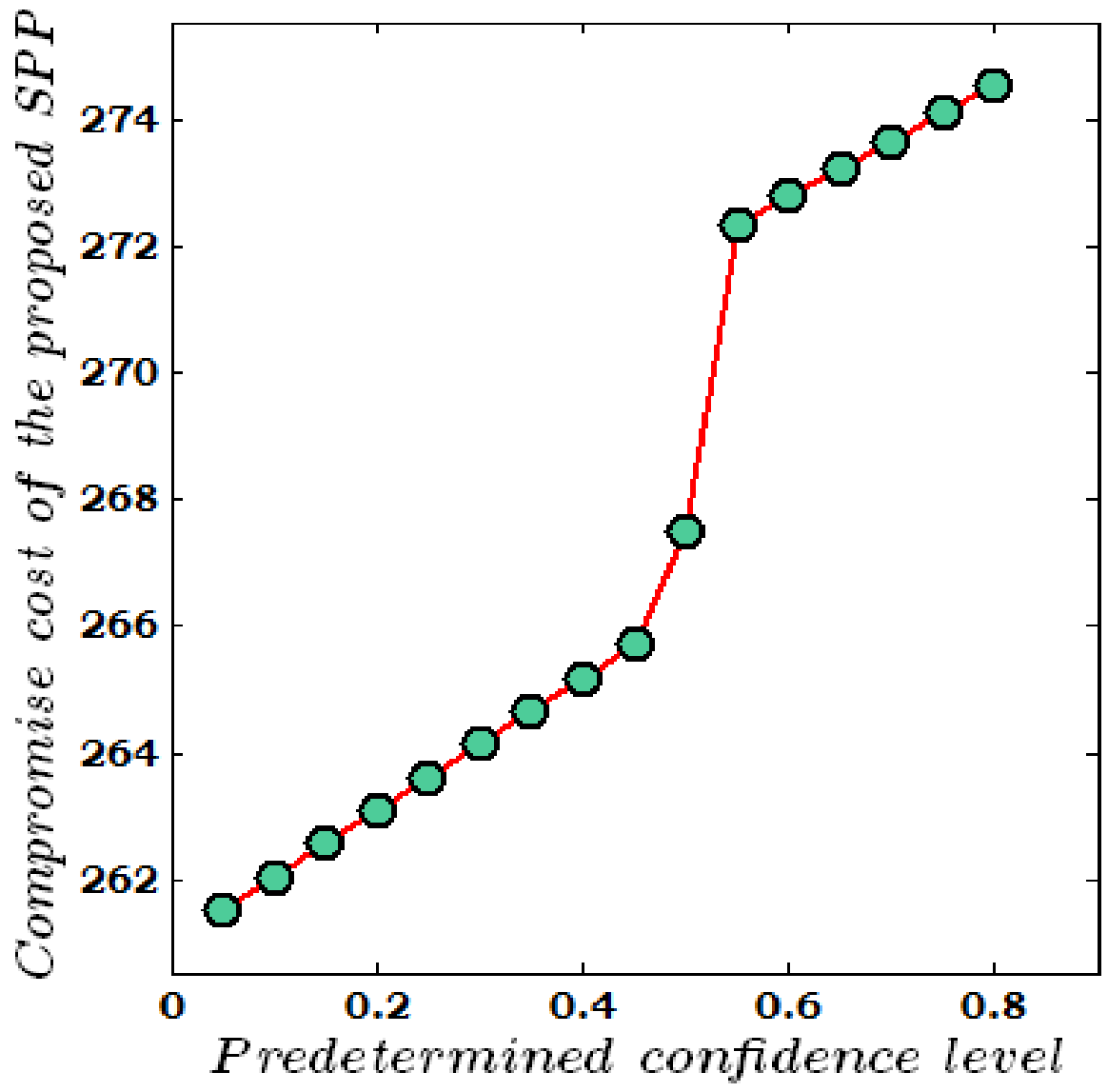

Figure 2.3 Compromise cost of the WCDN $G$ generated by solving model (2.31) at different confidence levels

The compromise solutions of the model (2.31) for $G$, at different predetermined confidence levels is also determined, which are presented in Table 2.9. Here, we observe that the compromise minimum cost of $G$ gradually increases as the confidence levels are increased. These compromise solutions are displayed graphically at different confidence levels in Fig. 2.3.

Table 2.9 Compromise cost of the proposed SPP for the WCDN $G$ at different credibility levels

| Predetermined confidence levels | Compromise cost of the proposed SPP | Interval cost of the proposed SPP | Predetermined confidence levels | Compromise cost of the proposed SPP | Interval cost of the proposed SPP |
|---|---|---|---|---|---|
| 0.05 | 261.5228 | [259.5900, 263.4556] | 0.45 | 265.7050 | [264.3100, 267.1000] |
| 0.10 | 262.0456 | [259.5900, 263.9111] | 0.50 | 267.5167 | [264.9000, 270.1333] |
| 0.15 | 262.5683 | [260.7700, 264.3667] | 0.55 | 272.3533 | [270.5667, 274.1400] |
| 0.20 | 263.0911 | [261.3600, 264.8222] | 0.60 | 272.7900 | [271.0000, 274.5800] |
| 0.25 | 263.6139 | [261.9500, 265.2778] | 0.65 | 273.2267 | [271.4333, 275.0200] |
| 0.30 | 264.1367 | [262.5400, 265.7333] | 0.70 | 273.6633 | [271.8667, 275.4600] |
| 0.35 | 264.6594 | [263.1300, 266.1889] | 0.75 | 274.1000 | [272.3000, 275.9000] |
| 0.40 | 265.1822 | [266.6444, 263.7200] | 0.80 | 274.5367 | [272.7333, 276.3400] |

*For the model (2.31), the predetermined confidence levels of $\delta^U$ and $\delta^L$ are considered the same, and increased by 0.05 for each execution of the model (2.31).

### 2.6.3 MSTP with Interval Type-2 Fuzzy Parameters

We consider a WCUN $H = (V_H, E_H)$ which is a complete graph of 6 vertices with 15 edges. Here, each edge $e_{ij}$ is associated with an imprecise cost represented by IT2FV. The network, $H$ and its associated edge costs are respectively, shown in Fig. 2.4 and Table 2.10.

To solve the chance-constrained model of MSTP, the corresponding predetermined confidence levels $\delta^U$ and $\delta^L$ for both UMFs and LMFs of the IT2FVs associated with the edges of $H$ are set to 0.9. Subsequently, we solve the models (2.33) and (2.34), to determine the MSTs and their optimized costs with respect to the UMFs and LMFs of the IT2FVs, associated with $H$. Accordingly, the generated solutions of the models (2.33) and (2.34) are respectively, obtained as 346.6400 and 344.1000. With the purpose of generating a compromise solution within the interval $[344.1000, 346.6400]$, model (2.35) is solved for $H$. Table 2.11 displays the results generated by the models (2.33), (2.34) and (2.35). Here, it is observed that the edges selected to generate the minimum spanning trees are the same However, the optimized cost of the MSTP is different when solved by each of the models.

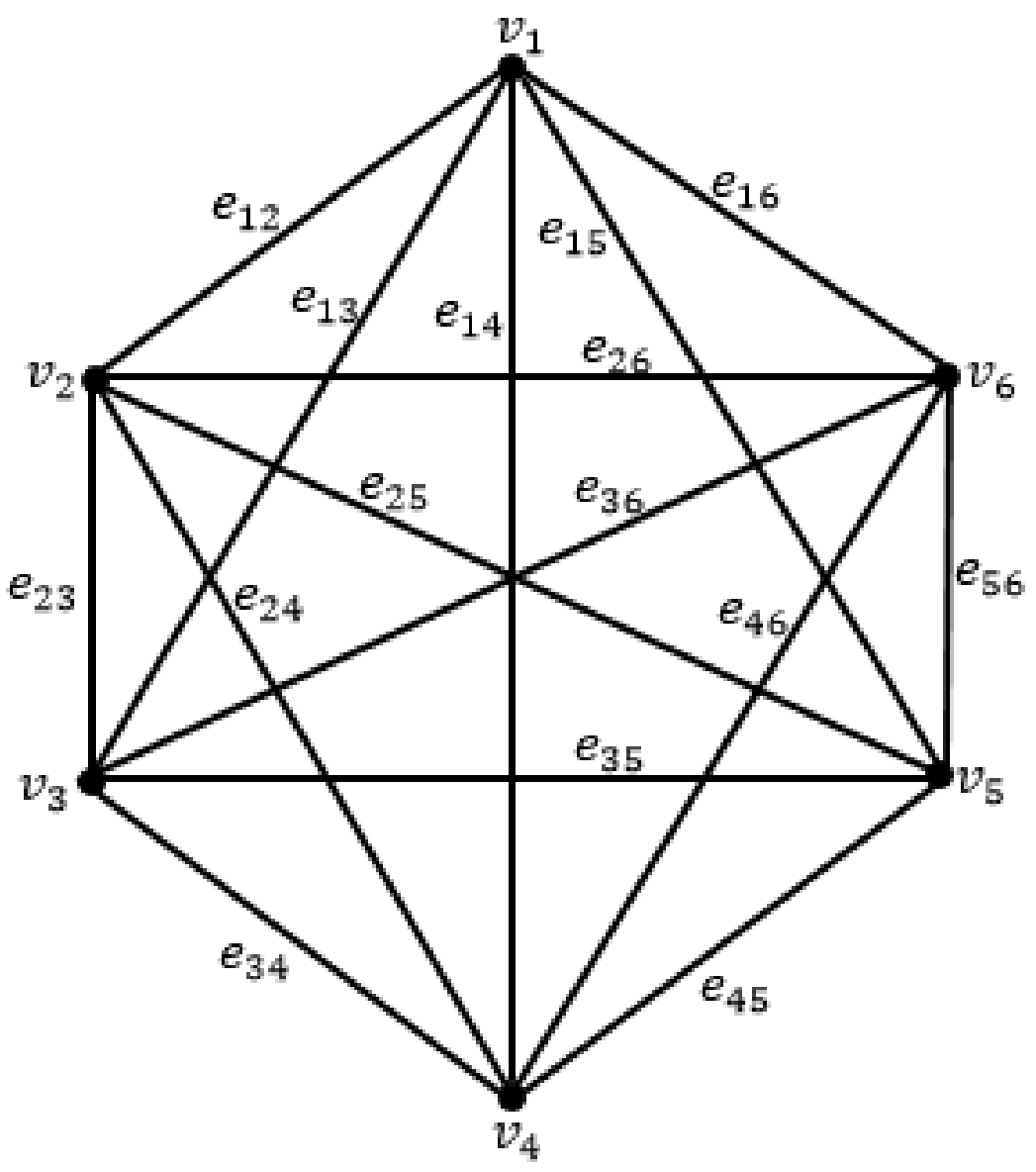

Figure 2.4 A WCUN $H$ with interval type-2 fuzzy edge costs

Table 2.10 Interval type-2 fuzzy edge costs of the WCUN $H$

| Edge | Edge cost $\tilde{d}_{ij} = (d_{ij}^U, d_{ij}^L)$ | Edge | Edge cost $\tilde{d}_{ij} = (d_{ij}^U, d_{ij}^L)$ |
|---|---|---|---|
| $e_{12}$ | $((68,69.5,71.2,74;\ 1.0), (69,69.7,70.2,73;\ 0.9))$ | $e_{26}$ | $((68,69.2,72,74; 1.0; ), (68.2,69.7,71,73; 0.9))$ |
| $e_{13}$ | $((72,73.2,74.9,76.8; 1.0), (72.7,73.9,74.2,75.5; 0.9))$ | $e_{34}$ | $((69,70.1,73,75; 1.0), (70,71.3,72,73.9; 0.9))$ |
| $e_{14}$ | $((74,75.2,76.9,78; 1.0), (75,76.2,76,77; 0.9))$ | $e_{35}$ | $((70,71.2,73,74.9; 1.0), (71,72,72.5,73.2; 0.9))$ |
| $e_{15}$ | $((75,76,77.7,79; 1.0), (76,76.2,76.7,78; 0.9))$ | $e_{36}$ | $((71,72.9,73,76; 1.0), (72,73,73.9,75; 0.9))$ |
| $e_{16}$ | $((68,69,71,72; 1.0), (69.7,69.7,70.7,71.2; 0.9))$ | $e_{45}$ | $((72,73,74.2,76; 1.0), (73,73.7,74,75; 0.9))$ |
| $e_{23}$ | $((65,67,69,70; 1.0), (66,67.2,68.1,69; 0.9))$ | $e_{46}$ | $((68,69,74,75; 1.0), (69,70.9,73,74; 0.9))$ |
| $e_{24}$ | $((59,60,63,65; 1.0), (61,62,62.7,64; 0.9))$ | $e_{56}$ | $((76,78,81,83; 1.0), (77,78.2,80,81.6; 0.9))$ |
| $e_{25}$ | $((63,65,66.2,67.2; 1.0), (64,65.2,66,66.9; 0.9))$ | $--$ | $--$ |

In order to verify the results of Table 2.11, we determine $F_{\tilde{a}_{ij}^U}$ and $F_{\tilde{a}_{ij}^L}$ (as mentioned in the models (2.33) and (2.34)) corresponding to each edge $e_{ij}$ of $H$. Moreover, to determine a compromise value ($F^{comp}$) between $F_{\tilde{a}_{ij}^U}$ and $F_{\tilde{a}_{ij}^L}$, we evaluate $\left(F_{\tilde{a}_{ij}^U} + F_{\tilde{a}_{ij}^L}\right)/2$ for every $e_{ij}$. Consequently, three independent runs of Kruskal's algorithm (Jr. J.B. Kruskal 1956) are executed, by considering the corresponding values of $F_{\tilde{a}_{ij}^U}$ , $F_{\tilde{a}_{ij}^L}$ and $F^{comp}$ of $H$ in each run. Considering Table 2.11, we observe that the results generated by Kruskal's algorithm with respect to $F_{\tilde{a}_{ij}^U}$, $F_{\tilde{a}_{ij}^L}$ and $F^{comp}$ matches exactly with the corresponding results of the models (2.33), (2.34) and (2.35).

Similar to Table 2.9, in Table 2.12, we report the compromise solutions of the model (2.35) for $H$, at different predetermined confidence levels. Here, we have observed that the solution generated by the model (2.35) progressively increases while increasing the confidence levels. Fig. 2.5, graphically depicts those compromise solutions for $H$ at different confidence levels.

## 2.7 Conclusion

In this chapter, we have presented a chance-constrained method to solve LPP whose associated parameters are considered as IT2FVs. Here, the CCM model is developed

Table 2.11 Results generated by models (2.33), (2.34) and (2.35) for the WCUN $H$

| Model | Minimum cost of the MSTP | Minimum spanning tree |
|---|---|---|
| (2.33) | 346.6400 | $e_{12}, e_{16}, e_{23}, e_{24}, e_{25}$ |
| (2.34) | 344.1000 | $e_{12}, e_{16}, e_{23}, e_{24}, e_{25}$ |
| (2.35) | 345.3700 | $e_{12}, e_{16}, e_{23}, e_{24}, e_{25}$ |

using the generalized credibility measure, and UMFs and LMFs of IT2FVs. The CCM is applied to solve three network optimization problems: (i) STP, (ii) SPP and (iii) MSTP whose associated parameters are characterized as IT2FVs. Particularly, for STP, the constraints corresponding to availability, demand and conveyance capacity are considered as imprecise and represented by IT2FV. For SPP and MSTP, the coefficients of the objective function which correspond to the edge costs of a network are imprecise and essentially represented by IT2FV. The crisp equivalents of those network problems are eventually solved at some predetermined confidence levels. Moreover, the results generated by the crisp equivalent models of SPP and MSTP are also verified with the results obtained by Dijkstra's and Kruskal's algorithms, respectively. It is worth pointing out that the proposed CCM models for STP, SPP and MSTP are formulated if

Table 2.12 Compromise cost of the proposed MSTP for the WCUN $H$ at different credibility levels

| Predetermined confidence levels | Compromise cost of the proposed MST | Interval cost of the proposed MST | Predetermined confidence levels | Compromise cost of the proposed MST | Interval cost of the proposed MST |
|---|---|---|---|---|---|
| 0.05 | 326.2961 | [323.7000, 328.8222] | 0.45 | 331.6400 | [329.3000, 333.8000] |
| 0.10 | 326.9922 | [324.4000, 329.4444] | 0.50 | 334.4556 | [ 330.0000, 338.4111] |
| 0.15 | 327.6883 | [325.1000, 330.0667] | 0.55 | 340.1511 | [339.1222, 341.1800] |
| 0.20 | 328.3844 | [ 325.8000, 330.6889] | 0.60 | 340.8967 | [339.8333, 341.9600] |
| 0.25 | 329.0806 | [326.5000, 331.3111] | 0.65 | 341.6422 | [340.5444, 342.7400] |
| 0.30 | 329.7433 | [327.2000, 331.9333] | 0.70 | 342.3878 | [341.2556, 343.5200] |
| 0.35 | 330.3756 | [327.9000, 332.5556] | 0.75 | 343.1333 | [341.9667, 344.3000] |
| 0.40 | 331.0078 | [328.6000, 333.1778] | 0.80 | 343.8789 | [342.6778, 345.0800] |

*For the model (2.35), the predetermined confidence levels of $\delta^U$and $\delta^L$ are considered the same, and increased by 0.05 for each execution of the model (2.35).

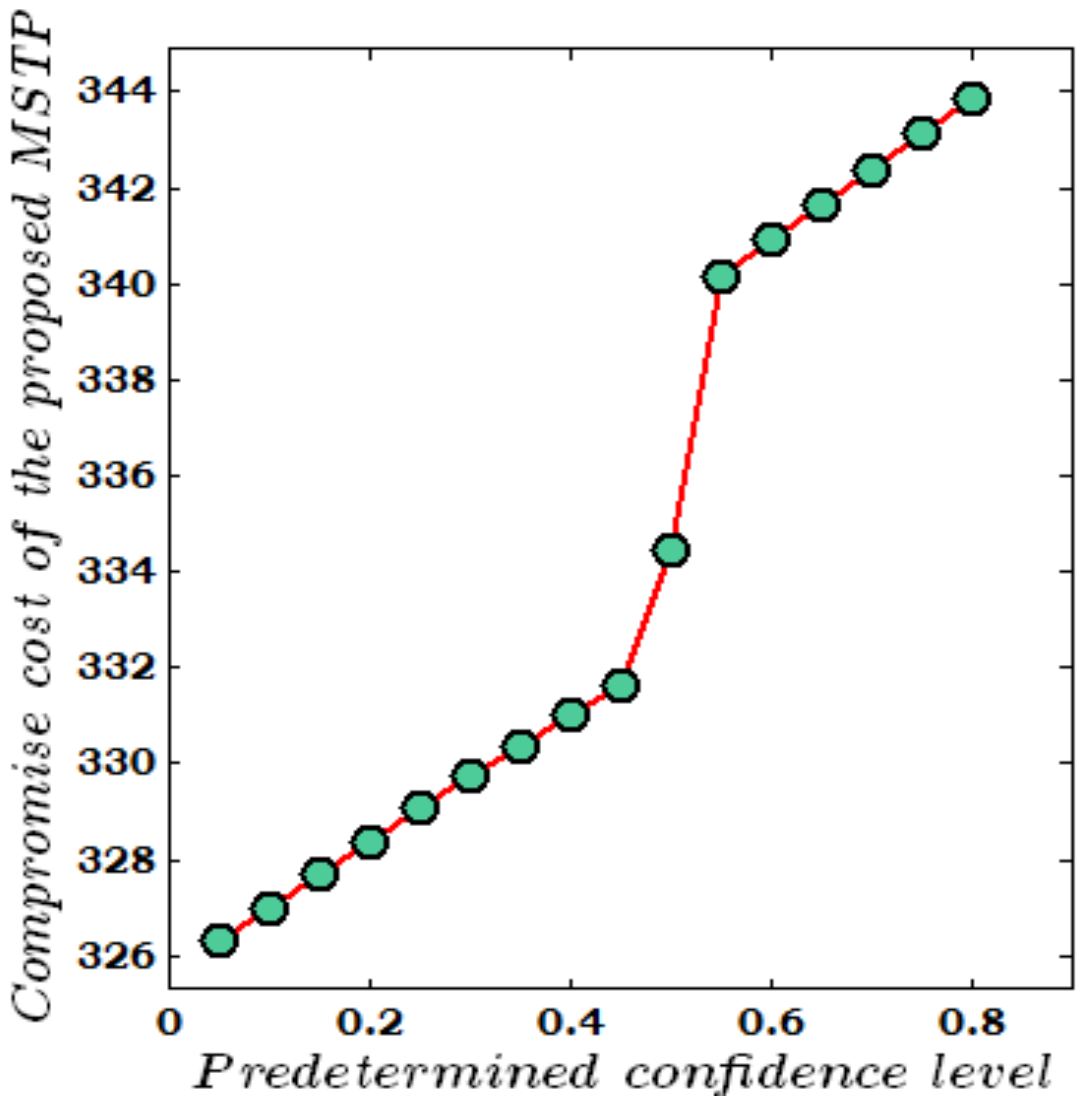

Figure 2.5 Compromise cost of the WCUN $H$ generated by solving model (2.35) at different confidence levels

a decision maker aims to optimize a critical value of an objective function subject to some constraints, which should hold at some chance levels (predetermined confidence levels) (B. Liu 2002). Therefore, as there exists no such uncertain (interval type-2 fuzzy) CCMs of STP, SPP and MSTP, we cannot compare the results of the proposed models in the process of decision making. The usage of the model is solely dependent on the decision maker’s preference.

In future, we would like to apply the proposed CCM to different optimization problems including portfolio selection, inventory management and supply chain management. In addition, the proposed CCM model can be modified for the generalized T2FVs to solve the different single and multi-objective optimization problems.

# Chapter 3

# Genetic Algorithm with Varying Population for Random Fuzzy Maximum Flow Problem

# Chapter 3

# Genetic Algorithm with Varying Population for Random Fuzzy Maximum Flow Problem

## 3.1 Introduction

Maximum flow problem (MFP) is one of the popular network optimization problems which has captivated many researchers due to its wide application domains including VLSI design, electrical power distribution system, logistic network, transportation network, computer network, etc. The genesis of MFP has its connection with the Soviet railway project which was underway in Eastern Europe (Harris and Ross 1955). The first algorithm to solve the maximum flow problem is presented by Ford and Fulkerson (1956). Subsequently, the MFP is further investigated by Ford and Fulkerson (1962). Thereafter, various researchers (E.A. Dinic 1970; Edmonds and Karp 1972; A.V. Karzanov 1974; R.E. Tarjan 1984; Goldberg and Tarjan 1986; Ahuja and Orlin 1989; Hachtel and Somenzi 1997) put forward different efficient maximum flow algorithms. Of late, Gu and Xu (2007) have extended the maximal flow algorithms of H.N. Gabow (1985) and A.V. Karzanov (1974) on symbolic graphs. In the study of Munakata and Hashier (1993), the same problem is addressed by using a genetic algorithm, where each solution is represented by a flow matrix. Later, a priority based GA is presented by Gen et al. (2008) to solve the MFP.

In practice, the network capacities are usually uncertain in nature. There are many articles in the literature, where the network capacities are considered as random or fuzzy variables. A random network is first studied by Frank and Hakimi (1965), where the stochastic phenomenon, involved in the capacities of a communication network, are modelled. Later, Nawathe and Rao (1980) determined the maximum flow of a probabilistic communication network. Besides, contributions of different authors (R. Hall 1986; Fu and Rilett 1998) on random networks also exist in the literature.

In various decision making problems, quite frequently we encounter situations, where the input parameters are imprecise due to the ambiguity of expert's judgment. Fuzzy mathematical programming becomes one of the alternatives to model such imprecision, which has been effectively used by some researchers (B. Bilgen 2010; Sadjadi et al. 2010) to solve various decision making problems. Considering the fuzzy maximum flow problem, Kim and Roush (1982) introduced MFP with fuzzy capacities for the first time. Since then, Chanas and Kołodzijczyk (1982, 1984, 1986) revisited the problem. Later, Hernandes et al. (2007b) modify the algorithm of Ford and Fulkerson (1962) by considering the capacities of the network as fuzzy parameters.

The decision makers in certain decision making problems may need to take decisions, where both randomness and fuzziness co-exist. In this context, the fuzzy random variable is introduced by H. Kwakernaak (1978), and developed by Puri and Ralescu (1985), Liu and Liu (2000), etc. Here, the parameters associated with a fuzzy variable are assumed to be random. However, B. Liu (2001) incorporated a new scenario, where the parameters associated with a random variable are fuzzy. For example, the price of an item may depend on its demand which can be regarded as a random variable following the normal distribution with expected value $\mu$ and standard deviation $\sigma$. Such a conclusion may be drawn, since there can be historical evidence about the demand for an item varying due to the seasonal impact. Again, $\mu$ can be imprecise, since the demand of an item depends on the expense related to its manufacturing and transportation which possibly fluctuate over a period of time. Therefore, in such situations, the price of an item can be represented by a random fuzzy variable. In the literature, some studies (Liu and Liu 2003; X. Huang 2007b; Katagiri et al. 2012) on the application of random fuzzy variables on different optimization problems can be observed.

In spite of several contributions on MFP, there are some gaps in the literature which are listed below.

- A maximum flow problem for a network in random fuzzy environment is yet to be studied in the literature.
- A varying population GA with an improved lifetime allocation strategy (iLAS) and uncertain crossover probability is not yet considered as a solution methodology for MFP.

To alleviate these lacunas, in our study we have formulated the expected value model (EVM) and chance-constrained model (CCM) for MFP in random fuzzy environment. The crisp transformations of each of these models are solved using the proposed varying population genetic algorithm with indeterminate crossover (VPGAwIC). The crossover probability used in VPGAwIC is indeterminate and defined as a function of parent's linguistic age under the framework of uncertainty theory (B. Liu 2007). The results of the proposed VPGAwIC are compared with Gen et al. (2008).

The remaining part of the chapter is organized as follows. The EVM and the CCM for random fuzzy maximum flow problem are formulated in Section 3.2. The corresponding crisp equivalent models are proposed in Section 3.3. The proposed VPGAwIC is presented in Section 3.4. Results of VPGAwIC for both crisp and random fuzzy instances of MFP are provided in Section 3.5. Finally, the study is concluded in Section 3.6.

## 3.2 Maximum Flow Problem

The maximum flow problem (MFP) deals with maximization of a feasible flow from source node to sink node in a network. The MFP with random fuzzy capacities is discussed in the following subsections.

### 3.2.1 Problem Formulation

Let $G = (V_G, E_G)$ be the directed connected network, where $V_G = \{v_1, v_2, \dots, v_n\}$ and $E_G$ are respectively, the sets of $n$ nodes (vertices) and $m$ edges of $G$. A pair of nodes $\langle v_i, v_j \rangle$, represents an edge $e_{ij}$ directed from $v_i$ to $v_j$. Each edge $e_{ij}$ is associated with a non-negative finite quantity, known as the capacity of $e_{ij}$. The capacity of an edge $e_{ij}$ indicates the maximum amount of an item that can flow from $v_i$ to $v_j$. In our study, we formulate random fuzzy maximum flow problem (RFMFP) in (3.1) by considering the capacity $\tilde{\tilde{\zeta}}_{C_{ij}}$ of each $e_{ij}$ as a random fuzzy variable.

$$\begin{cases} Maximize\, Z = \bar{f} \\ subject\ to \\ \sum_{j=1}^{n} x_{ij} - \sum_{k=1}^{n} x_{ki} = \begin{cases} \bar{f} & ; if\ i = 1 \\ 0 & ; if\ i = (2,3,\dots,n-1) \\ -\bar{f} & ; if\ i = n \end{cases} \\ 0 \le x_{ij} \le \tilde{\tilde{\zeta}}_{C_{ij}}, i,j = 1,2,\dots,n, \end{cases} \tag{3.1}$$

where the first constraint determines that for a particular node $v_i$, the total inward flow must be equal to the total outward flow except for the source node $v_1$ and sink node $v_n$. There are only outward flow from $v_1$ and inward flow to $v_n$. The decision variable $x_{ij}$ in the second constraint of (3.1) indicates that the flow through each edge $e_{ij}$ should not exceed the capacity $\tilde{\tilde{\zeta}}_{C_{ij}}$. A feasible flow of a network satisfies both the constraints of (3.1), and has a net flow of $\overline{f}$.

### 3.2.2 Expected Value Model (EVM)

In EVM (Liu and Liu 2002), the expected value of the objective function is optimized with respect to a set of expected constraints. The EVM of a constrained optimization problem with uncertain parameters is defined below in (3.2).

$$\begin{cases} \underset{x \in \Re^p}{Maximize}\ E\left[f(\tilde{\tilde{\zeta}}, x)\right] \\ subject\ to \\ E\left[g_j\left(\tilde{\tilde{\zeta}}, x\right)\right] \le 0, j = 1,2,\dots,l, \end{cases} \tag{3.2}$$

where $x = (x_1, x_2, \dots, x_p)$ is a $p$-dimensional decision vector, $\tilde{\tilde{\zeta}} = \left(\tilde{\tilde{\zeta}}_1, \tilde{\tilde{\zeta}}_2, \dots, \tilde{\tilde{\zeta}}_q\right)$ is a $q$-dimensional random fuzzy vector, $E\left[f\left(\tilde{\tilde{\zeta}}, x\right)\right]$ is the expected value of the objective function $f(\tilde{\bar{\zeta}}, x)$ and $E\left[g_j\left(\tilde{\bar{\zeta}}, x\right) \leq 0\right]$ is the expected value of the $j^{th}$ constraint $g_j\left(\tilde{\bar{\zeta}}, x\right) \leq 0, j = 1,2, \ldots, l$.

**Definition 3.2.1** (Liu and Liu 2002): A solution $x$ is said to be a feasible solution of (3.2), if $E\left[g_j\left(\tilde{\bar{\zeta}}, x\right)\right] \leq 0, j = 1,2, \ldots, l$.

**Definition 3.2.2** (Liu and Liu 2002): A feasible solution $x^*$ in (3.2) is said to be an optimal solution, if $E\left[f\left(\tilde{\bar{\zeta}}, x^*\right)\right] \geq E\left[f(\tilde{\bar{\zeta}}, x)\right]$ for any feasible solution $x$.

In the proposed RFMFP, defined in (3.1), if the flow of the network is maximized with respect to the expected values of the random fuzzy capacities of the network, then the corresponding EVM of (3.1) is defined as follows.

$$\begin{cases} Maximize\, \bar{f} \\ subject\ to \\ \sum_{j=1}^{n} x_{ij} - \sum_{k=1}^{n} x_{ki} = \begin{cases} \bar{f} & ; if\ i = 1 \\ 0 & ; if\ i = (2,3, \ldots, n-1) \\ -\bar{f} & ; if\ i = n \end{cases} \\ 0 \leq E\left[x_{ij} \leq \tilde{\bar{\zeta}}_{C_{ij}}\right], i, j = 1,2, \ldots, n. \end{cases} \tag{3.3}$$

In (3.3), the decision variable $x_{ij}$ in the second constraint determines the flow through $e_{ij}$ which will not exceed its corresponding expected capacity value.

### 3.2.3 Chance-constrained Model (CCM)

Chance-constrained model, introduced by Charnes and Cooper (1959), is another approach to model uncertainty in optimization problems. A CCM corresponding to an optimization problem with uncertain parameters is formulated in (3.4).

$$\begin{cases} Maximize\, \overline{Z} \\ subject\ to \\ Ch\left\{f\left(\tilde{\bar{\zeta}}, x\right) \geq \overline{Z}\right\}(\eta) \geq \tau \\ Ch\left\{g_j\left(\tilde{\bar{\zeta}}, x\right) \leq 0\right\}(\alpha_j) \geq \beta_j,\ j = 1,2, \ldots, l, \end{cases} \tag{3.4}$$

where $\eta$, $\tau$, $\alpha_j$, $\beta_j$ are the predetermined confidence levels, $j = 1,2, \ldots, l$.

If the associated uncertain parameters in (3.4) are random fuzzy variable, then following Definition 1.3.16 and Remark 1.3.6, definitions 3.2.3 and 3.2.4 are defined as follow.

**Definition 3.2.3**: A solution $x$ is said to be a feasible solution of (3.4) if and only if the credibility measure of a fuzzy event $\left\{\theta \in \Theta \mid Cr\left\{g_j\left(\tilde{\bar{\zeta}},\ x\right) \leq 0\right\} \geq \beta_j\right\}$ is at least $\alpha_j$ for $j = 1,2, \ldots, l$.

**Definition 3.2.4**: A feasible solution $x^*$ is said to be optimal to (3.4) if for any feasible solution $x$, $\left[Cr\left\{\theta \in \Theta | \, Pr\left\{f\left(\tilde{\tilde{\zeta}},\ x^*\right) \geq \overline{Z}\right\} \geq \tau\right\} \geq \eta\right] \geq \left[Cr\left\{\theta \in \Theta | \, Pr\left\{f\left(\tilde{\tilde{\zeta}},\ x\right) \geq \overline{Z}\right\} \geq \tau\right\} \geq \eta\right]$.

If a decision maker prefers to optimize RFMFP under chance constraints, then the corresponding CCM of the model (3.1) is formulated below in (3.5).

$$\begin{cases} Maximize\ \bar{f} \\ subject\ to \\ \sum_{j=1}^{n} x_{ij} - \sum_{k=1}^{n} x_{ki} = \begin{cases} \bar{f}\ ; if\ i = 1 \\ 0\ ; if\ i = (2,3, \dots, n-1) \\ -\bar{f}\ ; if\ i = n \end{cases} \\ 0 \leq Cr\left\{\theta \in \Theta | \, Pr\left\{x_{ij} \leq \tilde{\tilde{\zeta}}_{C_{ij}}(\theta)\right\} \geq \alpha_{ij}\right\} \geq \beta_{ij},\ i,j = 1,2, \dots, n, \end{cases} \tag{3.5}$$

where $\alpha_{ij}$ and $\beta_{ij}$ are the predetermined confidence levels.

## 3.3 Crisp Equivalents of EVM and CCM

To solve the corresponding EVM and CCM for RFMFP, formulated above in (3.3) and (3.5), we need to transform the models to their respective crisp equivalents. These crisp equivalent models are presented below in theorems 3.3.1 and 3.3.2.

**Theorem 3.3.1**: Let $\tilde{\tilde{\zeta}}_{C_{ij}}$ be a random fuzzy variable which represents the capacity of an edge $e_{ij}$ for a network. Further, if each $\tilde{\tilde{\zeta}}_{C_{ij}}$ is characterized as $\tilde{\tilde{\zeta}}_{C_{ij}}(\theta) = \mathcal{N}\left(\tilde{\zeta}_{C_{ij}}(\theta), \sigma_{C_{ij}}\right)$, such that the mean $\tilde{\zeta}_{C_{ij}}(\theta)$ is a fuzzy variable with variance $\sigma_{C_{ij}}$. Then the crisp equivalent of (3.3) can be defined as follows.

$$\begin{cases} Maximize\ Z = \bar{f} \\ \quad subject\ to \\ \quad \sum_{j=1}^{n} x_{ij} - \sum_{k=1}^{n} x_{ki} = \begin{cases} \bar{f} & ; if\ i = 1 \\ 0 & ; if\ i = (2,3, \dots, n-1) \\ -\bar{f} & ; if\ i = n \end{cases} \\ \quad 0 \leq x_{ij} \leq E\left[\tilde{\tilde{\zeta}}_{C_{ij}}(\theta)\right], i,j = 1,2, \dots, n, \end{cases} \tag{3.6}$$

where

(a) $E\left[\tilde{\tilde{\zeta}}_{C_{ij}}(\theta)\right] = \frac{1}{4}\left(\rho_{ij}^1 + \rho_{ij}^2 + \rho_{ij}^3 + \rho_{ij}^4\right)$, if $\tilde{\zeta}_{C_{ij}}(\theta)$ is a trapezoidal fuzzy variable which is characterized by $\mathcal{Tr}\left[\rho_{ij}^1, \rho_{ij}^2, \rho_{ij}^3, \rho_{ij}^4\right]$, $\rho_{ij}^1 < \rho_{ij}^2 \leq \rho_{ij}^3 < \rho_{ij}^4$ with inverse credibility distribution

$$\Psi_{\tilde{\zeta}_{C_{ij}}(\theta)}^{-1}\left(\beta_{ij}\right) = \begin{cases} \left(1 - 2\beta_{ij}\right)\rho_{ij}^1 + 2\beta_{ij}\rho_{ij}^2 & ; if\ \beta_{ij} \leq 0.5 \\ \left(2 - 2\beta_{ij}\right)\rho_{ij}^3 + \left(2\beta_{ij} - 1\right)\rho_{ij}^4 & ; if\ \beta_{ij} > 0.5. \end{cases}$$

(b) $E\left[\tilde{\tilde{\zeta}}_{C_{ij}}(\theta)\right] = \rho_{ij}$, if $\tilde{\zeta}_{C_{ij}}(\theta)$ is a Gaussian fuzzy variable which is characterized by $\mathcal{N}(\rho_{ij}, \delta_{ij})$ $\rho_{ij}, \delta_{ij} > 0$ with the following inverse credibility distribution.

$$\Psi^{-1}_{\tilde{\zeta}_{C_{ij}}(\theta)}(\beta_{ij}) = \begin{cases} \rho_{ij} - \delta_{ij}\sqrt{-\ln(2\beta_{ij})} & ; if\ \beta_{ij} \le 0.5 \\ \rho_{ij} + \delta_{ij}\sqrt{-\ln(2 - 2\beta_{ij})} & ; if\ \beta_{ij} > 0.5. \end{cases}$$

**Proof**: From the linearity of the expected value operator, model (3.3) can be represented as

$$\begin{cases} Maximize\ Z = \bar{f} \\ \quad subject\ to \\ \quad \sum_{j=1}^{n} x_{ij} - \sum_{k=1}^{n} x_{ki} = \begin{cases} \bar{f} & ; if\ i = 1 \\ 0 & ; if\ i = (2,3,\dots,n-1) \\ -\bar{f} & ; if\ i = n \end{cases} \\ \quad 0 \le x_{ij} \le E\left[\tilde{\tilde{\zeta}}_{C_{ij}}(\theta)\right], i,j = 1,2,\dots,n. \end{cases}$$

Let $\tilde{\tilde{\zeta}}_{C_{ij}}$ be a random fuzzy variable, then for any $\theta \in \Theta$, $\tilde{\tilde{\zeta}}_{C_{ij}}(\theta) = \mathcal{N}\left(\tilde{\zeta}_{C_{ij}}(\theta), \sigma_{C_{ij}}\right)$ is a random variable when normally distributed around its mean value $\tilde{\zeta}_{C_{ij}}(\theta)$ with crisp variance $\sigma_{C_{ij}}$. Then

$$E\left[\tilde{\tilde{\zeta}}_{C_{ij}}(\theta)\right] = \int_0^{+\infty} Pr\left\{\tilde{\tilde{\zeta}}_{C_{ij}}(\theta) \ge x_{ij}\right\} dx_{ij} - \int_{-\infty}^{0} Pr\left\{\tilde{\tilde{\zeta}}_{C_{ij}}(\theta) \le x_{ij}\right\} dx_{ij} = \tilde{\zeta}_{C_{ij}}(\theta).$$

*For part (a)*

If $\tilde{\zeta}_{C_{ij}}(\theta)$ is considered as a trapezoidal fuzzy variable and represented as $\tilde{\zeta}_{C_{ij}}(\theta) = \mathcal{T}r[\rho^1_{ij}, \rho^2_{ij}, \rho^3_{ij}, \rho^4_{ij}]$, $\rho^1_{ij} < \rho^2_{ij} \le \rho^3_{ij} < \rho^4_{ij}$, then from (1.7), the inverse credibility distribution of $\tilde{\zeta}_{C_{ij}}(\theta)$ can be defined as

$$\Psi^{-1}_{\tilde{\zeta}_{C_{ij}}(\theta)}(\beta_{ij}) = \begin{cases} (1 - 2\beta_{ij})\rho^1_{ij} + 2\beta_{ij}\rho^2_{ij} & ; if\ \beta_{ij} \le 0.5 \\ (2 - 2\beta_{ij})\rho^3_{ij} + (2\beta_{ij} - 1)\rho^4_{ij} & ; if\ \beta_{ij} > 0.5. \end{cases}$$

Subsequently, following (1.1) and (1.2) we get

$$\begin{aligned} E\left[\tilde{\zeta}_{C_{ij}}(\theta)\right] &= \int_0^{+\infty} Cr\left\{\tilde{\zeta}_{C_{ij}}(\theta) \ge x_{ij}\right\} dx_{ij} - \int_{-\infty}^{0} Cr\left\{\tilde{\zeta}_{C_{ij}}(\theta) \le x_{ij}\right\} dx_{ij} \\ &= \int_0^{0.5}\left[(1 - 2\beta_{ij})\rho^1_{ij} + 2\beta_{ij}\rho^2_{ij}\right] d\beta_{ij} + \int_{0.5}^{1}\left[(2 - 2\beta_{ij})\rho^3_{ij} + (2\beta_{ij} - 1)\rho^4_{ij}\right] d\beta_{ij} \\ &= \frac{1}{4}\left(\rho^1_{ij} + \rho^2_{ij} + \rho^3_{ij} + \rho^4_{ij}\right). \end{aligned}$$

*For part (b)*

Here, we consider $\tilde{\zeta}_{C_{ij}}(\theta)$ as a Gaussian fuzzy variable which is expressed as $\mathcal{N}(\rho_{ij}, \delta_{ij})$, $\rho_{ij}, \delta_{ij} > 0$. Accordingly, by following (1.11), the inverse credibility distribution of $\tilde{\zeta}_{C_{ij}}(\theta)$ is represented as

$$\Psi^{-1}_{\tilde{\zeta}_{C_{ij}}(\theta)}(\beta_{ij}) = \begin{cases} \rho_{ij} - \delta_{ij}\sqrt{-\ln(2\beta_{ij})} & ; if\ \beta_{ij} \le 0.5 \\ \rho_{ij} + \delta_{ij}\sqrt{-\ln(2-2\beta_{ij})} & ; if\ \beta_{ij} > 0.5. \end{cases}$$

Therefore, from (1.1) and (1.2) we have

$$E\left[\tilde{\zeta}_{C_{ij}}(\theta)\right] = \int_0^{+\infty} Cr\left\{\tilde{\zeta}_{C_{ij}}(\theta) \ge x_{ij}\right\} dx_{ij} - \int_{-\infty}^{0} Cr\left\{\tilde{\zeta}_{C_{ij}}(\theta) \le x_{ij}\right\} dx_{ij}$$

$$= \int_0^{0.5}\left[\rho_{ij} - \delta_{ij}\sqrt{-\ln(2\beta_{ij})}\right] d\beta_{ij} + \int_{0.5}^{1}\left[\rho_{ij} + \delta_{ij}\sqrt{-\ln(2-2\beta_{ij})}\right] d\beta_{ij} = \rho_{ij}.$$

Hence from Theorem 3.3.1, if the mean of the random fuzzy variable is a trapezoidal fuzzy variable, then (3.7) represents the crisp equivalent of model (3.3) and when the mean of the random fuzzy variable is Gaussian fuzzy variable, the corresponding crisp equivalent of model (3.3) is represented in (3.8).

$$\begin{cases} Maximize\ Z = \bar{f} \\ subject\ to \\ \sum_{j=1}^{n} x_{ij} - \sum_{k=1}^{n} x_{ki} = \begin{cases} \bar{f} & ; if\ i = 1 \\ 0 & ; if\ i = (2,3,\dots,n-1) \\ -\bar{f} & ; if\ i = n \end{cases} \\ 0 \le x_{ij} \le \frac{1}{4}\left(\rho^1_{ij} + \rho^2_{ij} + \rho^3_{ij} + \rho^4_{ij}\right),\ i,j = 1,2,\dots,n. \end{cases} \tag{3.7}$$

$$\begin{cases} Maximize\ Z = \bar{f} \\ subject\ to \\ \sum_{j=1}^{n} x_{ij} - \sum_{k=1}^{n} x_{ki} = \begin{cases} \bar{f} & ; if\ i = 1 \\ 0 & ; if\ i = (2,3,\dots,n-1) \\ -\bar{f} & ; if\ i = n \end{cases} \\ 0 \le x_{ij} \le \rho_{ij},\ i,j = 1,2,\dots,n. \end{cases} \tag{3.8}$$

**Theorem 3.3.2**: Let, $\tilde{\tilde{\zeta}}_{C_{ij}}$ be a random fuzzy variable which represents the corresponding capacity of an edge, $e_{ij}$ for a network. If, $\tilde{\tilde{\zeta}}_{C_{ij}}$ is characterized by $\tilde{\tilde{\zeta}}_{C_{ij}}(\theta) = \mathcal{N}\left(\tilde{\zeta}_{C_{ij}}(\theta), \sigma_{C_{ij}}\right)$ such that $\tilde{\tilde{\zeta}}_{C_{ij}}(\theta)$ is normally distributed around fuzzy mean $\tilde{\zeta}_{C_{ij}}(\theta)$ and crisp variance $\sigma_{C_{ij}}$. Then, if

$\tilde{\zeta}_{C_{ij}}(\theta) = \mathcal{T}r\left[\rho^1_{ij}, \rho^2_{ij}, \rho^3_{ij}, \rho^4_{ij}\right]$, $\rho^1_{ij} < \rho^2_{ij} \le \rho^3_{ij} < \rho^4_{ij}$ is a trapezoidal fuzzy variable which is characterized by the credibility distribution presented below following eq. (1.8).

$$Cr\{\tilde{\zeta}_{C_{ij}}(\theta) \geq x_{ij}\} = \begin{cases} 1 & ; if\ x_{ij} < \rho_{ij}^1 \\ \frac{2\rho_{ij}^2 - \rho_{ij}^1 - x_{ij}}{2\left(\rho_{ij}^2 - \rho_{ij}^1\right)} & ; \rho_{ij}^1 \leq x_{ij} < \rho_{ij}^2 \\ \frac{1}{2} & ; if\ \rho_{ij}^2 \leq x_{ij} < \rho_{ij}^3 \\ \frac{\rho_{ij}^4 - x_{ij}}{2\left(\rho_{ij}^4 - \rho_{ij}^3\right)} & ; if\ \rho_{ij}^3 \leq x_{ij} < \rho_{ij}^4 \\ 0 & ; if\ x_{ij} \geq \rho_{ij}^4. \end{cases}$$

Then considering the model (3.5), we have

$Cr\left\{\theta \in \Theta \middle| Pr\left\{x_{ij} \leq \tilde{\tilde{\zeta}}_{C_{ij}}(\theta)\right\} \geq \alpha_{ij}\right\} \geq \beta_{ij}$ if and only if

$$x_{ij} \leq \begin{cases} 2\rho_{ij}^2 - \rho_{ij}^1 - 2\beta_{ij}\left(\rho_{ij}^2 - \rho_{ij}^1\right) + \wp & ; if\ \rho_{ij}^1 \leq \mathcal{B} \leq \rho_{ij}^2,\ \beta_{ij} > 0.5 \\ \rho_{ij}^4 - 2\beta_{ij}\left(\rho_{ij}^4 - \rho_{ij}^3\right) + \wp & ; if\ \rho_{ij}^3 \leq \mathcal{B} \leq \rho_{ij}^4,\ \beta_{ij} \leq 0.5. \end{cases}$$

$\tilde{\zeta}_{C_{ij}}(\theta) = \mathcal{N}\left(\rho_{ij}, \delta_{ij}\right)$, $\rho_{ij}, \delta_{ij} > 0$ is a Gaussian fuzzy variable which is characterized by the credibility distribution shown below following eq. (1.12).

$$Cr\{\tilde{\zeta}_{C_{ij}}(\theta) \geq x_{ij}\} = \begin{cases} 1 - \frac{1}{2} exp\left\{-\left(\frac{\rho_{ij} - x_{ij}}{\delta_{ij}}\right)^2\right\} & ; if\ x_{ij} \leq \rho_{ij} \\ \frac{1}{2} exp\left\{-\left(\frac{x_{ij} - \rho_{ij}}{\delta_{ij}}\right)^2\right\} & ; if\ x_{ij} > \rho_{ij}. \end{cases}$$

Then, for the model (3.5) we have, $Cr\left\{\theta \in \Theta \middle| Pr\left\{x_{ij} \leq \tilde{\tilde{\zeta}}_{C_{ij}}(\theta)\right\} \geq \alpha_{ij}\right\} \geq \beta_{ij}$ if and only if

$$x_{ij} \leq \begin{cases} \rho_{ij} - \delta_{ij}\sqrt{-ln\left(2 - 2\beta_{ij}\right)} + \wp & ; if\ \mathcal{B} \leq \rho_{ij}, \beta_{ij} > 0.5 \\ \rho_{ij} + \delta_{ij}\sqrt{-ln\left(2\beta_{ij}\right)} + \wp & ; if\ \mathcal{B} > \rho_{ij}, \beta_{ij} \leq 0.5. \end{cases}$$

Here, $\mathcal{B} = x_{ij} - \varphi^{-1}\left(1 - \alpha_{ij}\right)\sqrt{\sigma_{C_{ij}}}$, $\wp = \varphi^{-1}\left(1 - \alpha_{ij}\right)\sqrt{\sigma_{C_{ij}}}$, $\varphi$ is a standardized normal distribution and $\alpha_{ij}, \beta_{ij} \in [0,1]$ are the predetermined chance levels.

**Proof:** Let for any $\theta \in \Theta$, $\tilde{\tilde{\zeta}}_{C_{ij}}(\theta)$ is a random variable and normally distributed over fuzzy mean $\tilde{\zeta}_{C_{ij}}(\theta)$ with variance $\sigma_{C_{ij}}$. Then

$$Pr\left\{x_{ij} \leq \tilde{\tilde{\zeta}}_{C_{ij}}(\theta)\right\} \geq \alpha_{ij} \Leftrightarrow Pr\left\{\tilde{\tilde{\zeta}}_{C_{ij}}(\theta) \geq x_{ij}\right\} \geq \alpha_{ij} \Leftrightarrow Pr\left\{\frac{\tilde{\tilde{\zeta}}_{C_{ij}}(\theta) - \tilde{\zeta}_{C_{ij}}(\theta)}{\sqrt{\sigma_{C_{ij}}}} \geq \frac{x_{ij} - \tilde{\zeta}_{C_{ij}}(\theta)}{\sqrt{\sigma_{C_{ij}}}}\right\} \geq \alpha_{ij}$$

$$\Leftrightarrow 1 - Pr\left\{\frac{\tilde{\tilde{\zeta}}_{C_{ij}}(\theta) - \tilde{\zeta}_{C_{ij}}(\theta)}{\sqrt{\sigma_{C_{ij}}}} \le \frac{x_{ij} - \tilde{\zeta}_{C_{ij}}(\theta)}{\sqrt{\sigma_{C_{ij}}}}\right\} \ge \alpha_{ij} \Leftrightarrow Pr\left\{\frac{\tilde{\tilde{\zeta}}_{C_{ij}}(\theta) - \tilde{\zeta}_{C_{ij}}(\theta)}{\sqrt{\sigma_{C_{ij}}}} \le \frac{x_{ij} - \tilde{\zeta}_{C_{ij}}(\theta)}{\sqrt{\sigma_{C_{ij}}}}\right\} \le$$

$$1 - \alpha_{ij} \Leftrightarrow \varphi\left[\frac{x_{ij} - \tilde{\zeta}_{C_{ij}}(\theta)}{\sqrt{\sigma_{C_{ij}}}}\right] \le 1 - \alpha_{ij} \Leftrightarrow \tilde{\zeta}_{C_{ij}}(\theta) + \varphi^{-1}(1 - \alpha_{ij})\sqrt{\sigma_{C_{ij}}} \ge x_{ij}.$$

Consequently, for given chance levels, $\alpha_{ij},\ \beta_{ij} \in [0,1]$, we get

$Cr\left\{\theta \in \Theta | \tilde{\zeta}_{C_{ij}}(\theta) + \varphi^{-1}(1 - \alpha_{ij})\sqrt{\sigma_{C_{ij}}} \ge x_{ij}\right\} \ge \beta_{ij}$ from $Cr\left\{\theta \in \Theta | Pr\left\{x_{ij} \le \tilde{\tilde{\zeta}}_{C_{ij}}(\theta)\right\} \ge \alpha_{ij}\right\} \ge \beta_{ij}$ .

*For part (a)*

We consider $\tilde{\zeta}_{C_{ij}}(\theta)$ as a trapezoidal fuzzy variable, so that $\tilde{\zeta}_{C_{ij}}(\theta) = \mathcal{Tr}[\rho_{ij}^1, \rho_{ij}^2, \rho_{ij}^3, \rho_{ij}^4]$, $\rho_{ij}^1 < \rho_{ij}^2 \le \rho_{ij}^3 < \rho_{ij}^4$. Therefore,

$$Cr\left\{\theta \in \Theta | \tilde{\zeta}_{C_{ij}}(\theta) + \varphi^{-1}(1 - \alpha_{ij})\sqrt{\sigma_{C_{ij}}} \ge \tilde{\tilde{\zeta}}_{C_{ij}}(\theta) \ge x_{ij}\right\} \ge \beta_{ij}$$

$$\Leftrightarrow \beta_{ij} \le \begin{cases} 1 & ; if\ \mathcal{B} < \rho_{ij}^1 \\ \frac{2\rho_{ij}^2 - \rho_{ij}^1 - \mathcal{B}}{2(\rho_{ij}^2 - \rho_{ij}^1)} & ; if\ \rho_{ij}^1 \le \mathcal{B} < \rho_{ij}^2 \\ \frac{1}{2} & ; if\ \rho_{ij}^2 \le \mathcal{B} < \rho_{ij}^3 \\ \frac{\rho_{ij}^4 - \mathcal{B}}{2(\rho_{ij}^4 - \rho_{ij}^3)} & ; if\ \rho_{ij}^3 \le \mathcal{B} < \rho_{ij}^4 \\ 0 & ; if\ \mathcal{B} \ge \rho_{ij}^4. \end{cases}$$

Moreover, from Lemma 1.3.1 it implies

$$x_{ij} \le \begin{cases} 2\rho_{ij}^2 - \rho_{ij}^1 - 2\beta_{ij}(\rho_{ij}^2 - \rho_{ij}^1) + \wp & ; if\ \rho_{ij}^1 \le \mathcal{B} \le \rho_{ij}^2,\ \beta_{ij} > 0.5 \\ \rho_{ij}^4 - 2\beta_{ij}(\rho_{ij}^4 - \rho_{ij}^3) + \wp & ; if\ \rho_{ij}^3 \le \mathcal{B} \le \rho_{ij}^4,\ \beta_{ij} \le 0.5. \end{cases}$$

*For part (b)*

We consider $\tilde{\zeta}_{C_{ij}}(\theta)$ as a Gaussian fuzzy variable, so that $\tilde{\zeta}_{C_{ij}}(\theta) = \mathcal{N}(\rho_{ij}, \delta_{ij})$, $\rho_{ij}, \delta_{ij} > 0$, then we get

$$Cr\left\{\theta \in \Theta | \tilde{\zeta}_{C_{ij}}(\theta) + \varphi^{-1}(1 - \alpha_{ij})\sqrt{\sigma_{C_{ij}}} \ge \tilde{\tilde{\zeta}}_{C_{ij}}(\theta) \ge x_{ij}\right\} \ge \beta_{ij}$$

$$\Leftrightarrow \beta_{ij} \le \begin{cases} 1 - \frac{1}{2} exp\left\{-\left(\frac{\rho_{ij} - \mathcal{B}}{\delta_{ij}}\right)^2\right\} & ; if\ \mathcal{B} \le \rho_{ij} \\ \frac{1}{2} exp\left\{-\left(\frac{\mathcal{B} - \rho_{ij}}{\delta_{ij}}\right)^2\right\} & ; if\ \mathcal{B} > \rho_{ij}. \end{cases}$$

Again, from Lemma 1.3.2 it follows

$$x_{ij} \leq \begin{cases} \rho_{ij} - \delta_{ij}\sqrt{-ln\left(2 - 2\beta_{ij}\right)} + \wp & ; if\ \mathcal{B} \leq \rho_{ij}, \beta_{ij} > 0.5 \\ \rho_{ij} + \delta_{ij}\sqrt{-ln\left(2\beta_{ij}\right)} + \wp & ; if\ \mathcal{B} > \rho_{ij}, \beta_{ij} \leq 0.5, \end{cases}$$

where $\mathcal{B} = x_{ij} - \varphi^{-1}\left(1 - \alpha_{ij}\right)\sqrt{\sigma_{C_{ij}}}$ and $\wp = \varphi^{-1}\left(1 - \alpha_{ij}\right)\sqrt{\sigma_{C_{ij}}}$.

Hence, it follows from Theorem 3.3.2, the crisp equivalent model of (3.5) becomes (3.9) if the $\tilde{\zeta}_{C_{ij}}(\theta)$ is a trapezoidal fuzzy variable and when $\tilde{\zeta}_{C_{ij}}(\theta)$ is a Gaussian fuzzy variable, (3.10) becomes the crisp equivalent model of (3.5).

$$\begin{cases} Maximize\ Z = \bar{f} \\ subject\ to \\ \sum_{j=1}^{n} x_{ij} - \sum_{k=1}^{n} x_{ki} = \begin{cases} \bar{f} & ; \text{if } i = 1 \\ 0 & ; \text{if } i = (2,3, \dots, n-1) \\ -\bar{f} & ; \text{if } i = n \end{cases} \\ 0 \leq x_{ij} \leq \mathcal{S}_{ij},\ i,j = 1,2, \dots, n, \end{cases} \tag{3.9}$$

where $\mathcal{S}_{ij} = \begin{cases} 2\rho_{ij}^2 - \rho_{ij}^1 - 2\beta_{ij}\left(\rho_{ij}^2 - \rho_{ij}^1\right) + \wp & ; if\ \rho_{ij}^1 \leq \mathcal{B} \leq \rho_{ij}^2,\ \beta_{ij} > 0.5 \\ \rho_{ij}^4 - 2\beta_{ij}\left(\rho_{ij}^4 - \rho_{ij}^3\right) + \wp & ; if\ \rho_{ij}^3 \leq \mathcal{B} \leq \rho_{ij}^4,\ \beta_{ij} \leq 0.5. \end{cases}$

$$\begin{cases} Maximize\ Z = \bar{f} \\ subject\ to \\ \sum_{j=1}^{n} x_{ij} - \sum_{k=1}^{n} x_{ki} = \begin{cases} \bar{f} & ; if\ i = 1 \\ 0 & ; if\ i = (2,3, \dots, n-1) \\ -\bar{f} & ; if\ i = n \end{cases} \\ 0 \leq x_{ij} \leq \mathcal{U}_{ij},\ i,j = 1,2, \dots, n, \end{cases} \tag{3.10}$$

where $\mathcal{U}_{ij} = \begin{cases} \rho_{ij} - \delta_{ij}\sqrt{-ln\left(2 - 2\beta_{ij}\right)} + \wp & ; if\ \mathcal{B} \leq \rho_{ij}, \beta_{ij} > 0.5 \\ \rho_{ij} + \delta_{ij}\sqrt{-ln\left(2\beta_{ij}\right)} + \wp & ; if\ \mathcal{B} > \rho_{ij}, \beta_{ij} \leq 0.5. \end{cases}$

## 3.4 Proposed Genetic Algorithm

Motivated by the studies of Last and Eyal (2005), and Gen et al. (2008), we have proposed a varying population genetic algorithm with indeterminate crossover (VPGAwIC) to solve MFP. Here, each chromosome is assigned a lifetime and an age. We have defined a new strategy to determine the lifetime of a chromosome. The age of each chromosome increases with generations. A chromosome is considered as dead and so is discarded from the population if its age is greater than its lifetime. The ages of chromosomes are categorized as one of the linguistic variables *Young*, *Middle* and *Old*, which follow uncertainty distributions (B. Liu 2007). The crossover probability in the VPGAwIC is defined as a function of parent's age. In the proposed GA, the offspring are combined with their parent population for the next generation. The number of offspring generated from a parent population is determined by the reproduction ratio

(Z. Michalewicz 1992). In a particular generation, the total population size does not exceed a predetermined threshold.

To describe the proposed algorithm VPGAwIC, the following notations are used.

$T$: Generation count

$age$: Age of a chromosome in terms to $T$

$lifetime$: Maximum age, a chromosome can stay alive in terms of $T$

$Pop(T)$: Population at generation $T$

$C(T)$: Offspring population generated by $Pop(T)$

$Dead(T)$: Set of chromosomes from $Pop(T)$, whose $age > lifetime$

$g_{max}$: Maximum generation number

$\rho$: Reproduction ratio

$\xi_{p_c}$: Uncertain crossover probability

$p_m$: Mutation probability

$PopSizeTh$: Maximum number of chromosomes in a generation

$Maxfit_T$: Maximum fitness of population at $T^{th}$ generation

$Minfit_T$: Minimum fitness of population at $T^{th}$ generation

$Avgfit_T$: Average fitness value of population at $T^{th}$ generation

$MaxSameCount$: Maximum allowed number of consecutive generations with same $Avgfit_T$

$C_P$: Permissible flow of a network path $P$

$c_{ij}$: Maximum capacity of an edge

$F_R$: Fitness value of a chromosome $R$ (maximum flow of a chromosome)

$CM\{cm_{ij},\ i,j = 1,2,\dots,n\}$: Capacity matrix of a network, where each element $cm_{ij}$ represents the capacity of an edge $e_{ij}$

$L_i$: Set of nodes adjacent to node $v_i$

$Visited_m$: Determine whether a node $v_m$ is included in a path $P$,

i.e., $Visited_m = \begin{cases} true \text{ ; if visited, i.e., } v_m \in P \\ false \text{ ; if not visited, i.e., } v_m \notin P \end{cases}$

$s$: Source node of the network

$t$: Sink node of the network

$Len$: Current length of the generated path $P$

### 3.4.1 Chromosome Representation

In a population $Pop(T)$, $i^{th}$ chromosome $X_i$ represents a sequence of $n$ nodes of the network. For every chromosome, the genes are assigned priority, and a path can be constructed based on the priorities of the genes using priority-based encoding (Lin and Gen 2007).

### 3.4.2 Population Initialization

A set of $N$ chromosomes are randomly generated which constitute initial population, $Pop(0)$. In this study, the size of each chromosome is fixed and represents all nodes of the network.

### 3.4.3 Fitness Evaluation

The fitness is the maximum flow of every newly generated chromosome which is evaluated by the *Overall-Flow* algorithm. The algorithms, *Single-Path-Growth* and *Overall-Flow*, have similar functionalities compare to *One-path growth* and *Overall-path growth* algorithms, respectively as proposed by Gen et al. (2008). However, the algorithm, *Single-Path-Growth* is modified using a cycle breaking mechanism while constructing a path of the network. Whereas, in the *Overall-Flow* algorithm, once the maximum flow of a newly generated chromosome is determined, its lifetime is computed using the proposed improved lifetime allocation strategy (iLAS).

We have presented both the *Single-Path-Growth* and *Overall-Flow* algorithms as follows. The *Single-Path-Growth* algorithm generates a path from source node to sink node. Here, when a vertex $m$ is included in a path, its corresponding flag, i.e., $Visited_m$ is set to $true$. In this process, if there exists a cycle in a network, a vertex with its corresponding flag set as $true$, is never included while computing the path.

**Algorithm 1**: *Single-Path-Growth*
**Input**: A chromosome $R$, number of nodes $n$, $L_i$, $t$
**Output**: A network path $P$ from $s$

***Single-Path-Growth*()**
**begin**
Initialize $Visited_m = false$ for all $m \in (1, n)$, where $s = 1$, $t = n$
$i = 1$, $Len = 1, P[Len] = i, Visited_1 = true$ // 1 is the source node
**while** $L_1 \neq \{\emptyset\}$
**while** $L_i = \{\emptyset\}$
**if** $Len = 1$ $then$
**return** $P = \{\emptyset\}$ // if the node is $s$
**end if**
$r = P[Len]$ //remove the last node from the path $P$
$Len = Len - 1$

$R[r] = 0$ // set the priority of the removed node to 0 so that it //is not selected again if an intermediate node has no outgoing //edges from it.
**if** $Len > 1$ **then** $i = P[Len]$
**else**
**return** $P$ // if the node is $t$
**end if**
**end while**
Set $j' = 0$
$j'$ = node with highest priority among non-visited nodes of $L_i$
**if** $j' > 0$ $and$ $R[j'] \neq 0$
$Len = Len + 1$, $P[Len] = j'$, $i = j'$ // Add $i$ to path and //make $i$ to $j'$
$Visited_{j'} = true$
**else**
$r = P[Len]$, $Len = Len - 1$
$Visited_r = true$
**if** $Len > 1$ $then$
$i = P[Len]$
**end if**
**end if**
**if** $P[Len] = t$ **then**
**return** $P$
**end if**
**end while**
**end**

**Algorithm 2**: *Overall-Flow*
**Input**: Capacity matrix $(CM = \{cm_{ij},\ i,j = 1,2,\dots,n\})$, adjacent list of $i^{th}$ node $L_i$, population $Pop(T)$, source node $s$, destination node $t$
**Output**: Determine $F_R$, $lifetime$ of each chromosome $R$ in $Pop(T)$

***Overall-Flow*()**
**begin**
**for** each chromosome $R$ in $Pop(T)$
**if** $F_R = -1$ then
Set $age$=0
**while** $L_1 \neq \{\emptyset\}$ // $L_1$ represents the source node $s$.
$P$= *Single-Path-Growth* () // Evaluate path $P$ and determine the //sink node $a$ of $P$
**if** $a = t$ **then**
$C_p$= **min**$\{cm_{ij} | e_{ij} \in P\}$
$F_R = F_R + C_p$

$$cm_{ij} = \begin{cases} cm_{ij} - C_p & ; if\ e_{ij} \in P \\ cm_{ij} & ; otherwise \end{cases}$$

**if** $cm_{ij} = 0$ **then**

$L_i = L_i \backslash v_j$

**end if**

**else**

$L_i = L_i \backslash a\ \forall\ i \in \{1,2, \dots, n\}$ // update the set of nodes in $L_i$

**end if**

**end while**

**end if**

**end for**

Compute $Maxfit_T$, $Avgfit_T$ and $Minfit_T$ for $Pop(T)$

Compute $lifetime$ to every newly generated chromosome by iLAS.

**end**

### 3.4.4 Age and Lifetime of Chromosome

In a generation, the age of a newly born chromosome $X_i$ is initialized to zero and its lifetime is computed following the proposed improved lifetime allocation strategy (iLAS), which is a modified version of the bilinear allocation strategy (bLAS) (Z. Michalewicz 1992). The iLAS is defined below in (3.11).

$$lifetime(X_i) = \begin{cases} minLT + \kappa \dfrac{F_{X_i} - Minfit_T}{Avgfit_T - Minfit_T} & ; if\ Avgfit_T \geq F_{X_i} \\ \dfrac{1}{2}(minLT + maxLT) + \kappa \dfrac{F_{X_i} + Maxfit_T - (2 * Avgfit_T)}{2 * (Maxfit_T - Avgfit_T)} & ; otherwise, \end{cases} \tag{3.11}$$

where

$F_{X_i}$: Fitness value of a chromosome (solution) of $X_i$

$maxLT$: Maximum lifetime of a chromosome

$minLT$: Minimum lifetime of a chromosome

$$\kappa = \frac{1}{2}(maxLT - minLT) \tag{3.12}$$

The age of a chromosome increases by one with generation. Once the age of a chromosome exceeds its lifetime, the chromosome is considered as dead. A chromosome is $Young$, $Middle$ or $Old$ according to its age. In this study, the linguistic variables associated with the age of chromosomes are uncertain variables (B. Liu 2007), whose uncertainty distributions are depicted in Fig. 3.1. Here, $Young$ and $Old$ are represented by linear uncertainty distribution, whereas $Middle$ is represented by zigzag uncertainty distribution (cf. definitions 1.3.29 and 1.3.30 of Section 1.3.11).

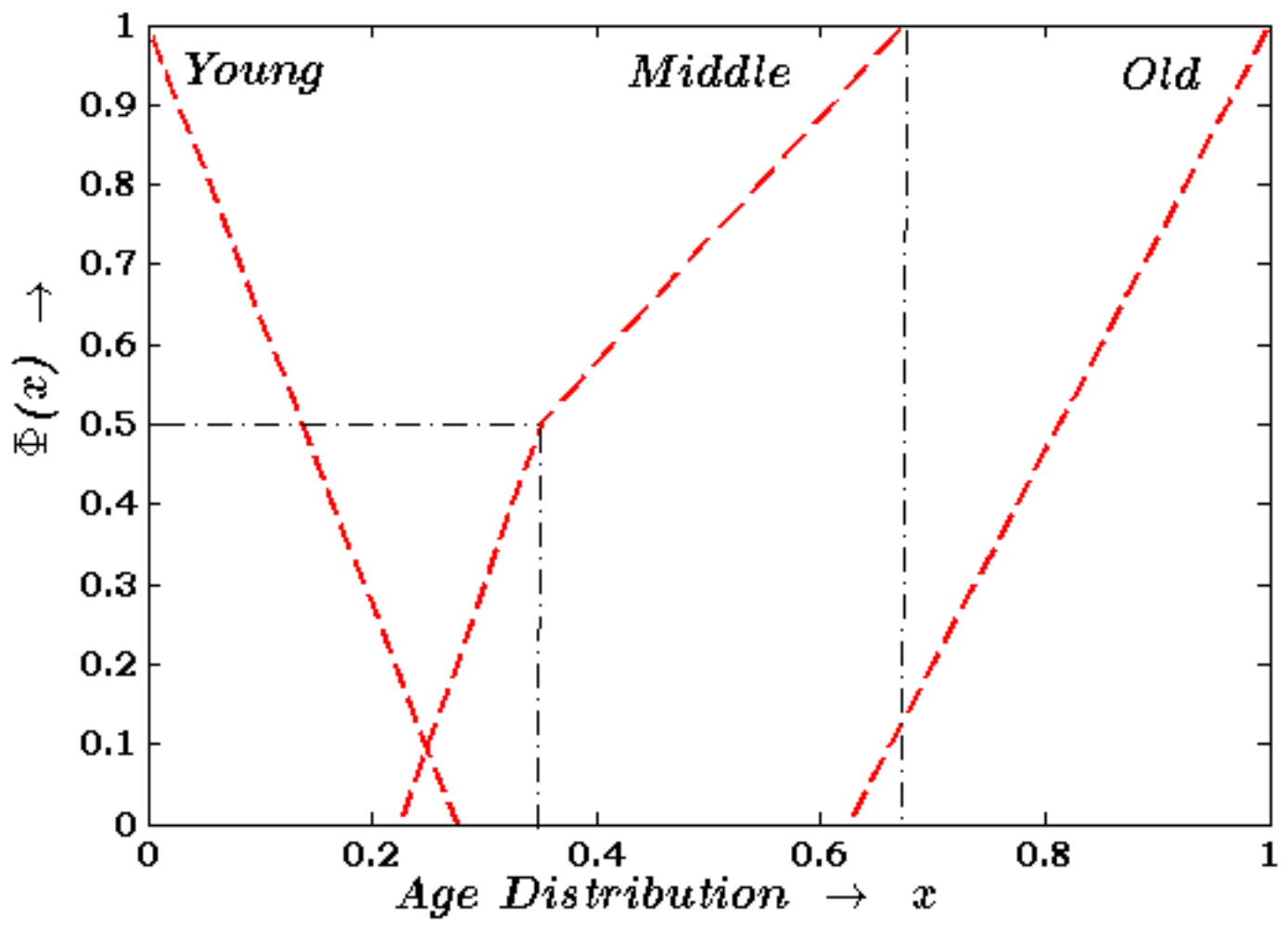

Figure 3.1 Uncertainty distributions of chromosome's age

The average of the lifetimes of the chromosomes is the average lifetime of the corresponding population. Subsequently, the $relative\ age$ of a chromosome is defined as the ratio between its age and an average lifetime of the corresponding population. Thereafter, the *relative ages* of the chromosomes are normalized, and the age interval of a chromosome $X_i$ is essentially determined as an uncertain variable, which is categorized as $Young$, $Middle$ or $Old$. In particular, the age interval for $Young$ and $Old$ are linear uncertain variables and represented by $\mathcal{L}(0, 0.275)$ and $\mathcal{L}(0.650, 1.0)$, respectively. Whereas, $Middle$ in represented by a zigzag uncertain variable $\mathcal{Z}(0.225, 0.350, 0.675)$.

### 3.4.5 Crossover

Crossover between two chromosomes, $X_i$ and $X_j$ is performed by following the steps as discussed below.

(i) **Crossover probability** $(\xi_{p_c})$: Once the age intervals of the parents $X_i$ and $X_j$ are identified, the corresponding crossover probability $\xi_{p_c}$ is estimated as an uncertain variable (i.e., $L$, $M$ or $H$) using the uncertain rule base as reported in Table 3.1. Here, $L$, $M$ and $H$, respectively represents $Low$, $Medium$ and $High$. In other words, the indeterminacy in the crossover probability for a pair of chromosomes $(X_i,\ X_j)$ can be either $Low$, $Medium$ or $High$ based on the age intervals of $X_i$ and $X_j$. Here, we have considered $L$ and $H$ as linear uncertain variables which are respectively, $\mathcal{L}(0, 0.250)$ and $\mathcal{L}(0.70,1.0)$. Whereas, $M$ is a zigzag uncertain variable represented by $\mathcal{Z}(0.150, 0.350, 0.750)$. Uncertainty distributions of $L$, $M$ and $H$ are depicted in Fig. 3.2.

Table 3.1 Uncertain rule base for crossover probability $\xi_{p_c}$

| $X_i$ \ $X_j$ | $Young$ | $Middle$ | $Old$ |
|---|---|---|---|
| $Young$ | $L$ | $M$ | $L$ |
| $Middle$ | $M$ | $H$ | $M$ |
| $Old$ | $L$ | $M$ | $L$ |

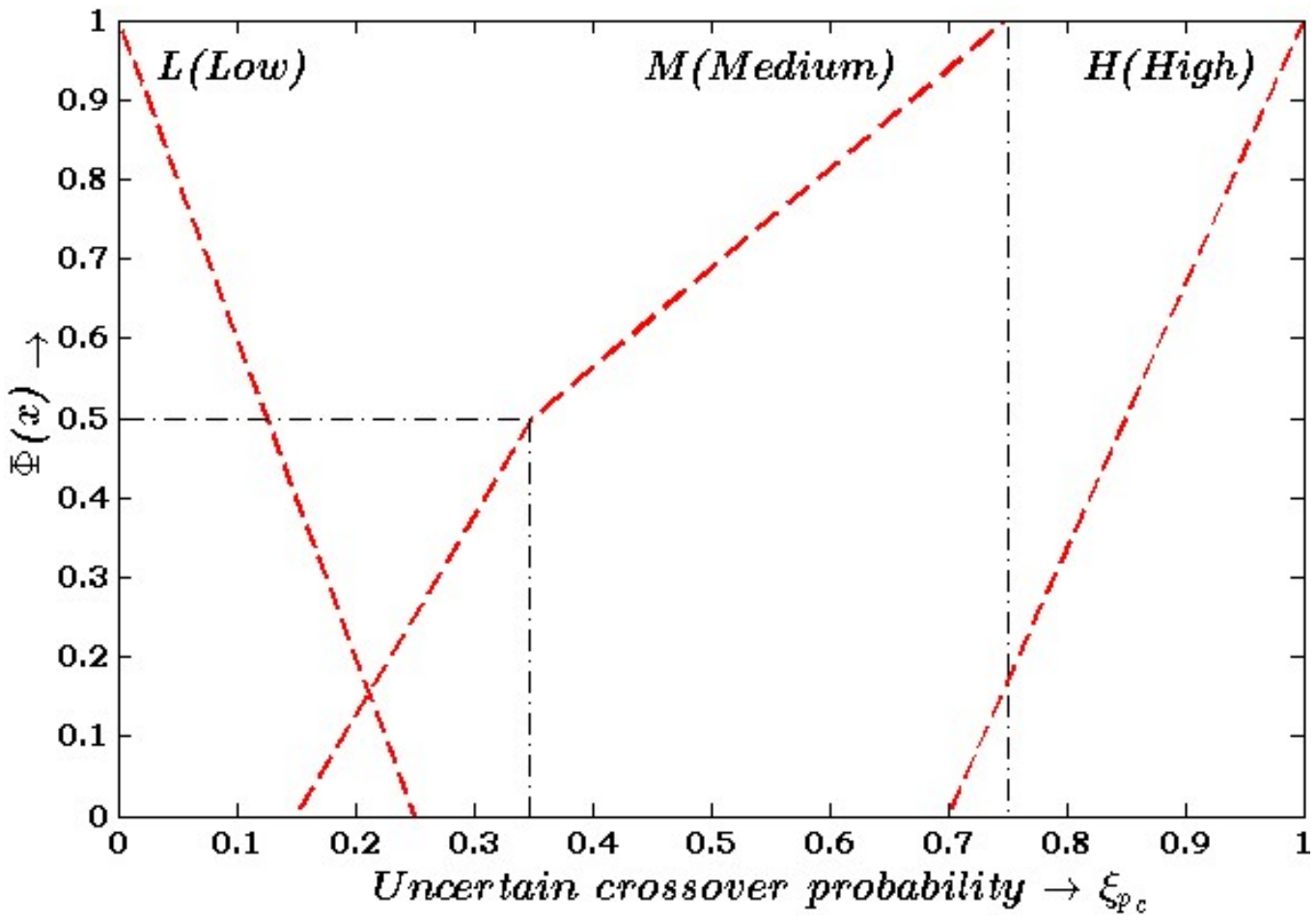

Figure 3.2 Uncertainty distributions of crossover probability $\xi_{p_c}$

(ii) **Crossover process**: The crossover process is defined in two steps which are as follow.

(a) For each chromosome pair $(X_i, X_j)$, a random number $r$ is generated within the interval $[0,1]$.

(b) For any value of $\gamma \in [0,1]$, if the crossover probability $\xi_{p_c}$for a pair of chromosomes $X_i$ and $X_j$ is

(1) $L$, then if $\mathcal{M}\{r \leq \xi_{p_c}\} \geq \gamma$, crossover is performed between $X_i$ and $X_j$.

(2) $H$, then if $\mathcal{M}\{r \geq \xi_{p_c}\} \geq \gamma$, crossover is performed between $X_i$ and $X_j$.

(3) $M$, then if $\mathcal{M}\{r \geq \xi_{p_c}\} \geq \gamma$, crossover is performed between $X_i$ and $X_j$.

In our study, $\gamma$ is set to $0.25$, since a lower value of $\gamma$ considerably increases the crossover probability and so the number of crossover operations. In order to determine the uncertainty distributions, i.e., $\mathcal{M}\{r \leq \xi_{p_c}\}$ and $\mathcal{M}\{r \geq \xi_{p_c}\}$, we have followed the definitions 1.3.29, 1.3.30 and 1.3.31, and Theorem 1.3.6 as presented in Section 1.3.11. In this study, we have used weight mapping crossover (WMX) (Gen et al. 2006) while performing the crossover between $X_i$ and $X_j$.

### 3.4.6 Mutation

Mutation is performed based on mutation probability $p_m$. For this purpose, a random number $q$ within the range $[0,1]$ is generated. If $q < p_m$, then the mutation is performed. Here, we have used insertion mutation (Gen et al. 2008) in VPGAwIC.

### 3.4.7 Selection

After crossover and mutation are performed on parent population $Pop(T)$, a total of $\lfloor \rho * |Pop(T)| \rfloor$ individuals are generated in the offspring population $C(T)$, where $\rho$ is the reproduction ratio (Z. Michalewicz 1992). In parent population, those solutions with ages greater than their lifetimes are considered as dead and are rejected. Thus $\{Pop(T) - Dead(T)\} \cup C(T)$ along with two elitist solutions are used for next generation $Pop(T+1)$. If the population size at $Pop(T+1)$ is greater than $PopSizeTh$, we randomly eliminate some chromosomes from $Pop(T+1)$ to maintain its size. However, this process ensures that the elitist solutions are preserved in $Pop(T+1)$.

### 3.4.8 Elitist Selection

From the parent population $Pop(T)$, two best solutions are selected for crossover to generate their offspring. This confirms that even if the best solutions in $Pop(T)$ die in $Pop(T+1)$, their offspring are preserved in the next generation.

### 3.4.9 Stopping Criteria

The execution of the algorithm terminates when either or both the following two conditions are satisfied.

(i) Number of generations reaches $g_{max}$.

(ii) Number of generations giving the same average fitness values reaches, $MaxSameCount$.

**Algorithm 3**: VPGAwIC

**Input**: $V$, $E$ and the capacities associated with all the edges of the given network

**Output**: Maximum flow of the network $G$

$\boldsymbol{VPGAwIC}()$

**begin**

Set $T = 0$, $SameCount$=0
Generate initial population $Pop(T)$ by *priority based encoding*
for each chromosome $R$, $F_R = -1$ in $Pop(T)$
Execute *Overall-Flow* on $Pop(T)$ // Determine the maximum flow and lifetime of every newly generated chromosome in $Pop(T)$.

**while** $SameCount < MaxSameCount$ AND $T < g_{max}$

Determine the normalize *relative age* for each chromosome in $Pop(T)$

Set $C(T) = \{\emptyset\}$

Select two best solutions from $Pop(T)$, perform WMX crossover on them and insert their offspring to $C(T)$.

**if** $\big((|C(T)| + |Pop(T)|) \leq PopSizeTh\big)$

Determine uncertain crossover probability $\xi_{p_c}$ for each selected pair of parents from $Pop(T)$.

Generate offspring by performing WMX crossover with probability $\xi_{p_c}$

For each offspring, perform insertion mutation with probability $p_m$

Add newly generated offspring to $C(T)$

**end if**

Set $Dead(T) = \{\emptyset\}$

For each chromosome in $C(T)$ set $F_R = -1$.

Increment age of each chromosome by 1 in $Pop(T)$.

Construct $Dead(T)$ if $age > lifetime$ for all chromosomes of $Pop(T)$. //Find the set of dead chromosomes

$Pop(T+1) = \{Pop(T) - Dead(T)\} \cup C(T)$.

Execute *Overall-Flow*() on $Pop(T+1)$. // Determine the maximum flow and lifetime of every newly generated chromosome in $Pop(T+1)$.

**if** $Avgfit_{T+1} = Avgfit_T$ $then$

$SameCount = SameCount + 1$

**else**

$SameCount = 0$

**end if**

$T = T + 1$

**end while**

Return best chromosome of $Pop(T)$

**end**

## 3.5 Results and Discussion

The performance of the proposed VPGAwIC is tested using six crisp instances of the maximum flow problem. Those maximum flow instances are named as $mflow_v_e$, where $v$ and $e$ are the number of nodes and edges, respectively. For each of these instances, the results of the proposed algorithm, VPGAwIC are compared with the genetic algorithm (GA) proposed by Gen et al. (2008). For comparison purpose, we have renamed the GA of Gen et al. (2008) as maximum flow genetic algorithm (MFGA).

The platform we have used for simulation is a personal computer with Intel $(R)$ Core$(TM)$ i3 @ 2.93GHz and 4 GB memory. In Section 3.5.1, the results generated by the algorithms, for the crisp instances are compared. Out of the six maximum flow instances, the description of $mflow_11_22$ and $mflow_25_49$ are given by Gen et al. (2008). The instance $mflow_20_300$ is taken from the location: ftp://dimacs.rutgers.edu/pub/netflow/maxflow/solver-4 which is defined in the file $ww.max$. While, the instances $mflow_102_290$, $mflow_902_2670$ and $mflow_2502_7450$ are accessed from the location ftp://dimacs.rutgers.edu/pub/netflow/maxflow/solver-3. These instances are maintained, respectively in the files, $m10.max$, $m30.max$ and $m50.max$. While executing the MFGA on $mflow_20_300$, *Single-Path-Growth* mechanism is used instead of *One-path growth* (Gen et al. 2008). In $mflow_20_300$ there exist cycles, and so a path from source to destination cannot be constructed if *One-path growth* is incorporated with the MFGA. In this section, the better experimental results in the tables are shown in bold. In Section 3.5.2, the results of the VPGAwIC are presented for the crisp equivalents of the EVM and the CCM for the random fuzzy maximum flow problem. These results are then compared with that generated by LINGO 11.0. Finally, a sensitivity analysis is also presented by executing the VPGAwIC at different confidence levels, for the crisp equivalent of the CCM of RFMFP.

### 3.5.1 Crisp Instances

In this subsection, the experimental results of the MFGA and the proposed VPGAwIC are compared for six maximum flow instances. The parameter setting for experimentation of MFGA and VPGAwIC are presented in Table 3.2. For VPGAwIC, the initial population size is set to $50$, and at subsequent generations, the population varies with a maximum threshold ($PopSizeTh$) of $5000$.

Due to stochastic nature of the algorithms considered in this chapter, to compare the performances of the existing MFGA and the proposed VPGAwIC, the present study considers $50$ runs for each of the algorithms on every instance. We have computed the values, $max\ fitness$, $mean\ fitness$ with standard deviation ($sd$) and $mean\ execution\ time$ in milliseconds ($ms$) for all the instances. These results are reported in Table 3.3, where it is observed that for the instances, $mflow_11_22$, $mflow_25_49$ and $mflow_20_300$, $max\ fitness$ and $mean\ fitness$ (after $50$ runs of the algorithm) do not differ for both the MFGA and VPGAwIC. Therefore, the values of $sd$ for these instances are zero. However, for the maximum flow instances, $mflow_102_290$, $mflow_902_2670$ and $mflow_2502_7450$, $max\ fitness$ and $mean\ fitness$ are better in case of the VPGAwIC as compare to the MFGA.

Table 3.2 Parameter setting for experimentation of the algorithms

| $Parameters$ | $Parameter\ values$ | |
|---|---|---|
| | MFGA | VPGAwIC |
| Population size | 100 | 50* (initial value) |
| Crossover probability ($p_c$) | 0.9 | uncertain |
| Mutation probability ($p_m$) | 0.015 | |
| Maximum number of chromosomes in a generation ($PopSizeTh$) | N/A | 5000 |
| Reproduction ratio ($\rho$) | N/A | 0.45 |
| Maximum lifetime ($minLT$) | N/A | 1 |
| Maximum lifetime ($maxLT$) | N/A | 7 |
| Maximum generation ($g_{max}$) | 500 | |
| Maximum allowed number of consecutive generations with average fitness ($MaxSameCount$) | 20 | |

*VPGAwIC has variable population size which can reach to the maximum limit of 5000.

Moreover, from Table 3.3, it is observed that the computational overhead for the VPGAwIC is more compared to the MFGA. This is due to the fact that in VPGAwIC, population size varies with every generation with a maximum value of 5000 which is set as a maximum threshold ($PopSizeTh$).

Table 3.3 $max\ fitness$, $mean\ fitness$ with $sd$ and $mean\ execution\ time$ for the six instances of maximum flow problem

| Instances | MFGA | | | VPGAwIC | | |
|---|---|---|---|---|---|---|
| | $max\ fitness$ | $(mean\ fitness, sd)$ | $mean\ execu-tion\ time\ (ms)$ | $max\ fitness$ | $(mean\ fitness, sd)$ | $mean\ execu-tion\ time\ (ms)$ |
| $mflow_11_22$ | **160.00** | (**160.00**, 0.00) | **350.00** | **160.00** | (**160.00**, 0.00) | 439.48 |
| $mflow_25_49$ | **90.00** | (**90.00**, 0.00) | **1330.24** | **90.00** | (**90.00**, 0.00) | 1498.24 |
| $mflow_20_300$ | **77962.00** | (**77962.00**, 0.00) | **7188.84** | **77962.00** | (**77962.00**, 0.00) | **25254.76** |
| $mflow_102_290$ | 9841447.00 | (9750162.38, 72512.82) | **22311.46** | **9860177.00** | (**9859140.52**, 5129.25) | 58912.74 |
| $mflow_902_2670$ | 26607221.00 | (26130086.92, 280175.14) | **372716.56** | **27595199.00** | (**27145504.16**, 280246.57) | 3503659.90 |
| $mflow_2502_7450$ | 40427368.00 | (39879257.60, 313029.69) | **3195557.46** | **41892528.00** | (**41174774.66**, 374237.46) | **16427026.10** |

The impact of setting different values of the parameters, $PopSizeTh$, $\rho$ and $maxLT$ on the proposed VPGAwIC, are shown, respectively in Table 3.4, Table 3.5 and Table 3.6. While varying each of these parameters in these tables, all the remaining associated parameters of the VPGAwIC are set to the values as mentioned in Table 3.2. With the increase of the values of these parameters, maximum fitness increases with the progressive increment of the execution time of the algorithm. It happens due to the varying population size of the proposed algorithm. These tables also show the choice of the values of the parameters which are highlighted as bold. The values as shown in these tables are generated by executing the proposed algorithm on the instance $mflow_102_290$. However, for all other instances, the choice of the values of these parameters as shown in Table 3.4-Table 3.6, also holds good for the proposed algorithm.

Table 3.4 Performances of the proposed $VPGAwIC$ on the instance $mflow_102_290$ at different values of $PopSizeTh$

| $PopSizeTh$ | $max\ fitness$ | $execution\ time(ms)$ |
|---|---|---|
| 1000 | 9315176.00 | 10534 |
| 1500 | 9516036.00 | 17372 |
| 2000 | 9519463.00 | 20572 |
| 2500 | 9639336.00 | 25297 |
| 3000 | 9640149.00 | 30734 |
| 3500 | 9822090.00 | 35949 |
| 4000 | 9841447.00 | 40728 |
| 4500 | 9841447.00 | 42419 |
| **5000** | 9860177.00 | 48936 |
| 5500 | 9860177.00 | 52308 |
| 6000 | 9860177.00 | 55946 |
| 6500 | 9860177.00 | 64495 |
| 7000 | 9860177.00 | 71023 |
| 7500 | 9860177.00 | 77809 |
| 8000 | 9860177.00 | 78975 |

To compare the performance of the proposed improved lifetime allocation strategy ($iLAS$), defined in (3.11), with the bilinear allocation strategy ($bLAS$) (Z. Michalewicz 1992), we have replaced $iLAS$ in $VPGAwIC$ with $bLAS$ and have executed $VPGAwIC$ on all the six instances for 50 runs. The results for these instances are reported in Table 3.7. Here, $VPGAwIC$- $bLAS$ implies, $VPGAwIC$ is executed with $bLAS$, whereas in $VPGAwIC$- $iLAS$, $bLAS$ is replaced with $iLAS$.

In Table 3.7, it is observed that VPGAwIC- iLAS performs better than VPGAwIC-bLAS for all the larger instances ($mflow_102_290$, $mflow_902_2670$, $mflow_2502_7450$) as far as $mean\ fitness$ is concerned. However, the computational time for VPGAwIC-bLAS is better than VPGAwIC- iLAS except for $mflow_20_300$ and $mflow_2502_7450$.

Table 3.5 Performances of the proposed $VPGAwIC$ on the instance $mflow_102_290$ at different values of $\rho$

| $\rho$ | $max\ fitness$ | $execution\ time(ms)$ | $\rho$ | $max\ fitness$ | $execution\ time(ms)$ |
|---|---|---|---|---|---|
| 0.10 | 9516036.00 | 165 | 0.55 | 9860177.00 | 151681 |
| 0.15 | 9519463.00 | 202 | 0.60 | 9860177.00 | 177396 |
| 0.20 | 9638732.00 | 239 | 0.65 | 9860177.00 | 174505 |
| 0.25 | 9640149.00 | 78872 | 0.70 | 9860177.00 | 170731 |
| 0.30 | 9642336.00 | 81959 | 0.75 | 9860177.00 | 175434 |
| 0.35 | 9822090.00 | 89868 | 0.80 | 9860177.00 | 176272 |
| 0.40 | 9841447.00 | 101725 | 0.85 | 9860177.00 | 177161 |
| **0.45** | 9860177.00 | 105317 | 0.90 | 9860177.00 | 175257 |
| 0.50 | 9860177.00 | 108727 | 0.95 | 9860177.00 | 174498 |

Table 3.6 Performances of the proposed $VPGAwIC$ on the instance $mflow_102_290$ at different values of $maxLT$

| $maxLT$ | $max\ fitness$ | $execution\ time(ms)$ | $maxLT$ | $max\ fitness$ | $execution\ time(ms)$ |
|---|---|---|---|---|---|
| 2 | 9315176.00 | 173 | **7** | 9860177.00 | 72990 |
| 3 | 9513059.00 | 196 | 8 | 9860177.00 | 77875 |
| 4 | 9516036.00 | 237 | 9 | 9860177.00 | 83124 |
| 5 | 9638732.00 | 91310 | 10 | 9860177.00 | 87379 |
| 6 | 9834265.00 | 92550 | 11 | 9860177.00 | 89293 |

The convergence plots of the algorithms are also presented for three larger instances: $mflow_102_290$, $mflow_902_2670$ and $mflow_2502_7450$ in Fig 3.3. The representation of each graph in Fig. 3.3 is done by plotting the $mean\ fitness$ at every generation. It is noticed, that for all the three instances, the convergence is better for the VPGAwIC-iLAS, although more generations are required for the VPGAwIC-iLAS to converge compare to its counterparts.

Table 3.8 shows the $median$ and $IQR$ of $max\ fitness$ after $50$ runs of the algorithms, for the six instances. Here, the performance of MFGA is outperformed by VPGAwIC for instances, $mflow_102_290$, $mflow_902_2670$ and $mflow_2502_7450$. Moreover, while comparing the $median$ and interquartile range$(IQR)$ of $max\ fitness$, of all the instances, for VPGAwIC-bLAS and VPGAwIC-iLAS in Table 3.9, it is observed that VPGAwIC-iLAS generates superior results for $mflow_902_2670$ and $mflow_2502_7450$. However, for the remaining instances, the results are similar. For most of the instances in tables 3.8 and 3.9, the values of $IQR$ is zero, since identical values of $max\ fitness$ are generated after every execution of the algorithms, for most of the test instances. It essentially means that for all the instances, except $mflow_902_2670$ and $mflow_2502_7450$, the first and third quartiles of 50 observations of the $max\ fitness$ overlap with the median value. This fact can also be visually interpreted in Fig. 3.4 (a)-(c).

Table 3.7 Comparison of results for the proposed $VPGAwIC$ with different allocation strategies for the six instances of maximum flow problem

| Instances | VPGAwIC- bLAS | | | VPGAwIC- iLAS | | |
|---|---|---|---|---|---|---|
| | $max$ $fitness$ | $(mean$ $fitness, sd)$ | $mean$ $execu-tion\ time$ $(ms)$ | $max$ $fitness$ | $(mean\ fitness,$ $sd)$ | $mean$ $execu-tion\ time$ $(ms)$ |
| $mflow_$ 11_22 | $\mathbf{160.00}$ | $(\mathbf{160.00}$, 0.00) | $\mathbf{417.04}$ | $\mathbf{160.00}$ | $(\mathbf{160.00}$, 0.00) | 439.48 |
| $mflow_$ 25_49 | $\mathbf{90.00}$ | $(\mathbf{90.00}$, 0.00) | $\mathbf{1425.34}$ | $\mathbf{90.00}$ | $(\mathbf{90.00}$, 0.00) | 1498.24 |
| $mflow_$ 20_300 | $\mathbf{77962.00}$ | $(\mathbf{77962.00}$, 0.00) | 37859.16 | $\mathbf{77962.00}$ | $(\mathbf{77962.00}$, 0.00) | $\mathbf{25254.76}$ |
| $mflow_$ 102_290 | $\mathbf{9860177.00}$ | (9854682.96, 17587.44) | $\mathbf{55971.48}$ | $\mathbf{9860177.00}$ | $(\mathbf{9859140.52}$, 5129.25) | 58912.74 |
| $mflow_$ 902_2670 | 27575843.00 | (26964672.16, 302750.85) | $\mathbf{3469381.26}$ | $\mathbf{27595199.00}$ | $(\mathbf{27145504.16}$, 280246.57) | 3503659.90 |
| $mflow_$ 2502_7450 | 41491424.00 | (40925464.12, 343045.10) | 17140499.86 | $\mathbf{41892528.00}$ | $(\mathbf{41174774.66}$, 374237.46) | $\mathbf{16427026.10}$ |

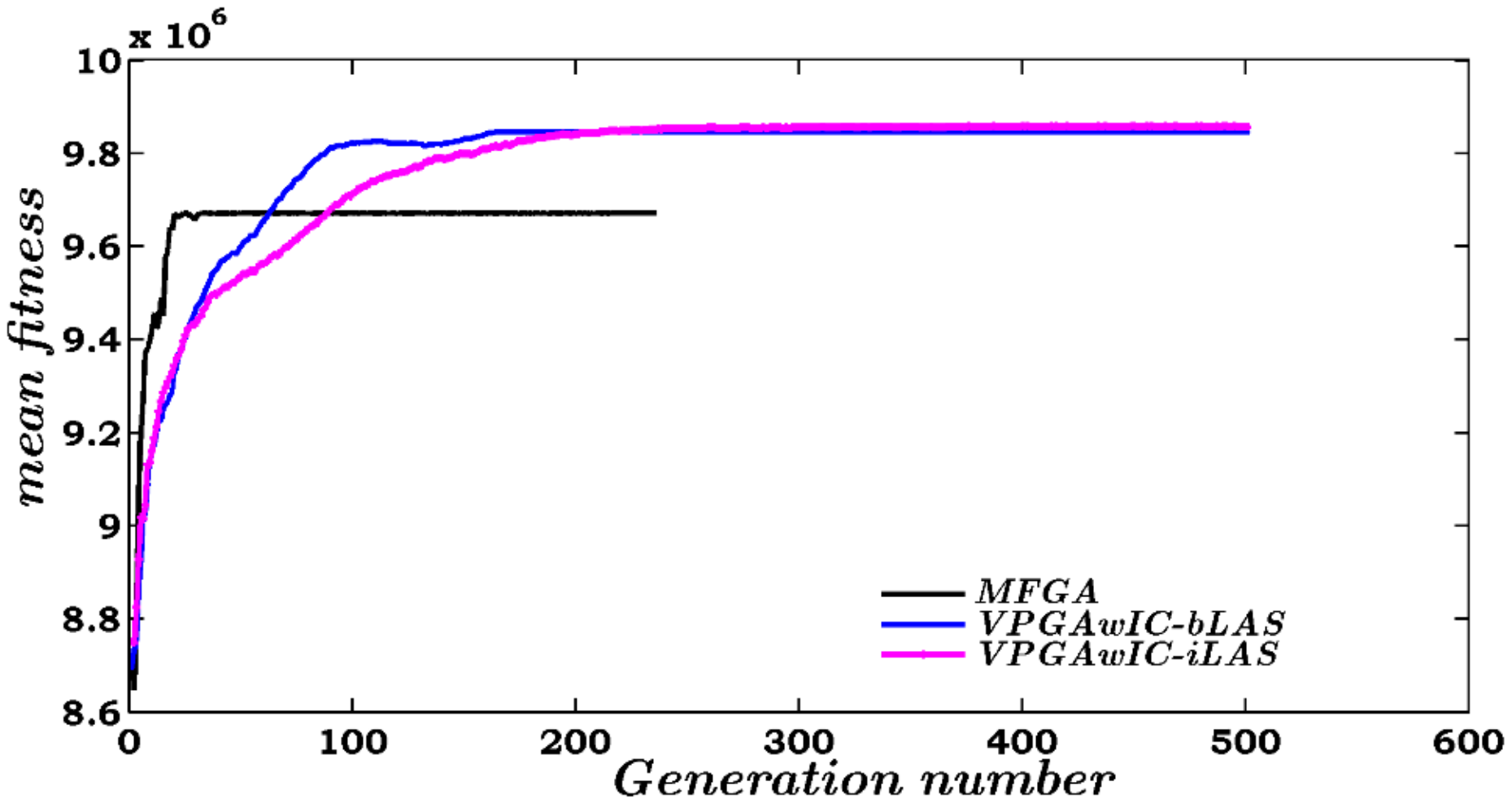

(a)

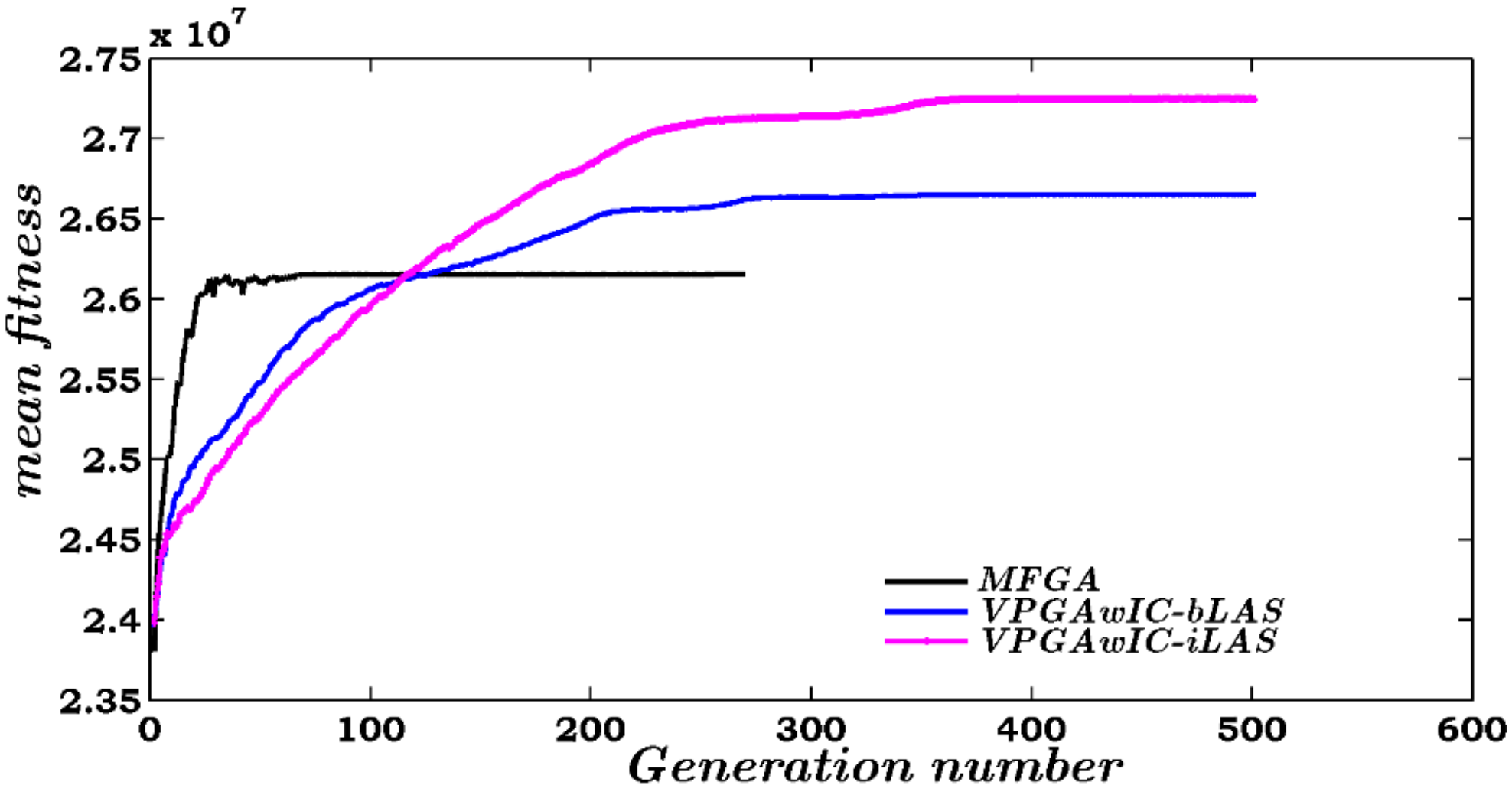

(b)

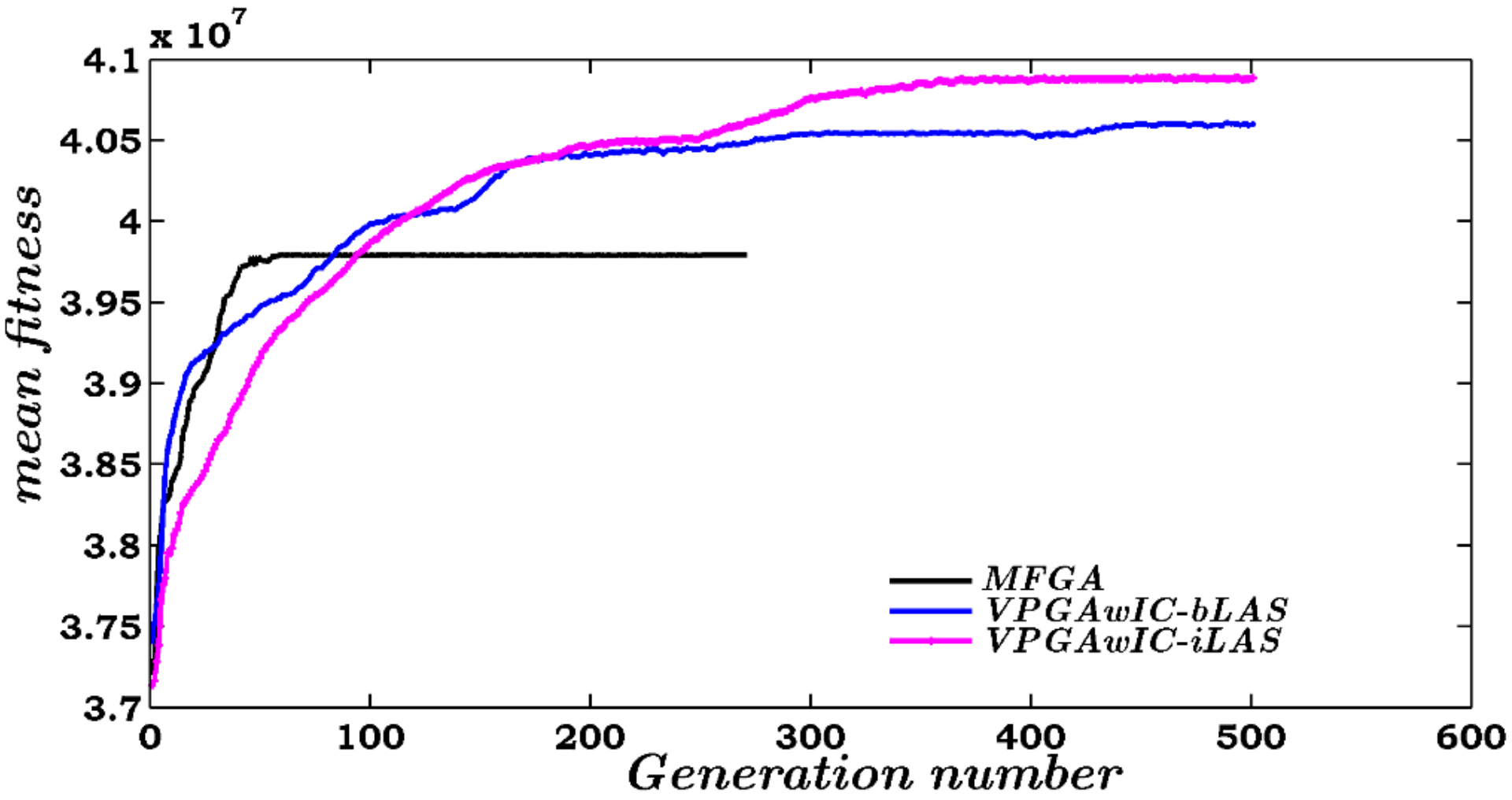

(c)

Figure 3.3 Convergence plots of MFGA, VPGAwIC-bLAS and VPGAwIC-iLAS for the maximum flow instances: (a) $mflow_102_290$ (b) $mflow_902_2670$ and (c) $mflow_2502_7450$

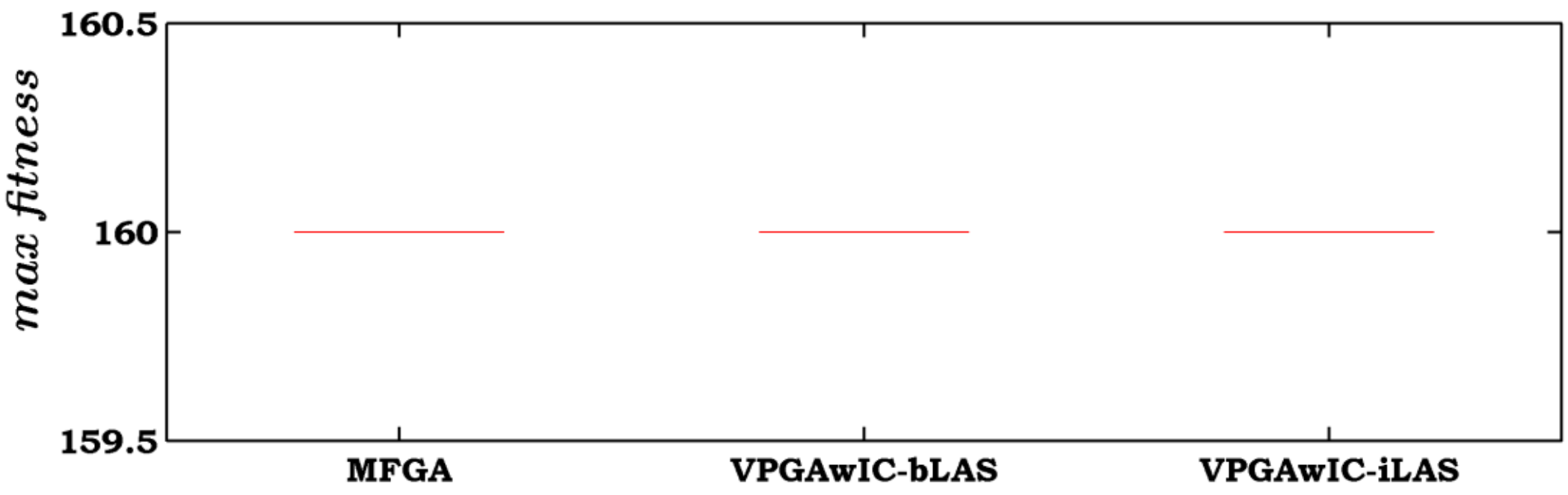

(a)

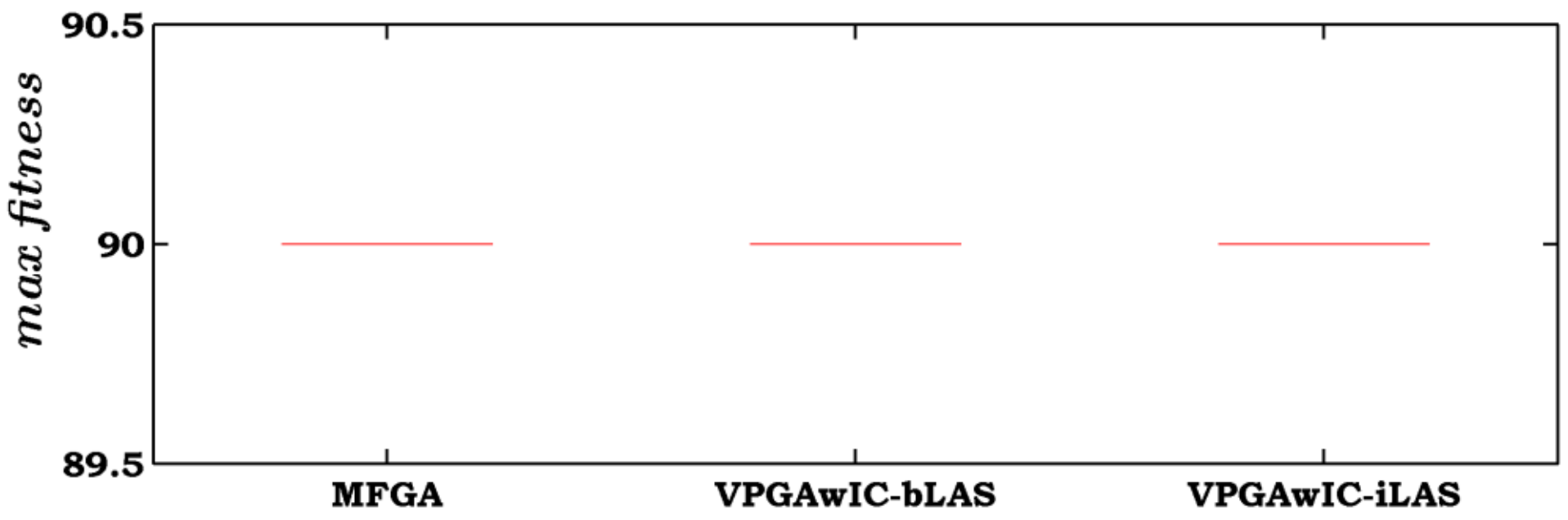

(b)

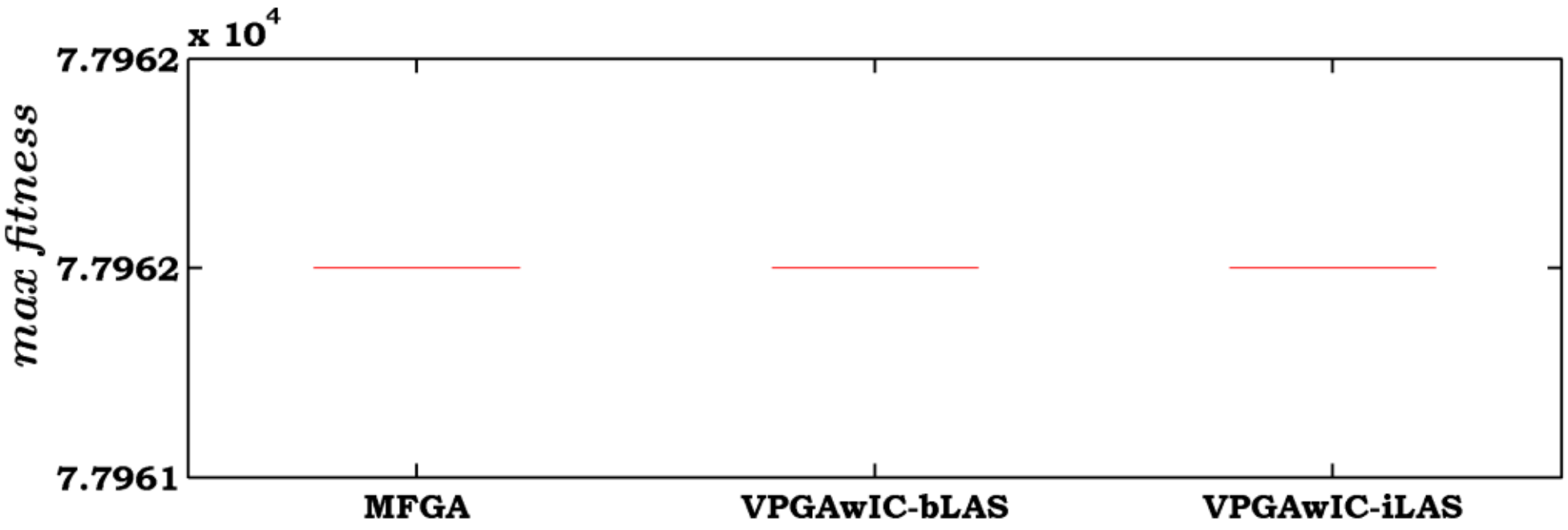

(c)

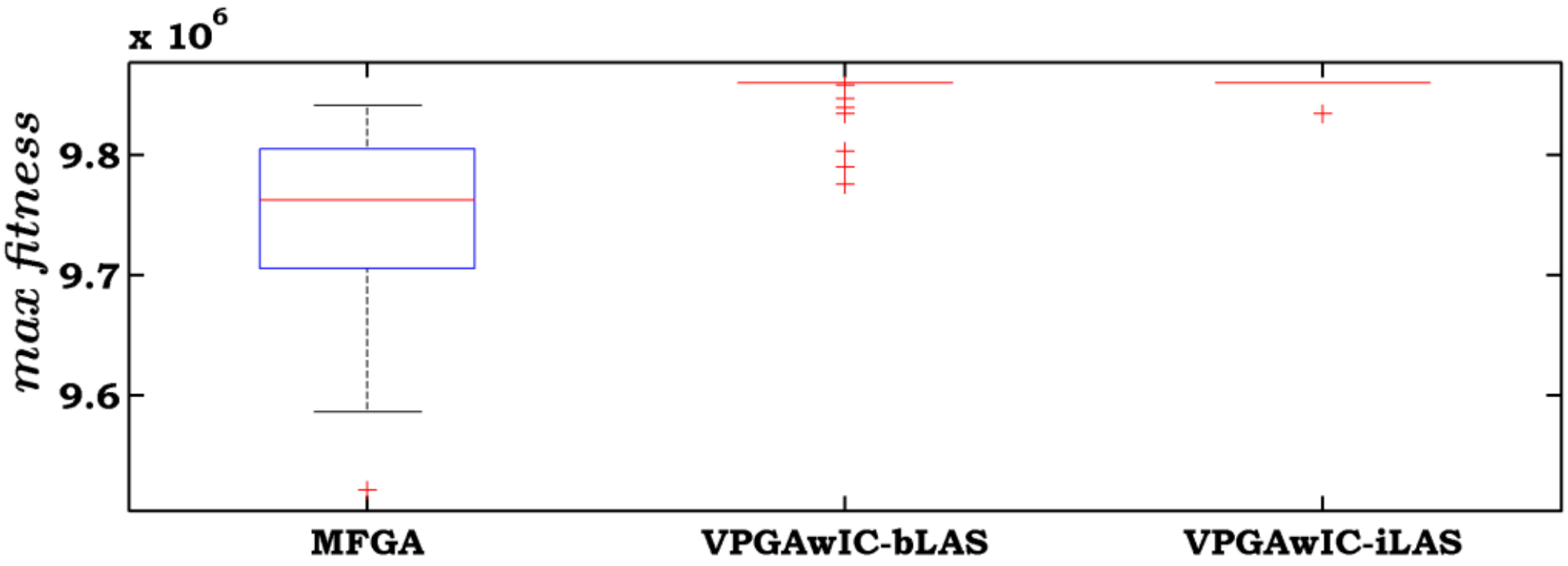

(d)

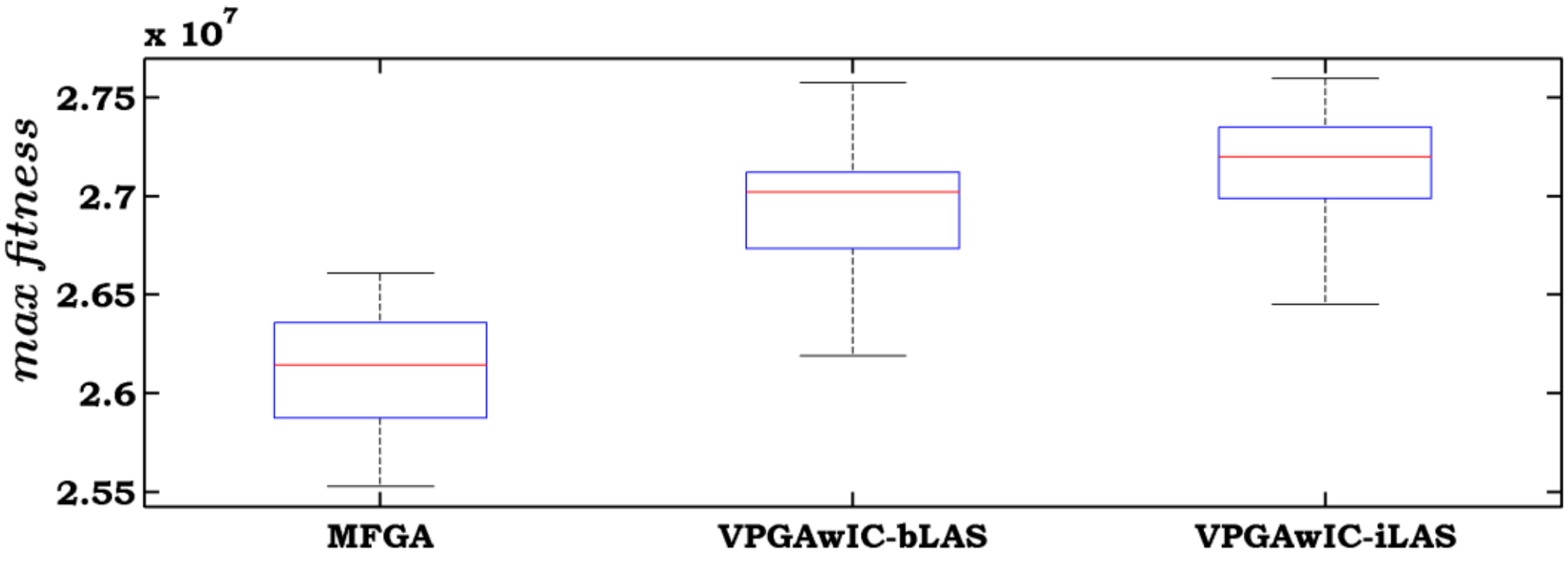

(e)

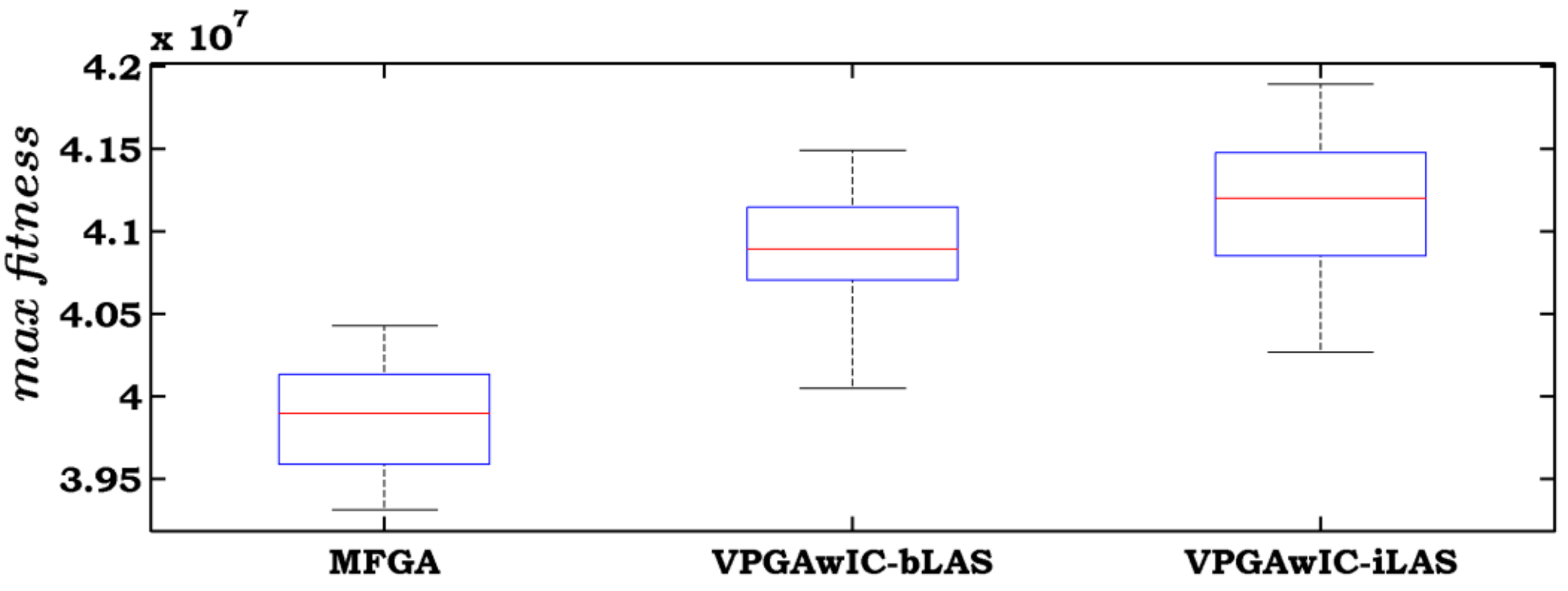

(f)

Figure 3.4 Boxplots of the maximum flow instances: (a) mflow_11_22 (b) mflow_25_49 (c) mflow_20_300 (d) mflow_102_290 *(e)* mflow_902_2670 and (f) mflow_2502_7450

Table 3.8 Median of maximum fitness of the six instances of maximum flow problem for the $MFGA$ and proposed $VPGAwIC$

| Instances | $(median, IQR)$ | |
|---|---|---|
| | MFGA | VPGAwIC |
| $mflow_11_22$ | ($\mathbf{160.00}$, 0.00) | ($\mathbf{160.00}$, 0.00) |
| $mflow_25_49$ | ($\mathbf{90.00}$, 0.00) | ($\mathbf{90.00}$, 0.00) |
| $mflow_20_300$ | ($\mathbf{77962.00}$, 0.00) | ($\mathbf{77962.00}$, 0.00) |
| $mflow_102_290$ | (9762344.50, 99139.00) | ($\mathbf{9860177.00}$, 0.00) |
| $mflow_902_2670$ | (26144623.50, 483453.00) | ($\mathbf{27198010.50}$, 362394.00) |
| $mflow_2502_7450$ | (39900146.50, 543049.00) | ($\mathbf{41203131.50}$, 622122.00) |

Fig. 3.4 displays the box plots corresponding to all the maximum flow instances. Here, the boxplots of $mflow_11_22$, $mflow_25_49$ and $mflow_20_300$ in Fig. 3.4 (a)-(c) display only the median values with no quartiles or outliers, for all the algorithms (MFGA, VPGAwIC-bLAS and VPGAwIC-iLAS). Since, for these instances, the algorithms generate identical $max\ fitness$ after each of their respective execution for 50 runs. Therefore, with respect to all the algorithms, the $IQR$ becomes zero for these instances. Similarly, for $mflow_102_290$, the box plots in Fig. 3.4(d) corresponding to VPGAwIC-bLAS and VPGAwIC-iLAS are represented only by the median and few outliers, since for most of the runs, VPGAwIC-bLAS and VPGAwIC- iLAS generate identical values of $max\ fitness$. Consequently, the corresponding $IQR$s of $mflow_102_290$, for VPGAwIC-bLAS and VPGAwIC- iLAS become zero. Whereas, the $max\ fitness$ differs with every execution of MFGA and therefore the box plot corresponding to the MFGA is constructed with unique first and third quartiles for $mflow_102_290$. For $mflow_902_2670$ and $mflow_2502_7450$, all the algorithms generate boxplots (cf. Fig. 3.4 (e)-(f)) with distinct first and third quartiles. For each of these instances, the median generated by the VPGAwIC-iLAS is greater compare to the VPGAwIC-bLAS and the MFGA.

Table 3.9 Median of maximum fitness of the six instances of maximum flow problem for the $VPGAwIC\text{-}bLAS$ and $VPGAwIC\text{-}iLAS$

| Instances | $(median, IQR)$ | |
|---|---|---|
| | VPGAwIC-bLAS | VPGAwIC-iLAS |
| $mflow_11_22$ | ($\mathbf{160.00}$, 0.00) | ($\mathbf{160.00}$, 0.00) |
| $mflow_25_49$ | ($\mathbf{90.00}$, 0.00) | ($\mathbf{90.00}$, 0.00) |
| $mflow_20_300$ | ($\mathbf{77962.00}$, 0.00) | ($\mathbf{77962.00}$, 0.00) |
| $mflow_102_290$ | ($\mathbf{9860177.00}$, 0.00) | ($\mathbf{9860177.00}$, 0.00) |
| $mflow_902_2670$ | (27021787.50, 384489.00) | ($\mathbf{27198010.50}$, 362394.00) |
| $mflow_2502_7450$ | (40892217.50, 444743.00) | ($\mathbf{41203131.50}$, 622122.00) |

### 3.5.2 Random Fuzzy Instances

In this subsection, we have considered a directed connected network $G = (V_G, E_G)$ with a source vertex $v_1$ and a sink vertex $v_{10}$, which are respectively, designated as $s$ and $t$. The capacity associated with an edge $e_{ij}$ in $G$ is an uncertain quantity and is represented by a random fuzzy variable. Two random fuzzy instances, namely $mflow_TrRF_10_15$ and $mflow_GRF_10_15$ are considered for which the capacities are represented by trapezoidal random fuzzy variables (TrRFVs) and Gaussian random fuzzy variables (GRFVs), respectively. The corresponding capacities of each of these instances are reported in Table 3.10 and Table 3.11. The graphical representation of both the instances is the same and depicted in Fig. 3.5.

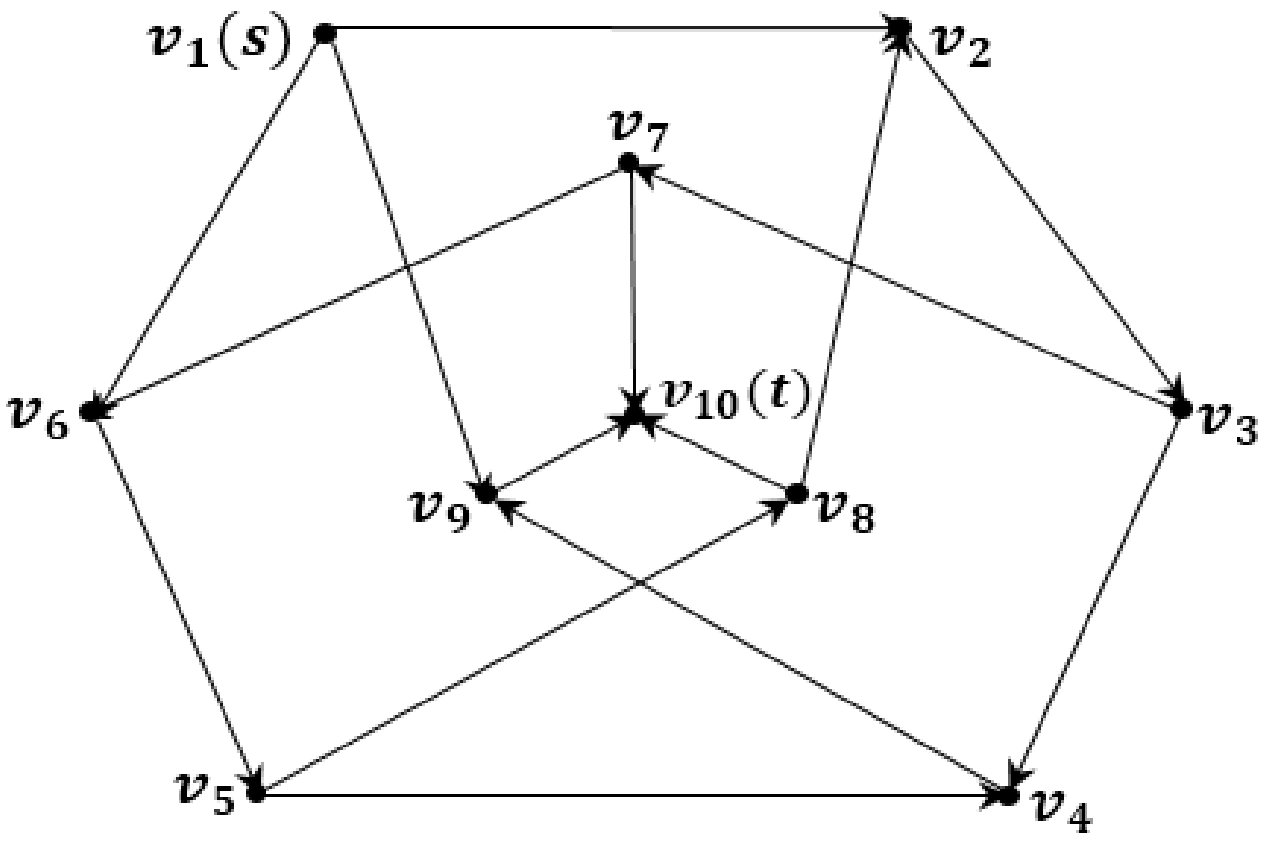

Figure 3.5 A directed connected network $G$

To determine the maximum flow of RFMFP for $mflow_TrRF_10_15$ and $mflow_GRF_10_15$, we formulate the EVM and the CCM following the respective models (3.3) and (3.5). Subsequently, these models are then transformed to their crisp equivalents as defined in (3.7)-(3.10). Specifically, for $mflow_TrRF_10_15$, the crisp equivalents of (3.3) and (3.5) are respectively, (3.7) and (3.9). Similarly, for $mflow_GRF_10_15$, the corresponding crisp equivalents of the EVM and the CCM are shown in (3.8) and (3.10).

Table 3.10 Random fuzzy capacities expressed as TrRFV of the instance $mflow_TrRF_10_15$

| Edges | Trapezoidal random fuzzy edge capacities |
|---|---|
| $e_{12}$ | $\tilde{\tilde{\zeta}}_{C_{12}} = \mathcal{N}(\tilde{\zeta}_{C_{12}}, 43.883), \tilde{\zeta}_{C_{12}} = \mathcal{Tr}[64, 89, 107, 135]$ |
| $e_{16}$ | $\tilde{\tilde{\zeta}}_{C_{16}} = \mathcal{N}(\tilde{\zeta}_{C_{16}}, 83.140), \tilde{\zeta}_{C_{16}} = \mathcal{Tr}[62, 81, 120, 127]$ |
| $e_{19}$ | $\tilde{\tilde{\zeta}}_{C_{19}} = \mathcal{N}(\tilde{\zeta}_{C_{19}}, 32.704), \tilde{\zeta}_{C_{19}} = \mathcal{Tr}[71, 80, 117, 130]$ |
| $e_{23}$ | $\tilde{\tilde{\zeta}}_{23} = \mathcal{N}(\tilde{\zeta}_{C_{23}}, 78.189), \tilde{\zeta}_{C_{23}} = \mathcal{Tr}[59, 88, 118, 126]$ |
| $e_{34}$ | $\tilde{\tilde{\zeta}}_{34} = \mathcal{N}(\tilde{\zeta}_{C_{34}}, 88.550), \tilde{\zeta}_{C_{34}} = \mathcal{Tr}[55, 86, 122, 134]$ |
| $e_{37}$ | $\tilde{\tilde{\zeta}}_{37} = \mathcal{N}(\tilde{\zeta}_{C_{37}}, 71.310), \tilde{\zeta}_{C_{37}} = \mathcal{Tr}[65, 87, 124, 129]$ |
| $e_{49}$ | $\tilde{\tilde{\zeta}}_{49} = \mathcal{N}(\tilde{\zeta}_{C_{49}}, 29.229), \tilde{\zeta}_{C_{49}} = \mathcal{Tr}[67, 75, 114, 136]$ |
| $e_{54}$ | $\tilde{\tilde{\zeta}}_{54} = \mathcal{N}(\tilde{\zeta}_{C_{54}}, 77.536), \tilde{\zeta}_{C_{54}} = \mathcal{Tr}[69, 93, 105, 131]$ |
| $e_{58}$ | $\tilde{\tilde{\zeta}}_{58} = \mathcal{N}(\tilde{\zeta}_{C_{58}}, 62.234), \tilde{\zeta}_{C_{58}} = \mathcal{Tr}[59, 94, 103, 123]$ |
| $e_{65}$ | $\tilde{\tilde{\zeta}}_{65} = \mathcal{N}(\tilde{\zeta}_{C_{65}}, 19.025), \tilde{\zeta}_{C_{65}} = \mathcal{Tr}[55, 81, 112, 139]$ |
| $e_{76}$ | $\tilde{\tilde{\zeta}}_{76} = \mathcal{N}(\tilde{\zeta}_{C_{76}}, 37.978), \tilde{\zeta}_{C_{76}} = \mathcal{Tr}[68, 89, 125, 137]$ |
| $e_{710}$ | $\tilde{\tilde{\zeta}}_{710} = \mathcal{N}(\tilde{\zeta}_{C_{710}}, 56.674), \tilde{\zeta}_{C_{710}} = \mathcal{Tr}[72, 95, 119, 133]$ |
| $e_{82}$ | $\tilde{\tilde{\zeta}}_{82} = \mathcal{N}(\tilde{\zeta}_{C_{82}}, 17.736), \tilde{\zeta}_{C_{82}} = \mathcal{Tr}[57, 84, 121, 128]$ |
| $e_{810}$ | $\tilde{\tilde{\zeta}}_{810} = \mathcal{N}(\tilde{\zeta}_{C_{810}}, 45.288), \tilde{\zeta}_{C_{810}} = \mathcal{Tr}[66, 98, 110, 138]$ |
| $e_{910}$ | $\tilde{\tilde{\zeta}}_{910} = \mathcal{N}(\tilde{\zeta}_{C_{910}}, 58.982), \tilde{\zeta}_{C_{910}} = \mathcal{Tr}[58, 80, 115, 140]$ |

The solutions of the EVM for random fuzzy instances are determined by solving models (3.7) and (3.8), and that of the CCM are obtained by solving models (3.9) and (3.10). Each of these models is solved by VPGAwIC, and the results are compared with the solutions obtained by solving the models with a standard optimization solver, LINGO 11.0. These solutions are organized in Table 3.12.

Table 3.11 Random fuzzy capacities expressed as GRFV of the instance $mflow_GRF_10_15$

| Edges | Gaussian random fuzzy edge capacities |
|---|---|
| $e_{12}$ | $\tilde{\tilde{\zeta}}_{C_{12}} = \mathcal{N}(\tilde{\zeta}_{C_{12}}, 43.883), \tilde{\zeta}_{C_{12}} = \mathcal{N}(60, 19.319)$ |
| $e_{16}$ | $\tilde{\tilde{\zeta}}_{C_{16}} = \mathcal{N}(\tilde{\zeta}_{C_{16}}, 83.140), \tilde{\zeta}_{C_{16}} = \mathcal{N}(62, 18.989)$ |
| $e_{19}$ | $\tilde{\tilde{\zeta}}_{C_{19}} = \mathcal{N}(\tilde{\zeta}_{C_{19}}, 32.704), \tilde{\zeta}_{C_{19}} = \mathcal{N}(84,11.652)$ |
| $e_{23}$ | $\tilde{\tilde{\zeta}}_{23} = \mathcal{N}(\tilde{\zeta}_{C_{23}}, 78.189), \tilde{\zeta}_{C_{23}} = \mathcal{N}(125, 19.095)$ |
| $e_{34}$ | $\tilde{\tilde{\zeta}}_{34} = \mathcal{N}(\tilde{\zeta}_{C_{34}}, 88.550), \tilde{\zeta}_{C_{34}} = \mathcal{N}(118, 24.846)$ |
| $e_{37}$ | $\tilde{\tilde{\zeta}}_{37} = \mathcal{N}(\tilde{\zeta}_{C_{37}}, 71.310), \tilde{\zeta}_{C_{37}} = \mathcal{N}(131, 22.953)$ |
| $e_{49}$ | $\tilde{\tilde{\zeta}}_{49} = \mathcal{N}(\tilde{\zeta}_{C_{49}}, 29.229), \tilde{\zeta}_{C_{49}} = \mathcal{N}(110, 27.589)$ |
| $e_{54}$ | $\tilde{\tilde{\zeta}}_{54} = \mathcal{N}(\tilde{\zeta}_{C_{54}}, 77.536), \tilde{\zeta}_{C_{54}} = \mathcal{N}(86, 12.809)$ |
| $e_{58}$ | $\tilde{\tilde{\zeta}}_{58} = \mathcal{N}(\tilde{\zeta}_{C_{58}}, 62.234), \tilde{\zeta}_{C_{58}} = \mathcal{N}(114, 25.925)$ |
| $e_{65}$ | $\tilde{\tilde{\zeta}}_{65} = \mathcal{N}(\tilde{\zeta}_{C_{65}}, 19.025), \tilde{\zeta}_{C_{65}} = \mathcal{N}(135, 15.318)$ |
| $e_{76}$ | $\tilde{\tilde{\zeta}}_{76} = \mathcal{N}(\tilde{\zeta}_{C_{76}}, 37.978), \tilde{\zeta}_{C_{76}} = \mathcal{N}(133, 23.008)$ |
| $e_{710}$ | $\tilde{\tilde{\zeta}}_{710} = \mathcal{N}(\tilde{\zeta}_{C_{710}}, 56.674), \tilde{\zeta}_{C_{710}} = \mathcal{N}(97, 22.650)$ |
| $e_{82}$ | $\tilde{\tilde{\zeta}}_{82} = \mathcal{N}(\tilde{\zeta}_{C_{82}}, 17.736), \tilde{\zeta}_{C_{82}} = \mathcal{N}(117, 29.241)$ |
| $e_{810}$ | $\tilde{\tilde{\zeta}}_{810} = \mathcal{N}(\tilde{\zeta}_{C_{810}}, 45.288), \tilde{\zeta}_{C_{810}} = \mathcal{N}(81, 19.314)$ |
| $e_{910}$ | $\tilde{\tilde{\zeta}}_{910} = \mathcal{N}(\tilde{\zeta}_{C_{910}}, 58.982), \tilde{\zeta}_{C_{910}} = \mathcal{N}(137, 23.138)$ |

Table 3.12 Comparison of results for crisp equivalents of the EVM and the CCM

| Random fuzzy instances | | | | | |
|---|---|---|---|---|---|
| $mflow_TrRF_10_15$ | | | $mflow_GRF_10_15$ | | |
| Models | LINGO | VPGAwIC | Models | LINGO | VPGAwIC |
| EVM (3.7) | 290.75 | 290.75 | EVM (3.8) | 206.00 | 206.00 |
| CCM (3.9) | 160.18 | 160.18 | CCM (3.10) | 115.15 | 115.15 |

We have maintained similar parameter settings of VPGAwIC as reported in Table 3.2 while solving the models (3.7)-(3.10) of $G$. In Table 3.12, the results corresponding to the models, (3.9) and (3.10) are obtained by setting the predetermined confidence levels $\alpha_{ij}(Pr)$ and $\beta_{ij}(Cr)$, respectively to 0.8997 and 0.9 corresponding to each $e_{ij}$ of $G$. It is observed from Table 3.12, that LINGO and VPGAwIC generate similar results for both the random fuzzy instances. Moreover, the convergence plots of VPGAwIC for the models (3.7)-(3.10) are depicted in Fig. 3.6. In each of these plots, maximum fitness values and mean fitness values are displayed for each generation. Here, it is observed that the mean fitness converges with the maximum fitness for all these plots.

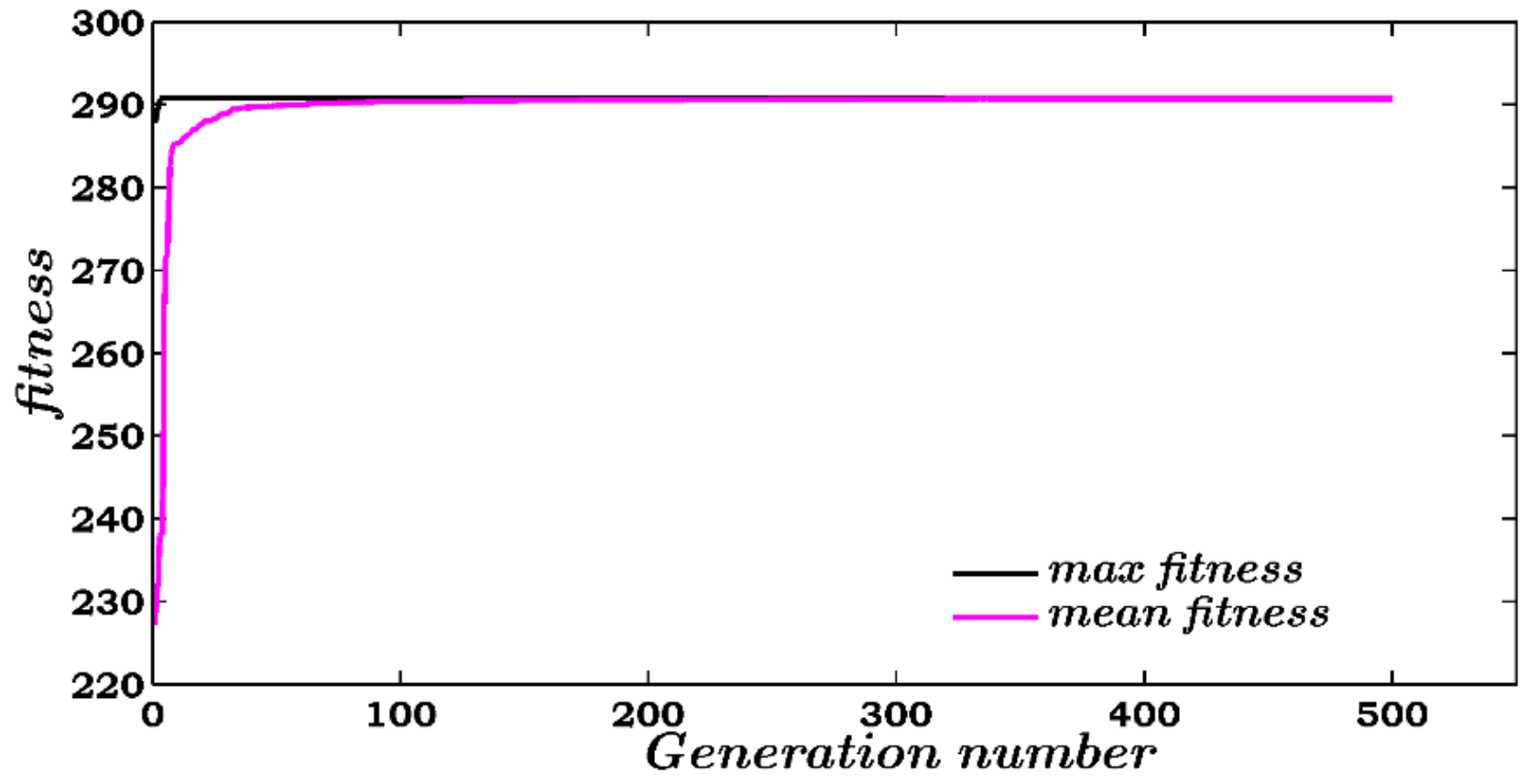

(a)

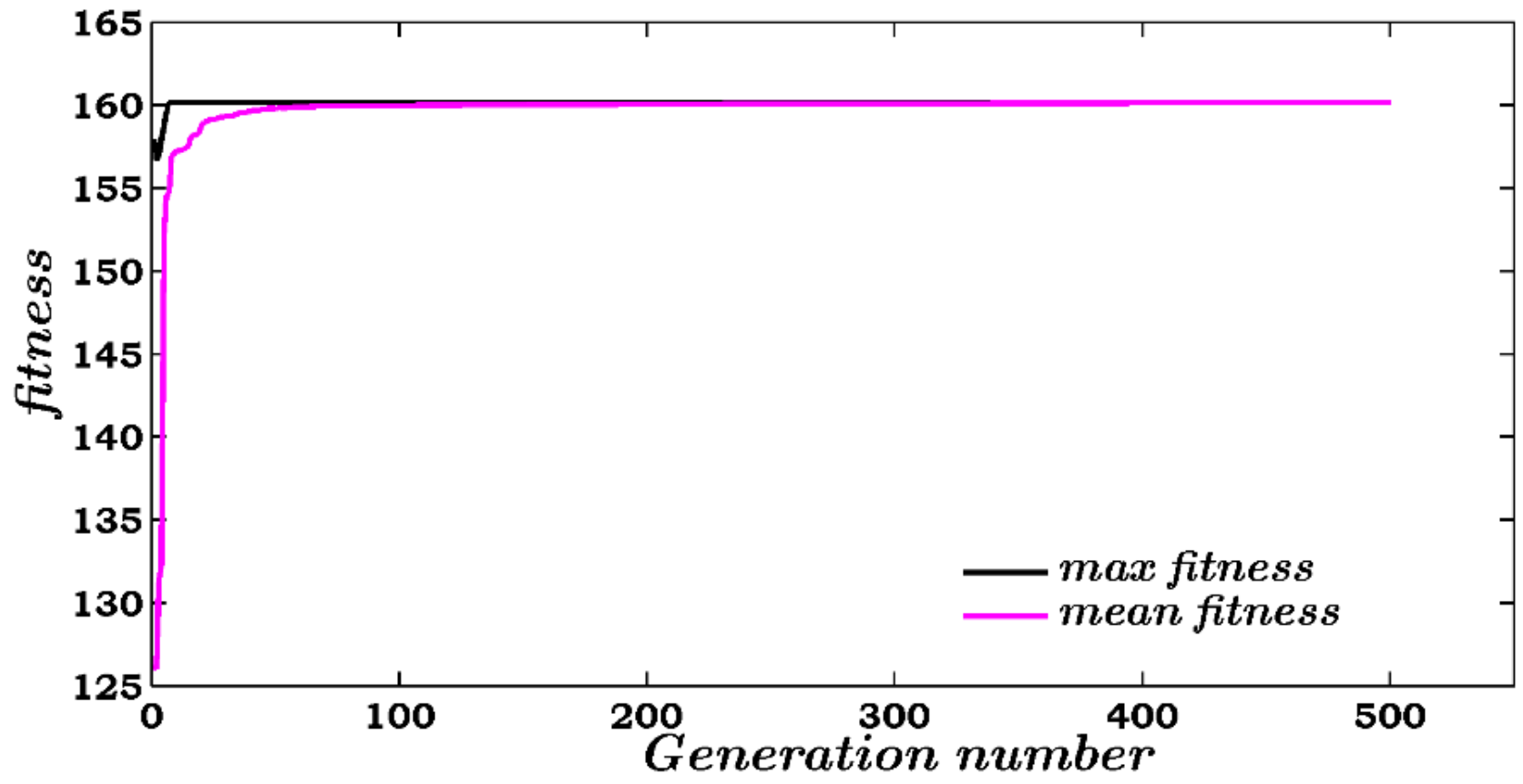

(b)

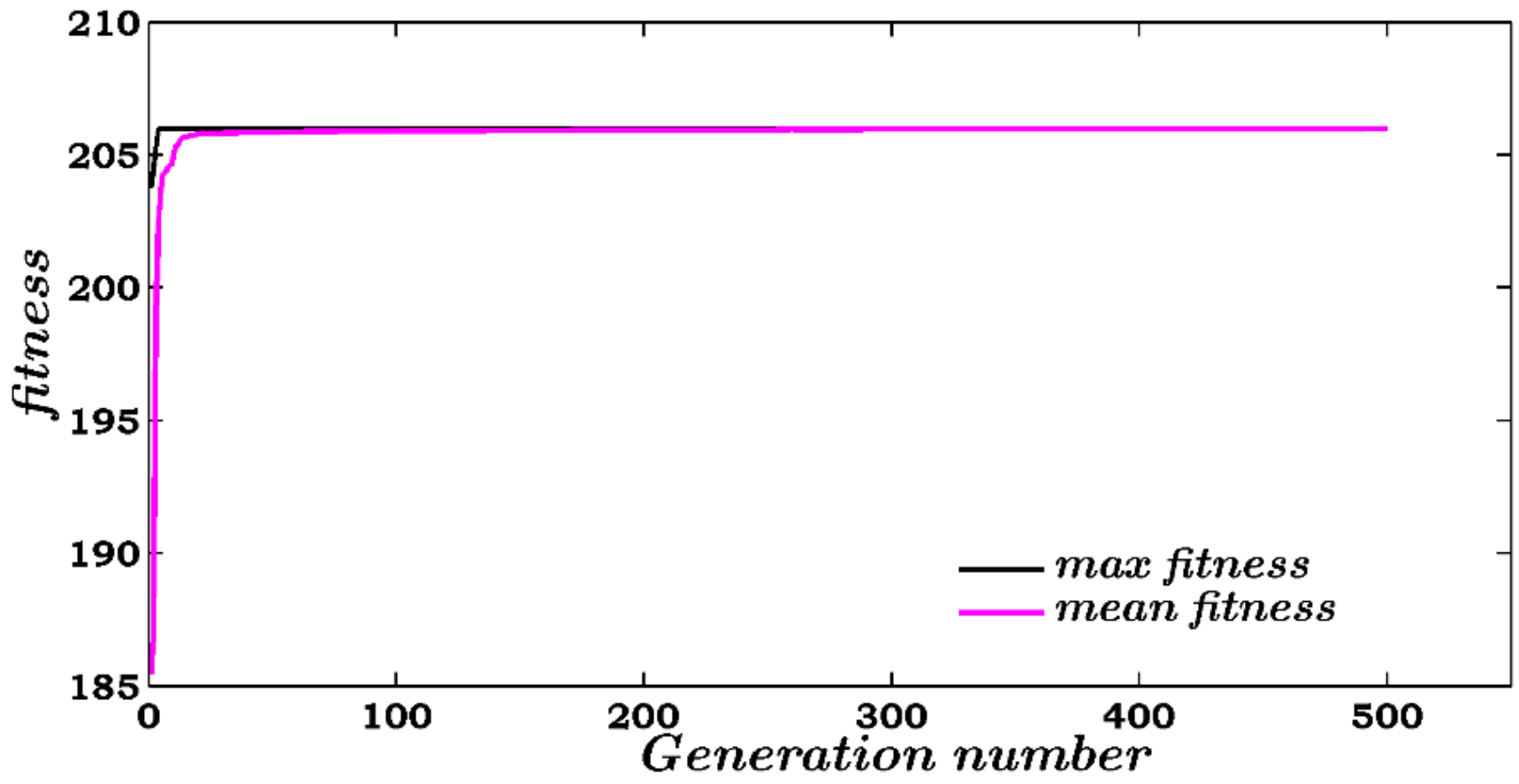

(c)

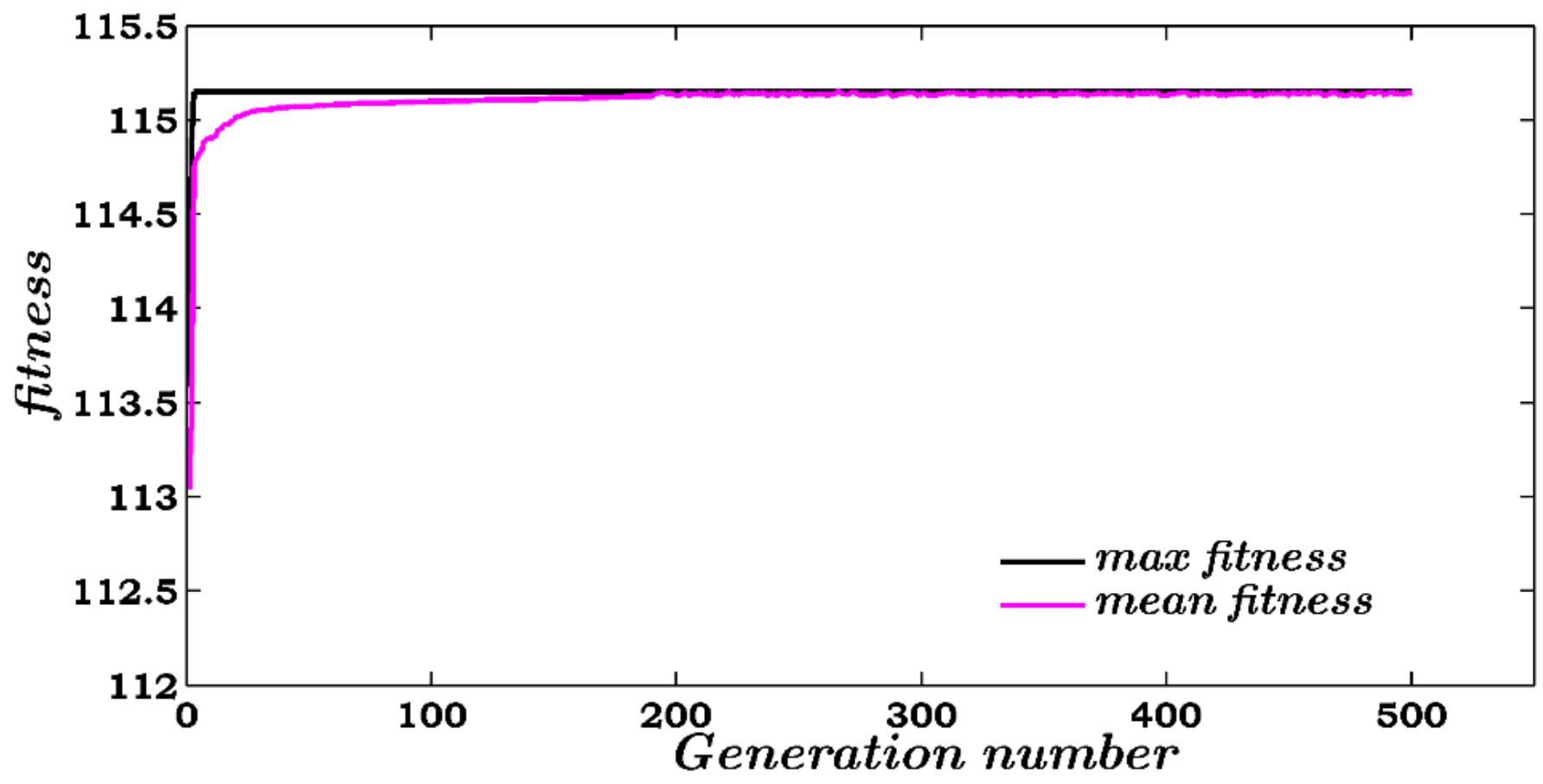

(d)

Figure 3.6 Convergence plot of the proposed $VPGAwIC$ for the random fuzzy maximum flow instances: (a) $mflow_TrRF_10_15$ using model (3.7) (b) $mflow_TrRF_10_15$ using model (3.9) (c) $mflow_GRF_10_15$ using model (3.8) (d) $mflow_GRF_10_15$ using model (3.10)

Table 3.13 Solutions of the model (3.9) at different confidence levels of $Cr$ and $Pr$ for the random fuzzy maximum flow instance $mflow_TrRF_10_15$

| Confidence levels | | CCM (3.9) | | Confidence levels | | CCM (3.9) | |
|---|---|---|---|---|---|---|---|
| $Cr$ | $Pr$ | VPGAwIC | LINGO | $Cr$ | $Pr$ | VPGAwIC | LINGO |
| 0.1 | 0.1003 | 407.71 | 407.71 | 0.5 | 0.6985 | 313.47 | 313.47 |
| | 0.3015 | 389.71 | 389.71 | | 0.8997 | 296.59 | 296.59 |
| | 0.5000 | 377.40 | 377.40 | 0.7 | 0.1003 | 244.69 | 244.69 |
| | 0.6985 | 365.09 | 365.09 | | 0.3015 | 229.06 | 229.06 |
| | 0.8997 | 347.09 | 347.09 | | 0.5000 | 218.20 | 218.20 |
| 0.3 | 0.1003 | 382.61 | 382.61 | | 0.6985 | 207.34 | 207.34 |
| | 0.3015 | 365.73 | 365.73 | | 0.8997 | 188.18 | 188.18 |
| | 0.5000 | 354.20 | 354.20 | 0.9 | 0.1003 | 214.13 | 214.13 |
| | 0.6985 | 342.67 | 342.67 | | 0.3015 | 198.26 | 198.26 |
| | 0.8997 | 325.79 | 325.79 | | 0.5000 | 187.40 | 187.40 |
| 0.5 | 0.1003 | 353.41 | 353.41 | | 0.6985 | 176.54 | 176.54 |
| | 0.3015 | 336.53 | 336.53 | | 0.8997 | 160.18 | 160.18 |
| | 0.5000 | 325.00 | 325.00 | | $--$ | $--$ | $--$ |

Subsequently, we have also performed sensitivity analysis of the models (3.9) and (3.10), respectively for the instances $mflow_TrRF_10_15$ and $mflow_GRF_10_15$, by altering the confidence levels (i.e., $Cr$ and $Pr$). The solutions of these models are then presented, respectively in tables 3.13 and 3.14. Here, for a particular $Cr$, if $Pr$ increases, the solutions of both the models, i.e., (3.9) and (3.10) are decreased. Similar observation is also noticed if $Cr$ increases for a particular $Pr$. Further, the solutions of the models in Table 3.13 and Table 3.14, as generated by LINGO are global optimal, which is the same as generated by the proposed VPGAwIC. Therefore, we can claim that our proposed algorithm has generated the global optimal solutions for the random fuzzy instances.

Table 3.14 Solutions of the model (3.10) at different confidence levels of $Cr$ and $Pr$ for the random fuzzy maximum flow instance $mflow_GRF_10_15$

| Confidence levels | | CCM (3.10) | | Confidence levels | | CCM (3.10) | |
|---|---|---|---|---|---|---|---|
| $Cr$ | $Pr$ | VPGAwIC | LINGO | $Cr$ | $Pr$ | VPGAwIC | LINGO |
| 0.1 | 0.1003 | 296.85 | 296.85 | 0.5 | 0.6985 | 194.84 | 194.84 |
| | 0.3015 | 280.54 | 280.54 | | 0.8997 | 178.53 | 178.53 |
| | 0.5000 | 269.38 | 269.38 | 0.7 | 0.1003 | 197.76 | 197.76 |
| | 0.6985 | 258.22 | 258.22 | | 0.3015 | 181.46 | 181.46 |
| | 0.8997 | 241.91 | 241.91 | | 0.5000 | 170.29 | 170.29 |
| 0.3 | 0.1003 | 269.18 | 269.18 | | 0.6985 | 159.14 | 159.14 |
| | 0.3015 | 252.86 | 252.86 | | 0.8997 | 142.82 | 142.82 |
| | 0.5000 | 241.71 | 241.71 | 0.9 | 0.1003 | 170.09 | 170.09 |
| | 0.6985 | 230.54 | 230.54 | | 0.3015 | 153.78 | 153.78 |
| | 0.8997 | 214.24 | 214.24 | | 0.5000 | 142.62 | 142.62 |
| 0.5 | 0.1003 | 233.47 | 233.47 | | 0.6985 | 131.46 | 131.46 |
| | 0.3015 | 217.15 | 217.15 | | 0.8997 | 115.15 | 115.15 |
| | 0.5000 | 206.00 | 206.00 | | — — | — — | — — |

## 3.6 Conclusion

This chapter proposed a maximum flow problem under random fuzzy environment and defined the EVM and the CCM of the problem, and also proposed their corresponding crisp transformations. A modified GA, VPGAwIC has been developed considering the crossover probability as a function of uncertain variables. Since there is no selection operator in VPGAwIC, we have proposed iLAS which is an improved version of bLAS (Z. Michalewicz 1992). For performance analysis, the results of VPGAwIC are compared with MFGA by considering six maximum flow instances. Here, it is observed that as the instance size increases, VPGAwIC generates better results than its counterpart. However, the computational overhead of the algorithm has been compromised for the instances due to the variation in population size at every subsequent generation with a maximum limit of 5000. VPGAwIC is also used to solve the crisp equivalents of the EVM and the CCM for the proposed random fuzzy maximum flow problem.

Conventional algorithms will always generate optimal solutions. So, the aim of this study is not to compare the proposed VPGAwIC with any conventional algorithms for maximum flow problem. However, many real-life problems can be modelled as different variants of the maximum flow problem (e.g., maximum flow problem with minimum quantities (Haugland et al. 2011, Thielen and Westphal 2013), maximum flow-minimum cost bi-objective problem, etc.), where the complexity of the problem potentially increases due to additional constraints and/or objectives. In those cases, genetic algorithms can be designed/redesigned to generate near-optimal solutions of those problems including their larger instances. However, in this context, conventional algorithms may not be able to generate the solutions.

In future, this study will provide a base while determining efficient solution procedure for maximum-flow based combinatorial optimization problems under uncertain environment. The maximum flow problem can also be extended to maximum flow minimum cost bi-objective problem under uncertain environment, which can be solved by considering a multi-objective version of VPGAwIC in future.

# Part-II

# Uncertain Multi-Objective Optimization Problems

- **Chapter 4**
- **Chapter 5**
- **Chapter 6**

# Chapter 4
# Multi-criteria Shortest Path Problem on Rough Graph

# Chapter 4

# Multi-criteria Shortest Path for Rough Graph

## 4.1 Introduction

In the field of network optimization, shortest path problem (SPP) is essentially one of the frequently encountered fundamental problems. An SPP aims to determine a minimum weighted path between two designated vertices within a network. This concept has numerous applications in various domains, e.g., transportation, scheduling, routing, communication, etc. As a result, SPP has been evolved as an area of growing interest for many researchers. In the context of single objective network optimization, several popular and efficient algorithms are available on SSP (R. Bellman 1958; E.W. Dijkstra 1959; S.E. Dreyfus 1969; D.E. Knuth 1977; Ahuja et al. 1993). Considering the fluid-dynamic approach to modelling and optimization, Cascone et al. (2010) focuses the most suitable research of packet routing to optimize the flow of data traffic on some telecommunication networks. Further, Cutolo et al. (2012) implemented a variant of $k$-shortest path algorithm on a fluid-dynamic model of a road network. Considering the dynamical weights generated by the fluid-dynamic model, the authors represent the evaluation of traffic inside road networks. Based on the fluid-dynamic model, optimal trajectories of emergency vehicles have been studied by Manzo et al. (2012), for analysing the critical situations like road accidents in congested and urban road networks.

In real-life scenarios, one needs to simultaneously consider several criteria including distance, time, cost, capacity, demand, etc., while exploring the shortest path in a network. As a result, there has been a concern among researchers to solve multi-criteria shortest path problems (MSPPs). Among the existing contributions on MSPP, P. Hansen (1980) first introduced a study on bi-objective SPP. Thereafter, several contributions on MSPP (Kostreva and Wiecek 1993; Papadimitriou and Yannakakis 2000; Gandibleux et al. 2006; Chen and Nie 2013; Sedeño-Noda and Raith 2015) are observed in the literature.

The researchers of all the above mentioned studies have addressed the problem under the deterministic environment, where the relevant problem parameters (distance, time, cost, capacity, etc.) have specific values. However, due to the complexities in socio-economic conditions including inadequate information, lack of evidence, multiple input sources, fluctuating nature of input parameters and measurement inaccuracy, there exists imprecision (i.e., uncertainty) among these parameter values. In order to process and represent those data hidden within the imprecise or ill-defined information, many researchers have proposed several improved theories like fuzzy set (L.A. Zadeh 1965), type-2 fuzzy set (L.A. Zadeh 1975a, b), rough set (Z. Pawlak 1982) and uncertainty theory (B. Liu 2007). The fuzzy shortest path problem was first introduced by Dubois and Prade (1980). Since then, there have been very few researches in the field of multi-objective fuzzy SPP. Mahdavi et al. (2011) proposed two algorithms and a dynamic programming model to solve the bi-objective fuzzy shortest path problem. Kumar and Sastry (2013) presented an algorithm to compute Pareto optimal paths of a fuzzy bi-objective shortest path problem.

Since the inception of rough set theory (Z. Pawlak 1982), many researchers (Z. Pawlak 1991; L. Polkowski 2002; Pawlak and Skowron 2007; W. Zhu 2009) contributed to its theoretical development, and used it to several application domains like data envelopment analysis (Xu et al. 2009; Shafiee and Shams-e-alam 2011), multi-criteria decision analysis (Dembczyński et al. 2009) and medical diagnosis (Hirano and Tsumoto 2005; S. Tsumoto 2007). Besides those contributions, one of the significant studies, conducted by Słowiński and Vanderpooten (2000), is related to the idea of modifying the indiscernibility relation in rough set theory (Z. Pawlak 1982), where the authors presented the concept of a binary similarity relation which is reflexive, non-symmetric and non-transitive. Enlightened by the study conducted by Słowiński and Vanderpooten (2000), B. Liu (2002) conceptually proposed rough space and eventually defined rough variable as a function which maps rough space to a set of real numbers. Subsequently, B. Liu (2004) presented the axiomatic definition of rough variable. Moreover, B. Liu (2002, 2004) showed that the arithmetic operations on rough variables coincide with interval arithmetic defined by Alefeld and Herzberger (1983), and E. Hansen (1992). As an example of a rough variable, let us consider the travelling cost $(c_{sd})$ between two cities $s$ and $d$ is estimated by four experts, which are respectively represented as intervals of $[25, 40.8]$, $[30.2, 42]$, $[20.5, 50.5]$ and $[27, 45]$. From such opinion of the experts, it is clear that every element within the interval $\underline{c}_{sd}$ = [30.2 40.8] is the member of all four different intervals as defined by the experts. Hence the elements of $\underline{c}_{sd}$ are certainly regarded as the lower approximation of $c_{sd}$. Again the elements of the interval $\overline{c}_{sd}$=$[20.5, 50.5]$, partly belong to the intervals defined by experts, and hence are considered as the upper approximation of $c_{sd}$. Then, $c_{sd}$ is expressed as a rough variable $[30.2, 40.8]$ $[\ 20.5, 50.5]$.

Chance-constrained programming (CCP) is an approach to formulate optimization problems under uncertain environment, where constraint set is assumed to be imprecise. In other words, the constraint set of CCP does not define a deterministic feasible set. However, for a given confidence level of a decision maker, the constraints of CCP should necessarily hold. Charnes and Cooper (1959) first presented CCP with stochastic constraints. Later, motivated by the study of Liu and Iwamura (1998) which extended the CCP to a fuzzy environment, Yang and Liu (2007) applied CCP on fuzzy solid transportation problem. Besides, Kundu et al. (2013a, 2017b) employed CCP to formulate the solid transportation problem with some input parameters represented by rough variable.

Considering all the above mentioned studies, to the best of our knowledge, a study on shortest path problem with multiple objectives, for a weighted connected directed network (WCDN) under rough environment, is yet to be conducted. In this study, we consider a WCDN with multiple criteria which are represented by rough variables (cf. Example A.1 of Appendix A). The main contribution of this chapter are mentioned below.

(i) A modified rough Dijkstra's (MRD) algorithm is proposed to explore the shortest path of a WCDN with respect to individual criterion as well as a compromise shortest path, which ensembles the criteria of the network.

(ii) A rough chance-constrained model of MSPP has been proposed which is eventually solved by

a. Goal attainment method (F.W. Gembicki 1974)
b. Nondominated sorting genetic algorithm II (NSGA-II) (Deb et al. 2002)

The weight vectors considered in the MRD algorithm and the goal attainment method, reflect the DM's preference for the criteria. Here, the weight vectors are calculated using the argument-dependent ordered weighted averaging $(OWA)$ operator proposed by Z. Xu (2006).

The remaining part of the chapter is organized as follows. The proposed MRD algorithm is presented in Section 4.2 followed by Section 4.3, where the model of MSPP is formulated using rough chance-constrained programming technique. Moreover, in section 4.3, we have discussed the implementation of goal attainment method to solve the models of MSPP. In Section 4.4, we provide the numerical illustration of the results for the proposed model of MSPP with rough parameters, using the methodologies discussed in sections 4.2 and 4.3, and by NSGA-II (cf. Section 1.3.16.1). The simulated results of MRD algorithm and NSGA-II are analyzed in Section 4.5. Finally, we conclude our study in Section 4.6.

## 4.2 Proposed Modified Rough Dijkstra's (MRD) Algorithm

We consider a WCDN, having source $(s)$ and sink $(t)$ vertices with all its edges associated with three different uncertain criteria, namely distance, cost and time. These uncertain criteria are represented by rough variable. We apply our proposed MRD algorithm on the WCDN and explore the pessimistic and optimistic shortest paths corresponding to each criterion as well as the compromise pessimistic and optimistic shortest path by considering all the criteria simultaneously. The determination of shortest paths is done at two different trust labels $Tm_1$ and $Tm_2$ such that, $0 < Tm_1 \leq 0.5$ and $0.5 < Tm_2 \leq 1.0$. This is due to the fact that, for any rough variable, its pessimistic value is less or equal to its optimistic value if the trust label is set within (0,0.5]. Further, if the trust label is within (0.5,1.0], then for a rough variable, its pessimistic value is greater or equal to its optimistic value. Here, for every edge of a WCDN, we determine the argument-dependent OWA weights (Z. Xu 2006) with respect to three uncertain criteria to explore the compromise pessimistic and optimistic shortest paths.

Let $G(V_G, E_G)$ be a WCDN, where $V_G$ and $E_G$ are the set of vertices and edges of $G$. The weights of each edge $e$ ($e \in E_G$) connecting vertices $v$ and $v'$ are associated with three uncertain criteria: distance ($P^1$), cost ($P^2$) and time ($P^3$) which are expressed as rough variables, where $P^k = \left([P_2^k, P_3^k], [P_1^k, P_4^k]\right), P_1^k \leq P_2^k < P_3^k \leq P_4^k \quad \forall k \in \{1,2,3\}$. For an edge $e$, each $P_j^k$ determines an extreme bound of an approximation set of the representing rough variable of $k^{th}$ criterion. Since the associated weights of $G$ are represented by rough variable, these weights are also termed as rough weights. Here, if $j = 2,3 \; \forall k \in \{1,2,3\}$, then $P_j^k$ determines an extreme bound of the lower approximation set of $P^k$. Again, if $j = 1,4 \; \forall k \in \{1,2,3\}$, then $P_j^k$ determines an extreme bound of the upper approximation set of $P^k$. The steps of the MRD algorithm are outlined below.

**Input**: A WCDN, $G$ with all the edges associated with three criteria, $P^1, P^2$ and $P^3$ which are represented by rough variables.

**Output**:

(i) Pessimistic and optimistic shortest paths and their corresponding path weights from source $(s)$ to sink $(t)$ vertices with respect to each $P^k$

(ii) Compromise pessimistic and optimistic shortest paths with their effective weights.

at trust measures, $Tm_1$and $Tm_2$.

**Step 1**: Set the values of the trust measure $Tm_1(\in (0,0.5])$ and $Tm_2(\in (0.5,1.0])$ for the rough weights associated with $G$.

**Step 2**: Set the label $Lb(s)$ of the source vertex $s$ such that

$$Lb(s)=\begin{bmatrix}\left(0^{P^1}_{pessi}, \; s, \; 0^{P^2}_{pessi}, \; s, \; 0^{P^3}_{pessi}, \; s, \; 0^{Comp}_{pessi}, \; s\right)^{Tm_l}, \\ \left(0^{P^1}_{opti}, \; s, \; 0^{P^2}_{opti}, \; s, \; 0^{P^3}_{opti}, \; s, \; 0^{Comp}_{opti}, \; s\right)^{Tm_l}\end{bmatrix},$$

where $0^{P^k}_{pessi}$ and $0^{P^k}_{opti}$ determine the pessimistic and optimistic values of criterion $P^k$ for a trust measure, $Tm_l \in \{Tm_1, Tm_2\} \; \forall k \in \{1,2,3\}$. Since $s$ is the source vertex, then $s$ itself becomes its predecessor vertex, and the pessimistic and optimistic shortest paths from the current vertex to its predecessor vertex, i.e., $s$, is set to $0^{Pk}_{pessi}$ and $0^{Pk}_{opti}$, respectively $\forall k \in \{1,2,3\}$.

Set the labels of the remaining vertices of $G$ as

$Lb(v_i)$=$\begin{bmatrix} \left(\infty^{P1}_{pessi}, __, \infty^{P2}_{pessi}, __, \infty^{P3}_{pessi}, __, \infty^{Comp}_{pessi}, __\right)^{Tm_l}, \\ \left(\infty^{P1}_{opti}, __, \infty^{P2}_{opti}, __, \infty^{P3}_{opti}, __, \infty^{Comp}_{opti}, __\right)^{Tm_l} \end{bmatrix}$, $\forall v_i \in V_G \backslash \{s\}$ and $Tm_l \in \{Tm_1, Tm_2\}$.

Initially, since the predecessor vertices of all $v_i$'s are unknown, therefore

- The shortest distance from each $v_i$ to its immediate predecessor vertex is set as $\infty^{Pk}_{pessi}$ and $\infty^{Pk}_{opti}$ for pessimistic and optimistic shortest paths, respectively $\forall k \in \{1,2,3\}$.
- The symbol '__' in $Lb(v_i)$ is used to designate the unknown predecessor vertex of $v_i$.

**Step 3**: For a vertex $v$, find its successor vertex set $Sr(v)$ and the set of edges $E_{v\,v'}$, $\forall v' \in Sr(v)$, where $Sr(v) \subseteq V_G$ and $E_{v\,v'} \subseteq E_G$.

**Step 4**: For a trust label $Tm_l (\in \{Tm_1, Tm_2\})$, consider each edge $e \in E_{v\,v'}$ to calculate the $Tm_l$-pessimistic $\left(\{e^{P1}_{pessi}, e^{P2}_{pessi}, e^{P3}_{pessi}\}\right)$ and $Tm_l$-optimistic $\left(\{e^{P1}_{opti}, e^{P2}_{opti}, e^{P3}_{opti}\}\right)$ values corresponding to its associated edge weight $e^{P^k}$ $(k = \{1,2,3\})$ using Step 4.1 and Step 4.2, respectively.

**Step 4.1**:

$$e_{inf}(Tm_l)=\begin{cases} (1-2Tm_l)e^{P^k_1} + 2Tm_l e^{P^k_4} & ; if\ Tm_l \le \left(\dfrac{\left(e^{P^k_2}-e^{P^k_1}\right)}{2\left(e^{P^k_4}-e^{P^k_1}\right)}\right) \\ 2(1-Tm_l)e^{P^k_1} + (2Tm_l-1)e^{P^k_4} & ; if\ Tm_l \ge \left(\dfrac{\left(e^{P^k_3}+e^{P^k_4}-2e^{P^k_1}\right)}{2\left(e^{P^k_4}-e^{P^k_1}\right)}\right) \\ \left(\dfrac{e^{P^k_1}\left(e^{P^k_3}-e^{P^k_2}\right)+e^{P^k_2}\left(e^{P^k_4}-e^{P^k_1}\right)+2Tm_l\left(e^{P^k_3}-e^{P^k_2}\right)\left(e^{P^k_4}-e^{P^k_1}\right)}{\left(e^{P^k_3}-e^{P^k_2}\right)+\left(e^{P^k_4}-e^{P^k_1}\right)}\right) & ; otherwise \end{cases}$$

**Step 4.2**:

$$e_{sup}(Tm_l)=\begin{cases} (1-2Tm_l)e^{P^k_4} + 2Tm_l e^{P^k_1} & ; if\ Tm_l \le \left(\dfrac{\left(e^{P^k_4}-e^{P^k_3}\right)}{2\left(e^{P^k_4}-e^{P^k_1}\right)}\right) \\ 2(1-Tm_l)e^{P^k_4} + (2Tm_l-1)e^{P^k_1} & ; if\ Tm_l \ge \left(\dfrac{\left(2e^{P^k_4}-e^{P^k_2}-e^{P^k_1}\right)}{2\left(e^{P^k_4}-e^{P^k_1}\right)}\right) \\ \left(\dfrac{e^{P^k_4}\left(e^{P^k_3}-e^{P^k_2}\right)+e^{P^k_3}\left(e^{P^k_4}-e^{P^k_1}\right)-2Tm_l\left(e^{P^k_3}-e^{P^k_2}\right)\left(e^{P^k_4}-e^{P^k_1}\right)}{\left(e^{P^k_3}-e^{P^k_2}\right)+\left(e^{P^k_4}-e^{P^k_1}\right)}\right) & ; otherwise \end{cases}$$

$\forall k \in \{1,2,3\}$ such that, $e_{inf}(Tm_l) \in \{e_{pessi}^{P^1}, e_{pessi}^{P^2}, e_{pessi}^{P^3}\}$ and $e_{sup}(Tm_l) \in \{e_{opti}^{P^1}, e_{opti}^{P^2}, e_{opti}^{P^3}\}$.

**Step 5**: For each edge $e \in E_{v\,v'}$ consider in Step 4, we calculate the compromise weights, $e_{pessi}^{Comp}$ and $e_{opti}^{Comp}$ for each $Tm_l$ by using argument-dependent OWA weights (Z. Xu 2006) (cf. Section 1.3.12 and Example 1.3.15), following Step 5.1 through Step 5.5.

**Step 5.1**: Calculate the means, $\mu_{pessi}$ and $\mu_{opti}$ corresponding to pessimistic $\left(\{e_{pessi}^{P^1}, e_{pessi}^{P^2}, e_{pessi}^{P^3}\}\right)$ and optimistic $\left(\{e_{opti}^{P^1}, e_{opti}^{P^2}, e_{opti}^{P^3}\}\right)$ edge weights.

**Step 5.2**: Considering the argument lists, $\{e_{pessi}^{P^1}, e_{pessi}^{P^2}, e_{pessi}^{P^3}\}$ and $\{e_{opti}^{P^1}, e_{opti}^{P^2}, e_{opti}^{P^3}\}$, determine a permutation sequence, $(\pi(1), \pi(2), \pi(3))$ of (1,2,3) such that $\left(e_{\mathcal{J}}^{P^k}\right)_{\pi(i-1)} \geq \left(e_{\mathcal{J}}^{P^k}\right)_{\pi(i)}$, $i = 2, 3$ and calculate the similarity degree $s\left(\left(e_{\mathcal{J}}^{P^k}\right)_{\pi(i)}, \mu_{\mathcal{J}}\right)$ of $i^{th}$ largest $\left(e_{\mathcal{J}}^{P^k}\right)_{\pi(i)}$,where $s\left(\left(e_{\mathcal{J}}^{P^k}\right)_{\pi(i)}, \mu_{\mathcal{J}}\right) = 1 - \frac{\left|\left(e_{\mathcal{J}}^{P^k}\right)_{\pi(i)} - \mu_{\mathcal{J}}\right|}{\sum_{i=1}^{3}\left(\left|e_{\mathcal{J}}^{P^i} - \mu_{\mathcal{J}}\right|\right)}$, $\mathcal{J} \in \{pessi, opti\}$ and $\forall k \in \{1,2,3\}$.

**Step 5.3**: Calculate the pessimistic and optimistic weight vectors, $\left(\omega_{\mathcal{J}} = \left[(\omega_1)_{\mathcal{J}}, (\omega_2)_{\mathcal{J}}, (\omega_3)_{\mathcal{J}}\right]^T\right)$ for $\mathcal{J} \in \{pessi, opti\}$, such that

$$(\omega_j)_{\mathcal{J}} = \frac{s\left(\left(e_{\mathcal{J}}^{P^k}\right)_{\pi(j)}, \mu_{\mathcal{J}}\right)}{\sum_{j=1}^{3} s\left(\left(e_{\mathcal{J}}^{P^k}\right)_{\pi(j)}, \mu_{\mathcal{J}}\right)}, (\omega_j)_{\mathcal{J}} \in [0,1], \forall k \in \{1,2,3\}.$$

**Step 5.4**: Evaluate the compromise pessimistic and optimistic edge weights, $\mathrm{OWA}\left(e_{\mathcal{J}}^{P^1}, e_{\mathcal{J}}^{P^2}, e_{\mathcal{J}}^{P^3}\right)_{\mathcal{J}}$, where $\mathrm{OWA}\left(e_{\mathcal{J}}^{P^1}, e_{\mathcal{J}}^{P^2}, e_{\mathcal{J}}^{P^3}\right)_{\mathcal{J}} = \sum_{j=1}^{3} (\omega_j)_{\mathcal{J}} \times \left(e_{\mathcal{J}}^{P^k}\right)_{\pi(j)}$, $\mathcal{J} \in \{pessi, opti\}$ and $\forall k \in \{1,2,3\}$.

**Step 5.5**: Set $e_{\mathcal{J}}^{Comp} = \mathrm{OWA}\left(e_{\mathcal{J}}^{P^1}, e_{\mathcal{J}}^{P^2}, e_{\mathcal{J}}^{P^3}\right)_{\mathcal{J}}$, $\mathcal{J} \in \{pessi, opti\}$.

**Step 6**: For each edge $e \in E_{v\,v'}$ the label of $v$ is determined as

$Lb(v) = \begin{bmatrix} \left(v_{pessi}^{P^1},\ q_1,\ v_{pessi}^{P^2},\ q_2,\ v_{pessi}^{P^3},\ q_3,\ v_{pessi}^{Comp},\ q_{Comp}\right)^{Tm_l}, \\ \left(v_{opti}^{P^1},\ r_1,\ v_{opti}^{P^2},\ r_2,\ v_{opti}^{P^3},\ r_3,\ v_{opti}^{Comp},\ r_{Comp}\right)^{Tm_l} \end{bmatrix}$, where $v_{pessi}^{P^1}, v_{pessi}^{P^2}$ and $v_{pessi}^{Comp}$ are the pessimistic path lengths from $s$ to $v$ and the corresponding paths incorporate $q_1, q_2, q_3$ and $q_{Comp}$ as the immediate predecessors of $v$ for the trust measure $Tm_l$. Similarly, $v_{opti}^{P^1}, v_{opti}^{P^2}, v_{opti}^{P^3}$ and $v_{opti}^{Comp}$ are the optimistic path lengths from $s$ to $v$ and $r_1, r_2, r_3$ and $r_{Comp}$ are the immediate predecessors of $v$ for a trust $Tm_l$.

If the label of $v'$ is set as

$$Lb(v')=\begin{bmatrix}\left(v'^{P^1}_{pessi},\ q'_1,\ v'^{P^2}_{pessi},\ q'_2,\ v'^{P^3}_{pessi},\ q'_3,\ v'^{Comp}_{pessi},\ q'_{Comp}\right)^{Tm_l}, \\ \left(v'^{P^1}_{opti},\ r'_1,\ v'^{P^2}_{opti},\ r'_2,\ v'^{P^3}_{opti},\ r'_3,\ v'^{Comp}_{opti},\ r'_{Comp}\right)^{Tm_l}\end{bmatrix}$$

then considering $v$ as the immediate predecessor of $v'$, we re-label $v'$as $Lb(v')$, where

$$Lb(v')=\begin{bmatrix}\left(v''^{P^1}_{pessi},\ q''_1,\ v''^{P^2}_{pessi},\ q''_2,\ v''^{P^3}_{pessi},\ q''_3,\ v''^{Comp}_{pessi},\ q''_{Comp}\right)^{Tm_l}, \\ \left(v''^{P^1}_{opti},\ r''_1,\ v''^{P^2}_{opti},\ r''_2,\ v''^{P^3}_{opti},\ r''_3,\ v''^{Comp}_{opti},\ r''_{Comp}\right)^{Tm_l}\end{bmatrix}$$

such that

$$if\ \ minimum\left(\left(v^{\wp}_{\mathcal{J}}+e^{\wp}_{\mathcal{J}}\right)^{Tm_l},\left(v'^{\wp}_{\mathcal{J}}\right)^{Tm_l}\right)=\left(v^{\wp}_{\mathcal{J}}+e^{\wp}_{\mathcal{J}}\right)^{Tm_l}$$

$then\ \left(v''^{\wp}_{\mathcal{J}}\right)^{Tm_l}=\left(v^{\wp}_{\mathcal{J}}+e^{\wp}_{\mathcal{J}}\right)^{Tm_l}$ and $V''_i=v$

$else$

$\left(v''^{\wp}_{\mathcal{J}}\right)^{Tm_l}=\left(v'^{\wp}_{\mathcal{J}}\right)^{Tm_l}$and $V''_i=\ V'_i$,

where $\wp\in\{P^1,P^2,P^3,Comp\}$, $V''_i\in\left\{q''_i,q''_{Comp},r''_i,r''_{Comp}\right\}$, $Tm_l\in\{Tm_1,Tm_2\}$, $V''_i\in\left\{q'_i,q'_{Comp},r'_i,r'_{Comp}\right\}$, $i=1,2,3$, $e^{\wp}_{\mathcal{J}}$ belongs to pessimistic and optimistic values with respect to the elements of $\wp$ associated with the edge connecting $v$ and $v'$, and $\mathcal{J}\in\{pessi,opti\}$.

**Step 7**: Repeat Step 3 through Step 6 until the sink vertex $t$ is reached with no successor vertex set, i.e., $|Sr(v)|=\emptyset$, where $v\in V_G$.

**Step 8**: Construct $\left[Path^{\wp}_{\mathcal{J}}\right]^{Tm_l}$ paths and their respective path weights by backtracking from $t$ to $s$, where $\wp\in\{P^1,P^2,P^3,Comp\}$, $\mathcal{J}\in\{pessi,opti\}$ and $Tm_l\in\{Tm_1,Tm_2\}$.

## 4.3 Proposed Multi-criteria Rough Chance-constrained Shortest Path Problem (MRCCSPP)

Chance-constrained programming technique provides the flexibility of modelling uncertain decision systems with uncertain constraints sets by assuming the constraints are satisfied at some confidence levels. The confidence level provided by a DM is considered as an appropriate safety margin. In this context, Charnes and Cooper (1959) first introduced the concept of chance-constrained programming under stochastic environment. Thereafter, various applications (Liu and Iwamura 1998; Yang and Liu 2007; Kundu et al. 2013a) of chance-constrained programming are observed in the literature. However, considering the existing works on chance-constrained programming with rough parameters, it has been found that there is no investigation done on applying rough chance-constrained programming technique on multi-objective shortest path problems. Therefore, in this section, we have used rough chance-constrained programming (RCCP) approach to model an MSPP.

We consider a WCDN, $G(V_G, E_G)$, where $V_G$ and $E_G$ are respectively, the set of vertices and edges of $G$. An edge $e_{ij} \in E_G$, joining a pair of vertices, $(v_i,\ v_j)$ is associated with three criteria: distance $(P_{ij}^1)$, cost $(P_{ij}^2)$ and time $(P_{ij}^3)$. Considering these criteria as rough variables, a model of MSPP is formulated in $M_1$.

$M_1$:

$$\begin{cases} Min\ Z_1 = \sum_{e_{ij}\in E_G} P_{ij}^1 x_{ij} \\ Min\ Z_2 = \sum_{e_{ij}\in E_G} P_{ij}^2 x_{ij} \\ Min\ Z_3 = \sum_{e_{ij}\in E_G} P_{ij}^3 x_{ij} \\ subject\ to \\ \sum_{e_{ij}\in E_G} x_{ij} - \sum_{e_{ij}\in E_G} x_{ij} = \begin{cases} 1 & ; if\ v_i = s \\ 0 & ; if\ v_i \notin \{s,t\} \\ -1 & ; if\ v_i = t \end{cases} \\ x_{ij} = \begin{cases} 1 & ; if\ e_{ij} \in E_G \\ 0 & ; otherwise, \end{cases} \end{cases} \tag{4.1}$$

where $x_{ij}$ is the decision variable, and $s$ and $t$ are respectively, the source and the sink vertices of a WCDN. Here, each $P_{ij}^k$ is expressed as $\left([P_{ij}^{k2},\ P_{ij}^{k3}], [P_{ij}^{k1},\ P_{ij}^{k4}]\right)$ such that $P_{ij}^{k1} \le P_{ij}^{k2} < P_{ij}^{k3} \le P_{ij}^{k4}, k = 1,2,3$. Since all the criteria: $P_{ij}^1,\ P_{ij}^2$ and $P_{ij}^3$ are considered as rough variables, the objective functions of $M_1$ shown in (4.1) will also become rough variables, i.e., $Z_k = \left([Z_{pk}^2,\ Z_{pk}^3], [Z_{pk}^1,\ Z_{pk}^4]\right), \forall k \in \{1, 2, 3\}$. The pessimistic and optimistic values of the rough variables are calculated at trust measures $Tm_1 (\in (0.0,0.5])$ and $Tm_2 (\in (0.5,1.0])$ while solving the models of MRCCSPP. The algorithmic steps to model and solve MRCCSPP by goal attainment method are presented below.

**Input**: A WCDN, $G(V_G, E_G)$, where $V_G$ and $E_G$ are the set of vertices and edges such that for each $v_i, v_j \in V_G$ the weights of each edge $e_{ij} \in E_G$ are represented by three criteria: $P_{i,j}^1, P_{i,j}^2$ and $P_{i,j}^3$, which are expressed as rough variables, i.e., $P_{ij}^k = \left([P_{ij}^{k2}, P_{ij}^{k3}], [P_{ij}^{k1}, P_{ij}^{k4}]\right), P_{ij}^{k1} \le P_{ij}^{k2} < P_{ij}^{k3} \le P_{ij}^{k4}, \forall k = 1,2,3.$

**Output**: Pessimistic and optimistic shortest paths and their corresponding effective weights with respect to each criterion $P^1, P^2$ and $P^3$ of $G$, and the compromise pessimistic and optimistic shortest paths with their corresponding path weights with respect to all the criteria.

**Step 1**: Set the trust measure $Tm$=$Tm_1$, where $Tm_1 \in (0.0,0.5]$.

**Step 2**: For each criterion $P^k$, we formulate and solve the models $M_{pk}^r,\ r = 1,2,3,4$, $k = 1,2,3$ as

$M^r_{pk}$:

$$\begin{cases} Min\ Z^r_{pk} = \sum_{e_{ij}\in E_G} {P_{ij}}^{kr} x_{ij} \\ subject\ to \\ constraints\ of\ (4.1) \\ k = 1,2,3, r = 1,2,3,4. \end{cases} \tag{4.2}$$

**Step 3**: To find the pessimistic shortest paths and their corresponding path weights, formulate a rough chance-constrained model of $M_1$ to minimize the smallest objectives $\overline{Z_1}$, $\overline{Z_2}$ and $\overline{Z_3}$ with respect to $P^1, P^2$ and $P^3$ satisfying $Tr\{Z_k \le \overline{Z_k}\} \ge Tm$, where the value of $Tm$ is the predetermined confidence level (trust level) set by the decision maker. Accordingly, the following model minimizes the $Tm$-pessimistic value, $(Z_k)^{Tm}_{pessi}$ of $Z_k, k = 1,2,3$.

$M_{k+1}$:

$$\begin{cases} Min(Min\ \overline{Z_k}) \\ subject\ to \\ Tr\{Z_k \le \overline{Z_k}\} \ge Tm \\ with\ constraints\ of\ (4.1)\ \forall\ k \in \{1,2,3\}. \end{cases} \tag{4.3}$$

**Step 4**: Following the definition of $Tm$-pessimistic value of a rough variable (cf. Definition 1.3.21), the crisp equivalent of the rough CCM of Step 3 becomes

$M_{k+4}$:

$$\begin{cases} Min\left((Z_k)^{Tm}_{pessi}\right) = \begin{cases} (1-2Tm)Z^1_{pk} + 2Tm\, Z^4_{pk} & ; if\ Tm \le \left(\frac{Z^2_{pk} - Z^1_{pk}}{2\left(Z^4_{pk} - Z^1_{pk}\right)}\right) \\ 2(1-Tm)Z^1_{pk} + (2Tm-1)Z^4_{pk} & ; if\ Tm \ge \left(\frac{Z^3_{pk} + Z^4_{pk} - 2Z^1_{pk}}{2\left(Z^4_{pk} - Z^1_{pk}\right)}\right) \\ \frac{Z^1_{pk}\left(Z^3_{pk} - Z^2_{pk}\right) + Z^2_{pk}\left(Z^4_{pk} - Z^1_{pk}\right) + \left(2Tm\left(Z^3_{pk} - Z^2_{pk}\right) \times \left(Z^4_{pk} - Z^1_{pk}\right)\right)}{\left(Z^3_{pk} - Z^2_{pk}\right) + \left(Z^4_{pk} - Z^1_{pk}\right)} & ; otherwise \end{cases} \\ subject\ to \\ the\ constraints\ of\ (4.1)\ \forall\ k \in \{1,2,3\}. \end{cases} \tag{4.4}$$

**Step 5**: To find the optimistic shortest paths and their corresponding path weights, formulate a rough chance-constrained model of $M_1$ to minimize the greatest objectives $\underline{Z_1}$, $\underline{Z_2}$ and $\underline{Z_3}$ with respect to $P^1$, $P^2$ and $P^3$ satisfying $Tr\left\{Z_k \ge \underline{Z_k}\right\} \ge Tm$ at trust level, $Tm$, i.e., we minimize the $Tm$-optimistic value, $(Z_k)^{Tm}_{opti}$ of $Z_k, k = 1,2,3$.

$M_{k+7}$:

$$\begin{cases} Min\left(Max\ \underline{Z_k}\right) \\ subject\ to \\ Tr\left\{Z_k \ge \underline{Z_k}\right\} \ge Tm \\ with\ the\ constraints\ of\ (4.1)\ \forall\ k \in \{1,2,3\}. \end{cases} \tag{4.5}$$

**Step 6**: From the definition of $Tm$-optimistic value (cf. Definition 1.3.21), the crisp equivalent of the rough chance-constrained model of Step 5 becomes

$M_{k+10}$:

$$\begin{cases} Min\left((Z_k)_{opti}^{Tm}\right) = \begin{cases} (1-2Tm)Z_{pk}^4 + 2Tm\, Z_{pk}^1 \quad ; if\ Tm \le \left(\frac{Z_{pk}^4 - Z_{pk}^3}{2\left(Z_{pk}^4 - Z_{pk}^1\right)}\right) \\ 2(1-Tm)Z_{pk}^4 + (2Tm-1)Z_{pk}^1 \quad ; if\ Tm \ge \left(\frac{2Z_{pk}^4 - Z_{pk}^2 - Z_{pk}^1}{2\left(Z_{pk}^4 - Z_{pk}^1\right)}\right) \\ \frac{Z_{pk}^4\left(Z_{pk}^3 - Z_{pk}^2\right) + Z_{pk}^3\left(Z_{pk}^4 - Z_{pk}^1\right) - \left(2Tm\left(Z_{pk}^3 - Z_{pk}^2\right)\times\left(Z_{pk}^4 - Z_{pk}^1\right)\right)}{\left(Z_{pk}^3 - Z_{pk}^2\right) + \left(Z_{pk}^4 - Z_{pk}^1\right)} \quad ; otherwise \end{cases} \\ subject\ to \\ \quad constraints\ of\ (4.1)\ \forall\ k \in \{1,2,3\}. \end{cases} \tag{4.6}$$

**Step 7**: Calculate the argument-dependent OWA weights (Z. Xu 2006) (cf. Section 1.3.12 and Example 1.3.15) of $Tm$-pessimistic and $Tm$-optimistic objective vectors using Step 7.1 through Step 7.4.

**Step 7.1**: For $Tm$-pessimistic and $Tm$-optimistic objective vectors calculate the means, $\mu_{pessi}^{Tm}$ and $\mu_{opti}^{Tm}$ for the objective vectors, $\gamma_{pessi}= \left[(Z_1)_{pessi}^{Tm},\ (Z_2)_{pessi}^{Tm},\ (Z_3)_{pessi}^{Tm}\right]$ and $\gamma_{opti}= \left[(Z_1)_{opti}^{Tm},\ (Z_2)_{opti}^{Tm},\ (Z_3)_{opti}^{Tm}\right]$, respectively, where $\mu_{pessi}^{Tm}= \frac{1}{3}\sum_{k=1}^{3}(Z_k)_{pessi}^{Tm}$ and $\mu_{opti}^{Tm}= \frac{1}{3}\sum_{k=1}^{3}(Z_k)_{opti}^{Tm}$.

**Step 7.2**: Let $\{\varphi(1), \varphi(2), \varphi(3)\}$ be a permutation of $(1,2,3)$ such that $\gamma'_{\mathcal{J}} = \left[\left(Z_{\varphi(1)}\right)_{\mathcal{J}}^{Tm}, \left(Z_{\varphi(2)}\right)_{\mathcal{J}}^{Tm}, \left(Z_{\varphi(3)}\right)_{\mathcal{J}}^{Tm}\right]$ be a permutation of $\gamma_{\mathcal{J}}$, where $\left(Z_{\varphi(t-1)}\right)_{\mathcal{J}}^{Tm} \ge \left(Z_{\varphi(t)}\right)_{\mathcal{J}}^{Tm} \forall\ t \in \{2,3\}$ and $\mathcal{J} \in \{pessi, opti\}$.

**Step 7.3**: Calculate the similarity degree, $s\left(\left(Z_{\varphi(k)}\right)_{\mathcal{J}}^{Tm}, \mu_{\mathcal{J}}^{Tm}\right)$ with respect to $k^{th}$ largest objective of $\gamma'_{\mathcal{J}}$ and the corresponding mean, so that $s\left(\left(Z_{\varphi(k)}\right)_{\mathcal{J}}^{Tm}, \mu_{\mathcal{J}}^{Tm}\right) = 1 - \frac{\left|\left(\left(Z_{\varphi(k)}\right)_{pessi}^{Tm} - \mu_{\mathcal{J}}^{Tm}\right)\right|}{\sum_{k=1}^{3}\left|\left((Z_k)_{\mathcal{J}}^{Tm} - \mu_{\mathcal{J}}^{Tm}\right)\right|}$ $\forall\ k \in \{1,2,3\}$ and $\mathcal{J} \in \{pessi, opti\}$.

**Step 7.4**: Determine the corresponding pessimistic and optimistic weight vectors, $\left(\omega_{\mathcal{J}} = \left[(\omega_1)_{\mathcal{J}}, (\omega_2)_{\mathcal{J}}, (\omega_3)_{\mathcal{J}}\right]^T\right)$ such that, $(\omega_k)_{\mathcal{J}} = \frac{s\left(\left(Z_{\varphi(k)}\right)_{\mathcal{J}}^{Tm},\ \mu_{\mathcal{J}}^{Tm}\right)}{\sum_{k=1}^{3} s\left(\left(Z_{\varphi(k)}\right)_{\mathcal{J}}^{Tm},\ \mu_{\mathcal{J}}^{Tm}\right)}$, $k = 1,2,3$, where $(\omega_k)_{\mathcal{J}} \in [0,1]$ and $\mathcal{J} \in \{pessi, opti\}$.

**Step 8**: For the weight vectors, $\omega_{pessi}$ and $\omega_{opti}$ the pessimistic and optimistic compromise models of $M_{k+4}$ and $M_{k+10}$, $k = 1,2,3$ correspondingly becomes $M'$ and $M''$, as presented below.

$M'$:

$$\begin{cases} Min\ \delta_1 \\ subject\ to \\ (Z_k)^{Tm}_{pessi} - (\omega_k)_{pessi}\delta_1 \leq \left((Z_k)^{Tm}_{pessi}\right)^* \\ constraints\ of\ (4.1) \\ \forall(\omega_k)_{pessi} \in [0,1], k = 1,2,3 \end{cases} \tag{4.7}$$

$M''$:

$$\begin{cases} Min\ \delta_2 \\ subject\ to \\ (Z_k)^{Tm}_{opti} - (\omega_k)_{opti}\delta_2 \leq \left((Z_k)^{Tm}_{opti}\right)^* \\ constraints\ of\ (4.1) \\ \forall(\omega_k)_{opti} \in [0,1], k = 1,2,3, \end{cases} \tag{4.8}$$

where $\left((Z_k)^{Tm}_{pessi}\right)^*$ and $\left((Z_k)^{Tm}_{opti}\right)^*$ are the goals associated with each of the objectives $(Z_k)^{Tm}_{pessi}$ and $(Z_k)^{Tm}_{opti}$, respectively $\forall\, k \in \{1, 2, 3\}$. In our study, these goals are obtained by solving each objective of $M_{k+4}$ and $M_{k+10}$, $k = 1,2,3$ individually. The compromise models in (4.7) and (4.8) are formulated using goal attainment method (cf. Section 1.3.15.3). Subsequently, we determine the compromise objective vectors, $(Z)^{Tm}_{pessi} = \begin{bmatrix} (Z_1)^{Tm}_{pessi} \\ (Z_2)^{Tm}_{pessi} \\ (Z_3)^{Tm}_{pessi} \end{bmatrix}$ and $(Z)^{Tm}_{opti} = \begin{bmatrix} (Z_1)^{Tm}_{opti} \\ (Z_2)^{Tm}_{opti} \\ (Z_3)^{Tm}_{opti} \end{bmatrix}$ by solving the models $M'$ and $M''$, respectively.

**Step 9**: Set $Tm$=$Tm_2$, where $Tm_2 \in (0.5,1.0]$ and repeat Step 3 through Step 8.

## 4.4 Numerical Illustration

In this section, we have considered a WCDN $G(V_G, E_G)$ depicted in Fig. 4.1. $V_G$ and $E_G$ are respectively, the set of vertices and edges of $G$. The weights of each edge $e_{ij} \in E_G$ are represented by three criteria: distance $(P^1_{ij})$, cost $(P^2_{ij})$ and time $(P^3_{ij})$. Each of these criterion is expressed as rough variable, i.e., $P^k_{ij} = \left([P^{k2}_{ij}, P^{k3}_{ij}], [P^{k1}_{ij}, P^{k4}_{ij}]\right)$, $k = \{1,2,3\}$ so that $P^{k1}_{ij} \leq P^{k2}_{ij} < P^{k3}_{ij} \leq P^{k4}_{ij}$. Vertices, $v_1$ and $v_{11}$ of $G$ are respectively, designated as source $(s)$ and sink $(t)$ vertices. The edge weights of $G$ are presented in Table 4.1.

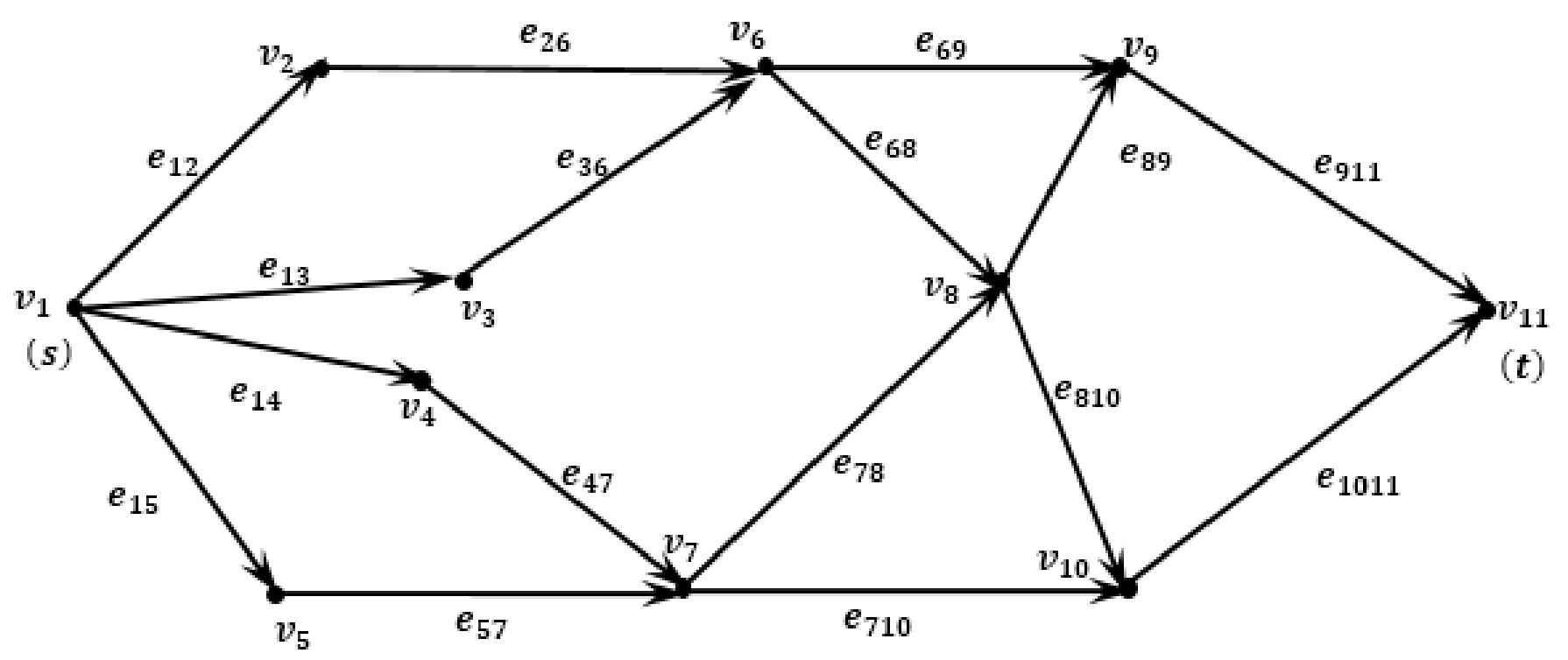

Figure 4.1 A weighted connected directed network $G$

Table 4.1 Edge weights of the WCDN $G$: distance, cost and time represented by rough variable

| Edges $(e_{ij})$ | Distance $\left(P_{ij}^1 = \left([P_{ij}^{12}, P_{ij}^{13}], [P_{ij}^{11}, P_{ij}^{14}]\right)\right)$ | Cost $\left(P_{ij}^2 = \left([P_{ij}^{22}, P_{ij}^{23}], [P_{ij}^{21}, P_{ij}^{24}]\right)\right)$ | Time $\left(P_{ij}^3 = \left([P_{ij}^{32}, P_{ij}^{33}], [P_{ij}^{31}, P_{ij}^{34}]\right)\right)$ |
|---|---|---|---|
| $e_{12}$ | [32,34], [30,35] | [23,24], [22,28] | [26,27], [25,29] |
| $e_{13}$ | [34,36], [33,38] | [24,26], [21,29] | [15,17], [14,18] |
| $e_{14}$ | [34,35], [33,38] | [19,20], [18,22] | [24,25], [20,26] |
| $e_{15}$ | [32,34], [31,36] | [21,22], [20,23] | [17,18], [16,19] |
| $e_{26}$ | [12,14], [11,17] | [23,24], [22,26] | [26,27], [25,28] |
| $e_{36}$ | [21,23], [19,26] | [27,29], [26,32] | [15,17], [12,18] |
| $e_{47}$ | [9,11], [7,12] | [7,8], [5,9] | [14,17], [13,18] |
| $e_{57}$ | [8,10], [6,12] | [9,10], [7,11] | [15,16], [14,17] |
| $e_{68}$ | [5,6], [4,7] | [9,11], [7,12] | [4,5], [3,6] |
| $e_{69}$ | [11,14], [10,15] | [21,26], [20,27] | [24,25], [23,28] |
| $e_{78}$ | [21,24], [20,25] | [4,6], [3,7] | [12,14], [11,19] |
| $e_{710}$ | [25,26], [23,28] | [31,37], [30,39] | [29,32], [27,35] |
| $e_{89}$ | [25,27], [24,29] | [42,46], [41,47] | [6,7], [4,9] |
| $e_{810}$ | [25,27], [23,32] | [12,14], [10,17] | [23,27], [21,31] |
| $e_{911}$ | [31,36], [30,37] | [26,27], [21,28] | [34,35], [32,37] |
| $e_{1011}$ | [30,35], [29,36] | [24,25], [22,27] | [33,36], [31,37] |

The MRD algorithm (cf. Section 4.2) is applied to $G$, and the pessimistic and optimistic shortest paths with their corresponding path weights are determined at trust levels $Tm_1$ and $Tm_2$, where $Tm_1 = 0.45$ and $Tm_2 = 0.85$. The related results are shown in Table 4.2. Moreover, the compromise pessimistic and optimistic paths and their path lengths are also reported in this table. Here, we observe that, when the trust level is set to 0.45, then the optimistic path weights become greater than the corresponding pessimistic path weights, but at trust level 0.85, the pessimistic path weights become greater than the respective optimistic path weights of $G$. This fact characteristically implies the property of rough variable (B. Liu 2002, 2004). In other words, for a rough variable $\zeta$, $\zeta_{inf}(Tm) \leq \zeta_{sup}(Tm)\ \forall\ Tm \in (0.0,0.5]$ and $\zeta_{inf}(Tm) \geq \zeta_{sup}(Tm)\ \forall\ Tm \in (0.5,1.0]$, where $\zeta_{inf}(Tm)$ and $\zeta_{sup}(Tm)$ are respectively, the pessimistic and optimistic values of $\zeta$, and $Tm$ is its trust level.

Table 4.2 Pessimistic and optimistic shortest paths of the WCDN $G$ when solved by MRD algorithm

| Pessimistic Shortest Paths of $G$ | | | Optimistic Shortest Paths of $G$ | | |
|---|---|---|---|---|---|
| $\left[Path_{pessi}^{P^1}\right]^{0.45}$ | $v_1 - v_2 - v_6 - v_9 - v_{11}$ | 91.3351 | $\left[Path_{opti}^{P^1}\right]^{0.45}$ | $v_1 - v_2 - v_6 - v_9 - v_{11}$ | 92.8792 |
| $\left[Path_{pessi}^{P^1}\right]^{0.85}$ | $v_1 - v_2 - v_6 - v_9 - v_{11}$ | 98.4113 | $\left[Path_{opti}^{P^1}\right]^{0.85}$ | $v_1 - v_2 - v_6 - v_9 - v_{11}$ | 86.3458 |
| $\left[Path_{pessi}^{P^2}\right]^{0.45}$ | $v_1 - v_4 - v_7 - v_8 - v_{10} - v_{11}$ | 69.0790 | $\left[Path_{opti}^{P^2}\right]^{0.45}$ | $v_1 - v_4 - v_7 - v_8 - v_{10} - v_{11}$ | 70.1433 |
| $\left[Path_{pessi}^{P^2}\right]^{0.85}$ | $v_1 - v_4 - v_7 - v_8 - v_{10} - v_{11}$ | 75.0933 | $\left[Path_{opti}^{P^2}\right]^{0.85}$ | $v_1 - v_4 - v_7 - v_8 - v_{10} - v_{11}$ | 64.8289 |
| $\left[Path_{pessi}^{P^3}\right]^{0.45}$ | $v_1 - v_3 - v_6 - v_8 - v_9 - v_{11}$ | 76.7251 | $\left[Path_{opti}^{P^3}\right]^{0.45}$ | $v_1 - v_3 - v_6 - v_8 - v_9 - v_{11}$ | 77.7749 |
| $\left[Path_{pessi}^{P^3}\right]^{0.85}$ | $v_1 - v_3 - v_6 - v_8 - v_9 - v_{11}$ | 81.8333 | $\left[Path_{opti}^{P^3}\right]^{0.85}$ | $v_1 - v_3 - v_6 - v_8 - v_9 - v_{11}$ | 71.7667 |
| $\left[Path_{pessi}^{Comp}\right]^{0.45}$ | $v_1 - v_5 - v_7 - v_{10} - v_{11}$ | 94.7430 | $\left[Path_{opti}^{Comp}\right]^{0.45}$ | $v_1 - v_5 - v_7 - v_{10} - v_{11}$ | 95.9998 |
| $\left[Path_{pessi}^{Comp}\right]^{0.85}$ | $v_1 - v_5 - v_7 - v_{10} - v_{11}$ | 100.4138 | $\left[Path_{opti}^{Comp}\right]^{0.85}$ | $v_1 - v_5 - v_7 - v_{10} - v_{11}$ | 89.1021 |

The proposed MRCCSPP is also applied on $G$ to determine its pessimistic and optimistic shortest paths. We have used LINGO 11.0 to solve all the models of MRCCSPP. Here, solving models $M_5, M_6$ and $M_7$ of Step 4 and $M_{11}, M_{12}$ and $M_{13}$ of Step 6 of MRCCSPP (cf. Section 4.3), for $G$, at trust levels, $Tm = 0.45$ and $Tm = 0.85$, we get the pessimistic and optimistic shortest paths. Specifically, models $M_5$ and $M_{11}$ are solved to determine the corresponding pessimistic and optimistic shortest paths with respect to criterion $P^1$. Similarly, considering the criterion $P^2$, the optimized values of $M_6$ and $M_{12}$, respectively generate pessimistic and optimistic shortest paths. Whereas, the pessimistic and optimistic shortest paths for $M_7$ and $M_{13}$ are respectively generated when $P^3$ is considered. Table 4.3 presents all the pessimistic and optimistic shortest paths and their corresponding path weights while solving the corresponding models of MRCCSPP as discussed above, at different trust levels of 0.45 and 0.85. Further, to determine the compromise pessimistic and optimistic shortest paths, the corresponding models $M'$ and $M''$ of MRCCSPP are solved for $G$ at the same trust levels. These results are summarized in Table 4.4. Besides, Table 4.4 also displays the argument-dependent OWA weight vectors $[\omega_1, \omega_2, \omega_3]^T$ of $M'$ and $M''$ for every compromise pessimistic and optimistic shortest paths at different trust levels. For the pessimistic and optimistic cases each of the corresponding decision variable, $x_{ij}$ in Table 4.4 assign value, either 1 or 0. If $x_{ij}$=1 then the corresponding edge, $e_{ij}$ is included in the shortest path. Whereas, if $x_{ij}$=0 then $e_{ij}$ is excluded from the shortest path. The compromise objective vectors, $(Z)_{pessi}^{Tm}$ and $(Z)_{opti}^{Tm}$ for $G$, at trust measures 0.45 and 0.85 are also shown in Table 4.5.

Table 4.3 Optimum values for pessimistic models $M_5, M_6, M_7$ and optimistic models $M_{11}, M_{12}, M_{13}$ for the WCDN $G$ at different trust levels of $Tm$

| $Tm = 0.45$ | | | | | |
|---|---|---|---|---|---|
| Pessimistic Paths of $G$ | | | Optimistic Paths of $G$ | | |
| $M_5$ | $v_1 - v_2 - v_6 - v_9 - v_{11}$ | 91.3829 | $M_{11}$ | $v_1 - v_2 - v_6 - v_9 - v_{11}$ | 92.9600 |
| $M_6$ | $v_1 - v_4 - v_7 - v_8 - v_{10} - v_{11}$ | 69.0710 | $M_{12}$ | $v_1 - v_4 - v_7 - v_8 - v_{10} - v_{11}$ | 70.1548 |
| $M_7$ | $v_1 - v_3 - v_6 - v_8 - v_9 - v_{11}$ | 76.7300 | $M_{13}$ | $v_1 - v_3 - v_6 - v_8 - v_9 - v_{11}$ | 77.8033 |
| $Tm = 0.85$ | | | | | |
| $M_5$ | $v_1 - v_2 - v_6 - v_9 - v_{11}$ | 97.6914 | $M_{11}$ | $v_1 - v_2 - v_6 - v_9 - v_{11}$ | 86.6514 |
| $M_6$ | $v_1 - v_4 - v_7 - v_8 - v_{10} - v_{11}$ | 74.8000 | $M_{12}$ | $v_1 - v_4 - v_7 - v_8 - v_{10} - v_{11}$ | 65.2000 |
| $M_7$ | $v_1 - v_3 - v_6 - v_8 - v_9 - v_{11}$ | 81.1000 | $M_{13}$ | $v_1 - v_3 - v_6 - v_8 - v_9 - v_{11}$ | 71.9000 |

Similar to Table 4.2, it is observed from Table 4.3, the pessimistic shortest path weights are smaller than optimistic values when the trust level $Tm$ is 0.45. Whereas, when $Tm$ is set to 0.85, the pessimistic shortest path weights become greater than optimistic shortest path weights.

Models (4.4) and (4.6) corresponding to Step 4 and Step 6 of MRCCSPP are also solved using NSGA-II for $G$. Here, jMetal 4.5 (Durillo and Nebro 2011) framework is used to implement NSGA-II on these models. While executing these models the trust levels are considered as 0.45 and 0.85, and the parameter setting of NSGA-II is set as follows. Population size =100, maximum generation= 250, crossover probability=0.9, mutation probability=0.03. The selection, crossover and mutation operators are considered as binary tournament selection, single point crossover and bit-flip mutation, respectively. The corresponding nondominated solutions of these models for $G$ are shown in Fig 4.2.

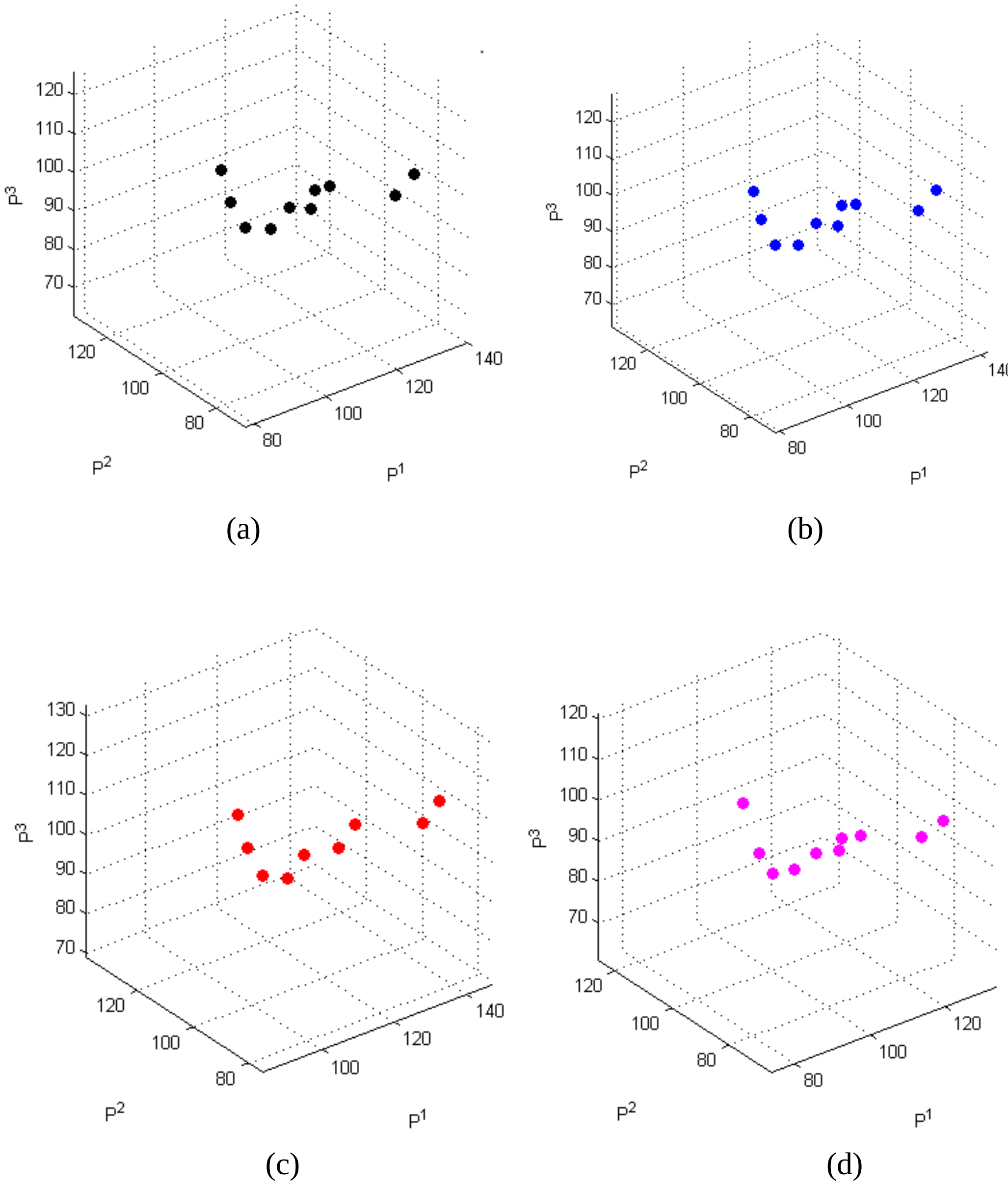

Figure 4.2 Nondominated solutions of the WCDN $G$ by solving the corresponding (a) pessimistic shortest path model (4.4) at 0.45 trust level (b) optimistic shortest path model (4.6) at 0.45 trust level (c) pessimistic shortest path model (4.4) at 0.85 trust level (d) optimistic shortest path model (4.6) at 0.85 trust level

## 4.5 Results and Discussion

In this section, the results generated by the MRD algorithm and the models of MRCCSPP for $G$ are compared. These results are shown in Table 4.6. Here, it is observed that the compromise pessimistic and optimistic path weights generated by execution of MRD algorithm on $G$, lies within the range of compromise objective vectors, obtained by solving $M'$ and $M''$ of MRCCSPP.

Table 4.4 Optimum results for the models $M'$ and $M''$ for the WCDN $G$

| Decision Variables and OWA weight vectors | $Tm = 0.45$ | | $Tm = 0.85$ | |
|---|---|---|---|---|
| | Pessimistic Case | Optimistic Case | Pessimistic Case | Optimistic Case |
| | $M'$: $Min\ \delta_1$ | $M''$: $Min\ \delta_2$ | $M'$: $Min\ \delta_1$ | $M''$: $Min\ \delta_2$ |
| $x_{12}$ | 0 | 0 | 0 | 0 |
| $x_{13}$ | 0 | 0 | 0 | 1 |
| $x_{14}$ | 0 | 0 | 0 | 0 |
| $x_{15}$ | 1 | 1 | 1 | 0 |
| $x_{26}$ | 0 | 0 | 0 | 0 |
| $x_{36}$ | 0 | 0 | 0 | 1 |
| $x_{47}$ | 0 | 0 | 0 | 0 |
| $x_{57}$ | 1 | 1 | 1 | 0 |
| $x_{68}$ | 0 | 0 | 0 | 0 |
| $x_{69}$ | 0 | 0 | 0 | 1 |
| $x_{78}$ | 0 | 0 | 0 | 0 |
| $x_{710}$ | 1 | 1 | 1 | 0 |
| $x_{89}$ | 0 | 0 | 0 | 0 |
| $x_{810}$ | 0 | 0 | 0 | 0 |
| $x_{911}$ | 0 | 0 | 0 | 1 |
| $x_{10,11}$ | 1 | 1 | 1 | 0 |
| $\omega_1$ | 0.2500 | 0.2500 | 0.2500 | 0.2500 |
| $\omega_2$ | 0.4527 | 0.4506 | 0.4348 | 0.4444 |
| $\omega_3$ | 0.2973 | 0.2994 | 0.3152 | 0.3056 |
| **Optimize units** | 8.0000 | 8.0000 | 66.3071 | 70.2070 |

Table 4.5 Compromise objective vectors of the WCDN $G$ at different trust levels of $Tm$

| $Tm = 0.45$ | | | | | | $Tm = 0.85$ | | | | | |
|---|---|---|---|---|---|---|---|---|---|---|---|
| $(Z)_{pessi}^{Tm}$ | | | $(Z)_{opti}^{Tm}$ | | | $(Z)_{pessi}^{Tm}$ | | | $(Z)_{pessi}^{Tm}$ | | |
| $(Z_1)_{pessi}^{Tm}$ | $(Z_2)_{pessi}^{Tm}$ | $(Z_3)_{pessi}^{Tm}$ | $(Z_1)_{opti}^{Tm}$ | $(Z_2)_{opti}^{Tm}$ | $(Z_3)_{opti}^{Tm}$ | $(Z_1)_{pessi}^{Tm}$ | $(Z_2)_{pessi}^{Tm}$ | $(Z_3)_{pessi}^{Tm}$ | $(Z_1)_{opti}^{Tm}$ | $(Z_2)_{opti}^{Tm}$ | $(Z_3)_{opti}^{Tm}$ |
| 99.4545 | 88.8700 | 97.4286 | 100.8485 | 90.1300 | 98.5714 | 105.0303 | 93.7000 | 102.0000 | 97.7333 | 96.4000 | 87.0000 |

For the simulation purposes, we have considered a personal computer with Intel $(R)$ Core$(TM)$ i3 @ 2.93GHz and 4 GB memory as a test platform. Here, we have executed the MRD algorithm using Matlab R2013b, on large instances of WCDNs. For every WCDN, the edge weights corresponding to the criteria, $P^1, P^2$ and $P^3$ are generated randomly. Table 4.7 shows the computational times of MRD algorithm for large WCDNs. It has been observed that the computation time for each of these WCDNs is less than 9 sec.

Table 4.6 Comparisons of results obtained by MRD algorithm and MRCCSPP for the WCDN $G$

| | $Tm = 0.45$ | | $Tm = 0.85$ | |
|---|---|---|---|---|
| | Path weights by MRD algorithm | Path weights by MRCCSPP | Path weights by MRD algorithm | Path weights by MRCCSPP |
| | Compromise path length | Optimal vector $(Z^*)^{Tm}_{pessi}$ | Compromise path length | Optimal vector $(Z^*)^{Tm}_{opti}$ |
| $Pessimistic\ Case$ | 94.7430 | 99.4545 | 100.4138 | 105.0303 |
| | | 88.8700 | | 93.7000 |
| | | 97.4286 | | 102.0000 |
| $Optimistic\ Case$ | 95.9998 | 100.8485 | 89.1021 | 97.7333 |
| | | 90.1300 | | 96.4000 |
| | | 98.5714 | | 87.0000 |

In order to study the performance metrics: hypervolume ($HV$) and spread ($\Delta$) for our proposed problem, we have compared these metrics, for $G$ (cf. Fig. 4.1) and $R$ (cf. Fig. A.2 in Appendix A), by executing NSGA-II on both these WCDNs. The associated input parameters of $R$ are provided in Table A.2 in Appendix A. For this purpose, we have used the jMetal4.5 framework to solve the models (4.4) and (4.6) for $G$ and $R$, using NSGA-II, at two different trust measures. Accordingly, for the experimental analysis, we have considered four unique combinations, each for $G$ and $R$. Two combinations are set to solve the pessimistic model (4.4), at trust levels 0.45 and 0.85. Whereas, the remaining two combinations are used to solve the optimistic model (4.6) at corresponding trust levels 0.45 and 0.85. Due to stochastic behaviour of NSGA-II, we have considered 15 executions of NSGA-II for every combination of the models (4.4) and (4.6), for $G$ and $R$. For each of these executions, the parameter setting of NSGA-II are considered as the same, as mentioned above.

For most of the multi-objective problems, the set of optimal solutions in the Pareto front ($PF$) are usually not available. Therefore, we approximate the $PF$ by generating a reference front by collecting all the best nondominated solutions from every execution of NSGA-II. Subsequently, for each execution of the algorithm, the values of the performance metrics are evaluated with respect to the reference front. Consequently, we determine maximum, minimum, mean and standard deviation ($sd$) for $HV$, $\Delta$ and execution time, considering all the executions of NSGA-II for each combination.

The related results of $HV$ and $\Delta$ are shown in Table 4.8-Table 4.11, whereas the corresponding results of execution time are presented in tables 4.12 and 4.13. In each of these tables, the superior values are displayed in bold. It is observed that the mean of $HV$ and $\Delta$, respectively increases and decreases for $R$ as compared to $G$, with respect to both pessimistic and optimistic shortest path problems. However, for both the pessimistic and optimistic cases, the execution time for $G$ becomes better than $R$.

Table 4.7 Computation time for MRD algorithm for larger WCDNs with rough parameters

| Vertices | Edges | Computation Time ($sec$) |
|---|---|---|
| 100 | 700 | 0.5189 |
| | 1000 | 0.7734 |
| | 3000 | 0.9736 |
| | 4900 | 1.2471 |
| 200 | 9000 | 0.8479 |
| | 12000 | 1.0798 |
| | 15000 | 1.3683 |
| | 18000 | 1.9231 |
| 300 | 27000 | 2.8664 |
| | 30000 | 3.0158 |
| | 33000 | 3.3789 |
| | 35000 | 4.1538 |
| 400 | 42000 | 4.3987 |
| | 45000 | 4.5414 |
| | 48000 | 4.9872 |
| | 51000 | 5.3844 |
| 500 | 53000 | 5.3510 |
| | 58000 | 5.7271 |
| | 65000 | 6.0575 |
| | 80000 | 8.7187 |

Table 4.8 Statistical measures of $HV$ after 15 runs of NSGA-II on the pessimistic model (4.4) of the WCDNs $G$ and $R$

| | $HV$ | | | |
|---|---|---|---|---|
| | $Tm = 0.45$ | | $Tm = 0.85$ | |
| | $G$ | $R$ | $G$ | $R$ |
| | $Pessimistic\ Case$ | $Pessimistic\ Case$ | $Pessimistic\ Case$ | $Pessimistic\ Case$ |
| $Minimum$ | 0.2028 | **0.2345** | 0.2927 | **0.7947** |
| $Maximum$ | 0.2671 | **0.2990** | 0.3256 | **0.7818** |
| $Mean$ | 0.2492 | **0.2796** | 0.3101 | **0.7619** |
| $sd$ | **0.0289** | 0.0719 | **0.0172** | 0.0517 |

## 4.6 Conclusion

In this chapter, we have proposed two approaches to solve MSPP for a WCDN with rough parameters. The first one is the MRD algorithm, and the other is the MRCCSPP. We have determined the compromise pessimistic and optimistic shortest paths, at two different trust measures, using the proposed methods, and compare the related results. Simulation of the MRD algorithm for larger instances of WDCNs is also studied. Moreover, the crisp equivalent of the proposed models for pessimistic and optimistic shortest paths are also solved by NSGA–II, and the related performance metrics ($HV$ and $\Delta$) and *execution time* are studied for the two WCDNs, $G$ and $H$.

Table 4.9 Statistical measures of $HV$ after 15 runs of NSGA-II on the optimistic model (4.6) of the WCDNs $G$ and $R$

| | $HV$ | | | |
|---|---|---|---|---|
| | $Tm = 0.45$ | | $Tm = 0.85$ | |
| | $G$ | $R$ | $G$ | $R$ |
| | $Optimistic\ Case$ | $Optimistic\ Case$ | $Optimistic\ Case$ | $Optimistic\ Case$ |
| $Minimum$ | 0.3082 | $\mathbf{0.6594}$ | 0.2455 | $\mathbf{0.8134}$ |
| $Maximum$ | 0.3239 | $\mathbf{0.8988}$ | 0.2792 | $\mathbf{0.8621}$ |
| $Mean$ | 0.3179 | $\mathbf{0.8584}$ | 0.2693 | $\mathbf{0.8560}$ |
| $sd$ | $\mathbf{0.0657}$ | 0.0945 | $\mathbf{0.0734}$ | 0.0929 |

Table 4.10 Statistical measures of $\Delta$ after 15 runs of NSGA-II on the pessimistic model (4.4) of the WCDNs $G$ and $R$

| | $\Delta$ | | | |
|---|---|---|---|---|
| | $Tm = 0.45$ | | $Tm = 0.85$ | |
| | $G$ | $R$ | $G$ | $R$ |
| | $Pessimistic\ Case$ | $Pessimistic\ Case$ | $Pessimistic\ Case$ | $Pessimistic\ Case$ |
| $Minimum$ | 1.8345 | $\mathbf{1.5299}$ | 1.9207 | $\mathbf{1.5859}$ |
| $Maximum$ | 1.9790 | $\mathbf{1.9670}$ | $\mathbf{1.9854}$ | 1.9978 |
| $Mean$ | 1.9503 | $\mathbf{1.8520}$ | 1.9650 | $\mathbf{1.7903}$ |
| $sd$ | $\mathbf{0.0372}$ | 0.1192 | $\mathbf{0.0166}$ | 0.1069 |

Table 4.11 Statistical measures of $\Delta$ after 15 runs of NSGA-II on the optmistic model (4.6) of the WCDNs $G$ and $R$

| | $\Delta$ | | | |
|---|---|---|---|---|
| | $Tm = 0.45$ | | $Tm = 0.85$ | |
| | $G$ | $R$ | $G$ | $R$ |
| | $Optimistic\ Case$ | $Optimistic\ Case$ | $Optimistic\ Case$ | $Optimistic\ Case$ |
| $Minimum$ | 1.6054 | $\mathbf{1.5139}$ | 1.9080 | $\mathbf{1.5582}$ |
| $Maximum$ | 1.9836 | $\mathbf{1.9198}$ | 1.9739 | $\mathbf{1.8381}$ |
| $Mean$ | 1.9434 | $\mathbf{1.8747}$ | 1.9128 | $\mathbf{1.7705}$ |
| $sd$ | $\mathbf{0.0981}$ | 0.1294 | $\mathbf{0.0169}$ | 0.1143 |

Table 4.12 Statistical measures of execution time after 15 runs of NSGA-II on the pessimistic model (4.4) of the WCDNs $G$ and $R$

| | Execution time (sec.) | | | |
|---|---|---|---|---|
| | $Tm = 0.45$ | | $Tm = 0.85$ | |
| | $G$ | $R$ | $G$ | $R$ |
| | $Pessimistic\ Case$ | $Pessimistic\ Case$ | $Pessimistic\ Case$ | $Pessimistic\ Case$ |
| $Minimum$ | $\mathbf{11.9172}$ | 15.4102 | $\mathbf{10.6260}$ | 16.0722 |
| $Maximum$ | $\mathbf{14.0885}$ | 16.9502 | $\mathbf{14.2977}$ | 18.5532 |
| $Mean$ | $\mathbf{13.2367}$ | 16.2143 | $\mathbf{12.6547}$ | 16.9874 |
| $sd$ | $\mathbf{0.5228}$ | 0.5240 | 0.8317 | $\mathbf{0.5712}$ |

Table 4.13 Statistical measures of execution time after 15 runs of NSGA-II on the optimistic model (4.6) of the WCDNs $G$ and $R$

| | Execution time (sec.) | | | |
|---|---|---|---|---|
| | $Tm = 0.45$ | | $Tm = 0.85$ | |
| | $G$ | $R$ | $G$ | $R$ |
| | $Optimistic\ Case$ | $Optimistic\ Case$ | $Optimistic\ Case$ | $Optimistic\ Case$ |
| $Minimum$ | $\mathbf{12.2733}$ | 15.7480 | $\mathbf{12.1042}$ | 15.2633 |
| $Maximum$ | $\mathbf{13.8537}$ | 17.2769 | $\mathbf{14.0986}$ | 17.7376 |
| $Mean$ | $\mathbf{12.9270}$ | 16.6081 | $\mathbf{12.9908}$ | 16.3516 |
| $sd$ | $\mathbf{0.4661}$ | 0.6065 | $\mathbf{0.5473}$ | 0.5970 |

In future, modelling of different combinatorial optimization problems like $k$-shortest paths, maximal-flow problem and scheduling problems under rough, rough fuzzy, rough random and fuzzy random environments can be considered as our subject of interest.

# Chapter 5

# Uncertain Multi-objective Multi-item Fixed Charge Solid Transportation Problem with Budget Constraint

# Chapter 5

# Uncertain Multi-objective Multi-item Fixed Charged Solid Transportation with Budget Constraint

## 5.1 Introduction

The notion of classical transportation problem (TP) is to determine an optimal solution (transportation plan) such that the transportation cost is minimized. The noteworthy work of F.L. Hitchcock (1941) introduced the term transportation problem, by modelling it as a conventional optimization problem with two-dimensional properties, i.e., supply and demand. However, apart from supply and demand constraints, in real-world scenarios, often we need to consider the mode of transportation (e.g., goods train, cargo flights and trucks), types of goods, etc. Under such circumstances, a TP is extended to a solid transportation problem (STP), where apart from source and destination constraints, an additional constraint, related to the modes of transportation (convenience) or types of goods is considered. E.D. Schell (1955), first extended the classical TP to solid transportation problem. Later, Bhatia et al. (1976) minimized the shipping time of an STP. Afterwards, Jiménez and Verdegay (1998), in their work, dealt with an STP and solved the problem by considering the supply quantities, demand quantities and conveyance capacities as interval values instead of point values.

Hirsch and Dantzig (1968) presented another variant of TP, called fixed charge transportation problem (FTP). In FTP, the objective is to determine an optimal transportation plan that minimizes the overall cost between sources and destinations. The overall cost has two components: variable shipping cost and an independent fixed charge. The shipping cost depends directly on the quantity of the transported item(s) from sources to destinations whereas, the fixed charge, which commonly appears due to the expenditure related to permit fees, property tax or toll charges, etc., is associated with every feasible transportation plan. Several researchers (Kennington and Unger 1976; Sun et al. 1998; Adlakha and Kowalski 2003) have proposed different solution approaches for FTP.

Most of the real-life decision making problems, can be well expressed with multiple conflicting criteria and are usually formulated as multi-objective optimization problems. Moreover, the associated parameters for such problems may be imprecise due to insufficient or inexact information owing to incompleteness or lack of evidence, statistical analysis, etc. Therefore, to process and represent imprecise or ill-defined data for decision making problems, many researchers have presented a number of improved theories, e.g., fuzzy set (L.A. Zadeh 1965), type-2 fuzzy set (L.A. Zadeh 1975a, b), rough set (Z. Pawlak 1982), etc. Likewise, the uncertainty theory proposed by B. Liu (2007) deals with the belief degrees of human being, which can appropriately evaluate personal belief degree of experts in terms of uncertain measure.

In the literature, there are many studies done in the field of TP with uncertain input parameters. Kaur and Kumar (2012) described a new solution technique for TP by considering the transportation costs as generalized trapezoidal fuzzy numbers. In the multi-objective domain, Bit et al. (1993) solved the multi-objective STP using Zimmermann's fuzzy programming technique (H.-J. Zimmermann 1978). Later, Gen et al. (1995) proposed a bi-objective STP with fuzzy parameters. Recently, Kundu et al. (2017a) used interval type-2 fuzzy multi-criteria group decision making approach to solve a transportation mode selection problem. Considering the budget limitation of a TP, Kundu et al. (2013b) discussed a multi-objective STP by imposing budget constraints at destinations with fuzzy random hybrid parameters. Later, Baidya and Bera (2014) proposed a budget constraint STP by representing transportation cost, availability and demand, and conveyance capacity as interval numbers. Besides, Kundu et al. (2014b) also investigated a fuzzy multi-criteria decision making (MCDM) approach to determine the most preferred transportation mode, to solve an STP. Recently, Das et al. (2017) solved the green STP under type-2 fuzzy environment. Considering fixed charge TP in imprecise domain, Liu et al. (2008) presented an FTP in a fuzzy environment and solved it by genetic algorithm. Later, Pramanik et al. (2015b) revisited the FTP by considering the shipping cost, fixed charge, availability and demand in a two-stage supply chain under type-2 Gaussian fuzzy transportation network.

Uncertainty theory, founded by B. Liu (2007) and refined by B. Liu (2009, 2010), is a branch of axiomatic mathematics for modelling human uncertainty. Nowadays, uncertainty theory has been applied to many areas such as economics (Yang and Gao 2016; Yang and Gao 2017), management (Gao and Yao 2015; Gao et al. 2017) and finance (Chen and Gao 2013; Guo and Gao 2017). Similarly, under the framework of uncertainty theory, different variants of TP are also studied in the literature (Sheng and Yao 2012a, 2012b; Cui and Sheng 2013; Mou et al. 2013; Guo et al. 2015; Chen et al. 2017). In Table 5.1, we have furnished some recent reviews of different variants of TP under diverse uncertain environments.

In spite of all the developments of TP, there are several gaps in the literature which are listed below.

(i) To the best of our knowledge, none has considered, a multi-objective multi-item profit maximization and time minimization fixed charge STP model with budget constraint.

Table 5.1 Some recent studies on variants of transportation problem under diverse uncertain frameworks

| Author(s) (year) | Uncertain Environment | Objective(s) | Uncertain Programming Model(s) | Model variants | | | | |
|---|---|---|---|---|---|---|---|---|
| | | | | TP | STP | FTP | Items | Budget Constraint |
| Sheng and Yao (2012a) | Uncertainty Theory | Single | Expected-constrained | Yes | No | Yes | Single | No |
| Sheng and Yao (2012b) | Uncertainty Theory | Single | Expected-constrained | Yes | No | No | Single | No |
| Mou et al. (2013) | Uncertainty Theory | Multi | Expected-constrained | Yes | No | No | Single | No |
| Kundu et al. (2013b) | Random, Fuzzy, Random fuzzy | Multi | Chance-constrained, Expected value | Yes | Yes | No | Single | Yes |
| Kundu et al. (2013c) | Fuzzy | Multi | Expected value, Minimum of Fuzzy Number | Yes | Yes | No | Multi | No |
| Kundu et al. (2014a) | Type-2 Fuzzy | Single | Chance-constrained | Yes | No | Yes | Single | No |
| Giri et al. (2015) | Fuzzy | Single | Chance-constrained | Yes | Yes | Yes | Multi | No |
| Sinha et al. (2016) | Interval Type-2 Fuzzy | Multi | Expected value | Yes | Yes | No | Single | No |
| Das et al. (2016) | Rough Interval | Single | Expected value, Chance-constrained | Yes | Yes | No | Single | No |
| H. Dalman (2016) | Uncertainty Theory | Multi | Expected-constrained | Yes | Yes | No | Multi | No |
| Gao and Kar (2017) | Uncertainty Theory | Single | Expected value, Chance-constrained | Yes | Yes | No | Single | No |
| Kundu et al. (2017b) | Rough | Single | Chance-constrained | Yes | Yes | No | Single | No |
| Liu et al. (2017) | Uncertainty Theory | Single | Expected value, Chance-constrained | Yes | Yes | Yes | Multi | No |
| **Proposed model** | Uncertainty Theory | Multi | Expected value, Chance-constrained, Dependent chance-constrained | Yes | Yes | Yes | Multi | Yes |

(ii) Use of Liu's uncertainty theory to formulate the multi-objective solid transportation problem using dependent chance-constrained programming technique is yet to be studied.

(iii) Implementation and analysis of results for EVM, CCM and DCCM for any fixed charge STP with budget constraint, using zigzag and normal uncertain variables, under the uncertain framework, has not yet been done.

In order to address the above mentioned lacunas, in the present study, we have considered uncertain multi-objective multi-item fixed charge STP with budget constraint (UMMFSTPwB) at destinations, under the framework of uncertainty theory

(B. Liu 2007). The problem is formulated in three different models: expected value model (EVM), chance-constrained model (CCM) and dependent chance-constrained model (DCCM). Consecutively, each model is solved by three different multi-objective compromise techniques: linear weighted method, global criterion method and fuzzy programming method.

The subsequent sections of this chapter are organized as follow. The uncertain programming models, EVM, CCM and DCCM of the proposed UMMFSTPwB are presented in Section 5.2. The crisp equivalent for each of the models of UMMFSTPwB is formulated in Section 5.3. Some related theorems of the three different compromise multi-objective solution methodologies are explained in Section 5.4. In Section 5.5, we present the numerical examples to illustrate the model, and the results are discussed with comparative analysis. Finally, the epilogue of our study is presented in Section 5.6.

## 5.2 Problem Definition

A solid transportation problem is concerned with transportation of homogeneous products from source $i$ to destination $j$ via conveyance $k$ $(\geq 2)$ (trucks, cargo van, goods train, etc.) for which we need to find an optimal transportation plan so that the total transportation cost is minimized. A balanced condition in STP assumes that the total supply at source depots, total demand at destinations and total conveyance capacities are equal. But, in reality, we may encounter the following situations in the context of TP. They are mentioned as follow.

(i) Adequate quantities of the item at the sources to satisfy the demands at destinations.

(ii) The conveyances should have the ability to suffice the demand at destinations.

(iii) There may be a requirement of multiple items or products at destinations for which we may need to consider shipment of heterogeneous items from source $i$ to destination $j$ via conveyance $k$.

(iv) Limited budget at destinations.

(v) Besides transportation cost, fixed charge cost can also exist when a transportation activity is initiated from source $i$ to destination $j$.

(vi) In some situations, often we require to optimize several objectives at the same time which are conflicting in nature. In the context of transportation problem, these objectives may be the minimization of total transportation cost, total delivery time, deterioration of breakable goods, maximization of profit, etc.

In order to consider some of the practical situations mentioned above, in our study, we have presented a profit maximization and time minimization of multi-item fixed charge solid transportation problem with budget constraint. The following notations are used to formulate a multi-objective multi-item fixed charge solid transportation problem with budget constraint (MMFSTPwB) model.

- $m$ : Number of origins/source depots.
- $n$ : Number of destinations/demand points.
- $K$ : Number of conveyances utilize for transportation of products.
- $r$ : Number of different types of items to be transported.

  For $i = 1, 2, \dots, m$, $j = 1, 2, \dots, n$, $k = 1, 2, \dots, K$ and $p = 1, 2, \dots, r$
- $a_i^p$: Amount of item $p$ available at source $i$.
- $b_j^p$: Demand of item $p$ at destination $j$.
- $e_k$: Transportation capacity of conveyance $k$.
- $c_{ijk}^p$: Cost for transporting one unit of item $p$ from source $i$ to destination $j$ by conveyance $k$.
- $f_{ijk}^p$: Fixed charge with respect to the transportation activity for transporting item $p$ from source $i$ to destination $j$ by conveyance $k$.
- $x_{ijk}^p$: Quantity of item $p$ transported from source $i$ to destination $j$ by conveyance $k$.
- $t_{ijk}^p$: Time required for transporting item $p$ with respect to transportation activity from source $i$ to destination $j$ by conveyance $k$.
- $s_j^p$: Selling price per unit of item $p$ at the destination $j$.
- $v_i^p$: Purchasing cost per unit of item $p$ at the source $i$.
- $B_j$: Budget at destination $j$.

To initiate a transportation activity from source $i$ to destination $j$ we need to reserve conveyance $k$, leading to a fixed charge. In other words, a fixed charge will be added to the direct transportation cost, if $x_{ijk}^p > 0$. Therefore, in our problem, the fixed charge is incorporated using a binary decision variable $y_{ijk}^p$, where $y_{ijk}^p = \begin{cases} 1 & ; if\ x_{ijk}^p > 0 \\ 0 & ; otherwise \end{cases}$ $\forall\, p, i, j, k$. Accordingly, we present the mathematical model of MMFSTPwB in (5.1).

In model (5.1), the first objective is to maximize the overall profit earned after transporting all the required items from each source to every destination points through the available conveyances. The second objective minimizes the total transportation time for shipment of all the required items from all sources to all destinations. The significance of the constraints of the model (5.1) is discussed below.

The first constraint determines that the total quantity of item $p$ which is shipped from source $i$ does not exceed the availability $a_i^p$. The second constraint determines the shipment of a total quantity of item $p$ at destination $j$, satisfies at least the demand $b_j^p$.

The third constraint implies that the quantity of item $p$ shipped from source $i$ to destination $j$ using conveyance $k$ does not exceed $e_k$. Finally, the fourth constraint indicates that the total expenditure at destination $j$ including the purchase price of item $p$ at source $i$, the transportation cost and fixed charge for transporting item $p$ from source $i$ to destination $j$ via conveyance $k$ is not more than the allocated budget $B_j$.

$$\begin{cases} Max\ Z_1 = \sum_{p=1}^{r}\sum_{i=1}^{m}\sum_{j=1}^{n}\sum_{k=1}^{K}\{(s_j^p - v_i^p - c_{ijk}^p)x_{ijk}^p - f_{ijk}^p y_{ijk}^p\} \\ Min\ Z_2 = \sum_{p=1}^{r}\sum_{i=1}^{m}\sum_{j=1}^{n}\sum_{k=1}^{K}\{t_{ijk}^p y_{ijk}^p\} \\ subject\ to \\ \sum_{j=1}^{n}\sum_{k=1}^{K} x_{ijk}^p \le a_i^p, i = 1,2,\dots,m, p = 1,2,\dots,r \\ \sum_{i=1}^{m}\sum_{k=1}^{K} x_{ijk}^p \ge b_j^p\ , j = 1,2,\dots,n, p = 1,2,\dots,r \\ \sum_{p=1}^{r}\sum_{i=1}^{m}\sum_{j=1}^{n} x_{ijk}^p \le e_k, k = 1,2,\dots,K \\ \sum_{p=1}^{r}\sum_{i=1}^{m}\sum_{k=1}^{K}\{(v_i^p + c_{ijk}^p)x_{ijk}^p + f_{ijk}^p y_{ijk}^p\} \le B_j, j = 1,2,\dots,n \\ x_{ijk}^p \ge 0, \quad y_{ijk}^p = \begin{cases} 1 & ; x_{ijk}^p > 0 \\ 0 & ; otherwise \end{cases} \quad \forall\, p,i,j,k. \end{cases} \tag{5.1}$$

Estimation of an effective transportation plan generally depends on previous records. But, often the data of these previous records may be ill-defined due to uncertainty in the judgement, lack of evidence, or fluctuation in market price, etc. For example, the transportation cost and transportation time cannot be exact since these parameters depend on labour charges, toll tax, fluctuation in fuel price, traffic congestion, etc. Similarly, the supply at source may be inexact due to the variation in the availabilities of manpower, raw materials, demand of products, etc. Under such circumstances, uncertain programming, proposed by B. Liu (2007), becomes one of the relevant techniques to deal with uncertain information. Uncertain programming is mathematical programming which models decision making approaches in uncertain environments using human belief degrees.

In order to propose the formulation of MMFSTPwB problem under the framework of uncertainty theory, we categorically termed the problem as uncertain MMFSTPwB, i.e., uncertain multi-objective multi-item fixed charge solid transportation problem with budget constraint (UMMFSTPwB). The UMMFSTPwB is essentially converted to uncertain programming models, i.e., expected value model (EVM), chance-constrained model (CCM) and dependent chance-constrained model (DCCM). Here, the associated uncertain parameters of UMMFSTPwB are the transportation cost, fixed charge, transportation time, availability of items at sources, demand of items at destinations, conveyance capacities, and budget at destinations. These parameters are respectively, represented by $\xi_{c_{ijk}^p}$, $\xi_{f_{ijk}^p}$, $\xi_{t_{ijk}^p}$, $\xi_{a_i^p}$, $\xi_{b_j^p}$, $\xi_{e_k}$ and $\xi_{B_j}$.

The corresponding formulations of expected value model (EVM), chance-constrained model (CCM) and dependent chance-constrained model (DCCM) are discussed below. These models of UMMFSTPwB along with the solution techniques are graphically represented in Fig. 5.1.

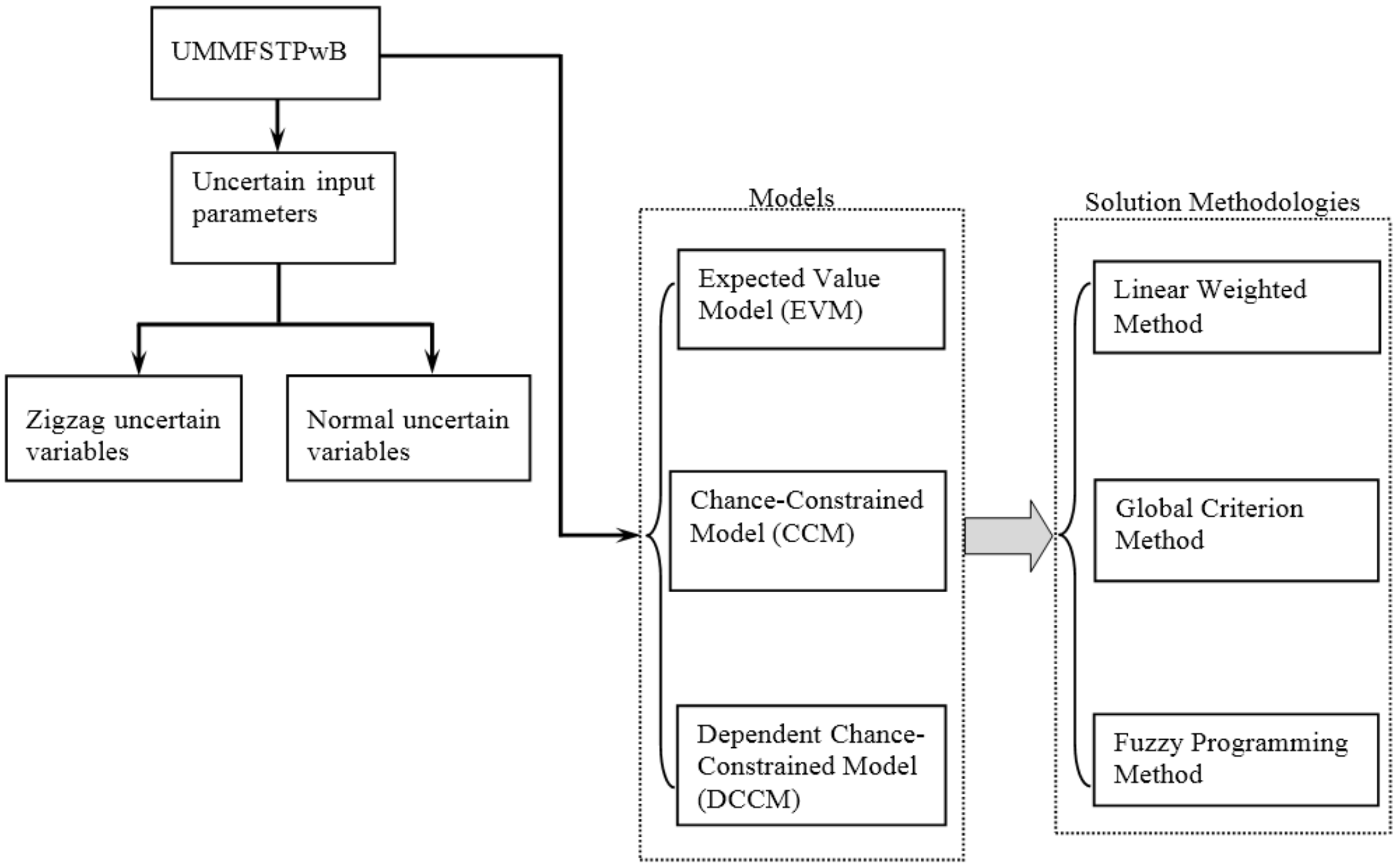

Figure 5.1 Graphical representation of UMMFSTPwB: from model formulation to solution techniques

### 5.2.1 Expected Value Model

Liu and Liu (2002) presented the expected value model of uncertain programming. The model optimizes the expected value of an objective function under the expected constraints. Here, we formulate EVM for UMMFSTPwB and present the model in (5.2).

In model (5.2), the first objective is to maximize the overall expected profit. The second objective is to minimize the expected total transportation time. These objectives are subjected to some expected constraints.

$$\begin{cases} Max\ E[Z_1] = E\left[\sum_{p=1}^{r}\sum_{i=1}^{m}\sum_{j=1}^{n}\sum_{k=1}^{K}\left\{\left(s_j^p - v_i^p - \xi_{c_{ijk}^p}\right)x_{ijk}^p - \xi_{f_{ijk}^p}y_{ijk}^p\right\}\right] \\ Min\ E[Z_2] = E\left[\sum_{p=1}^{r}\sum_{i=1}^{m}\sum_{j=1}^{n}\sum_{k=1}^{K}\xi_{t_{ijk}^p}y_{ijk}^p\right] \\ subject\ to \\ \quad E\left[\sum_{j=1}^{n}\sum_{k=1}^{K}x_{ijk}^p - \xi_{a_i^p}\right] \le 0, i = 1,2,\dots,m, p = 1,2,\dots,r \\ \quad E\left[\sum_{i=1}^{m}\sum_{k=1}^{K}x_{ijk}^p - \xi_{b_j^p}\right] \ge 0, j = 1,2,\dots,n, p = 1,2,\dots,r \\ \quad E\left[\sum_{p=1}^{r}\sum_{i=1}^{m}\sum_{j=1}^{n}x_{ijk}^p - \xi_{e_k}\right] \le 0, k = 1,2,\dots,K \\ \quad E\left[\sum_{p=1}^{r}\sum_{i=1}^{m}\sum_{k=1}^{K}\left\{\left(v_i^p + \xi_{c_{ijk}^p}\right)x_{ijk}^p + \xi_{f_{ijk}^p}y_{ijk}^p\right\} - \xi_{B_j}\right] \le 0, j = 1,2,\dots,n \\ \quad x_{ijk}^p \ge 0,\ y_{ijk}^p = \begin{cases} 1 & ; if\ x_{ijk}^p > 0 \\ 0 & ; otherwise \end{cases} \quad \forall\, p,i,j,k. \end{cases} \tag{5.2}$$

### 5.2.2 Chance-constrained Model

Chance-constrained programming is an alternate method to deal with optimization of a problem under uncertain environment. The basic idea of CCM (Charnes and Cooper 1959; B. Liu 2002) is that it allows violation of constraints, but we need to ensure that the constraints should hold at some chance level (confidence level). Here, we develop the chance-constrained model for UMMFSTPwB and present it in (5.3).

$$\begin{cases} Max\ \bar{Z}_1 \\ Min\ \bar{Z}_2 \\ subject\ to \\ \quad \mathcal{M}\left\{\sum_{p=1}^{r}\sum_{i=1}^{m}\sum_{j=1}^{n}\sum_{k=1}^{K}\left[\left(s_j^p - v_i^p - \xi_{c_{ijk}^p}\right)x_{ijk}^p - \xi_{f_{ijk}^p}y_{ijk}^p\right] \ge \bar{Z}_1\right\} \ge \alpha_1 \\ \quad \mathcal{M}\left\{\sum_{p=1}^{r}\sum_{i=1}^{m}\sum_{j=1}^{n}\sum_{k=1}^{K}\left[\xi_{t_{ijk}^p}y_{ijk}^p\right] \le \bar{Z}_2\right\} \ge \alpha_2 \\ \quad \mathcal{M}\left\{\sum_{j=1}^{n}\sum_{k=1}^{K}x_{ijk}^p - \xi_{a_i^p} \le 0\right\} \ge \beta_i^p, i = 1,2,\dots,m, p = 1,2,\dots,r \\ \quad \mathcal{M}\left\{\sum_{i=1}^{n}\sum_{k=1}^{K}x_{ijk}^p - \xi_{b_j^p} \ge 0\right\} \ge \gamma_j^p, j = 1,2,\dots,n, p = 1,2,\dots,r \\ \quad \mathcal{M}\left\{\sum_{p=1}^{r}\sum_{i=1}^{m}\sum_{j=1}^{n}x_{ijk}^p - \xi_{e_k} \le 0\right\} \ge \delta_k, k = 1,2,\dots,K \\ \quad \mathcal{M}\left\{\sum_{p=1}^{r}\sum_{i=1}^{m}\sum_{k=1}^{K}\left\{\left(v_i^p + \xi_{c_{ijk}^p}\right)x_{ijk}^p + \xi_{f_{ijk}^p}y_{ijk}^p\right\} - \xi_{B_j} \le 0\right\} \ge \rho_j, j = 1,2,\dots,n \\ \quad x_{ijk}^p \ge 0,\ y_{ijk}^p = \begin{cases} 1 & ; if\ x_{ijk}^p > 0 \\ 0 & ; otherwise \end{cases} \quad \forall\, p,i,j,k. \end{cases} \tag{5.3}$$

In CCM, $\alpha_1$, $\alpha_2$, $\beta_i^p$, $\gamma_j^p$, $\delta_k$ and $\rho_j$ are predetermined confidence levels. The objectives $\bar{Z}_1$ and $\bar{Z}_2$ determine the critical values corresponding to the first and second constraints. The first constraint determines the $\alpha_1$-optimistic value of the overall profit corresponding to an $\alpha_1$-transportation plan and the second constraint determines the $\alpha_2$-pessimistic value of the shipping/transportation time with respect to an $\alpha_2$-transportation plan. Moreover, the remaining constraints (third to sixth) of model (5.3), also hold at their corresponding chance levels, which are at least $\beta_i^p$, $\gamma_j^p$, $\delta_k$ and $\rho_j$.

### 5.2.3 Dependent Chance-constrained Model (DCCM)

The main idea of dependent chance programming is to optimize the chance of an uncertain event. For the proposed problem, a DM has to satisfy predetermined values of minimum profit margin and maximum transportation time limit, and then maximize the uncertain measures (confidence level) that satisfy the following.

- Total profit is not less than a predetermined profit margin.
- Total transportation time is not more than the predetermined time.

We formulate the dependent chance programming model under chance constraints in model (5.4) to obtain the most suitable transportation plan.

$$\begin{cases} Max\ v_{Z_1'} = \mathcal{M}\left\{\sum_{p=1}^{r}\sum_{i=1}^{m}\sum_{j=1}^{n}\sum_{k=1}^{K}\left[\left(s_j^p - v_i^p - \xi_{c_{ijk}^p}\right)x_{ijk}^p - \xi_{f_{ijk}^p}y_{ijk}^p\right] \geq Z_1'\right\} \\ Max\ v_{Z_2'} = \mathcal{M}\left\{\sum_{p=1}^{r}\sum_{i=1}^{m}\sum_{j=1}^{n}\sum_{k=1}^{K}\left[\xi_{t_{ijk}^p}y_{ijk}^p\right] \leq Z_2'\right\} \\ subject\ to\ third\ to\ seventh\ constraints\ of\ (5.3), \end{cases} \quad (5.4)$$

where $Z_1'$ and $Z_2'$ are respectively, the predetermined values of minimum profit and maximum transportation time.

## 5.3 Crisp Equivalents of the Models

The crisp equivalent of EVM is provided below in Theorem 5.3.1.

**Theorem 5.3.1**: Let $\xi_{c_{ijk}^p}, \xi_{f_{ijk}^p}, \xi_{t_{ijk}^p}, \xi_{a_i^p}, \xi_{b_j^p}, \xi_{e_k}$ and $\xi_{B_j}$ are the independent uncertain variables, having uncertainty distributions $\Phi_{\xi_{c_{ijk}^p}}, \Phi_{\xi_{f_{ijk}^p}}, \Phi_{\xi_{t_{ijk}^p}}, \Phi_{\xi_{a_i^p}}, \Phi_{\xi_{b_j^p}}, \Phi_{\xi_{e_k}}$ and $\Phi_{\xi_{B_j}}$, respectively. Then the crisp equivalent of EVM is presented below in model (5.5).

$$\begin{cases} Max\ E[Z_1] = \sum_{p=1}^{r}\sum_{i=1}^{m}\sum_{j=1}^{n}\sum_{k=1}^{K}\left\{\left(s_j^p - v_i^p - \int_0^1 \Phi^{-1}_{\xi_{c_{ijk}^p}}(\alpha_1)d\alpha_1\right)x_{ijk}^p - \int_0^1 \Phi^{-1}_{\xi_{f_{ijk}^p}}(\alpha_1)d\alpha_1\, y_{ijk}^p\right\} \\ Min\ E[Z_2] = \sum_{p=1}^{r}\sum_{i=1}^{m}\sum_{j=1}^{n}\sum_{k=1}^{K}\int_0^1 \Phi^{-1}_{\xi_{t_{ijk}^p}}(\alpha_2)d\alpha_2\, y_{ijk}^p \\ subject\ to \\ \quad \sum_{j=1}^{n}\sum_{k=1}^{K}x_{ijk}^p - \int_0^1 \Phi^{-1}_{\xi_{a_i^p}}\left(1-\beta_i^p\right)d\beta_i^p \leq 0, i = 1,2,\dots,m, p = 1,2,\dots,r \\ \quad \sum_{i=1}^{m}\sum_{k=1}^{K}x_{ijk}^p - \int_0^1 \Phi^{-1}_{\xi_{b_j^p}}\left(\gamma_j^p\right)d\gamma_j^p \geq 0, j = 1,2,\dots,n, p = 1,2,\dots,r \\ \quad \sum_{p=1}^{r}\sum_{i=1}^{m}\sum_{j=1}^{n}x_{ijk}^p - \int_0^1 \Phi^{-1}_{\xi_{e_k}}(1-\delta_k)\,d\delta_k \leq 0, k = 1,2,\dots,K \\ \quad \sum_{p=1}^{r}\sum_{i=1}^{m}\sum_{k=1}^{K}\left\{\left(v_i^p + \int_0^1 \Phi^{-1}_{\xi_{c_{ijk}^p}}(\rho_j)\,d\rho_j\right)x_{ijk}^p + \int_0^1 \Phi^{-1}_{\xi_{f_{ijk}^p}}(\rho_j)\,d\rho_j y_{ijk}^p\right\} - \int_0^1 \Phi^{-1}_{\xi_{B_j}}\left(1-\rho_j\right)d\rho_j \leq 0, \\ \qquad\qquad j = 1,2,\dots,n \\ \quad x_{ijk}^p \geq 0,\ y_{ijk}^p = \begin{cases} 1 & ; if\ x_{ijk}^p > 0 \\ 0 & ; otherwise \end{cases} \quad \forall\, p,i,j,k. \end{cases} \quad (5.5)$$

**Proof** From the linearity property of the expected value operator, the crisp equivalent of EVM in (5.5) is formulated as follows.

$$\begin{cases} Max\ E[Z_1] = \sum_{p=1}^{r}\sum_{i=1}^{m}\sum_{j=1}^{n}\sum_{k=1}^{K}\left\{\left(s_j^p - v_i^p - E\left[\xi_{c_{ijk}^p}\right]\right)x_{ijk}^p - E\left[\xi_{f_{ijk}^p}\right]y_{ijk}^p\right\} \\ Min\ E[Z_2] = \sum_{p=1}^{r}\sum_{i=1}^{m}\sum_{j=1}^{n}\sum_{k=1}^{K} \mathrm{E}\left[\xi_{t_{ijk}^p}\right]y_{ijk}^p \\ subject\ to \\ \quad \sum_{j=1}^{n}\sum_{k=1}^{K}x_{ijk}^p - E\left[\xi_{a_i^p}\right] \le 0, i = 1,2,\dots,m, p = 1,2,\dots,r \\ \quad \sum_{i=1}^{m}\sum_{k=1}^{K}x_{ijk}^p - E\left[\xi_{b_j^p}\right] \ge 0, j = 1,2,\dots,n, p = 1,2,\dots,r \\ \quad \sum_{p=1}^{r}\sum_{i=1}^{m}\sum_{j=1}^{n}x_{ijk}^p - E\left[\xi_{e_k}\right] \le 0, k = 1,2,\dots,K \\ \quad \sum_{p=1}^{r}\sum_{i=1}^{m}\sum_{k=1}^{K}\left\{\left(v_i^p + E\left[\xi_{c_{ijk}^p}\right]\right)x_{ijk}^p + E\left[\xi_{f_{ijk}^p}\right]y_{ijk}^p\right\} - E\left[\xi_{B_j}\right] \le 0, j = 1,2,\dots,n \\ \quad x_{ijk}^p \ge 0,\ y_{ijk}^p = \begin{cases} 1 & ; if\ x_{ijk}^p > 0 \\ 0 & ; otherwise \end{cases} \quad \forall\, p,i,j,k. \end{cases} \tag{5.6}$$

Following Theorem 1.3.9, model (5.6) can be written as model (5.5).

**Corollary 5.3.1**: If $\xi_{c_{ijk}^p}, \xi_{f_{ijk}^p}, \xi_{t_{ijk}^p}, \xi_{a_i^p}, \xi_{b_j^p}, \xi_{e_k}$ and $\xi_{B_j}$ are the independent zigzag uncertain variables of the form $\mathcal{Z}(g,h,l)$ with $g,h,l \in \Re$ and $g < h < l$, then the crisp equivalent of model (5.5) can be written as shown in model (5.7).

$$\begin{cases} Max\ E[Z_1] = \sum_{p=1}^{r}\sum_{i=1}^{m}\sum_{j=1}^{n}\sum_{k=1}^{K}\left\{\left(s_j^p - v_i^p - \frac{\left(g_{c_{ijk}^p}+2h_{c_{ijk}^p}+l_{c_{ijk}^p}\right)}{4}\right)x_{ijk}^p - \frac{\left(g_{f_{ijk}^p}+2h_{f_{ijk}^p}+l_{f_{ijk}^p}\right)}{4}y_{ijk}^p\right\} \\ Min\ E[Z_2] = \sum_{p=1}^{r}\sum_{i=1}^{m}\sum_{j=1}^{n}\sum_{k=1}^{K}\frac{\left(g_{t_{ijk}^p}+2h_{t_{ijk}^p}+l_{t_{ijk}^p}\right)}{4}y_{ijk}^p \\ subject\ to \\ \quad \sum_{j=1}^{n}\sum_{k=1}^{K}x_{ijk}^p - \frac{\left(g_{a_i^p}+2h_{a_i^p}+l_{a_i^p}\right)}{4} \le 0, i = 1,2,\dots,m, p = 1,2,\dots,r \\ \quad \sum_{i=1}^{m}\sum_{k=1}^{K}x_{ijk}^p - \frac{\left(g_{b_j^p}+2h_{b_j^p}+l_{b_j^p}\right)}{4} \ge 0, j = 1,2,\dots,n, p = 1,2,\dots,r \\ \quad \sum_{p=1}^{r}\sum_{i=1}^{m}\sum_{j=1}^{n}x_{ijk}^p - \frac{\left(g_{e_k}+2h_{e_k}+l_{e_k}\right)}{4} \le 0, k = 1,2,\dots,K \\ \quad \sum_{p=1}^{r}\sum_{i=1}^{m}\sum_{k=1}^{K}\left\{\left(v_i^p + \frac{\left(g_{c_{ijk}^p}+2h_{c_{ijk}^p}+l_{c_{ijk}^p}\right)}{4}\right)x_{ijk}^p + \frac{\left(g_{f_{ijk}^p}+2h_{f_{ijk}^p}+l_{f_{ijk}^p}\right)}{4}y_{ijk}^p\right\} - \frac{\left(g_{B_j}+2h_{B_j}+l_{B_j}\right)}{4} \le 0, \\ \qquad j = 1,2,\dots,n \\ \quad x_{ijk}^p \ge 0,\ y_{ijk}^p = \begin{cases} 1 & ; if\ x_{ijk}^p > 0 \\ 0 & ; otherwise \end{cases} \quad \forall\, p,i,j,k. \end{cases} \tag{5.7}$$

**Corollary 5.3.2**: If $\xi_{c_{ijk}^p}, \xi_{f_{ijk}^p}, \xi_{t_{ijk}^p}, \xi_{a_i^p}, \xi_{b_j^p}, \xi_{e_k^p}$ and $\xi_{B_j^p}$ are the independent normal uncertain variables of the form $\mathcal{N}(\mu,\sigma)$, where $\mu,\sigma \in \Re$ and $\sigma > 0$, then model (5.5) can be presented as follows.

$$
\begin{cases}
Max\ E[Z_1] = \sum_{p=1}^{r}\sum_{i=1}^{m}\sum_{j=1}^{n}\sum_{k=1}^{K}\left\{\left(s_j^p - v_i^p - \mu_{c_{ijk}^p}\right)x_{ijk}^p - \mu_{f_{ijk}^p}y_{ijk}^p\right\} \\
Min\ E[Z_2] = \sum_{p=1}^{r}\sum_{i=1}^{m}\sum_{j=1}^{n}\sum_{k=1}^{K}\mu_{t_{ijk}^p}y_{ijk}^p \\
subject\ to \\
\quad \sum_{j=1}^{n}\sum_{k=1}^{K}x_{ijk}^p - \mu_{a_i^p} \le 0, i = 1,2,\dots,m, p = 1,2,\dots,r \\
\quad \sum_{i=1}^{m}\sum_{k=1}^{K}x_{ijk}^p - \mu_{b_j^p} \ge 0\ , j = 1,2,\dots,n, p = 1,2,\dots,r \\
\quad \sum_{p=1}^{r}\sum_{i=1}^{m}\sum_{j=1}^{n}x_{ijk}^p - \mu_{e_k} \le 0,\ k = 1,2,\dots,K \\
\quad \sum_{p=1}^{r}\sum_{i=1}^{m}\sum_{k=1}^{K}\left\{\left(v_i^p + \mu_{c_{ijk}^p}\right)x_{ijk}^p + \mu_{f_{ijk}^p}y_{ijk}^p\right\} - \mu_{B_j} \le 0,\ j = 1,2,\dots,n \\
\quad x_{ijk}^p \ge 0,\ y_{ijk}^p = \begin{cases} 1 & ; if\ x_{ijk}^p > 0 \\ 0 & ; otherwise \end{cases} \qquad \forall\, p,i,j,k.
\end{cases} \tag{5.8}
$$

Some related theorems for the crisp equivalents of CCM and DCCM, and their corresponding proofs and corollaries are presented in Appendix C.

## 5.4 Methodologies for Crisp Equivalence

In this section, we have considered three different multi-objective programming techniques: (i) linear weighted method, (ii) global criterion method and (iii) fuzzy programming method. These methods are used to generate the compromise solutions of the crisp equivalents of EVM, CCM and DCCM of the proposed UMMFSTPwB.

### 5.4.1 Linear Weighted Method

The linear weighted method (cf. Section 1.3.15.1) converts a multi-objective optimization problem to its equivalent SOOP by the weighted sum of the objective functions, where the weights are the relative importance of the objectives as determined by the DM. Here, the crisp equivalent multi-objective models, (5.5) and (C1) (cf. Appendix C) are converted to their equivalent compromise SOOPs by the linear weighted method and are represented in models (5.9) and (5.10), respectively. For DCCM, model (C5) (cf. Appendix C), is converted to equivalent compromise SOOP as presented in the model (5.11).

$$
\begin{cases}
Min\ \{-\lambda_1 E[Z_1] + \lambda_2 E[Z_2]\} \\
subject\ to\ the\ constraints\ of\ (5.5) \\
\lambda_1 + \lambda_2 = 1,\ \lambda_1, \lambda_2 \in [0,1].
\end{cases} \tag{5.9}
$$

$$
\begin{cases}
Min\ \{-\lambda_1 \bar{Z}_1 + \lambda_2 \bar{Z}_2\} \\
subject\ to\ the\ constraints\ of\ (\mathrm{C1})\ in\ \text{Appendix C} \\
\lambda_1 + \lambda_2 = 1,\ \lambda_1, \lambda_2 \in [0,1].
\end{cases} \tag{5.10}
$$

$$
\begin{cases}
Max\ \{\lambda_1 v_{Z_1'} + \lambda_2 v_{Z_2'}\} \\
subject\ to\ the\ constraints\ of\ (\mathrm{C1})\ in\ \text{Appendix C} \\
\lambda_1 + \lambda_2 = 1,\ \lambda_1, \lambda_2 \in [0,1].
\end{cases} \tag{5.11}
$$

**Theorem 5.4.1**: A feasible solution of the crisp equivalent of EVM in (5.5) is

(i) an optimal solution of the compromise model (5.9), if it is Pareto optimal to the multi-objective model (5.5),

(ii) a Pareto optimal solution of the multi-objective model (5.5), if it is an optimal solution of the compromise model (5.9).

**Proof**

(i) Let $t^*$ is the optimal solution of the compromise model (5.9), which is not Pareto optimal to multi-objective model (5.5). Then, there exists a Pareto optimal solution $t$, which dominates $t^*$ or $t \prec t^*$. It follows, $(-\lambda_1 E[Z_1^t] + \lambda_2 E[Z_2^t]) < \left(-\lambda_1 E\left[Z_1^{t^*}\right] + \lambda_2 E\left[Z_2^{t^*}\right]\right)$, where $\lambda_1 + \lambda_2 = 1,\ \lambda_1, \lambda_2 \in [0,1]$. This eventually implies that $t^*$ is not an optimal solution of model (5.9) which contradicts our previous hypothesis that $t^*$ is the optimal solution of model (5.9).

(ii) Let $t^*$ be the Pareto optimal solution of model (5.5), which is not an optimal solution of the model (5.9). Then there exist an optimal solution $t'$ of model (5.9) such that $\left(-\lambda_1 E\left[Z_1^{t'}\right] + \lambda_2 E\left[Z_2^{t'}\right]\right) < \left(-\lambda_1 E\left[Z_1^{t^*}\right] + \lambda_2 E\left[Z_2^{t^*}\right]\right)$, where $\lambda_1, \lambda_2 \in [0,1]$ and $\lambda_1 + \lambda_2 = 1$. This implies $t'$ is Pareto optimal to model (5.5). It contradicts our initial hypothesis that $t^*$ is the Pareto optimal solution of model (5.5).

Similar proofs can be done for the multi-objective model (C1) and its compromise model (5.10) of CCM, and for the model (C5) and its corresponding compromise model (5.11) of DCCM.

## 5.4.2 Global Criterion Method

The global criterion method (cf. Section 1.3.15.5) transforms a multi-objective optimization problem to its equivalent SOOP and minimizes the sum of deviation of the ideal solutions from the corresponding objective functions. Here, we use the global criterion method in $L_2$ norm to convert the crisp equivalent multi-objective models (5.5), (C1) and (C5) to their corresponding equivalent compromise SOOPs, as presented below in (5.12),(5.13) and (5.14).

$$\begin{cases} Min\left\{\sqrt{\left(\frac{E[Z_1]^{max}-E[Z_1]}{E[Z_1]^{max}}\right)^2 + \left(\frac{E[Z_2]-E[Z_2]^{min}}{E[Z_2]^{min}}\right)^2}\right\} \\ subject\ to\ the\ constraints\ of\ (5.5), \end{cases} \tag{5.12}$$

where $E[Z_1]^{max}$ and $E[Z_2]^{min}$ are the ideal objective values of the crisp equivalent of EVM in (5.5).

$$\begin{cases} Min\left\{\sqrt{\left(\frac{\bar{Z}_1^{max}-\bar{Z}_1}{\bar{Z}_1^{max}}\right)^2 + \left(\frac{\bar{Z}_2-\bar{Z}_2^{min}}{\bar{Z}_2^{min}}\right)^2}\right\} \\ subject\ to\ the\ constraints\ of\ (C1)\ in\ \text{Appendix C}, \end{cases} \tag{5.13}$$

where $\overline{Z}_1^{max}$ and $\overline{Z}_2^{min}$ are the ideal objective values of the crisp equivalent of CCM in (C1).

$$\begin{cases} Min\left\{\sqrt{\left(\frac{v_{Z_1'}^{max}-v_{Z_1'}}{v_{Z_1'}^{max}}\right)^2+\left(\frac{v_{Z_2'}^{max}-v_{Z_2'}}{v_{Z_2'}^{max}}\right)^2}\right\} \\ subject\ to\ the\ constraints\ of\ (\text{C1})\ in\ \text{Appendix C}, \end{cases} \tag{5.14}$$

where $v_{Z_1'}^{max}$ and $v_{Z_2'}^{max}$ are the ideal objective values of the crisp equivalent of DCCM in (C5).

**Theorem 5.4.2**: A feasible solution of the crisp equivalent of EVM in (5.5) is

(i) an optimal solution of the compromise model (5.12), if it is Pareto optimal to the multi-objective model (5.5),

(ii) a Pareto optimal solution of the multi-objective model (5.5), if it is an optimal solution of compromise model (5.12).

**Proof**

(i) Let $t^*$ is the optimal solution of compromise model (5.12) which is not Pareto optimal to multi-objective model (5.5). Then, there exists a Pareto optimal solution $t$, such that $t$ dominates ($\prec$) $t^*$. This implies

$$\sqrt{\left(\frac{E[Z_1]^{max}-E[Z_1^t]}{E[Z_1]^{max}}\right)^2+\left(\frac{E[Z_2^t]-E[Z_2]^{min}}{E[Z_2]^{min}}\right)^2} < \sqrt{\left(\frac{E[Z_1]^{max}-E[Z_1^{t^*}]}{E[Z_1]^{max}}\right)^2+\left(\frac{E[Z_2^{t^*}]-E[Z_2]^{min}}{E[Z_2]^{min}}\right)^2}.$$

It implies that $t^*$ is not the optimal solution of (5.12), which directly contradicts our previous assumption that $t^*$ is the optimal solution of (5.12).

(ii) Let $t^*$ be the Pareto optimal solution of (5.5) which is not an optimal solution of (5.12). Then there exists an optimal solution $t'$ of (5.12) such that

$$\sqrt{\left(\frac{E[Z_1]^{max}-E[Z_1^{t'}]}{E[Z_1]^{max}}\right)^2+\left(\frac{E[Z_2^{t'}]-E[Z_2]^{min}}{E[Z_2]^{min}}\right)^2} < \sqrt{\left(\frac{E[Z_1]^{max}-E[Z_1^{t^*}]}{E[Z_1]^{max}}\right)^2+\left(\frac{E[Z_2^{t^*}]-E[Z_2]^{min}}{E[Z_2]^{min}}\right)^2}.$$

Therefore, it follows that $t'$ is Pareto optimal to (5.5), which contradicts with our previous assumption that $t^*$ is the Pareto optimal solution of the model (5.5).

Similar proofs can be done for the multi-objective model (C1) and its compromise model (5.13) of CCM, and for the model (C5) and its compromise model (5.14) of DCCM.

### 5.4.3 Fuzzy Programming Method

H.-J. Zimmermann (1978) presented the fuzzy programming method (cf. Section 1.3.15.4) for solving the multi-objective problem. Here, we have applied the fuzzy programming method on crisp equivalent multi-objective models, (5.5) for EVM presented above, (C1) for CCM and (C5) for DCCM, and convert those models to their equivalent compromise SOOPs by introducing an auxiliary variable $\lambda$. These compromise models are represented, respectively in (5.15), (5.16) and (5.17).

$$\begin{cases} Max\ \lambda \\ subject\ to \\ \left(\frac{E[Z_1]-E[Z_1]^{LB_1}}{E[Z_1]^{UB_1}-E[Z_1]^{LB_1}}\right) \geq \lambda \\ \left(\frac{E[Z_2]^{UB_2}-E[Z_2]}{E[Z_2]^{UB_2}-E[Z_2]^{LB_2}}\right) \geq \lambda \\ constraints\ of\ (5.5), \end{cases} \tag{5.15}$$

where $E[Z_1]^{UB_1}$ and $E[Z_2]^{LB_2}$ are respectively, the expected optimal solutions of $E[Z_1]$ and $E[Z_2]$, $E[Z_1]^{LB_1}$ is the expected lower bound of $E[Z_1]$ corresponding to $E[Z_2]^{LB_2}$, and $E[Z_2]^{UB_2}$ is the expected upper bound of $E[Z_2]$ corresponding to $E[Z_1]^{UB_1}$.

$$\begin{cases} Max\ \lambda \\ subject\ to \\ \left(\frac{\overline{Z}_1-\overline{Z}_1^{\mathrm{LB}_1}}{\overline{Z}_1^{\mathrm{UB}_1}-\overline{Z}_1^{\mathrm{LB}_1}}\right) \geq \lambda \\ \left(\frac{\overline{Z}_2^{\mathrm{UB}_2}-\overline{Z}_2}{\overline{Z}_2^{\mathrm{UB}_2}-\overline{Z}_2^{\mathrm{LB}_2}}\right) \geq \lambda \\ constraints\ of\ (\mathrm{C1})\ in\ \text{Appendix C}, \end{cases} \tag{5.16}$$

where $\overline{\mathrm{Z}}_1^{\mathrm{UB}_1}$ and $\overline{\mathrm{Z}}_2^{\mathrm{LB}_2}$ are respectively, the optimal solutions of $\overline{Z}_1$ and $\overline{Z}_2$, $\overline{Z}_1^{\mathrm{LB}_1}$ is the lower bound of $\overline{Z}_1$ corresponding to $\overline{\mathrm{Z}}_2^{\mathrm{LB}_2}$, and $\overline{\mathrm{Z}}_2^{\mathrm{UB}_2}$ is the upper bound of $\overline{Z}_2$ corresponding to $\overline{\mathrm{Z}}_1^{\mathrm{UB}_1}$.

$$\begin{cases} Max\ \lambda \\ subject\ to \\ \left(\frac{v_{Z_1'}-v_{Z_1'}^{\mathrm{LB}_1}}{v_{Z_1'}^{\mathrm{UB}_1}-v_{Z_1'}^{\mathrm{LB}_1}}\right) \geq \lambda \\ \left(\frac{v_{Z_2'}-v_{Z_2'}^{\mathrm{LB}_2}}{v_{Z_2'}^{\mathrm{UB}_2}-v_{Z_2'}^{\mathrm{LB}_2}}\right) \geq \lambda \\ constraints\ of\ (\mathrm{C1})\ in\ \text{Appendix C}, \end{cases} \tag{5.17}$$

where $v_{Z_1'}^{\mathrm{UB}_1}$ and $v_{Z_2'}^{\mathrm{UB}_2}$ are respectively, the optimal solutions of $v_{Z_1'}$ and $v_{Z_2'}$, $v_{Z_1'}^{\mathrm{LB}_1}$ is the lower bound of $v_{Z_1'}$ corresponding to $v_{Z_2'}^{\mathrm{UB}_2}$, and $v_{Z_2'}^{\mathrm{LB}_2}$ is the lower bound of $v_{Z_2'}$ corresponding to $v_{Z_1'}^{\mathrm{UB}_1}$.

**Theorem 5.4.3**: A feasible solution of the crisp equivalent of EVM in (5.5) is

(i) an optimal solution of compromise model (5.15), if it is Pareto optimal to multi-objective model (5.5),

(ii) a Pareto optimal solution of the multi-objective model (5.5), if it is optimal to compromise model (5.15).

**Proof**

(i) Let $t^*$ is the optimal solution of compromise model (5.15) which is not the Pareto optimal to multi-objective model (5.5). Then, there exists a Pareto optimal solution $t$, such that $t$ dominates $(\prec)$ $t^*$. It implies

$$\frac{E[Z_1^t]-E[Z_1]^{LB_1}}{E[Z_1]^{UB_1}-E[Z_1]^{LB_1}} > \frac{E\left[Z_1^{t^*}\right]-E[Z_1]^{LB_1}}{E[Z_1]^{UB_1}-E[Z_1]^{LB_1}} \text{ and } \frac{E[Z_2]^{UB_2}-E[Z_2^t]}{E[Z_2]^{UB_2}-E[Z_2]^{LB_2}} < \frac{E[Z_2]^{UB_2}-E\left[Z_2^{t^*}\right]}{E[Z_2]^{UB_2}-E[Z_2]^{LB_2}}$$

$\Leftrightarrow \eta_1(E[Z_1^t]) > \eta_1\left(E\left[Z_1^{t^*}\right]\right)$ and $\eta_2(E[Z_2^t]) < \eta_2\left(E\left[Z_2^{t^*}\right]\right)$,

where $$\eta_1(E[Z_1^T]) = \begin{cases} 1 & ; if\ E[Z_1]^{UB_1} \le E[Z_1^T] \\ \frac{E[Z_1^T]-E[Z_1]^{LB_1}}{E[Z_1]^{UB_1}-E[Z_1]^{LB_1}} & ; if\ E[Z_1]^{LB_1} < E[Z_1^T] < E[Z_1]^{UB_1} \\ 0 & ; if\ E[Z_1^T] \le E[Z_1]^{LB_1} \end{cases}$$

and

$$\eta_2(E[Z_2^T]) = \begin{cases} 1 & ; if\ E[Z_2^T] \le E[Z_2]^{LB_2} \\ \frac{E[Z_2]^{UB_2}-E[Z_2^T]}{E[Z_2]^{UB_2}-E[Z_2]^{LB_2}} & ; if\ E[Z_2]^{LB_2} < E[Z_2^T] < E[Z_2]^{UB_2} \\ 0 & ; if\ E[Z_2]^{UB_2} \le E[Z_2^T] \end{cases}$$

for $T \in \{t, t^*\}$.

This means there exists a $\lambda$, such that $\lambda > \lambda^*$. Therefore, it follows $t^*$ is not the optimal solution of the model (5.15) which contradicts our initial assumption that $t^*$ is the optimal solution of the model (5.15).

(ii) Let $t^*$ be the Pareto optimal solution of model (5.5), which is not an optimal solution of model (15.1). Then there exists an optimal solution $t'$ of (5.15) such that $\eta_1\left(E\left[Z_1^{t'}\right]\right) > \eta_1\left(E\left[Z_1^{t^*}\right]\right)$ and $\eta_2\left(E\left[Z_2^{t'}\right]\right) < \eta_2\left(E\left[Z_2^{t^*}\right]\right)$. Therefore,

$\frac{E\left[Z_1^{t'}\right]-E[Z_1]^{LB_1}}{E[Z_1]^{UB_1}-E[Z_1]^{LB_1}} > \frac{E\left[Z_1^{t^*}\right]-E[Z_1]^{LB_1}}{E[Z_1]^{UB_1}-E[Z_1]^{LB_1}}$ and $\frac{E[Z_2]^{UB_2}-E\left[Z_2^{t'}\right]}{E[Z_2]^{UB_2}-E[Z_2]^{LB_2}} < \frac{E[Z_2]^{UB_2}-E\left[Z_2^{t^*}\right]}{E[Z_2]^{UB_2}-E[Z_2]^{LB_2}}$. Hence, $t^*$ is not the Pareto optimal solution of model (5.5) which contradicts our initial hypothesis that $t^*$ is the Pareto optimal solution of (5.5).

Similar proofs can be done for the multi-objective model (C1) and its compromise model (5.16) of CCM, and for the multi-objective model (C5) and its respective compromise model (5.17) of DCCM.

## 5.5 Results and Discussion

This section presents the results of three models: (i) EVM, (ii) CCM and (iii) DCCM of the proposed UMMFSTPwB problem using three compromise programming methods: (i) linear weighted method, (ii) global criterion method and (iii) fuzzy programming method. For numerical illustration of the proposed UMMFSTPwB, we have considered shipment of two items from two sources to three destinations using two types of conveyance, i.e., $r = 2$, $m = 2$, $n = 3$ and $K = 2$. The values of all the input parameters are provided in 9 tables, Table D.1 through Table D.9 in Appendix D. We have used the standard optimization software, LINGO 11.0, to determine the results of three models.

We have presented the results for the crisp equivalent of EVM and CCM followed by the results of DCCM along with their transportation plans. For CCM, $cl_1$ and $cl_2$ represent all the chance levels having the values within the interval $[0,0.5)$ and $[0.5, 1.0]$, respectively. For $cl_1$, the values of chance levels are set as $\alpha_1 = 0.4, \alpha_2 = 0.4,\ \beta_i^p = 0.45, \gamma_j^p = 0.35, \delta_k = 0.45$ and $\rho_j = 0.4\ \forall\ i,j,k,p.$ Whereas, for $cl_2$, the chance level values are set as $\alpha_1 = 0.9, \alpha_2 = 0.9, \beta_i^p = 0.8, \gamma_j^p = 0.75,\ \delta_k = 0.85$ and $\rho_j = 0.8\ \forall\ i,j,k,p.$ EVM and CCM optimizes the objective values, and DCCM maximizes the satisfaction levels. Hence, we have combined the results of EVM and CCM, but the results of DCCM are presented separately.

For EVM, the models, (5.7) and (5.8) are solved to generate the results, respectively for zigzag and normal uncertain variables. In case of CCM, the models, (C2) and (C3) are solved for zigzag uncertain variables at chance levels $cl_1$ and $cl_2$, respectively, and model (C4) is solved for normal uncertain variables at chance level $cl_2$ . The results of DCCM are obtained by solving models (C6) and (C7), respectively for zigzag and normal uncertain variables.

Table 5.2 shows the results of EVM and CCM for zigzag and normal uncertain variables. Here, the linear weighted method is used to generate the results of the corresponding models with different values of the weights, $\lambda_1$ and $\lambda_2$. From Table 5.2, it is observed that, as $\lambda_1$ decreases, both the maximization functions, $E[Z_1]$ and $\bar{Z}_1$, proportionately decrease their values. Similarly, the minimization functions, $E[Z_2]$ and $\bar{Z}_2$, decrease their values when $\lambda_2$ increases. Moreover, from Table 5.2 it is observed, for both the zigzag and normal uncertain variables, the results of EVM and CCM are nondominated to each other at different values of $\lambda_1$ and $\lambda_2$.

The results generated by three solution methodologies to solve the crisp equivalent models of EVM and CCM are reported in Table 5.3. Here, the values of the weights for the linear weighted method, i.e., $\lambda_1$ and $\lambda_2$ are considered the same. From Table 5.3, it is observed that, in the case of EVM, for both uncertain variables (zigzag and normal uncertain variables), the solutions generated by global criterion method and fuzzy programming method, which are the same, and the solution generated by the linear weighted method are nondominated to each other. However, for CCM and for zigzag uncertain variables, the solutions generated by the linear weighted method and global criterion method are nondominated to each other, and the solutions generated by the linear weighted method and fuzzy programming method are nondominated to each other but the solution generated by global criterion method dominates the solution of fuzzy programming method. While, for CCM with normal uncertain variables, the solutions generated by the linear weighted method, global criterion method and fuzzy programming method are nondominated to each other. Here, Table 5.4 shows the transportation plans corresponding to the results provided in Table 5.3, for EVM and CCM.

Table 5.2 Optimum results of EVM and CCM for zigzag and normal uncertain variables using linear weighted method

| | Weights | | EVM | | CCM | | | |
|---|---|---|---|---|---|---|---|---|
| | | | | | $cl_1$ | | $cl_2$ | |
| | $\lambda_1$ | $\lambda_2$ | $E[Z_1]$ | $E[Z_2]$ | $\bar{Z}_1$ | $\bar{Z}_2$ | $\bar{Z}_1$ | $\bar{Z}_2$ |
| Zigzag uncertain variables | 1.0 | 0.0 | 526.125 | 87.500 | 572.327 | 86.719 | 346.414 | 102.481 |
| | 0.7 | 0.3 | 526.125 | 87.500 | 570.600 | 85.120 | 346.414 | 102.481 |
| | 0.5 | 0.5 | 513.250 | 75.250 | 570.600 | 85.120 | 344.440 | 100.800 |
| | 0.3 | 0.7 | 509.419 | 73.742 | 554.440 | 72.940 | 326.175 | 93.154 |
| | 0.0 | 1.0 | 334.375 | 69.700 | 315.980 | 62.420 | 268.580 | 90.740 |
| Normal uncertain variables | | | | | $cl_2$ | | | |
| | | | | | $\bar{Z}_1$ | | $\bar{Z}_2$ | |
| | 1.0 | 0.0 | 576.185 | 89.290 | 467.172 | | 98.741 | |
| | 0.7 | 0.3 | 573.769 | 87.912 | 456.383 | | 97.279 | |
| | 0.5 | 0.5 | 572.848 | 87.800 | 456.147 | | 95.584 | |
| | 0.3 | 0.7 | 535.400 | 65.114 | 404.117 | | 84.650 | |
| | 0.0 | 1.0 | 427.529 | 59.600 | 347.127 | | 81.525 | |

Table 5.3 Comparative results of three different solution methodologies

| | Solution method | Model | | | | | |
|---|---|---|---|---|---|---|---|
| | | EVM | | CCM | | | |
| | | | | $cl_1$ | | $cl_2$ | |
| | | $E[Z_1]$ | $E[Z_2]$ | $\bar{Z}_1$ | $\bar{Z}_2$ | $\bar{Z}_1$ | $\bar{Z}_2$ |
| Zigzag uncertain variables | Linear weighted method (with equal weights) | 513.250 | 75.250 | 570.600 | 85.120 | 344.440 | 100.800 |
| | Global criterion method | 495.750 | 72.425 | 542.220 | 70.180 | 315.500 | 94.860 |
| | Fuzzy programming method | 495.750 | 72.425 | 483.558 | 70.180 | 313.372 | 94.860 |
| Normal uncertain variables | | | | $cl_2$ | | | |
| | | | | $\bar{Z}_1$ | | $\bar{Z}_2$ | |
| | Linear weighted method (with equal weights) | 572.848 | 87.800 | 456.147 | | 95.584 | |
| | Global criterion method | 516.736 | 69.200 | 404.117 | | 84.652 | |
| | Fuzzy programming method | 516.736 | 69.200 | 409.800 | | 87.736 | |

For DCCM, the results for three solution methodologies are shown in Table 5.5. Similar to Table 5.3, the associated weights of the linear weighted method in Table 5.5 are considered equal. Moreover, to solve the crisp equivalents of DCCM, the predetermined limits of $Z_1'$ and $Z_2'$, are set to, $325$ and $94$, respectively, for the model $(C6)$, and $419$ and $83$, respectively, for the model $(C7)$. It is observed from Table 5.5 that, for zigzag uncertain variables, the solution generated by the global criterion method dominates each of the solutions generated by the linear weighted method and fuzzy programming method. Moreover, the solution generated by the fuzzy programming method dominates the solution generated by the linear weighted method. However, for normal uncertain variables, the solutions generated by the linear weighted method and global criterion method, which are the same, and the solution generated by the fuzzy programming method are nondominated to each other. Here, Table 5.6 shows the transportation plans corresponding to the results presented in Table 5.5, of DCCM, for zigzag and normal uncertain variables.

Table 5.4 Optimal transportation plan of the EVM and CCM for normal uncertain variables using three different solution methodologies

| | Model | | Solution method | | |
|---|---|---|---|---|---|
| | | | Linear weighted method (with equal weights) | Global criterion method | Fuzzy programming method |
| Zigzag uncertain variables | EVM | | $x^1_{121}$=19.00, $x^1_{132}$=13.50, $x^1_{212}$=31.50, $x^1_{232}$=5.75, $x^2_{121}$ =8.60,$x^2_{131}$=17.50, $x^2_{212}$= 13.73, $x^2_{222}$=15.00 | $x^1_{112}$=13.50, $x^1_{121}$=19.00, $x^1_{212}$=18.00, $x^1_{232}$=19.25, $x^2_{131}$=26.00,$x^2_{212}$=13.75, $x^2_{222}$=15.00 | $x^1_{112}$=13.50, $x^1_{121}$=19.00, $x^1_{212}$=18.00, $x^1_{232}$=19.25, $x^2_{131}$=26.00,$x^2_{212}$=13.75, $x^2_{222}$=15.00 |
| | CCM | $cl_1$ | $x^1_{121}$=18.60, $x^1_{132}$= 13.70,$x^1_{212}$=32.70, $x^1_{232}$=4.70, $x^2_{121}$=8.60, $x^2_{131}$=17.50, $x^2_{212}$= 23.70, $x^2_{222}$=5.50 | $x^1_{112}$=13.70, $x^1_{121}$=18.60, $x^1_{212}$=19.00, $x^1_{232}$=18.40, $x^2_{131}$=26.10,$x^2_{212}$=15.10, $x^2_{222}$=14.10 | $x^1_{112}$=16.20, $x^1_{121}$=16.10, $x^1_{212}$=11.428, $x^1_{232}$= 18.40, $x^2_{131}$=26.10,$x^2_{212}$= 15.10, $x^2_{222}$=14.10 |
| | | $cl_2$ | $x^1_{121}$=19.00, $x^1_{132}$= 12.40,$x^1_{212}$=72.10, $x^1_{232}$=8.10, $x^2_{121}$=5.40, $x^2_{131}$=20. 00, $x^2_{212}$= 16.10, $x^2_{222}$=11.10 | $x^1_{121}$=20.60, $x^1_{132}$=10.80, $x^1_{212}$=25.50, $x^1_{232}$=9.70, $x^2_{122}$ =3.30,$x^2_{131}$=22.10, $x^2_{212}$= 14.00, $x^2_{222}$=13.20 | $x^1_{121}$=20.60, $x^1_{132}$=10.80, $x^1_{212}$=25.50, $x^1_{232}$=9.70, $x^2_{122}$=3.30,$x^2_{131}$=21.43, $x^2_{212}$=14.00, $x^2_{222}$=13.20 |
| Normal uncertain variables | EVM | | $x^1_{121}$=17.84, $x^1_{132}$=14.15, $x^1_{212}$=25.12, $x^1_{232}$=9.87, $x^2_{112}$ =11.00,$x^2_{121}$=14.00, $x^2_{211}$= 8.15, $x^2_{232}$=19.84 | $x^1_{112}$=16.00, $x^1_{121}$=16.00, $x^1_{212}$=12.15, $x^1_{232}$=22.84, $x^2_{121}$=21.21,$x^2_{212}$=12.00, $x^2_{232}$=16.00 | $x^1_{112}$=16.00, $x^1_{121}$=16.00, $x^1_{212}$=12.15, $x^1_{232}$=22.84, $x^2_{121}$=21.21,$x^2_{212}$=12.00, $x^2_{232}$=16.00 |
| | CCM | $cl_2$ | $x^1_{121}$=16.46, $x^1_{132}$=7.15, $x^1_{212}$=22.92, $x^1_{232}$= 11.41, $x^2_{112}$=9. 17,$x^2_{121}$= 14.50, $x^2_{212}$=11. 07, $x^2_{232}$=16.56 | $x^1_{112}$=10.89, $x^1_{121}$=16. 46, $x^1_{212}$=15.61, $x^1_{232}$=18.72, $x^2_{131}$=23. 67,$x^2_{212}$=13.14, $x^2_{222}$=14.50 | $x^1_{121}$=16.46, $x^1_{132}$=14. 34 ,$x^1_{212}$=30. 10, $x^1_{232}$=4.22, $x^2_{131}$=19.53,$x^2_{212}$=13.14, $x^2_{222}$=14.50 |

We have analyzed the sensitivity of CCM with input parameters of UMMFSTPwB, consider as zigzag and normal uncertain variables. The corresponding results of the CCM are reported in Table 5.7. Here, the models for both the uncertain variables (zigzag and normal uncertain variables) are solved, using the global criterion method, by changing the values of the chance levels, $\alpha_1$ and $\alpha_2$. For zigzag uncertain variables, model $(C2)$ is solved when the values of $\alpha_1$ and $\alpha_2$ change their respective values between 0.1 to 0.4. In this case, all other chance levels except $\alpha_1$ and $\alpha_2$, are set to, $\beta_i^p = 0.45, \gamma_j^p = 0.35$, $\delta_k = 0.45$ and $\rho_j = 0.4$. When the values of $\alpha_1$ and $\alpha_2$ are changed within the range from 0.5 to 1.0, model $(C3)$ is solved with the remaining chance levels set to, $\beta_i^p = 0.8, \gamma_j^p = 0.75$, $\delta_k = 0.85$ and $\rho_j = 0.8$. However, for normal uncertain variables, model $(C4)$ is solved at different values of $\alpha_1$ and $\alpha_2$ by considering the values of the remaining chance levels, the same, as set for the model $(C3)$.

Table 5.5 Results of DCCM for zigzag and normal uncertain variables

| | Solution method | Maximum satisfaction levels | |
|---|---|---|---|
| | | $\upsilon_{Z_1'}$ | $\upsilon_{Z_2'}$ |
| Zigzag uncertain variables | Linear weighted method (with equal weights) | 0.848 | 0.798 |
| | Global criterion method | 0.866 | 0.869 |
| | Fuzzy programming method | 0.857 | 0.868 |
| Normal uncertain variables | Linear weighted method (with equal weights) | 0.815 | 0.856 |
| | Global criterion method | 0.815 | 0.856 |
| | Fuzzy programming method | 0.830 | 0.747 |

From Table 5.7, we observe that, for all the models, and for both the uncertain variables (zigzag and normal uncertain variables), as the values of $\alpha_1$ and $\alpha_2$ are increased, the corresponding values of maximizing objective $\bar{Z}_1$ and minimizing objective $\bar{Z}_2$, increases and decreases, respectively.

Table 5.6 Optimal transportation plan for DCCM for normal uncertain variables

| | Solution method | | |
|---|---|---|---|
| | Linear weighted method (with equal weights) | Global criterion method | Fuzzy programming method |
| Zigzag uncertain variables | $x_{112}^1$=12.40, $x_{121}^1$=19.00, $x_{212}^1$=14.70, $x_{232}^1$=20.50, $x_{121}^2$=5.40,$x_{131}^2$=20.00, $x_{212}^2$=20.00, $x_{222}^2$=11.10 | $x_{121}^1$=20.60, $x_{132}^1$=10.80, $x_{212}^1$=25.50, $x_{232}^1$=9.70, $x_{122}^2$=3.30,$x_{131}^2$=22.10, $x_{212}^2$=14.00, $x_{222}^2$=13.20 | $x_{121}^1$=20.26, $x_{132}^1$=11.13, $x_{212}^1$=25.31, $x_{232}^1$=9. 36, $x_{122}^2$=3.30,$x_{131}^2$=22.10, $x_{212}^2$=14.0, $x_{222}^2$=13.20 |
| Normal uncertain variables | $x_{112}^1$=10.89, $x_{121}^1$=16.46, $x_{212}^1$=15.61, $x_{232}^1$=18.72, $x_{131}^2$=23. 68,$x_{212}^2$=13.13, $x_{222}^2$=14.50 | $x_{112}^1$=10.89, $x_{121}^1$=16.46, $x_{212}^1$=15.61, $x_{232}^1$=18.72, $x_{131}^2$=23. 68,$x_{212}^2$=13.1, $x_{222}^2$=14.50 | $x_{121}^1$=16.46, $x_{132}^1$=14.34, $x_{212}^1$=30.10, $x_{232}^1$=4.22, $x_{131}^2$=19.53,$x_{212}^2$=13.13, $x_{222}^2$=14.50 |

Subsequently, the values of $\bar{Z}_1$ and $\bar{Z}_2$ are also displayed graphically at different values of $\alpha_1$ and $\alpha_2$, for the models (C2) and (C3), for zigzag uncertain variables in Fig. 5.2 (a), and for the model (C4), for normal uncertain variables in Fig. 5.2 (b).

## 5.6 Conclusion

In this chapter we have designed, an uncertain multi-objective multi-item fixed charge solid transportation problem with budget constraint (UMMFSTPwB). The uncertain model has been developed by considering different uncertain parameters as zigzag and normal uncertain variables. We have formulated the three models, EVM, CCM and DCCM, for UMMFSTPwB to transform it into deterministic ones. The equivalent multi-objective deterministic models are solved using the linear weighted method, global criterion method and fuzzy programming method. Further, the features of these models are studied and some related theorems are also established. The models are numerically illustrated by analyzing the corresponding results.

Table 5.7 Results for CCM for zigzag and normal uncertain variables at different chance levels of $\alpha_1$ and $\alpha_2$

| Chance Levels | | Zigzag uncertain variables | | | | Normal uncertain variables | |
|---|---|---|---|---|---|---|---|
| | | Objective values of model (C2) | | Objective values of model (C3) | | Objective values of model (C4) | |
| $\alpha_1$ | $\alpha_2$ | $\bar{Z}_1$ | $\bar{Z}_2$ | $\bar{Z}_1$ | $\bar{Z}_2$ | $\bar{Z}_1$ | $\bar{Z}_2$ |
| 0.1 | 0.1 | 579.420 | 68.820 | -- | -- | 582.691 | 66.948 |
| 0.2 | 0.2 | 538.440 | 71.740 | -- | -- | 549.106 | 70.215 |
| 0.3 | 0.3 | 503.620 | 76.780 | -- | -- | 526.784 | 72.387 |
| 0.4 | 0.4 | 466.960 | 80.340 | -- | -- | 508.486 | 74.166 |
| 0.5 | 0.5 | -- | -- | 430.300 | 83.900 | 491.694 | 75.800 |
| 0.6 | 0.6 | -- | -- | 401.600 | 86.640 | 475.533 | 77.433 |
| 0.7 | 0.7 | -- | -- | 372.900 | 89.380 | 457.922 | 79.213 |
| 0.8 | 0.8 | -- | -- | 344.200 | 92.120 | 436.439 | 81.385 |
| 0.9 | 0.9 | -- | -- | 315.500 | 94.860 | 404.117 | 84.651 |

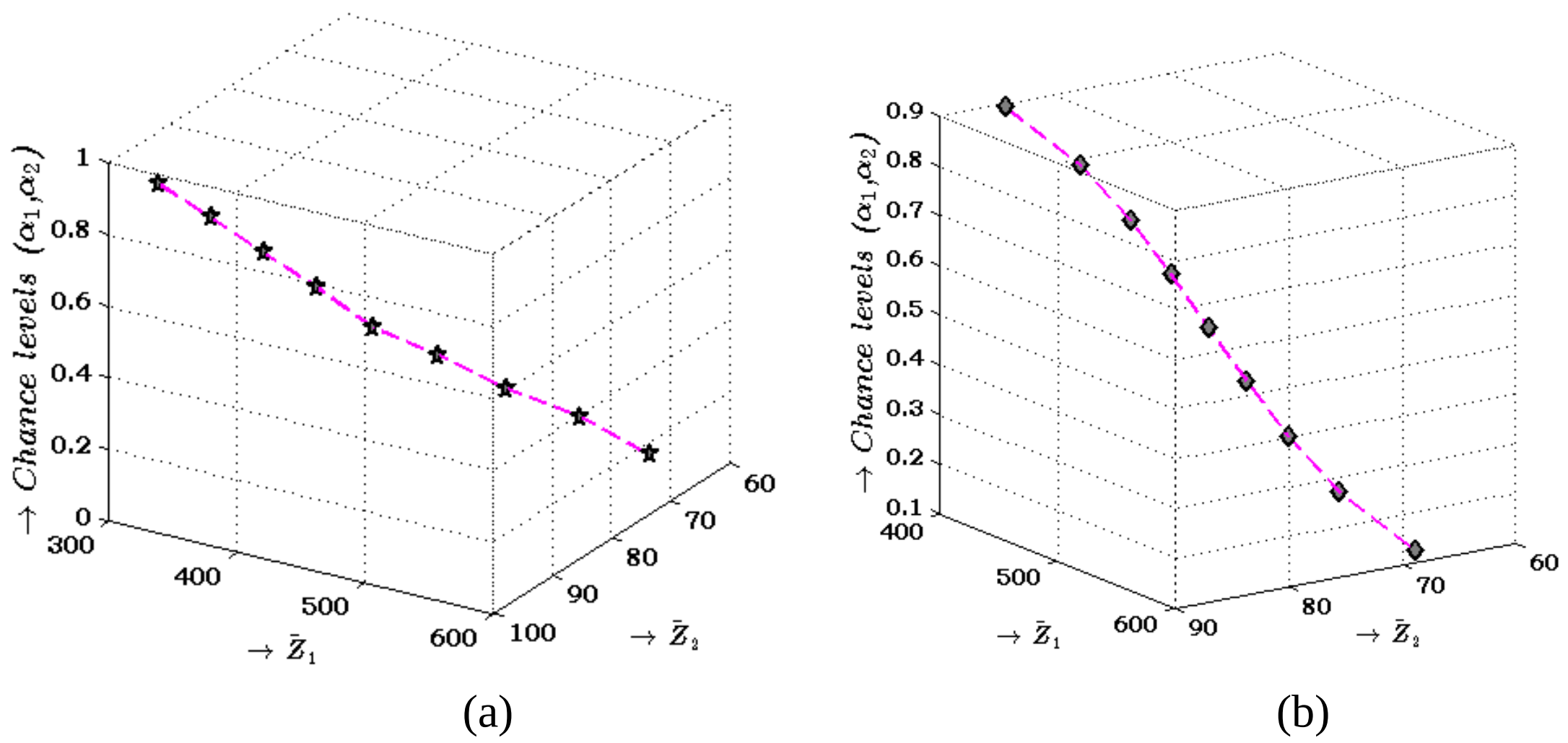

Figure 5.2 Uncertainty distribution of CCM at different chance levels of $\alpha_1$ and $\alpha_2$, for (a) zigzag uncertain variables (b) normal uncertain variables

In future, the models we have developed can be extended to include price discounts and breakable/deteriorating items. The models can also be extended for the uncertain random environment where the associated indeterminate parameters can be represented by uncertain and random variables.

# Chapter 6

# Multi-objective Rough Fuzzy Quadratic Minimum Spanning Tree Problem

# Chapter 6

# Multi-objective Rough Fuzzy Quadratic Minimum Spanning Tree Problem

## 6.1 Introduction

Minimum spanning tree (MST) problem is one of the most fundamental and important problems in network optimization which has diverse applications in different engineering and scientific domains. The MST problem is to find a minimally connected sub-graph which spans all the vertices of a connected finite network. The quadratic minimum spanning tree (QMST) problem is a variant of MST introduced by Assad and Xu (1992), which determines a minimum spanning tree of a weighted connected network by considering two types of weights: linear and quadratic. The linear weights are associated with each edge and the quadratic weights, generally arise due to the interaction effect between any pair of edges. The QMST problem has different applications (Cordone and Passeri 2012) in network design including telecommunication networks, oil or water transmission networks, transportation networks, etc. In telecommunication networks, the linear weights represent the costs to establish radio or cable connections, while the quadratic weights represent the cost of a conversion device between different types of connected cables. In a utility network (oil or water transmission networks) or a transportation network, the linear weights represent the cost of installing different pipes, and the quadratic weights represent the interaction/interfacing costs for installing valves or bending losses in T-joints. In a transportation network, the linear and quadratic weights, respectively represent the cost of constructing roads and turn penalties. In QMST, the interfacing costs mostly exist for adjacent edges. However, sometimes these interfacing costs may also exist for any pair of edges, but such topological design has little resemblance with physical layout, as in the case of fiber-optic networks (Assad and Xu 1992; Cordone and Passeri 2012).

During the last few decades, QMST has been investigated by many researchers. Assad and Xu (1992) incorporated the branch-and-bound method to determine the solutions of the problem. Zhou and Gen (1998) proposed a genetic algorithm for QMST. Palubeckis et al. (2010) presented the multi-start simulated annealing, hybrid genetic algorithm and iterative tabu search algorithm to solve QMST problem. Sundar and Singh (2010) solved it by an artificial bee colony algorithm. To solve the QMST problem, Öncan and Punnen (2010) developed an efficient local search algorithm.

Cordone and Passeri (2012) presented a tabu search and variable neighbourhood search algorithms to solve the problem. A reformulation for QMST was introduced by Pereira et al. (2013), which can compute stronger linear programming bounds. Lozano et al. (2013) proposed the solution methodology for the QMST by incorporating the strategic oscillation algorithm with tabu search. The authors presented an algorithm based on dynamic column and row generation to evaluate the bounds. To solve the problem, Rostami and Malucelli (2015) modelled some new mixed 0-1 linear formulations using a reformulation-linearization technique. Later, Pereira et al. (2015) provided a new formulation of the QMST and solved the problem using two parallel branch-and-bound algorithms. Recently, Ćustić et al. (2018) investigated the polynomially and NP-hard solvable instances of a connected network to determine QMST and its three variants: (i) quadratic bottleneck spanning tree problem, (ii) minimum spanning tree problem with conflict pair constraints and (iii) bottleneck spanning tree problem with conflict pair constraints.

All the above mentioned studies on QMST are considered as a single objective optimization problem, where the linear weights are added with the quadratic weights. The conflicting nature of the linear and the quadratic weights have been studied by Maia et al. (2013). Here, the authors have considered the linear and quadratic weights as two separate objectives for the problem. In their study, a Pareto local search algorithm has been proposed for a bi-objective QMST, where the quadratic weights are considered for any pair of adjacent edges.

In many decision making problems, the associated parameters are not always exact but are imprecise in nature due to inadequate information, lack of evidence, multiple input sources, fluctuating nature of input parameters, measurement inaccuracy, etc. Therefore, in order to process and represent the useful data hidden within the imprecise or ill-defined information, many researchers have proposed several improved theories like probability theory, fuzzy set (L.A. Zadeh 1965), rough set (Z. Pawlak 1982), and uncertainty theory (B. Liu 2007). Some applications (Zhang et al. 2003; Feng et al. 2010; Sun et al. 2014a; Wang et al. 2016; Majumder et al. 2018; Zhou et al. 2018; Rani and Garg 2018) of those theories are also observed in the literature.

Considering the uncertain MST problem, the stochastic MST has been studied by various researchers (Swamy and Shmoys 2006; Torkestani and Meybodi 2012). Similarly, some significant contributions on fuzzy MST (Janik and Kasperski 2008; Zhou et al. 2016a; Gao et al. 2018) also exist in the literature. However, unlike uncertain MST, not many investigations have been done for QMST with uncertain parameters. At present, there are only two relevant studies on QMST problem under uncertain environment. In the first study, Gao and Lu (2005) defined a QMST with linear and quadratic fuzzy weights, and solved the problem using a genetic algorithm. Later, Zhou et al. (2014b) defined a chance-constrained model for a QMST, where the linear and quadratic weights are represented by uncertain variables (B. Liu 2007).

In some decision making process, a DM may need to take decisions under a two-fold uncertain environment, where both roughness and fuzziness co-exist. In this context, the concepts of rough fuzzy sets and fuzzy rough sets, introduced by Dubois and Prade (1990), play an important role. In their study, the authors combined the fuzzy set and rough set with an aim to fulfil two different objectives. The first one is to determine the lower and upper approximations of the fuzzy set, and the second one is to replace the equivalence relation of the rough set with the fuzzy similarity relation. Since then, some important contributions on the hybridization of fuzzy and rough sets have been observed both in theory (Li and Zhang 2008; Yang and Hinde 2010; Y. Cheng 2015; M. Diker 2018; Qiao and Hu 2018) and in applications (Wu et al. 2003; Liu et al. 2014b; Sun et al. 2014b; Lin et al. 2018; Zhang et al. 2018) domains. Motivated by the study of Dubois and Prade (1990), B. Liu (2002) introduced the concept of rough fuzzy variable and developed it further in B. Liu (2004). In reality, the parameters in some decision making problems may be uncertain and can be represented by rough fuzzy variable. For instance, the transportation cost $(c_t)$ depends upon fuel price, toll tax, vehicle maintenance cost, etc., each of which fluctuates from time to time. In general, the possible parameter values are provided by experts in terms of approximate intervals. As an example, let the transportation cost while traversing a route, in a particular month, as estimated by experts is "*between* \$ 70- \$ 80", which can be represented by a rough variable (B. Liu 2002). However, each point within the intervals of a rough variable may not be equipossible and can be "*around* \$ 74". It may be because $c_t$ may differ throughout the months in a year due to the fluctuation of fuel price, toll charges, etc. Therefore, such type of linguistic information can be approximated by rough fuzzy variables, and accordingly $c_t$ can be represented by a rough fuzzy transportation cost $\zeta_{c_t}$, where $\zeta_{c_t} = [\xi, \xi + 2][\xi - 2, \xi + 4], \xi = (72, 74, 76\,)$.

Considering the existing studies on QMST and some recent studies on rough fuzzy set, we observe that there are some lacunas in the literature which are mentioned below.

(i) Any optimization problem with associated parameters as rough fuzzy variables is yet to be studied in the literature.
(ii) A bi-objective QMST problem under rough fuzzy uncertain environments is not yet considered in the literature.

In order to address the above mentioned lacunas, in the present study, we have considered a bi-objective rough fuzzy quadratic minimum spanning tree problem (b-RFQMSTP), by considering the linear and quadratic weights as rough fuzzy variables. The bi-objective QMST problem is modelled using rough fuzzy chance-constrained programming (RFCCP) technique. It is solved by epsilon-constraint method (Hames et al. 1971), which is a classical multi-objective programming technique, and two multi-objective genetic algorithms (MOGAs): (i) nondominated sorting algorithm II (NSGA-II) (Deb et al. 2002) and (ii) multi-objective cross generational elitist selection, heterogeneous recombination, and cataclysmic mutation (MOCHC) (Nebro et al. 2007).

The rest of the chapter is organized as follows. Some related theorems of a multi-objective optimization problem with rough fuzzy parameters are presented in Section 6.2. The proposed chance-constrained model and its crisp equivalents are formulated in Section 6.3. A numerical example is illustrated in Section 6.4. The discussions of the results, of the proposed model, are provided in Section 6.5. In Section 6.6, we have presented a case study to numerically illustrate the proposed b-RFQMSTP. Finally, we conclude the study in Section 6.7.

## 6.2 Multi-objective Model with Rough Fuzzy Coefficients

A rough fuzzy multi-objective problem (RFMOP) deals with optimizing a vector of conflicting objectives subject to a set of constraints, where the associated coefficients of the objectives and constraints are rough fuzzy variables. An RFMOP model is defined below in (6.1).

$$\begin{cases} Minimize \;\; \left(f_1(\hat{\zeta}_1^T, \mathcal{X}), f_2(\hat{\zeta}_2^T, \mathcal{X}), \dots, f_m(\hat{\zeta}_m^T, \mathcal{X})\right) \\ subject\ to \\ g_i(\hat{\zeta}_i^T, \mathcal{X}) \le 0;\ i = 1,2, \dots, p, \end{cases} \tag{6.1}$$

where, $m$ is the number of objective functions, $\mathcal{X} = (x_1, x_2, \dots, x_n)^T$ is the decision vector. $\hat{\zeta}_l^T = \left(\hat{\zeta}_{l1}, \hat{\zeta}_{l2}, \dots, \hat{\zeta}_{ln}\right)^T l = 1,2, \dots, m$ and $\hat{\zeta}_i^T = \left(\hat{\zeta}_{i1}, \hat{\zeta}_{i2}, \dots, \hat{\zeta}_{in}\right)^T; i = 1,2, \dots, p$ are the rough fuzzy vectors.

Following Definition 1.3.26 (cf. Section 1.3.10), from (6.1), the primitive chance of $g_i(\hat{\zeta}_i^T, \mathcal{X}) \le 0$ can be stated as

$Ch\{g_i(\hat{\zeta}_i^T, \mathcal{X}) \le 0\}(\beta_i) \ge \alpha_i = Cr\{\theta \in \Theta | Tr\{g_i(\hat{\zeta}_i^T, \mathcal{X}) \le 0\} \ge \alpha_i\} \ge \beta_i, i = 1,2, \dots, p,$ and $Ch\{f_l(\hat{\zeta}_l^T, \mathcal{X}) \le \bar{\bar{f_l}}\}(\tau_l) \ge \rho_l = Cr\{\theta \in \Theta | Tr\{f_l(\hat{\zeta}_l^T, \mathcal{X}) \le \bar{\bar{f_l}}\} \ge \rho_l\} \ge \tau_l, l = 1,2, \dots, m.$

Furthermore, a decision vector $\mathcal{X}$ is said to be feasible to (6.1) if and only if the credibility measures of the fuzzy events, $\{\theta \in \Theta | Tr\{f_l(\hat{\zeta}_l^T, \mathcal{X}) \le \bar{\bar{f_l}}\} \ge \rho_l\}$ at confidence levels of at least $\tau_l$, are satisfied by a set of fuzzy event, $\{\theta \in \Theta | Tr\{g_i(\hat{\zeta}_i^T, \mathcal{X}) \le 0\} \ge \alpha_i\}$ having at least credibility measures $\beta_i$.

Hence, the chance-constrained multi-objective programming model can be formulated as presented in (6.2).

$$\begin{cases} Minimize \ \ (\bar{\bar{f}}_1, \bar{\bar{f}}_2, \dots, \bar{\bar{f}}_m) \\ subject\ to \\ Cr\{\theta \in \Theta | Tr\{f_l(\hat{\tilde{\zeta}}_l^T, \mathcal{X}) \le \bar{\bar{f}}_l\} \ge \rho_l\} \ge \tau_l;\ l = 1,2, \dots, m \\ Cr\{\theta \in \Theta | Tr\{g_i(\hat{\tilde{\zeta}}_i^T, \mathcal{X}) \le 0\} \ge \alpha_i\} \ge \beta_i; i = 1,2, \dots, p, \end{cases} \tag{6.2}$$

where $Cr\{\cdot\}$ and $Tr\{\cdot\}$ denote the credibility and trust measure of an event, respectively. $\rho_l, \tau_l, \alpha_i$ and $\beta_i$ are the predetermined confidence levels.

Moreover, if $\mathcal{X}_1$ and $\mathcal{X}_2$ are two solutions of the model (6.2) at the predetermined confidence levels $\rho_l$ and $\tau_l$, then $\mathcal{X}_1$ dominates $\mathcal{X}_2$, i.e., $\mathcal{X}_1 \prec \mathcal{X}_2$ at a trust level $\rho_l$ and credibility level $\tau_l$ if and only if following two conditions are true.

(i) $\{ Cr\{\theta \in \Theta | Tr\{f_l(\hat{\tilde{\zeta}}_l^T, \mathcal{X}_1) \le f_l^*\} \ge \rho_l\} \ge \tau_l\} \le \{ Cr\{\theta \in \Theta | Tr\{f_l(\hat{\tilde{\zeta}}_l^T, \mathcal{X}_2) \le \bar{\bar{f}}_l\} \ge \rho_l\} \ge \tau_l\}, \forall l = 1,2, \dots, m.$

(ii) There exists at least one $j$ such that $\{ Cr\{\theta \in \Theta | Tr\{f_j(\hat{\tilde{\zeta}}_j^T, \mathcal{X}_1) \le f_j^*\} \ge \rho_j\} \ge \tau_j\} < \{ Cr\{\theta \in \Theta | Tr\{f_j(\hat{\tilde{\zeta}}_j^T, \mathcal{X}_2) \le \bar{\bar{f}}_j\} \ge \rho_l\} \ge \tau_l\}$, where $j \in \{1,2, \dots, m\}$

Here, $f_l^*$ and $\bar{\bar{f}}_l$ are the value in the objective space corresponding to $\mathcal{X}_1$ and $\mathcal{X}_2$. Again, $\mathcal{X}_1$ and $\mathcal{X}_2$ are said to be nondominated solutions at $\rho_l$ and $\tau_l$ if neither $\mathcal{X}_1 \prec \mathcal{X}_2$ nor $\mathcal{X}_2 \prec \mathcal{X}_1$ at trust level $\rho_l$ and credibility level $\tau_l$.

**Remark 6.2.1**: A rough fuzzy vector $\hat{\tilde{\zeta}}_l$ is degenerated to a fuzzy vector, $\tilde{\zeta}_l$ if for $\rho_l > 0$, $Cr\{\theta \in \Theta | Tr\{f_l(\hat{\tilde{\zeta}}_l^T, \mathcal{X}) \le \bar{\bar{f}}_l\} \ge \rho_l\} \ge \tau_l$ is equivalent to a fuzzy event $Cr\{ \theta \in \Theta | f_l(\tilde{\zeta}_l^T, \mathcal{X})\} \ge \tau_l;\ l = 1,2, \dots, m$ which is a standard fuzzy chance constraint. Similarly, $Cr\{\theta \in \Theta | Tr\{g_i(\hat{\tilde{\zeta}}_i^T, \mathcal{X}) \le 0\} \ge \alpha_i\} \ge \beta_i$ also degenerates to $Cr\{ \theta \in \Theta | g_i(\tilde{\zeta}_i^T, \mathcal{X})\} \ge \beta_i$ at $\alpha_i > 0$ ; $i = 1,2, \dots, p$. Hence, the model (6.2) is transformed to an equivalent multi-objective fuzzy chance-constrained model as shown below.

$$\begin{cases} Minimize \ \ (\bar{\bar{f}}_1, \bar{\bar{f}}_2, \dots, \bar{\bar{f}}_m) \\ subject\ to \\ \quad Cr\{\theta \in \Theta | f_l(\tilde{\zeta}_l^T, \mathcal{X}) \le \bar{\bar{f}}_l\} \ge \tau_l;\ l = 1,2, \dots, m \\ \quad Cr\{ \theta \in \Theta | g_i(\tilde{\zeta}_l^T, \mathcal{X}) \le 0\} \ge \beta_i;\ i = 1,2, \dots, p. \end{cases} \tag{6.3}$$

**Remark 6.2.2**: A rough fuzzy vector $\hat{\tilde{\zeta}}_l$ is degenerated to a rough vector $\zeta_l$ if for $\tau_l > 0$, $Cr\{\theta \in \Theta | Tr\{f_l(\hat{\tilde{\zeta}}_l^T, \mathcal{X}) \le \bar{\bar{f}}_l\} \ge \rho_l\} \ge \tau_l$ become equivalent to a rough event $Tr\{f_l(\zeta_l^T, \mathcal{X}) \le \bar{\bar{f}}_l\} \ge \rho_l; l = 1,2, \dots, m$ which is a standard rough chance constraint. Similarly, $Cr\{\theta \in \Theta | Tr\{g_i(\hat{\tilde{\zeta}}_i^T, \mathcal{X}) \le 0\} \ge \alpha_i\} \ge \beta_i$ degenerates to $Tr\{g_i(\zeta_i^T, \mathcal{X}) \le 0\} \ge \alpha_i; i = 1,2, \dots, p$ at $\beta_i > 0$. Hence, the model (6.2) is transformed to an equivalent multi-objective rough chance-constrained model as presented in (6.4).

$$\begin{cases} Minimize \ \left(\bar{\bar{f}}_1, \bar{\bar{f}}_2, \dots, \bar{\bar{f}}_m\right) \\ subject\ to \\ \quad Tr\{f_l(\zeta_l^T, \mathcal{X}) \le \bar{\bar{f}}_l\} \ge \rho_l;\ l = 1,2,\dots,m \\ \quad Tr\{g_i(\zeta_i^T, \mathcal{X}) \le 0\} \ge \alpha_i; i = 1,2,\dots,p. \end{cases} \tag{6.4}$$

**Lemma 6.2.1**: For any $\theta \in \Theta$, a rough fuzzy vector $\hat{\zeta}_{ij}\chi$, degenerated to a fuzzy vector $\tilde{\zeta}_i(\theta)^T\chi$ which takes the form of a fuzzy variable $(u, v, w)$, where $\tilde{\zeta}_i(\theta) = [\tilde{\zeta}_{i1}(\theta), \tilde{\zeta}_{i2}(\theta), \dots, \tilde{\zeta}_{in}(\theta)]^T$ and $\chi = [x_1, x_2, \dots, x_n]^T$ are respectively a fuzzy vector and decision vector.

**Proof**: Without loss of generality, for any $\theta \in \Theta$, we have assumed that a fuzzy variable degenerated from a rough fuzzy variable becomes a triangular fuzzy variable. Let $\hat{\zeta}_{ij}$ be a rough fuzzy variable, then for any $\theta \in \Theta$ there exists a fuzzy variable $\tilde{\zeta}_{ij}(\theta) = \left(l_{ij}, r_{ij}, s_{ij}\right)$, $l_{ij} < r_{ij} < s_{ij}$ with credibility measures defined as

$$Cr\left(\tilde{\zeta}_{ij}(\theta) \le y\right) = \begin{cases} 1 & ; if\ s_{ij} < y \\ \frac{y + s_{ij} - 2r_{ij}}{2\left(s_{ij} - r_{ij}\right)} & ; if\ r_{ij} < y \le s_{ij} \\ \frac{y - l_{ij}}{2\left(r_{ij} - l_{ij}\right)} & ; if\ l_{ij} \le y \le r_{ij} \\ 0 & ; otherwise \end{cases}$$

and $$Cr\left(\tilde{\zeta}_{ij}(\theta) \ge y\right) = \begin{cases} 1 & ; if\ y < l_{ij} \\ \frac{2r_{ij} - l_{ij} - y}{2\left(r_{ij} - l_{ij}\right)} & ; if\ l_{ij} \le y < r_{ij} \\ \frac{s_{ij} - y}{2\left(s_{ij} - r_{ij}\right)} & ; if\ r_{ij} \le y \le s_{ij} \\ 0 & ; otherwise \end{cases}$$

Subsequently, it follows

$\tilde{\zeta}_i(\theta)^T\mathcal{X} =$
$\sum_{j=1}^{n} \tilde{\zeta}_{ij}(\theta)\mathcal{X} = \left(\sum_{j=1}^{n}\left(l_{ij}\mathcal{X}_j, r_{ij}\mathcal{X}_j, s_{ij}\mathcal{X}_j\right)\right) = \left(\sum_{j=1}^{n} l_{ij}\mathcal{X}_j, \sum_{j=1}^{n} r_{ij}\mathcal{X}_j, \sum_{j=1}^{n} s_{ij}\mathcal{X}_j\right)$
$= (u, v, w)$, where $u = \sum_{j=1}^{n} l_{ij}\mathcal{X}_j$, $v = \sum_{j=1}^{n} r_{ij}\mathcal{X}_j$ and $w = \sum_{j=1}^{n} s_{ij}\mathcal{X}_j$.

**Lemma 6.2.2**: For any $\theta \in \Theta$, a rough fuzzy vector $\hat{\zeta}_{ij}\mathcal{X}$, degenerated to a rough vector $\zeta_i(\theta)^T\mathcal{X}$ which takes the form of a rough variable $[m, n], [p, q]$, where $\zeta_i(\theta)^T = [\zeta_{i1}(\theta), \zeta_{i2}(\theta), \dots, \zeta_{in}(\theta)]^T$ and $\mathcal{X} = [x_1, x_2, \dots, x_n]^T$ are respectively, a rough vector and a decision vector.

**Proof**: For a given $\theta \in \Theta$, a rough fuzzy variable $\hat{\zeta}_{ij}$ when degenerated to a rough variable $\zeta_{ij}(\theta) = \left[u_{ij}, v_{ij}\right]\left[s_{ij}, t_{ij}\right]$, $s_{ij} \le u_{ij} \le v_{ij} \le t_{ij}$ with trust distributions

$$Tr\{\zeta_{ij}(\theta) \le z\} = \begin{cases} 0 & ; if\ z \le s_{ij} \\ \frac{z-s_{ij}}{2(t_{ij}-s_{ij})} & ; if\ s_{ij} \le z \le u_{ij} \\ \frac{1}{2}\left(\frac{z-u_{ij}}{v_{ij}-u_{ij}} + \frac{z-s_{ij}}{t_{ij}-s_{ij}}\right) & ; if\ u_{ij} \le z \le v_{ij} \\ \frac{1}{2}\left(\frac{z-s_{ij}}{t_{ij}-s_{ij}} + 1\right) & ; if\ v_{ij} \le z \le t_{ij} \\ 1 & ; if\ z \ge t_{ij} \end{cases}$$

and $$Tr\{\zeta_{ij}(\theta) \ge z\} = \begin{cases} 0 & ; if\ z \ge q \\ \frac{t_{ij}-z}{2(t_{ij}-s_{ij})} & ; if\ v_{ij} \le z \le t_{ij} \\ \frac{1}{2}\left(\frac{t_{ij}-z}{t_{ij}-s_{ij}} + \frac{v_{ij}-z}{v_{ij}-u_{ij}}\right) & ; if\ u_{ij} \le z \le v_{ij} \\ \frac{1}{2}\left(\frac{t_{ij}-z}{t_{ij}-s_{ij}} + 1\right) & ; if\ s_{ij} \le z \le u_{ij} \\ 1 & ; if\ z \le s_{ij} \end{cases}$$

Then, $\zeta_i(\theta)^T\mathcal{X} = \sum_{j=1}^{n} \zeta_{ij}(\theta)\mathcal{X} = \sum_{j=1}^{n}([u_{ij}\mathcal{X}_j, v_{ij}\mathcal{X}_j], [s_{ij}\mathcal{X}_j, t_{ij}\mathcal{X}_j])$ $= ([\sum_{j=1}^{n} u_{ij}\mathcal{X}_j, \sum_{j=1}^{n} v_{ij}\mathcal{X}_j], [\sum_{j=1}^{n} s_{ij}\mathcal{X}_j, \sum_{j=1}^{n} t_{ij}\mathcal{X}_j]) = [m,n],[p,q]$, where $p \le m \le n \le q$ such that $m = \sum_{j=1}^{n} u_{ij}\mathcal{X}_j$, $n = \sum_{j=1}^{n} v_{ij}\mathcal{X}_j$, $p = \sum_{j=1}^{n} s_{ij}\mathcal{X}_j$ and $q = \sum_{j=1}^{n} t_{ij}\mathcal{X}_j$.

**Theorem 6.2.1**: For given confidence levels $\alpha_i, \beta_i \in [0,1]$, when $p \le \bar{\bar{f}}_i \le m$, $Cr\{\theta \in \Theta | Tr\{\zeta_i(\theta)^T\mathcal{X} \le \bar{\bar{f}}_i\} \ge \alpha_i\} \ge \beta_i$ implies $Cr\{\theta \in \Theta | p + 2\alpha_i Q_1 \le \bar{\bar{f}}_i\} \ge \beta_i$, where $Q_1 = (q-p)$.

**Proof**: From the result of Lemma 6.2.2, $\zeta_i(\theta)^T\mathcal{X}$ is known to be a rough variable. Therefore, when $p \le \bar{\bar{f}}_i \le m$, the trust distribution of $\zeta_i(\theta)^T\mathcal{X}$ is given below.

$$Tr\{\zeta_i(\theta)^T\mathcal{X} \le \bar{\bar{f}}_i\} = \begin{cases} 0 & ; \bar{\bar{f}}_i \le p \\ 0.5\frac{\bar{\bar{f}}_i-p}{(q-p)} & ; p \le \bar{\bar{f}}_i \le m \\ 0.5\left(\frac{\bar{\bar{f}}_i-m}{n-m} + \frac{\bar{\bar{f}}_i-p}{q-p}\right) & ; m \le \bar{\bar{f}}_i \le n \\ 0.5\left(\frac{\bar{\bar{f}}_i-p}{q-p} + 1\right) & ; n \le \bar{\bar{f}}_i \le q \\ 1 & ; \bar{\bar{f}}_i \ge q \end{cases}$$

$$\Rightarrow \alpha_i \le \begin{cases} 0 & ; \bar{\bar{f}}_i \le p \\ 0.5\frac{\bar{\bar{f}}_i-p}{(q-p)} & ; p \le \bar{\bar{f}}_i \le m \\ 0.5\left(\frac{\bar{\bar{f}}_i-m}{n-m} + \frac{\bar{\bar{f}}_i-p}{q-p}\right) & ; m \le \bar{\bar{f}}_i \le n \\ 0.5\left(\frac{\bar{\bar{f}}_i-p}{q-p} + 1\right) & ; n \le \bar{\bar{f}}_i \le q \\ 1 & ; \bar{\bar{f}}_i \ge q, \end{cases}$$

where $\alpha_i$ is the predetermined confidence level of the trust distribution for the rough variable $\zeta_i(\theta)^T\mathcal{X}$. It immediately follows that

$\bar{\bar{f}}_i \geq p + 2\alpha_i(q-p)$ if $p \leq \bar{\bar{f}}_i \leq m$

$\bar{\bar{f}}_i \geq \frac{mq+pn+2\alpha_i(n-m)(q-p)-2pm}{n+q-(p+m)}$ if $m \leq \bar{\bar{f}}_i \leq n$

and

$\bar{\bar{f}}_i \geq 2p-q+2\alpha_i(q-p)$,if $n \leq \bar{\bar{f}}_i \leq q$.

Therefore, we get

$$Cr\{\theta \in \Theta | Tr\{\zeta_i(\theta)^T\mathcal{X} \leq \bar{\bar{f}}_i\} \geq \alpha_i\} \geq \beta_i$$

$$\Rightarrow Cr\left\{\theta \in \Theta | \bar{\bar{f}}_i \geq \begin{cases} p+2\alpha_i(q-p) & ; p \leq \bar{\bar{f}}_i \leq m \\ \frac{mq+pn+2\alpha_i(n-m)(q-p)-2pm}{n+q-(p+m)} & ; m \leq \bar{\bar{f}}_i \leq n \\ 2p-q+2\alpha_i(q-p) & ; n \leq \bar{\bar{f}}_i \leq q \\ q & ;\ q \leq \bar{\bar{f}}_i \end{cases}\right\} \geq \beta_i$$

$$\Rightarrow Cr\left\{\theta \in \Theta | \bar{\bar{f}}_i \geq \begin{cases} p+2\alpha_i Q_1 & ; p \leq \bar{\bar{f}}_i \leq m \\ \frac{mq+pn+2\alpha_i Q_2-2pm}{n+q-(p+m)} & ; m \leq \bar{\bar{f}}_i \leq n \\ 2p-q+2\alpha_i Q_1 & ; n \leq \bar{\bar{f}}_i \leq q \\ q & ; q \leq \bar{\bar{f}}_i, \end{cases}\right\} \geq \beta_i$$

where $Q_1 = (q-p)$ and $Q_2 = (n-m)(q-p)$.

Hence, when $p \leq \bar{\bar{f}}_i \leq m$, it follows from the above result that $Cr\{\theta \in \Theta | Tr\{\zeta_i(\theta)^T\mathcal{X} \leq \bar{\bar{f}}_i\} \geq \alpha_i\} \geq \beta_i \Rightarrow Cr\{\theta \in \Theta | p + 2\alpha_i Q_1 \leq \bar{\bar{f}}_i\} \geq \beta_i$, where $Q_1 = (q-p)$.

**Theorem 6.2.2**: For given confidence levels $\alpha_i, \beta_i \in [0,1]$, when $p \leq \bar{\bar{f}}_i \leq m$, $Cr\{\theta \in \Theta | Tr\{\zeta_i(\theta)^T\mathcal{X} \leq \bar{\bar{f}}_i\} \geq \alpha_i\} \geq \beta_i$ implies

$$\bar{\bar{f}}_i \geq \begin{cases} w + 2\alpha_i w_{Q_1} & ; if\ w \leq \bar{\bar{f}}_i - 2\alpha_i Q_1 \\ 2(v + \alpha_i v_{Q_1}) - (w + 2\alpha_i w_{Q_1}) + 2\beta_i\left((w + 2\alpha_i w_{Q_1}) - (v + 2\alpha_i v_{Q_1})\right) & ; if\ v \leq \bar{\bar{f}}_i - 2\alpha_i Q_1 \leq w \\ u + 2\alpha_i u_{Q_1} + 2\beta_i\left(\left((v + 2\alpha_i v_{Q_1}) - (u + 2\alpha_i u_{Q_1})\right)\right) & ; if\ u \leq \bar{\bar{f}}_i - 2\alpha_i Q_1 \leq v, \end{cases}$$

such that $Q_1 = (q-p)$ is a triangular fuzzy variable of the form $(u_{Q_1}, v_{Q_1}, w_{Q_1})$.

**Proof**: From the result of Theorem 6.2.1, $Cr\{\theta \in \Theta | Tr\{\zeta_i(\theta)^T\mathcal{X} \leq \bar{\bar{f}}_i\} \geq \alpha_i\} \geq \beta_i \Rightarrow Cr\{\theta \in \Theta | p + 2\alpha_i Q_1 \leq \bar{\bar{f}}_i\} \geq \beta_i$ when $p \leq \bar{\bar{f}}_i \leq m$ and $Q_1 = (q-p)$. Now $p$ and $Q_1$ are assumed to be triangular fuzzy variables of the form $(u, v, w)$ (cf. Lemma 6.2.1). Thus

$Cr\{\theta \in \Theta | p + 2\alpha_i Q_1 \leq \bar{\bar{f_i}}\} \geq \beta_i \quad \Rightarrow Cr\{\theta \in \Theta | p \leq \bar{\bar{f_i}} - 2\alpha_i Q_1\} \geq \beta_i$ when $p \leq \bar{\bar{f_i}} - 2\alpha_i Q_1 \leq m$

$$\Rightarrow \beta_i \leq \begin{cases} 1 & ; if\ w \leq \bar{\bar{f_i}} - 2\alpha_i Q_1 \\ \frac{\bar{\bar{f}}_i - 2\alpha_i Q_1 + w - 2v}{2(w-v)} & ; if\ v \leq \bar{\bar{f_i}} - 2\alpha_i Q_1 \leq w \\ \frac{\bar{\bar{f}}_i - 2\alpha_i Q_1 - u}{2(v-u)} & ; if\ u \leq \bar{\bar{f_i}} - 2\alpha_i Q_1 \leq v \\ 0 & ; otherwise \end{cases}$$

$$\Rightarrow \bar{\bar{f_i}} \geq \begin{cases} w + 2\alpha_i Q_1 & ; if\ w \leq \bar{\bar{f_i}} - 2\alpha_i Q_1 \\ 2v - w + 2\alpha_i Q_1 + 2\beta_i (w - v) & ; if\ v \leq \bar{\bar{f_i}} - 2\alpha_i Q_1 \leq w \\ u + 2\alpha_i Q_1 + 2\beta_i (v - u) & ; if\ u \leq \bar{\bar{f_i}} - 2\alpha_i Q_1 \leq v \end{cases}$$

$\Rightarrow$

$$\bar{\bar{f_i}} \geq \begin{cases} w + 2\alpha_i w_{Q_1} & ; if\ w \leq \bar{\bar{f_i}} - Q_1 \\ 2(v + \alpha_i v_{Q_1}) - (w + 2\alpha_i w_{Q_1}) + 2\beta_i \left((w + 2\alpha_i w_{Q_1}) - (v + 2\alpha_i v_{Q_1})\right) & ; if\ v \leq \bar{\bar{f_i}} - 2\alpha_i Q_1 \leq w \\ u + 2\alpha_i u_{Q_1} + 2\beta_i \left(\left(v + 2\alpha_i v_{Q_1}\right) - \left(u + 2\alpha_i u_{Q_1}\right)\right)\Big) & ; if\ u \leq \bar{\bar{f_i}} - 2\alpha_i Q_1 \leq v, \end{cases}$$

where $Q_1$ is a triangular fuzzy variable expressed as $(u_{Q_1}, v_{Q_1}, w_{Q_1})$.

**Theorem 6.2.3**: For any given confidence level $\alpha_i \in [0,1]$

(i) If $0 \leq \beta_i \leq 0.5$, then $Cr\{\theta \in \Theta | p \leq \bar{\bar{f_i}} - 2\alpha_i Q_1\} \geq \beta_i$ if and only if,
$u + 2\alpha_i u_{Q_1} + 2\beta_i \left\{\left(v + 2\alpha_i v_{Q_1}\right) - \left(u + 2\alpha_i u_{Q_1}\right)\right)\right\} \leq \bar{\bar{f}}_i$ for $u \leq \bar{\bar{f_i}} - 2\alpha_i Q_1 \leq v$

(ii) If $0.5 \leq \beta_k \leq 1$, then $Cr\{\theta \in \Theta | p \leq \bar{\bar{f_i}} - 2\alpha_i Q_1\} \geq \beta_i$ if and only if,
$2\left(v + \alpha_i v_{Q_1}\right) - \left(w + 2\alpha_i w_{Q_1}\right) + 2\beta_i \left(\left(w + 2\alpha_i w_{Q_1}\right) - \left(v + 2\alpha_i v_{Q_1}\right)\right) \leq \bar{\bar{f_i}}$
for $v \leq \bar{\bar{f_i}} - 2\alpha_i Q_1 \leq w$, where $Q_1 = (q - p)$is of the form $(u_{Q_1}, v_{Q_1}, w_{Q_1})$.

**Proof**: It follows directly from Remark 1.3.3 corresponding to Theorem 1.3.2 (cf. Section 1.3.2) and Theorem 6.2.2.

## 6.3 Quadratic Minimum Spanning Tree Problem with Rough Fuzzy Coefficients

In this section, a bi-objective rough fuzzy quadratic minimum spanning tree problem (b-RFQMSTP) is modelled for a weighted connected network (WCN) $G = (V_G, E_G)$, where $V_G$ and $E_G$ are the vertex set and edge set of $G$, respectively. Here, two types of weights are associated with $G$: (i) linear weight and (ii) quadratic weight. The linear weight $\xi_{e_i}$ of an edge $e_i (\in E_G)$ is the cost associated with $e_i$ which connects vertices $v_m$ and $v_n$. Whereas, the quadratic weight $\zeta_{e_i e_j} (i \neq j)$ implies the interaction (interfacing) cost between any pair of edges $e_i$ and $e_j$.

In this study, we model a bi-objective quadratic minimum spanning tree by considering the linear and quadratic weights of a network as two separate minimization objectives. Each of these linear and quadratic weights is represented by rough fuzzy variable. Here, we have used the chance-constrained programming technique to model the proposed b-RFQMSTP as shown in (6.5) and its equivalent crisp transformations are presented in the models (6.8) and (6.9).

$$\begin{cases} Minimize\ \left(\bar{\bar{f_1}}, \bar{\bar{f_2}}\right) \\ subject\ to \\ Cr\left\{\theta \in \Theta \middle| Tr\left\{f_1(\xi_s(\theta), x_s) \leq \bar{\bar{f_1}}\right\} \geq \alpha_1\right\} \geq \beta_1 \\ Cr\left\{\theta \in \Theta \middle| Tr\left\{f_2(\zeta_{st}(\theta), x_s, x_t) \leq \bar{\bar{f_2}}\right\} \geq \alpha_2\right\} \geq \beta_2 \\ \sum_{s=1}^{|E_G|} x_s = |V_G| - 1 \\ \sum_{s \in E_\kappa} x_s \leq |\kappa| - 1,\ \kappa \subset V_G, |\kappa| \geq 3 \\ \forall\ x_s, x_t \in \{0,1\};\ x_s \neq x_t \\ \alpha_1, \alpha_2, \beta_1, \beta_2 \in [0,1] \end{cases} \tag{6.5}$$

Here, $\alpha_1, \alpha_2, \beta_1$ and $\beta_2$ are predetermined confidence levels, and $|V_G|$ and $|E_G|$ are respectively, the order and size of $G$. Subsequently, model (6.5) is expressed as

$$\begin{cases} Minimize\ \left(\bar{\bar{f_1}}, \bar{\bar{f_2}}\right) \\ subject\ to \\ Cr\left\{\theta \in \Theta \middle| Tr\left\{\sum_{s=1}^{|E_G|} \xi_s(\theta) x_s \leq \bar{\bar{f_1}}\right\} \geq \alpha_1\right\} \geq \beta_1 \\ Cr\left\{\theta \in \Theta \middle| Tr\left\{\sum_{s=1}^{|E_G|} \sum_{t=1}^{|E_G|} \zeta_{st}(\theta) x_s x_t \leq \bar{\bar{f_2}}\right\} \geq \alpha_2\right\} \geq \beta_2 \\ \sum_{s=1}^{|E_G|} x_s = |V_G| - 1 \\ \sum_{s \in E_\kappa} x_s \leq |\kappa| - 1,\ \kappa \subset V_G, |\kappa| \geq 3 \\ \forall\ x_s, x_t \in \{0,1\};\ x_s \neq x_t \\ \alpha_1, \alpha_2, \beta_1, \beta_2 \in [0,1] \end{cases} \tag{6.6}$$

Let $L(\theta) = \sum_{s=1}^{|E_G|} \xi_s(\theta) x_s$ and $Q(\theta) = \sum_{s=1}^{|E_G|} \sum_{t=1}^{|E_G|} \zeta_{st}(\theta) x_s x_t$ then according to Lemma 6.2.2, $L(\theta)$ and $Q(\theta)$ are rough variables which are respectively, represented by $[m_1, n_1], [p_1, q_1]$ and $[m_2, n_2], [p_2, q_2]$. Here, according to Theorem 6.2.1, model (6.7) can be extended from model (6.6) by considering, $p_1 \leq L(\theta) \leq m_1$ and $p_2 \leq Q(\theta) \leq m_2$.

Again, $p_1 + 2\alpha_1(q_1 - p_1)$ and $p_2 + 2\alpha_2(q_2 - p_2)$ become the triangular fuzzy variables such that $p_1 = (u_1, v_1, w_1)$, $(q_1 - p_1) = (u_L, v_L, w_L)$ and $p_2 = (u_2, v_2, w_2)$ and $(q_2 - p_2) = (u_Q, v_Q, w_Q)$. Then by Theorem 6.2.2 and Theorem 6.2.3, the crisp equivalents of model (6.7) are presented in models (6.8) and (6.9) when $\beta_1$, $\beta_2$ belong to $(0, 0.5]$ and $(0.5,1]$, respectively.

$$\begin{cases} Minimize \ \left(\bar{\bar{f}}_1, \bar{\bar{f}}_2\right) \\ subject\ to \\ \quad Cr\{\theta \in \Theta | p_1 + 2\alpha_1(q_1 - p_1) \leq \bar{\bar{f}}_1\} \geq \beta_1 \\ \quad Cr\{\theta \in \Theta | p_2 + 2\alpha_2(q_2 - p_2) \leq \bar{\bar{f}}_2\} \geq \beta_2 \\ \quad \sum_{s=1}^{|E_G|} x_s = |V_G| - 1 \\ \quad \sum_{s \in E_\kappa} x_s \leq |\kappa| - 1, \ \kappa \subset V_G, |\kappa| \geq 3 \\ \quad \forall \ x_s, x_t \in \{0,1\};\ x_s \neq x_t \\ \quad \alpha_1, \alpha_2, \beta_1, \beta_2 \in [0,1] \end{cases} \tag{6.7}$$

$$\begin{cases} \bar{\bar{f}}_1 = Minimize\ u_1 + 2\alpha_1 u_L + 2\beta_1\left(\left(v_1 + 2\alpha_1 v_L\right) - \left(u_1 + 2\alpha_1 u_L\right)\right) \\ \bar{\bar{f}}_2 = Minimize\ u_2 + 2\alpha_2 u_Q + 2\beta_2\left(\left(v_2 + 2\alpha_2 v_Q\right) - \left(u_2 + 2\alpha_2 u_Q\right)\right) \\ subject\ to \\ \quad \sum_{s=1}^{|E_G|} x_s = |V_G| - 1 \\ \quad \sum_{s \in E_\kappa} x_s \leq |\kappa| - 1, \ \kappa \subset V_G, |\kappa| \geq 3 \\ \quad \forall \ x_s, x_t \in \{0,1\};\ x_s \neq x_t \\ \quad \alpha_1, \alpha_2 \in [0,1];\ \beta_1, \beta_2 \in (0,0.5] \end{cases} \tag{6.8}$$

and

$$\begin{cases} \bar{\bar{f}}_1 = Minimize\ 2(v_1 + \alpha_1 v_L) - (w_1 + 2\alpha_2 w_L) + 2\beta_1\left((w_1 + 2\alpha_1 w_L) - (v_1 + 2\alpha_1 v_L)\right) \\ \bar{\bar{f}}_2 = Minimize\ 2\left(v_2 + \alpha_2 v_Q\right) - \left(w_2 + 2\alpha_2 w_Q\right) + 2\beta_2\left(\left(w_2 + 2\alpha_2 w_Q\right) - \left(v_2 + 2\alpha_2 v_Q\right)\right) \\ subject\ to \\ \sum_{s=1}^{|E_G|} x_s = |V_G| - 1 \\ \sum_{s \in E_\kappa} x_s \leq |\kappa| - 1, \kappa \subset V_G, |\kappa| \geq 3 \\ \quad \forall \ x_s, x_t \in \{0,1\};\ x_s \neq x_t \\ \quad \alpha_1, \alpha_2 \in [0,1];\ \beta_1, \beta_2 \in (0.5,1]. \end{cases} \tag{6.9}$$

## 6.4 Numerical Illustration

We consider a WCUN $G = (V_G, E_G)$, as shown in Fig. 6.1, where $V_G$ is the vertex set and $E_G$ is the edge set of $G$. The corresponding linear and quadratic weights of $G$ are represented by rough fuzzy variable, and presented in Table 6.1 and Table 6.2, respectively.

Models (6.8) and (6.9), which are essentially the crisp equivalents of the model (6.5) are solved for $G$ with the epsilon-constraint method (cf. Section 1.3.15.2), using a standard optimization solver, LINGO 11.0. For the sake of convenience, the confidence levels of each of these crisp models, are considered as the same such that, $\alpha_1$ and $\alpha_2$ are represented by $\alpha$, and $\beta_1$and $\beta_2$ are represented by $\beta$, i.e., $\alpha_1 = \alpha_2 = \alpha$ and $\beta_1 = \beta_2 = \beta$. Table 6.3 shows the solutions in term of compromised decision vector $x$ and its corresponding objective values. The confidence levels $\alpha$ and $\beta$ for both the objectives of the models (6.8) and (6.9) are set as $\alpha = 0.9,\ \beta = 0.4$ and $\alpha = 0.9,\ \beta = 0.8$, respectively. Fig. 6.2 and Fig. 6.3 depict the QMSTs of $G$ corresponding to the optimal decision vectors shown in Table 6.3.

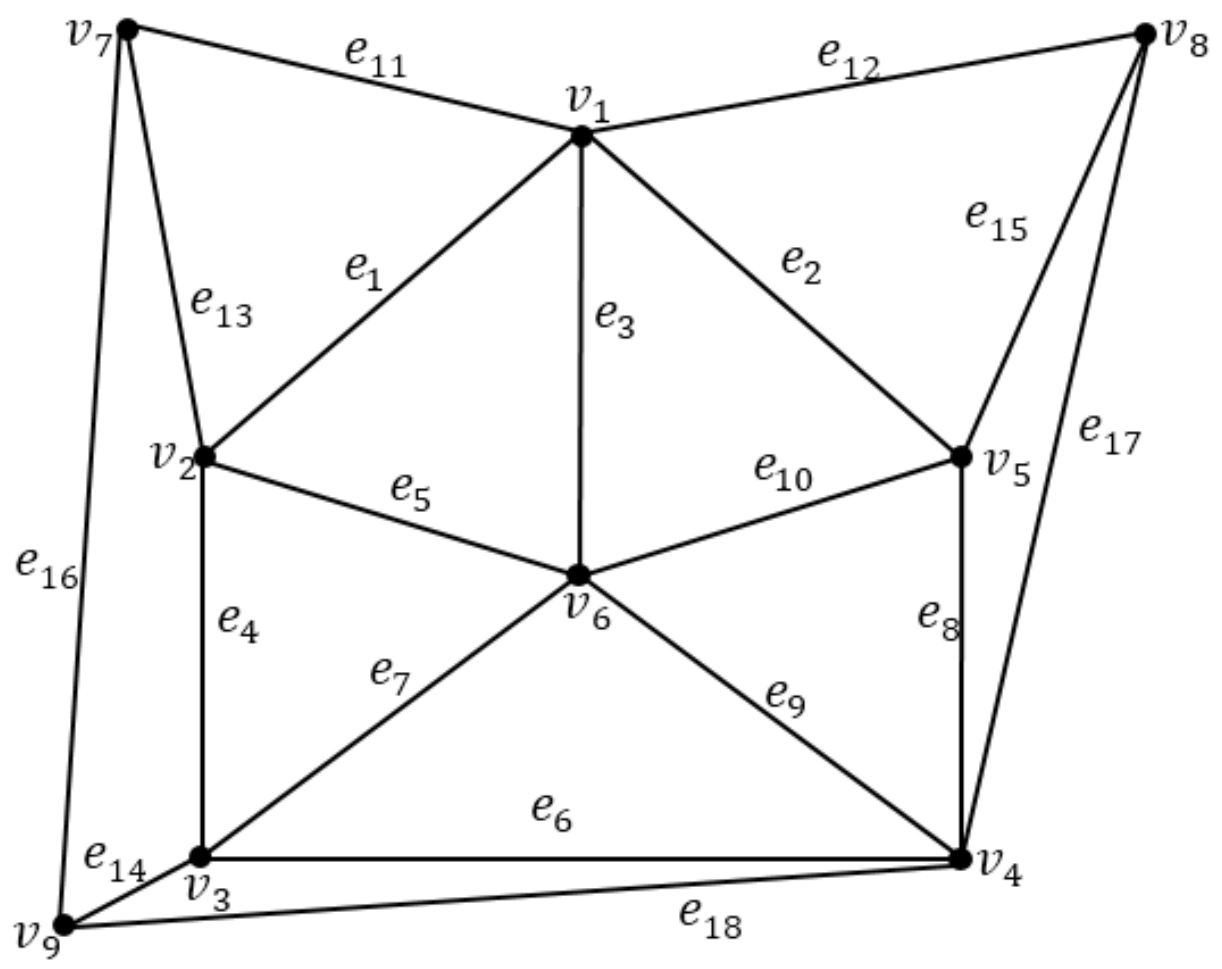

Figure 6.1 A weighted connected undirected network $G$

Table 6.1 Rough fuzzy linear weights of the WCUN $G$

| Edges | Rough fuzzy linear weights | Edges | Rough fuzzy linear weights |
|---|---|---|---|
| $e_1$ | $[\xi, \xi + 2][\xi - 1, \xi + 3]$ $\xi = (9, 11.5, 12.7)$ | $e_{10}$ | $[\xi, \xi + 0.5][\xi - 0.7, \xi + 0.9]$ $\xi = (9.4, 11.6, 12.8)$ |
| $e_2$ | $[\xi, \xi + 1][\xi - 1, \xi + 2]$ $\xi = (11, 13, 15)$ | $e_{11}$ | $[\xi, \xi + 0.9][\xi - 1.2, \xi + 1.8]$ $\xi = (11.5, 12.9, 13.8)$ |
| $e_3$ | $[\xi, \xi + 1.5][\xi - 0.7, \xi + 1.9]$ $\xi = (9.2, 10.5, 12.4)$ | $e_{12}$ | $[\xi, \xi + 0.5][\xi - 0.6, \xi + 0.95]$ $\xi = (8.9, 10.2, 12)$ |
| $e_4$ | $[\xi, \xi + 2][\xi - 2, \xi + 3]$ $\xi = (10.6, 14.1, 17.2)$ | $e_{13}$ | $[\xi, \xi + 1][\xi - 1.1, \xi + 1.5]$ $\xi = (10, 12, 13.5)$ |
| $e_5$ | $[\xi, \xi + 1][\xi - 2, \xi + 2]$ $\xi = (8.3, 10, 12.4)$ | $e_{14}$ | $[\xi, \xi + 0.4][\xi - 0.5, \xi + 0.8]$ $\xi = (9.8, 11.2, 12.4)$ |
| $e_5$ | $[\xi, \xi + 2][\xi - 2, \xi + 2.5]$ $\xi = (9.8, 11.7, 14)$ | $e_{15}$ | $[\xi, \xi + 1.2][\xi - 1.2, \xi + 1.4]$ $\xi = (11.9, 12.8, 14.5)$ |
| $e_7$ | $[\xi, \xi + 1.5][\xi - 0.5, \xi + 2]$ $\xi = (11, 14, 15)$ | $e_{16}$ | $[\xi, \xi + 1.1][\xi - 0.9, \xi + 1.3]$ $\xi = (11.5, 12, 13.5)$ |
| $e_8$ | $[\xi, \xi + 1][\xi - 0.5, \xi + 1.8]$ $\xi = (12, 14.7, 17.1)$ | $e_{17}$ | $[\xi, \xi + 0.6][\xi - 0.4, \xi + 0.67]$ $\xi = (9, 10.2, 12.9)$ |
| $e_9$ | $[\xi, \xi + 2.5][\xi - 1.5, \xi + 2.7]$ $\xi = (10, 12, 14)$ | $e_{18}$ | $[\xi, \xi + 0.78][\xi - 0.8, \xi + 0.8]$ $\xi = (10.2, 11.6, 12.4)$ |

For both the models, defined in (6.8) and (6.9), two different combinations of values for the confidence levels $\alpha$ and $\beta$ generate two distinct compromised binary decision vectors. A decision variable in a decision vector $x$ corresponds to an edge in $G$. If a decision variable is set to 1, the corresponding edge is considered in QMST of $G$; otherwise, the edge is not included in the QMST. Fig. 6.2 shows the QMST corresponding to the compromised decision vector reported in the first row of Table 6.3. Similarly, Fig. 6.3 displays the QMST corresponding to the compromised decision vector presented in the second row of Table 6.3.

Table 6.2 Rough fuzzy quadratic weights of the WCUN $G$

| Edges | Rough fuzzy quadratic weights | Edges | Rough fuzzy quadratic weights |
|---|---|---|---|
| $e_1e_2$ | $[\zeta,\zeta+0.7][\xi-0.2,\zeta+1]$ $\zeta=(9,11,13)$ | $e_7e_{10}$ | $[\zeta,\zeta+0.9][\zeta-0.7,\zeta+1.2]$ $\zeta=(10,12.5,14)$ |
| $e_1e_3$ | $[\zeta,\zeta+1.2][\zeta-0.5,\zeta+1.4]$ $\zeta=(8.5,10.2,11.5)$ | $e_1e_{10}$ | $[\zeta,\zeta+0.7][\zeta-0.5,\zeta+1.0]$ $\zeta=(9,11,13)$ |
| $e_2e_3$ | $[\zeta,\zeta+0.4][\zeta-0.2,\zeta+0.8]$ $\zeta=(9.5,10.7,11.6)$ | $e_2e_{15}$ | $[\zeta,\zeta+0.5][\zeta-0.5,\zeta+0.9]$ $\zeta=(10.9,11.5,12.1)$ |
| $e_2e_{10}$ | $[\zeta,\zeta+1][\zeta-0.8,\zeta+1.2]$ $\zeta=(9.8,10.8,11.5)$ | $e_9e_{17}$ | $[\zeta,\zeta+0.9][\zeta-0.4,\zeta+1.2]$ $\zeta=(10.8,12.2,13.5)$ |
| $e_5e_7$ | $[\zeta,\zeta+1.2][\zeta-0.3,\zeta+1.5]$ $\xi=(10.2,12.2,13.4)$ | $e_2e_{18}$ | $[\zeta,\zeta+0.7][\zeta-0.3,\zeta+0.8]$ $\zeta=(11.2,12,13.9)$ |
| $e_4e_5$ | $[\zeta,\zeta+0.9][\zeta-0.5,\zeta+1.6]$ $\zeta=(10.4,11.7,12.9)$ | $e_{11}e_{17}$ | $[\zeta,\zeta+1.2][\zeta-1.1,\zeta+1.1]$ $\zeta=(10,11.5,12.5)$ |
| $e_1e_5$ | $[\zeta,\zeta+1.2][\zeta-0.2,\zeta+1.5]$ $\zeta=(8.6,9.2,9.8)$ | $e_{11}e_{12}$ | $[\zeta,\zeta+0.7][\zeta-0.6,\zeta+0.8]$ $\zeta=(8,10,12)$ |
| $e_5e_9$ | $[\zeta,\zeta+0.9][\zeta-0.3,\zeta+1.2]$ $\zeta=(8.1,9.8,10.9)$ | $e_{12}e_{16}$ | $[\zeta,\zeta+1.5][\zeta-1.3,\zeta+1.7]$ $\zeta=(8.9,10,11.2)$ |
| $e_4e_6$ | $[\zeta,\zeta+2][\zeta-0.9,\zeta+2.6]$ $\zeta=(10.8,12.9,14.2)$ | $e_{13}e_{14}$ | $[\zeta,\zeta+1.2][\zeta-1.2,\zeta+1.4]$ $\zeta=(11.2,12.3,13.4)$ |
| $e_4e_9$ | $[\zeta,\zeta+2.1][\zeta-0.2,\zeta+2.4]$ $\zeta=(11,13,14)$ | $e_{13}e_{15}$ | $[\zeta,\zeta+1.6][\zeta-1.5,\zeta+1.7]$ $\zeta=(10,12.2,14)$ |
| $e_4e_{10}$ | $[\zeta,\zeta+1][\zeta-1.2,\zeta+1.5]$ $\zeta=(10.7,11.7,12.6)$ | $e_{14}e_{15}$ | $[\zeta,\zeta+1][\zeta-1,\zeta+1.2]$ $\zeta=(9,11,13)$ |
| $e_6e_{10}$ | $[\zeta,\zeta+1][\zeta-1,\zeta+2]$ $\zeta=(8,10,12)$ | $e_{10}e_{13}$ | $[\zeta,\zeta+0.2][\zeta-0.8,\zeta+1]$ $\zeta=(12.1,12.6,12.9)$ |
| $e_3e_6$ | $[\zeta,\zeta+0.9][\zeta-1,\zeta+1.9]$ $\zeta=(7.8,11,12)$ | $e_9e_{13}$ | $[\zeta,\zeta+2][\zeta-1,\zeta+3]$ $\zeta=(10,11,12)$ |
| $e_6e_9$ | $[\zeta,\zeta+1][\zeta-1,\zeta+2]$ $\zeta=(9,11,13)$ | $e_{14}e_{15}$ | $[\zeta,\zeta+1][\zeta-1,\zeta+2]$ $\zeta=(10.5,11.5,12.9)$ |
| $e_{16}e_{17}$ | $[\zeta,\zeta+1.5][\zeta-1.2,\zeta+1.9]$ $\zeta=(9.8,11.2,12.4)$ | $e_5e_{14}$ | $[\zeta,\zeta+1.2][\zeta-2,\zeta+2.4]$ $\zeta=(11,12.6,13.5)$ |
| $e_7e_8$ | $[\zeta,\zeta+2][\zeta-1,\zeta+3]$ $\zeta=(12,14,15)$ | $e_9e_{15}$ | $[\zeta,\zeta+1.8][\zeta-1.3,\zeta+2.9]$ $\zeta=(12,13.6,14.5)$ |
| $e_8e_9$ | $[\zeta,\zeta+0.7][\zeta-0.5,\zeta+0.9]$ $\zeta=(11.6,12.7,13.8)$ | $e_{10}e_{18}$ | $[\zeta,\zeta+1][\zeta-1.4,\zeta+1.9]$ $\zeta=(8.9,10.9,12.5)$ |
| $e_3e_{10}$ | $[\zeta,\zeta+1.4][\zeta-0.1,\zeta+1.5]$ $\zeta=(9.5,10.5,11.8)$ | $e_8e_{10}$ | $[\zeta,\zeta+0.5][\zeta-0.4,\zeta+0.6]$ $\zeta=(10.8,11.8,12.9)$ |

Table 6.3 Solutions of models (6.8) and (6.9) of the WCUN $G$

| Confidence Levels $(\alpha,\beta)$ | Objective values $\langle\bar{\bar{f}}_1,\bar{\bar{f}}_2\rangle$ | Optimal decision vector $x$ |
|---|---|---|
| $\alpha=0.9,\beta=0.4$ | $\langle 128.5600, 11.9400\rangle$ | $(1\ 0\ 0\ 0\ 1\ 0\ 0\ 0\ 1\ 1\ 1\ 0\ 1\ 1\ 0\ 1\ 0\ 0)^T$ |
| $\alpha=0.9,\beta=0.8$ | $\langle 129.5760, 16.4400\rangle$ | $(0\ 0\ 1\ 0\ 0\ 0\ 1\ 0\ 0\ 1\ 0\ 1\ 1\ 0\ 1\ 0\ 1\ 1)^T$ |

The models (6.8) and (6.9) are solved for $G$ by setting the pair of confidence levels of $\alpha$ and $\beta$, respectively to $(0.9,0.4)$ and $(0.9,0.8)$. We have used jMetal 4.5 (Durillo and Nebro 2011) framework, to solve each of these models by executing NSGA-II and MOCHC for 500 generations. The parameter setting of each of these algorithms is mentioned below.

- For NSGA-II: Population size = 100, maximum generations = 500, crossover probability = 0.9, mutation probability = 0.03.
- For MOCHC: Population size = 100, maximum generations = 500, crossover probability = 0.9, incest threshold count = 0.25, convergence value = 1, preserved population = 0.05.

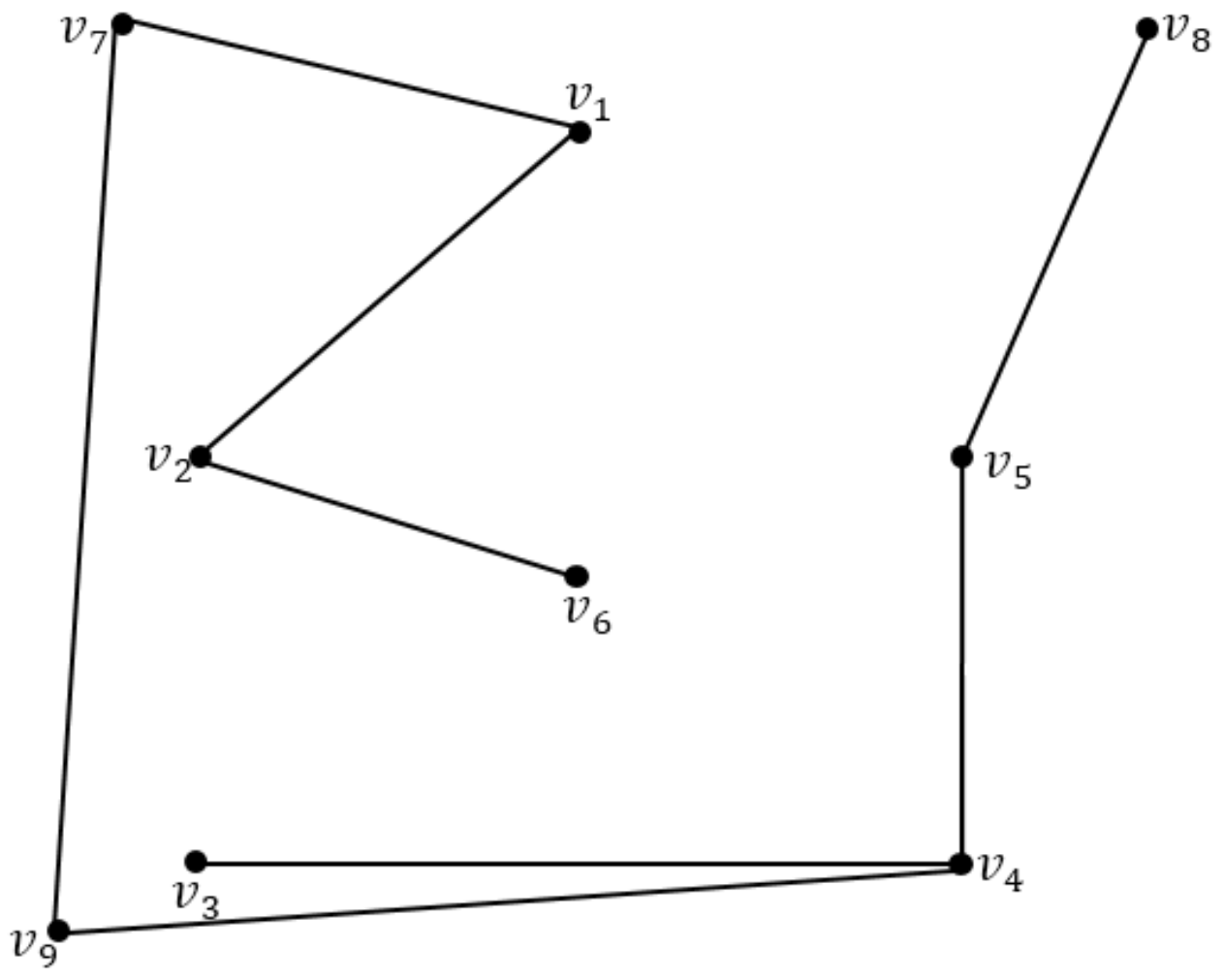

Figure 6.2 A rough fuzzy quadratic minimum spanning tree of the WCUN $G$ with $\alpha = 0.9$ and $\beta = 0.4$

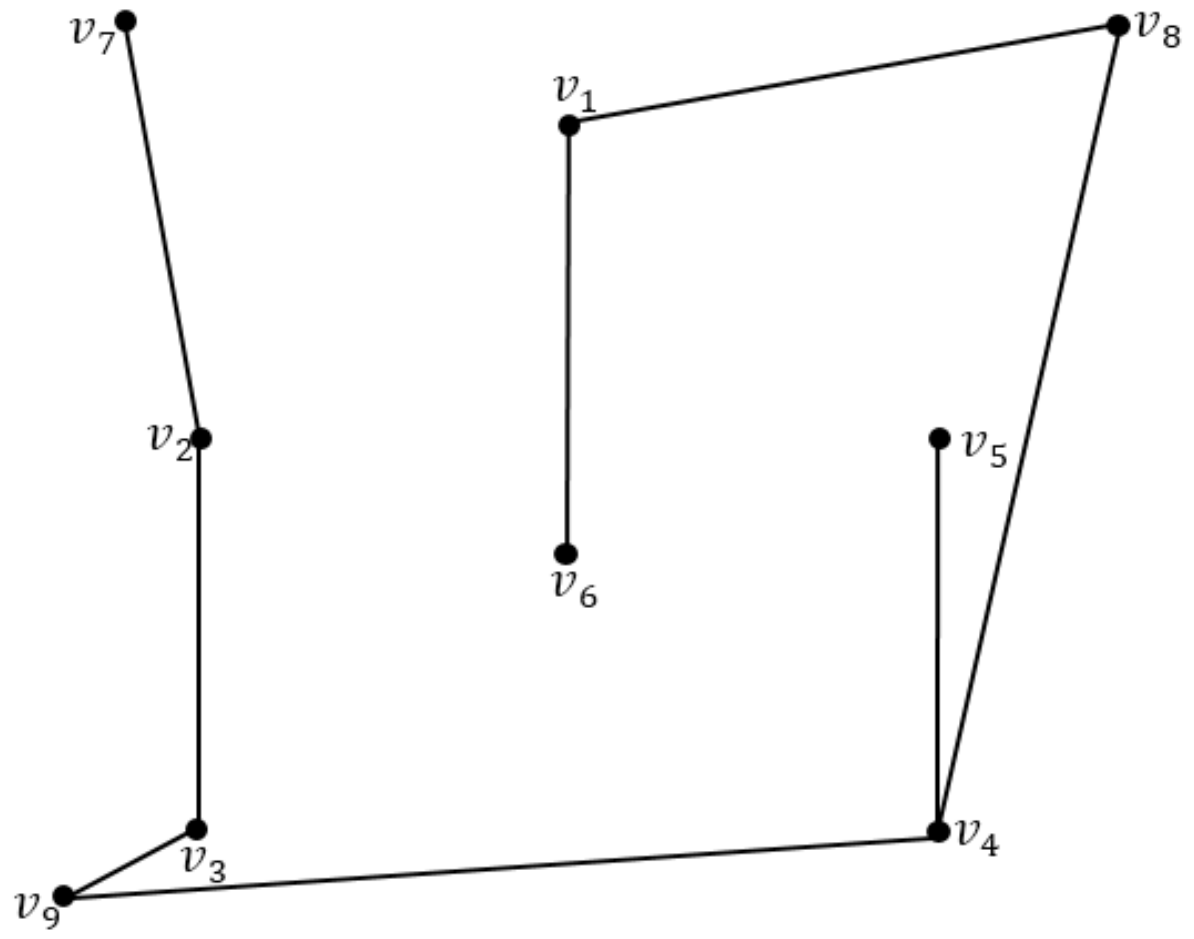

Figure 6.3 A rough fuzzy quadratic minimum spanning tree of the WCUN $G$ with $\alpha = 0.9$ and $\beta = 0.8$

The nondominated solutions of the model (6.8) are displayed in Fig. 6.4 (a) and Fig. 6.4 (b), and that of the model (6.9) are shown in Fig. 6.5 (a) and Fig. 6.5 (b). Specifically, the nondominated solutions generated by NSGA-II are displayed in Fig. 6.4 (a) and Fig. 6.5 (a). Whereas, Fig. 6.4 (b) and Fig. 6.5 (b) depict the nondominated solutions generated by MOCHC. The solutions generated by the epsilon-constraint method at different confidence levels are also shown in each of these figures, which are also nondominated as compared to the solutions of NSGA-II and MOCHC.

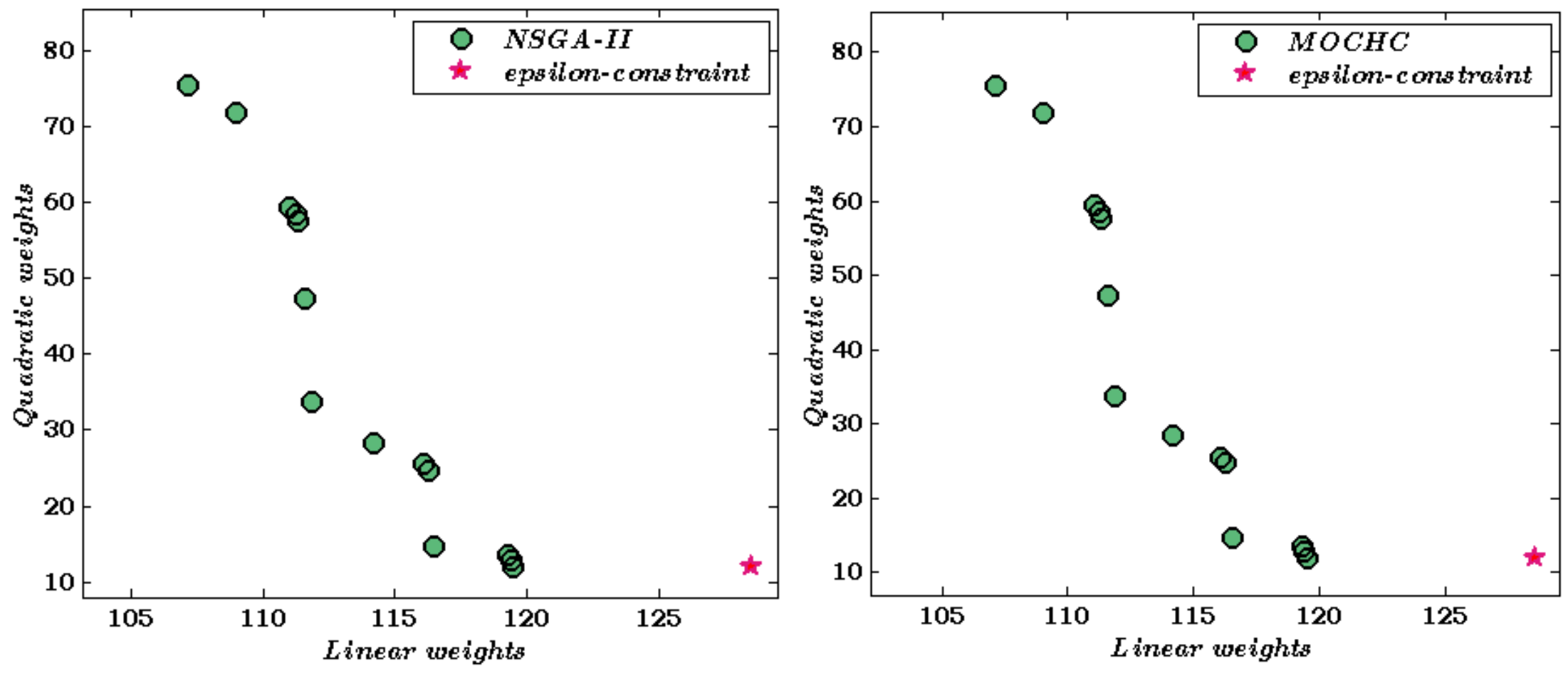

(a) (b)

Figure 6.4 Nondominated solutions of the model (6.8) for the WCUN $G$ at predetermined confidence levels, $\alpha = 0.9$ and $\beta = 0.4$ when solved by the epsilon-constraint method, and (a) NSGA-II and (b) MOCHC

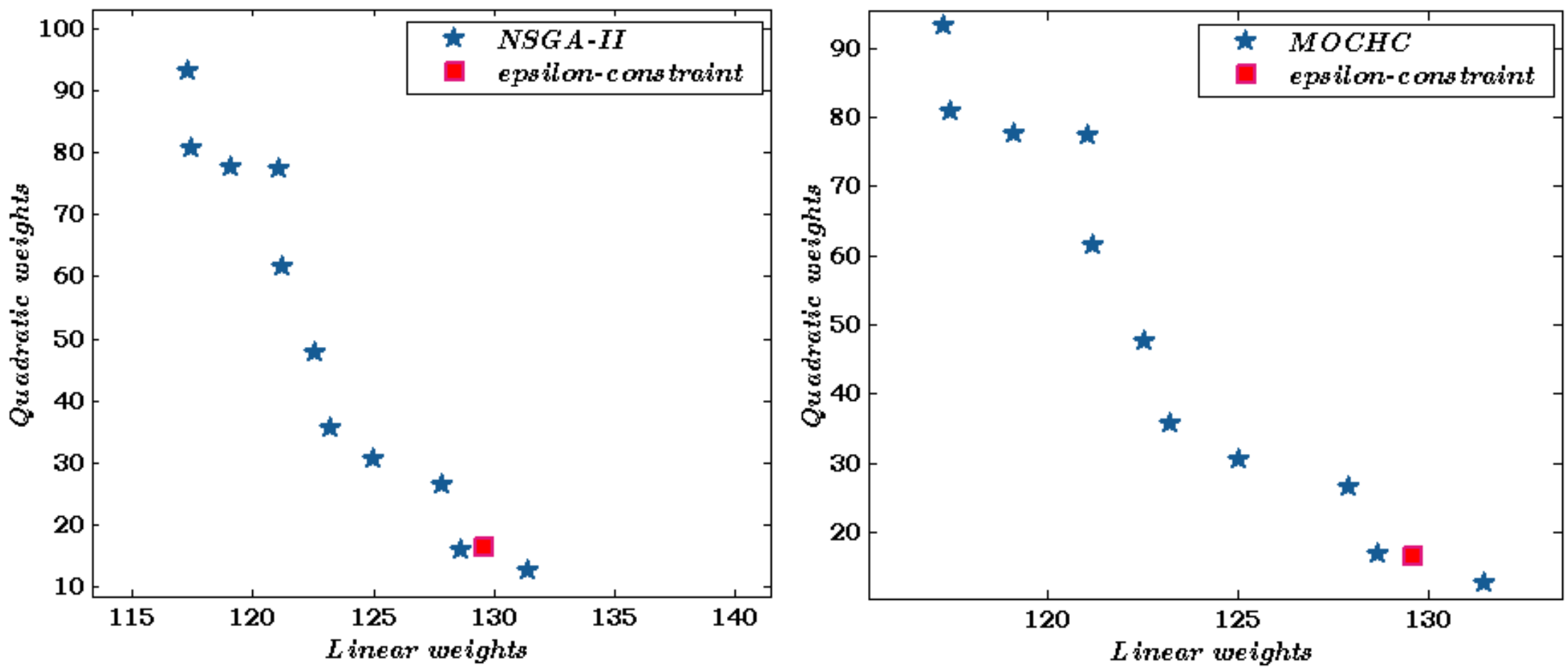

(a) (b)

Figure 6.5 Nondominated solutions of the model (6.9) for the WCUN $G$ at predetermined confidence levels, $\alpha = 0.9$ and $\beta = 0.8$ when solved by the epsilon-constraint method, and (a) NSGA-II and (b) MOCHC

## 6.5 Results and Discussion

In this subsection, we have simulated the models (6.8) and (6.9) for five large instances, generated randomly. Considering the wheel graph instances of QMST which are NP-hard in nature (Ćustić et al. 2018), in this study, we first randomly generate a wheel graph instance, and then include two additional edges between any two pair of vertices such that there exists no self-loop and parallel edges in the resulting instance. Each of these generated instances is named as $QMST_n_m$, where $n$ and $m$ are respectively, the numbers of vertices and edges of a connected network. For each instance, a linear weight is associated with each edge. Moreover, and for any pair of edges, there exists a quadratic weight. The linear and quadratic weights are generated randomly and represented by rough fuzzy variable. While simulating model (6.8) for each those five instances, the confidence levels $\alpha$ and $\beta$ are respectively, set to 0.9 and 0.4. Whereas, when the model (6.9) is considered, $\alpha$ and $\beta$ are respectively, set to 0.9 and 0.8. Subsequently, the random instances are solved with two MOGAs: NSGA-II and MOCHC. Moreover, the corresponding performance metrics, e.g., hypervolume ($HV$) (Zitzler and Thiele 1999), spread ($\Delta$) (Zhou et al. 2006), generational distance ($GD$) (Van Veldhuizen and Lamont 1998) and inverted generational distance ($IGD$) (Van Veldhuizen and Lamont 1998) are studied to make a comparison between the MOGAs. For any MOGA, larger values of $HV$ and smaller values of $\Delta$, $GD$ and $IGD$ are desirable. Moreover, for a particular MOGA, each of these performance metrics determines either the convergence or the diversity or the both. Particularly, the $HV$ and the $IGD$ metrics are used to measure both the convergence and diversity among the solutions. Furthermore, the diversity and the convergence among the solutions are measured by $\Delta$ and $GD$, respectively.

We have used jMetal 4.5 for the simulation of the two MOGAs: NSGA-II and MOCHC, for the instances. As the algorithms are stochastic in nature, we execute each of them for 100 times by considering 500 generations for each execution. Since there does not exist Pareto optimal solutions of any rough fuzzy bi-objective quadratic minimum spanning problem in the literature, therefore, we approximate the Pareto front of each of the instances by collecting their best solutions obtained from the first font of NSGA-II and MOCHC. The parameter settings for NSGA-II and MOCHC are considered the same as mentioned above in Section 6.4.

Subsequently, some statistical measures of the performance metrics are studied which are reported in Table 6.4 through Table 6.7, where the better values are highlighted as bold. Specifically, for all the performance metrics, the $mean$ and the standard deviation ($sd$) are shown in Table 6.4 and Table 6.6, whereas, the $median$ and the interquartile range ($IQR$) are reported in Table 6.5 and Table 6.7. Considering Table 6.4, MOCHC performs better than NSGA-II for all the five instances with respect to $\Delta$. Moreover, MOCHC becomes superior to NSGA-II relating to $GD$ and $IGD$, respectively for four and three instances. However, for $HV$, NSGA-II performs better than MOCHC, respectively for three instances. Similar to Table 6.4, in Table 6.5, MOCHC performs better than NSGA-II for all the instances as far as $\Delta$ is concerned. For $GD$ and $IGD$, MOCHC becomes superior to NSGA-II for three out of five instances. However, considering $HV$, NSGA-II performs better than MOCHC for three instances. Therefore, for most of the instances, in tables 6.4 and 6.5, we observe that MOCHC is superior to NSGA-II with respect to $\Delta$, $GD$ and $IGD$, and NSGA-II becomes superior to MOCHC with respect to $HV$.

Table 6.4 Mean and $sd$ of $HV$, $\Delta$, $GD$ and $IGD$ after 100 runs of NSGA-II and MOCHC on each of the five rough fuzzy instances of QMST at predetermined confidence levels, $\alpha = 0.9$ and $\beta = 0.4$

| MOGA | Rough fuzzy Instances generated randomly | $HV$ | | $\Delta$ | | $GD$ | | $IGD$ | |
|---|---|---|---|---|---|---|---|---|---|
| | | $mean$ | $sd$ | $mean$ | $sd$ | $mean$ | $sd$ | $mean$ | $sd$ |
| NSGA-II | $QMST$_10_20 | **7.52E-1** | 3.4E-2 | 1.48E+0 | 4.2E-2 | 3.45E-3 | 1.4E-3 | 5.43E-4 | 3.3E-4 |
| | $QMST$_20_40 | **7.69E-1** | 3.5E-2 | 1.45E+0 | 4.6E-2 | 4.24E-3 | 1.5E-3 | **1.69E-4** | 1.7E-4 |
| | $QMST$_30_60 | **7.35E-1** | 2.4E-2 | 1.73E+0 | 5.9E-2 | 3.28E-3 | 1.7E-3 | **5.49E-4** | 1.4E-4 |
| | $QMST$_40_80 | 7.87E-1 | 2.6E-2 | 1.59E+0 | 4.7E-2 | 3.97E-3 | 1.4E-3 | 2.51E-3 | 0.3E-3 |
| | $QMST$_50_100 | 8.07E-1 | 3.7E-2 | 1.69E+0 | 4.9E-2 | **1.80E-3** | 1.1E-3 | 2.15E-3 | 1.2E-3 |
| MOCHC | $QMST$_10_20 | 7.49E-1 | 3.2E-2 | **5.32E-1** | 4.1E-2 | **1.62E-3** | 0.7E-3 | **3.43E-4** | 1.9E-4 |
| | $QMST$_20_40 | 7.65E-1 | 3.8E-2 | **5.42E-1** | 5.2E-2 | **2.75E-3** | 1.9E-3 | 2.67E-4 | 4.3E-4 |
| | $QMST$_30_60 | 7.33E-1 | 2.2E-2 | **5.83E-1** | 4.9E-2 | **2.47E-3** | 1.6E-3 | 6.72E-4 | 2.7E-4 |
| | $QMST$_40_80 | **7.88E-1** | 1.7E-2 | **6.76E-1** | 4.5E-2 | **3.59E-3** | 2.3E-3 | **2.39E-3** | 0.8E-3 |
| | $QMST$_50_100 | **8.10E-1** | 2.7E-2 | **5.97E-1** | 4.3E-2 | 2.27E-3 | 0.7E-3 | **1.63E-3** | 2.7E-3 |

Table 6.5 Median and $IQR$ of $HV$, $\Delta$, $GD$ and $IGD$ after 100 runs of NSGA-II and MOCHC on each of the five rough fuzzy instances of QMST at predetermined confidence levels, $\alpha = 0.9$ and $\beta = 0.4$

| MOGAs | Rough fuzzy Instances generated randomly | $HV$ | | $\Delta$ | | $GD$ | | $IGD$ | |
|---|---|---|---|---|---|---|---|---|---|
| | | $median$ | $IQR$ | $median$ | $IQR$ | $median$ | $IQR$ | $median$ | $IQR$ |
| NSGA-II | $QMST$_10_20 | **7.87E-1** | 2.6E-2 | 1.36E+0 | 4.2E-2 | **1.79E-3** | 1.2E-3 | 5.37E-4 | 3.7E-4 |
| | $QMST$_20_40 | **7.96E-1** | 3.7E-2 | 1.53E+0 | 8.5E-2 | **2.85E-3** | 1.8E-3 | **1.98E-4** | 2.3E-4 |
| | $QMST$_30_60 | **7.67E-1** | 2.3E-2 | 1.72E+0 | 1.4E-2 | 2.98E-3 | 1.7E-3 | **5.83E-4** | 1.7E-4 |
| | $QMST$_40_80 | 7.73E-1 | 2.8E-2 | 1.68E+0 | 3.1E-2 | 5.87E-3 | 1.5E-3 | 2.58E-3 | 0.7E-4 |
| | $QMST$_50_100 | 8.14E-1 | 3.9E-2 | 1.87E+0 | 1.6E-2 | 1.63E-2 | 0.9E-3 | 2.49E-3 | 1.8E-3 |
| MOCHC | $QMST$_10_20 | 7.85E-1 | 3.4E-2 | **4.12E-1** | 5.4E-2 | 2.45E-3 | 1.5E-3 | **3.52E-4** | 2.4E-4 |
| | $QMST$_20_40 | 7.94E-1 | 4.8E-2 | **4.45E-1** | 1.2E-2 | 1.56E-2 | 7.2E-3 | 2.88E-4 | 4.7E-4 |
| | $QMST$_30_60 | 7.63E-1 | 2.5E-2 | **3.85E-1** | 3.5E-2 | **2.77E-3** | 1.4E-3 | 6.84E-4 | 2.7E-4 |
| | $QMST$_40_80 | **7.74E-1** | 2.3E-2 | **5.02E-1** | 5.7E-2 | **5.07E-3** | 2.6E-3 | **2.42E-3** | 1.5E-4 |
| | $QMST$_50_100 | **8.17E-1** | 3.4E-2 | **3.72E-1** | 6.9E-2 | **1.42E-2** | 7.8E-3 | **1.87E-3** | 3.3E-3 |

Considering Table 6.6, NSGA-II emerges as superior to MOCHC with respect to $GD$ and $IGD$, respectively for three and four instances. Conversely, compare to NSGA-II, MOCHC generate better $HV$ and $\Delta$, respectively for three and five instances. In Table 6.7, NSGA-II emerges as better than MOCHC, with respect to $GD$ and $IGD$ for three and four instances, respectively. On the other hand, MOCHC emerges as superior to NSGA-II, for $HV$ and $\Delta$, respectively for four and five instances. Therefore, for most of the instances in tables 6.6 and 6.7, MOCHC performs better than NSGA-II with respect to $HV$ and $\Delta$, while NSGA-II emerges as superior to MOCHC, for both $GD$ and $IGD$.

Table 6.6 Mean and $sd$ of $HV$, $\Delta$, $GD$ and $IGD$ after 100 runs of NSGA-II and MOCHC on each of the five rough fuzzy instances of QMST at predetermined confidence levels, $\alpha = 0.9$ and $\beta = 0.8$

| MOGAs | Rough fuzzy Instances generated randomly | $HV$ | | $\Delta$ | | $GD$ | | $IGD$ | |
|---|---|---|---|---|---|---|---|---|---|
| | | $mean$ | $sd$ | $mean$ | $sd$ | $mean$ | $sd$ | $mean$ | $sd$ |
| NSGA-II | $QMST$_10_20 | **7.72E-1** | 2.7E-2 | 1.52E+0 | 4.6E-2 | **1.94E-3** | 1.2E-3 | **2.63E-4** | 0.7E-4 |
| | $QMST$_20_40 | 7.87E-1 | 3.5E-2 | 1.72E+0 | 4.2E-2 | **2.97E-3** | 1.4E-3 | 4.34E-4 | 1.5E-4 |
| | $QMST$_30_60 | **7.85E-1** | 1.3E-2 | 1.62E+0 | 6.1E-2 | 2.73E-3 | 8.3E-4 | **2.32E-4** | 2.2E-4 |
| | $QMST$_40_80 | 7.45E-1 | 2.8E-2 | 1.04E+0 | 7.0E-2 | 3.37E-3 | 1.8E-3 | **7.04E-4** | 1.9E-4 |
| | $QMST$_50_100 | 7.93E-1 | 2.3E-2 | 1.91E+0 | 8.1E-2 | **6.85E-3** | 2.9E-3 | **3.27E-3** | 2.9E-3 |
| MOCHC | $QMST$_10_20 | 7.70E-1 | 1.6E-2 | **6.73E-1** | 4.9E-2 | 3.52E-3 | 1.2E-3 | 3.27E-4 | 0.8E-4 |
| | $QMST$_20_40 | **7.88E-1** | 2.2E-2 | **6.42E-1** | 5.4E-2 | 4.54E-3 | 2.2E-3 | **3.02E-4** | 1.2E-4 |
| | $QMST$_30_60 | 7.83E-1 | 1.1E-2 | **7.51E-1** | 3.2E-2 | **1.50E-3** | 4.9E-4 | 4.69E-4 | 3.8E-4 |
| | $QMST$_40_80 | **7.47E-1** | 3.0E-2 | **6.73E-1** | 4.1E-2 | **2.23E-3** | 0.5E-3 | 7.83E-4 | 1.5E-4 |
| | $QMST$_50_100 | **7.94E-1** | 3.3E-2 | **7.14E-1** | 3.7E-2 | 8.49E-3 | 3.2E-3 | 3.97E-3 | 1.9E-3 |

## 6.6 Case Study

To illustrate an application of the proposed b-RFQMSTP, we present an example related to an airline service network of a company. We summarize the problem as follows. An airline company is willing to start its service by considering ten major cities of a country. For this purpose, nineteen possible air routes are considered with which the cities can be connected. Moreover, to attract more passengers, the company offers a discount on total flight price, if a passenger avails two consecutive flight, in the course of a journey. Now, to optimize the cost, the management should consider a flight schedule for the next quarter so that the total cost of providing flight services between the cities and the possible amount of discounts, which are to be borne by the company, are minimized. In this context, the company also identifies several factors like fuel price, labour cost, maintenance and overhaul costs of aircraft, which influence the cost of providing air service and essentially fluctuates from time to time. It is quite common for the decision makers (project managers) to be uncertain about estimating the expenses related to flight service and the discounts on flight fares. Therefore, the estimations of the decision makers on the cost of air service and the discount amount are considered as rough fuzzy variables.

Table 6.7 Median and $IQR$ of $HV$, $\Delta$, $GD$ and $IGD$ after 100 runs of NSGA-II and MOCHC on each of the five rough fuzzy instances of QMST at predetermined confidence levels, $\alpha = 0.9$ and $\beta = 0.8$

| MOGAs | Rough fuzzy Instances generated randomly | $HV$ | | $\Delta$ | | $GD$ | | $IGD$ | |
|---|---|---|---|---|---|---|---|---|---|
| | | $median$ | $IQR$ | $median$ | $IQR$ | $median$ | $IQR$ | $median$ | $IQR$ |
| NSGA-II | $QMST$_10_20 | 7.68E-1 | 1.4E-2 | 1.64E+0 | 5.5E-2 | **3.57E-3** | 1.3E-3 | **3.49E-4** | 1.7E-4 |
| | $QMST$_20_40 | 7.63E-1 | 2.1E-2 | 1.72E+0 | 6.2E-2 | **2.60E-3** | 1.7E-3 | **3.98E-4** | 1.4E-4 |
| | $QMST$_30_60 | **7.72E-1** | 1.5E-2 | 9.34E+0 | 1.8E-1 | 2.79E-3 | 1.5E-3 | **1.52E-4** | 2.7E-4 |
| | $QMST$_40_80 | 7.43E-1 | 2.6E-2 | 1.85E+0 | 1.7E-1 | 3.07E-3 | 1.2E-3 | 7.84E-4 | 3.5E-4 |
| | $QMST$_50_100 | 7.74E-1 | 1.9E-2 | 1.62E+0 | 1.1E-1 | **6.57E-3** | 3.6E-3 | **3.18E-3** | 2.8E-3 |
| MOCHC | $QMST$_10_20 | **7.69E-1** | 2.7E-2 | **6.74E-1** | 4.6E-2 | 4.68E-3 | 1.2E-3 | 3.94E-4 | 1.9E-4 |
| | $QMST$_20_40 | **7.65E-1** | 3.1E-2 | **6.24E-1** | 7.1E-2 | 3.43E-3 | 1.2E-3 | 5.82E-4 | 2.1E-4 |
| | $QMST$_30_60 | 7.71E-1 | 1.7E-2 | **7.08E-1** | 5.6E-2 | **1.72E-3** | 7.6E-4 | 3.37E-4 | 3.5E-4 |
| | $QMST$_40_80 | **7.44E-1** | 2.2E-2 | **6.39E-1** | 5.2E-2 | **2.94E-3** | 2.3E-3 | **6.67E-4** | 3.2E-4 |
| | $QMST$_50_100 | **7.75E-1** | 1.8E-2 | **7.53E-1** | 6.9E-2 | 8.19E-3 | 2.7E-3 | 3.68E-3 | 2.9E-3 |

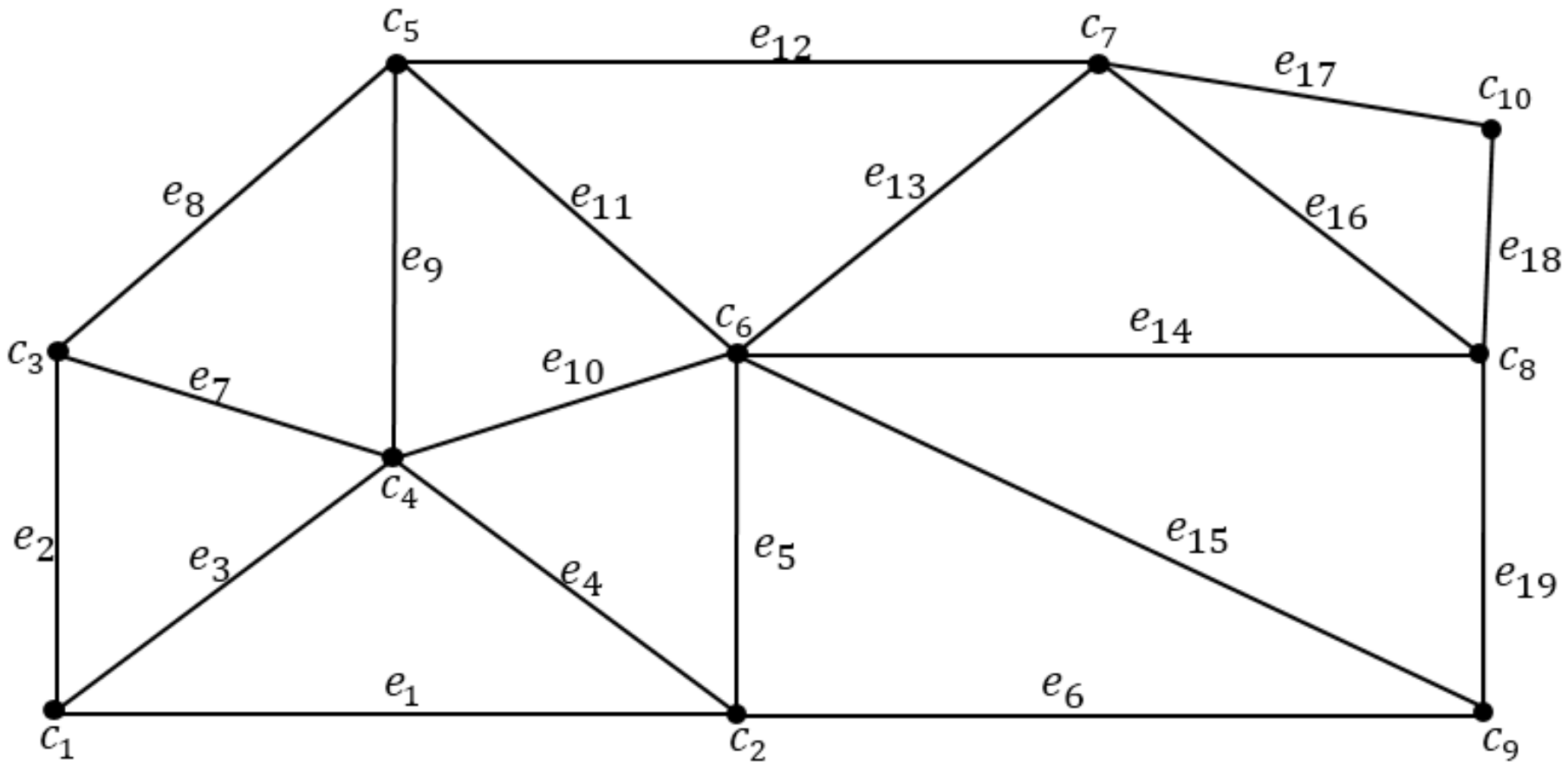

Figure 6.6 A weighted connected undirected network $W$

The project managers consider the possible air services between the cities as a weighted connected network (WCN), $W$=$(C_W, E_W)$ as shown in Fig. 6.6. Here, each vertex represents a city $c_i, i = 1,2, \dots, |C_W|$ and an edge $e_j, j = 1,2, \dots, |E_W|$ represents the possible air service between the pair of cities to be provided by the company. Each edge of $W$ is associated with a linear weight, which determines the possible cost to the company while providing the air service between a pair of cities. Moreover, the quadratic weight, for any pair of adjacent edges, determines the possible discount on the total flight fare of two consecutive flights. In this case study, these weights are considered randomly. The linear weights are reported in Table E.1. Whereas, the quadratic weights are represented in tables E.2 and E.3. These tables are presented in Appendix E.

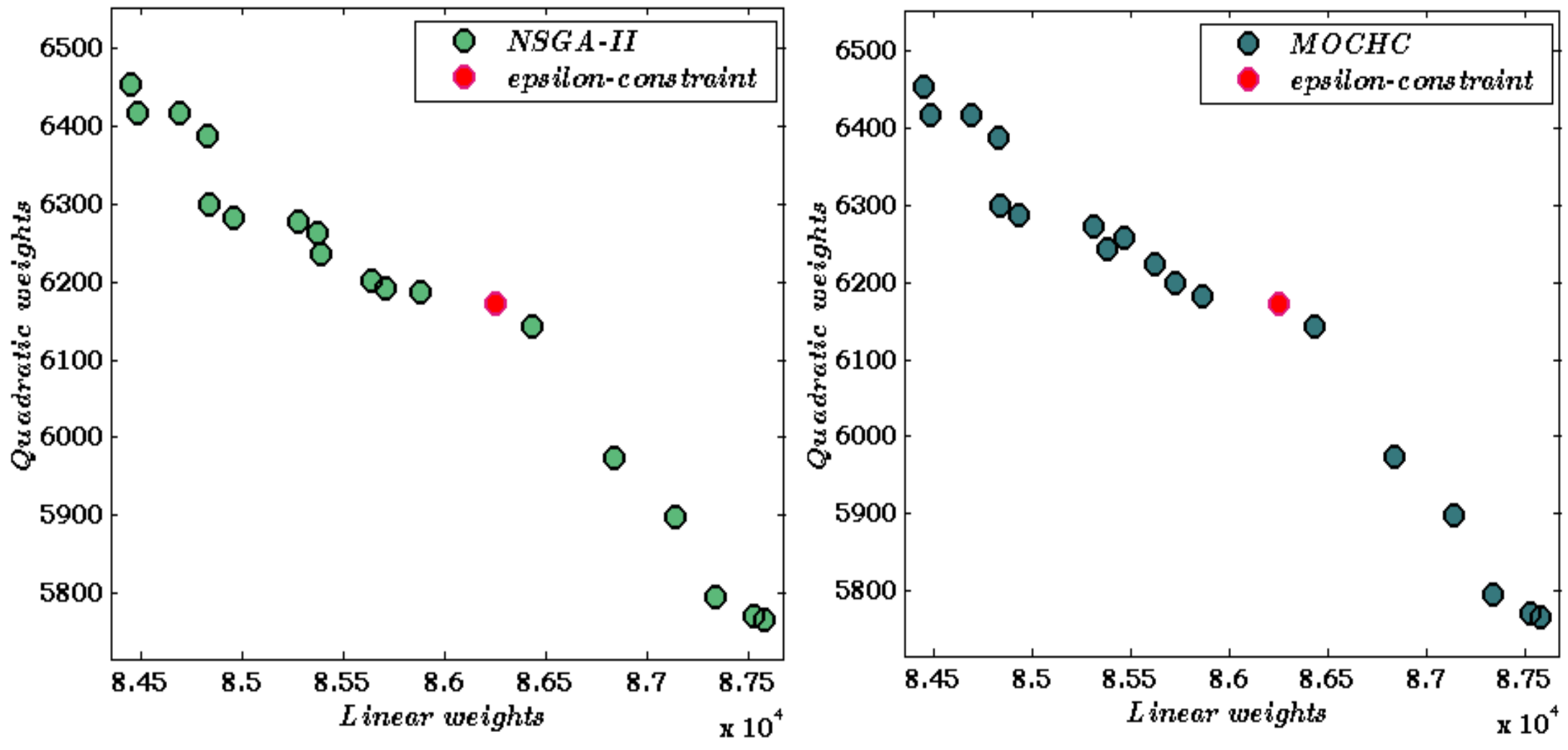

(a) (b)

Figure 6.7 Nondominated solutions of the model (6.9) for the WCUN $W$ at predetermined confidence levels, $\alpha = 0.9$ and $\beta = 0.8$, when solved by the epsilon-constraint method, and (a) NSGA-II and (b) MOCHC

In order to determine an optimize flight schedule, the project managers find a quadratic minimum spanning tree in $W$ using model (6.5). Assuming, the predetermined confidence levels of the decision makers as 0.9 for $\alpha_1$ and $\alpha_2$, and 0.8 for $\beta_1$ and $\beta_2$, the corresponding crisp equivalent model (6.9) is solved using the epsilon-constraint method, and the corresponding results are shown in Table 6.8. Since $\alpha_1$ and $\alpha_2$ have the same value, we represent $\alpha_1$ and $\alpha_2$ as $\alpha$. Similarly, $\beta_1$ and $\beta_2$ are represented by $\beta$. Here, $\bar{\bar{f}}_1$ represents the total possible expense for providing air services between all the cities and $\bar{\bar{f}}_2$ determines the total possible discount to be offered by the company. The confidence levels $\alpha$ and $\beta$ of the decision makers are respectively, set to 0.9 and 0.8.

Table 6.8 A compromise solution of the model (6.9) for the WCUN $W$

| Objective values $\langle \bar{\bar{f}}_1, \bar{\bar{f}}_2 \rangle$ | Optimal decision vector $x$ |
|---|---|
| $\langle 86250.00, 6172.00 \rangle$ | $(1\,1\,0\,0\,0\,1\,1\,0\,1\,0\,0\,1\,0\,1\,0\,0\,0\,1\,0\,1)^T$ |

Model (6.9) is also solved with two MOGAS: NSGA-II and MOCHC for $W$. The corresponding nondominated solutions are shown in Fig.6.7. Here, each solution represents a possible flight schedule, which the decision makers can select. From Fig. 6.7, we observe that the solution generated by the epsilon-constraint method is nondominated with the solutions generated by NSGA-II and MOCHC. Here, it is to be mentioned, that the project managers want to determine an optimize flight schedule by considering (i) the total possible expense for providing air service between all the cities and (ii) the total possible discount to be offered by the company. Multiple nondominated compromise solutions are shown in Fig. 6.7, among which, one of the solutions, reported in Table 6.8, is depicted in Fig. 6.8.

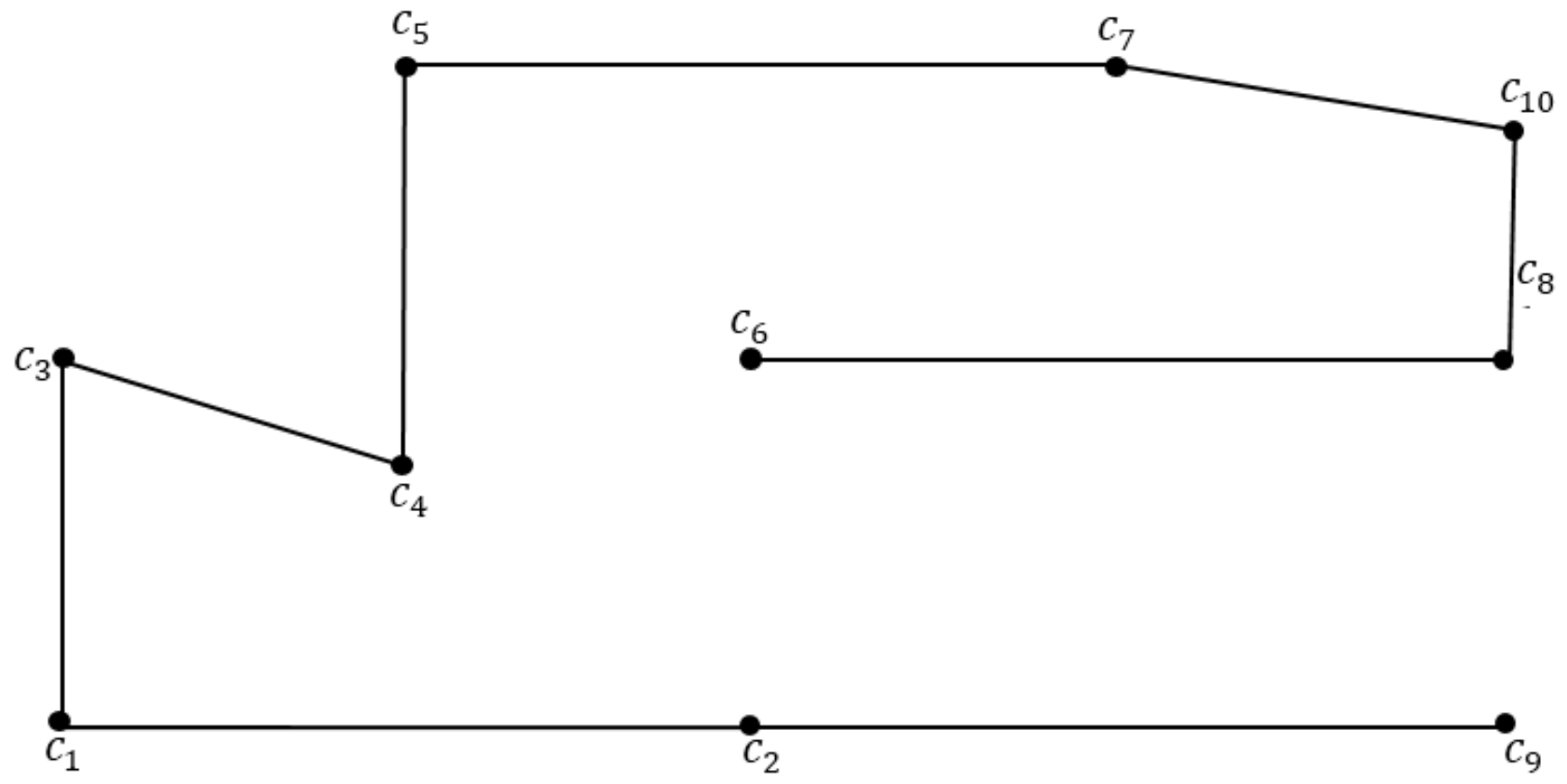

Figure 6.8 A possible flight schedule for the compromise solution of the WCUN $W$ as reported in Table 6.8

## 6.7 Conclusion

In this chapter, we have proposed a b-RFQMSTP which is modelled using the rough fuzzy chance-constrained programming technique. The linear and quadratic weights of the proposed model are considered as two different objectives which are optimized simultaneously. Till now, any study under such rough fuzzy uncertain environment is yet to be conducted in the literature, where the associated parameters are essentially considered as rough fuzzy variables. The crisp equivalent of the proposed model is solved by the epsilon-constraint method and by two MOGAs: NSGA-II and MOCHC. For each of the MOGAs, the proposed model is also solved for five randomly generated rough fuzzy instances, at two different combinations of the confidence levels: (i) $\alpha = 0.9$ and $\beta = 0.4$ and (ii) $\alpha = 0.9$ and $\beta = 0.8$. The performance of these MOGAs are analyzed for all the five instances with respect to four different performance metrics: (i) $HV$, (ii) $\Delta$, (iii) $GD$ and (iv) $IGD$. Moreover, in this study, we have also presented an example to numerically illustrate an application of the proposed model.

In future, the rough fuzzy chance-constrained programming model can also be extended to model different network optimization problems including shortest path, maximum flow and transportation. Moreover, optimization problems in the domains of portfolio, inventory, supply chain, etc., can also be addressed using rough fuzzy chance-constrained programming.

# Chapter 7
# Conclusion and Future Scope

# Chapter 7

# Conclusion and Future Scope

## 7.1 Conclusion

In this dissertation, we have proposed two single and three multi-objective network optimization problems under different uncertain paradigms including type-2 fuzzy, random fuzzy, rough, rough fuzzy, and uncertainty theory. Those uncertain network models are formulated using uncertain programming techniques and are solved by different classical and evolutionary algorithms. The main contribution of this thesis is presented in five chapters, which are again categorized into two parts. Part-I consists of Chapter 2 and Chapter 3. While, Part-II contains Chapter 4, Chapter 5 and Chapter 6.

Chapter 1 contains introduction of the thesis, and conclusion and future scope of the thesis are discussed in Chapter 7. In Chapter 2, we have proposed a linear programming problem with interval type-2 fuzzy parameters. The formulation of the LPP is based on chance-constrained programming and credibility measure of IT2FV. The applicability of the proposed method is demonstrated by solving three network problems: (i) solid transportation problem (STP), (ii) shortest path problem (SPP) and (iii) minimum spanning tree problem (MSTP).

The study in Chapter 3 deals with random fuzzy maximum flow problem (MFP), where we have proposed the corresponding expected value model (EVM) and chance-constrained model (CCM) of the problem. We have also developed the varying population genetic algorithm with indeterminate crossover (VPGAwIC) to solve the crisp equivalent of the MFP. In the proposed algorithm, the selection of a chromosome depends on its lifetime. For this purpose, in our study, we have also proposed an improved lifetime allocation strategy (iLAS). The proposed algorithm is compared with other similar algorithms, and their performances are analyzed.

Chapter 4 deals with uncertain multi-criteria shortest path problem (MSPP) under rough environment. The equivalent CCM is formulated for the proposed problem. To solve the problem, we have developed the modified rough Dijkstra's (MRD) algorithm. The MSPP is solved with a classical multi-objective solution technique based on the goal attainment method and a multi-objective genetic algorithm (MOGA), nondominated sorting algorithm II (NSGA-II). The performances of these algorithms are also analyzed.

In Chapter 5, we have proposed an uncertain multi-objective multi-item fixed charge solid transportation problem with budget constraint. The model is formulated under the framework of Liu's uncertainty theory. Three equivalent uncertain programming models: (i) EVM, (ii) CCM and (iii) dependent chance-constrained model (DCCM) are developed for the proposed model, which are eventually solved by three classical multi-objective solution methods: (i) linear weighted method, (ii) global criterion method and (iii) fuzzy programming method. For each of these models, the corresponding results are also compared.

Chapter 6 deals with a bi-objective quadratic minimum spanning tree problem. Here, we have considered a rough fuzzy hybrid framework to model the proposed bi-objective rough fuzzy quadratic minimum spanning tree problem (b-RFQMSTP). The proposed model is solved with the epsilon-constraint method and two MOGAs: NSGA-II and multi-objective cross generational, elitist selection, heterogeneous recombination and cataclysmic mutation (MOCHC). Subsequently, the performances of the MOGAs are analysed with respect to four performance metrics: hypervolume ($HV$), spread ($\Delta$), generational distance ($GD$) and inverted generational distance ($IGD$). Moreover, we have provided a case study to numerically illustrate the proposed b-RFQMSTP.

## 7.2 Future Scope

Several future directions of the research works, proposed in our dissertation are reported below.

In Chapter 2, the proposed linear programming for solving single objective optimization problem based on interval type-2 fuzzy parameters can be extended by considering the critical values (Qin et al. 2011) of discrete type-2 fuzzy variables. Our proposed method can also be applied to other optimization problems like inventory, portfolio and supply chain problems. Moreover, the proposed model can also be extended to solve these problems for multi-objective versions.

The study presented in Chapter 3 can be considered as the basis for future studies for determining efficient solution procedure for maximum-flow based combinatorial optimization problems under different uncertain environment. The computational overhead of the proposed VPGAwIC algorithm can be improved and implemented to solve other combinatorial optimization problems. Further, the maximum flow problem can also be extended to bi-objective maximum flow minimum cost problem under uncertain environment, which can be solved by multi-objective version of the proposed VPGAwIC.

In Chapter 4, the research work related to multi-objective shortest path problem under rough environment can be extended to address multi-objective version of bottleneck SPP, minimum deviation SPP, $k$-shortest path problem ($k$-SPP), etc., under different uncertain frameworks. Different multi-objective evolutionary algorithms can also be used to solve those problems.

In Chapter 5, the proposed variant of multi-objective transportation problem developed under Liu's uncertainty theory can still be extended by considering a multi-objective multi-item fixed charge solid multi-stage transportation problem with product blending, price discounts and breakable/deteriorating items as additional constraints. Till now, for any transportation problem, it is assumed that a transportation activity between source and destination takes place through the shortest route. However, in reality, different routes are available between source and destinations which may have different criteria, such as safety (particularly, in roadways or railways) and carbon emission. Therefore, solid transportation problem, which is a 3D-transportation problem, can be extended to model and solve the 4D-transportation problem considering the different routes for transportation in addition to the existing three parameters: (i) availability at sources, (ii) demand at destinations and (iii) capacity of conveyances.

In Chapter 6, the bi-objective quadratic spanning tree problem can be extended to multi-objective version of minimum spanning tree problem with conflict pairs (Darmann et al. 2011; Zhang et al. 2011), quadratic bottleneck spanning tree problem (Zhang and Punnen 2013), capacitated MSTP (Chandy and Lo 1973) and degree-constrained MSTP (Garey and Johnson 1990) under different uncertain environments. Different multi-objective evolutionary algorithms can also be considered as possible solution methodologies of those problems.

# Appendix A

In this section, we present an example of a WCDN, with edge weights represented as rough variable. Moreover, we present a WCDN $H$, where each edge is associated with three criteria which are considered as rough variables.

**Example A.1**: Consider a WCDN of four cities. A city is represented by a vertex, and an edge connecting two cities is represented by the cost incurred while travelling from city $v_i$ to $v_j$. The cost of travelling from $v_i$ to $v_j$ depends on several factors like fuel price, toll charges, labour costs, etc., and since each of these factors fluctuate from time to time, it is not always possible to determine exact travelling cost between a pair of cities. Suppose that four experts are chosen to determine the possible cost $c_{ij}$ while traversing from $v_i$ to $v_j$ during a certain time period. The experts feedback are obtained as intervals which are, $[100, 150]$, $[108,146.90]$, $[110,155.50]$ and $[105.50,149.50]$, respectively.

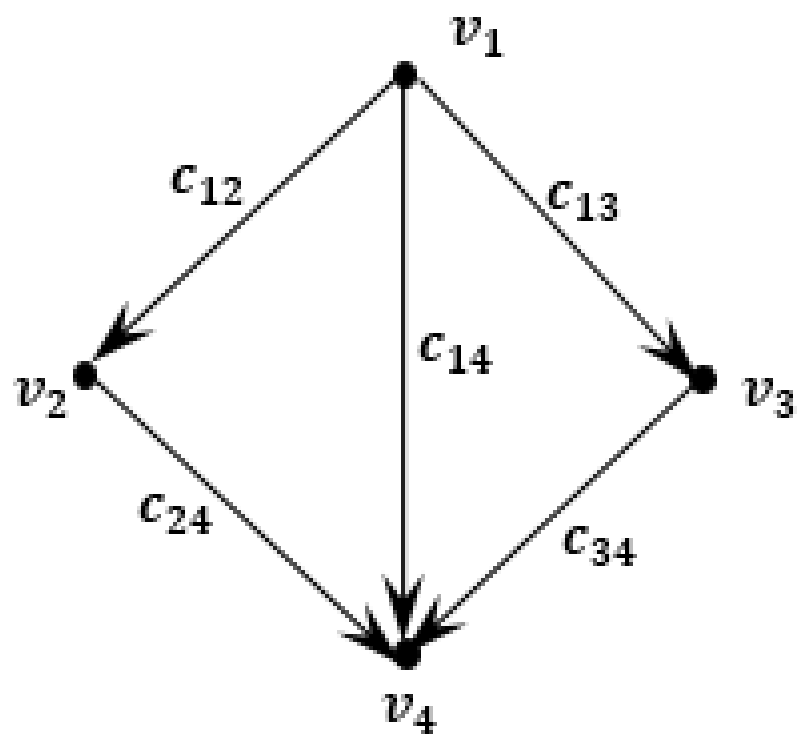

Figure A.1 A WCDN $S$ with associated edge weights represented by rough variable

Table A.1 Edge costs of $S$ estimated by experts

| Edge Cost | Expert 1 | Expert 2 | Expert 3 | Expert 4 | Rough Costs |
|---|---|---|---|---|---|
| $c_{12}$ | $[100, 150]$ | $[108,146.90]$ | $[110,155.50]$ | $[105.50,149.50]$ | $[110,146.90]$, $[100,155.50]$ |
| $c_{13}$ | $[116,127.45]$ | $[122.43,131.45]$ | $[113.40,135]$ | $[123.71,136.23]$ | $[123.71,127.45]$, $[113.40, 136.23]$ |
| $c_{14}$ | $[67,81]$ | $[55,73]$ | $[61,72]$ | $[56.98,69.86]$ | $[67, 69.86]$, $[55,81]$ |
| $c_{24}$ | $[161,174]$ | $[157,176.23]$ | $[154,168]$ | $[162,170]$ | $[162,168]$, $[154,176.23]$ |
| $c_{34}$ | $[117, 124.56]$ | $[112.43,126]$ | $[119.78, 129.50]$ | $[120.25, 128.75]$ | $[120.25,124.56]$, $[112.43, 129.50]$ |

Considering all these intervals, the lower approximation of $c_{ij}$ becomes $[110, 146.90]$, i.e., every member of $[110,146.90]$ are certain values of $c_{ij}$. In other words, the interval selected as the lower approximation of $c_{ij}$, satisfies the consent of all the experts.

Further, the interval, $[100,155.50]$ becomes the upper approximation of $c_{ij}$ which include all the feedback received from the experts such that the elements of $[100,155.50]$ become the possible values of $c_{ij}$. Therefore, $c_{ij}$ can be represented by a rough variable $([110,146.90],[100,155.50])$. Fig. A.1 depicts a WCDN $S$ for four cities and the corresponding costs as determined by the experts are shown in Table A.1.

**Example A.2**: Here, we present a WCDN $R$, where each edge is associated with multiple criteria and are represented by rough variable. Fig. A.2 depicts the WCDN $R$ and its corresponding edge weights are reported in Table A.2.

Table A.2 Edge weights of $R$: distance, cost and time represented by rough variable

| Edges $(e_{ij})$ | Distance $\left(P_{ij}^1 = ([P_2^1, P_3^1], [P_1^1, P_4^1])\right)$ | Cost $\left(P_{ij}^2 = ([P_2^2, P_3^2], [P_1^2, P_4^2])\right)$ | Time $\left(P_{ij}^3 = ([P_2^3, P_3^3], [P_1^3, P_4^3])\right)$ |
|---|---|---|---|
| $e_{12}$ | $[9,11],[8,15]$ | $[21,26],[19,27]$ | $[24,27],[23,29]$ |
| $e_{13}$ | $[17,21],[14,25]$ | $[14,15],[12,17]$ | $[11,12],[10,14]$ |
| $e_{14}$ | $[30,36],[27,38]$ | $[9,11],[7,12]$ | $[30,31],[24,32]$ |
| $e_{25}$ | $[26,28],[22,33]$ | $[14,19],[11,20]$ | $[27,28],[26,29]$ |
| $e_{37}$ | $[8,12],[6,17]$ | $[15,18],[14,21]$ | $[10,11],[9,15]$ |
| $e_{45}$ | $[31,35],[29,39]$ | $[27,29],[24,30]$ | $[26,28],[21,32]$ |
| $e_{46}$ | $[27,29],[24,36]$ | $[12,14],[10,15]$ | $[24,26],[20,27]$ |
| $e_{58}$ | $[36,38],[42,40]$ | $[18,19],[17,21]$ | $[18,21],[17,29]$ |
| $e_{59}$ | $[31,33],[29,37]$ | $[21,26],[18,27]$ | $[21,22],[20,23]$ |
| $e_{69}$ | $[11,17],[10,21]$ | $[18,19],[14,21]$ | $[18,19],[17,21]$ |
| $e_{710}$ | $[30,35],[28,38]$ | $[27,29],[24,32]$ | $[14,15],[11,17]$ |
| $e_{815}$ | $[25,28],[24,31]$ | $[30,31],[29,34]$ | $[21,24],[19,25]$ |
| $e_{97}$ | $[18,23],[15,27]$ | $[8,11],[7,14]$ | $[15,17],[14,18]$ |
| $e_{98}$ | $[27,29],[23,34]$ | $[10,15],[9,17]$ | $[20,21],[17,22]$ |
| $e_{912}$ | $[5,8],[4,10]$ | $[18,21],[15,23]$ | $[31,32],[29,35]$ |
| $e_{1013}$ | $[11,17],[9,20]$ | $[15,18],[12,19]$ | $[12,14],[10,17]$ |
| $e_{119}$ | $[21,26],[18,29]$ | $[25,28],[21,30]$ | $[28,29],[27,32]$ |
| $e_{1112}$ | $[28,30],[24,35]$ | $[22,24],[20,25]$ | $[15,18],[14,17]$ |
| $e_{1213}$ | $[7,9],[5,11]$ | $[12,14],[11,15]$ | $[14,17],[12,19]$ |
| $e_{1214}$ | $[17,18],[15,20]$ | $[28,32],[23,34]$ | $[11,12],[10,21]$ |
| $e_{1311}$ | $[28,32],[26,37]$ | $[31,34],[29,35]$ | $[18,19],[17,21]$ |
| $e_{1314}$ | $[7,9],[3,14]$ | $[27,29],[26,30]$ | $[25,28],[24,34]$ |
| $e_{1415}$ | $[21,27],[22,32]$ | $[24,26],[23,28]$ | $[18,20],[17,24]$ |
| $e_{1416}$ | $[35,38],[33,39]$ | $[9,11],[8,12]$ | $[28,29],[27,30]$ |
| $e_{1512}$ | $[5,8],[4,11]$ | $[28,29],[25,31]$ | $[25,26],[24,27]$ |
| $e_{1516}$ | $[28,29],[24,31]$ | $[36,37],[33,39]$ | $[21,25],[20,28]$ |
| $e_{1517}$ | $[41,46],[39,47]$ | $[42,45],[41,46]$ | $[37,39],[34,46]$ |
| $e_{1617}$ | $[36,39],[31,42]$ | $[39,42],[37,45]$ | $[40,42],[39,48]$ |

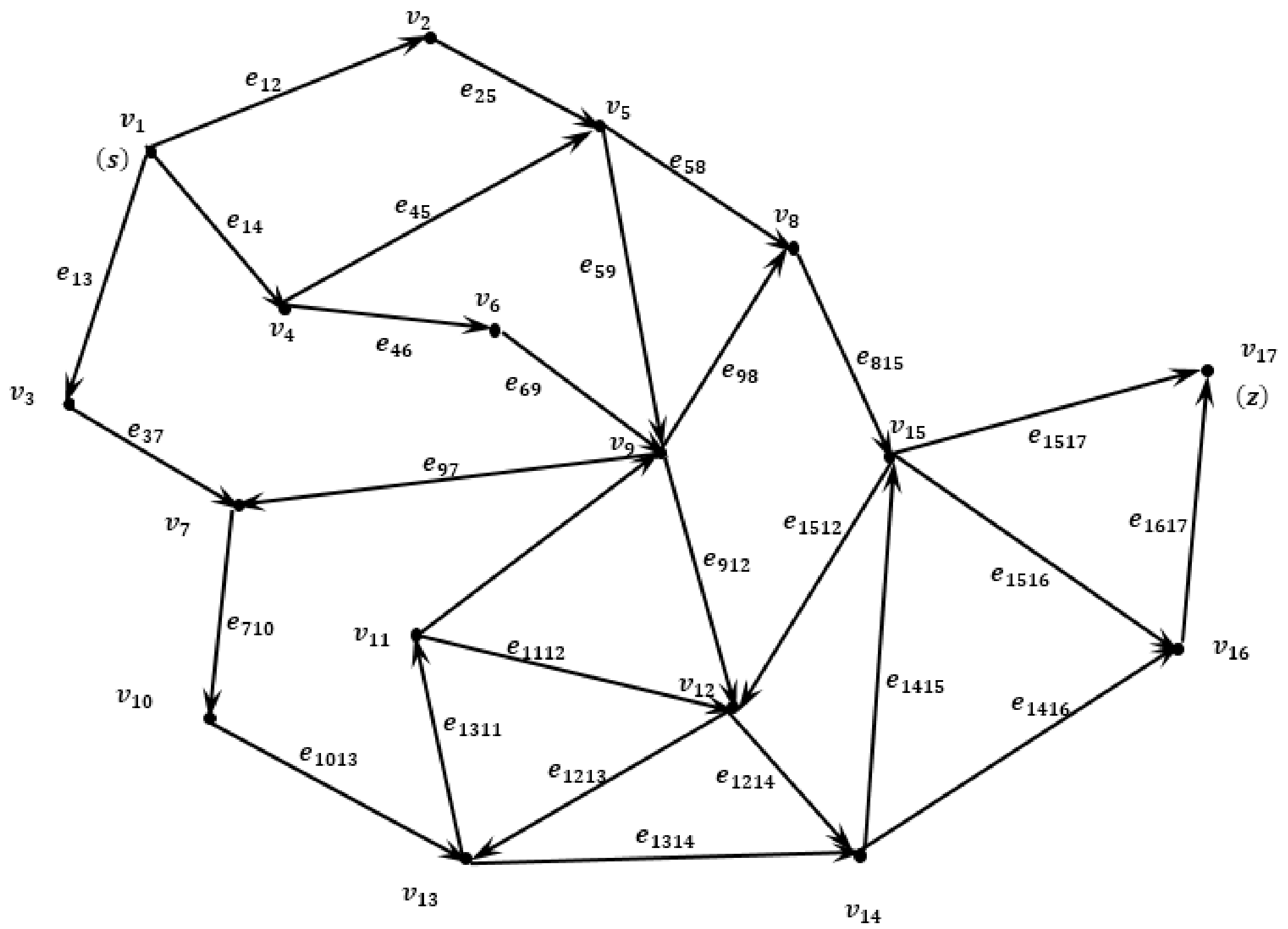

Figure A.2 A WCDN $R$ with associated edge weights represented by rough variable

# Appendix B

In this section, some theorems related to uncertain programming are revisited.

**Theorem B.1** (B. Liu 2010): Let $\zeta_1, \zeta_2, . . . , \zeta_n$ are independent uncertain variables with uncertainty distributions $\Phi_1, \Phi_2, ..., \Phi_n$, respectively, and $f_1(x), f_2(x), . . ., f_n(x)$, $\overline{f}(x)$ are real-valued functions. Then

$$\mathcal{M}\{\sum_{i=1}^{n} \zeta_i f_i(x) \leq \overline{f}(x)\} \geq \alpha$$

holds if and only if,

$$\sum_{i=1}^{n} \Phi_i^{-1}(\alpha) f_i^{+}(x) - \sum_{i=1}^{n} \Phi_i^{-1}(1-\alpha) f_i^{-}(x) \leq \overline{f}(x), \tag{B1}$$

where $f_i^{+}(x) = \begin{cases} f_i(x) \,; if \; f_i(x) > 0 \\ 0 \quad ; if \; f_i(x) \leq 0 \end{cases}$ (B2)

and

$$f_i^{-}(x) = \begin{cases} 0 \quad ; if \; f_i(x) \geq 0 \\ -f_i(x) \,; if \; f_i(x) < 0 \end{cases} \text{ for } i = 1,2, \ldots, n. \tag{B3}$$

If $f_1(x), f_2(x), . . ., f_n(x)$ are all non-negative, then (B1) becomes $\sum_{i=1}^{n} \Phi_i^{-1}(\alpha) f_i(x) \leq \overline{f}(x)$, and if $f_1(x)$, $f_2(x), . . ., f_n(x)$ are all non-positive, then (B1) becomes $\sum_{i=1}^{n} \Phi_i^{-1}(1-\alpha) f_i(x) \leq \overline{f}(x)$.

**Theorem B.2** (B. Liu 2010): Let $x_1, x_2, . . ., x_n$ are non-negative decision variables and $\zeta_1, \zeta_2, . . ., \zeta_n$ are independent zigzag uncertain variables which are represented by $\mathcal{Z}(g_1, h_1, l_1), \mathcal{Z}(g_2, h_2, l_2), . . ., \mathcal{Z}(g_n, h_n, l_n)$, respectively. Then

$$\mathcal{M}\{\sum_{i=1}^{n} \zeta_i x_i \leq c\} = \begin{cases} 0 & ; if \; c < \sum_{i=1}^{n} g_i x_i \\ \frac{c - \sum_{i=1}^{n} g_i x_i}{2\sum_{i=1}^{n}(h_i - g_i) x_i} & ; if \; c \in [\sum_{i=1}^{n} g_i x_i, \sum_{i=1}^{n} h_i x_i] \\ \frac{c + \sum_{i=1}^{n}(l_i - 2h_i) x_i}{2\sum_{i=1}^{n}(l_i - h_i) x_i} & ; if \; c \in [\sum_{i=1}^{n} h_i x_i, \sum_{i=1}^{n} l_i x_i] \\ 1 & ; if \; c > \sum_{i=1}^{n} h_i x_i \end{cases} \tag{B4}$$

**Theorem B.3** (B. Liu 2010): Let $x_1, x_2, . . ., x_n$ are non-negative decision variables and $\zeta_1, \zeta_2, . . ., \zeta_n$ are independent normal uncertain variables which are denoted as $\mathcal{N}(\rho_1, \sigma_1), \mathcal{N}(\rho_2, \sigma_2), . . ., \mathcal{N}(\rho_n, \sigma_n)$, respectively. Then

$$\mathcal{M}\{\sum_{i=1}^{n} \zeta_i x_i \leq c\} = \left(1 + \exp\left(\frac{\pi(\sum_{i=1}^{n} \rho_i x_i - c)}{\sqrt{3}\sum_{i=1}^{n} \sigma_i x_i}\right)\right)^{-1}. \tag{B5}$$

**Theorem B.4** (B. Liu 2010): Let $\zeta$ be an uncertain variable with continuous uncertainty distribution $\Phi$. Then, for any real number $x$, we have

$$\mathcal{M}\{\zeta \leq x\} = \Phi(x), \mathcal{M}\{\zeta \geq x\} = 1 - \Phi(x). \tag{B6}$$

# Appendix C

In this section, we state the relevant theorems to formulate the crisp equivalents of the chance-constrained model (CCM) and dependent chance-constrained model (DCCM) of UMMFSTPwB.

### Crisp Equivalents of Chance-constrained Model (CCM)

**Lemma C.1**: If $a$ and $r$ are positive real numbers, $\xi$ is an independent uncertain variable with uncertainty distribution $\Phi$ and $\alpha$ is the chance level. Then, $\mathcal{M}\{a-\xi \geq r\} \geq \alpha$ holds if and only if $a-\Phi^{-1}(\alpha) \geq r$.

**Proof**: $\mathcal{M}\{a-\xi \geq r\} \geq \alpha \;\Leftrightarrow\; \mathcal{M}\{\xi \leq a-r\} \geq \alpha \;\Leftrightarrow\; a-r \geq \Phi^{-1}(\alpha) \;\Leftrightarrow\; a-\Phi^{-1}(\alpha) \geq r$.

**Theorem C.1**: Let $\xi_{c^p_{ijk}}, \xi_{f^p_{ijk}}, \xi_{t^p_{ijk}}, \xi_{a^p_i}, \xi_{b^p_j}, \xi_{e_k}$ and $\xi_{B_j}$ are the independent uncertain variables, respectively associated with uncertainty distributions $\Phi_{\xi_{c^p_{ijk}}}, \Phi_{\xi_{f^p_{ijk}}}, \Phi_{\xi_{t^p_{ijk}}}, \Phi_{\xi_{a^p_i}}, \Phi_{\xi_{b^p_j}}, \Phi_{\xi_{e_k}}$ and $\Phi_{\xi_{B_j}}$ then the crisp equivalent of chance-constrained model (CCM) in model (5.3) can be equivalently formulated as model (C1).

$$\begin{cases}
Max\ \bar{Z}_1 = \left[\sum_{p=1}^{r}\sum_{i=1}^{m}\sum_{j=1}^{n}\sum_{k=1}^{K}\left\{\left(s_j^p - v_i^p - \Phi^{-1}_{\xi_{c^p_{ijk}}}(\alpha_1)\right)x^p_{ijk} - \Phi^{-1}_{\xi_{f^p_{ijk}}}(\alpha_1)y^p_{ijk}\right\}\right] \\
Min\ \bar{Z}_2 = \sum_{p=1}^{r}\sum_{i=1}^{m}\sum_{j=1}^{n}\sum_{k=1}^{K}\Phi^{-1}_{\xi_{t^p_{ijk}}}(\alpha_2)y^p_{ijk} \\
\quad subject\ to \qquad\qquad\qquad\qquad\qquad\qquad\qquad\qquad (\mathrm{C1}) \\
\quad \sum_{j=1}^{n}\sum_{k=1}^{K}x^p_{ijk} - \Phi^{-1}_{\xi_{a^p_i}}\left(1-\beta^p_i\right) \leq 0, i = 1,2,\dots,m, p = 1,2,\dots,r \\
\quad \sum_{i=1}^{m}\sum_{k=1}^{K}x^p_{ijk} - \Phi^{-1}_{\xi_{b^p_j}}\left(\gamma^p_j\right) \geq 0\,, j = 1,2,\dots,m, p = 1,2,\dots,r \\
\quad \sum_{p=1}^{r}\sum_{i=1}^{m}\sum_{j=1}^{n}x^p_{ijk} - \Phi^{-1}_{\xi_{e_k}}(1-\delta_k) \leq 0, k = 1,2,\dots,K \\
\quad \sum_{p=1}^{r}\sum_{i=1}^{m}\sum_{k=1}^{K}\left\{\left(v^p_i + \Phi^{-1}_{\xi_{c^p_{ijk}}}(\rho_j)\right)x^p_{ijk} + \Phi^{-1}_{\xi_{f^p_{ijk}}}(\rho_j)y^p_{ijk}\right\} - \Phi^{-1}_{B_j}\left(1-\rho_j\right) \leq 0, \\
\qquad\qquad\qquad\qquad\qquad\qquad\qquad\qquad j = 1,2,\dots,n \\
\quad x^p_{ijk} \geq 0, \qquad y^p_{ijk} = \begin{cases} 1 & ; if\ x^p_{ijk} > 0 \\ 0 & ; otherwise \end{cases} \qquad \forall\, p,i,j,k.
\end{cases}$$

**Proof**: Considering the CCM of UMMFSTPwB presented in the model (5.3), the corresponding constraints can be written as follow.

(i) The constraint $\mathcal{M}\left\{\sum_{p=1}^{r}\sum_{i=1}^{m}\sum_{j=1}^{n}\sum_{k=1}^{K}\left[\left(s_j^p - v_i^p - \xi_{c_{ijk}^p}\right)x_{ijk}^p - \left(\xi_{f_{ijk}^p}\right)y_{ijk}^p\right] \geq \bar{Z}_1\right\} \geq \alpha_1$ can be rewritten as $\mathcal{M}\{Z_1 \geq \bar{Z}_1\} \geq \alpha_1$, since $\xi_{c_{ijk}^p}$ and $\xi_{f_{ijk}^p}$ are the independent uncertain variables with regular uncertainty distributions $\Phi_{\xi_{c_{ijk}^p}}$ and $\Phi_{\xi_{f_{ijk}^p}}$, respectively. Then according to the Theorem 1.3.8 provided in Section 1.3.11 and Lemma C.1, $\mathcal{M}\{Z_1 \geq \bar{Z}_1\} \geq \alpha_1$ can be reformulated as

$$\sum_{p=1}^{r}\sum_{i=1}^{m}\sum_{j=1}^{n}\sum_{k=1}^{K}\left[\left(s_j^p - v_i^p - \Phi_{c_{ijk}^p}^{-1}(\alpha_1)\right)x_{ijk}^p - \Phi_{f_{ijk}^p}^{-1}(\alpha_1)y_{ijk}^p\right] \geq \bar{Z}_1.$$

In similar way, constraint $\mathcal{M}\left\{\sum_{p=1}^{r}\sum_{i=1}^{m}\sum_{j=1}^{n}\sum_{k=1}^{K}\left[\xi_{t_{ijk}^p}y_{ijk}^p\right] \leq \bar{Z}_2\right\} \geq \alpha_2$ can be restructured as $\sum_{p=1}^{r}\sum_{i=1}^{m}\sum_{j=1}^{n}\sum_{k=1}^{K}\Phi_{t_{ijk}^p}^{-1}(\alpha_2)y_{ijk}^p \leq \bar{Z}_2$.

(ii) Constraint $\mathcal{M}\left\{\sum_{j=1}^{n}\sum_{k=1}^{K}x_{ijk}^p - \xi_{a_i^p} \leq 0\right\} \geq \beta_i^p \Leftrightarrow \mathcal{M}\left\{\xi_{a_i^p} \geq \sum_{j=1}^{n}\sum_{k=1}^{K}x_{ijk}^p\right\} \geq \beta_i^p$. Since, $\Phi_{\xi_{a_i^p}}$ is the uncertainty distribution of $\xi_{a_i^p}$ then from Theorem B.4 (cf. Appendix B) $\mathcal{M}\left\{\xi_{a_i^p} \geq \sum_{j=1}^{n}\sum_{k=1}^{K}x_{ijk}^p\right\} \geq \beta_i^p \Leftrightarrow$
$1 - \Phi_{\xi_{a_i^p}}\left(\sum_{j=1}^{n}\sum_{k=1}^{K}x_{ijk}^p\right) \geq \beta_i^p \Leftrightarrow \sum_{j=1}^{n}\sum_{k=1}^{K}x_{ijk}^p - \Phi_{a_i^p}^{-1}\left(1 - \beta_i^p\right) \leq 0$.

(iii) Constraint $\mathcal{M}\left\{\sum_{i=1}^{m}\sum_{k=1}^{K}x_{ijk}^p - \xi_{b_j^p} \geq 0\right\} \geq \gamma_j^p \Leftrightarrow \mathcal{M}\left\{\xi_{b_j^p} \leq \sum_{i=1}^{n}\sum_{k=1}^{K}x_{ijk}^p\right\} \geq \gamma_j^p$. Since, $\Phi_{\xi_{b_j^p}}$ is the uncertainty distribution of $\xi_{b_j^p}$ then from Theorem B.4, $\mathcal{M}\left\{\xi_{b_j^p} \leq \sum_{i=1}^{m}\sum_{k=1}^{K}x_{ijk}^p\right\} \geq \gamma_j^p \Leftrightarrow \Phi_{\xi_{b_j^p}}\left(\sum_{j=1}^{n}\sum_{k=1}^{K}x_{ijk}^p\right) \geq \gamma_j^p$
$\Leftrightarrow \sum_{j=1}^{n}\sum_{k=1}^{K}x_{ijk}^p - \Phi_{\xi_{b_j^p}}^{-1}\left(\gamma_j^p\right) \geq 0$.

Similarly, constraint $\mathcal{M}\left\{\sum_{p=1}^{r}\sum_{i=1}^{m}\sum_{j=1}^{n}x_{ijk}^p - \xi_{e_k} \leq 0\right\} \geq \delta_k$ can be equivalently transformed into $\sum_{p=1}^{r}\sum_{i=1}^{m}\sum_{j=1}^{n}x_{ijk}^p - \Phi_{e_k}^{-1}(1 - \delta_k) \leq 0$.

(iv) From Theorem B.1 and Theorem B.4, the crisp transformation of constraint $\mathcal{M}\left\{\left\{\sum_{p=1}^{r}\sum_{i=1}^{m}\sum_{k=1}^{K}\left(v_i^p + \xi_{c_{ijk}^p}\right)x_{ijk}^p + \xi_{f_{ijk}^p}y_{ijk}^p\right\} - \xi_{B_j^p} \leq 0\right\} \geq \rho_j$ is equivalently becomes $\left\{\sum_{p=1}^{r}\sum_{i=1}^{m}\sum_{k=1}^{K}\left(v_i^p + \Phi_{c_{ijk}^p}^{-1}(\rho_j)\right)x_{ijk}^p + \Phi_{f_{ijk}^p}^{-1}(\rho_j)\,y_{ijk}^p\right\} - \Phi_{B_j}^{-1}\left(1 - \rho_j\right) \leq 0$.

Therefore, considering (i), (ii), (iii) and (iv) shown above, the crisp equivalent of model (5.3) follows directly, the model (C1).

**Corollary C.1**: If $\xi_{c_{ijk}^p}$, $\xi_{f_{ijk}^p}$, $\xi_{t_{ijk}^p}$, $\xi_{a_i^p}$, $\xi_{b_j^p}$, $\xi_{e_k^p}$ and $\xi_{B_j^p}$ are the independent zigzag uncertain variables of the form $\mathcal{Z}(g,h,l)$ with $g < h < l$. Then, according to Theorem C.1 and the inverse uncertainty distribution of zigzag uncertain variables, we can conclude the following.

(i) For all chance levels <0.5, model (C1) becomes

$$\begin{cases} Max\ \bar{Z}_1 = \sum_{p=1}^{r}\sum_{i=1}^{m}\sum_{j=1}^{n}\sum_{k=1}^{K}\left[\left(s_j^p - v_i^p - \left((1-2\alpha_1)g_{c_{ijk}^p} + 2\alpha_1 h_{c_{ijk}^p}\right)\right)x_{ijk}^p - \left((1-2\alpha_1)g_{f_{ijk}^p} + 2\alpha_1 h_{f_{ijk}^p}\right)y_{ijk}^p\right] \\ Min\ \bar{Z}_2 = \sum_{p=1}^{r}\sum_{i=1}^{m}\sum_{j=1}^{n}\sum_{k=1}^{K}\left[\left((1-2\alpha_2)g_{t_{ijk}^p} + 2\alpha_2 h_{t_{ijk}^p}\right)y_{ijk}^p\right] \\ subject\ to \\ \sum_{j=1}^{n}\sum_{k=1}^{K} x_{ijk}^p - \left(2\beta_i^p h_{a_i^p} + (1-2\beta_i^p) l_{a_i^p}\right) \le 0, i = 1,2,\dots,m, p = 1,2,\dots,r \\ \sum_{i=1}^{m}\sum_{k=1}^{K} x_{ijk}^p - \left((1-2\gamma_j^p) g_{b_j^p} + 2\gamma_j^p h_{b_j^p}\right) \ge 0, j = 1,2,\dots,n, p = 1,2,\dots,r \\ \sum_{p=1}^{r}\sum_{i=1}^{m}\sum_{j=1}^{n} x_{ijk}^p - \left(2\delta_k h_{e_k} + (1-2\delta_k) l_{e_k}\right) \le 0, k = 1,2,\dots,K \\ \sum_{p=1}^{r}\sum_{i=1}^{m}\sum_{k=1}^{K}\left[v_i^p + \left((1-2\rho_j) g_{c_{ijk}^p} + 2\rho_j^p h_{c_{ijk}^p}\right)\right]x_{ijk}^p + \left[\left((1-2\rho_j) g_{f_{ijk}^p} + 2\rho_j h_{f_{ijk}^p}\right)\right]y_{ijk}^p \\ \qquad - \left(2\rho_j h_{B_j} + (1-2\rho_j) l_{B_j}\right) \le 0, j = 1,2,\dots,n \\ x_{ijk}^p \ge 0, \quad y_{ijk}^p = \begin{cases} 1 & ; if\ x_{ijk}^p > 0 \\ 0 & ; otherwise \end{cases} \quad \forall\, p,i,j,k. \end{cases} \tag{C2}$$

(ii) For all chance levels $\ge$ 0.5, model (C1) can be described as given in (C3).

$$\begin{cases} Max\ \bar{Z}_1 = \sum_{p=1}^{r}\sum_{i=1}^{m}\sum_{j=1}^{n}\sum_{k=1}^{K}\left[\left(s_j^p - v_i^p - \left((2-2\alpha_1)h_{c_{ijk}^p} + (2\alpha_1 - 1) l_{c_{ijk}^p}\right)\right)x_{ijk}^p - \left((2-2\alpha_1)h_{f_{ijk}^p} + (2\alpha_1 - 1) l_{f_{ijk}^p}\right)y_{ijk}^p\right] \\ Min\ \bar{Z}_2 = \sum_{p=1}^{r}\sum_{i=1}^{m}\sum_{j=1}^{n}\sum_{k=1}^{K}\left[\left((2-2\alpha_2)h_{t_{ijk}^p} + (2\alpha_2 - 1) l_{t_{ijk}^p}\right)y_{ijk}^p\right] \\ subject\ to \\ \sum_{j=1}^{n}\sum_{k=1}^{K} x_{ijk}^p - \left((2\beta_i^p - 1) g_{a_i^p} + (2-2\beta_i^p) h_{a_i^p}\right) \le 0, i = 1,2,\dots,m, p = 1,2,\dots,r \\ \sum_{i=1}^{m}\sum_{k=1}^{K} x_{ijk}^p - \left((2-2\gamma_j^p) h_{b_j^p} + (2\gamma_j^p - 1) l_{b_j^p}\right) \ge 0, j = 1,2,\dots,n, p = 1,2,\dots,r \\ \sum_{p=1}^{r}\sum_{i=1}^{m}\sum_{j=1}^{n} x_{ijk}^p - \left((2\delta_k - 1) g_{e_k} + (2-2\delta_k) h_{e_k}\right) \le 0, k = 1,2,\dots,K \\ \sum_{p=1}^{r}\sum_{i=1}^{m}\sum_{k=1}^{K}\left[v_i^p + \left((2-2\rho_j) h_{c_{ijk}^p} + (2\rho_j - 1) l_{c_{ijk}^p}\right)\right]x_{ijk}^p + \left[(2-2\rho_j) h_{f_{ijk}^p} + (2\rho_j - 1) l_{f_{ijk}^p}\right]y_{ijk}^p \\ \qquad - \left((2\rho_j - 1) g_{B_j} + (2-2\rho_j) h_{B_j}\right) \le 0, j = 1,2,\dots,n \\ x_{ijk}^p \ge 0, \quad y_{ijk}^p = \begin{cases} 1 & ; if\ x_{ijk}^p > 0 \\ 0 & ; otherwise \end{cases} \quad \forall\, p,i,j,k. \end{cases} \tag{C3}$$

**Corollary C.2**: If $\xi_{c_{ijk}^p}, \xi_{f_{ijk}^p}, \xi_{t_{ijk}^p}, \xi_{a_i^p}, \xi_{b_j^p}, \xi_{e_k}$ and $\xi_{B_j}$ are independent normal uncertain variables of the form $\mathcal{N}(\mu, \sigma)$, such that $\mu, \sigma \in \Re$ and $\sigma > 0$. Then, according to Theorem C.1 and the inverse uncertainty distribution of normal uncertain variables, model $(\mathrm{C1})$ can be written as follows.

$$
\begin{cases}
Max\ \bar{Z}_1 = \sum_{p=1}^{r}\sum_{i=1}^{m}\sum_{j=1}^{n}\sum_{k=1}^{K}\left[\left(s_j^p - v_i^p - \left(\mu_{c_{ijk}^p} + \frac{\sqrt{3}\sigma_{c_{ijk}^p}}{\pi} ln\frac{\alpha_1}{1-\alpha_1}\right)\right)x_{ijk}^p - \left(\mu_{f_{ijk}^p} + \frac{\sqrt{3}\sigma_{f_{ijk}^p}}{\pi} ln\frac{\alpha_1}{1-\alpha_1}\right)y_{ijk}^p\right] \\
Min\ \bar{Z}_2 = \sum_{p=1}^{r}\sum_{i=1}^{m}\sum_{j=1}^{n}\sum_{k=1}^{K}\left[\left(\mu_{t_{ijk}^p} + \frac{\sqrt{3}\sigma_{t_{ijk}^p}}{\pi} ln\frac{\alpha_2}{1-\alpha_2}\right)y_{ijk}^p\right] \\
subject\ to \\
\sum_{j=1}^{n}\sum_{k=1}^{K} x_{ijk}^p - \left(\mu_{a_i^p} - \frac{\sqrt{3}\sigma_{a_i^p}}{\pi} ln\frac{\beta_i^p}{1-\beta_i^p}\right) \le 0, i = 1,2,\dots,m, p = 1,2,\dots,r \\
\sum_{i=1}^{m}\sum_{k=1}^{K} x_{ijk}^p - \left(\mu_{b_j^p} + \frac{\sqrt{3}\sigma_{b_j^p}}{\pi} ln\frac{\gamma_j^p}{1-\gamma_j^p}\right) \ge 0, j = 1,2,\dots,n, p = 1,2,\dots,r \\
\sum_{p=1}^{r}\sum_{i=1}^{m}\sum_{j=1}^{n} x_{ijk}^p - \left(\mu_{e_k} - \frac{\sqrt{3}\sigma_{e_k}}{\pi} ln\frac{\delta_k}{1-\delta_k}\right) \le 0, k = 1,2,\dots,K \\
\sum_{p=1}^{r}\sum_{i=1}^{m}\sum_{k=1}^{K}\left[v_i^p + \mu_{c_{ijk}^p} + \frac{\sqrt{3}\sigma_{c_{ijk}^p}}{\pi} ln\frac{\rho_j}{1-\rho_j}\right]x_{ijk}^p + \left[\mu_{f_{ijk}^p} + \frac{\sqrt{3}\sigma_{f_{ijk}^p}}{\pi} ln\frac{\rho_j}{1-\rho_j}\right]y_{ijk}^p - \left(\mu_{B_j} - \frac{\sqrt{3}\sigma_{B_j}}{\pi} ln\frac{\rho_j}{1-\rho_j}\right) \le 0, \\
\qquad j = 1,2,\dots,n \\
x_{ijk}^p \ge 0, \quad y_{ijk}^p = \begin{cases} 1 & ; if\ x_{ijk}^p > 0 \\ 0 & ; otherwise \end{cases} \quad \forall\, p, i, j, k.
\end{cases}
\tag{C4}
$$

**Crisp Equivalents of Dependent Chance-Constrained Model (DCCM)**

For DCCM, the following model in $(\mathrm{C5})$ is considered as a general case of crisp equivalent for DCCM corresponding to models $(\mathrm{C6})$ and $(\mathrm{C7})$, respectively for zigzag and normal uncertain variables.

$$
\begin{cases}
Max\ v_{Z_1'} \\
Max\ v_{Z_2'} \\
\quad subject\ to\ the\ constraints\ of\ (\mathrm{C1}).
\end{cases}
\tag{C5}
$$

**Theorem C.2**: Let $\xi_{c_{ijk}^p}, \xi_{f_{ijk}^p}, \xi_{t_{ijk}^p}, \xi_{a_i^p}, \xi_{b_j^p}, \xi_{e_k}$ and $\xi_{B_j}$ are the independent zigzag uncertain variables denoted as $\mathcal{Z}(g_c, h_c, l_c)$ with $0.5 \le \eta \le 1$ and $c \in \left\{\xi_{c_{ijk}^p}, \xi_{f_{ijk}^p}, \xi_{t_{ijk}^p}, \xi_{a_i^p}, \xi_{b_j^p}, \xi_{e_k}, \xi_{B_j}\right\}$, where $\eta \in \{\beta_i^p, \gamma_j^p, \delta_k, \rho_j\}$. Then, the crisp equivalent of DCCM, presented in model (5.4), is equivalent to model $(\mathrm{C6})$.

$$
\begin{cases}
Max\ v_{Z_1'} \\
Max\ v_{Z_2'} \\
\quad subject\ to\ the\ constraints\ of\ (\mathrm{C3}),
\end{cases}
\tag{C6}
$$

where

$$\nu_{Z_1'} = \begin{cases} 1 & ; if\ Z_1' \le \bar{g} \\ \frac{2\bar{h}-\bar{g}-Z_1'}{2(\bar{h}-\bar{g})} & ; if\ \bar{g} < Z_1' \le \bar{h} \\ \frac{\bar{l}-Z_1'}{2(\bar{l}-\bar{h})} & ; if\ \bar{h} < Z_1' \le \bar{l} \\ 0 & ; if\ Z_1' > \bar{l} \end{cases} \text{ and } \nu_{Z_2'} = \begin{cases} 0 & ; if\ Z_2' \le \bar{\bar{g}} \\ \frac{Z_2'-\bar{\bar{g}}}{2(\bar{\bar{h}}-\bar{\bar{g}})} & ; if\ \bar{\bar{g}} < Z_2' \le \bar{\bar{h}} \\ \frac{Z_2'+\bar{\bar{l}}-2\bar{\bar{h}}}{2(\bar{\bar{l}}-\bar{\bar{h}})} & ; if\ \bar{\bar{h}} < Z_2' \le \bar{\bar{l}} \\ 1 & ; if\ Z_2' > \bar{\bar{l}}, \end{cases}$$

such that

$$\bar{g} = \sum_{p=1}^{r}\sum_{i=1}^{m}\sum_{j=1}^{n}\sum_{k=1}^{K}\left[\left(s_j^p - v_i^p - l_{\xi_{c_{ijk}^p}}\right)x_{ijk}^p - \left(l_{\xi_{f_{ijk}^p}}\right)y_{ijk}^p\right],$$

$$\bar{h} = \sum_{p=1}^{r}\sum_{i=1}^{m}\sum_{j=1}^{n}\sum_{k=1}^{K}\left[\left(s_j^p - v_i^p - h_{\xi_{c_{ijk}^p}}\right)x_{ijk}^p - \left(h_{\xi_{f_{ijk}^p}}\right)y_{ijk}^p\right],$$

$$\bar{l} = \sum_{p=1}^{r}\sum_{i=1}^{m}\sum_{j=1}^{n}\sum_{k=1}^{K}\left[\left(s_j^p - v_i^p - g_{\xi_{c_{ijk}^p}}\right)x_{ijk}^p - \left(g_{\xi_{f_{ijk}^p}}\right)y_{ijk}^p\right]$$

$$\bar{\bar{g}} = \sum_{p=1}^{r}\sum_{i=1}^{m}\sum_{j=1}^{n}\sum_{k=1}^{K}\left[\left(g_{\xi_{t_{ijk}^p}}\right)y_{ijk}^p\right], \quad \bar{\bar{h}} = \sum_{p=1}^{r}\sum_{i=1}^{m}\sum_{j=1}^{n}\sum_{k=1}^{K}\left[\left(h_{\xi_{t_{ijk}^p}}\right)y_{ijk}^p\right]$$

and $\bar{\bar{l}} = \sum_{p=1}^{r}\sum_{i=1}^{m}\sum_{j=1}^{n}\sum_{k=1}^{K}\left[\left(l_{\xi_{t_{ijk}^p}}\right)y_{ijk}^p\right]$.

**Proof**: Considering the objective $\mathcal{M}\left\{\sum_{p=1}^{r}\sum_{i=1}^{m}\sum_{j=1}^{n}\sum_{k=1}^{K}\left[\left(s_j^p - v_i^p - \xi_{c_{ijk}^p}\right)x_{ijk}^p - \left(\xi_{f_{ijk}^p}\right)y_{ijk}^p\right] \ge Z_1'\right\}$, $\xi_{c_{ijk}^p}$ and $\xi_{f_{ijk}^p}$ are independent zigzag uncertain variables. $x_{ijk}^p$, $s_j^p$ and $v_i^p$ are greater or equal to zero, and $y_{ijk}^p$ are binary variables. Consequently, $\mathcal{M}\left\{\sum_{p=1}^{r}\sum_{i=1}^{m}\sum_{j=1}^{n}\sum_{k=1}^{K}\left[\left(s_j^p - v_i^p - \xi_{c_{ijk}^p}\right)x_{ijk}^p - \left(\xi_{f_{ijk}^p}\right)y_{ijk}^p\right] \ge Z_1'\right\}$ follows zigzag uncertainty distribution and therefore is a zigzag uncertain variable say $\mathcal{Z}(\bar{g}, \bar{h}, \bar{l})$, where $\bar{g}, \bar{h}$ and $\bar{l}$ are given above in (C6). Then, from Definition 1.3.30 (cf. Section 1.3.11), and theorems B.2 and B.4, we write

$$\mathcal{M}\left\{\sum_{p=1}^{r}\sum_{i=1}^{m}\sum_{j=1}^{n}\sum_{k=1}^{K}\left[\left(s_j^p - v_i^p - \xi_{c_{ijk}^p}\right)x_{ijk}^p - \left(\xi_{f_{ijk}^p}\right)y_{ijk}^p\right] \ge Z_1'\right\}$$

$$= 1 - \mathcal{M}\{\mathcal{Z}(\bar{g}, \bar{h}, \bar{l}) \le Z_1'\} = \nu_{Z_1'} = \begin{cases} 1 & ; if\ Z_1' \le \bar{g} \\ \frac{2\bar{h}-\bar{g}-Z_1'}{2(\bar{h}-\bar{g})} & ; if\ \bar{g} < Z_1' \le \bar{h} \\ \frac{\bar{l}-Z_1'}{2(\bar{l}-\bar{h})} & ; if\ \bar{h} < Z_1' \le \bar{l} \\ 0 & ; if\ Z_1' > \bar{l}. \end{cases}$$

Similarly, for the second objective of model (5.4), $\xi_{t_{ijk}^p}$ are independent zigzag uncertain variables. Hence, $\mathcal{M}\left\{\sum_{p=1}^{r}\sum_{i=1}^{m}\sum_{j=1}^{n}\sum_{k=1}^{K}\left[\left(\xi_{t_{ijk}^p}\right)y_{ijk}^p\right] \le Z_2'\right\}$ follows zigzag uncertainty distribution for a zigzag uncertain variable say $\mathcal{Z}(\bar{\bar{g}}, \bar{\bar{h}}, \bar{\bar{l}})$, where $\bar{\bar{g}}, \bar{\bar{h}}$ and $\bar{\bar{l}}$ are defined above in model (C6). So, from Definition 1.3.30 and Theorem B.2 we have

$$\mathcal{M}\left\{\sum_{p=1}^{r}\sum_{i=1}^{m}\sum_{j=1}^{n}\sum_{k=1}^{K}\left[\left(\xi_{t_{ijk}^{p}}\right)y_{ijk}^{p}\right]\leq Z_2'\right\}=\nu_{Z_2'}=\begin{cases}0 & ;if\ Z_2'\leq\bar{\bar{g}}\\ \frac{Z_2'-\bar{\bar{g}}}{2(\bar{\bar{h}}-\bar{\bar{g}})} & ;if\bar{\bar{g}}<Z_2'\leq\bar{\bar{h}}\\ \frac{Z_2'+\bar{\bar{l}}-2\bar{\bar{h}}}{2(\bar{\bar{l}}-\bar{\bar{h}})} & ;if\ \bar{\bar{h}}<Z_2'\leq\bar{\bar{l}}\\ 1 & ;if\ Z_2'>\bar{\bar{l}}.\end{cases}$$

Moreover, from Corollary C.1 (ii) the crisp transformations of the constraint set of model (5.4) become the same to that of the constraint set of model (C3). Hence, it directly follows model (C6).

**Theorem C.3**: Let $\xi_{c_{ijk}^{p}}$, $\xi_{f_{ijk}^{p}}$, $\xi_{t_{ijk}^{p}}$, $\xi_{a_i^p}$, $\xi_{b_j^p}$, $\xi_{e_k^p}$ and $\xi_{B_j^p}$ are the independent normal uncertain variables of the form $\mathcal{N}(\mu_q,\sigma_q)$ with $q\in\left\{\xi_{c_{ijk}^{p}},\xi_{f_{ijk}^{p}},\xi_{t_{ijk}^{p}},\xi_{a_i^p},\xi_{b_j^p},\xi_{e_k^p},\xi_{B_j^p}\right\}$ Then the crisp equivalent of model (5.4) is given in model (C7).

$$\begin{cases}Max\ \nu_{Z_1'}=1-\left(1+exp\left(\frac{\pi(\mu_1-Z_1')}{\sqrt{3}\,\sigma_1}\right)\right)^{-1}\\ Max\ \nu_{Z_2'}=\left(1+exp\left(\frac{\pi(\mu_2-Z_2')}{\sqrt{3}\,\sigma_2}\right)\right)^{-1}\\ subject\ to\ the\ constraints\ of\ (C4).\end{cases}\tag{C7}$$

**Proof**: Considering the first objective of model (5.4), i.e., $\mathcal{M}\left\{\sum_{p=1}^{r}\sum_{i=1}^{m}\sum_{j=1}^{n}\sum_{k=1}^{K}\left[\left(s_j^p-v_i^p-\xi_{c_{ijk}^{p}}\right)x_{ijk}^{p}-\left(\xi_{f_{ijk}^{p}}\right)y_{ijk}^{p}\right]\geq Z_1'\right\}$, $\xi_{c_{ijk}^{p}}$ and $\xi_{f_{ijk}^{p}}$ are independent normal uncertain variables. $x_{ijk}^{p}$, $s_j^p$ and $v_i^p$ are greater or equal to zero, and $y_{ijk}^{p}$ are binary variables. Then $\sum_{p=1}^{r}\sum_{i=1}^{m}\sum_{j=1}^{n}\sum_{k=1}^{K}\left[\left(s_j^p-v_i^p-\xi_{c_{ijk}^{p}}\right)x_{ijk}^{p}-\left(\xi_{f_{ijk}^{p}}\right)y_{ijk}^{p}\right]$ can be considered as a normal uncertain variable $\mathcal{N}(\mu_1,\sigma_1)$, such that $\mu_1$ and $\sigma_1$ are respectively, $\sum_{p=1}^{r}\sum_{i=1}^{m}\sum_{j=1}^{n}\sum_{k=1}^{K}\left[\left(s_j^p-v_i^p-\mu_{\xi_{c_{ijk}^{p}}}\right)x_{ijk}^{p}+\left(\mu_{\xi_{f_{ijk}^{p}}}\right)y_{ijk}^{p}\right]$ and $\sum_{p=1}^{r}\sum_{i=1}^{m}\sum_{j=1}^{n}\sum_{k=1}^{K}\left[\sigma_{\xi_{c_{ijk}^{p}}}x_{ijk}^{p}+\sigma_{\xi_{f_{ijk}^{p}}}y_{ijk}^{p}\right]$.

Therefore, from Definition 1.3.31 (cf. Section 1.3.11), and theorems B.3 and B.4, $\mathcal{M}\left\{\sum_{p=1}^{r}\sum_{i=1}^{m}\sum_{j=1}^{n}\sum_{k=1}^{K}\left[\left(s_j^p-v_i^p-\xi_{c_{ijk}^{p}}\right)x_{ijk}^{p}-\left(\xi_{f_{ijk}^{p}}\right)y_{ijk}^{p}\right]\geq Z_1'\right\}=1-\left(1+exp\left(\frac{\pi(\mu_1-Z_1')}{\sqrt{3}\,\sigma_1}\right)\right)^{-1}$.

Similarly, for the second objective of model (5.4), $\xi_{t_{ijk}^{p}}$ are the independent normal uncertain variables. Therefore, $\mathcal{M}\left\{\sum_{p=1}^{r}\sum_{i=1}^{m}\sum_{j=1}^{n}\sum_{k=1}^{K}\left[\xi_{t_{ijk}^{p}}y_{ijk}^{p}\right]\leq Z_2'\right\}$ is a normal uncertain variable, $\mathcal{N}(\mu_2, \sigma_2)$ such that $\mu_2$ and $\sigma_2$ are respectively, $\sum_{p=1}^{r}\sum_{i=1}^{m}\sum_{j=1}^{n}\sum_{k=1}^{K}\left[\mu_{\xi_{t_{ijk}^{p}}} y_{ijk}^{p}\right]$ and $\sum_{p=1}^{r}\sum_{i=1}^{m}\sum_{j=1}^{n}\sum_{k=1}^{K}\left[\sigma_{\xi_{t_{ijk}^{p}}} y_{ijk}^{p}\right]$. Accordingly, from Definition 1.3.31, and theorems B.3 and B.4, $\mathcal{M}\left\{\sum_{p=1}^{r}\sum_{i=1}^{m}\sum_{j=1}^{n}\sum_{k=1}^{K}\left[\xi_{t_{ijk}^{p}} y_{ijk}^{p}\right] \leq Z_2'\right\} = \left(1 + \exp\left(\frac{\pi(\mu_2 - Z_2')}{\sqrt{3}\,\sigma_2}\right)\right)^{-1}$. Further, from Corollary C.2 the crisp transformations of the constraints of model (5.4) become the same to that of the constraint set of model (C4). Hence, the model (C7) follows directly.

# Appendix D

The input parameters related to UMMFSTPwB are reported below in Table D.1 through Table D.9. The parameters shown in Table D.1 and Table D.2 are crisp. Whereas, the parameters presented in Table D.3-Table D.9 are uncertain. These uncertain parameters are represented by (i) zigzag uncertain variables and (ii) normal uncertain variables.

Table D.1 Unit purchase costs of items 1 and 2 at two different sources

| $i$ | 1 | 2 |
|---|---|---|
| $v_i^1$ | 10 | 8 |
| $v_i^2$ | 8 | 9 |

Table D.2 Unit selling prices of items 1 and 2 at three different destinations

| $j$ | 1 | 2 | 3 |
|---|---|---|---|
| $s_j^1$ | 22 | 17 | 19 |
| $s_j^1$ | 24 | 16 | 20 |

Table D.3 Transportation cost for item 1 and item 2 represented by zigzag and normal uncertain variables

| | Item $p$ | $\xi_{c_{ij1}^p}$ | 1 | 2 | 3 | $\xi_{c_{ij2}^p}$ | 1 | 2 | 3 |
|---|---|---|---|---|---|---|---|---|---|
| Zigzag uncertain variables | 1 | 1 | $\mathcal{Z}(8,10,14)$ | $\mathcal{Z}(5,6,7)$ | $\mathcal{Z}(8,11,13)$ | 1 | $\mathcal{Z}(4,7,10)$ | $\mathcal{Z}(4,6,8)$ | $\mathcal{Z}(6,7,8)$ |
| | | 2 | $\mathcal{Z}(8\ 10,12)$ | $\mathcal{Z}(9,12,13)$ | $\mathcal{Z}(10,11,14)$ | 2 | $\mathcal{Z}(4,5,6)$ | $\mathcal{Z}(4,7,10)$ | $\mathcal{Z}(5,6,7)$ |
| | 2 | 1 | $\mathcal{Z}(7,8,12)$ | $\mathcal{Z}(5,6,7)$ | $\mathcal{Z}(5,8,9)$ | 1 | $\mathcal{Z}(5,7,10)$ | $\mathcal{Z}(3,7,9)$ | $\mathcal{Z}(4,8,11)$ |
| | | 2 | $\mathcal{Z}(7,9,12)$ | $\mathcal{Z}(7,11,15)$ | $\mathcal{Z}(9,12,14)$ | 2 | $\mathcal{Z}(6,7,8)$ | $\mathcal{Z}(4,5,6)$ | $\mathcal{Z}(5,7,10)$ |
| Normal uncertain variables | 1 | 1 | $\mathcal{N}(11,1.23)$ | $\mathcal{N}(7,0.82)$ | $\mathcal{N}(12,0.86)$ | 1 | $\mathcal{N}(7,0.52)$ | $\mathcal{N}(5,0.37)$ | $\mathcal{N}(6,0.49)$ |
| | | 2 | $\mathcal{N}(8,1.02)$ | $\mathcal{N}(9,1.32)$ | $\mathcal{N}(11,2.18)$ | 2 | $\mathcal{N}(5,0.68)$ | $\mathcal{N}(7,0.71)$ | $\mathcal{N}(4,1.04)$ |
| | 2 | 1 | $\mathcal{N}(9,0.59)$ | $\mathcal{N}(6,0.62)$ | $\mathcal{N}(7,0.42)$ | 1 | $\mathcal{N}(5,0.43)$ | $\mathcal{N}(8,0.68)$ | $\mathcal{N}(9,0.49)$ |
| | | 2 | $\mathcal{N}(7,0.39)$ | $\mathcal{N}(10,0.83)$ | $\mathcal{N}(11,1.19)$ | 2 | $\mathcal{N}(6,0.42)$ | $\mathcal{N}(5,0.12)$ | $\mathcal{N}(4,0.18)$ |

Table D.4 Fixed charge cost for item 1 and item 2 represented by zigzag and normal uncertain variables

| | Item $p$ | $\xi_{c_{ij1}^p}$ | 1 | 2 | 3 | $\xi_{c_{ij2}^p}$ | 1 | 2 | 3 |
|---|---|---|---|---|---|---|---|---|---|
| Zigzag uncertain variables | 1 | 1 | $\mathcal{Z}(22,25,28)$ | $\mathcal{Z}(13,14,16)$ | $\mathcal{Z}(22,26,31)$ | **1** | $\mathcal{Z}(15,17,18)$ | $\mathcal{Z}(18,21,23)$ | $\mathcal{Z}(11,13,14)$ |
| | | 2 | $\mathcal{Z}(18,22,24)$ | $\mathcal{Z}(15,18,19)$ | $\mathcal{Z}(20,22,25)$ | **2** | $\mathcal{Z}(10,11,14)$ | $\mathcal{Z}(14,15,20)$ | $\mathcal{Z}(10,11,14)$ |
| | 2 | 1 | $\mathcal{Z}(20,23,24)$ | $\mathcal{Z}(14,1517)$ | $\mathcal{Z}(24,26,29)$ | **1** | $\mathcal{Z}(14,17,19)$ | $\mathcal{Z}(19,20,24)$ | $\mathcal{Z}(12,14,17)$ |
| | | 2 | $\mathcal{Z}(19,22,25)$ | $\mathcal{Z}(21,24,26)$ | $\mathcal{Z}(18,20,23)$ | **2** | $\mathcal{Z}(11,12,14)$ | $\mathcal{Z}(12,13,16)$ | $\mathcal{Z}(15,18,24)$ |
| Normal uncertain variables | 1 | 1 | $N(25,1.28)$ | $N(16,1.72)$ | $N(26,0.87)$ | **1** | $N(16,0.59)$ | $N(22,1.91)$ | $N(13,1.35)$ |
| | | 2 | $N(22,0.85)$ | $N(18,1.69)$ | $N(21,2.07)$ | **2** | $N(12,1.71)$ | $N(15,0.27)$ | $N(12,0.59)$ |
| | 2 | 1 | $N(25,0.95)$ | $N(15,1.25)$ | $N(17,0.50)$ | **1** | $N(16,0.67)$ | $N(20,1.74)$ | $N(14,0.63)$ |
| | | 2 | $\mathcal{N}(22,0.67)$ | $\mathcal{N}(24,1.83)$ | $\mathcal{N}(20,1.26)$ | **2** | $\mathcal{N}(12,0.45)$ | $\mathcal{N}(15,0.31)$ | $\mathcal{N}(12,0.80)$ |

Table D.5 Transportation time for item 1 and item 2 represented by zigzag and normal uncertain variables

| | Item $p$ | $\xi_{c_{ij1}^p}$ | 1 | 2 | 3 | $\xi_{c_{ij2}^p}$ | 1 | 2 | 3 |
|---|---|---|---|---|---|---|---|---|---|
| Zigzag uncertain variables | 1 | 1 | $\mathcal{Z}(12.0,14.0,15.0)$ | $\mathcal{Z}(6.7,8.6.2,11.2)$ | $\mathcal{Z}(9.5,11.2,12.1)$ | 1 | $\mathcal{Z}(8.2,9.4,10.5)$ | $\mathcal{Z}(10.2,12.4,14.7)$ | $\mathcal{Z}(8.4,12.8,14.8)$ |
| | | 2 | $\mathcal{Z}(8.0,9.0,11.0)$ | $\mathcal{Z}(12.2,14.6,16.3)$ | $\mathcal{Z}(11.9,14.4,17.2)$ | 2 | $\mathcal{Z}(6.8,7.8,10.4)$ | $\mathcal{Z}(11.2,12.1,15.8)$ | $\mathcal{Z}(10.2,13.9,14.1)$ |
| | 2 | 1 | $\mathcal{Z}(12.1,15.4,18.7)$ | $\mathcal{Z}(8.1,13.2,14.5)$ | $\mathcal{Z}(10.0,11.8,13.1)$ | 1 | $\mathcal{Z}(10.0,14.1,17.4)$ | $\mathcal{Z}(7.0,7.5,8.5)$ | $\mathcal{Z}(10.2,13.6,14.2)$ |
| | | 2 | $\mathcal{Z}(11.9,13.2,17.4)$ | $\mathcal{Z}(10.0,12.0,14.0)$ | $\mathcal{Z}(10.0,13.7,15.0)$ | 2 | $\mathcal{Z}(9.0,11.0,13.5)$ | $\mathcal{Z}(8.0,10.5,12.0)$ | $\mathcal{Z}(13.0,14.6,16.0)$ |
| Normal uncertain variables | 1 | 1 | $\mathcal{N}(14.0,0.27)$ | $\mathcal{N}(9.2,0.53)$ | $\mathcal{N}(11.7,0.49)$ | 1 | $\mathcal{N}(8.6,1.2)$ | $\mathcal{N}(12.2,0.6)$ | $\mathcal{N}(11.2,1.6)$ |
| | | 2 | $\mathcal{N}(11.6,1.21)$ | $\mathcal{N}(12.0,0.14)$ | $\mathcal{N}(14.4,0.75)$ | 2 | $\mathcal{N}(12.2,1.0)$ | $\mathcal{N}(9.8,0.4)$ | $\mathcal{N}(9.0,1.17)$ |
| | 2 | 1 | $\mathcal{N}(14.4,0.70)$ | $\mathcal{N}(9.0,0.8)$ | $\mathcal{N}(13.2,1.03)$ | 1 | $\mathcal{N}(14.0,1.44)$ | $\mathcal{N}(10.7,0.70)$ | $\mathcal{N}(13.6,1.27)$ |
| | | 2 | $\mathcal{N}(13.2,0.40)$ | $\mathcal{N}(11.6,1.41)$ | $\mathcal{N}(13.2,0.25)$ | 2 | $\mathcal{N}(11.2,0.79)$ | $\mathcal{N}(12.4,1.59)$ | $\mathcal{N}(10.0,0.75)$ |

Table D.6 Available amounts of item 1 and item 2 represented by zigzag and normal uncertain variables

| | $i$ | 1 | 2 | $i$ | 1 | 2 |
|---|---|---|---|---|---|---|
| Zigzag uncertain variables | $\xi_{a_i^1}$ | $\mathcal{Z}(31, 32, 35)$ | $\mathcal{Z}(34, 37, 41)$ | $\xi_{a_i^2}$ | $\mathcal{Z}(25, 26, 27)$ | $\mathcal{Z}(26, 29, 31)$ |
| Normal uncertain variables | $\xi_{a_i^1}$ | $\mathcal{N}(32,1.57)$ | $\mathcal{N}(35,0.87)$ | $\xi_{a_i^2}$ | $\mathcal{N}(25,1.73)$ | $\mathcal{N}(28,0.47)$ |

Table D.7 Demand for item 1 and item 2 represented by zigzag and normal uncertain variables

| | $j$ | 1 | 2 | 3 | $j$ | 1 | 2 | 3 |
|---|---|---|---|---|---|---|---|---|
| Zigzag uncertain variables | $\xi_{b_j^1}$ | $\mathcal{Z}(19,21,24)$ | $\mathcal{Z}(14,17,21)$ | $\mathcal{Z}(17,19,22)$ | $\xi_{b_j^2}$ | $\mathcal{Z}(11,13,15)$ | $\mathcal{Z}(12,15,18)$ | $\mathcal{Z}(14,19,21)$ |
| Normal uncertain variables | $\xi_{b_j^1}$ | $\mathcal{N}(20,0.39)$ | $\mathcal{N}(16,0.76)$ | $\mathcal{N}(18,0.94)$ | $\xi_{b_j^2}$ | $\mathcal{N}(12,0.47)$ | $\mathcal{N}(14,0.83)$ | $\mathcal{N}(15,0.62)$ |

Table D.8 Transportation capacities of two conveyances expressed as zigzag and normal uncertain variables

| | $k$ | 1 | 2 |
|---|---|---|---|
| Zigzag uncertain variables | $\xi_{e_k}$ | $\mathcal{Z}(67,70,73)$ | $\mathcal{Z}(75,80,83)$ |
| Normal uncertain variables | $\xi_{e_k}$ | $\mathcal{N}(70,1.69)$ | $\mathcal{N}(80,1.77)$ |

Table D.9 Budget availability at destinations represented by zigzag and normal uncertain variables

| | $j$ | 1 | 2 | 3 |
|---|---|---|---|---|
| Zigzag uncertain variables | $\xi_{B_j}$ | $\mathcal{Z}(820,825,830)$ | $\mathcal{Z}(800,810,815)$ | $\mathcal{Z}(890,894,900)$ |
| Normal uncertain variables | $\xi_{B_j}$ | $\mathcal{N}(650,7.80)$ | $\mathcal{N}(600,9.04)$ | $\mathcal{N}(640,10.31)$ |

# Appendix E

### Data Tables for Input Parameters

The rough fuzzy input parameters related to the $W$ as shown in Fig. 6.6, are reported below in the following tables.

Table E.1 The rough fuzzy linear weights representing the possible costs to the company while providing the air service between a pair of cities

| Edges | Rough fuzzy linear weights | Edges | Rough fuzzy linear weights |
|---|---|---|---|
| $e_1$ | $[\xi, \xi + 1][\xi - 1, \xi + 2]$ $\xi = (25, 26.5, 27.2)$ | $e_{11}$ | $[\xi, \xi + 0.8][\xi - 1, \xi + 1.2]$ $\xi = (34.2, 36, 37.2)$ |
| $e_2$ | $[\xi, \xi + 2][\xi - 0.7, \xi + 2.5]$ $\xi = (26.1, 28.3, 29.2)$ | $e_{12}$ | $[\xi, \xi + 1][\xi - 1, \xi + 1.5]$ $\xi = (31.2, 32.7, 33.5)$ |
| $e_3$ | $[\xi, \xi + 1.2][\xi - 0.5, \xi + 1.9]$ $\xi = (30.7,32.8, 34)$ | $e_{13}$ | $[\xi, \xi + 0.2][\xi - 0.2, \xi + 1.4]$ $\xi = (36.1,37.2, 37.9)$ |
| $e_4$ | $[\xi, \xi + 1.5][\xi - 2, \xi + 2.1]$ $\xi = (28.4, 29.7, 30.2)$ | $e_{14}$ | $[\xi, \xi + 0.5][\xi - 1.2, \xi + 1.3]$ $\xi = (29.5, 31.4, 32.6)$ |
| $e_5$ | $[\xi, \xi + 1.7][\xi - 1.8, \xi + 2.3]$ $\xi = (26.1, 28.9, 31.1)$ | $e_{15}$ | $[\xi, \xi + 1.2][\xi - 1.3, \xi + 1.4]$ $\xi = (28.4, 29.3, 31.2)$ |
| $e_6$ | $[\xi, \xi + 1][\xi - 0.3, \xi + 1.5]$ $\xi = (31.6, 34.5, 35.2)$ | $e_{16}$ | $[\xi, \xi + 1.2][\xi - 1.2, \xi + 1.7]$ $\xi = (31.1, 33.6, 35)$ |
| $e_7$ | $[\xi, \xi + 0.7][\xi - 1, \xi + 1.6]$ $\xi = (32.4, 33, 34.1)$ | $e_{17}$ | $[\xi, \xi + 1.7][\xi - 1, \xi + 1.9]$ $\xi = (33.2, 35.8, 36.4)$ |
| $e_8$ | $[\xi, \xi + 0.7][\xi - 1, \xi + 1.4]$ $\xi = (28.2, 29, 30.4)$ | $e_{18}$ | $[\xi, \xi + 1.2][\xi - 0.8, \xi + 1.4]$ $\xi = (31.6, 32.7, 34.5)$ |
| $e_9$ | $[\xi, \xi + 0.2][\xi - 1, \xi + 2]$ $\xi = (41.6, 43.5, 45.4)$ | $e_{19}$ | $[\xi, \xi + 1.2][\xi - 1, \xi + 2]$ $\xi = (29.1, 30.9, 32.5)$ |
| $e_{10}$ | $[\xi, \xi + 0.9][\xi - 1.2, \xi + 2]$ $\xi = (36.4, 37.2, 38.4)$ | $--$ | $--$ |

Table E.2 The rough fuzzy quadratic weights of a pair of adjacent edges representing the possible discount on the total flight fare of two consecutive flights

| Edges | Rough fuzzy quadratic weights | Edges | Rough fuzzy quadratic weights |
|---|---|---|---|
| $e_1e_2$ | $[\xi,\xi+0.3][\xi-1,\xi+1.7]$ $\xi=(10.8,11.5,12.9)$ | $e_3e_4$ | $[\xi,\xi+0.6][\xi-0.9,\xi+1.8]$ $\xi=(11.4,12.4,14.6)$ |
| $e_1e_3$ | $[\xi,\xi+0.7][\xi-1.2,\xi+1.2]$ $\xi=(12.1,13.7,14.5)$ | $e_4e_7$ | $[\xi,\xi+0.2][\xi-0.9,\xi+1.3]$ $\xi=(11.5,12.8,14.9)$ |
| $e_2e_3$ | $[\xi,\xi+0.2][\xi-0.9,\xi+1.7]$ $\xi=(11.7,12.8,14.2)$ | $e_4e_9$ | $[\xi,\xi+0.7][\xi-0.8,\xi+1.6]$ $\xi=(11.5,14.8,15.7)$ |
| $e_1e_4$ | $[\xi,\xi+1.2][\xi-1.1,\xi+1.5]$ $\xi=(10.5,11.3,12.2)$ | $e_4e_{10}$ | $[\xi,\xi+0.2][\xi-0.7,\xi+1.7]$ $\xi=(12.7,14.6,15.2)$ |
| $e_1e_5$ | $[\xi,\xi+1.1][\xi-1.5,\xi+2.1]$ $\xi=(10.6,11.5,13.2)$ | $e_7e_9$ | $[\xi,\xi+0.6][\xi-0.9,\xi+1.9]$ $\xi=(14.2,15.6,16.2)$ |
| $e_1e_6$ | $[\xi,\xi+0.3][\xi-1,\xi+1.7]$ $\xi=(11.6,14.5,15.2)$ | $e_7e_{10}$ | $[\xi,\xi+1][\xi-1.2,\xi+1.4]$ $\xi=(13.2,14.6,15.7)$ |
| $e_4e_5$ | $[\xi,\xi+1.1][\xi-1.7,\xi+1.5]$ $\xi=(10.4,11.7,13.1)$ | $e_9e_{10}$ | $[\xi,\xi+1][\xi-1.5,\xi+1.7]$ $\xi=(14.2,15.9,17.2)$ |
| $e_4e_6$ | $[\xi,\xi+0.3][\xi-1,\xi+1.4]$ $\xi=(11.5,12.8,14.4)$ | $e_8e_9$ | $[\xi,\xi+0.2][\xi-0.2,\xi+1.3]$ $\xi=(12.7,14.2,15.9)$ |
| $e_5e_6$ | $[\xi,\xi+0.8][\xi-1,\xi+1.7]$ $\xi=(11.6,12.5,14.8)$ | $e_8e_{11}$ | $[\xi,\xi+0.6][\xi-0.8,\xi+1.7]$ $\xi=(12.7,13.9,15.7)$ |
| $e_2e_7$ | $[\xi,\xi+0.7][\xi-1.4,\xi+1.5]$ $\xi=(10.7,12.7,13.6)$ | $e_8e_{12}$ | $[\xi,\xi+0.4][\xi-0.9,\xi+1.3]$ $\xi=(10.7,12.9,14.6)$ |
| $e_2e_8$ | $[\xi,\xi+0.5][\xi-0.9,\xi+1.2]$ $\xi=(10.5,12.4,14.8)$ | $e_9e_{11}$ | $[\xi,\xi+0.2][\xi-1.3,\xi+1]$ $\xi=(12.7,14.9,16.9)$ |
| $e_7e_8$ | $[\xi,\xi+0.2][\xi-0.8,\xi+1.3]$ $\xi=(12.5,14.4,15.1)$ | $e_9e_{12}$ | $[\xi,\xi+1.5][\xi-0.7,\xi+1.7]$ $\xi=(14.2,15.8,17.4)$ |
| $e_3e_7$ | $[\xi,\xi+0.7][\xi-0.2,\xi+1.6]$ $\xi=(12.1,13.4,14.3)$ | $e_{11}e_{12}$ | $[\xi,\xi+0.5][\xi-0.8,\xi+1.2]$ $\xi=(12.6,13.7,14.5)$ |
| $e_3e_9$ | $[\xi,\xi+0.6][\xi-0.5,\xi+1.7]$ $\xi=(13.1,15.4,16.7)$ | $e_5e_{10}$ | $[\xi,\xi+1.1][\xi-1.6,\xi+1.8]$ $\xi=(12.1,13.9,14.9)$ |
| $e_3e_{10}$ | $[\xi,\xi+0.3][\xi-0.8,\xi+1.6]$ $\xi=(12.1,14.4,15.8)$ | $e_5e_{11}$ | $[\xi,\xi+0.1][\xi-0.6,\xi+0.9]$ $\xi=(10.1,12.7,14.1)$ |

Table E.3 Continuation of Table E.2

| Edges | Rough fuzzy quadratic weights | Edges | Rough fuzzy quadratic weights |
|---|---|---|---|
| $e_5e_{13}$ | $[\xi,\xi+0.6][\xi-1.2,\xi+1.6]$ $\xi=(12.3,13.7,14.8)$ | $e_{12}e_{17}$ | $[\xi,\xi+0.6][\xi-1.2,\xi+1.5]$ $\xi=(11.9,14.2,15.7)$ |
| $e_5e_{14}$ | $[\xi,\xi+0.4][\xi-1.3,\xi+1.7]$ $\xi=(10.4,12.6,13.7)$ | $e_{13}e_{16}$ | $[\xi,\xi+1.2][\xi-1.4,\xi+1.9]$ $\xi=(12.4,14.4,15.9)$ |
| $e_5e_{15}$ | $[\xi,\xi+0.5][\xi-1.7,\xi+1.8]$ $\xi=(11.4,13.6,14.5)$ | $e_{13}e_{17}$ | $[\xi,\xi+0.6][\xi-1.2,\xi+1.7]$ $\xi=(12.5,13.8,14.9)$ |
| $e_{10}e_{11}$ | $[\xi,\xi+0.4][\xi-1.3,\xi+1.6]$ $\xi=(12.4,14.7,15.7)$ | $e_{16}e_{17}$ | $[\xi,\xi+0.2][\xi-1,\xi+1.3]$ $\xi=(12.6,13.9,15.2)$ |
| $e_{10}e_{13}$ | $[\xi,\xi+1.2][\xi-1.1,\xi+1.9]$ $\xi=(11.4,14.9,16.7)$ | $e_{14}e_{16}$ | $[\xi,\xi+0.9][\xi-1.2,\xi+1.4]$ $\xi=(10.3,12.8,14.6)$ |
| $e_{10}e_{14}$ | $[\xi,\xi+0.7][\xi-1.6,\xi+1.9]$ $\xi=(10.7,13.7,15.6)$ | $e_{14}e_{18}$ | $[\xi,\xi+0.5][\xi-0.1,\xi+1.3]$ $\xi=(11.5,12.6,15.8)$ |
| $e_{10}e_{15}$ | $[\xi,\xi+0.7][\xi-1.9,\xi+2.2]$ $\xi=(11.7,13.9,15.3)$ | $e_{14}e_{19}$ | $[\xi,\xi+0.3][\xi-0.7,\xi+0.9]$ $\xi=(11.5,13.4,14.8)$ |
| $e_{11}e_{13}$ | $[\xi,\xi+0.4][\xi-1.3,\xi+1.4]$ $\xi=(12.5,14.6,15.8)$ | $e_{16}e_{18}$ | $[\xi,\xi+0.5][\xi-0.7,\xi+1.5]$ $\xi=(10.1,12.9,14.4)$ |
| $e_{11}e_{14}$ | $[\xi,\xi+1.4][\xi-0.3,\xi+1.8]$ $\xi=(11.2,13.9,15.2)$ | $e_{16}e_{19}$ | $[\xi,\xi+0.7][\xi-0.2,\xi+1.4]$ $\xi=(11.7,14.4,15.8)$ |
| $e_{11}e_{15}$ | $[\xi,\xi+0.7][\xi-1.3,\xi+1.8]$ $\xi=(10.2,13.1,14.4)$ | $e_{18}e_{19}$ | $[\xi,\xi+0.5][\xi-0.8,\xi+1.4]$ $\xi=(11.5,12.8,14.6)$ |
| $e_{13}e_{14}$ | $[\xi,\xi+0.3][\xi-1,\xi+1.7]$ $\xi=(11.2,13.9,14.7)$ | $e_6e_{15}$ | $[\xi,\xi+0.8][\xi-0.9,\xi+1.6]$ $\xi=(11.6,12.7,15.1)$ |
| $e_{13}e_{15}$ | $[\xi,\xi+0.5][\xi-1.1,\xi+1.6]$ $\xi=(12.6,14.1,14.9)$ | $e_6e_{19}$ | $[\xi,\xi+0.7][\xi-0.3,\xi+1.7]$ $\xi=(12.1,14.8,15.6)$ |
| $e_{14}e_{15}$ | $[\xi,\xi+0.7][\xi-0.3,\xi+1.6]$ $\xi=(10.2,12.6,14.8)$ | $e_{15}e_{19}$ | $[\xi,\xi+0.3][\xi-1.8,\xi+2]$ $\xi=(10.5,11.8,14.2)$ |
| $e_{12}e_{13}$ | $[\xi,\xi+0.6][\xi-1.5,\xi+1.8]$ $\xi=(12.5,14.3,15.9)$ | $e_{17}e_{18}$ | $[\xi,\xi+0.6][\xi-0.2,\xi+1.4]$ $\xi=(12.7,14.7,15.9)$ |
| $e_{12}e_{16}$ | $[\xi,\xi+0.2][\xi-1.5,\xi+1.6]$ $\xi=(10.6,11.5,13.2)$ | $--$ | $--$ |

# References

Adlakha V., Kowalski K. (2003) A simple heuristic for solving small fixed-charge transportation problem. *Omega* 31 (3): 205-211.

Agarwal P.K., Edelsbrunner H., Schwarzkopf O., Welzl E. (1991) Euclidean minimum spanning trees and bi-chromatic closest pairs. *Discrete & Computational Geometry* 6 (3): 407–422.

Aggarwal S., Gupta C. (2016) Solving intuitionistic fuzzy solid transportation problem via new ranking method based on signed distance. *International Journal of Uncertainty, Fuzziness and Knowledge-Based Systems* 24 (4): 483-501.

Ahuja R.K., Orlin J.B. (1989) A fast and simple algorithm for the maximum flow problem. *Operations Research* 37 (5): 748-759.

Ahuja R.K., Magnanti T.L., Orlin J.B. (1993) *Network flows: Theory, algorithms, and applications*. Prentice-Hall, Upper Saddle River, NJ.

Alefeld G., Herzberger J. (1983) *Introduction to interval computations*. Academic Press, New York.

Aliev R. A., Pedrycz W., Guirimov B., Aliev R. R., Ilhan U., Babagil M., Mammadli S. (2011) Type-2 fuzzy neural networks with fuzzy clustering and differential evolution optimization. *Information Sciences* 181 (9): 1591–1608.

Ammar E.E., Youness E.A. (2005) Study on multi-objective transportation problem with fuzzy numbers. *Applied Mathematics and Computation* 166 (2): 241-253.

Andersen K.A., Jörnsten K., Lind M. (1996) On bi-criterion minimal spanning trees: An approximation. *Computers & Operations. Research* 23 (12): 1171–1182.

Aneja Y.P., Nair K.P.K. (1979) Bi-criteria transportation problem. *Management Science* 25 (1): 73-78.

Anusuya V., Sathya R. (2014) A new approach for solving type-2 fuzzy shortest path problem. *Annals of Pure and Applied Mathematics* 8 (1): 83-92.

Appa G.M. (1973) The transportation problem and its variants. *Journal of the Operational Research Society* 24 (1): 79-99.

Arsham H., Kahn A.B. (1989) A simplex-type algorithm for general transportation problems: An alternative to stepping-stone. *Journal of the Operational Research Society* 40 (6): 581-590.

Assad A., Xu W. (1992) The quadratic minimum spanning tree problem. *Naval Research Logistics* 39 (3): 399-417.

Altassan K.M., EI-Sherbiny M.M., Abid A.D. (2014) Artificial immune algorithm for solving fixed charge transportation problem. *Applied Mathematics Information Sciences* 8 (2): 751-759.

Baidya A., Bera U.K. (2014) An interval valued solid transportation problem with budget constraint in different interval approaches. *Journal of Transportation Security* 7 (2): 147–155.

Barr R.S., Glover F., Klingman D. (1981) A new optimization method for large scale fixed charge transportation problems. *Operations Research* 29 (3): 448–463.

Bellman R. (1958) On a routing problem. *Quarterly of Applied Mathematics* 16 (1): 87-90.

Bellman R.E., Zadeh L.A. (1970) Decision making in a fuzzy environment. *Management Science* 17 (4): 141-164.

Bhatia H.L. (1981) Indefinite quadratic solid transportation problem. *Journal of Information Optimization Sciences* 2 (3): 297-303.

Bhatia H.L., Swarup K., Puri M.C. (1976) Time minimizing solid transportation problem. *Mathematische Operationsforschung Statistik* 7 (3): 395-403.

Bilgen B. (2010) Application of fuzzy mathematical programming approach to the production allocation and distribution supply chain network problem. *Expert Systems with Applications* 37 (6): 4488–4495.

Bit A.K., Biswal M.P., Alam S.S. (1993) Fuzzy programming approach to multi-objective solid transportation problem. *Fuzzy Sets and Systems* 57 (2): 183–194.

Borüvka O. (1926) O jistém problem minimálním. *Práce Moravské Přírodovědecké Společnosti* III (3): 37-58.

Buson E., Roberti R., Toth P. (2014) A reduced-cost iterated local search heuristic for the fixed-charge transportation problem. *Operations Research* 62 (5): 1095-1106.

Cabot A.V., Erenguc S.S. (1984) Some branch and bound procedures for fixed cost transportation problems. *Naval Research Logistics Quarterly* 31 (1): 145–154.

Cabot A.V., Erenguc S.S. (1986) Improved penalties for fixed cost linear programs using Lagrangean relaxation. *Management Science* 32 (1): 856–869.

Carey M., Henrickson C. (1986) Bounds on expected performance of networks with links subject to failure. *Networks* 14 (3): 439-456.

Cascone A., Marigo A., Piccoli B., Rarità L. (2010) Decentralized optimal routing for packets flow on data networks. *Discrete and Continuous Dynamical Systems - Series B (DCDS-B)* 13 (1): 59-78.

Chakraborty D., Jana D.K., Roy T.K. (2014) A new approach to solve intuitionistic fuzzy optimization problem using possibility, necessity, and credibility measures. *International Journal of Engineering Mathematics* 2014, Article ID 593185, 12 pages doi: 10.1155/2014/593185.

Chanas S., Kołodziejczyk W. (1982) Maximum flow in a network with fuzzy arc capacities. *Fuzzy Sets and Systems* 8 (2): 165-173.

Chanas S., Kołodziejczyk W. (1984) Real-valued flows in a network with fuzzy arc capacities. *Fuzzy Sets and Systems* 13 (2): 139-151.

Chanas S., Kołodziejczyk W. (1986) Integer flows in network with fuzzy capacity constraints. *Networks* 16 (1): 17-31.

Chanas S., Kołodziejczyk W., Machaj A. (1984) A fuzzy approach to the transportation problem. *Fuzzy Sets and Systems* 13 (3): 211-221.

Chanas S., Kuchta D. (1996) A concept of the optimal solution of the transportation problem with fuzzy cost coefficients. *Fuzzy Sets and Systems* 82 (3): 299-305.

Chandy K.M., Lo T. (1973) The capacitated minimum spanning tree. *Networks* 3: 173-182.

Chang P.-T., Lee E.S. (1994) Ranking fuzzy sets based on the concept of existence. *Computers & Mathematics with Applications* 27 (9-10): 1-21.

Chang P.-T., Lee E. S. (1999) Fuzzy decision networks and deconvolution. *Computers & Mathematics with Applications* 37 (11): 53–63.

Charnes A., Cooper W.W. (1959) Chance-constrained programming. *Management Science* 6 (1): 73–79.

Chen B. Y., Lam W. H. K., Sumalee A., Li, Z. (2012) Reliable shortest path finding in stochastic networks with spatial correlated link travel times. *International Journal of Geographical information Science* 26 (2): 365–386.

Chen L., Peng J., Zhang B. (2017) Uncertain goal programming models for bi-criteria solid transportation problem. *Applied Soft Computing* 51: 49-59.

Chen L., Peng J., Zhang B., Rosyida I. (2016) Diversified models for portfolio selection based on uncertain semivariance. *International Journal of Systems Science* 48 (3): 637-648.

Chen P., Nie Y. (M.) (2013) Bi-criterion shortest path problem with a general nonadditive cost. *Transportation Research Part B: Methodological* 57: 419–435.

Chen S.-M., Lee .L-W. (2010) Fuzzy multiple criteria hierarchical group decision making based on interval type-2 fuzzy sets. *IEEE Transactions on Systems, Man, and Cybernetics - Part A: Systems and Humans* 40 (5): 1120–1128.

Chen T.-Y. (2013) An interactive method for multiple criteria group decision analysis based on interval type-2 fuzzy sets and its application to medical decision making. *Fuzzy Optimization and Decision Making* 12 (3): 323–356.

Chen T.-Y. (2014) A PROMETHEE-based outranking method for multiple criteria decision analysis with interval type-2 fuzzy sets. *Soft Computing* 18 (5): 923-940.

Cheng H., Huang W., Cai J. (2013) Solving a fully fuzzy linear programming problem through compromise programming. *Journal of Applied Mathematics* 2013, Article ID 726296, 10 pages doi: 10.1155/2013/726296.

Cheng Y. (2015) Forward approximation and backward approximation in fuzzy rough sets. *Neurocomputing* 148: 340-353

Clímaco J.C.N., Pascoal M.M.B. (2012) Multi-criteria path and tree problems: discussion on exact algorithms and applications. *International Transactions in Operational Research* 19 (1–2): 63–98.

Cordone R., Passeri G. (2012) Solving the minimum quadratic spanning tree problem. *Applied Mathematics and Computation* 218 (23): 11597-11612.

Corley H.W., Moon I.D. (1985) Shortest paths in networks with vector weights. *Journal of Optimization Theory and Applications* 46 (1): 79–86.

Cui Q., Sheng Y. (2013) Uncertain programming model for solid transportation problem. *Information* 15 (3): 342-348.

Ćustić A., Zhang R., Punnen A.P. (2018) The quadratic minimum spanning tree problem and its variations. *Discrete Optimization* 27: 73- 87.

Cutolo A., Nicola C.D., Manzo R., Rarità L. (2012) Optimal paths on urban networks using travelling times prevision. *Modelling and Simulation in Engineering* 2012, Article ID: 564168, 9 pages.

Dalman H. (2016) Uncertain programming model for multi-item solid transportation problem. *International Journal of Machine Learning and Cybernetics* doi: 10.1007/s13042-016-0538-7.

Dantzig G.B. (1951) Application of the simplex method to a transportation problem. In: *Activity Analysis of Production and Allocation* (T.C. Koopmans, ed.). John Wiley & Sons, New York, pp. 359-373.

Darmann A., Pferschy U., Schauer J., Woeginger G.J. (2011) Paths, trees and matchings under disjunctive constraints. *Discrete Applied Mathematics* 159 (16): 1726-1735.

Das A., Bera U.K., Maiti M. (2016) A profit maximizing solid transportation model under rough interval approach. *IEEE Transactions on Fuzzy Systems* 25 (3): 485-498.

Das A., Bera U.K., Maiti M. (2017) Defuzzification and application of trapezoidal type-2 fuzzy variables to green solid transportation problem. *Soft Computing*, doi: 10.1007/s00500-017-2491-0.

Das S.K., Goswami A., Alam S.S. (1999) Multi-objective transportation problem with interval cost, source and destination parameters. *European Journal of Operational Research* 117 (1): 100-112.

De Almeida T.A., Yamakami A., Takahashi M. T. (2005a) An evolutionary approach to solve minimum spanning tree problem with fuzzy parameters. In: *Proceedings of the International Conference on Computational Intelligence for Modelling, Control and Automation and International Conference on Intelligent Agents, Web Technologies and Internet Commerce, CIMCA-IAWTIC'06*, Vienna, pp. 203-208.

De Almeida T.A., Prado F.M.S., Souza V.N., Yamakami A., Takahashi M.T.(2005b) A genetic algorithm to solve minimum spanning tree problem with fuzzy parameters using possibility measure. In: *Proceedings of the North American Fuzzy Information Processing Society*, *NAFIPS 2005*, Detroit, United States, pp. 627-632.

Deb K., Pratap A., Agarwal S., Meyarivan T. (2002) A fast and elitist multi-objective genetic algorithm: NSGA-II. *IEEE Transactions on Evolutionary Computation* 6 (2): 182- 197.

Dembczyński K., Greco S., Słowiński R. (2009) Rough set approach to multiple criteria classification with imprecise evaluations and assignments. *European Journal of Operational Research* 198 (2): 626-636.

Dhamdhere, K., Ravi R., Singh M. (2005) On two-stage stochastic minimum spanning trees. In: *Integer Programming and Combinatorial Optimization*, *IPCO 2005* (M. Jünger and V. Kaibel, eds.). Lecture Notes in Computer Science, Springer, Berlin, Heidelberg, vol. 3509, pp. 321–334.

Deng Y., Chen Y., Zhang Y., Mahadevan S. (2012) Fuzzy Dijkstra algorithm for shortest path problem under uncertain environment. *Applied Soft Computing* 12 (3): 1231-1237.

Dijkstra E.W. (1959) A note on two problems in connexion with graphs. *Numerische Mathematlk* 1: 269-271.

Diker M. (2018) Textures and fuzzy unit operations in rough set theory: An approach to fuzzy rough set models. *Fuzzy Sets and Systems* 336: 27-53

Ding S. (2015) The $\alpha$-maximum flow model with uncertain capacities. *Applied Mathematical Modelling* 39 (7): 2056-2063.

Dinic E.A. (1970) Algorithm for solution of a problem of maximum flow in networks with power estimation. *Soviet Math Doklady* 11 (5): 1277-1280.

Di Puglia Pugliese L., Guerriero F., Santos J.L. (2015) Dynamic programming for spanning tree problems: application to the multi-objective case. *Optimization Letters* 9 (3): 437–450.

Dou Y., Zhu L., Wang H.S. (2012) Solving the fuzzy shortest path problem using multi-criteria decision method based on vague similarity measure. *Applied Soft Computing* 12 (6): 1621-1631.

Doulliez P. (1971) Probability distribution function for the capacity of a multi-terminal network. *RAIRO - Operations Research* 5 (V1): 39-49.

Dreyfus S.E. (1969) An appraisal of some shortest-path algorithms. *Operations Research* 17 (3): 395-412.

Dubois D., Prade H. (1980) *Fuzzy sets and systems: Theory and applications*. Academic Press, New York.

Dubois D., Prade H. (1990) Rough fuzzy sets and fuzzy rough sets. *International Journal of General Systems* 17 (2-3): 191–208.

Durillo J.J., Nebro A.J. (2011) jMetal: a java framework for multi-objective optimization. *Advances in Engineering Software* 42 (10): 760-771.

Dursun P. Bozdağ E. (2014) Chance-constrained programming models for constrained shortest path problem with fuzzy parameters. *Journal of Multiple-Valued Logic and Soft Computing* 22 (4-6): 599-618.

Ebrahimnejad A., Karimnejad Z., Alrezaamiri H. (2015) Particle swarm optimization algorithm for solving shortest path problems with mixed fuzzy arc weights. *International Journal of Applied Decision Sciences* 8 (2): 203-222.

Eckenrode R.T. (1965) Weighting multiple criteria. *Management Science* 12 (3): 180-192.

Edmonds J., Karp R.M. (1972) Theoretical improvements in algorithmic efficiency for network flow problems. *Journal of the Association for Computing Machinery* 19 (2): 248-264.

Ehrgott M. (2005) *Multi-criteria optimization*. Springer-Verlag, Berlin Heidelberg.

Enke D., Mehdiyev N. (2013) Type-2 fuzzy clustering and a type-2 fuzzy inference neural network for the prediction of short-term interest rates. *Procedia Computer Science* 20: 115-120.

Ericsson M., Resende M.G.C., Pardalos P.M. (2001) A genetic algorithm for the weight setting problem in OSPF routing. *Journal of Combinatorial Optimization* 6 (3): 299–333.

Eshelman L.J. (1991) The CHC adaptive search algorithm: How to have safe search when engaging in in nontraditional genetic recombination. In: *Foundations of Genetic Algorithms* (G.J.E. Rawlins, ed.). Elsevier, vol. 1, pp. 265-283.

Euler L. (1736) Solutio problematis ad geometriam situs pertinentis. *Commentarii academiae scientiarum Petropolitanae* 8: 128-140.

Evans J.R. (1976) Maximum flow in probabilistic graphs-the discrete case. *Networks* 6 (2): 161:183.

Feng L., Wang G.-Y., Li X.-X. (2010) Knowledge acquisition in vague objective information systems based on rough sets. *Expert Systems* 27 (2): 129-142

Figueroa-García J.C (2012) A general model for linear programming with interval type-2 fuzzy technological coefficients. In: *Proceedings of the IEEE Annual Meeting of the North American Fuzzy Information Processing Society*, *NAFIPS '12*, Berkeley, pp. 1–4.

Figueroa-García J.C., Hernández G. (2014) A method for solving linear programming models with interval type-2 fuzzy constraints. *Pesquisa Operacional* 34 (1): 73–89.

Fishman G.S. (1987) The distribution of maximum flow with applications to multistate reliability systems. *Operations Research* 35 (4): 607-618.

Floyd R. W. (1962) Algorithm 97: Shortest path. *Communications of the Association for Computing Machinery* 5 (6): 345.

Ford L.R., Fulkerson D.R. (1956) Maximal flow through a network. *Canadian Journal of Mathematics* 8 (3): 399-404.

Ford L.R., Fulkerson D.R. (1962) *Flows in networks*. Princeton University Press, Princeton, NJ.

Frank H., Hakimi S.L. (1965) Probabilistic flows through a communication network. *IEEE Transactions on Circuit Theory* 12 (3): 413-414.

Frieze A. M. (1985) On the value of a random minimum spanning tree problem. *Discrete Applied Mathematics* 10 (1): 47–56.

Fu L., Rilett L.R. (1998) Expected shortest paths in dynamic and stochastic traffic networks. *Transportation Research Part B: Methodological* 32 (7): 499-516.

Fulkerson D.R., Dantzig G.B. (1955) Computations of maximum flow in networks. *Naval Research Logistics Quarterly* 2 (4): 277-283.

Gabow H.N. (1977) Two algorithms for generating weighted spanning trees in order. *SIAM Journal on Computing* 6 (1): 139–150.

Gabow H.N. (1985) Scaling algorithms for network problems. *Journal of Computer and System Sciences* 31 (2): 148-168.

Gandibleux X, Beugnies F, Randriamasy S (2006) Martins' algorithm revisited for multi-objective shortest path problems with a MaxMin cost function. *4OR* 4 (1): 47–59.

Gao Y. (2011) Shortest path problem with uncertain arc lengths. *Computers & Mathematics with applications* 62 (6): 2591-2600.

Gao J., Lu M. (2005) Fuzzy quadratic minimum spanning tree problem. *Applied Mathematics and Computation* 164 (3): 773-788.

Gao S.P., Liu S.Y. (2004) Two-phase fuzzy algorithms for multi-objective transportation problem. *Journal of Fuzzy Mathematics* 12 (1): 147-156.

Gao J., Yang X., Liu D. (2017) Uncertain Shapley value of coalitional game with application to supply chain alliance. *Applied Soft Computing* 56: 551-556.

Gao J., Yao K. (2015) Some concepts and theorems of uncertain random process. *International Journal of Intelligent Systems* 30 (1): 52-65.

Gao W., Zhang Q., Lu Z., Wu D., Du X. (2018) Modelling and application of fuzzy adaptive minimum spanning tree in tourism agglomeration area division. *Knowledge-Based Systems* 143: 317-326

Gao X., Jia L. (2017) Degree constrained minimum spanning tree problem with uncertain edge weights. *Applied Soft Computing* 56: 580-588.

Gao X., Jia L., Kar S. (2017) Degree constrained minimum spanning tree problem of uncertain random network. *Journal of Ambient Intelligence and Humanized Computing* 8 (5): 747-757.

Gao Y., Kar S. (2017) Uncertain solid transportation problem with product blending. *International Journal of Fuzzy Systems* https://doi.org/10.1007/s40815-016-0282-x.

Garey M.R., Johnson D.S. (1990) *Computers and intractability: A guide to the theory of NP-completeness*. W.H. Freeman & Co., New York, NY, USA.

Gass S.I. (1990) On solving the transportation problem. *Journal of the Operational Research Society* 41 (4): 291-297.

Gembicki F.W. (1974) *Vector optimization for control with performance and parameter sensitivity indices*. Ph.D. Dissertation, Case Western Reserve University, Cleveland, OH.

Gen M., Altiparamk F., Lin L. (2006) A genetic algorithm for two-stage transportation problem using priority-based encoding. *OR Spectrum* 28 (3): 337-354.

Gen M., Cheng R. (1997) *Genetic algorithms and engineering design*, 1st ed. Wiley, New York.

Gen M., Cheng R., Lin L. (2008) *Network models and optimization*, 1st ed. Springer-Verlag, London.

Gen M., Cheng R., Wang D. (1997) Genetic algorithms for solving shortest path problems. In: *Proceedings of the IEEE International Conference on Evolutionary Computation, ICEC'97*, Indianapolis, IN, USA, pp. 488–493.

Gen M., Ida K., Li Y., Kubota E. (1995) Solving bi-criteria solid transportation problem with fuzzy numbers by a genetic algorithm. *Computers and Industrial Engineering* 29 (1-4): 537-541.

Giri P.K., Maity M.K., Maiti M. (2015) Fully fuzzy fixed charge multi-item solid transportation problem. *Applied Soft Computing* 27: 77–91.

Goldberg A.V., Tarjan R.E. (1986) A new approach to the maximum flow problem. In: *Proceedings of the 18th Annual ACM Symposium on the Theory of Computing, STOC'86*, Berkeley, pp. 136–146.

Graham R.L., Hell P. (1985) On the history of the minimum spanning tree problem. *IEEE Annals of the History of Computing* 7 (1): 43–57.

Gu T., Xu Z. (2007) The symbolic algorithms for maximum flow in networks. *Computers & Operations Research* 34 (3): 799-816.

Guo C., Gao J. (2017) Optimal dealer pricing under transaction uncertainty. *Journal of Intelligent Manufacturing* 28 (3): 657-665.

Guo H., Wang X., Zhou S. (2015) A transportation problem with uncertain costs and random supplies. *International Journal of e-Navigation and Maritime Economy* 2: 1-11.

Gupta A., Warburton A. (1987) Approximation methods for multiple criteria travelling salesman problems. In: *Toward Interactive and Intelligent Decision Support Systems* (Y. Sawaragi, K. Inoue, H. Nakayama, eds.). Lecture Notes in Economics and Mathematical Systems. Springer, Berlin, vol. 285, pp. 211-217.

Hachtel G.D., Somenzi F. (1997) A symbolic algorithm for maximum flow in 0-1 networks. *Formal Methods in System Design* 10 (2-3): 207–219.

Hadley G., Whitin T.M. (1963) *Analysis of inventory systems*. Prentice Hall.

Haley K.B. (1962) The solid transportation problem. *Operational Research* 10 (4): 448-463.

Han L., Wang Y. (2005) A novel genetic algorithm for multi-criteria minimum spanning tree problem. In: *Computational Intelligence and Security, CIS 2005* (Y. Hao, J. Liu, Y. Wang, Y.-M. Cheung, H. Yin, L. Jiao, J. Ma, Y.-C. Jiao, eds.). Lecture Notes in Computer Science, Springer Berlin Heidelberg, Berlin, Heidelberg, vol. 3801, pp. 297-302.

Han S., Peng Z., Wang S. (2014) The maximum flow problem of uncertain network. *Information Sciences* 265: 167-175.

Haimes Y.Y., Lasdon L.S., Wismer D.A. (1971) On a bi-criterion formulation of the problems of integrated system identification and system optimization. *IEEE Transaction on Systems, Man, and Cybernetics* 1 (3): 296-297.

Hall R. (1986) The fastest path through a network with random time-dependent travel time. *Transportation Science* 20 (3): 182-188.

Hansen E. (1992) *Global optimization using interval analysis*. Marcel Dekker, New York.

Hansen P. (1980) Bi-criterion path problems. In: *Multiple Criteria Decision Making Theory and Application* (G. Fandel, T. Gal, eds.). Lecture notes in Economics and Mathematical Systems, Springer, Berlin Heidelberg, vol. 177, pp. 109–127.

Harris R. (1998) Introduction to decision making. *VirtualSalt*, http://www.virtualsalt.com/crebook5.htm.

Harris T. E., Ross F. S. (1955). *Fundamentals of a method for evaluating rail net capacities*. Research Memorandum, AD Number: AD093458, Rand Corporation.

Hassanzadeh R., Mahdavi I., Mahdavi-Amiri N., Tajdin A. (2013) A genetic algorithm for solving fuzzy shortest path problems with mixed fuzzy arc lengths. *Mathematical and Computer Modelling* 57 (1–2): 84-99.

Hasuike T. (2010) Fuzzy random shortest path problem using conditional Value at Risk. In: *Proceedings of the International Conference on System Science and Engineering, ICSSE 2010*, Taipei, Taiwan, pp. 439-444.

Haugland D., Eleyat M., Hetland, M.L. (2011). The maximum flow problem with minimum lot sizes. In: *International Conference on Computational Logistics, ICCL 2011* (J.W. Böse, H. Hu, C. Jahn, X. Shi, R. Stahlbock, S. Voß, eds.). Lecture Notes in Computer Science, Hamburg, Germany, vol. 6971, pp. 170–182.

Henig M.I. (1986) The shortest path problem with two objective functions. *European Journal of Operational Research* 25 (2): 281-291.

Hernandes F., Lamata M.T., Verdegay J.L., Yamakami A. (2007a) The shortest path problem on networks with fuzzy parameters. *Fuzzy Sets and Systems* 158 (14): 1561-1570.

Hernandes F., Lamata M.T., Takahashi M.T. (2007b) An Algorithm for the fuzzy maximum flow problem. In: *IEEE International Fuzzy Systems Conference, FUZZ-IEEE 2007*, London, pp. 1-6.

Hirano S., Tsumoto S. (2005) Rough representation of a region of interest in medical images. *International Journal of Approximate Reasoning* 40 (1-2): 23-34.

Hirsch W.M., Dantzig G.B. (1968) The fixed charge problem. *Naval Research Logistics Quarterly* 15 (3): 413-424.

Hitchcock F.L. (1941) The distribution of product from several sources to numerous localities. *Journal of Mathematical Physics* 20 (1-4): 224–230.

Holland H. J. (1975) *Adaptation in natural and artificial systems.* University of Michigan Press, Ann Arbor, MI.

Huang X. (2007a) Chance-constrained programming models for capital budgeting with NPV as fuzzy parameters. *Journal of Computational and Applied Mathematics* 198 (1): 149-159.

Huang X. (2007b) Optimal project selection with random fuzzy parameters. *International Journal of Production Economics* 106 (2): 513–522.

Huang X. (2012) A risk index model for portfolio selection with returns subject to experts' estimations. *Fuzzy Optimization and Decision Making* 11 (4): 451–463.

Iori M., Martello S., Pretolani D. (2010) An aggregate label setting policy for the multi-objective shortest path problem. *European Journal of Operational Research* 207 (3): 1489-1496.

Janiak A., Kasperski A. (2008) The minimum spanning tree problem with fuzzy costs. *Fuzzy Optimization and Decision Making* 7 (2): 105–118.

Jarník V. (1930) O jistém problem minimálním. *Práce Moravské Přírodovědecké Společnosti* IV (3): 57-63.

Ji X., Yang L., Shao Z. (2006) Chance-constrained maximum flow problem with fuzzy arc capacities. In: *International Conference on Intelligent Computing*, *ICIC 2006* (D.-S. Huang, K. Li, G.W. Irwin, eds.). Lecture Notes in Computer Science, Springer, Berlin, Heidelberg, vol. 4114, pp. 11–19.

Ji X., Iwamura K., Shao Z. (2007) New models for shortest path problem with fuzzy arc lengths. *Applied Mathematical Modelling* 31 (2): 259-269.

Jian L., Xiaojing X., Kaiquan S. (2008) Rough similarity degree and rough close degree in rough fuzzy sets and the applications. *Journal of Systems Engineering and Electronics* 19 (5): 945-951.

Jiménez F., Verdegay J.L. (1998) Uncertain solid transportation problems. *Fuzzy Sets and Systems* 100 (1–3): 45-57.

Jiménez F., Verdegay J. (1999) An evolutionary algorithm for interval solid transportation problems. *Evolutionary Computation* 7 (1): 103-107.

Isermann H. (1979) The enumeration of all efficient solutions for a linear multi-objective transportation problem. *Naval Research Logistics Quarterly* 26 (1): 123-139.

Ishii H., Shiode S., Nishida T., Namasuya Y. (1981) Stochastic spanning tree problem. *Discrete Applied Mathematics* 3 (4): 263–273.

Ishii H., Matsutomi T. (1995) Confidence regional method of stochastic spanning tree problem. *Mathematical and Computer Modelling* 22 (10): 77–82.

Itoh T., Ishii H. (1996) An approach based on necessity measure to the fuzzy spanning tree problems. *Journal of the Operations Research Society of Japan* 39 (2): 247–257.

Jiménez F., Verdegay J.L. (1999) Solving fuzzy solid transportation problems by an evolutionary algorithm based parametric approach. *European Journal of Operational Research* 117 (3): 485-510.

Jungnickel D. (1999) *Graphs, Networks and Algorithms*. Springer-Verlag, Berlin.

Karger D.R., Klein P.N., Tarjan R.E. (1995) A randomized linear-time algorithm to find minimum spanning trees. *Journal of the Association for Computing Machinery* 42 (2): 321–328.

Karzanov A.V. (1974) Determining the maximal flow in a network by the method of preflows. *Soviet Math Doklady* 15 (2): 434-437.

Katagiri H., Kato K., Hasuike T. (2012) A random fuzzy minimum spanning tree problem through a possibility-based value at risk model. *Expert Systems with Applications* 39 (12): 10639-10646.

Katagiri H., Mermri E.B., Sakawa M., Kato K. (2004) A study on fuzzy random minimum spanning tree problems through possibilistic programming and the expectation optimization model. In: *Proceedings of the 47th IEEE International Midwest Symposium on Circuits and Systems*, Hiroshima, Japan, pp. III-49-III-52.

Kaufmann A., Gupta M.M. (1985) *Introduction to fuzzy arithmetic: Theory and applications*. Van Nostrand Reinhold, New York.

Kaur A., Kumar A. (2012) A new approach for solving fuzzy transportation problems using generalized trapezoidal fuzzy numbers. *Applied Soft Computing* 12 (3): 1201−1213.

Kennington J., Unger E. (1976) A new branch-and-bound algorithm for the fixed-charge transportation problem. *Management Science* 22 (10): 1116–1126.

Keshavarz E., Khorram E. (2009) A fuzzy shortest path with the highest reliability. *Journal of Computational and Applied Mathematics* 230 (1): 204-212.

Kim K., Roush F. (1982) Fuzzy flows on networks. *Fuzzy Sets and Systems* 8 (1): 35-38.

Klein C.M. (1991) Fuzzy shortest paths. *Fuzzy Sets and Systems* 39 (1): 27-41.

Knuth D.E. (1977) A generalization of Dijkstra's algorithm. *Information Processing Letters* 6 (1): 1–5.

König D. (1936) *Theorie der endlichen und unendlichen Graphen*. Akademische Verlagsgesellschaft, Leipzig.

Koopmans T.C. (1949) Optimum utilization of the transportation system. *Econometrica: Journal of the Econometric Society* 17: 136-146.

Kostreva M.M., Wiecek M.M. (1993) Time dependency in multiple objective dynamic programming. *Journal of Mathematical Analysis and Applications* 173 (1): 289–307.

Kowalski K., Lev B., Shen W., Tu Y. (2014) A fast and simple branching algorithm for solving small scale fixed-charge transportation problem. *Operations Research Perspectives* 1 (1): 1-5.

Kruskal Jr. J.B. (1956) On the shortest spanning subtree of a graph and traveling salesman problem. In: *Proceeding of the American Mathematical Society* 7 (1): 48–50.

Kumar A., Kaur M. (2010) An algorithm for solving fuzzy maximal flow problems using generalized trapezoidal fuzzy numbers. *International Journal of Applied Science and Engineering* 8 (2): 109–118.

Kumar A., Kaur M. (2011) Solution of fuzzy maximal flow problems using fuzzy linear programming. *International Journal of Computational and Mathematical Sciences* 5 (2): 62–67.

Kumar M.K., Sastry V.N. (2013) A new algorithm to compute Pareto-optimal paths in a multi-objective fuzzy weighted network. *Opsearch* 50 (3): 297-318.

Kumar R., Jha S., Singh S. (2017) Shortest path problem in network with type-2 triangular fuzzy arc length. *Journal of Applied Research on Industrial Engineering* 4 (1): 1-7.

Kundu P., Kar S., Maiti M. (2013a) Some solid transportation models with crisp and rough costs. *International Journal of Mathematical and Computational Sciences* 7 (1): 14-21.

Kundu P., Kar S., Maiti M. (2013b) Multi-objective solid transportation problems with budget constraint in uncertain environment. *International Journal of Systems Science* 45 (8): 1668-1682.

Kundu P., Kar S., Maiti M. (2013c) Multi-objective multi-item solid transportation problem in fuzzy environment. *Applied Mathematical Modelling* 37 (4): 2028-2038.

Kundu P., Kar S., Maiti M. (2014a) Fixed charge transportation problem with type-2 fuzzy variables. *Information Sciences* 255: 170-186.

Kundu P., Kar S., Maiti M. (2014b) A fuzzy MCDM method and an application to solid transportation problem with mode preference. *Soft Computing* 18 (9): 1853-1864.

Kundu P., Kar S., Maiti M. (2017a) A fuzzy multi-criteria group decision making based on ranking interval type-2 fuzzy variables and an application to transportation mode selection problem. *Soft Computing* 21 (11): 3051–3062.

Kundu P., Kar M.B., Kar S., Pal T., Maiti M. (2017b) A solid transportation model with product blending and parameters as rough variables. *Soft Computing* 21 (9): 2297–2306.

Kwakernaak H. (1978) Fuzzy random variables—I. definitions and theorems. *Information Sciences* 15 (1): 1–29.

Lamar B.W., Wallace C.A. (1997) Revised-modified penalties for fixed charge transportation problems. *Management Science* 43 (10): 1431–1436.

Last M., Eyal S. (2005) A fuzzy-based lifetime extension of genetic algorithms. *Fuzzy Sets and Systems* 149 (1): 131–147.

Le H., Gogne Z. (2010) Fuzzy linear programming with possibility and necessity relation. In: *Fuzzy Information and Engineering 2010* (B. Cao, G. Wang, S. Guo, S. Chen, eds.). Advances in Intelligent and Soft Computing, Springer, Berlin, vol. 78, pp. 305–311.

Li T.-J., Zhang W.-X. (2008) Rough fuzzy approximations on two universes of discourse. *Information Sciences* 178 (3): 892–906.

Li Y., Zou C.Y., Zhang S., Vai M.I. (2013) Research on multi-objective minimum spanning tree algorithm based on ant algorithm. *Research Journal of Applied Sciences, Engineering and Technology* 5 (21): 5051-5056.

Lin L., Gen M. (2007) A bi-criteria shortest path routing problems by hybrid genetic algorithm in communication networks. *2007 IEEE Congress on Evolutionary Computation, CEC 2007,* Singapore, pp. 4577-4582.

Lin Y., Li Y., Wang C., Chen J. (2018) Attribute reduction for multi-label learning with fuzzy rough set. *Knowledge-Based Systems* doi: 10.1016/j.knosys.2018.04.004.

Liou T.-S., Wang M.-J. J. (1992) Ranking fuzzy numbers with integral value. *Fuzzy Sets and Systems* 50 (3): 247-255.

Liu B. (1997) Dependent-chance programming: A class of stochastic programming. *Computers and Mathematics with Applications* 34 (12): 89-104.

Liu B. (2001) *Random fuzzy variables and random fuzzy programming*. Technical Report.

Liu B. (2002) *Theory and practice of uncertain programming*, 1st ed. Springer-Verlag, Berlin.

Liu B. (2004) *Uncertainty theory: An introduction to its axiomatic foundation*, 1st ed. Springer-Verlag, Berlin.

Liu B. (2007) *Uncertainty theory*, 2nd edn. Springer-Verlag, Berlin.

Liu B. (2009) Some research problems in uncertainty theory. *Journal of Uncertain Systems* 3 (1): 3-10.

Liu B. (2010) *Uncertainty theory: A branch of mathematics for modeling human uncertainty*. Springer-Verlag, Berlin.

Liu B. (2012) Why is there a need for uncertainty theory? *Journal of Uncertain System* 6 (1): 3–10.

Liu B., Iwamura K. (1998) Chance-constrained programming with fuzzy parameters. *Fuzzy Sets and Systems* 94 (2): 227-237.

Liu B., Liu Y.-K. (2002) Expected value of fuzzy variable and fuzzy expected value models. *IEEE Transactions on Fuzzy Systems* 10 (4): 445–450.

Liu S.T., Kao C. (2004) Network Flow Problems with Fuzzy Arc Lengths. *IEEE Transactions on Systems, Man and Cybernetics* 34(1): 765-769.

Liu Y.-K., Liu B. (2003) Expected value operator of random fuzzy variable and random fuzzy expected value models. *International Journal of Uncertainty, Fuzziness and Knowledge-Based Systems* 11 (2): 195-215.

Liu W., Yang C. (2007) Fuzzy random degree constrained minimum spanning tree problem. *International Journal of Uncertainty, Fuzziness and Knowledge-Based Systems* 15 (2): 107-115.

Liu L, Yang X, Mu H, Jiao Y (2008) The fuzzy fixed charge transportation problem and genetic algorithm. In: *Proceedings of the Fifth International Conference on Fuzzy*

*Systems and Knowledge Discovery, FSKD'08*, IEEE Computer Society Washington, DC, USA, pp. 208–212.

Liu L., Zhang B., Ma W. (2017) Uncertain programming models for fixed charge multi-item solid transportation problem. *Soft Computing* doi: 10.1007/s00500-017-2718-0.

Liu P., Yang L., Wang L., Li S. (2014a) A solid transportation problem with type-2 fuzzy variables. *Applied Soft Computing* 24: 543–558.

Liu W., Yang C. (2007) Fuzzy random degree constrained minimum spanning tree problem. *International Journal of Uncertainty, Fuzziness and Knowledge-Based Systems* 15 (2): 107-115.

Liu Y., Ha M. (2010) Expected value of function of uncertain variables. *Journal of Uncertain Systems* 4 (3): 181-186.

Liu Y., Lin Y., Zhao H.-h. (2014b) Variable precision intuitionistic fuzzy rough set model and applications based on conflict distance. *Expert Systems* 32 (2): 220-227

Liu Y.-K., Liu B (2000) *Fuzzy random variables: A scalar expected value*. Technical Report.

Liu Z.Q., Liu Y.-K. (2010) Type-2 fuzzy variables and their arithmetic. *Soft Computing* 14 (7): 729–747.

Lotfi M.M., Tavakkoli-Moghaddam R. (2013) A genetic algorithm using priority-based encoding with new operators for fixed charge transportation problems. *Applied Soft Computing* 13 (5): 2711-2726.

Lozano M., Glover F., García-Martínez C., Rodríguez F.J., Martí R. (2013) Tabu search with strategic oscillation for the quadratic minimum spanning tree. *IIE Transactions* 46 (4): 414-428

Mahapatra D.R., Roy S.K., Biswal M.P. (2010) Multi-objective stochastic transportation problem involving log-normal. *Journal of Physical Science* 14: 63-76.

Mahdavi I., Nourifar R., Heidarzade A., Amiri N.M. (2009) A dynamic programming approach for finding shortest chains in a fuzzy network. *Applied Soft Computing* 9 (2): 503-511.

Mahdavi I., Mahdavi-Amiri N., Nejati S. (2011) Algorithms for bi-objective shortest path problems in fuzzy networks. *Iranian Journal of Fuzzy Systems* 8 (4): 9-37.

Maia S.M.D.M., Goldbarg E.F.G., Goldbarg M.C. (2013) On the bi-objective adjacent only quadratic spanning tree problem. *Electronic Notes in Discrete Mathematics* 41 (5): 535-542.

McDonald R., Pereira F., Ribarov K., Hajič J. (2005) Non-projective dependency parsing using spanning tree algorithms. In: *Proceedings of the Conference on Human Language Technology and Empirical Methods in Natural Language Processing*, pp. 523–530. Association for Computational Linguistics, Stroudsburg, PA, USA.

Majumder S., Kundu P., Kar S., Pal T. (2018) Uncertain multi-objective multi-item fixed charge solid transportation problem. *Soft Computing* doi: 10.1007/s00500-017-2987-7.

Manzo R., Piccoli B., Rarità L. (2012) Optimal distribution of traffic flows in emergency cases. *European Journal of Applied Mathematics* 23 (4): 515-535.

Mendel J.M., John R.I. (2002) Type-2 fuzzy sets made simple. *IEEE Transactions on Fuzzy Systems* 10 (2): 307–315.

Mendel J.M. John R.I., Liu, F.L. (2006) Interval type-2 fuzzy logical systems made simple. *IEEE Transactions on Fuzzy Systems* 14 (6): 808–821.

Michalewicz Z. (1992) *Genetic Algorithms + Data Structures = Evolution Programs*, 1st edn. Springer-Verlag, Berlin.

Midya S., Roy S.K. (2014) Solving single-sink, fixed-charge, multi-objective, multi-index stochastic transportation problem. *American Journal of Mathematical and Management Sciences* 33 (4): 300-314.

Mizumoto M., Tanaka K. (1981) Fuzzy sets and type-2 under algebraic product and algebraic sum. *Fuzzy Sets and Systems* 5 (3): 277-290.

Moradkhan M.D., Browne W.N. (2006) A knowledge-based evolution strategy for the multi-objective minimum spanning tree problem. In: *2006 IEEE International Conference on Evolutionary Computation*, pp. 1391–1398.

Mou D., Zhao W., Chen X. (2013) Transportation problem with uncertain truck times and unit costs. *Industrial Engineering and Management Systems* 12 (1): 30-35.

Munakata T., Hashier D.J. (1993) A genetic algorithm applied to the maximum flow problem. In: *Proceeding of the 5th International Conference on Genetic Algorithms*, San Francisco, CA, pp. 488–493.

Nawathe S.P., Rao B.V. (1980) Maximum flow in probabilistic communication networks. *International Journal of Circuit Theory and Applications* 8 (2): 167-177.

Nebro J., Alba E., Molina G., Chicano F., Luna F., Durillo J.J. (2007) Optimal antenna placement using a new multi-objective CHC algorithm. In: *Proceedings of the 9th annual conference on Genetic and evolutionary computation*, *GECCO '07,* New York, NY, pp. 876-883.

Nahmias S. (1978) Fuzzy variable. *Fuzzy Sets and Systems* 1 (2): 97–101.

Nešetřil J., Nešetřilová H. (2012) The origins of minimum spanning tree algorithms – Borüvka and Jarník. *Documenta Mathematica* Extra Volume ISMP 127-141.

Nie Y., Wu X. (2009). Shortest path problem considering on-time arrival probability. *Transportation Research Part B: Methodological* 43 (6): 597–613.

Ojha A., Das B., Mondal S., Maiti M. (2009) An entropy-based solid transportation problem for general fuzzy costs and time with fuzzy equality. *Mathematical and Computer Modelling* 50 (1–2): 166-178.

Ojha A., Das B., Mondal S. K., Maiti M. (2014) A transportation problem with fuzzy stochastic cost. *Applied Mathematical Modelling* 38 (4): 1464-1481.

Okeda S. (2004) Fuzzy shortest path problems incorporating interactivity among paths. *Fuzzy Sets and Systems* 142 (3): 335-357.

Okada S., Gen M. (1994) Fuzzy shortest path problem. *Computers & Industrial Engineering* 27 (1–4): 465-468.

Öncan T. (2007) Design of capacitated minimum spanning tree with uncertain cost and demand parameters. *Information Sciences* 177 (20): 4354–4367.

Öncan T., Punnen A.P. (2010) The quadratic minimum spanning tree problem: A lower bounding procedure and an efficient search algorithm. *Computers & Operations Research* 37 (10): 1762-1773.

Palekar U.S., Karwan M.H., Zionts S. (1990) A branch-and-bound method for the fixed charge transportation problem. *Management Science* 36 (9): 1092–1105.

Palubeckis G., Rubliauskas D., Targamadzè A. (2010) Metaheuristic approaches for the quadratic minimum spanning tree problem. *Information Technology and Control* 39 (4): 257–268.

Pandian P., Anuradha D. (2010) A new approach for solving solid transportation problems. *Applied Mathematical Sciences* 4 (72): 3603-3610.

Pandian P., Natarajan G. (2010) A new method for finding an optimal solution for transportation problems. *International Journal of Mathematical Sciences and Engineering Applications* 4 (2): 59-65.

Papadimitriou C.H., Yannakakis M. (2000) On the approximability of trade-offs and optimal access of web sources. In: *Proceedings 41st Annual IEEE Symposium on Foundations of Computer Science*, Redondo Beach, CA, USA, pp. 86-92.

Pawlak Z. (1982) Rough sets. *International Journal of Computer and Information Sciences* 11 (5): 341-356.

Pawlak Z. (1991) *Rough sets - theoretical aspects of reasoning about data*, 1st edn. Kluwer Academatic Publishers, Springer, Netherlands.

Pawlak Z., Skowron A. (2007) Rough sets: Some extensions. *Information Sciences* 177 (1): 28-40.

Peidro D., Vasant P. (2011) Transportation planning with modified S-curve membership functions using an interactive fuzzy multi-objective approach. *Applied Soft Computing* 11 (2): 2656-2663.

Pereira D.L., Gendreau M., Cunha A.S.D. (2013) Stronger lower bounds for the quadratic minimum spanning tree problem with adjacency costs. *Electronic Notes in Discrete Mathematics* 41 (5): 229–236.

Pereira D.L., Gendreau M., Cunha A.S.D. (2015) Lower bounds and exact algorithms for the quadratic minimum spanning tree problem. *Computers & Operations Research* 63: 149-160.

Pettie S., Ramachandran V. (2002) An optimal minimum spanning tree algorithm. *Journal of the Association for Computing Machinery* 49 (1): 16–34.

Polkowski L. (2002) *Rough sets mathematical foundations*, 1st edn. Physica-Verlag, Heidelberg.

Prim R.C. (1957) Shortest connection networks and some generalizations. *Bell System Technical Journal* 36 (6): 1389–1401.

Pramanik S., Jana D.K., Maiti M. (2015a) A fixed charge multi-objective solid transportation problem in random fuzzy environment. *Journal of Intelligent & Fuzzy Systems* 28 (6): 2643-2654.

Pramanik S., Jana D.K., Mondal S.K., Maiti M. (2015b) A fixed charge transportation problem in two-stage supply chain network in Gaussian type-2 fuzzy environments. *Information Sciences* 325: 190-214.

Pramanik S., Banerjee D., Giri B.C. (2015) Multi-objective chance-constrained transportation problem with fuzzy parameters. *Global Journal of Advanced Research* 2 (1): 49-63.

Pulat P.S., Huarng F., Ravindran A. (1992) An Algorithm for bi-criteria Integer Network Flow Problem. In: *Proceedings on Tenth International Conference on Multiple Criteria Decision Making*, Taipei, Taiwan, pp. 305-318.

Puri M.L., Ralescu D.A. (1985) The concept of normality for fuzzy random variables. *The Annals of Probability* 13 (4): 1373–1379.

Qiao J., Hu B.Q. (2018) On ($\odot$, &)-fuzzy rough sets based on residuated and co-residuated lattices. *Fuzzy Sets and Systems* 336: 54-86.

Qin R., Liu Y.K., Liu Z.Q. (2011) Methods of critical value reduction for type-2 fuzzy variables and their applications. *Journal of Computational and Applied Mathematics* 235 (5): 1454–1481.

Ramakrishnan C.S. (1988) An improvement to Goyal's modified VAM for the unbalanced transportation problem. *Journal of the Operational Research Society* 39 (6): 609-610.

Rani D., Garg H. (2018) Complex intuitionistic fuzzy power aggregation operators and their applications in multi-criteria decision-making. *Expert Systems*, doi: 10.1111/exsy.12325.

Rao S.S. (2006) *Engineering optimization-theory and practice*, 3rd edn. New Age International Publishers, New Delhi.

Reinfeld N.V., Vogel W. (1958) *Mathematical programming*. Prentice-Hall, Englewood Cliffs, NJ.

Romeijn H.E., Sargut F.Z. (2011) The stochastic transportation problem with single sourcing. *European Journal of Operational Research* 214 (2): 262-272.

Rostami B., Malucelli F. (2015) Lower bounds for the quadratic minimum spanning tree problem based on reduced cost computation. *Computers & Operations Research* 64: 178-188

Sadjadi S.J., Ghazanfari M., Yousefli A. (2010) Fuzzy pricing and marketing planning model: A possibilistic geometric programming approach. *Expert Systems with Applications* 37 (2): 3392–3397.

Sakawa M., Kato K., Nishizaki I. (2003) An interactive fuzzy satisficing method for multi-objective stochastic liner programing problems through an expectation model. *European Journal of Operational Research* 145 (3): 665–672.

Schell E.D. (1955) Distribution of a product by several properties. In: *Proceedings 2nd Symposium in Linear Programming*, DCS/Comptroller, HQUS Air Force, Washington, DC, pp. 615–642.

Sedeño-Noda A., Raith A. (2015) A Dijkstra-like method computing all extreme supported nondominated solutions of the bi-objective shortest path problem. *Computers & Operations Research* 57: 83–94.

Shafiee M., Shams-e-alam N. (2011) Supply chain performance evaluation with rough data envelopment analysis. In: *International Conference on Business and Economics Research*, Kuala Lumpur, Malaysia, pp. 57-61.

Sheng Y., Gao J. (2014) Chance distribution of the maximum flow of uncertain random network. *Journal of Uncertainty Analysis and Applications* 2-15, doi: 10.1186/s40467-014-0015-3.

Sheng Y., Gao Y. (2016) Shortest path problem of uncertain random network. *Computers and Industrial Engineering*. 99: 97-105.

Sheng Y., Qin Z., Shi G. (2017) Minimum spanning tree problem of uncertain random network. *Journal of Intelligent Manufacturing* 28 (3): 565-574.

Sheng Y., Yao K. (2012a) Fixed charge transportation problem and its uncertain programming model. *Industrial Engineering & Management Systems* 11 (2): 183-187.

Sheng Y., Yao K. (2012b) A transportation model with uncertain costs and demands. *Information: An International Interdisciplinary Journal* 15 (8): 3179-3186.

Shi G., Sheng Y, Ralescu D.A. (2017a) The maximum flow problem of uncertain random network. *Journal of Ambient Intelligence and Humanized Computing* 8 (5): 667-675.

Shi N., Zhou S., Wang F., Tao Y., Liu L. (2017b) The multi-criteria constrained shortest path problem. *Transportation Research Part E: Logistics and Transportation Review* 101: 13-29.

Shih W. (1987) Modified stepping stone method as a teaching aid for capacitated transportation problems. *Decision Sciences* 18: 662-676.

Sinha B., Das A., Bera U.K. (2016) Profit maximization solid transportation problem with trapezoidal interval type-2 fuzzy numbers. *International Journal of Applied and Computational Mathematics* 2 (1): 41-56.

Słowiński R., Vanderpooten D. (2000) A generalized definition of rough approximations based on similarity. *IEEE Transactions on Knowledge and Data Engineering* 12 (2): 331-336.

Steiner S., Radzik T. (2008) Computing all efficient solutions of the bi-objective minimum spanning tree problem. *Computers & Operations Research* 35 (1): 198–211.

Sun M., Aronson J.E., Mckeown P.G., Drinka D. (1998) A tabu search heuristic procedure for the fixed charge transportation problem. *European Journal of Operational Research* 106 (2-3): 441–456.

Sun B., Ma W., Gong Z. (2014a) Dominance-based rough set theory over interval-valued information systems. *Expert Systems* 31 (2): 185-197.

Sun B., Ma W., Zhao H. (2014b) Decision-theoretic rough fuzzy set model and application. *Information Sciences* 283: 180-196

Sutcliffe C., Board J., Cheshire P. (1984) Goal programming and allocating children to secondary schools in reading. *The Journal of the Operational Research Society* 35 (8): 719-730.

Sheng Y., Yao K. (2012a) Fixed charge transportation problem and its uncertain programming model. *Industrial Engineering & Management Systems* 11 (2): 183-187.

Sheng Y., Yao K. (2012b) A transportation model with uncertain costs and demands. *Information* 15 (8): 3179-3186.

Sundar S., Singh A. (2010) A swarm intelligence approach to the quadratic minimum spanning tree problem. *Information Sciences* 180 (17): 3182-3191.

Swamy C., Shmoys D.B. (2006) Approximation algorithms for 2-stage stochastic optimization problems. *Association for Computing Machinery Special Interest Group on Algorithms and Computation Theory* 37 (1): 33–46.

Tajdin A., Mahdavi I., Mahdavi-Amiri N., Sadeghpour-Gildeh B. (2010) Computing a fuzzy shortest path in a network with mixed fuzzy arc lengths using $\alpha$-cuts. *Computers & Mathematics with Applications* 60 (4): 989-1002.

Tarjan R.E. (1984) A simple version of Karzanov's blocking flow algorithm. *Operations Research Letters* 2 (6): 265-268.

Tayi G.K. (1986) Bi-criteria transportation problem: An alternate approach. *Socio-Economic Planning Sciences* 20 (3): 127-130.

Thielen, C. Westphal S. (2013) Complexity and approximability of the maximum flow problem with minimum quantities. *Networks* 62 (2): 125-131.

Torkestani J.A. (2013) Degree constrained minimum spanning tree problem: a learning automata approach. *The Journal of Supercomputing* 64 (1): 226–249.

Torkestani J. A., Meybodi M. R. (2012) A learning automata-based heuristic algorithm for solving the minimum spanning tree problem in stochastic graphs. *The Journal of Supercomputing* 59 (2): 1035–1054.

Tsumoto S. (2007) Medical differential diagnosis from the viewpoint of rough sets. *Information Sciences* 10: 28-34.

Van Veldhuizen D.A., Lamont G.B. (1998) *Multi-objective evolutionary algorithm research: A history and analysis*. Technical report TR-98-03, Department of electrical and computer engineering, Graduate school of engineering, Air force institute of technology, Wright-Patterson, AFB, OH.

Vignaux G., Michalewicz Z. (1991) A genetic algorithm for the linear transportation problem. *IEEE Transactions on Systems, Man and Cybernetics: Systems* 21 (2): 445-452.

Wang D., Zheng J., Ma G., Song X., Liu Y. (2016) Risk prediction of product-harm events using rough sets and multiple classifier fusion: an experimental study of listed companies in China. *Expert Systems* 33 (3): 254-274.

Wang P. (1982) Fuzzy contactability and fuzzy variables. *Fuzzy Sets and Systems* 8 (1): 81–92.

Wen M., Iwamura K. (2008) Facility location–allocation problem in random fuzzy environment: Using $(\alpha,\beta)$-cost minimization model under the Hurewicz criterion. *Computers and Mathematics with Applications* 55 (4): 704-713.

Williams A.C. (1963) A stochastic transportation problem. *Operations Research* 11 (5): 759-770.

Wilson D. (1975) A mean cost approximation for transportation problems with stochastic demand. *Naval Research Logistics Quarterly* 22 (1): 181-187.

Wu D., Mendel J. M. (2007) Uncertainty measures for interval type-2 fuzzy sets. *Information Sciences* 177: 5378–5393.

Wu X. L., Liu Y. K. (2012) Optimizing fuzzy portfolio selection problems by parametric quadratic programming. *Fuzzy Optimization and Decision Making* 11 (4): 411–449.

Wu W.-Z., Zhang W.-X., Li H.-Z. (2003) Knowledge acquisition in incomplete fuzzy information systems via the rough set approach. *Expert Systems* 20 (5): 280-286.

Xu J., Li B., Wu D. (2009) Rough data envelopment analysis and its application to supply chain performance evaluation. *International Journal of Production Economics* 122 (2): 628-638.

Xu Z. (2006) Dependent OWA operators. In: *International Conference on Modeling Decisions for Artificial Intelligence, MDAI 2006* (V. Torra, Y. Narukawa, A. Valls, J.

Domingo-Ferrer, eds.). Lecture Notes in Computer Science, Springer, Berlin, Heidelberg, vol. 3885, pp. 172-178.

Yager R.R. (1981) A procedure for ordering fuzzy subsets of the unit interval. *Information Sciences* 24 (2): 143-161.

Yager R.R. (1988) On ordered weighted averaging aggregation operators in multi-criteria decision making. *IEEE Transactions on Systems, Man, and Cybernetics* 18 (1): 183-190.

Yang L., Feng Y. (2007) A bi-criteria solid transportation problem with fixed charge under stochastic environment. *Applied Mathematical Modelling* 31 (12): 2668-2683.

Yang L., Liu L. (2007) Fuzzy fixed charge solid transportation problem and algorithm. *Applied Soft Computing* 7 (3): 879-889.

Yang X., Gao J. (2016) Linear quadratic uncertain differential game with application to resource extraction problem. *IEEE Transactions on Fuzzy Systems* 24 (4): 819-826.

Yang X., Gao J. (2017) Bayesian equilibria for uncertain bimatrix game with asymmetric information. *Journal of Intelligent Manufacturing* 28 (3): 515-525.

Yang Y., Hinde C. (2010) A new extension of fuzzy sets using rough sets: R-fuzzy sets. *Information Sciences* 180 (3): 354-365.

Yi X., Miao Y., Zhou J., Wang Y. (2016) Some novel inequalities for fuzzy variables on the variance and its rational upper bound. *Journal of Inequalities and Applications* 2016 (41), doi: 10.1186/s13660-016-0975-6.

Zadeh L.A. (1965) Fuzzy sets. *Information Control* 8 (3): 338–353.

Zadeh L.A. (1975a) The concept of a linguistic variable and its application to approximate reasoning – I. *Information Sciences* 8 (3): 199–249.

Zadeh L.A. (1975b) The concept of a linguistic variable and its application to approximate reasoning – II. *Information Sciences* 8 (4): 301–357.

Zadeh L.A. (1978) Fuzzy sets as a basis for a theory of possibility. *Fuzzy Sets and Systems* 100 (1): 9-34.

Zhang B., Peng J., Li S., Chen L. (2016) Fixed charge solid transportation problem in uncertain environment and its algorithm. *Computers & Industrial Engineering* 102: 186-197.

Zhang M., Xu L.D., Zhang W.-X., Li H.-Z. (2003) A rough set approach to knowledge reduction based on inclusion degree and evidence reasoning theory. *Expert Systems* 20 (5): 298-304.

Zhang X., Mei C., Chen D., Yang Y. (2018) A fuzzy rough set-based feature selection method using representative instances. *Knowledge-Based Systems* doi:10.1016/j.knosys.2018.03.031.

Zhang R., Kabadi S.N., Punnen A.P. (2011) The minimum spanning tree problem with conflict constraints and its variations. *Discrete Optimization* 8 (2): 191-205.

Zhang R., Punnen A.P. (2013) Quadratic bottleneck knapsack problems. *Journal of Heuristics* 19 (4): 573-589.

Zhang X., Wang Q., Zhou J. (2013a) Two uncertain programming models for inverse minimum spanning tree problem. *Industrial Engineering & Management Systems* 12 (1): 9-15.

Zhang X., Wang Q., Zhou J. (2013b) A chance-constrained programming model for inverse spanning tree problem with uncertain edge weights. *International Journal of Advancements in Computing Technology* 5 (6): 76-83.

Zhou A., Jin Y., Zhang Q., Sendhoff B., Tsang E. (2006) Combining model-based and genetics-based offspring generation for multi-objective optimization using a convergence criterion. In: *2006 IEEE International Conference on Evolutionary Computation*, CEC 2006, Vancouver, BC, Canada, pp. 892-899.

Zhou G., Gen M. (1998) An effective genetic algorithm approach to the quadratic minimum spanning tree problem. *Computers & Operations Research* 25 (3): 229-247.

Zhou G., Gen M. (1999) Genetic algorithm approach on multi-criteria minimum spanning tree problem. *European Journal of Operational Research* 114 (1): 141–152.

Zhou J., Chen L., Wang K. (2015) Path optimality conditions for minimum spanning tree problem with uncertain edge weights. *International Journal of Uncertainty, Fuzziness and Knowledge-Based Systems* 23 (1): 49-71.

Zhou J., Wang Q., Zhang X. (2013) The Inverse Spanning Tree of a Fuzzy Graph Based on Credibility Measure. *Journal of Communications* 8 (9): 566-571.

Zhou J., Yang F., Wang K. (2014a) An inverse shortest path problem on an uncertain graph. *Journal of Networks* 9 (9): 2353-2359.

Zhou J., He X., Wang K. (2014b) Uncertain Quadratic Minimum Spanning Tree Problem. *Journal of Communications* 9 (5): 385-390.

Zhou J., Chen L., Wang K., Yang F. (2016a) Fuzzy $\alpha$-minimum spanning tree problem: Definition and solutions. *International Journal of General Systems* 45 (3): 311-335.

Zhou J., Yang F., Wang K. (2016b). Fuzzy arithmetic for LR fuzzy numbers with applications to fuzzy programming. *Journal of Intelligent & Fuzzy Systems* 30 (1): 71-87.

Zhou J., Yi X., Wang K., Liu J. (2016c) Uncertain distribution-minimum spanning tree problem. *International Journal of Uncertainty, Fuzziness and Knowledge-Based Systems* 24 (4): 537-560.

Zhou L., Lü K., Liu W., Ren C. (2018) Decision-making under uncertainty through extending influence diagrams with interval-valued parameters. *Expert Systems*, doi: 10.1111/exsy.12277.

Zhu W. (2009) Relationship between generalized rough sets based on binary relation and covering. *Information Sciences* 179 (3): 210-225.

Zimmermann H.-J. (1978) Fuzzy programming and linear programming with several objective functions. *Fuzzy Sets and Systems* 1 (1): 45–55.

Zitzler E., Thiele L. (1999) Multi-objective evolutionary algorithms: a comparative case study and the strength Pareto approach. *IEEE Transactions on Evolutionary Computation* 3 (4): 257-271.

Zockaie A., Nie Y., Mahmassani H. S. (2014) A simulation-based method for finding minimum travel time budget paths in stochastic networks with correlated link times. *Transportation Research Record: Journal of the Transportation Research Board* 2467: 140–148. Washington, D.C.: Transportation Research Board of the National Academies.

# List of Publications

1. **Saibal Majumder**, Bishwajit Saha, Pragya Anand, Samarjit Kar, Tandra Pal (2018) Uncertainty based genetic algorithm with varying population for random fuzzy maximum flow problem. *Expert Systems, Wiley,* doi: 10.1111/exsy.12264 (***Science Citation Index Expanded***).
2. **Saibal Majumder**, Pradip Kundu, Samarjit Kar, Tandra Pal (2018) Uncertain multi-objective multi-item fixed charge solid transportation problem with budget constraint. *Soft Computing, Springer Berlin Heidelberg,* doi: 10.1007/s00500-017-2987-7 (***Science Citation Index Expanded***).
3. **Saibal Majumder**, Samarjit Kar (2017) Multi-criteria shortest path for rough graph. *Journal of Ambient Intelligence and Humanized Computing, Springer Berlin Heidelberg,* doi: 10.1007/s12652-017-0601-6 (***Science Citation Index Expanded***).
4. Pradip Kundu, **Saibal Majumder**, Samarjit Kar, Manoranjan Maiti (2018) A method to solve linear programming problem with interval type-2 fuzzy parameters. *Fuzzy Optimization and Decision Making, Springer US,* doi: 10.1007/s10700-018-9287-2 (***Science Citation Index Expanded***).
5. **Saibal Majumder**, Samarjit Kar, Tandra Pal (2018) Rough-fuzzy quadratic minimum spanning tree problem. *Expert Systems, Wiley,* doi: 10.1111/exsy.12364 (***Science Citation Index Expanded***).